\documentclass[11pt]{article}

\usepackage[T1]{fontenc}
\usepackage[utf8]{inputenc}
\usepackage{doatlas-report}
\usepackage{times}
\usepackage{microtype}
\usepackage{amsmath,amssymb}
\usepackage{graphicx}
\usepackage{booktabs}
\usepackage{array}
\usepackage{wrapfig}
\usepackage{needspace}
\usepackage{xcolor}
\usepackage{hyperref}
\usepackage{url}

\hypersetup{
  hidelinks,
  pdftitle={DoAtlas-2: A Foundation for Self-Evolving Causal Biomedical Discovery},
  pdfauthor={DoAtlas Team, MBZUAI},
  pdfsubject={Technical report},
  pdfkeywords={causal discovery, biomedical AI, scientific discovery}
}

\title{DoAtlas-2: A Foundation for Self-Evolving Causal Biomedical Discovery}
\shorttitle{DoAtlas-2 Technical Report}
\author{Yulong~Li$^{*1,2}$, Rong~Xia$^{*1,2}$, Yuxuan~Zhang$^{*1,2}$, Jianxu~Chen$^{1,2}$,
  Xiwei~Liu$^{1,2}$, Haochen~Xue$^{1,2}$, Maosheng~Li$^{1,2}$, Yuhang~Liu$^{1,2}$, Yibo~Yuan$^{1,2}$,
  Yutong~Xie$^{2}$, Chong~Li$^{1}$, Jionglong~Su$^{1}$, Hagai~Rossman$^{2}$, Eran~Segal$^{2}$, Imran~Razzak$^{\dagger,1,2}$}
\titlenote{$^{*}$Equal contribution. $^{\dagger}$Corresponding authors.}
\contact{\href{mailto:yulong.li@mbzuai.ac.ae}{Yulong.Li@mbzuai.ac.ae},
  \href{mailto:Imran.Razzak@mbzuai.ac.ae}{Imran.Razzak@mbzuai.ac.ae}}
\affiliation{ $^1$DoAtlas Team $\cdot$ MedOmni\quad\quad $^2$MBZUAI}
\date{September 2026}
\reportlogos{\includegraphics[height=1.9cm]{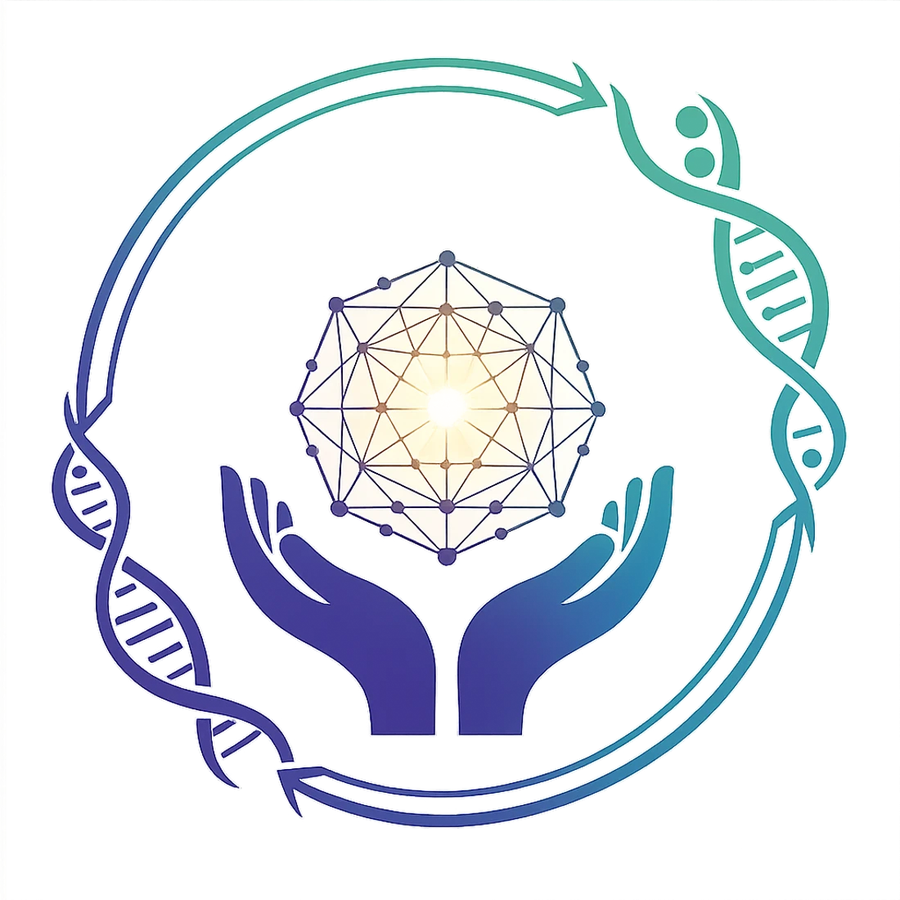}}
  {\includegraphics[height=1.15cm]{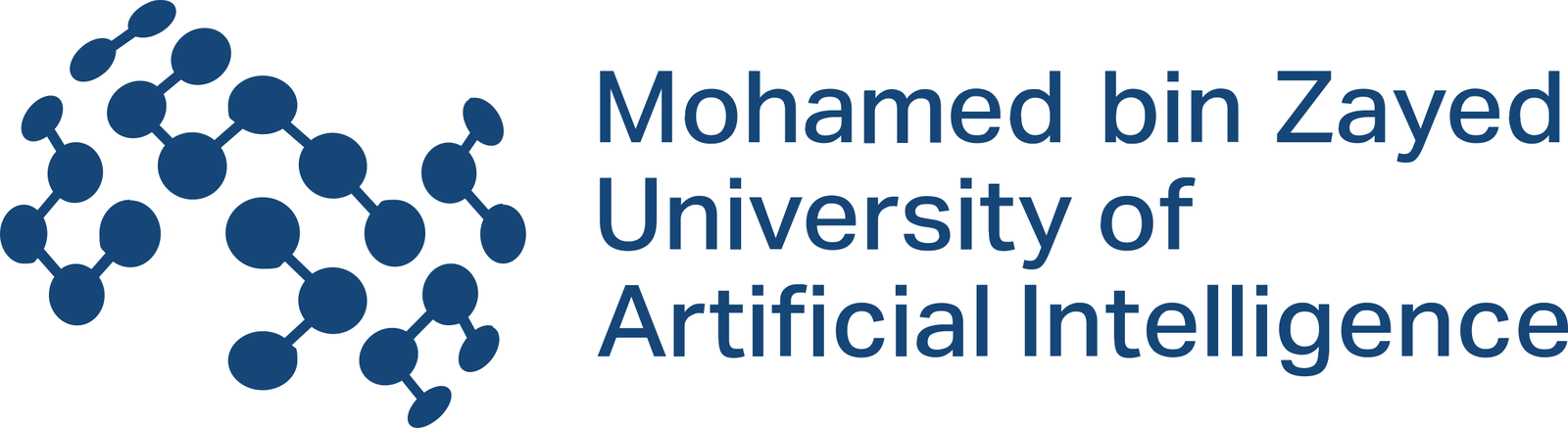}}

\begin{document}

\maketitle

\begin{abstract}
We introduce DoAtlas-2, a foundation for self-evolving causal biomedical
discovery that organizes knowledge around causal mechanisms and advances through
external evidence from human populations. DoAtlas-2 integrates 771 research
resources covering more than 720,000 participants in 48 countries, from
longitudinal clinical phenotypes, medical imaging, and continuous physiological
signals to eight molecular layers, together with a contextualized evidence
network of approximately 4.7 million literature-derived records over 93,566
concepts and 149,383 candidate causal relations. DoAtlas-2 autonomously
formulates research questions from evidence gaps and unresolved mechanisms,
prespecifies their causal designs, and generates validated analyses.
Supporting, challenging, and unresolved results continuously revise mechanistic
interpretations, the causal evidence state, and the discovery frontier, so that
DoAtlas-2 self-evolves within a closed loop of hypothesis generation, empirical
testing, and renewed discovery. DoAtlas-2 has systematically evaluated 2,031
research questions. In the Human Phenotype Project (HPP), it formulated 4,014
candidate pathway questions across vascular, early-glycemic, and
hepatic-metabolic systems, and screening of the first 1,079 yielded statistical
support for 756. Representative studies identify blood pressure as a
convergence node linking adiposity, hepatic, and lipid phenotypes to vascular
outcomes, and show that an adiposity--inflammation--blood-pressure pathway is
largely attenuated by joint adjustment for body mass index (BMI) and smoking. The discovered
vascular network constitutes a completely interpretable predictive foundation,
admitting exact attribution of every prediction and closed-form
mediation effects. DoAtlas-2 thereby unifies causal mechanism discovery,
population-evidence testing, and interpretable prediction within one
continuously evolving foundation.
\end{abstract}

\section{Overview}

\looseness=-1
Biomedical discovery depends on new evidence revising existing scientific
judgments and determining what should be studied next~\citep{king2009automation}. Biomedical AI now spans knowledge organization, hypothesis generation, study design, and causal
analysis \citep{zhang2025biomedicalkg,gottweis2026coscientist,ghareeb2026robin}. Yet these paradigms remain anchored to a researcher-specified objective, disease scope, or starting question, and do not convert external validation outcomes into continuous feedback that revises causal state and subsequent research direction. Growth in knowledge scale and completion of individual tasks therefore do not constitute an evolution in scientific capability. When external evidence revises scientific state, interpretability must extend across the entire evolution of knowledge rather than remain a post
hoc explanation of an individual result \citep{ghassemi2021falsehope}. The reason each mechanism is supported, bounded, contested, or revised, and the reason that change redirects subsequent research, must rest on a testable causal basis~\citep{pearl2009causal,bareinboim2016datafusion}. Causal interpretability thereby provides the scientific basis for evidence-driven self-evolution.

DoAtlas-2 begins from a contextualized causal evidence state. Evidence gaps,
conflicting relations, and unresolved paths jointly define the research
frontier, from which DoAtlas-2 autonomously formulates and prioritizes testable
research questions. Selected mechanisms are evaluated in the Human Phenotype Project (HPP) \citep{reicher2025deep} and
independent data; supportive, challenging, or unresolved evidence revises mechanism state and
redefines subsequent research questions and evidence-acquisition priorities.
The next research round begins from this revised causal state, allowing each
external outcome to reshape the subsequent discovery frontier. This continuous
cycle of inquiry, validation, revision, and renewed discovery connects otherwise
isolated scientific tasks into a self-evolving process driven by external
evidence across successive rounds of scientific inquiry.

This loop operates over a large-scale causal evidence base. As of September
22, 2026, DoAtlas-2 contains 93,566 canonical nodes, 149,383 candidate
causal relations, and 4,696,704 literature-derived evidence records, and supports
4,014 candidate pathway questions across vascular, early-glycemic, and
hepatic-metabolic systems. We evaluate this foundation through two decisive demonstrations: whether
evidence-guided evolution reveals biomedically meaningful long-chain causal
pathways; and whether those findings revise the causal evidence state and
redirect subsequent research questions. Both demonstrations require every mechanism and its revision to
have an explicit causal basis.

DoAtlas-2 moves biomedical AI from one-shot hypothesis generation and static
evidence accumulation toward cumulative causal mechanism discovery driven by
external evidence, in which mechanism discovery, scientific explanation, and
evidence evolution form a unified scientific process. Biomedical knowledge
thereby shifts from isolated associations and black-box predictions to an
evolving system of causal mechanisms.
Sections~\ref{sec:self-evolution-loop}--\ref{sec:local-foundation} report the
discovery loop across 2,031 evaluated research questions, representative
mechanistic findings in which blood pressure links adiposity, hepatic, and
lipid phenotypes to vascular outcomes, and an interpretable
predictive foundation.

\section*{Data, Research Scope, and Resource Footprint}

\begin{figure*}[t]
  \centering
  \includegraphics[width=0.95\textwidth]{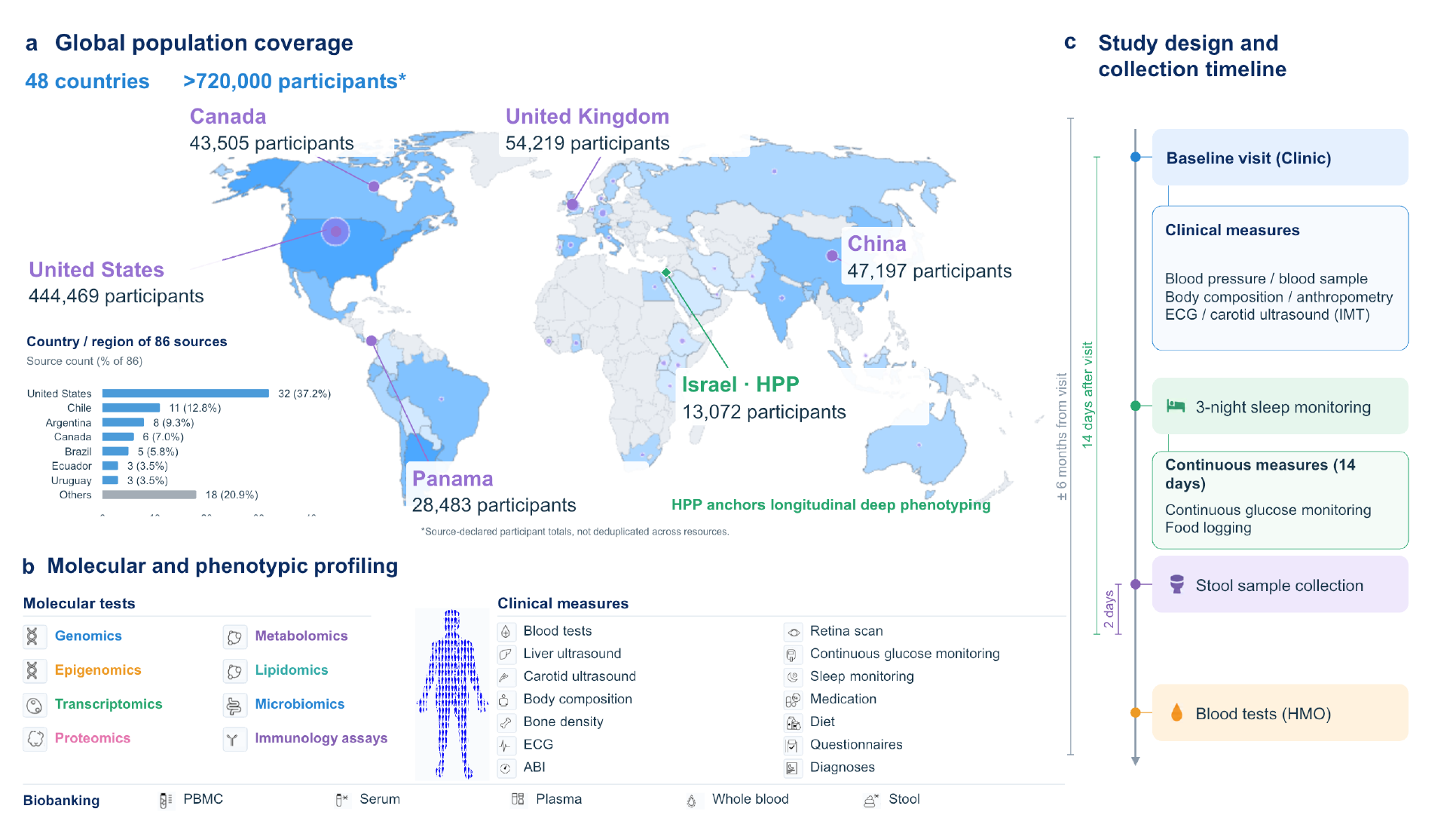}

  \caption{\textbf{Population coverage, molecular profiling and study design of DoAtlas-2.}
  \textbf{(a)} Resources span 48 countries and $>$720,000 source-reported
  participants. The inset shows the country/region distribution of
  86 sources. Israel HPP anchors longitudinal deep phenotyping.
  \textbf{(b)} Molecular assays, clinical measures and biobanking.
  \textbf{(c)} Study design and collection timeline.
  Participant totals are not deduplicated across resources.}

  \label{fig:data-foundation}
  \vspace{-1.5em}
\end{figure*}

DoAtlas-2's self-evolution is grounded in data resources spanning
populations, organ systems, and molecular scales. The data landscape
brings together 771 research resources, covering more than 720,000 study
participants across 48 countries. Its measurements extend from population
surveys, longitudinal cohorts, clinical records, and continuous physiological
signals to medical imaging, genomics, epigenomics, transcriptomics, proteomics, metabolomics, lipidomics, microbiomics, and immunology assays. Spanning cardiovascular, glycemic,
hepatic, renal, sleep and respiratory, muscular, and body-composition systems, these resources contain more than 40 million resource-specific measurement dimensions, allowing candidate causal mechanisms to be constrained by cross-population replication, longitudinal phenotyping, and molecular evidence across scales (Figure~\ref{fig:data-foundation}).

As the core longitudinal observational cohort of DoAtlas-2, HPP connects long-term follow-up, deep phenotyping, and
multi-omics measurements at the participant level \citep{reicher2025deep}. By December 2024, HPP had
enrolled 27,916 participants, of whom 13,072 had completed their first
deep-phenotyping visit. Designed for 25 years of follow-up, the cohort tracks
health through annual questionnaires and periodic in-person reassessments;
current observations span baseline, one-year follow-up, a second in-person
visit, and three-year follow-up. Across 37 observational domains, 3,067
measurement fields comprise 63,004,814 field-level measurements, while
population structure is characterized through grandparental birthplaces
spanning 61 countries and five principal ancestry groups. Clinical assays,
lifestyle, anthropometry, continuous glucose, sleep, activity, medical imaging,
and multi-omics biosamples form participant-linked, cross-system longitudinal
records. These include 109,788 participant-days of continuous glucose
monitoring, 574,000 participant-days of food, medication, sleep, and activity
logging, 41,899 participant-days of sleep monitoring, 17,278 liver-ultrasound
examinations, 17,278 carotid-ultrasound examinations, 16,683 dual-energy X-ray absorptiometry (DXA) scans, and
10,954 fundus-imaging records. HPP thereby provides temporally structured
deep-phenotype evidence for candidate mechanisms and serves as the core
longitudinal environment for causal-state revision and held-out evaluation.

\Needspace{22\baselineskip}

\begin{wraptable}[18]{r}{0.50\textwidth}
  \centering
  \vspace{-0.8em}
  \caption{\textbf{Research-resource footprint.}}
  \vspace{-0.5em}
  \label{tab:research-resource-footprint}
  \small
  \setlength{\tabcolsep}{1.6pt}
  \renewcommand{\arraystretch}{1.0}
  \begin{tabular*}{\linewidth}{@{\extracolsep{\fill}}lrrrr@{}}
    \toprule
    \multicolumn{5}{@{}l@{}}{8,724 candidate RQs; 2,031 completed.} \\
    \midrule
      & Terminal & Complete & Tokens & Time \\
    \midrule
    Overall & $\approx$62,677 & $\approx$60,646
      & $\approx$1.587\,T & 70\,d \\
    Mean & 30.86 & 29.86 & 782\,M & 5.26\,h \\
    Median & 30.50 & 30.00 & 770\,M & 4.83\,h \\
    Range & 24--40 & 19--39 & 430--1,460\,M & 3.64--8.71\,h \\
    \midrule
    \multicolumn{5}{@{}l@{}}{Input / output tokens:} \\
    \multicolumn{5}{@{}l@{}}{Overall: $\approx$1.578\,T / $\approx$9.458\,B;} \\
    \multicolumn{5}{@{}l@{}}{mean per completed RQ: 777\,M / 4.7\,M.} \\
    \bottomrule
  \end{tabular*}

  \vspace{0.35em}
  \begin{minipage}{\linewidth}
    \footnotesize
     Mean, median and range are per completed RQ.
    M/B/T: million/billion/trillion; d/h: days/hours.
    Terminal: experimental units reaching terminal disposition.
    Complete: units completing the prespecified computation and producing
    an analytic terminal record. 
  \end{minipage}
\vspace{-1em}
\end{wraptable}

\enlargethispage{\baselineskip}
As of September 22, 2026, DoAtlas-2 had established a research space
comprising 8,724 candidate research questions (RQs) and completed systematic
evaluation of 2,031 independent RQs.
Each completed RQ yielded a mean of 30.86 experimental units reaching terminal
disposition, of which 29.86 completed the prespecified computation and produced
terminal records. All evaluations used GPT-5.6 Sol with xhigh reasoning and
consumed a mean of 781,551,295 model tokens per RQ. With downstream experiments
executed concurrently within each RQ, the mean end-to-end evaluation time was
5.26 hours. Across the study, this corresponded to approximately 62,677
terminal experimental units, 60,646 computation-complete terminal units, and
1.587 trillion model tokens over a 70-day run (Table~\ref{tab:research-resource-footprint}).

\section{The DoAtlas-2 Self-Evolving Causal Foundation}
\label{sec:foundation}

DoAtlas-2 represents each research round by $S_t=(E_t,Q_t,A_t)$:
the contextualized causal evidence state, the discovery frontier and the
external evidence requirements. Evidence gaps and conflicting findings generate
candidate mechanism questions with traceable parent results. Selected questions
specify the study population, temporal structure, causal estimand and
identification conditions before external outcomes are revealed \citep{nosek2018preregistration}. Research agents
report all prespecified outcomes, including diagnostics and non-estimable states. External scientific selection records $\Sigma_t$ drive the transition
$S_t\xrightarrow{\Sigma_t}S_{t+1}$: results revise evidence support and scope,
reshaping questions and evidence needs for the next round.
Supported paths motivate extension; counterevidence motivates competing
mechanisms; context restrictions and unresolved results identify boundary tests
or additional evidence requirements. An attributable joint revision of
$E_t$, $Q_t$ and $A_t$ constitutes one self-evolution transition.
Each change preserves its source, applicable context and remaining uncertainty.

\begin{figure}[tp]
\centering
\includegraphics[width=\linewidth]{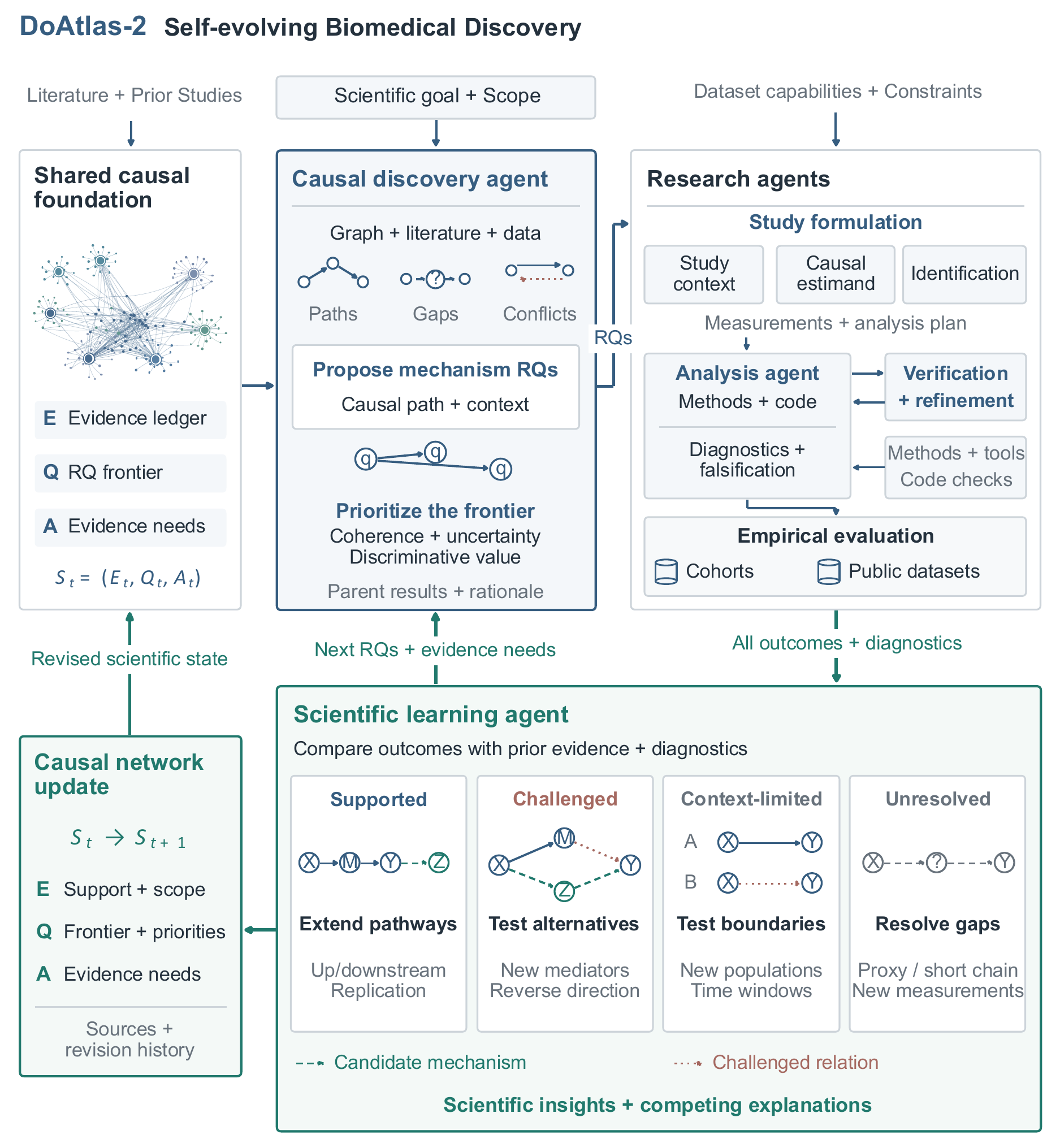}
\caption{\small\textbf{Agent architecture for evidence-driven causal discovery.}
The causal discovery agent proposes mechanism questions; research agents turn
them into studies (Appendix~\ref{app:rq-designs}) and report outcomes,
uncertainty and diagnostics; the scientific learning agent revises evidence
support and scope, the discovery frontier and evidence needs. The four
response motifs show context-dependent follow-up; A/B denote contexts. Teal
dashed/red dotted edges, candidate/challenged relations; the network thumbnail
shows 183 concepts and 372 candidate relations.}
\label{fig:scientific-loop}
\vspace{-1.5em}
\end{figure}

Long-chain composition requires compatible directions, contexts, measurements
and temporal structures.
Figure~\ref{fig:scientific-loop} summarizes the agent architecture.
Appendix~\ref{app:foundation} gives the detailed definitions of question
selection, study design, external adjudication and state revision.

\section{Evidence for the Self-Evolution Loop}
\label{sec:self-evolution-loop}

\begin{figure*}[t]
  \centering
  \includegraphics[width=\textwidth]{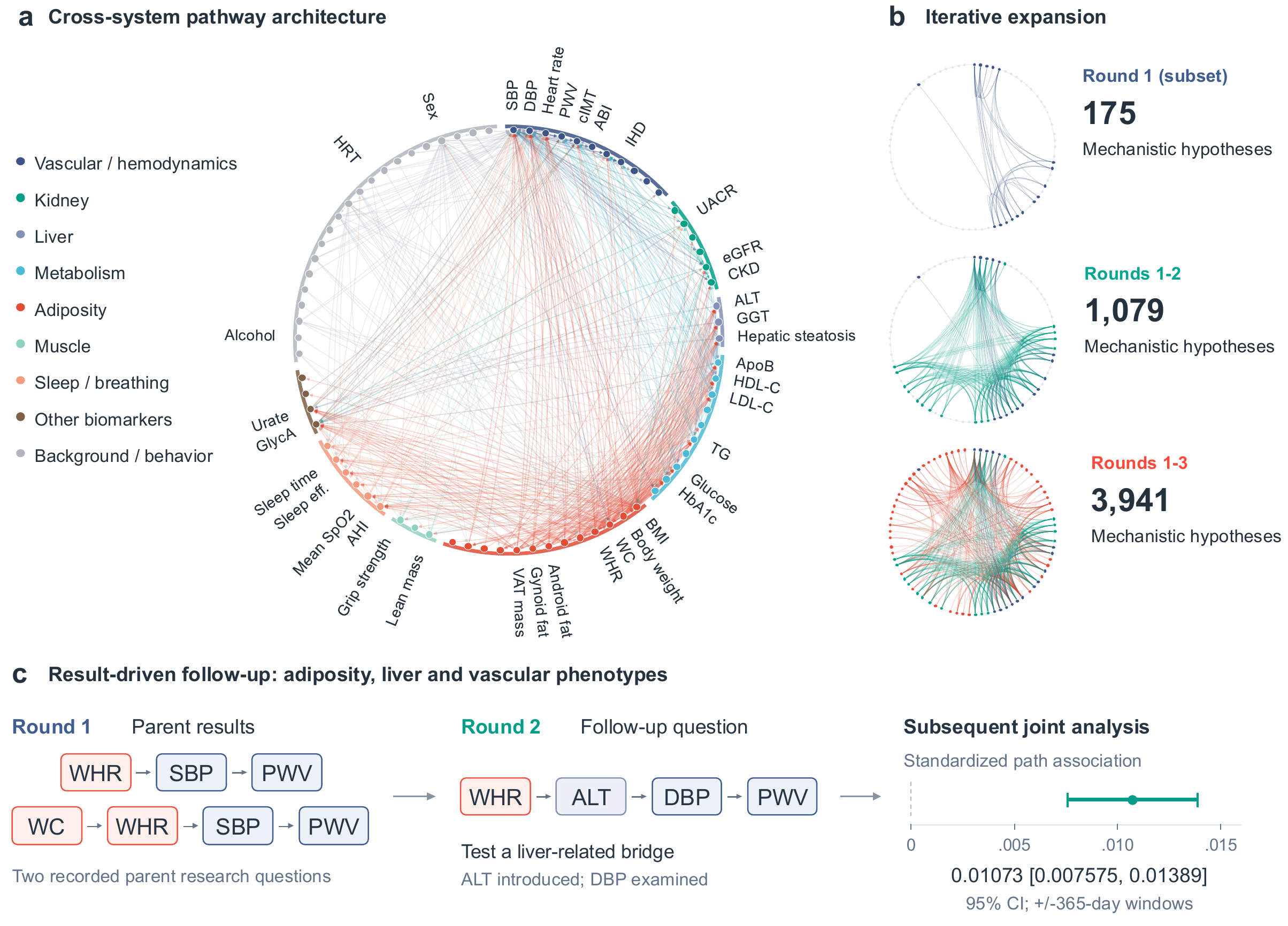}

  \caption{\textbf{Cross-system pathway atlas and iterative hypothesis expansion.}
  \textbf{(a)} Network of phenotype and biomarker concepts, colored by domain.
  \textbf{(b)} Expansion of hypotheses, concepts and connections across research rounds.
  \textbf{(c)} Example of result-driven follow-up linking adiposity, liver and vascular phenotypes.
  The point and horizontal line show the standardized path association and its 95\% confidence interval.
  Arrows represent hypothesized relationships or research progression; estimates are observational.}

  \label{fig:pathway-atlas}
  \label{fig:findings:network-overview}
  \vspace{-1.5em}
\end{figure*}

DoAtlas-2 has established a research space of 8,724 candidate research
questions and systematically evaluated 2,031 of them
(Table~\ref{tab:research-resource-footprint}), covering 88,731 result
estimates. Across three research rounds, the cumulative candidate catalog grew
from 175 to 1,079 and then to 4,014 pathway questions covering 3,941 distinct
pathways (Figure~\ref{fig:findings:network-overview}b), connecting 77 concepts
and 517 candidate directed relations across seven physiological domains, other
biomarkers and background or behavioral variables
(Figure~\ref{fig:findings:network-overview}a). Unadjusted screening in the first two rounds
covered 1,079 research questions, of which 756 received screening-level
statistical support. Appendix~\ref{app:loop:scope} specifies the scope and
counting units.

All 904 Round-2
questions record Round-1 parents, linking to 61 distinct Round-1 questions.
Of these follow-ups, 326 test serial metabolic--hemodynamic pathways,
264 test competing mediator pathways that bypass blood pressure and 142 test
hemodynamic convergence; another 108 and 64 introduce liver- and sleep-related
phenotypes, respectively.
Round-1 studies of adiposity, blood pressure, and vascular endpoints thus
extend into systematic tests of mediator pathways and competing explanations.
Appendix~\ref{app:loop:scope} reports the round-level counts and parent--child
links; Appendix~\ref{app:loop:attribution} traces how each result leads to
specific successor questions.

The HPP results revise the causal evidence state, contributing 2,169
edge-level evidence records, 725 of which set relation states, so that 200 of
the 259 relations with HPP evidence acquire a population-derived state:
81 supported, 2 challenged and 117 unresolved. Of the 81 supported relations,
16 match a directional literature reference and 49 are supported by population
evidence alone, with no counterpart in the literature-derived evidence network.
Where the literature is divided, HPP evidence provides a population direction:
BMI is positively related to low-density lipoprotein (LDL) cholesterol
(7 increasing versus 4 decreasing literature reports) and high-density
lipoprotein (HDL) cholesterol inversely to diastolic blood pressure (DBP;
2 versus 3). An unresolved state indicates that current
population evidence cannot yet determine a relation's direction, rather than
that the relation is absent: 38 unresolved relations are supported in at least
one evidence context, 33 of them in cross-sectional analyses, but remain
imprecise in the others, mostly under temporally ordered designs, and the other
79 have imprecise or inconsistent estimates across contexts. These relations
form the frontier of subsequent research rounds, which call for larger
temporally ordered samples or more precise designs to address the remaining uncertainty (Appendix~\ref{app:loop}).

Figure~\ref{fig:findings:network-overview}c illustrates one cross-round follow-up. The
Round-1 WHR--SBP--PWV and WC--WHR--SBP--PWV results are parents
of the Round-2 WHR--ALT--DBP--PWV question. This follow-up introduces ALT
and examines the DBP branch to test a liver-related pathway between adiposity
and vascular phenotypes. Subsequent joint analysis with $\pm365$-day
measurement windows estimated a standardized path association of 0.01073
(95\% CI [0.007575, 0.01389]; Section~\ref{sec:findings:vascular-convergence}),
connecting prior results, a successor question, and its estimate.
Appendix~\ref{app:loop:case-h1} records the lineage and distinguishes this
joint model from the earlier Round-2 estimate of the same pathway (H1) in
Figure~\ref{fig:findings:experimental-atlas}c.

Prior results guide follow-up questions about competing mechanisms
and endpoint context. For adiposity--ApoB--blood-pressure--vascular pathways,
screening contrasts between serial and nested shorter paths and across
vascular endpoints motivate reassessment of blood-pressure paths, alternative
sleep- or lipid-related mediators, and the same mediator sequence for PWV
and carotid intima-media thickness (cIMT;
Appendices~\ref{app:loop:case-neck-apob}--\ref{app:loop:case-whr-endpoint}).
Round 3 treats intermediate phenotypes in existing paths as outcomes:
the first 30 pathway proposals span 15 outcome constructs. The
HbA1c--SBP--UACR hypothesis, for example, motivates comparisons of
blood-pressure-mediated, non-blood-pressure, and common-cause explanations
for urinary albumin-to-creatinine ratio
(Appendix~\ref{app:loop:case-glycemic-renal}). These cases show how preceding
results shape the specific questions and comparisons proposed for follow-up
research; Section~\ref{sec:key-findings} presents representative biomedical
findings.

\section{Key Findings}
\label{sec:key-findings}

Representative analyses show recurring blood-pressure-related structures in
adiposity--vascular associations, with sleep-oxygenation pathways additionally
linking to cIMT and lower-limb PWV. We present 15 pathways with complete serial
estimates as concrete examples, forming a cross-system association network
of 19 phenotype nodes and 29 connections
(Figure~\ref{fig:findings:experimental-atlas}b). Serial estimates are products of
the adjacent edge coefficients, on the standardized scale unless stated
otherwise, and all models account for age and sex.
Figure~\ref{fig:findings:experimental-atlas}c,d reports pathway estimates and
component associations, respectively;
Figure~\ref{fig:findings:experimental-atlas}e--j reports reverse-direction,
within-person, off-path and confounding-sensitivity analyses;
Figure~\ref{fig:findings:conditional-validation} presents temporal, population,
and covariate comparisons. Further results and
detailed analyses are provided in Appendix~\ref{app:findings} and the
accompanying online supplementary material.

\subsection{Blood-pressure convergence of metabolic and vascular phenotypes}
\label{sec:findings:vascular-convergence}

Five representative shorter paths retaining a blood-pressure node receive
statistical support, linking adiposity to cIMT or lower-limb pulse-wave
velocity (PWV) (B2 estimates in Figure~\ref{fig:findings:experimental-atlas}c:
WHR--DBP--PWV in H1, Hip--DBP--PWV in H2, Waist--SBP--cIMT in H3,
WHR--SBP--cIMT in H4 and VAT--SBP--PWV in L2).
In a joint model using $\pm365$-day windows between adjacent measurements,
the standardized WHR--ALT--DBP--PWV path product is 0.01073
(95\% CI [0.007575, 0.01389]; Figure~\ref{fig:findings:network-overview}c).
The full-path minus WHR--ALT--PWV contrast is
0.008542 (95\% CI [0.002897, 0.01419]; Holm-adjusted $P=0.01207$, \citealp{holm1979simple}),
estimated on a common sample and scale. The hip-related network supports
parallel liver and lipid branches: within hip-anchored $\pm365$-day windows,
the Hip--ALT--DBP--PWV and Hip--ApoB--DBP--PWV path products are 0.01174
(95\% CI [0.005539, 0.02063]) and 0.004717 (95\% CI [0.001192, 0.009364]),
respectively. Both branches retain support after Holm correction.
Hepatic and lipid branches converge on diastolic blood pressure and
lower-limb PWV, making blood pressure the shared intermediate linking
metabolic routes to vascular phenotypes.

In two independent cohorts of 912 \citep{fukuda2014ggtbmjopen,fukuda2014ggtdata} and 879
\citep{ning2022arterial} adults with brachial--ankle PWV (baPWV), the
metabolic-to-pressure routes replicate in the 912-adult cohort (BMI--TG--DBP,
GGT--TG--SBP, ALT--DBP and
TG--DBP to baPWV; Holm-adjusted $P\le0.0012$). In the 879-adult cohort, the
BMI--TG and DBP--baPWV associations are positive (standardized
DBP--baPWV coefficient 0.29, 95\% CI [0.22, 0.36]), and all serial
estimates share the HPP sign (Appendix~\ref{app:findings:replication}). Nocturnal oxygen saturation (SpO$_2$) paths O1 and O2 reach vascular
outcomes, cIMT and PWV (Figure~\ref{fig:findings:experimental-atlas}b).
The standardized Waist--SpO$_2$--SBP--cIMT path product (O1) is $-0.002524$
(95\% CI [$-0.004037$, $-0.001117$]); the SpO$_2$--GlycA--PWV product
(O2) is 0.02437 (95\% CI [0.004872, 0.04387]) on the measurement
scale. Both intervals exclude zero
(Figure~\ref{fig:findings:experimental-atlas}c, serial-product estimates S for O1 and O2).

Component estimates specify the association directions
(Figure~\ref{fig:findings:experimental-atlas}d, component edges E1--E3 of O1
and E1--E2 of O2). Lower nocturnal SpO$_2$ and obstructive sleep apnea
accompany higher blood pressure in clinical studies
\citep{nieto2000association,lavie2000obstructive}, and HPP reproduces this
association without adiposity adjustment: higher mean nocturnal SpO$_2$
accompanies lower SBP (age-, sex- and cohort-adjusted 95\% CI
[$-0.08928$, $-0.04359$]). Adiposity accounts for this inverse association.
Waist circumference is a shared determinant of both quantities: it is
negatively associated with nocturnal SpO$_2$ (E1: $-0.3492$, 95\% CI
[$-0.3768$, $-0.3216$]) and raises SBP, so participants with lower SpO$_2$
are also those with larger waists and higher pressure. Conditional on waist
circumference, the SpO$_2$--SBP association becomes weakly positive (E2:
0.04372, 95\% CI [0.01863, 0.06881]); by the omitted-variable relation
\citep{cinelli2020sensitivity}, a
standardized waist--SBP association of about 0.3 accounts for the change from
negative to positive. The pattern is consistent across HPP analyses: all 9
SpO$_2$--SBP intervals excluding zero are negative without an adiposity term
and all 10 are positive with waist, BMI, neck circumference or visceral fat in
the model, whereas SpO$_2$--DBP associations conditional on adiposity remain
predominantly negative (6 of the 7 intervals excluding zero; 34 of 43 point
estimates). In O2, both the SpO$_2$--GlycA and
GlycA--PWV conditional associations are negative.

\subsection{Temporal support for DBP paths and differences in ApoB associations}
\label{sec:findings:conditional-support}

ALT--DBP--PWV and TG--DBP--PWV both receive statistical support under strict
temporal ordering (Figure~\ref{fig:findings:conditional-validation}b).
In both sequences, the hepatic and lipid markers precede diastolic pressure,
supporting a metabolic-to-hemodynamic order within the liver and lipid modules. Selected ApoB-related sequences show sex and age differences. The
VAT--ApoB--SBP--PWV serial estimate differs significantly between sexes (Holm-adjusted $P=0.005$;
Figure~\ref{fig:findings:conditional-validation}d).
Weight--ApoB--DBP--PWV has a smaller serial estimate at older ages: the $+10$
minus $-10$ year contrast around the reference age is $\Delta=-0.02275$
(95\% CI [$-0.04126$, $-0.00828$]; Holm-adjusted $P=0.006$;
Figure~\ref{fig:findings:conditional-validation}e), indicating a weaker
lipid--pressure serial association with peripheral vascular phenotypes in older
participants.

\subsection{Covariate-dependent adiposity--inflammation--blood-pressure link}
\label{sec:findings:adjustment-revision}

Joint adjustment for BMI and smoking significantly attenuates the
WHR--GlycA--SBP--PWV serial association (R1). Under this adjustment, the
serial estimate decreases from 0.01218 to 0.002022
on a common sample and standardization scale, allowing direct comparison between the estimates. The formal difference is
$\Delta=-0.01016$ (95\% CI [$-0.01537$, $-0.004936$];
Figure~\ref{fig:findings:conditional-validation}c).
The attenuation indicates that the adiposity--inflammation--blood-pressure
pathway reflects overall adiposity and smoking more than the fat distribution
indexed by the waist-to-hip ratio.

\newsavebox{\primaryFigureBox}
\begin{figure}[t]
  \centering
  \begin{lrbox}{\primaryFigureBox}
  \begin{minipage}{\textwidth}
  \centering
  \includegraphics[width=\textwidth]{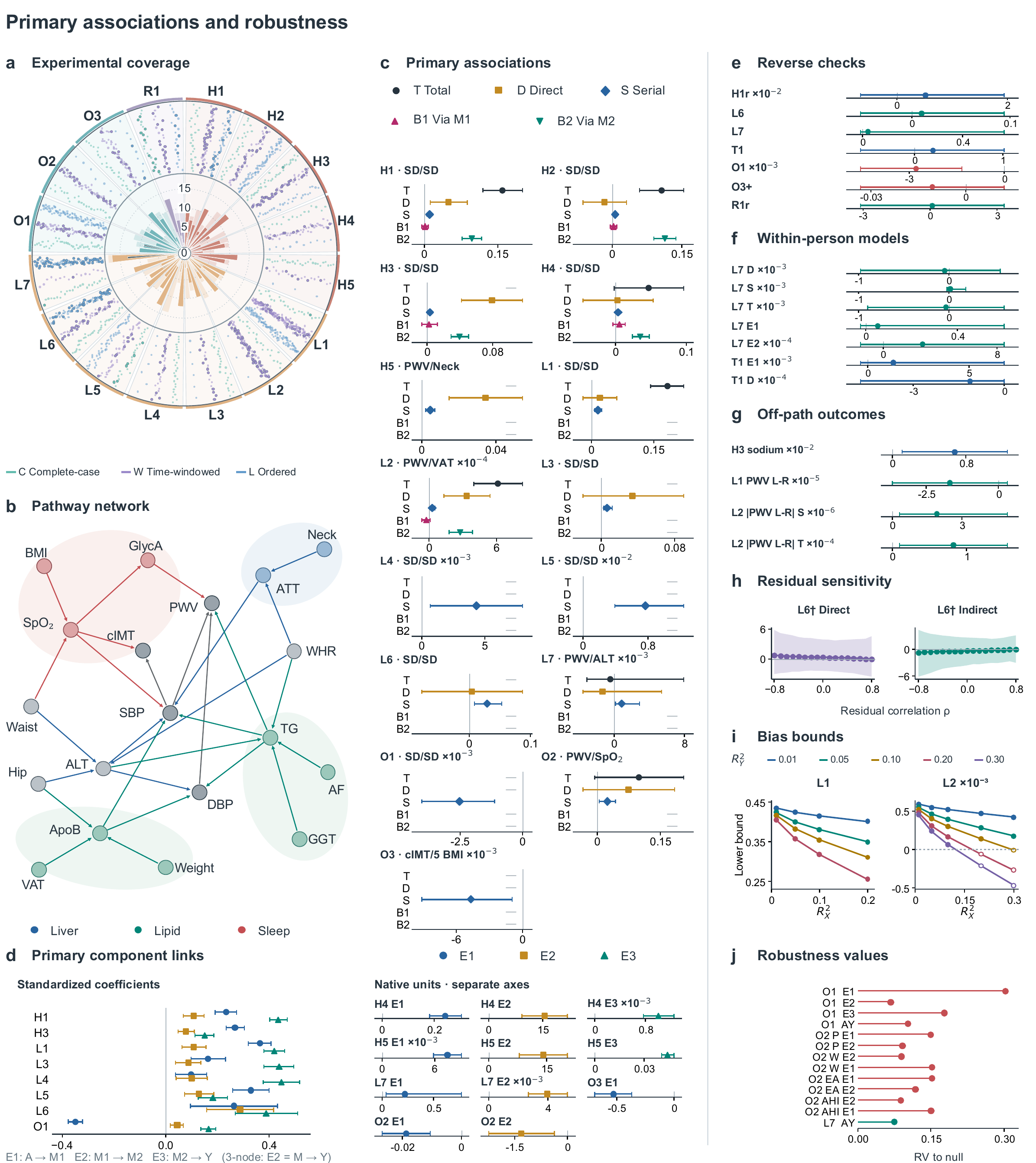}
  \caption{\textbf{Primary associations and robustness.}
  (a,b) Coverage/network; (c,d) pathway/link estimates;
  (e--g) reverse, within-person and off-path analyses;
  (h--j) confounding sensitivity.
  Bars: 95\% CIs. Definitions: Appendix~\ref{app:findings:experimental-atlas}.}
  \label{fig:findings:experimental-atlas}
  \end{minipage}
  \end{lrbox}
  \typeout{FIGURE3_CONTENT_HEIGHT=\the\dimexpr\ht\primaryFigureBox+\dp\primaryFigureBox\relax}
  \typeout{FIGURE3_WITH_GAP_HEIGHT=\the\dimexpr\ht\primaryFigureBox+\dp\primaryFigureBox+\textfloatsep\relax}
  \typeout{FIGURE3_THREE_QUARTER_HEIGHT=\the\dimexpr\textheight*3/4\relax}
  \ifdim\dimexpr\ht\primaryFigureBox+\dp\primaryFigureBox+\textfloatsep\relax>0.75\textheight
  \fi
  \usebox{\primaryFigureBox}
  \vspace{-1.5em}
\end{figure}

\newsavebox{\conditionsFigureBox}
\begin{figure}[t]
  \centering
  \begin{lrbox}{\conditionsFigureBox}
  \begin{minipage}{\textwidth}
  \centering
  \includegraphics[width=\textwidth]{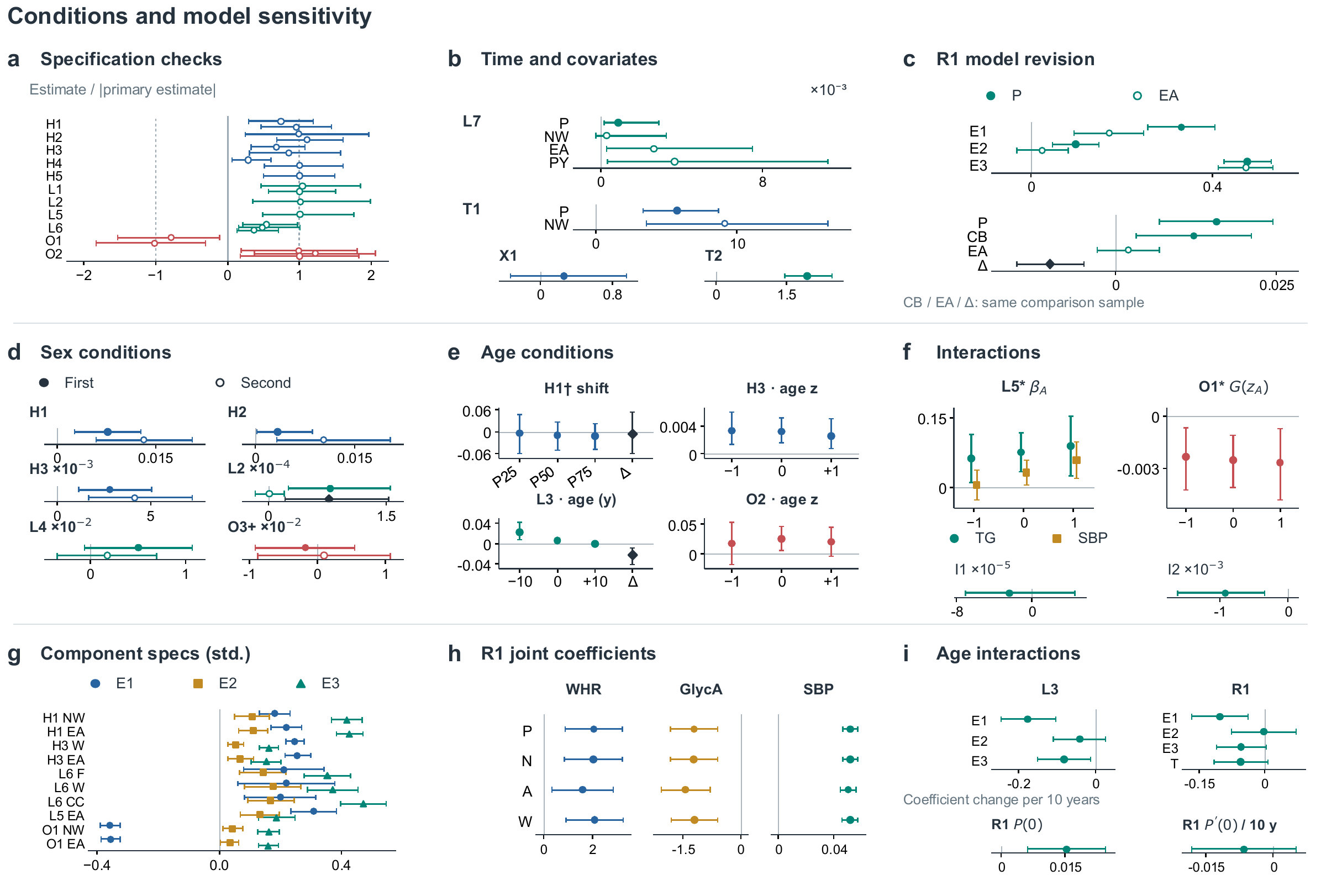}
  \caption{\textbf{Conditions and model sensitivity.}
  (a--c) Specification, timing and adjustment; (d--f) sex, age and interactions;
  (g--i) component, joint-coefficient and age-interaction sensitivity.
  128 estimates with 95\% intervals; definitions:
  Appendix~\ref{app:findings:conditional-validation}.}
  \label{fig:findings:conditional-validation}
  \vspace{-1.5em}
  \end{minipage}
  \end{lrbox}
  \typeout{FIGURE4_CONTENT_HEIGHT=\the\dimexpr\ht\conditionsFigureBox+\dp\conditionsFigureBox\relax}
  \typeout{FIGURE4_WITH_GAP_HEIGHT=\the\dimexpr\ht\conditionsFigureBox+\dp\conditionsFigureBox+\textfloatsep\relax}
  \typeout{FIGURE4_HALF_TEXT_HEIGHT=\the\dimexpr\textheight/2\relax}
  \ifdim\dimexpr\ht\conditionsFigureBox+\dp\conditionsFigureBox+\textfloatsep\relax>0.5\textheight
    \PackageError{figure-layout}{The conditions figure exceeds half the text-page height}{Reduce the figure or caption height.}
  \fi
  \usebox{\conditionsFigureBox}
\end{figure}

\section{From Causal Evidence to an Interpretable Foundation Model}
\label{sec:local-foundation}
\label{sec:findings:local-foundation}

The self-evolution of DoAtlas-2 yields a completely interpretable predictive
foundation. The vascular network of Section~\ref{sec:key-findings} (19 nodes,
29 connections, 15 pathways) is the model structure itself: every prediction
is an explicit closed-form expression, and every term belongs to a named
variable, a source relationship or an intercept. The decomposition of a
prediction into node, interaction and branch contributions is an exact
identity, and independent recomputation reproduces the model output for all
14,079 real predictions with a maximum discrepancy of $1.42\times10^{-14}$.
For temporally ordered questions, the same network yields closed-form direct
and mediated effects. Interpretability follows directly from the construction
of the model and its explicit mathematical structure.

\paragraph{Explicit predictive representation.}
For outcome task $t$, the prediction is
\begin{equation}
\begin{aligned}
 \eta_t(x,m)={}&b_t+\sum_j f_{tj}(x_j,m_j)
 +\sum_r a_r(x,m)\,w_{tr}\,q_r(x_{S_r})
 +\sum_j d_{tj}(1-m_j),\\
 q_r(x_{S_r})={}&\beta_{r0}+\sum_\ell\beta_{r\ell}\,\phi_{r\ell}(x_{S_r}).
\end{aligned}
\label{eq:foundation:task-head}
\end{equation}
Here $x$ contains the network phenotypes and covariates, and $m_j$ indicates
whether variable $j$ is observed; $f_{tj}$ is a univariate term; $q_r$ is the
relationship equation specified by its source study, with input set $S_r$ and
basis $\phi_{r\ell}$ carrying the declared transformations and interactions;
$w_{tr}$ weights the relationship for outcome $t$, and $a_r$ is its
applicability indicator. Each source term $q_r$ traces to a specific
relationship in Section~\ref{sec:key-findings}; $b_t$, $f_{tj}$ and $d_{tj}$
are task-specific fitted terms. Continuous outcomes use
$\widehat y_t=\eta_t$; 60-month mortality risk is
\begin{equation}
 \widehat R_{60}=1-\prod_{k=1}^{5}\bigl[1-\sigma(\alpha_k+\eta_t)\bigr],
\label{eq:foundation:mortality-link}
\end{equation}
where $\sigma$ is the logistic function and $\alpha_k$ is the intercept of
annual interval $k$.

\paragraph{The complete equation system of the network.}
Each source context yields one source equation for every directed connection
of its pathway (the extended L2 context is sex-stratified): target node
$Z_s$ takes as inputs all upstream nodes $Z_0,\dots,Z_{s-1}$, age, sex and
the covariates declared by the source study. Let
$S_r^{\mathrm c},S_r^{\mathrm{cat}}\subseteq S_r$ be the continuous and
categorical inputs of equation $r$, $S_r^{\mathrm{sq}}$ the inputs with a
declared quadratic term, and $\mathcal P_r$ its declared interaction pairs.
Continuous inputs enter as training-standardized values
$z_v=(x_v-\mu_v)/s_v$; categorical inputs expand into indicators relative
to a reference level $h_v^{0}$; quadratic terms are centered at their
training means:
\begin{equation}
\begin{aligned}
 q_r(x_{S_r})={}&\beta_{r0}
 +\sum_{v\in S_r^{\mathrm c}}\beta_{rv}z_v
 +\sum_{v\in S_r^{\mathrm{sq}}}\beta_{rv^2}\bigl(z_v^{2}-\overline{z_v^{2}}\bigr)\\
 &+\sum_{v\in S_r^{\mathrm{cat}}}\sum_{h\neq h_v^{0}}\beta_{rvh}\,\mathbf 1[x_v=h]
 +\sum_{(u,v)\in\mathcal P_r}\beta_{ruv}\,z_u z_v .
\end{aligned}
\label{eq:foundation:source-basis}
\end{equation}
Univariate terms $f_{tj}$ use the same standardization, with a linear term,
centered hinge terms $(z_j-\kappa)_+$ at the three training quartiles, and
indicators for categorical inputs. The 15 pathways and 22 source contexts
give 63 equations, of which 61 are estimable in HPP; the source coefficients
$\beta$ are estimated in HPP and frozen, and $b_t$, $f_{tj}$, $d_{tj}$ and
the source weights $w_{tr}$ are fitted per outcome task.
Table~\ref{tab:foundation:source-sequences} in
Appendix~\ref{app:disease-foundation-construction} lists the node chain,
source covariates and equation count of each pathway.

\paragraph{Node, interaction and branch contributions.}
Let $V_{r\ell}$ be the variable set of $\phi_{r\ell}$; then
\begin{equation}
\begin{aligned}
 C_{tj}(x,m)&=f_{tj}(x_j,m_j)
 +\sum_{r,\ell:\,V_{r\ell}=\{j\}}
 a_r\,w_{tr}\,\beta_{r\ell}\,\phi_{r\ell}(x_{S_r}),\\
 C^{\mathrm{int}}_{tJ}(x,m)&=
 \sum_{r,\ell:\,V_{r\ell}=J}
 a_r\,w_{tr}\,\beta_{r\ell}\,\phi_{r\ell}(x_{S_r}),\qquad |J|\geq2.
\end{aligned}
\label{eq:foundation:node-contributions}
\end{equation}
Covariate terms follow the same rule. The prediction satisfies the identity
\begin{equation}
 \eta_t=B_t+\sum_jC_{tj}+\sum_{J:|J|\geq2}C^{\mathrm{int}}_{tJ}+M_t,\qquad
 B_t=b_t+\sum_r a_rw_{tr}\beta_{r0},\quad
 M_t=\sum_jd_{tj}(1-m_j).
\label{eq:foundation:prediction-accounting}
\end{equation}
For a source pathway $b$ of Section~\ref{sec:key-findings} (a branch), let
$\mathcal R_b$ be its set of relationship equations; the branch contribution
is
\begin{equation}
 C^{\mathrm{path}}_{tb}(x,m)=\sum_{r\in\mathcal R_b}
 a_r(x,m)\,w_{tr}\,q_r(x_{S_r}).
\label{eq:foundation:path-contribution}
\end{equation}
The node view and the branch view are two alternative summaries of the same
prediction. Contributions keep the outcome's original units; mortality
contributions are on the logit-hazard scale. Independent recomputation
reproduces the model output for 12,492 HPP predictions and 1,587 National Health and Nutrition Examination Survey (NHANES) \citep{nchs2024nhanes}
predictions, with maximum discrepancies of $1.42\times10^{-14}$ and
$5.55\times10^{-16}$.

\paragraph{Effect decomposition of the BMI--SpO$_2$--cIMT pathway.}
Longitudinal analysis of this pathway
(O3 in Figure~\ref{fig:findings:experimental-atlas}) decomposes the total
BMI--cIMT contrast into a
positive direct and a negative SpO$_2$-mediated component. Let
$A$ denote BMI, $M$ SpO$_2$ and $Y$ cIMT; following
\citet{vanderweele2009mediation}, the conditional means are
\begin{equation}
\begin{aligned}
 \mu_M(a,c)&=\alpha_0+\alpha_Aa+\gamma_M^{\top}c,\\
 \mu_Y(a,m,c)&=\theta_0+\theta_Aa+\theta_Mm+\theta_{AM}am+\gamma_Y^{\top}c.
\end{aligned}
\label{eq:foundation:o3-conditional-model}
\end{equation}
The adjustment vector $C_e$ contains age, sex, cohort, prior smoking, education
and waist circumference. The interaction makes each conditional slope depend
on the other variable:
\begin{equation}
 \frac{\partial\mu_Y}{\partial a}=\theta_A+\theta_{AM}m,
 \qquad
 \frac{\partial\mu_Y}{\partial m}=\theta_M+\theta_{AM}a.
\label{eq:foundation:o3-conditional-slopes}
\end{equation}

The population pathway effect is defined as
\begin{equation}
 g(a,a')=E_{C_e}\!\left[\int\mu_Y(a,m,C_e)\,dF_{M\mid A=a',C_e}(m)\right].
\label{eq:foundation:o3-pathway-estimand}
\end{equation}
With $\Delta a=a_1-a_0$ and $\overline M_0=E_{C_e}[\mu_M(a_0,C_e)]$,
\begin{equation}
\begin{aligned}
 D&=g(a_1,a_0)-g(a_0,a_0)
   =\Delta a\left(\theta_A+\theta_{AM}\overline M_0\right),\\
 I&=g(a_1,a_1)-g(a_1,a_0)
   =\alpha_A\Delta a\left(\theta_M+\theta_{AM}a_1\right),\\
 T&=D+I.
\end{aligned}
\label{eq:foundation:o3-component-forms}
\end{equation}
The mediated component equals the BMI-induced change in SpO$_2$ multiplied
by the conditional SpO$_2$ slope for cIMT, retaining the exposure--mediator
interaction.

Under the declared causal identification conditions, a BMI contrast of
$22.921\rightarrow28.397$ kg/m$^2$ in 425 HPP participants gives
\begin{equation}
 T=\underbrace{0.02678}_{D\,:\,\text{direct}}
   +\underbrace{(-0.00372)}_{I\,:\,\text{SpO}_2\text{-mediated}}
   =0.02306\ \text{mm}.
\label{eq:foundation:o3-numerical-decomposition}
\end{equation}
The total contrast has 95\% CI $[0.00400, 0.04285]$ mm; the direct and
mediated components have $[0.00737, 0.04683]$ and $[-0.00883, -0.00055]$ mm.
The BMI contrast corresponds to a 0.44848 percentage-point decrease in the
modeled SpO$_2$ mean, while the conditional SpO$_2$ slope for cIMT at the
higher BMI is positive; their product is the negative mediated component,
which partially offsets the positive direct component. Fixing SpO$_2$ at
96\% gives a controlled direct contrast of 0.02689 mm
(95\% CI $[0.00743, 0.04737]$), corresponding to a conditional BMI slope of
0.00491 mm per kg/m$^2$ (95\% CI $[0.00136, 0.00865]$).

\paragraph{Sensitivity to covariate adjustment.}
Expanded adjustment adds prior cIMT and alcohol; in the 294 participants with
complete covariates, the total contrast is 0.01962 mm
(95\% CI [$-0.00090$, $0.03747$]), and removing prior cIMT on the same 294
participants gives 0.03338 mm, a paired difference of 0.01376 mm
(95\% CI $[0.00554, 0.02572]$).

\Needspace{16\baselineskip}
\begin{wraptable}{r}{0.50\textwidth}
  \centering
  \caption{\textbf{HPP phenotype prediction on held-out participants.}}
  \label{tab:foundation:real-performance}
  \small
  \setlength{\tabcolsep}{2pt}
  \begin{tabular*}{\linewidth}{@{\extracolsep{\fill}}lrrrr@{}}
    \toprule
    Task & Test $n$ & NRMSE & $R^2$ & Cov.\ (\%) \\
    \midrule
    Future PWV & 166 & 0.959 & 0.378 & 89.2 \\
    Future cIMT & 172 & 0.926 & 0.176 & 83.1 \\
    Queried ATT & 1,380 & 0.915 & 0.155 & 92.0 \\
    Queried ApoB & 297 & 0.861 & 0.203 & 93.3 \\
    Queried GlycA & 297 & 0.782 & 0.332 & 87.2 \\
    Queried SpO$_2$ & 1,177 & 0.790 & 0.346 & 90.7 \\
    \bottomrule
  \end{tabular*}

  \vspace{0.35em}
  \begin{minipage}{\linewidth}
    \footnotesize
    Normalized root-mean-square error (NRMSE) uses the training-label standard deviation; coverage (Cov.) is
    measured for nominal 90\% prediction intervals. Queried outcomes are
    hidden from their own inputs.
  \end{minipage}
\end{wraptable}

\paragraph{Multi-phenotype prediction under incomplete inputs.}
Across the six HPP tasks in 10,401 participants, held-out $R^2$ is 0.378 for
future PWV and 0.176 for future cIMT
(Table~\ref{tab:foundation:real-performance}).
With 70\% additional random masking of the inputs, the five-seed mean
held-out $R^2$ is 0.248 for PWV and 0.136 for cIMT, and every prediction
retains its complete contribution decomposition.

\paragraph{Cross-population transfer of frozen relationship parameters.}
The predictive foundation with frozen HPP relationship parameters reaches an area
under the receiver operating characteristic curve (AUC) of 0.798
(95\% CI $[0.730, 0.859]$) and an inverse probability of censoring weighted
(IPCW) Brier score \citep{graf1999assessment} of 0.0473
(95\% CI $[0.0382, 0.0572]$) for 60-month all-cause mortality \citep{nchs2022lmf} on the NHANES
temporal test set (1,587 participants, 110 deaths). The relationship
coefficients keep their HPP estimates, and the remaining parameters of the
mortality task are estimated on NHANES development data; NHANES measures 9
of the 19 network nodes. On the same NHANES split, a target-domain refit of the predictive
foundation reaches AUC 0.799, matching a generalized additive model (0.798; paired $\Delta$AUC
$+0.001$ $[-0.002, 0.005]$) and a gradient-boosted tree (0.803; $-0.003$
$[-0.049, 0.033]$).

\section{A New Paradigm for Cumulative Causal Discovery}
\label{sec:new-paradigm}

DoAtlas-2 turns the biomedical foundation from static knowledge and isolated
outputs into a causal evidence state that evolves under external evidence:
every revision is attributable to results of population studies,
and the candidate questions, frozen designs and external selection records
form the complete denominator. The discovered vascular network itself
becomes the interpretable predictive foundation: every prediction is an
explicit closed-form expression that decomposes term by term as an exact
identity, and the same network yields closed-form direct and mediated
effects for temporally ordered questions. The vascular predictive foundation of
Section~\ref{sec:local-foundation} is the first disease-specific foundation
constructed here. Cumulative causal discovery, driven by
external evidence and reconstructable at every step, thereby becomes the
scientific working mode of biomedical AI.

\label{sec:main-text-end}
\bibliographystyle{iclr2027_conference}
\bibliography{references}

\clearpage
\appendix
\counterwithin{table}{section}
\counterwithin{figure}{section}
\counterwithin{equation}{section}

\section{Abbreviations and Measurement Definitions}
\label{app:data}

Table~\ref{tab:data:abbreviations} lists the abbreviations used in the paper
and, for the principal phenotypes, their measurement and unit in HPP. Pathway identifiers
and the panel symbols of
Figures~\ref{fig:findings:experimental-atlas}
and~\ref{fig:findings:conditional-validation} are defined with the figures in
Appendix~\ref{app:findings} to facilitate interpretation of the reported mechanistic findings.

\begin{table}[H]
  \centering
  \caption{\textbf{Abbreviations and measurement definitions.}}
  \label{tab:data:abbreviations}
  \footnotesize
  \setlength{\tabcolsep}{4pt}
  \renewcommand{\arraystretch}{0.95}
  \begin{tabular}{@{}l>{\raggedright\arraybackslash}p{4.3cm}>{\raggedright\arraybackslash}p{5.4cm}l@{}}
    \toprule
    Abbrev. & Full name & Measurement in HPP & Unit \\
    \midrule
    \multicolumn{4}{@{}l}{\emph{Phenotypes}} \\
    WHR & Waist-to-hip ratio & Waist circumference divided by hip circumference & ratio \\
    Waist (WC) & Waist circumference & Circumference at the narrowest part of the abdomen & cm \\
    Hip & Hip circumference & Circumference at the widest part of the hips and buttocks & cm \\
    Neck & Neck circumference & Circumference midway between the top of the shoulders and the base of the chin & cm \\
    Weight & Body weight & Anthropometry & kg \\
    BMI & Body mass index & Weight divided by squared height & kg/m$^2$ \\
    AF & Android fat & DXA android-region fat mass & g \\
    VAT & Visceral adipose tissue & DXA total-scan visceral fat mass & g \\
    ALT & Alanine aminotransferase & Clinical blood test & U/L \\
    GGT & $\gamma$-Glutamyl transferase & Clinical blood test & U/L \\
    TG & Triglycerides & Clinical blood test & mg/dL \\
    HbA1c & Glycated hemoglobin & Clinical blood test & \% \\
    LDL(-C), HDL(-C) & Low-, high-density lipoprotein cholesterol & Clinical blood test & mg/dL \\
    Sodium & Blood sodium & Clinical blood test & mmol/L \\
    ApoB & Apolipoprotein B & NMR metabolomics & g/L \\
    GlycA & Glycoprotein acetyls & NMR metabolomics & mmol/L \\
    ATT & Liver ultrasound attenuation & Attenuation coefficient of plane-wave liver ultrasound & dB/cm/MHz \\
    SpO$_2$ & Oxygen saturation & Mean nocturnal saturation during sleep monitoring & \% \\
    SBP, DBP & Systolic, diastolic blood pressure & Mean of available seated readings & mmHg \\
    PWV & Pulse wave velocity & Lower-limb thigh-to-ankle PWV, left, right or bilateral mean & m/s \\
    cIMT & Carotid intima--media thickness & Carotid ultrasound, left, right or bilateral mean & mm \\
    UACR & Urinary albumin-to-creatinine ratio & Urine test & mg/g \\
    \midrule
    \multicolumn{4}{@{}l}{\emph{Other abbreviations}} \\
    ABI & \multicolumn{3}{l}{Ankle--brachial index} \\
    AHI & \multicolumn{3}{l}{Apnea--hypopnea index} \\
    baPWV & \multicolumn{3}{l}{Brachial--ankle pulse wave velocity (external cohorts, Appendix~\ref{app:findings:replication})} \\
    DXA & \multicolumn{3}{l}{Dual-energy X-ray absorptiometry} \\
    ECG & \multicolumn{3}{l}{Electrocardiogram} \\
    HMO & \multicolumn{3}{l}{Health maintenance organization} \\
    IMT & \multicolumn{3}{l}{Intima--media thickness} \\
    NMR & \multicolumn{3}{l}{Nuclear magnetic resonance} \\
    HPP & \multicolumn{3}{l}{Human Phenotype Project} \\
    NHANES & \multicolumn{3}{l}{National Health and Nutrition Examination Survey} \\
    PBMC & \multicolumn{3}{l}{Peripheral blood mononuclear cell} \\
    RQ & \multicolumn{3}{l}{Research question} \\
    CI & \multicolumn{3}{l}{Confidence interval} \\
    CKD & \multicolumn{3}{l}{Chronic kidney disease} \\
    eGFR & \multicolumn{3}{l}{Estimated glomerular filtration rate} \\
    IHD & \multicolumn{3}{l}{Ischemic heart disease} \\
    SD & \multicolumn{3}{l}{Standard deviation} \\
    SoS & \multicolumn{3}{l}{Speed of sound (liver ultrasound)} \\
    AUC & \multicolumn{3}{l}{Area under the receiver operating characteristic curve} \\
    IPCW & \multicolumn{3}{l}{Inverse probability of censoring weighting} \\
    NRMSE & \multicolumn{3}{l}{Normalized root-mean-square error} \\
    \bottomrule
  \end{tabular}
\end{table}

\section{The DoAtlas-2 Self-Evolving Causal Foundation}
\label{app:foundation}

\subsection{Scientific state and notation}
\label{app:foundation:state}

DoAtlas-2 represents the scientific state of round $t$ as $S_t=(E_t,Q_t,A_t)$.
The causal evidence state $E_t$ binds each causal claim to its research
context and evidence source and retains its estimation uncertainty; supporting
and conflicting results coexist under their respective conditions of
applicability. The discovery frontier $Q_t$ contains unresolved research
questions and candidate long-chain causal pathways. The external evidence
requirements $A_t$ specify the measurements, study populations and temporal
structures still required to change the current scientific judgment.

\subsection{Discovery and selection of research questions}
\label{app:foundation:rq-discovery}
\label{app:foundation:prioritization}

The causal discovery agent examines how new results in $E_t$ extend,
intersect with or conflict with existing evidence. A candidate research
question is $q=(m,c)$, where $m$ is a testable causal-mechanism proposition
and $c$ its context of applicability. Each question retains its parent
results, the rationale for its generation and its pathway evidence, so that
it traces to the findings that produced it. Candidate questions lie within a
prespecified scientific scope $\Omega_t^Q$. $\operatorname{Priority}_t(q\mid S_t)$
weighs causal coherence, unresolved scientific uncertainty and expected
discriminative value, and $Q_t^\star\subseteq\Omega_t^Q$ denotes the questions
selected for evaluation in round $t$. All candidate questions and their
selection status are retained, so that every round has a complete
denominator.

\subsection{Study design and evidence requirements}
\label{app:foundation:study-design}
\label{app:foundation:evidence-requirements}

The full design
$D(q)=(\mathcal P_q,X_q,Y_q,\tau_q,\psi_q,\mathcal I_q)$ of
$q\in Q_t^\star$ grounds the population $\mathcal P_q$, exposure $X_q$ and
outcome $Y_q$ in HPP measurements and declares windows $\tau_q$, causal
estimand $\psi_q$ and identification conditions $\mathcal I_q$. Screening
designs fix the hypothesized path and its variables, and their estimates stay
outside $E_t$. Analysis units realize $D(q)$, each with its own population, windows and
adjustment set. They include reverse-direction, falsification, sensitivity and
heterogeneity analyses. Each unit declares its evidence class. Strictly
ordered units require measurement times that follow the path order for every
analyzed participant, and weakly ordered units declare measurement windows and
treat their temporal order as ambiguous; together they form the temporally
ordered designs. Cross-sectional units, most with same-day or symmetric
windows, assert no temporal order. Primary outputs set
relation states (Appendix~\ref{app:foundation:evidence-rules}); secondary and
other outputs add evidence without setting one.

An analysis unit ends with an estimate, a non-estimable result with its
reason, or a mismatch between the designed and the available measurement. The
non-estimability reason names the evidence still required: another data
resource for a missing measurement, a larger population or longer follow-up
for insufficient support, and measured confounders or an instrument for
non-identification. A non-informative estimate requires a larger sample or a
more precise design. A question that only intervention, genetic instruments or
external experiments can answer defines an external validation requirement for subsequent rounds of scientific inquiry.

\subsection{Long-chain causal pathways}
\label{app:foundation:long-chain}

A long-chain causal pathway $\pi=(Z_0\rightarrow\cdots\rightarrow Z_K;c)$
composes segments $Z_{k-1}\rightarrow Z_k$, each evidenced by the records in
$E_t$ for its ordered concept pair. Composition in a context $c$ fixing a
population and an evidence class is compatible when each shared node is one
measured variable in both adjacent segments and each segment has a record in
that evidence class whose population covers that of $c$. Strictly ordered
composition requires
$\operatorname{time}(Z_0)<\cdots<\operatorname{time}(Z_K)$ for each
participant.

The serial path association under design $D$ is
$\beta_\pi^{D}=\prod_k\beta_k^{D}$, where $\beta_k^{D}$ is the coefficient of
$Z_{k-1}$ in the equation for $Z_k$ or its conditional slope at declared input
values. Serial estimates and path products in the main text take this form,
each computed with its interval within one design. Each segment carries the
relation state of its concept pair, and the path carries no state. The
causal pathway estimand $\psi_\pi(c)$ is given by a design declaring
identification conditions jointly for all segments.

\subsection{External scientific selection}
\label{app:foundation:selection}

External scientific selection evaluates each $q\in Q_t^\star$ on the results
of its screening or full design $D(q)$. With $\mathcal U(q)$ the analysis units of $D(q)$ and
$\mathcal U_t(q)\subseteq\mathcal U(q)$ those executed on HPP data by round
$t$, each $u\in\mathcal U_t(q)$ has a result $v_{t,u}$: an estimate of each
output with its uncertainty, or the reason no estimate exists. The results and
selection record of $q$ are
\begin{equation}
  V_t(q)=\{v_{t,u}:u\in\mathcal U_t(q)\},
  \qquad
  \sigma_t(q)=\Pi_t\bigl(q,D(q),V_t(q)\bigr),
  \label{eq:external-selection}
\end{equation}
where the adjudication policy $\Pi_t$ applies admission and record-state
rules of Appendix~\ref{app:foundation:evidence-rules}.
$\sigma_t(q)$ lists each unit's result and each admitted estimate of $q$ with
relation and evidence class; a primary estimate carries a record
state determined by the estimate, its design and the literature direction
reference.

\subsection{Evidence-state rules}
\label{app:foundation:evidence-rules}

Admission to $E_{t+1}$ requires an HPP estimate to be an edge-level
coefficient from an analysis unit with an estimated result. It needs a point
estimate $\hat\psi_i$, an interval $[L_i,U_i]$, and a scale and estimand
declared in its design; its endpoints must map to one directed concept pair
$x\to y$, and its sample must support the reported interval.

Precision classifies record $i$ by its interval $[L_i,U_i]$ relative to the
null value $\psi^0_i$ and to the equivalence margin declared in its design.
The record is an informative non-null estimate when its interval excludes
$\psi^0_i$ and meets the multiplicity criterion of its design, an informative
null when its interval contains $\psi^0_i$ and lies within the equivalence
margin, and non-informative otherwise. The HPP direction $s_i$
is the sign of $\hat\psi_i-\psi^0_i$ in the orientation of the concepts. The
literature direction reference $\rho_{xy}$ pools the literature reports on
both endpoint concepts and their synonyms: it is increasing or decreasing when
all directional reports agree, conflicted when both directions occur,
non-directional when the reports state no direction and absent when no report
exists.

An evidence context holds the primary records of one relation within one
evidence class. Its state aggregates these records, and the relation state
aggregates the evidence-context states (Table~\ref{tab:appb:evidence-states})
without weighting by sample size or evidence class. An unresolved state means
that the direction of the relation is not yet established in every evidence
context.

\begin{table}[H]
  \centering
  \caption{\textbf{Record, evidence-context and relation states.} Rows apply
  in order.}
  \label{tab:appb:evidence-states}
  \small
  \setlength{\tabcolsep}{4pt}
  \begin{tabular}{@{}>{\raggedright\arraybackslash}p{3.3cm}>{\raggedright\arraybackslash}p{4.6cm}>{\raggedright\arraybackslash}p{2.3cm}>{\raggedright\arraybackslash}p{3.3cm}@{}}
    \toprule
    State & Record & Evidence context & Relation \\
    \midrule
    No population-derived state & -- & -- & No admitted primary record \\
    Supported & Informative non-null; $s_i$ equals a directional $\rho_{xy}$,
      or $\rho_{xy}$ is conflicted, non-directional or absent
      & All records supported & All contexts supported \\
    Challenged & Informative non-null with $s_i$ opposing a directional
      $\rho_{xy}$, or informative null
      & All records challenged & All contexts challenged \\
    Context-dependent & -- & -- & At least one supported and one challenged context \\
    Unresolved & Non-informative & Otherwise & Otherwise \\
    \bottomrule
  \end{tabular}
\end{table}

\subsection{Joint state revision}
\label{app:foundation:revision}
\label{app:foundation:interpretability}

Selection records of round $t$ form $\Sigma_t=\{\sigma_t(q):q\in Q_t^\star\}$,
and the transition $S_t\xrightarrow{\Sigma_t}S_{t+1}$ factors into three
updates:
\begin{equation}
  \begin{aligned}
    E_{t+1}&=\Phi^E_t\bigl(E_t,\Sigma_t\bigr),\\
    Q_{t+1}&=\Phi^Q_t\bigl(Q_t,E_{t+1},\Sigma_t\bigr),\\
    A_{t+1}&=\Phi^A_t\bigl(A_t,E_{t+1},Q_{t+1},\Sigma_t\bigr).
  \end{aligned}
  \label{eq:joint-state-transition}
\end{equation}
Within $\Phi^E_t$, the context and relation aggregation of
Table~\ref{tab:appb:evidence-states} sets relation states.

Change sets $\Delta^E_t$, $\Delta^Q_t$ and $\Delta^A_t$ collect the elements
of $E$, $Q$ and $A$ added or revised from round $t$ to round $t+1$, and
$\Delta_t$ is their union. The source set
$\operatorname{Src}(\delta)\subseteq\Sigma_t$ of a change $\delta\in\Delta_t$
comprises the selection records cited as its origin. The revision of round
$t$ is attributable when $\operatorname{Src}(\delta)\neq\varnothing$ for every
$\delta\in\Delta_t$, and joint when all three change sets are non-empty.

\section{Scientific Evaluation of Self-Evolution}
\label{app:loop}

\subsection{Scope and counting units}
\label{app:loop:scope}

A research question (RQ) is an ordered concept sequence to be tested,
with its context. Candidate RQs form the research space, and an RQ is
evaluated when its prespecified analyses have been completed on HPP data. This
report includes 88,731 result estimates. Screening is an association
test that precedes full evaluation, with standardized exposure, mediator and
outcome and no covariate adjustment; screening measures the vascular outcomes on the
left side. A screened question receives
screening-level statistical support when its indirect association and each of
its links remain significant under false-discovery-rate control and its path is
free of measurement artifacts, such as two measurements of one construct or one
instrument. A pathway is a distinct ordered concept
sequence, and a candidate directed relation is an ordered pair of adjacent
concepts in a pathway; pathways share concepts and relations, so the counts
are reported separately and are not additive. A parent--child link connects a
question to a parent it records: Round-2 questions record Round-1 parent
questions, and Round-3 questions record parent results.

Table~\ref{tab:loop:round-scope} reports the catalog, screening and
parent--child links by round. The 4,014 Round-3 pathway questions comprise
1,546 vascular, 1,329 early-glycemic and 1,139 hepatic-metabolic questions.
Of the 86 supported Round-1 questions, 61 became parents of Round-2 questions,
and none of the 89 unsupported questions did; each Round-2 question records two
or three parents.

\begin{table}[H]
  \centering
  \caption{\textbf{Catalog, screening and parent--child links by round.}}
  \label{tab:loop:round-scope}
  \small
  \begin{tabular}{@{}lrrr@{}}
    \toprule
     & Round 1 & Round 2 & Round 3 \\
    \midrule
    Catalog pathways in the round & 175 & 904 & 3,078 \\
    New distinct pathways & 175 & 904 & 2,862 \\
    Cumulative distinct pathways & 175 & 1,079 & 3,941 \\
    Cumulative concepts & 17 & 33 & 77 \\
    Cumulative candidate directed relations & 44 & 271 & 517 \\
    Cumulative outcome concepts & 3 & 4 & 53 \\
    \midrule
    Screened questions & 175 & 904 & 4,014 \\
    Screening-level support & 86 & 670 & 1,736 \\
    Fully evaluated questions in this report & -- & 700 & 285 \\
    \midrule
    Questions recording parents & -- & 904 & 2,762 \\
    Distinct parent questions & -- & 61 & -- \\
    Distinct parent results & -- & -- & 2,616 \\
    Parent--child links & -- & 2,535 & 5,371 \\
    \bottomrule
  \end{tabular}
\end{table}

\subsection{From results to successor questions}
\label{app:loop:attribution}

The 904 Round-2 questions fall into five families by how they extend
Round-1 results (Table~\ref{tab:loop:successor-families}). Of the 2,535
parent--child links, 2,294 share the parent's outcome and 1,441 share at least
one directed relation with the parent. Successor questions insert a mediator
upstream of blood pressure, replace blood pressure with a competing
mediator, or test blood pressure as the shared intermediate node.

An evaluation of the 904 Round-2 screening results identifies 120 pathway
contrasts. In 81, a serial bridge is unsupported while its nested shorter path
is supported; in 30, the same exposure--mediator structure reaches different
conclusions for different vascular outcomes; in 9, the serial path is supported
while its nested shorter path is not. Ten of these contrasts receive a
scientific interpretation: 4 endpoint divergences, 3 serial signals with
unsupported nested paths and 3 ApoB bridges. The first seven contrasts yield seven
successor-question proposals, all deepening existing Round-2 questions: 4
require a paired comparison with the ankle--brachial index outcome and 3 a
comparison with the nested shorter path. The 3 ApoB bridges remain open
directions, and a further evaluation of them yields 11 contextual
interpretations and 11 successor-question proposals, 10 deepening existing
questions and 1 a new screening question
(Appendices~\ref{app:loop:case-neck-apob} and~\ref{app:loop:case-whr-endpoint}).

The first 30 Round-3 pathway proposals have 15 outcome constructs: 5 were
intermediate phenotypes in earlier rounds (HbA1c, TG, GlycA, hepatic steatosis
and apnea--hypopnea index), 3 are earlier outcomes (SBP, PWV and cIMT) and 7
did not appear in earlier rounds (UACR, estimated glomerular filtration rate, chronic kidney
disease, ischemic heart disease, stroke, ischemic stroke and deep vein
thrombosis), with UACR the outcome of 6 proposals. The proposals comprise 19
single relations, 8 three-node paths and 3 five-node long chains, together with
8 requests for common-cause, fork and measurement diagnostics to assess alternative explanations and artifacts.

\begin{table}[H]
  \centering
  \caption{\textbf{Five families of Round-2 successor questions.} Parents:
  distinct Round-1 parent questions; a parent can have successors in several
  families. Support: screening-level statistical support.}
  \label{tab:loop:successor-families}
  \footnotesize
  \setlength{\tabcolsep}{3pt}
  \begin{tabular}{@{}>{\raggedright\arraybackslash}p{2.3cm}r>{\raggedright\arraybackslash}p{2.7cm}>{\raggedright\arraybackslash}p{2.5cm}>{\raggedright\arraybackslash}p{1.9cm}rrr@{}}
    \toprule
    Family & $n$ & Path form & Main intermediates & Outcomes & Parents & Links & Support \\
    \midrule
    Serial metabolic--hemodynamic & 326
      & Exposure $\to$ mediator $\to$ SBP/DBP $\to$ outcome
      & GlycA, TG, HbA1c, HDL, LDL, ApoB & PWV 166, cIMT 160 & 44 & 892 & 252 (77.3\%) \\
    Competing mediator bypassing blood pressure & 264
      & Exposure $\to$ mediator $\to$ outcome
      & Glucose, GlycA, HbA1c, TG, heart rate, HDL & PWV 137, cIMT 127 & 42 & 735 & 166 (62.9\%) \\
    Hemodynamic convergence & 142
      & Exposure $\to$ SBP/DBP $\to$ outcome
      & DBP, SBP & Ankle--brachial index 54, PWV 47, cIMT 41 & 41 & 415 & 120 (84.5\%) \\
    Liver-related & 108
      & Exposure $\to$ liver phenotype $\to$ SBP/DBP $\to$ outcome
      & Hepatic steatosis, ALT, GGT & PWV 54, cIMT 54 & 28 & 309 & 95 (88.0\%) \\
    Sleep-related & 64
      & Exposure $\to$ sleep phenotype $\to$ SBP/DBP $\to$ outcome
      & Total sleep time, apnea--hypopnea index, mean SpO$_2$, sleep efficiency
      & PWV 32, cIMT 32 & 17 & 184 & 37 (57.8\%) \\
    \midrule
    Total & 904 & & & & 61 & 2,535 & 670 (74.1\%) \\
    \bottomrule
  \end{tabular}
\end{table}

\subsection{Cases of result-driven follow-up}

\subsubsection{Adiposity--liver--vascular follow-up}
\label{app:loop:case-h1}

The two Round-1 parents received screening-level support, with standardized
indirect associations of 0.2124 (WHR--SBP--PWV) and 0.0836 (WC--WHR--SBP--PWV).
The Round-2 successor WHR--ALT--DBP--PWV belongs to the liver-related family
(Table~\ref{tab:loop:successor-families}); it shares the WHR and PWV nodes with
both parents but none of their directed relations.

The pathway has three primary estimates obtained under three distinct
analytical designs (Table~\ref{tab:loop:h1-estimates}).
Round-2 screening gave a standardized indirect association of 0.0149 in 7,452
participants without covariate adjustment. The Round-2 primary design, H1 in
Figure~\ref{fig:findings:experimental-atlas}c, used symmetric windows of
$\pm180$, $\pm90$ and $\pm30$ days between adjacent measurements and gave a
serial estimate of 0.01099, matching the product $0.2323\times0.1088\times0.4347$
of its component coefficients. The joint model of
Figure~\ref{fig:findings:network-overview}c and
Section~\ref{sec:findings:vascular-convergence} used $\pm365$-day windows for all
three relations and estimated the ALT branch, the DBP branch and the direct
association in one model.

The two full analyses agree on the branches. In the Round-2 primary design, the
WHR--ALT--PWV branch product conditional on DBP is 0.001912 (95\% CI
[$-0.005126$, $0.008951$]) and the WHR--DBP--PWV branch is 0.09709
([0.07721, 0.117]); in the joint model, the ALT bypass product is 0.00219
([$-0.002308$, $0.006688$]) and the DBP bypass product is 0.09771
([0.08289, 0.1125]). Appendix~\ref{app:findings:h1-joint} reports the window,
negative-control, sex, extended-adjustment, interaction and
unmeasured-confounding analyses of the joint model.

\begin{table}[H]
  \centering
  \caption{\textbf{Primary estimates of the WHR--ALT--DBP--PWV pathway.}}
  \label{tab:loop:h1-estimates}
  \footnotesize
  \setlength{\tabcolsep}{4pt}
  \begin{tabular}{@{}>{\raggedright\arraybackslash}p{2.9cm}>{\raggedright\arraybackslash}p{2.0cm}r>{\raggedright\arraybackslash}p{2.7cm}>{\raggedright\arraybackslash}p{3.9cm}@{}}
    \toprule
    Analysis & Windows & Estimate & 95\% CI & Model \\
    \midrule
    Round-2 screening & -- & 0.0149 & -- & Standardized indirect association without covariate adjustment; $n=7{,}452$, $q=7\times10^{-16}$ \\
    Round-2 primary design (H1, Figure~\ref{fig:findings:experimental-atlas}c) & $\pm180$, $\pm90$, $\pm30$ d & 0.01099 & [0.006451, 0.01552] & Serial estimate; adjusted for age and sex \\
    Joint model (Figure~\ref{fig:findings:network-overview}c) & $\pm365$ d each & 0.01073 & [0.007575, 0.01389] & Estimated jointly with the ALT branch, DBP branch and direct association; adjusted for age and sex \\
    \bottomrule
  \end{tabular}
\end{table}

\subsubsection{Neck circumference, ApoB and blood pressure}
\label{app:loop:case-neck-apob}

The serial questions Neck--ApoB--DBP--PWV and Neck--ApoB--SBP--PWV were not
supported at screening, with indirect associations of 0.0049 and 0.0043
($n=750$). Four of their nested shorter paths were supported: Neck--DBP--PWV and
Neck--SBP--PWV gave 0.1536 and 0.2289 ($n=2{,}665$), and the downstream
ApoB--DBP--PWV and ApoB--SBP--PWV gave 0.0826 and 0.0845 ($n=1{,}388$), indirect
associations that exceed the total association. Neck--ApoB--PWV was not
supported, whereas the serial paths with AHI, TG or GlycA instead of ApoB were
(Table~\ref{tab:loop:case-neck}). The DBP- and SBP-branch contrasts yield seven
successor-question proposals: six deepening existing Round-2 questions
(Neck--DBP--PWV, Neck--SBP--PWV, ApoB--DBP--PWV, ApoB--SBP--PWV,
Neck--AHI--DBP--PWV and Neck--AHI--SBP--PWV) and one new screening question,
Neck--AHI--DBP.

\begin{table}[H]
  \centering
  \caption{\textbf{Screening results of the neck-circumference case.} AHI, apnea--hypopnea index. Support: screening-level statistical support.}
  \label{tab:loop:case-neck}
  \small
  \begin{tabular}{@{}lrrrc@{}}
    \toprule
    Path & $n$ & Indirect association & $q$ & Support \\
    \midrule
    Neck--ApoB--DBP--PWV & 750 & 0.0049 & 0.066 & No \\
    Neck--DBP--PWV & 2,665 & 0.1536 & $6\times10^{-51}$ & Yes \\
    ApoB--DBP--PWV & 1,388 & 0.0826 & $1\times10^{-8}$ & Yes \\
    Neck--AHI--DBP--PWV & 1,910 & 0.0423 & $2\times10^{-16}$ & Yes \\
    Neck--TG--DBP--PWV & 1,967 & 0.0208 & $2\times10^{-8}$ & Yes \\
    Neck--GlycA--DBP--PWV & 750 & 0.0187 & $4\times10^{-4}$ & Yes \\
    Neck--ApoB--SBP--PWV & 750 & 0.0043 & 0.075 & No \\
    Neck--SBP--PWV & 2,665 & 0.2289 & $6\times10^{-83}$ & Yes \\
    ApoB--SBP--PWV & 1,388 & 0.0845 & $5\times10^{-8}$ & Yes \\
    Neck--AHI--SBP--PWV & 1,910 & 0.0290 & $1\times10^{-10}$ & Yes \\
    Neck--TG--SBP--PWV & 1,967 & 0.0139 & $9\times10^{-5}$ & Yes \\
    Neck--GlycA--SBP--PWV & 750 & 0.0113 & 0.012 & Yes \\
    Neck--ApoB--PWV & 751 & 0.0042 & 0.26 & No \\
    \bottomrule
  \end{tabular}
\end{table}

\subsubsection{Waist-to-hip ratio, ApoB and vascular endpoints}
\label{app:loop:case-whr-endpoint}

The serial question WHR--ApoB--DBP--cIMT was not supported at screening
(indirect association 0.0012, $n=1{,}117$), whereas its nested WHR--DBP--cIMT
was (0.0334, $n=7{,}925$). For cIMT, the downstream ApoB--DBP--cIMT and
the serial paths with TG or GlycA in place of ApoB were supported, and
WHR--ApoB--cIMT was not; the same mediator sequence was supported for PWV
(WHR--ApoB--DBP--PWV, 0.0073, $n=1{,}388$; Table~\ref{tab:loop:case-whr}). This
contrast yields four successor-question proposals, all deepening Round-2 questions: WHR--DBP--cIMT, ApoB--DBP--cIMT, WHR--TG--DBP--cIMT and
WHR--ApoB--DBP--PWV, carrying the mediator sequence to PWV.

\begin{table}[H]
  \centering
  \caption{\textbf{Screening results of the waist-to-hip-ratio case.} Support: screening-level statistical support.}
  \label{tab:loop:case-whr}
  \small
  \begin{tabular}{@{}lrrrc@{}}
    \toprule
    Path & $n$ & Indirect association & $q$ & Support \\
    \midrule
    WHR--ApoB--DBP--cIMT & 1,117 & 0.0012 & 0.073 & No \\
    WHR--ApoB--SBP--cIMT & 1,117 & 0.0021 & 0.062 & No \\
    WHR--DBP--cIMT & 7,925 & 0.0334 & $2\times10^{-19}$ & Yes \\
    ApoB--DBP--cIMT & 1,117 & 0.0208 & 0.0011 & Yes \\
    ApoB--SBP--cIMT & 1,117 & 0.0359 & $2\times10^{-4}$ & Yes \\
    WHR--TG--DBP--cIMT & 6,202 & 0.0064 & $4\times10^{-11}$ & Yes \\
    WHR--TG--SBP--cIMT & 6,200 & 0.0088 & $2\times10^{-15}$ & Yes \\
    WHR--GlycA--DBP--cIMT & 1,117 & 0.0050 & 0.0038 & Yes \\
    WHR--GlycA--SBP--cIMT & 1,117 & 0.0056 & 0.020 & Yes \\
    WHR--ApoB--cIMT & 1,118 & 0.0027 & 0.31 & No \\
    WHR--ApoB--DBP--PWV & 1,388 & 0.0073 & 0.0018 & Yes \\
    WHR--ApoB--SBP--PWV & 1,388 & 0.0067 & 0.0032 & Yes \\
    WHR--DBP--PWV & 9,174 & 0.1324 & $1\times10^{-138}$ & Yes \\
    WHR--ApoB--PWV & 1,390 & 0.0052 & 0.13 & No \\
    \bottomrule
  \end{tabular}
\end{table}

\subsubsection{HbA1c, SBP and UACR}
\label{app:loop:case-glycemic-renal}

The Round-3 HbA1c--SBP--UACR proposal inserts SBP between HbA1c and the new
renal outcome UACR. It asks whether the association of HbA1c with UACR is
consistent with a blood-pressure-mediated path or arises mainly outside blood
pressure; the competing explanation treats
blood pressure as a common cause or a parallel consequence rather than a serial
mediator. The proposal pairs the pathway with the single-relation question
HbA1c--UACR and requires the serial, direct and common-cause explanations to be
evaluated separately. Of the 29 Round-2 screening paths containing the
HbA1c--SBP link, 27 received screening-level support; the 2 unsupported paths
both start from total sleep time. Among the first 30 Round-3 pathway proposals, the other
five with UACR as the outcome are GlycA--UACR, oxygen desaturation index--UACR,
AHI--GlycA--UACR, HbA1c--UACR and VAT--HbA1c--UACR.

\subsection{Evidence-state revision}
\label{app:loop:state-revision}

The 88,731 result estimates in this report and the joint analyses of
Section~\ref{sec:key-findings} contribute 12,202 edge-level coefficient
records, 434 of them from the joint analyses; 11,241 map to directed concept
pairs, and 2,169 are admitted as evidence. The admitted records comprise 1,026
cross-sectional, 884 weakly ordered and 259 strictly ordered records, 1,177 of
them informative. The 725 primary records that set relation states come from
214 RQs and one joint analysis. Of the 259 relations with HPP evidence, 200
acquire a population-derived state; the HPP evidence of the other 59 consists
of secondary or unspecified records only.

Of the 81 supported relations, 16 match a directional literature reference,
49 have no counterpart in the literature-derived evidence network
(Table~\ref{tab:loop:no-counterpart}), and the remaining 16 have a conflicted
or non-directional literature reference. Where the literature is conflicted,
the HPP direction mostly agrees with the majority of literature reports
(Table~\ref{tab:loop:conflicted}). Each of the 2 challenged relations rests on
one directional literature report and one primary record: gynoid fat
mass--LDL cholesterol, with decreasing literature and a standardized HPP
estimate of 0.072 (95\% CI [0.035, 0.110], weakly ordered), and sleep
efficiency--heart rate, with increasing literature and an estimate of $-0.056$
on the original measurement scale ([$-0.095$, $-0.017$], cross-sectional). Of the
117 unresolved relations, 38 are supported in at least one evidence context:
33 in the cross-sectional context, 5 of which also appear in a temporally ordered
context, and 5 only in temporally ordered contexts. Another 78 are unresolved
in every context, 52 of them with no informative primary record, and the last
holds a challenged cross-sectional context and two unresolved temporally
ordered contexts.

\begin{table}[H]
  \centering
  \caption{\textbf{Supported relations with conflicted literature.} Literature
  reports: papers with increasing/decreasing claims; a paper can count in both
  columns. Contexts: S, strictly ordered; W, weakly ordered; C,
  cross-sectional.}
  \label{tab:loop:conflicted}
  \small
\begin{tabular}{@{}lrlcr@{}}
    \toprule
    Relation & Literature reports & HPP direction & Contexts & Primary records \\
    \midrule
    BMI--SBP & 29/1 & Increase & W & 2 \\
    BMI--DBP & 22/2 & Increase & C & 2 \\
    BMI--LDL & 7/4 & Increase & W & 2 \\
    Weight--SBP & 6/1 & Increase & W & 1 \\
    LDL--DBP & 4/1 & Increase & C & 1 \\
    Waist--HbA1c & 2/1 & Increase & S & 1 \\
    WHR--HbA1c & 2/1 & Increase & S, C & 4 \\
    BMI--heart rate & 2/1 & Increase & W & 3 \\
    Hip--DBP & 1/1 & Increase & C & 3 \\
    BMI--HDL & 1/9 & Decrease & W, C & 3 \\
    HDL--DBP & 2/3 & Decrease & W, C & 2 \\
    \bottomrule
  \end{tabular}
\end{table}

\begin{table}[H]
  \centering
  \caption{\textbf{Supported relations with no counterpart in the
  literature-derived evidence network.} Direction: HPP direction. Contexts as
  in Table~\ref{tab:loop:conflicted}. Rec.: primary records. ABI, ankle--brachial
  index; AHI, apnea--hypopnea index; liver SoS, liver ultrasound speed of sound.}
  \label{tab:loop:no-counterpart}
  \footnotesize
  \setlength{\tabcolsep}{3pt}
  \begin{tabular}{@{}lccr@{\hspace{12pt}}lccr@{}}
    \toprule
    Relation & Dir. & Contexts & Rec. & Relation & Dir. & Contexts & Rec. \\
    \midrule
    VAT--PWV & + & W, C & 15 & GlycA--DBP & + & C & 1 \\
    DBP--PWV & + & W, C & 11 & Gynoid fat--ABI & + & C & 1 \\
    Neck--PWV & + & W, C & 7 & Gynoid fat--DBP & + & C & 1 \\
    Total lean mass--cIMT & + & W, C & 6 & Gynoid fat--HbA1c & + & C & 1 \\
    BMI--ApoB & + & W, C & 5 & Heart rate--ABI & $-$ & C & 1 \\
    Neck--cIMT & + & W, C & 5 & Liver SoS--cIMT & + & C & 1 \\
    VAT--SBP & + & W, C & 5 & Liver SoS--DBP & $-$ & C & 1 \\
    VAT--TG & + & S, W, C & 3 & SpO$_2$--PWV & + & C & 1 \\
    Waist--ABI & + & C & 3 & SpO$_2$--SBP & + & W & 1 \\
    AF--heart rate & + & W, C & 2 & Neck--ABI & + & C & 1 \\
    BMI--urate & + & W, C & 2 & Neck--liver SoS & $-$ & C & 1 \\
    Gynoid fat--SBP & + & C & 2 & Neck--SBP & + & C & 1 \\
    Hip--ABI & + & C & 2 & Sleep efficiency--DBP & $-$ & C & 1 \\
    Liver SoS--PWV & $-$ & C & 2 & VAT--heart rate & + & W & 1 \\
    SpO$_2$--sleep heart rate & $-$ & C & 2 & VAT--liver SoS & $-$ & C & 1 \\
    TG--DBP & + & W, C & 2 & VAT--SpO$_2$ & $-$ & S & 1 \\
    TG--SBP & + & W, C & 2 & Waist--sleep efficiency & $-$ & C & 1 \\
    VAT--GlycA & + & W, C & 2 & Waist--total sleep time & $-$ & C & 1 \\
    WHR--ABI & + & W & 2 & Waist--urate & + & C & 1 \\
    WHR--DBP & + & W, C & 2 & WHR--ApoB & + & C & 1 \\
    WHR--GlycA & + & W, C & 2 & WHR--GGT & + & S & 1 \\
    WHR--ATT & + & W, C & 2 & Weight--ApoB & + & W & 1 \\
    AF--ATT & + & C & 1 & Weight--GlycA & + & C & 1 \\
    AHI--heart rate & + & C & 1 & Weight--heart rate & + & C & 1 \\
    GGT--TG & + & C & 1 &  &  &  &  \\
    \bottomrule
  \end{tabular}
\end{table}

\section{Complete Evidence for the Key Findings}
\label{app:findings}

\subsection{Definitions for \texorpdfstring{Figure~\ref{fig:findings:experimental-atlas}}{Figure 4}}
\label{app:findings:experimental-atlas}
\renewcommand{\topfraction}{0.95}\renewcommand{\bottomfraction}{0.6}
\renewcommand{\textfraction}{0.05}\renewcommand{\floatpagefraction}{0.8}
\setcounter{topnumber}{3}\setcounter{bottomnumber}{2}\setcounter{totalnumber}{4}
\makeatletter\setlength{\@fptop}{0pt}\setlength{\@fpsep}{\floatsep}\setlength{\@fpbot}{0pt plus 1fil}\makeatother
\suppressfloats[t]
\paragraph{Pathway codes and plotted values.}
Table~\ref{tab:findings:pathway-codes} lists the ordered path of each pathway
in Figure~\ref{fig:findings:experimental-atlas} together with the design class
and adjustment set of its primary analysis, and
Table~\ref{tab:findings:other-codes} lists the other codes.
Tables~\ref{tab:findings:fig4-c} to~\ref{tab:findings:fig4-j} list all 184
values plotted in panels c--j as axis readings with at most four significant
digits, together with the scale and axis factor $\times10^{k}$ of each axis; a reading multiplied by $10^{k}$ gives the estimate
on the scale of the underlying analysis.

\begin{table}[btp]
  \centering
  \caption{\textbf{Pathways and primary analyses in
  Figure~\ref{fig:findings:experimental-atlas}.} Design class: W, weakly ordered;
  L, strictly ordered. Age enters linearly unless age$^2$ is listed; cohort and study
  are HPP labels.}
  \label{tab:findings:pathway-codes}
  \footnotesize
  \setlength{\tabcolsep}{4pt}
  \renewcommand{\arraystretch}{0.92}
\begin{tabular}{@{}l>{\raggedright\arraybackslash}p{4.0cm}c>{\raggedright\arraybackslash}p{6.6cm}@{}}
    \toprule
    Code & Ordered path & Design & Adjustment set \\
    \midrule
    H1 & WHR--ALT--DBP--PWV & W & Age, sex \\
    H2 & Hip--ALT--DBP--PWV & W & Age, age$^2$, sex, study \\
    H3 & Waist--ALT--SBP--cIMT & W & Age, sex, cohort \\
    H4 & WHR--ATT--SBP--cIMT & W & Age, sex \\
    H5 & Neck--ATT--SBP--PWV & W & Age, age$^2$, sex \\
    L1 & AF--TG--DBP--PWV & W & Age, sex, study \\
    L2 & VAT--ApoB--SBP--PWV & W & Age, age$^2$, sex \\
    L3 & Weight--ApoB--DBP--PWV & W & Age, sex, study, height \\
    L4 & Hip--ApoB--DBP--PWV & W & Age, age$^2$, sex, cohort, height, current smoking \\
    L5 & WHR--TG--SBP--cIMT & W & Age, age$^2$, sex, cohort \\
    L6 & GGT--TG--SBP--PWV & W & Age, sex \\
    L7 & ALT--TG--PWV & L & Age, sex \\
    O1 & Waist--SpO$_2$--SBP--cIMT & W & Age, sex, cohort \\
    O2 & SpO$_2$--GlycA--PWV & W & Age, sex, prior BMI \\
    O3 & BMI--SpO$_2$--cIMT & L & Age, sex, cohort, current smoking, education \\
    \bottomrule
  \end{tabular}
\end{table}

\begin{wraptable}{r}{0.42\textwidth}
  \centering
  \caption{\textbf{Other codes in Figure~\ref{fig:findings:experimental-atlas}}
  with the research question (RQ) holding each model and the panels showing it.
  R1 is a research question in panel a; H1r and R1r are reverse models, and
  L6$^{\dagger}$ is strictly ordered.}
  \label{tab:findings:other-codes}
  \footnotesize
  \setlength{\tabcolsep}{3pt}
  \renewcommand{\arraystretch}{0.92}
  \begin{tabular*}{\linewidth}{@{\extracolsep{\fill}}llcc@{}}
    \toprule
    Code & Ordered path & RQ & Panels \\
    \midrule
    R1 & WHR--GlycA--SBP--PWV & R1 & a \\
    T1 & ALT--DBP--PWV & L7 & e, f \\
    H1r & PWV--ALT--WHR & H1 & e \\
    R1r & PWV--SBP--WHR & R1 & e \\
    O3$^{+}$ & BMI--SpO$_2$--SBP--cIMT & O3 & e \\
    L6$^{\dagger}$ & GGT--TG--PWV & L6 & h \\
    \bottomrule
  \end{tabular*}
\end{wraptable}
\paragraph{Coverage and network (a, b).}
Panel a has one sector per research question, labeled with its pathway code
and holding all analyses of that question, including models along other
paths such as T1 in the L7 sector. Outer dots are the reported outputs of
each question, one per estimate or value, in three bands per sector,
clockwise C, W and L; darker dots are outputs plotted individually.
Inner bars count analysis units per design class, clockwise C, W and L:
designed units (pale), units with results (colored) and units with
individually plotted outputs (dark core). C denotes cross-sectional
(complete-case) designs, W weakly ordered designs with declared measurement
windows and L strictly ordered designs
(Appendix~\ref{app:foundation:study-design}). Panel b joins pathways H1--H5,
L1--L7 and O1--O3 into one directed network with arrows in path order; colors
mark the liver (H), lipid (L) and sleep (O) modules, and grey marks nodes and
connections shared by modules, highlighting overlap among the three modules.

\needspace{18\baselineskip}
\begin{wraptable}{r}{0.5\textwidth}
  \centering
  \caption{\textbf{Parameters in Figure~\ref{fig:findings:experimental-atlas}c,d}
  from linear models along a four-node path $A\to M_1\to M_2\to Y$ or a
  three-node path $A\to M\to Y$, with intercepts and adjustment sets omitted.
  For L7, T is D$+$S.}
  \label{tab:findings:panel-symbols}
  \footnotesize
  \setlength{\tabcolsep}{3pt}
  \renewcommand{\arraystretch}{1.05}
  \begin{tabular*}{\linewidth}{@{\extracolsep{\fill}}lll@{}}
    \toprule
     & Four-node path & Three-node path \\
    \midrule
    Model & $M_1=\alpha A$ & $M=\alpha A$ \\
     & $M_2=\gamma_A A+\gamma_1 M_1$ & \\
     & $Y=\beta_A A+\beta_1 M_1+\beta_2 M_2$ & $Y=\beta_A A+\beta_M M$ \\
     & $Y=\tau A$ & $Y=\tau A$ \\
    \midrule
    T & $\tau$ & $\tau$ \\
    D & $\beta_A$ & $\beta_A$ \\
    S & E1$\cdot$E2$\cdot$E3 & E1$\cdot$E2 \\
    B1 & E1$\cdot\beta_1$ & \\
    B2 & $\gamma_A\cdot$E3 & \\
    E1 & $\alpha$ & $\alpha$ \\
    E2 & $\gamma_1$ & $\beta_M$ \\
    E3 & $\beta_2$ & \\
    \bottomrule
  \end{tabular*}
\end{wraptable}
\paragraph{Primary estimates (c, d).}
For a path $A\to M_1\to M_2\to Y$, each primary analysis fits the linear models
of Table~\ref{tab:findings:panel-symbols} with the adjustment set of
Table~\ref{tab:findings:pathway-codes}: T is the total association, D the
direct association, S the serial association through all intermediate nodes,
B1 and B2 the branches via $M_1$ and $M_2$, and E1--E3 the component
coefficients along the path. Estimates are in outcome SDs per exposure SD
(SD/SD), with variables standardized within each analysis, or in the units of
Table~\ref{tab:data:abbreviations}; the SD/SD estimates of H4 in panel c are
its measurement-unit estimates multiplied by SD($A$)/SD($Y$). In O3, BMI enters
as its deviation from the mean in units of 5\,kg/m$^2$, and the outcome model
includes a BMI$\times$SpO$_2$ interaction; E1 and S of O3 are per
5\,kg/m$^2$ of BMI, and S is the product of E1 and the SpO$_2$ coefficient at
mean BMI. Panel c plots T, D, S, B1 and B2 of each primary analysis with 95\%
CIs on one axis per pathway, with --- for parameters without a reported
interval (Table~\ref{tab:findings:fig4-c}). Panel d plots the component
coefficients E1--E3, or E1 and E2 for three-node paths, of the primary analyses
other than H2 and L2: standardized coefficients on a common axis and
coefficients in measurement units (H4, H5, L7, O2 and O3) on separate axes,
with E1 only for O3 (Table~\ref{tab:findings:fig4-d}).

\paragraph{Reverse, within-person and off-path analyses (e--g).}
Panel e plots S of seven reverse analyses on separate axes. H1r, R1r and the
reverse L7 and T1 models, PWV--TG--ALT and PWV--DBP--ALT, take PWV as the
exposure in strictly ordered designs with PWV measured first; L6, O1 and
O3$^{+}$ keep the forward models and reverse the measurement order along part
of the path, with PWV before SBP before same-day GGT and TG in L6, SBP before
SpO$_2$ in O1 and cIMT before SBP in O3$^{+}$. H1r is in WHR SDs per
interquartile PWV contrast, the reverse L7 and T1 models are in measurement
units and the others are in SD/SD. Panel f plots participant fixed-effects
versions of the strictly ordered L7 and T1 analyses in measurement units,
using within-participant variation only. Panel g plots, in measurement units,
associations with outcomes off the declared path: waist circumference with
blood sodium in place of ALT (H3), AF with the left-minus-right PWV difference
(L1), and S and T of the L2 path with the absolute left--right PWV difference
as outcome. Table~\ref{tab:findings:fig4-efg} lists the values of panels e--g.

\begin{table}[tbp]
  \centering
  \caption{\textbf{Values in Figure~\ref{fig:findings:experimental-atlas}c.} Axis readings with 95\% CIs below, in SD/SD
  unless a scale is given under the code; a reading times the axis factor beside the code gives
  the estimate; ---, no reported interval.}
  \label{tab:findings:fig4-c}
  \footnotesize
  \setlength{\tabcolsep}{1.5pt}
  \renewcommand{\arraystretch}{0.9}
\begin{tabular*}{\linewidth}{@{\extracolsep{\fill}}lccccc@{}}
    \toprule
    Code & T & D & S & B1 & B2 \\
    \midrule
    H1 & $0.1601$ & $0.05006$ & $0.01099$ & $0.001912$ & $0.09709$ \\
     & {\scriptsize $[0.1195,0.2006]$} & {\scriptsize $[0.01241,0.08772]$} & {\scriptsize $[0.006451,0.01552]$} & {\scriptsize $[-0.005126,0.008951]$} & {\scriptsize $[0.07721,0.117]$} \\[1pt]
    H2 & $0.107$ & $-0.01578$ & $0.006596$ & $0.002977$ & $0.1132$ \\
     & {\scriptsize $[0.05934,0.1537]$} & {\scriptsize $[-0.06288,0.03087]$} & {\scriptsize $[0.002991,0.01145]$} & {\scriptsize $[-0.004173,0.01004]$} & {\scriptsize $[0.09033,0.138]$} \\[1pt]
    H3 & --- & $0.07912$ & $0.003139$ & $0.001833$ & $0.03933$ \\
     &  & {\scriptsize $[0.04215,0.1165]$} & {\scriptsize $[0.001706,0.004976]$} & {\scriptsize $[-0.006972,0.01203]$} & {\scriptsize $[0.02907,0.05033]$} \\[1pt]
    H4 & $0.04668$ & $0.003047$ & $0.003715$ & $0.005376$ & $0.03455$ \\
\bottomrule
  \end{tabular*}
\end{table}

\begin{table}[tbp]
  \centering
  \caption{\textbf{Values in Figure~\ref{fig:findings:experimental-atlas}d.} Axis readings of the component
  coefficients with 95\% CIs: standardized coefficients in SD/SD, and coefficients in
  measurement units, each in units of the modeled node per unit of its predictor along the
  path of Table~\ref{tab:findings:pathway-codes} and per 5\,kg/m$^2$ of BMI for O3.
  $^{\mathrm{a}}$Axis factor $\times10^{-3}$.}
  \label{tab:findings:fig4-d}
  \footnotesize
  \setlength{\tabcolsep}{2pt}
  \renewcommand{\arraystretch}{0.95}
  \begin{tabular*}{\linewidth}{@{\extracolsep{\fill}}lrcrcrc@{}}
    \toprule
    & \multicolumn{2}{c}{E1} & \multicolumn{2}{c}{E2} & \multicolumn{2}{c}{E3} \\
    \cmidrule(lr){2-3}\cmidrule(lr){4-5}\cmidrule(l){6-7}
    Code & Value & 95\% CI & Value & 95\% CI & Value & 95\% CI \\
    \midrule
    \multicolumn{7}{@{}l}{\emph{Standardized coefficients (SD/SD)}} \\
    H1 & $0.2323$ & $[0.1915,\,0.2732]$ & $0.1088$ & $[0.06888,\,0.1487]$ & $0.4347$ & $[0.4011,\,0.4683]$ \\
    H3 & $0.2671$ & $[0.2329,\,0.3046]$ & $0.07835$ & $[0.04865,\,0.1115]$ & $0.15$ & $[0.1136,\,0.1868]$ \\
    L1 & $0.363$ & $[0.3198,\,0.4057]$ & $0.1086$ & $[0.06154,\,0.1551]$ & $0.4198$ & $[0.379,\,0.4598]$ \\
    L3 & $0.1639$ & $[0.09842,\,0.231]$ & $0.0873$ & $[0.03825,\,0.1351]$ & $0.4378$ & $[0.3792,\,0.4953]$ \\
    L4 & $0.09701$ & $[0.03659,\,0.1574]$ & $0.1002$ & $[0.04051,\,0.1599]$ & $0.4464$ & $[0.3762,\,0.5166]$ \\
    L5 & $0.3283$ & $[0.258,\,0.3985]$ & $0.1289$ & $[0.07333,\,0.1845]$ & $0.1819$ & $[0.125,\,0.2388]$ \\
    L6 & $0.2626$ & $[0.09452,\,0.4307]$ & $0.2872$ & $[0.1575,\,0.4169]$ & $0.3877$ & $[0.2672,\,0.5081]$ \\
    O1 & $-0.3492$ & $[-0.3768,\,-0.3216]$ & $0.04372$ & $[0.01863,\,0.06881]$ & $0.1653$ & $[0.1366,\,0.194]$ \\
    \midrule
    \multicolumn{7}{@{}l}{\emph{Measurement units}} \\
    H4 & $0.2422$ & $[0.1818,\,0.3026]$ & $15.55$ & $[9.269,\,21.82]$ & $1.005^{\mathrm{a}}$ & $[0.761,\,1.25]$ \\
    H5 & $7.484^{\mathrm{a}}$ & $[5.918,\,9.051]$ & $14.11$ & $[7.88,\,20.34]$ & $0.04444$ & $[0.04053,\,0.04834]$ \\
    L7 & $0.2205$ & $[0.03882,\,0.7663]$ & $3.957^{\mathrm{a}}$ & $[2.663,\,5.308]$ &  &  \\
    O2 & $-0.01867$ & $[-0.02692,\,-0.01042]$ & $-1.306$ & $[-2.231,\,-0.3806]$ &  &  \\
    O3 & $-0.5281$ & $[-0.6866,\,-0.3696]$ &  &  &  &  \\
    \bottomrule
  \end{tabular*}
\end{table}

\begin{table}[tbp]
  \centering
  \caption{\textbf{Values in Figure~\ref{fig:findings:experimental-atlas}e--g.} Axis readings with 95\%
  CIs; a reading times the axis factor of its scale gives the estimate. H1r is in WHR SDs
  per interquartile PWV contrast, and measurement-unit scales are the modeled node per unit
  of its predictor, in the units of Table~\ref{tab:data:abbreviations}. In g, Sodium is
  the coefficient of Waist in the model for blood sodium, PWV L--R that of AF in the model
  for the left-minus-right PWV difference, and S and T refer to the L2 path with the
  absolute left--right PWV difference as outcome.}
  \label{tab:findings:fig4-efg}
  \footnotesize
  \setlength{\tabcolsep}{3pt}
  \renewcommand{\arraystretch}{0.95}
\begin{tabular*}{\linewidth}{@{\extracolsep{\fill}}llllrc@{}}
    \toprule
    Code & Path & Parameter & Scale & Value & 95\% CI \\
    \midrule
    \multicolumn{6}{@{}l}{\emph{e, reverse analyses}} \\
    H1r & PWV--ALT--WHR & S & WHR SD/IQR $\times10^{-2}$ & $0.5207$ & $[-0.6551,\,1.947]$ \\
    \addlinespace[1.5pt]
    L6 & GGT--TG--SBP--PWV & S & SD/SD & $0.009925$ & $[-0.05225,\,0.09413]$ \\
    \addlinespace[1.5pt]
    L7 & PWV--TG--ALT & S & ALT/PWV & $0.02135$ & $[-0.008912,\,0.5664]$ \\
    \addlinespace[1.5pt]
    T1 & PWV--DBP--ALT & S & ALT/PWV & $0.2099$ & $[-0.6153,\,1.018]$ \\
    \addlinespace[1.5pt]
    O1 & Waist--SpO$_2$--SBP--cIMT & S & SD/SD $\times10^{-3}$ & $-2.781$ & $[-4.526,\,-1.337]$ \\
    \addlinespace[1.5pt]
    O3$^{+}$ & BMI--SpO$_2$--SBP--cIMT & S & SD/SD & $-0.01053$ & $[-0.03331,\,0.01226]$ \\
\bottomrule
  \end{tabular*}
\end{table}

\begin{table}[tbp]
  \begin{minipage}[t]{0.505\linewidth}
    \centering
    \caption{\textbf{Values in Figure~\ref{fig:findings:experimental-atlas}h.} Direct and indirect (serial)
    associations of L6$^{\dagger}$ in PWV SDs per SD of GGT with 95\% CIs, the shaded
    bands, at each assumed residual correlation $\rho$.}
    \label{tab:findings:fig4-h}
    \scriptsize
    \setlength{\tabcolsep}{2.7pt}
    \renewcommand{\arraystretch}{0.95}
\begin{tabular*}{\linewidth}{@{\extracolsep{\fill}}rrcrc@{}}
      \toprule
      & \multicolumn{2}{c}{Direct} & \multicolumn{2}{c}{Indirect} \\
      \cmidrule(lr){2-3}\cmidrule(l){4-5}
      $\rho$ & Estimate & 95\% CI & Estimate & 95\% CI \\
      \midrule
      $-0.8$ & $0.6919$ & $[-3.114,\,5.864]$ & $-0.7076$ & $[-6.136,\,4.41]$ \\
      $-0.7$ & $0.5867$ & $[-2.912,\,5.219]$ & $-0.6023$ & $[-5.744,\,3.899]$ \\
      $-0.6$ & $0.518$ & $[-2.818,\,4.985]$ & $-0.5337$ & $[-5.309,\,3.521]$ \\
      $-0.5$ & $0.4666$ & $[-2.623,\,4.753]$ & $-0.4822$ & $[-4.783,\,3.332]$ \\
      $-0.4$ & $0.4245$ & $[-2.625,\,4.412]$ & $-0.4402$ & $[-4.342,\,3.126]$ \\
      $-0.3$ & $0.3882$ & $[-2.472,\,4.237]$ & $-0.4038$ & $[-4.123,\,3.012]$ \\
      $-0.2$ & $0.3553$ & $[-2.396,\,4.208]$ & $-0.3709$ & $[-3.936,\,2.846]$ \\
      $-0.1$ & $0.3244$ & $[-2.444,\,4.03]$ & $-0.34$ & $[-3.799,\,2.748]$ \\
      $0.0$ & $0.2944$ & $[-2.455,\,3.912]$ & $-0.3101$ & $[-3.59,\,2.675]$ \\
      $0.1$ & $0.2645$ & $[-2.487,\,3.869]$ & $-0.2801$ & $[-3.311,\,2.631]$ \\
      $0.2$ & $0.2336$ & $[-2.6,\,3.706]$ & $-0.2492$ & $[-3.336,\,2.561]$ \\
      $0.3$ & $0.2007$ & $[-2.707,\,3.824]$ & $-0.2163$ & $[-3.168,\,2.526]$ \\
      $0.4$ & $0.1643$ & $[-2.84,\,3.866]$ & $-0.1799$ & $[-3.036,\,2.373]$ \\
      $0.5$ & $0.1223$ & $[-2.925,\,3.941]$ & $-0.1379$ & $[-3.068,\,2.475]$ \\
      $0.6$ & $0.07082$ & $[-3.175,\,3.999]$ & $-0.08645$ & $[-2.887,\,2.637]$ \\
      $0.7$ & $0.002185$ & $[-3.401,\,4.058]$ & $-0.01782$ & $[-2.674,\,2.929]$ \\
      $0.8$ & $-0.1031$ & $[-3.628,\,4.566]$ & $0.08747$ & $[-3.01,\,3.162]$ \\
      \bottomrule
    \end{tabular*}
  \end{minipage}\hfill
  \begin{minipage}[t]{0.465\linewidth}
    \centering
    \caption{\textbf{Values in Figure~\ref{fig:findings:experimental-atlas}i.} Lower bounds of the
    omitted-variable bias range by $R_Y^2$ (rows) and $R_X^2$ (columns), for L1 in SD/SD and
    for L2 in PWV/VAT with the axis factor $\times10^{-3}$; negative bounds are the open
    markers. Deterministic values without sampling intervals.}
    \label{tab:findings:fig4-i}
    \scriptsize
    \setlength{\tabcolsep}{2pt}
    \renewcommand{\arraystretch}{0.95}
    \begin{tabular*}{\linewidth}{@{\extracolsep{\fill}}llrrrrr@{}}
      \toprule
      & & \multicolumn{5}{c}{$R_X^2$} \\
      \cmidrule(l){3-7}
      Code & $R_Y^2$ & $0.01$ & $0.05$ & $0.10$ & $0.20$ & $0.30$ \\
      \midrule
      L1 & $0.01$ & $0.4349$ & $0.4241$ & $0.4154$ & $0.4014$ &  \\
       & $0.05$ & $0.4245$ & $0.4003$ & $0.3808$ & $0.3495$ &  \\
       & $0.10$ & $0.4167$ & $0.3824$ & $0.3548$ & $0.3106$ &  \\
       & $0.20$ & $0.4056$ & $0.3572$ & $0.3181$ & $0.2555$ &  \\
      \addlinespace[2pt]
      L2 & $0.01$ & $0.585$ & $0.5459$ & $0.5145$ & $0.464$ & $0.4172$ \\
       & $0.05$ & $0.5473$ & $0.4601$ & $0.3898$ & $0.277$ & $0.1723$ \\
       & $0.10$ & $0.5192$ & $0.3958$ & $0.2963$ & $0.1368$ & $-0.01123$ \\
       & $0.20$ & $0.4793$ & $0.3048$ & $0.1642$ & $-0.06144$ & $-0.2708$ \\
       & $0.30$ & $0.4488$ & $0.235$ & $0.06276$ & $-0.2136$ & $-0.4699$ \\
      \bottomrule
    \end{tabular*}
  \end{minipage}
\end{table}

\needspace{15\baselineskip}
\begin{wraptable}{r}{0.43\textwidth}
  \centering
  \caption{\textbf{Values in Figure~\ref{fig:findings:experimental-atlas}j.} Robustness values, top to bottom
  as plotted. P, primary specification; W, wider windows; EA, additional SBP adjustment;
  AHI, additional AHI adjustment. Deterministic values.}
  \label{tab:findings:fig4-j}
  \footnotesize
  \setlength{\tabcolsep}{2pt}
  \renewcommand{\arraystretch}{0.95}
  \begin{tabular*}{\linewidth}{@{\extracolsep{\fill}}llllr@{}}
    \toprule
    Code & Setting & Target & Association & RV \\
    \midrule
    O1 &  & E1 & Waist--SpO$_2$ & $0.302$ \\
     &  & E2 & SpO$_2$--SBP & $0.06707$ \\
     &  & E3 & SBP--cIMT & $0.1765$ \\
     &  & AY & Waist--cIMT & $0.102$ \\
    \addlinespace[1.5pt]
    O2 & P & E1 & SpO$_2$--GlycA & $0.1488$ \\
     & P & E2 & GlycA--PWV & $0.09105$ \\
     & W & E2 & GlycA--PWV & $0.08925$ \\
     & W & E1 & SpO$_2$--GlycA & $0.1516$ \\
     & EA & E1 & SpO$_2$--GlycA & $0.1513$ \\
     & EA & E2 & GlycA--PWV & $0.1178$ \\
     & AHI & E2 & GlycA--PWV & $0.0874$ \\
     & AHI & E1 & SpO$_2$--GlycA & $0.1496$ \\
    \addlinespace[1.5pt]
    L7 &  & AY & ALT--PWV & $0.0747$ \\
    \bottomrule
  \end{tabular*}
\end{wraptable}
\paragraph{Confounding sensitivity (h--j).}
Panel h plots the direct and indirect (serial) associations of
L6$^{\dagger}$, in PWV SDs per SD of GGT, against an assumed correlation
$\rho$ from $-0.8$ to 0.8 between the residuals of its TG and PWV models, with
95\% intervals as shaded bands (Table~\ref{tab:findings:fig4-h}). Panel i plots the lower bound of the
omitted-variable bias range \citep{cinelli2020sensitivity} for the
standardized pairwise DBP--PWV coefficient in the L1 research question and for
T of L2 in measurement units, against the partial $R^2$ of a hypothetical
confounder with the predictor, DBP or VAT, $R_X^2$, with one line per partial
$R^2$ with PWV, $R_Y^2$; both range from 0.01 to 0.20 for L1 and from 0.01 to
0.30 for L2, and open markers are lower bounds below zero (Table~\ref{tab:findings:fig4-i}). Panel j plots
robustness values (RV), the partial $R^2$ that a confounder would need with
both the predictor and the outcome of an association to reduce it to zero, for
the pairwise Waist--SpO$_2$, SpO$_2$--SBP, SBP--cIMT and Waist--cIMT analyses
of O1 (E1, E2, E3 and AY), for E1 and E2 of O2 under the primary specification
(P), wider windows (W, distinct from design class W), additional SBP
adjustment (EA) and additional AHI adjustment (AHI), and for the ALT--PWV
association in the L7 research question (AY)
(Table~\ref{tab:findings:fig4-j}). Panels i and j show
deterministic values without sampling intervals.

\FloatBarrier

\subsection{Definitions for \texorpdfstring{Figure~\ref{fig:findings:conditional-validation}}{Figure 5}}
\label{app:findings:conditional-validation}

\paragraph{Codes and plotted values.}
Figure~\ref{fig:findings:conditional-validation} uses the codes of
Tables~\ref{tab:findings:pathway-codes} and~\ref{tab:findings:other-codes};
Table~\ref{tab:findings:fig5-codes} lists the ordered path, research question,
design class and adjustment set of the models in the figure other than the
primary analyses of Table~\ref{tab:findings:pathway-codes}.
Tables~\ref{tab:findings:fig5-a} to~\ref{tab:findings:fig5-i} list all 128
plotted values as axis readings with their 95\% CIs and at most four
significant digits; a reading multiplied by the axis factor $\times10^{k}$
gives the estimate, and in panel a the reading is S divided by the absolute
value of the primary S (Table~\ref{tab:findings:fig5-a}). Diamonds mark four
formal contrasts: the R1 adjustment contrast (c), the L2 sex-category contrast
(d), and the H1$^{\dagger}$ and L3 age contrasts (e).

\begin{table}[tbp]
  \centering
  \caption{\textbf{Other models in Figure~\ref{fig:findings:conditional-validation}:} ordered path, research question
  (RQ), design class and adjustment set of the models other than the primary analyses of
  Table~\ref{tab:findings:pathway-codes}. Design class as in Figure~\ref{fig:findings:experimental-atlas}a: C,
  cross-sectional; W, weakly ordered; L, strictly ordered. Age enters linearly unless age$^2$ is listed; cohort
  and study are HPP labels.}
  \label{tab:findings:fig5-codes}
  \footnotesize
  \setlength{\tabcolsep}{3pt}
  \renewcommand{\arraystretch}{0.95}
  \begin{tabular}{@{}l>{\raggedright\arraybackslash}p{4.9cm}cc>{\raggedright\arraybackslash}p{4.3cm}c@{}}
    \toprule
    Code & Ordered path or model & RQ & Design & Adjustment set & Panels \\
    \midrule
    R1 & WHR--GlycA--SBP--PWV & R1 & W & Age, age$^2$, sex, cohort & c, i \\
    R1 joint & PWV on WHR, GlycA and SBP & R1 & C & Age, sex, study & h \\
    T1 & ALT--DBP--PWV & L7 & L & Age, sex & b \\
    T2 & TG--DBP--PWV & L7 & L & Age, sex & b \\
    X1 & ALT--TG--DBP--PWV & L7 & W & Age, sex, BMI & b \\
    H1$^{\dagger}$ & WHR--ALT--PWV & H1 & L & Age, age$^2$, sex, cohort & e \\
    O3$^{+}$ & BMI--SpO$_2$--SBP--cIMT & O3 & W & Age, age$^2$, sex, study, current and lifetime smoking & d \\
    L5$^{*}$ & WHR--cIMT with WHR$\times$TG or WHR$\times$SBP & L5 & C & Age, sex & f \\
    O1$^{*}$ & Waist--SpO$_2$--SBP--cIMT with Waist$\times$SpO$_2$ and Waist$\times$SBP & O1 & W & Age, sex, cohort & f \\
    I1 & ALT--TG--PWV with ALT$\times$TG & L7 & L & Age, sex & f \\
    I2 & ALT--DBP--PWV with ALT$\times$DBP & L7 & L & Age, sex & f \\
    \bottomrule
  \end{tabular}
\end{table}

\paragraph{Specification, timing and adjustment (a--c).}
Panel a plots the serial association S under 21 alternative specifications,
with the estimate and both interval endpoints divided by the absolute value of
the primary S of the same pathway in
Figure~\ref{fig:findings:experimental-atlas}c, so that the dashed line at $+1$
($-1$ for O1) marks the primary estimate. Each specification changes the
measurement windows, the adjustment set or the temporal structure and
otherwise follows the primary analysis (Table~\ref{tab:findings:fig5-a}): NW
and W narrow and widen, respectively, the windows between adjacent
measurements; EA adds covariates to the primary adjustment set, namely
smoking, alcohol use and moderate physical activity for H1; height, smoking
and alcohol use for H3; smoking for L5 and O1; and SBP for O2; H2, H4 and L1
add smoking and alcohol use, BMI and height, respectively, and a further O2
specification adds AHI. Smoking, alcohol use and physical activity are the
latest records on or before the day of the exposure measurement. The three L6
specifications change the temporal structure: CC measures PWV
contemporaneously with SBP, W widens the time ranges between SBP and the
same-day GGT and TG measurements and between SBP and PWV, and F requires SBP
to follow the day on which GGT and TG were measured. L3, L4, L7 and O3 are
absent from panel a: L3 and O3 have no alternative specification with an
interval, the alternative specification of L4 gives the same estimate as its
primary analysis, and all L7 specifications appear in panel b. Panel b plots S
under the timing and covariate specifications of the L7 research question in
measurement units (PWV/TG for T2 and PWV/ALT otherwise;
Table~\ref{tab:findings:fig5-b}). L7, T1 and T2 are strictly ordered analyses;
in L7 and T1, P is the primary specification, NW narrows the allowed intervals
between adjacent measurements, EA adds BMI, smoking and alcohol use, and PY
adds the latest PWV measured before ALT to the PWV model, restricting the
analysis to ALT measurements preceded by a PWV measurement; in X1, ALT and TG
come from the same blood test. Panel c compares the primary (P) and
extended-adjustment (EA) models of R1 (Table~\ref{tab:findings:fig5-c}). EA
adds BMI on the WHR measurement day and the latest smoking status on or before
that day to age, age$^2$, sex and cohort, and is fitted and restandardized on
the sample with both records. The upper axis shows the component coefficients
E1--E3; the lower axis shows S, where the common base CB refits the primary
adjustment set on the EA sample, and $\Delta=\mathrm{EA}-\mathrm{CB}$ is the
change in S from adding BMI and smoking, with the same sample and
standardization.

\begin{table}[tbp]
  \centering
  \caption{\textbf{Values in Figure~\ref{fig:findings:conditional-validation}a.} S of each alternative specification
  with its 95\% CI, in the scale and axis factor of Table~\ref{tab:findings:fig4-c}, and the plotted reading, S
  divided by the absolute value of the primary S in Figure~\ref{fig:findings:experimental-atlas}c; rows within each
  code follow the marks top to bottom. NW, narrow window; W, wide window or, for L6, wider SBP ranges
  (distinct from design class W); EA, extended adjustment; CC, PWV contemporaneous with SBP; F, SBP after
  the GGT/TG day.}
  \label{tab:findings:fig5-a}
  \footnotesize
  \setlength{\tabcolsep}{1pt}
  \renewcommand{\arraystretch}{0.95}
\begin{tabular*}{\linewidth}{@{\extracolsep{\fill}}lllrcrc@{}}
    \toprule
    & & & \multicolumn{2}{c}{S} & \multicolumn{2}{c}{Reading} \\
    \cmidrule(lr){4-5}\cmidrule(l){6-7}
    Code & Specification & Scale & Value & 95\% CI & Value & 95\% CI \\
    \midrule
    H1 & Narrow window (NW) & SD/SD & $0.008172$ & $[0.003172,0.01317]$ & $0.744$ & $[0.289,1.198]$ \\
     & + smoking, alcohol, activity (EA) &  & $0.01052$ & $[0.005153,0.0159]$ & $0.957$ & $[0.469,1.447]$ \\
    \addlinespace[1.5pt]
    H2 & + smoking, alcohol & SD/SD & $0.006543$ & $[0.001673,0.01296]$ & $0.992$ & $[0.254,1.965]$ \\
     & Wide window (W) &  & $0.007283$ & $[0.004549,0.01061]$ & $1.104$ & $[0.690,1.609]$ \\
    \addlinespace[1.5pt]
    H3 & Wide window (W) & SD/SD & $0.00214$ & $[0.00103,0.003406]$ & $0.682$ & $[0.328,1.085]$ \\
     & + height, smoking, alcohol (EA) &  & $0.002671$ & $[0.0009525,0.004954]$ & $0.851$ & $[0.303,1.578]$ \\
\bottomrule
  \end{tabular*}
\end{table}

\begin{table}[tbp]
  \begin{minipage}[t]{0.49\linewidth}
    \centering
    \caption{\textbf{Values in Figure~\ref{fig:findings:conditional-validation}b,} with the axis factor $\times10^{-3}$.
    P, primary; NW, narrower intervals; EA, added BMI, smoking and alcohol use; PY, added prior PWV.}
    \label{tab:findings:fig5-b}
    \footnotesize
    \setlength{\tabcolsep}{1.4pt}
    \renewcommand{\arraystretch}{0.95}
    \begin{tabular*}{\linewidth}{@{\extracolsep{\fill}}lllrc@{}}
      \toprule
      Code & Specification & Scale & Value & 95\% CI \\
      \midrule
      L7 & P & PWV/ALT & $0.8726$ & $[0.1607,2.892]$ \\
       & NW &  & $0.304$ & $[-0.247,3.227]$ \\
       & EA &  & $2.629$ & $[0.2741,7.509]$ \\
       & PY &  & $3.664$ & $[0.3396,11.25]$ \\
      \addlinespace[1.5pt]
      T1 & P & PWV/ALT & $5.739$ & $[3.361,8.694]$ \\
       & NW &  & $9.123$ & $[3.557,16.43]$ \\
      \addlinespace[1.5pt]
      X1 &  & PWV/ALT & $0.2614$ & $[-0.3314,0.9597]$ \\
      \addlinespace[1.5pt]
      T2 &  & PWV/TG & $1.942$ & $[1.466,2.483]$ \\
      \bottomrule
    \end{tabular*}
  \end{minipage}\hfill
  \begin{minipage}[t]{0.49\linewidth}
    \centering
    \caption{\textbf{Values in Figure~\ref{fig:findings:conditional-validation}c} for R1 in SD/SD. P, primary model;
    EA, extended adjustment; CB, primary adjustment set on the EA sample.}
    \label{tab:findings:fig5-c}
    \footnotesize
    \setlength{\tabcolsep}{2pt}
    \renewcommand{\arraystretch}{0.95}
    \begin{tabular*}{\linewidth}{@{\extracolsep{\fill}}llrc@{}}
      \toprule
      Parameter & Model & Value & 95\% CI \\
      \midrule
      E1 & P & $0.3324$ & $[0.2588,0.4059]$ \\
       & EA & $0.1725$ & $[0.09574,0.2492]$ \\
      E2 & P & $0.09871$ & $[0.04792,0.1495]$ \\
       & EA & $0.02471$ & $[-0.03203,0.08144]$ \\
      E3 & P & $0.478$ & $[0.4259,0.5302]$ \\
       & EA & $0.4745$ & $[0.4136,0.5353]$ \\
      \addlinespace[1.5pt]
      S & P & $0.01569$ & $[0.006851,0.02452]$ \\
       & CB & $0.01218$ & $[0.003217,0.02114]$ \\
       & EA & $0.002022$ & $[-0.002765,0.006808]$ \\
      $\Delta$ & EA $-$ CB & $-0.01016$ & $[-0.01537,-0.004936]$ \\
      \bottomrule
    \end{tabular*}
  \end{minipage}
\end{table}

\begin{table}[tbp]
  \begin{minipage}[t]{0.485\linewidth}
    \centering
    \caption{\textbf{Values in Figure~\ref{fig:findings:conditional-validation}d.} S in the two sex categories, First
    and Second, of each analysis, in SD/SD except L2 in PWV/VAT; a reading times the axis factor beside
    the code gives the estimate; $\Delta$S, First minus Second.}
    \label{tab:findings:fig5-d}
    \footnotesize
    \setlength{\tabcolsep}{2pt}
    \renewcommand{\arraystretch}{0.95}
\begin{tabular*}{\linewidth}{@{\extracolsep{\fill}}llrc@{}}
      \toprule
      Code & Category & Value & 95\% CI \\
      \midrule
      H1 & First & $0.007646$ & $[0.002633,0.01266]$ \\
       & Second & $0.01319$ & $[0.005866,0.02052]$ \\
      \addlinespace[1.5pt]
      H2 & First & $0.003355$ & $[0.0001803,0.008561]$ \\
       & Second & $0.01028$ & $[0.003169,0.02036]$ \\
      \addlinespace[1.5pt]
      H3\,{\scriptsize$\times10^{-3}$} & First & $2.818$ & $[1.14,5.049]$ \\
       & Second & $4.154$ & $[1.69,7.223]$ \\
      \addlinespace[1.5pt]
      L2\,{\scriptsize$\times10^{-4}$} & First & $0.786$ & $[0.2593,1.55]$ \\
       & Second & $0.01293$ & $[-0.1624,0.2022]$ \\
       & $\Delta$S & $0.7731$ & $[0.2087,1.533]$ \\
      \addlinespace[1.5pt]
      L4\,{\scriptsize$\times10^{-2}$} & First & $0.5041$ & $[-0.06028,1.069]$ \\
       & Second & $0.1737$ & $[-0.3496,0.6971]$ \\
      \addlinespace[1.5pt]
      O3$^{+}$\,{\scriptsize$\times10^{-2}$} & First & $-0.1787$ & $[-0.9097,0.5523]$ \\
       & Second & $0.09713$ & $[-0.8757,1.07]$ \\
      \bottomrule
    \end{tabular*}
  \end{minipage}\hfill
  \begin{minipage}[t]{0.485\linewidth}
    \centering
    \caption{\textbf{Values in Figure~\ref{fig:findings:conditional-validation}e.} Shift of H1$^{\dagger}$ in
    participant-weighted SDs of PWV at age percentiles, with $\Delta$ = P75 $-$ P25; H3 and O2 at
    standardized age $z$, O2 in PWV/SpO$_2$; L3 at $\pm10$ years from the participant-weighted mean age,
    with $\Delta = S_{+10}-S_{-10}$; SD/SD unless stated.}
    \label{tab:findings:fig5-e}
    \footnotesize
    \setlength{\tabcolsep}{2pt}
    \renewcommand{\arraystretch}{0.95}
    \begin{tabular*}{\linewidth}{@{\extracolsep{\fill}}llrc@{}}
      \toprule
      Code & Age point & Value & 95\% CI \\
      \midrule
      H1$^{\dagger}$ & P25 & $-0.004822$ & $[-0.06078,0.04757]$ \\
       & P50 & $-0.01025$ & $[-0.05129,0.02584]$ \\
       & P75 & $-0.01189$ & $[-0.0482,0.02309]$ \\
       & $\Delta$ & $-0.007071$ & $[-0.06151,0.05428]$ \\
      \addlinespace[1.5pt]
      H3 & $z=-1$ & $0.003317$ & $[0.001365,0.006012]$ \\
       & $z=0$ & $0.003187$ & $[0.00164,0.005166]$ \\
       & $z=+1$ & $0.002531$ & $[0.0007849,0.005034]$ \\
      \addlinespace[1.5pt]
      L3 & $-10$ y & $0.02229$ & $[0.007786,0.04081]$ \\
       & 0 y & $0.005946$ & $[0.002232,0.01105]$ \\
       & $+10$ y & $-0.0004561$ & $[-0.003684,0.001825]$ \\
       & $\Delta$ & $-0.02275$ & $[-0.04126,-0.00828]$ \\
      \addlinespace[1.5pt]
      O2 & $z=-1$ & $0.01721$ & $[-0.01834,0.05276]$ \\
       & $z=0$ & $0.0253$ & $[0.004964,0.04564]$ \\
       & $z=+1$ & $0.02052$ & $[-0.004507,0.04555]$ \\
      \bottomrule
    \end{tabular*}
  \end{minipage}
\end{table}

\paragraph{Sex, age and interactions (d--f).}
Panel d plots S in the two sex categories, First and Second, of six analyses
(Table~\ref{tab:findings:fig5-d}). H1, L2 and O3$^{+}$ are fitted separately
within each category; H2 and H3 add interactions of sex with every upstream
variable in pooled models, and L4 adds only interactions of sex with the path
variable immediately upstream in each equation. S is standardized by the SDs
of the pooled sample of both categories, except for L2, which is in
measurement units (PWV/VAT). L2 also gives the category contrast
$\Delta S=S_{\mathrm{First}}-S_{\mathrm{Second}}$. Panel e plots age-conditional
associations, each age point being an evaluation position in the model
(Table~\ref{tab:findings:fig5-e}). The H1$^{\dagger}$ model includes a
WHR$\times$ALT interaction and interactions of the path variables with age;
Shift is the difference in predicted PWV, with WHR fixed at its 75th
percentile, when ALT moves from its predicted mean at the 25th percentile of
WHR to that at the 75th percentile of WHR, in participant-weighted SDs of PWV,
evaluated at the 25th, 50th and 75th percentiles of age (P25, P50, P75), and
$\Delta$ is the difference between P75 and P25. H3, L3 and O2 add interactions
of age with every upstream variable to the primary models, and S is the
product of the path slopes at a given age: at standardized ages $-1$, 0 and
$+1$ for H3 and O2, and at $-10$, 0 and $+10$ years from the
participant-weighted mean age for L3, with $\Delta$ the difference between
$+10$ and $-10$ years. O2 is in measurement units (PWV/SpO$_2$), and H3 and L3
are in SD/SD. Panel f plots interaction models
(Table~\ref{tab:findings:fig5-f}). L5$^{*}$ comprises two cIMT models, one with
a WHR$\times$TG and one with a WHR$\times$SBP interaction, plotted as
WHR--cIMT slopes at the sample mean of TG or SBP minus one SD, at the mean and
at the mean plus one SD. O1$^{*}$ adds a Waist$\times$SpO$_2$ interaction to
the SBP equation and a Waist$\times$SBP interaction to the cIMT equation of
O1, and $G(z_A)=\mathrm{E1}\,(\mathrm{E2}+\eta z_A)(\mathrm{E3}+\kappa z_A)$ is
the local serial gradient at standardized waist $z_A$, where E1--E3 are the
component coefficients at mean waist (Table~\ref{tab:findings:panel-symbols})
and $\eta$ and $\kappa$ are the two interaction coefficients, respectively. I1
and I2 are the ALT$\times$TG and ALT$\times$DBP interaction coefficients in the
PWV models of the L7 and T1 analyses, respectively.

\begin{table}[tbp]
  \centering
  \caption{\textbf{Values in Figure~\ref{fig:findings:conditional-validation}f.} L5$^{*}$, WHR--cIMT slopes in cIMT/WHR at the
  sample mean of TG or SBP and at the mean $\pm1$ SD; O1$^{*}$, $G(z_A)$ in SD/SD; I1 and I2, interaction
  coefficients in PWV/(ALT$\cdot$TG) $\times10^{-5}$ and PWV/(ALT$\cdot$DBP) $\times10^{-3}$, respectively.}
  \label{tab:findings:fig5-f}
  \footnotesize
  \setlength{\tabcolsep}{3pt}
  \renewcommand{\arraystretch}{0.95}
\begin{tabular*}{\linewidth}{@{\extracolsep{\fill}}llrcllrc@{}}
    \toprule
    Code & Condition & Value & 95\% CI & Code & Condition & Value & 95\% CI \\
    \midrule
    L5$^{*}$ & TG $-1$ SD & $0.06246$ & $[0.0102,0.1147]$ & O1$^{*}$ & $z_A=-1$ & $-0.002339$ & $[-0.004273,-0.0006947]$ \\
     & TG mean & $0.07583$ & $[0.03412,0.1175]$ &  & $z_A=0$ & $-0.002533$ & $[-0.004115,-0.00112]$ \\
     & TG $+1$ SD & $0.0892$ & $[0.02457,0.1538]$ &  & $z_A=+1$ & $-0.002682$ & $[-0.004836,-0.0007045]$ \\
     & SBP $-1$ SD & $0.004611$ & $[-0.02797,0.03719]$ & I1 & ALT$\times$TG & $-2.405$ & $[-7.113,4.604]$ \\
     & SBP mean & $0.03167$ & $[0.004323,0.05901]$ & I2 & ALT$\times$DBP & $-0.9153$ & $[-1.608,-0.3361]$ \\
     & SBP $+1$ SD & $0.05872$ & $[0.01956,0.09789]$ &  &  &  &  \\
    \bottomrule
  \end{tabular*}
\end{table}

\begin{table}[tbp]
  \centering
  \caption{\textbf{Values in Figure~\ref{fig:findings:conditional-validation}g.} Standardized component coefficients
  E1--E3 in SD/SD under the ten specifications of Table~\ref{tab:findings:fig5-a}.}
  \label{tab:findings:fig5-g}
  \footnotesize
  \setlength{\tabcolsep}{2pt}
  \renewcommand{\arraystretch}{0.95}
  \begin{tabular*}{\linewidth}{@{\extracolsep{\fill}}llrcrcrc@{}}
    \toprule
    & & \multicolumn{2}{c}{E1} & \multicolumn{2}{c}{E2} & \multicolumn{2}{c}{E3} \\
    \cmidrule(lr){3-4}\cmidrule(lr){5-6}\cmidrule(l){7-8}
    Code & Specification & Value & 95\% CI & Value & 95\% CI & Value & 95\% CI \\
    \midrule
    H1 & NW & $0.1815$ & $[0.1306,0.2323]$ & $0.1078$ & $[0.05108,0.1645]$ & $0.4178$ & $[0.3669,0.4686]$ \\
     & EA & $0.2204$ & $[0.1703,0.2706]$ & $0.1117$ & $[0.06421,0.1592]$ & $0.4273$ & $[0.3848,0.4699]$ \\
    \addlinespace[1.5pt]
    H3 & W & $0.2476$ & $[0.2194,0.2781]$ & $0.05358$ & $[0.02766,0.08089]$ & $0.1613$ & $[0.1307,0.1938]$ \\
     & EA & $0.2552$ & $[0.2146,0.2996]$ & $0.06826$ & $[0.02722,0.1132]$ & $0.1533$ & $[0.103,0.2032]$ \\
    \addlinespace[1.5pt]
    L6 & F & $0.2119$ & $[0.07973,0.344]$ & $0.1429$ & $[0.0675,0.2183]$ & $0.354$ & $[0.278,0.43]$ \\
     & W & $0.2199$ & $[0.06138,0.3784]$ & $0.1749$ & $[0.08348,0.2663]$ & $0.3714$ & $[0.2882,0.4545]$ \\
     & CC & $0.2001$ & $[0.08283,0.3173]$ & $0.1682$ & $[0.09227,0.2442]$ & $0.4716$ & $[0.3968,0.5464]$ \\
    \addlinespace[1.5pt]
    L5 & EA & $0.3102$ & $[0.2355,0.3848]$ & $0.1331$ & $[0.07016,0.1961]$ & $0.1882$ & $[0.1274,0.249]$ \\
    \addlinespace[1.5pt]
    O1 & NW & $-0.3577$ & $[-0.3902,-0.3252]$ & $0.0441$ & $[0.01248,0.07571]$ & $0.162$ & $[0.1268,0.1972]$ \\
     & EA & $-0.3564$ & $[-0.388,-0.3249]$ & $0.03479$ & $[0.004801,0.06478]$ & $0.1606$ & $[0.1275,0.1938]$ \\
    \bottomrule
  \end{tabular*}
\end{table}

\begin{table}[tbp]
  \begin{minipage}[t]{0.485\linewidth}
    \centering
    \caption{\textbf{Values in Figure~\ref{fig:findings:conditional-validation}h.} Coefficients of WHR, GlycA and SBP
    in the R1 joint model, in PWV per unit of each predictor. P, primary window; N and W, narrower
    and wider windows; A, primary window with BMI.}
    \label{tab:findings:fig5-h}
    \footnotesize
    \setlength{\tabcolsep}{2pt}
    \renewcommand{\arraystretch}{0.95}
    \begin{tabular*}{\linewidth}{@{\extracolsep{\fill}}llrc@{}}
      \toprule
      Predictor & Window & Value & 95\% CI \\
      \midrule
      WHR & P & $2.051$ & $[0.8835,3.219]$ \\
       & N & $2.034$ & $[0.8615,3.207]$ \\
       & A & $1.59$ & $[0.3499,2.83]$ \\
       & W & $2.086$ & $[0.9171,3.254]$ \\
      \addlinespace[1.5pt]
      GlycA & P & $-1.215$ & $[-1.826,-0.6038]$ \\
       & N & $-1.218$ & $[-1.829,-0.6063]$ \\
       & A & $-1.433$ & $[-2.073,-0.7941]$ \\
       & W & $-1.206$ & $[-1.817,-0.5947]$ \\
      \addlinespace[1.5pt]
      SBP & P & $0.05166$ & $[0.04603,0.05728]$ \\
       & N & $0.05167$ & $[0.04604,0.05729]$ \\
       & A & $0.0502$ & $[0.04438,0.05603]$ \\
       & W & $0.05151$ & $[0.04586,0.05716]$ \\
      \bottomrule
    \end{tabular*}
  \end{minipage}\hfill
  \begin{minipage}[t]{0.485\linewidth}
    \centering
    \caption{\textbf{Values in Figure~\ref{fig:findings:conditional-validation}i.} E1--E3 and T, changes in
    standardized coefficients per 10 years; $P(0)$, R1 serial association at the mean age in the
    age-interaction model, in SD/SD; $P'(0)$, its change per 10 years.}
    \label{tab:findings:fig5-i}
    \footnotesize
    \setlength{\tabcolsep}{2pt}
    \renewcommand{\arraystretch}{0.95}
    \begin{tabular*}{\linewidth}{@{\extracolsep{\fill}}llrc@{}}
      \toprule
      Code & Parameter & Value & 95\% CI \\
      \midrule
      L3 & E1 & $-0.1762$ & $[-0.2444,-0.1039]$ \\
       & E2 & $-0.04082$ & $[-0.1087,0.025]$ \\
       & E3 & $-0.08134$ & $[-0.1495,-0.01241]$ \\
      \addlinespace[1.5pt]
      R1 & E1 & $-0.1018$ & $[-0.1659,-0.03776]$ \\
       & E2 & $-0.001401$ & $[-0.07545,0.07265]$ \\
       & E3 & $-0.05297$ & $[-0.1103,0.004404]$ \\
       & T & $-0.05388$ & $[-0.1155,0.007767]$ \\
       & $P(0)$ & $0.01528$ & $[0.006219,0.02434]$ \\
       & $P'(0)$ & $-0.006523$ & $[-0.01811,0.005064]$ \\
      \bottomrule
    \end{tabular*}
  \end{minipage}
\end{table}

\paragraph{Components, joint coefficients and age interactions (g--i).}
Panel g plots the standardized component coefficients E1--E3
(Tables~\ref{tab:findings:panel-symbols} and~\ref{tab:findings:fig5-g}) under
ten specifications, defined as in panel a: NW and EA for H1; W and EA for H3;
F, W and CC for L6; EA for L5; and NW and EA for O1. The corresponding primary
components are in Figure~\ref{fig:findings:experimental-atlas}d. Panel h plots
the coefficients of the R1 joint model (Table~\ref{tab:findings:fig5-h}), which
regresses PWV on WHR, GlycA and SBP together, with GlycA, SBP and PWV taken as
the measurements nearest the WHR measurement day; the coefficients are in
measurement units and give the conditional difference in PWV with the other
variables held fixed. P is the primary window of this model, N and W are
narrower and wider windows, and A adds BMI under the primary window; P and W
here differ from P in panels b and c and from the design class W. Panel i plots
age-interaction coefficients as changes in standardized coefficients per 10
years (Table~\ref{tab:findings:fig5-i}). E1--E3 of L3 come from the L3 age
model of panel e; E1--E3 and T of R1 come from the primary R1 model with
interactions of age with every upstream variable added, with age at the WHR
measurement centered at the sample mean of 52.11 years, and T is the change in
the total WHR--PWV slope. $P(0)$ is the R1 serial association at the mean age
in this model, and $P'(0)$ is its derivative with respect to age in decades,
the change in the serial association per 10 years at the mean age.

\FloatBarrier

\subsection{Sensitivity analyses of the WHR--ALT--DBP--PWV joint model}
\label{app:findings:h1-joint}

\paragraph{Joint model.}
The joint model of Figure~\ref{fig:findings:network-overview}c and
Section~\ref{sec:findings:vascular-convergence} describes WHR--ALT--DBP--PWV
with three linear equations (ALT on WHR; DBP on WHR and ALT; PWV on WHR, ALT
and DBP), each adjusted for age and sex, with the variables standardized within
the analysis sample and symmetric windows taken between adjacent measurements.
The coefficients $a$, $b$ and $c$ are those of WHR--ALT, ALT--DBP and DBP--PWV,
respectively; WHR--DBP and ALT--PWV are the two bypass coefficients; and
WHR--PWV is the direct association. The full serial product is $abc$; the ALT
bypass product is $a$ times the ALT--PWV coefficient, the DBP bypass product is
the WHR--DBP coefficient times $c$, and the total association is the direct
association plus the three products. In the notation of
Table~\ref{tab:findings:panel-symbols}, $a$, $b$ and $c$ are E1--E3, $abc$ is S,
the ALT and DBP bypass products are B1 and B2, and the direct association is D.
Tables~\ref{tab:findings:h1-windows} to~\ref{tab:findings:h1-interaction} give
the estimates of each analysis with 95\% CIs, with coefficients and products in
SD/SD.

\paragraph{Windows and negative control.}
The narrower window (NW) narrows the windows between adjacent measurements and
takes contemporaneous or nearest measurements, and the wider window (W) widens
the windows between adjacent measurements
(Table~\ref{tab:findings:h1-windows}). The negative control takes the
left-minus-right difference in lower-limb PWV as the outcome, requires
measurements on both sides and otherwise follows the primary model.

\begin{table}[tbp]
  \centering
  \caption{\textbf{Primary model, windows and negative control of the WHR--ALT--DBP--PWV joint model.} Estimates
  in SD/SD with 95\% CIs below; NW, narrower window; W, wider window; in the negative control, PWV denotes
  the left-minus-right PWV difference; ---, not estimated in that analysis.}
  \label{tab:findings:h1-windows}
  \footnotesize
  \setlength{\tabcolsep}{1pt}
  \renewcommand{\arraystretch}{0.95}
\begin{tabular*}{\linewidth}{@{\extracolsep{\fill}}lcccc@{}}
    \toprule
    Parameter & Primary & NW & W & Negative control \\
    \midrule
    $a$ (WHR--ALT) & $0.2111$ & $0.2326$ & $0.2096$ & $0.2081$ \\
     & {\scriptsize $[0.1797,0.2425]$} & {\scriptsize $[0.1919,0.2732]$} & {\scriptsize $[0.1801,0.239]$} & {\scriptsize $[0.1762,0.2401]$} \\[1pt]
    $b$ (ALT--DBP) & $0.1161$ & $0.1066$ & $0.1053$ & $0.1159$ \\
     & {\scriptsize $[0.08709,0.1451]$} & {\scriptsize $[0.06715,0.1461]$} & {\scriptsize $[0.07867,0.1319]$} & {\scriptsize $[0.08608,0.1457]$} \\[1pt]
    $c$ (DBP--PWV) & $0.4379$ & $0.436$ & $0.4306$ & $0.05205$ \\
     & {\scriptsize $[0.4143,0.4616]$} & {\scriptsize $[0.4027,0.4694]$} & {\scriptsize $[0.4084,0.4527]$} & {\scriptsize $[0.02458,0.07953]$} \\[1pt]
    WHR--DBP bypass & $0.2231$ & $0.2249$ & $0.2317$ & $0.2166$ \\
\bottomrule
  \end{tabular*}
\end{table}

\paragraph{Sex.}
The sex analysis lets each network coefficient vary between the two sex
categories, First and Second, within one model, and gives the category-specific
coefficients and products, the difference in $abc$ between the categories and a
joint test that $a$, $b$ and $c$ are equal across the categories
(Table~\ref{tab:findings:h1-sex}).

\begin{table}[tbp]
  \centering
  \caption{\textbf{Sex categories in the WHR--ALT--DBP--PWV joint model.} Estimates for First and Second in
  SD/SD with 95\%
  CIs, the difference in $abc$ between the categories and the joint test of equal $a$, $b$ and $c$.}
  \label{tab:findings:h1-sex}
  \footnotesize
  \setlength{\tabcolsep}{3pt}
  \renewcommand{\arraystretch}{0.95}
  \begin{tabular*}{\linewidth}{@{\extracolsep{\fill}}lrcrc@{}}
    \toprule
    & \multicolumn{2}{c}{First} & \multicolumn{2}{c}{Second} \\
    \cmidrule(lr){2-3}\cmidrule(l){4-5}
    Parameter & Value & 95\% CI & Value & 95\% CI \\
    \midrule
    $a$ (WHR--ALT) & $0.1727$ & $[0.1397,0.2057]$ & $0.2567$ & $[0.2031,0.3103]$ \\
    $b$ (ALT--DBP) & $0.1316$ & $[0.08263,0.1806]$ & $0.1068$ & $[0.07155,0.142]$ \\
    $c$ (DBP--PWV) & $0.4583$ & $[0.428,0.4886]$ & $0.4207$ & $[0.3847,0.4566]$ \\
    $abc$ & $0.01042$ & $[0.006185,0.01465]$ & $0.01153$ & $[0.006926,0.01613]$ \\
    ALT bypass product & $-0.002542$ & $[-0.009603,0.004519]$ & $0.006177$ & $[-0.0004949,0.01285]$ \\
    DBP bypass product & $0.0799$ & $[0.05996,0.09985]$ & $0.1183$ & $[0.09706,0.1395]$ \\
    WHR--PWV direct & $0.09119$ & $[0.05721,0.1252]$ & $-0.0002928$ & $[-0.03884,0.03826]$ \\
    \midrule
    \multicolumn{5}{@{}l}{$abc$, First $-$ Second: $-0.001112$ $[-0.007301,0.005076]$} \\
    \multicolumn{5}{@{}l}{Joint test of equal $a$, $b$ and $c$: $P=0.01884$} \\
    \bottomrule
  \end{tabular*}
\end{table}

\paragraph{Extended adjustment.}
The extended adjustment adds BMI and current smoking status to age and sex
(Table~\ref{tab:findings:h1-adjust}). The primary adjustment is fitted
separately on the full available sample (the primary model) and on the common
sample with both BMI and smoking records; on the common sample, the primary and
extended adjustments use the same sample and the same standardization, and
their difference is the change in $a$, $b$, $c$ and $abc$ from adding BMI and
smoking, with a joint test of the four changes.

\begin{table}[tbp]
  \centering
  \caption{\textbf{Extended adjustment of the WHR--ALT--DBP--PWV joint model.} Primary adjustment on the full
  sample and on the common sample, extended adjustment on the common sample, and the change (extended minus
  primary on the common sample); estimates in SD/SD with 95\% CIs below; ---, not estimated.}
  \label{tab:findings:h1-adjust}
  \footnotesize
  \setlength{\tabcolsep}{1.2pt}
  \renewcommand{\arraystretch}{0.95}
  \begin{tabular*}{\linewidth}{@{\extracolsep{\fill}}lcccc@{}}
    \toprule
    & \multicolumn{2}{c}{Primary adjustment} & Extended & Change \\
    \cmidrule(lr){2-3}
    Parameter & Full sample & Common sample & Common sample & Common sample \\
    \midrule
    $a$ (WHR--ALT) & $0.2111$ & $0.1975$ & $0.1084$ & $-0.08913$ \\
     & {\scriptsize $[0.1797,0.2425]$} & {\scriptsize $[0.1651,0.23]$} & {\scriptsize $[0.07453,0.1422]$} & {\scriptsize $[-0.104,-0.0743]$} \\[1pt]
    $b$ (ALT--DBP) & $0.1161$ & $0.1167$ & $0.06815$ & $-0.04851$ \\
     & {\scriptsize $[0.08709,0.1451]$} & {\scriptsize $[0.08529,0.148]$} & {\scriptsize $[0.04076,0.09554]$} & {\scriptsize $[-0.05953,-0.03749]$} \\[1pt]
    $c$ (DBP--PWV) & $0.4379$ & $0.4346$ & $0.448$ & $0.01341$ \\
     & {\scriptsize $[0.4143,0.4616]$} & {\scriptsize $[0.4092,0.46]$} & {\scriptsize $[0.4211,0.4749]$} & {\scriptsize $[0.006719,0.0201]$} \\[1pt]
    $abc$ & $0.01073$ & $0.01001$ & $0.003309$ & $-0.006704$ \\
     & {\scriptsize $[0.007575,0.01389]$} & {\scriptsize $[0.006832,0.0132]$} & {\scriptsize $[0.001611,0.005007]$} & {\scriptsize $[-0.008586,-0.004823]$} \\[1pt]
    ALT bypass product & $0.00219$ & $0.002267$ & $0.0019$ & --- \\
     & {\scriptsize $[-0.002308,0.006688]$} & {\scriptsize $[-0.002135,0.006668]$} & {\scriptsize $[-0.000555,0.004355]$} &  \\[1pt]
    DBP bypass product & $0.09771$ & $0.0942$ & $0.02988$ & --- \\
     & {\scriptsize $[0.08289,0.1125]$} & {\scriptsize $[0.07871,0.1097]$} & {\scriptsize $[0.01455,0.04521]$} &  \\[1pt]
    WHR--PWV direct & $0.04861$ & $0.0515$ & $0.07342$ & --- \\
     & {\scriptsize $[0.02241,0.07481]$} & {\scriptsize $[0.02347,0.07952]$} & {\scriptsize $[0.04325,0.1036]$} &  \\[1pt]
    \midrule
    \multicolumn{4}{@{}l}{Joint test of the four changes} & $P=8.455\times10^{-35}$ \\
    \bottomrule
  \end{tabular*}
\end{table}

\paragraph{Interactions and unmeasured confounding.}
The interaction analysis adds WHR$\times$ALT to the DBP equation and
WHR$\times$ALT, WHR$\times$DBP and ALT$\times$DBP to the PWV equation; each
interaction coefficient is the change in a standardized slope per 1-SD increase
in the modifier, $abc$ is evaluated at the means of the variables, and a joint
test covers the four interactions (Table~\ref{tab:findings:h1-interaction}).
The robustness value RV of $a$, $b$ or $c$ is the partial $R^2$ that one
unmeasured confounder would need with both the predictor and the outcome of that
coefficient to reduce its point estimate to zero; the weakest-edge summary is
the smallest of the three.

\begin{table}[tbp]
  \centering
  \caption{\textbf{Interactions and robustness values of the WHR--ALT--DBP--PWV joint model.} Interaction
  coefficients are changes in standardized slopes per SD of the modifier, and $abc$ is evaluated at the
  means; the robustness value RV is a partial $R^2$, and the weakest-edge summary is the smallest RV of $a$,
  $b$ and $c$.}
  \label{tab:findings:h1-interaction}
  \footnotesize
  \setlength{\tabcolsep}{3pt}
  \renewcommand{\arraystretch}{0.95}
  \begin{tabular}{@{}lrc@{}}
    \toprule
    Term & Value & 95\% CI \\
    \midrule
    WHR$\times$ALT, DBP equation & $0.02393$ & $[-0.004099,0.05196]$ \\
    WHR$\times$ALT, PWV equation & $-0.006987$ & $[-0.02949,0.01551]$ \\
    WHR$\times$DBP, PWV equation & $-0.0441$ & $[-0.06773,-0.02048]$ \\
    ALT$\times$DBP, PWV equation & $0.01668$ & $[-0.007582,0.04094]$ \\
    $abc$ at the means & $0.009722$ & $[0.006598,0.01285]$ \\
    Joint test of the four interactions & \multicolumn{2}{l}{$P=0.001917$} \\
    \midrule
    \multicolumn{3}{@{}l}{\emph{Robustness value (partial $R^2$)}} \\
    $a$ (WHR--ALT) & $0.1633$ & \\
    $b$ (ALT--DBP) & $0.1081$ & \\
    $c$ (DBP--PWV) & $0.3947$ & \\
    Weakest edge, smallest of the three & $0.1081$ & \\
    \bottomrule
  \end{tabular}
\end{table}

\FloatBarrier

\subsection{Replication in external cohorts}
\label{app:findings:replication}

\paragraph{External cohorts and tests.}
Both external cohorts are cross-sectional: 912 adults from a population study
\citep{fukuda2014ggtbmjopen,fukuda2014ggtdata} and 879 hospitalized patients with diabetes in
Chengdu \citep{ning2022arterial}. Seven tests re-estimate five HPP paths, H1,
L1, L6, T1 and T2 (Tables~\ref{tab:findings:pathway-codes}
and~\ref{tab:findings:fig5-codes}), in these cohorts
(Table~\ref{tab:findings:replication}). BMI replaces WHR and android fat;
baPWV replaces lower-limb PWV, taking the higher of the two sides in the
912-adult cohort and the mean of both sides in the 879-adult cohort; every model
is adjusted for age and sex, the path variables are standardized within each
model, and E1--E3 and S are defined as in
Table~\ref{tab:findings:panel-symbols}. A test counts as replicated when S has
the sign of the HPP estimate and is significant after Holm correction within its
cohort, and every component coefficient has the HPP sign with a 95\% CI
excluding zero. The 95\% CI and P of S come from a participant bootstrap;
${<}0.0012$ is the smallest Holm-adjusted P attainable at the bootstrap resolution.

\begin{table}[tbp]
  \centering
  \caption{\textbf{Seven tests in the external cohorts.} Component coefficients and serial estimates S,
  standardized within each model (SD/SD), with 95\% CIs below; P, Holm-adjusted P of S within each cohort,
  where the adjustment in the 912-adult cohort also covers an ALT--TG--baPWV test (L7, $P=0.0016$);
  $\checkmark$, replicated; ---, no E3 for three-node paths.}
  \label{tab:findings:replication}
  \footnotesize
  \setlength{\tabcolsep}{2pt}
  \renewcommand{\arraystretch}{0.95}
  \begin{tabular*}{\linewidth}{@{\extracolsep{\fill}}lcccccrc@{}}
    \toprule
    Path to baPWV & HPP & E1 & E2 & E3 & S & P & \\
    \midrule
    \multicolumn{8}{@{}l}{\emph{912-adult cohort}} \\
    BMI--ALT--DBP & H1 & $0.3772$ & $0.05499$ & $0.4468$ & $0.009268$ & $0.054$ &  \\
     & & {\scriptsize $[0.2775,0.4768]$} & {\scriptsize $[-0.00671,0.1167]$} & {\scriptsize $[0.3826,0.511]$} & {\scriptsize $[-0.0001411,0.01978]$} & & \\[1pt]
    BMI--TG--DBP & L1 & $0.2277$ & $0.1009$ & $0.4448$ & $0.01022$ & $0.0012$ & $\checkmark$ \\
     & & {\scriptsize $[0.1508,0.3046]$} & {\scriptsize $[0.04618,0.1556]$} & {\scriptsize $[0.3791,0.5104]$} & {\scriptsize $[0.004551,0.01834]$} & & \\[1pt]
    GGT--TG--SBP & L6 & $0.2731$ & $0.1405$ & $0.4535$ & $0.01741$ & ${<}0.0012$ & $\checkmark$ \\
     & & {\scriptsize $[0.1255,0.4207]$} & {\scriptsize $[0.05838,0.2227]$} & {\scriptsize $[0.3904,0.5166]$} & {\scriptsize $[0.009174,0.02936]$} & & \\[1pt]
    ALT--DBP & T1 & $0.1774$ & $0.4178$ & --- & $0.07413$ & ${<}0.0012$ & $\checkmark$ \\
     & & {\scriptsize $[0.1124,0.2425]$} & {\scriptsize $[0.3562,0.4794]$} &  & {\scriptsize $[0.0501,0.1025]$} & & \\[1pt]
    TG--DBP & T2 & $0.173$ & $0.4188$ & --- & $0.07247$ & ${<}0.0012$ & $\checkmark$ \\
     & & {\scriptsize $[0.08798,0.2581]$} & {\scriptsize $[0.3561,0.4815]$} &  & {\scriptsize $[0.04898,0.1012]$} & & \\[1pt]
    \midrule
    \multicolumn{8}{@{}l}{\emph{879-adult cohort}} \\
    BMI--TG--DBP & L1 & $0.1245$ & $0.01164$ & $0.289$ & $0.0004188$ & $1$ &  \\
     & & {\scriptsize $[0.0638,0.1852]$} & {\scriptsize $[-0.05452,0.0778]$} & {\scriptsize $[0.2205,0.3574]$} & {\scriptsize $[-0.002096,0.002983]$} & & \\[1pt]
    TG--DBP & T2 & $0.02246$ & $0.284$ & --- & $0.006378$ & $1$ &  \\
     & & {\scriptsize $[-0.04203,0.08695]$} & {\scriptsize $[0.2145,0.3535]$} &  & {\scriptsize $[-0.01159,0.02557]$} & & \\[1pt]
    \bottomrule
  \end{tabular*}
\end{table}

\FloatBarrier

\FloatBarrier
\section{Research-Agent Designs and HPP Results of Ten Research Questions}
\label{app:rq-designs}
\sloppy

For ten research questions, this appendix lists the analysis units designed by the
research agents, each with its design, its model and its HPP results; within each
question the units are grouped by analytic purpose. Intervals are at the level stated
for each unit; 97.5\% intervals are Bonferroni-adjusted for a family formed by the
path products of two units. A $P$ value printed as 0 lies below the recorded
numerical precision.

\subsection{Hand grip strength--GlycA--DBP--cIMT}
\label{app:rq-designs:f559}

\paragraph{Research question and measures.}
The research question is the association of grip strength with carotid intima-media thickness along the serial path through GlycA and diastolic blood pressure, with data from the HPP cohort. Exposure: grip strength (mean of the available left- and right-hand values); mediators: GlycA (glycoprotein acetyl signal measured by NMR metabolomics, a marker of systemic inflammation) and seated diastolic blood pressure (mean of the available values of two seated readings); outcome: carotid intima-media thickness (cIMT, mean of the available left and right sides). Unless stated otherwise, coefficients are SD changes per 1 SD after standardization within each unit's analysis sample.

\subsubsection*{Primary pathways}

\paragraph{1.\ Serial path: grip strength $\to$ GlycA $\to$ diastolic blood pressure $\to$ carotid IMT}
\textit{Design.} GlycA measured within 365 days before or after grip strength, diastolic blood pressure within 90 days before or after GlycA (same day allowed), and carotid IMT measured 30--1825 days after diastolic blood pressure (all bounds inclusive); one chain per participant. All four equations share the same sample, with adjustment for approximate age on the grip strength measurement date, sex, recruitment cohort, and the most recent BMI on or within 730 days before the earliest measurement date in the chain; continuous variables standardized within the analysis sample, with sex and cohort as categorical terms.

\textit{Model.} $M_1 = a_1 A + \gamma_1^\top C + \varepsilon_1$; $M_2 = a_2 A + d_{21} M_1 + \gamma_2^\top C + \varepsilon_2$; $Y = c' A + b_1 M_1 + b_2 M_2 + \gamma_3^\top C + \varepsilon_3$; $Y = c A + \gamma_T^\top C + \varepsilon_T$. All four are linear regressions (with intercept); $A$, $M_1$, $M_2$ and $Y$ are standardized grip strength (larger of left and right hand), GlycA, seated diastolic blood pressure (mean of two readings) and carotid IMT (mean of left and right sides), respectively, and $C$ denotes the covariates. $a_1$ is the SD change in GlycA per 1 SD higher grip strength; $d_{21}$ is the slope of diastolic blood pressure on GlycA adjusted for grip strength; $b_2$ is the slope of IMT on diastolic blood pressure adjusted for grip strength and GlycA; $c'$ is the slope of IMT on grip strength adjusted for both mediators; $c$ is the total slope of IMT on grip strength adjusted for $C$ only; the serial path product is $\theta = a_1 \cdot d_{21} \cdot b_2$.

\textit{Intervals.} $a_1$, $d_{21}$, $b_2$, $c'$, $c$: 95\% confidence intervals; $\theta$: bootstrap 95\% percentile interval. $P$ values not adjusted for multiple comparisons.
{\footnotesize\setlength{\tabcolsep}{3pt}
\begin{longtable}{@{}>{\raggedright\arraybackslash}p{0.14\linewidth}>{\raggedright\arraybackslash}p{0.29\linewidth}>{\raggedleft\arraybackslash}p{0.13\linewidth}>{\raggedright\arraybackslash}p{0.22\linewidth}>{\raggedleft\arraybackslash}p{0.14\linewidth}@{}}
\toprule
Block & Parameter & Estimate & Interval & $P$ \\
\midrule\endhead
Serial path index & $\theta = a_1 \cdot d_{21} \cdot b_2$: serial path product association index & $-0.002119$ & $[-0.006861$, $0.0006244]$ &  \\
Grip strength $\to$ IMT association & $c'$: standardized slope for grip strength $\to$ carotid IMT adjusted for GlycA and diastolic blood pressure ($A \to Y \mid M_1, M_2$) & $0.001609$ & $[-0.1537$, $0.157]$ & \mbox{$0.9838$} \\
 & $c$: adjusted total standardized slope for grip strength $\to$ carotid IMT ($A \to Y$) & $0.01867$ & $[-0.1384$, $0.1758]$ & \mbox{$0.8155$} \\
Path components & $a_1$: standardized slope for grip strength $\to$ GlycA ($A \to M_1$) & $-0.1114$ & $[-0.2644$, $0.04146]$ & \mbox{$0.1527$} \\
 & $d_{21}$: standardized slope for GlycA $\to$ seated diastolic blood pressure adjusted for grip strength ($M_1 \to M_2 \mid A$) & $0.1646$ & $[0.06322$, $0.266]$ & \mbox{$0.001521$} \\
 & $b_2$: standardized slope for seated diastolic blood pressure $\to$ carotid IMT adjusted for grip strength and GlycA ($M_2 \to Y \mid A, M_1$) & $0.1155$ & $[0.0342$, $0.1967]$ & \mbox{$0.005461$} \\
\bottomrule
\end{longtable}}

\paragraph{2.\ Exposure--outcome association: grip strength and carotid IMT ($\pm$365 days)}
\textit{Design.} Grip strength and carotid intima-media thickness (IMT) measured in the same period: each IMT measurement date serves as the anchor, with the closest grip strength measurement within 365 days before or after it (inclusive); the same participant may contribute multiple anchor dates, $n = 7909$. Grip strength and IMT each taken as the mean of the available left and right values. Adjustment for approximate age (IMT measurement year minus birth year), the closest BMI within $\pm$365 days of the anchor date, and sex; continuous variables kept on the original scale.

\textit{Model.} $Y = \beta_0 + \beta_A \cdot A + \beta_{\mathrm{age}} \cdot \mathrm{age} + \beta_{\mathrm{BMI}} \cdot \mathrm{BMI} + \gamma_{\mathrm{sex}} + \varepsilon$ (linear regression; $A$ is grip strength and $Y$ is carotid IMT, both on the original scale). $\beta_A$ is the difference in carotid IMT per 1 original unit higher grip strength with age, BMI and sex held fixed.

\textit{Intervals.} 95\% confidence interval.
{\footnotesize\setlength{\tabcolsep}{3pt}
\begin{longtable}{@{}>{\raggedright\arraybackslash}p{0.14\linewidth}>{\raggedright\arraybackslash}p{0.29\linewidth}>{\raggedleft\arraybackslash}p{0.13\linewidth}>{\raggedright\arraybackslash}p{0.22\linewidth}>{\raggedleft\arraybackslash}p{0.14\linewidth}@{}}
\toprule
Block & Parameter & Estimate & Interval & $P$ \\
\midrule\endhead
$\pm$365-day window & $\beta_A$: adjusted grip strength--carotid IMT association coefficient & $0.0005656$ & $[0.0002785$, $0.0008527]$ & \mbox{$0.000113$} \\
\bottomrule
\end{longtable}}

\subsubsection*{Adjacent segments and pairwise associations}

\paragraph{3.\ Pairwise association: grip strength $\to$ GlycA ($\pm$365 days)}
\textit{Design.} GlycA measured within 365 days before or after the grip strength measurement (inclusive, same day allowed); one measurement pair per participant. Adjustment for approximate age on the grip strength measurement date, sex, recruitment cohort, and the most recent BMI on or within 730 days before the earlier of the two measurement dates; continuous variables standardized within the analysis sample, with sex and cohort as categorical terms.

\textit{Model.} $M_1 = \beta_0 + \beta \cdot A + \gamma^\top C + \varepsilon$ (linear regression; $A$ is grip strength (larger of left and right hand) and $M_1$ is GlycA, both standardized, and $C$ denotes the covariates). $\beta$ is the SD difference in GlycA per 1 SD higher grip strength. Bias bound $= \sqrt{R^2_{Y\sim U} \cdot R^2_{D\sim U} / (1 - R^2_{D\sim U})} \cdot \hat{\sigma}_Y / \hat{\sigma}_D$, where $R^2_{D\sim U}$ and $R^2_{Y\sim U}$ are the partial $R^2$ values of a hypothetical omitted confounder $U$ with the coefficient's regressor $D$ and with the equation's outcome, $\hat{\sigma}_Y$ is the residual SD of the equation, and $\hat{\sigma}_D$ is the residual SD from regressing $D$ on the remaining terms of the equation; each of the two partial $R^2$ values takes 0.01, 0.05, 0.10, 0.20, 0.30 and 0.50.

\textit{Intervals.} $\beta$: 95\% confidence interval. Bias-bound grid: lower and upper limits are the coefficient $\pm$ the bias bound.
{\footnotesize\setlength{\tabcolsep}{3pt}
\begin{longtable}{@{}>{\raggedright\arraybackslash}p{0.14\linewidth}>{\raggedright\arraybackslash}p{0.29\linewidth}>{\raggedleft\arraybackslash}p{0.13\linewidth}>{\raggedright\arraybackslash}p{0.22\linewidth}>{\raggedleft\arraybackslash}p{0.14\linewidth}@{}}
\toprule
Block & Parameter & Estimate & Interval & $P$ \\
\midrule\endhead
Symmetric window, $\pm$365 days & $\beta$: adjusted standardized slope for grip strength $\to$ GlycA & $-0.2853$ & $[-0.3662$, $-0.2045]$ & \mbox{$6.545\times10^{-12}$} \\
\bottomrule
\end{longtable}}

{\scriptsize\setlength{\tabcolsep}{3pt}
\begin{longtable}{@{}lc@{}}
\toprule
Partial $R^2$ of U (regressor; outcome) & $\beta$ \\
\midrule\endhead
Coefficient & $-0.2853$ \\
0.01; 0.01 & $[-0.3013, -0.2694]$ \\
0.01; 0.05 & $[-0.321, -0.2496]$ \\
0.01; 0.10 & $[-0.3358, -0.2348]$ \\
0.01; 0.20 & $[-0.3567, -0.2139]$ \\
0.01; 0.30 & $[-0.3728, -0.1979]$ \\
0.01; 0.50 & $[-0.3982, -0.1724]$ \\
0.05; 0.01 & $[-0.3218, -0.2489]$ \\
0.05; 0.05 & $[-0.3668, -0.2038]$ \\
0.05; 0.10 & $[-0.4006, -0.1701]$ \\
0.05; 0.20 & $[-0.4483, -0.1224]$ \\
0.05; 0.30 & $[-0.4849, -0.08572]$ \\
0.05; 0.50 & $[-0.543, -0.02764]$ \\
0.10; 0.01 & $[-0.3383, -0.2324]$ \\
0.10; 0.05 & $[-0.4037, -0.1669]$ \\
0.10; 0.10 & $[-0.4528, -0.1179]$ \\
0.10; 0.20 & $[-0.5221, -0.04853]$ \\
0.10; 0.30 & $[-0.5754, 0.00469]$ \\
0.10; 0.50 & $[-0.6597, 0.08908]$ \\
0.20; 0.01 & $[-0.3648, -0.2059]$ \\
0.20; 0.05 & $[-0.4629, -0.1077]$ \\
0.20; 0.10 & $[-0.5365, -0.03417]$ \\
0.20; 0.20 & $[-0.6405, 0.06987]$ \\
0.20; 0.30 & $[-0.7204, 0.1497]$ \\
0.20; 0.50 & $[-0.847, 0.2763]$ \\
0.30; 0.01 & $[-0.3893, -0.1813]$ \\
0.30; 0.05 & $[-0.5179, -0.0528]$ \\
0.30; 0.10 & $[-0.6142, 0.04352]$ \\
0.30; 0.20 & $[-0.7504, 0.1797]$ \\
0.30; 0.30 & $[-0.8549, 0.2843]$ \\
0.30; 0.50 & $[-1.021, 0.45]$ \\
0.50; 0.01 & $[-0.4442, -0.1265]$ \\
0.50; 0.05 & $[-0.6405, 0.06987]$ \\
0.50; 0.10 & $[-0.7877, 0.217]$ \\
0.50; 0.20 & $[-0.9957, 0.4251]$ \\
0.50; 0.30 & $[-1.155, 0.5847]$ \\
0.50; 0.50 & $[-1.409, 0.8379]$ \\
\bottomrule
\end{longtable}}

\paragraph{4.\ Pairwise association: GlycA $\to$ diastolic blood pressure ($\pm$90 days)}
\textit{Design.} Seated blood pressure measured within 90 days before or after the GlycA measurement (inclusive, same day allowed); one measurement pair per participant, $n = 1544$. Adjustment for approximate age on the GlycA measurement date, sex, recruitment cohort, and the most recent BMI on or within 730 days before the earlier of the two measurement dates; continuous variables standardized within the analysis sample, with sex and cohort as categorical terms.

\textit{Model.} $M_2 = \beta_0 + \beta \cdot M_1 + \gamma^\top C + \varepsilon$ (linear regression; $M_1$ is GlycA and $M_2$ is seated diastolic blood pressure (mean of two readings), both standardized, and $C$ denotes the covariates). $\beta$ is the SD difference in diastolic blood pressure per 1 SD higher GlycA.

\textit{Intervals.} 95\% confidence interval.
{\footnotesize\setlength{\tabcolsep}{3pt}
\begin{longtable}{@{}>{\raggedright\arraybackslash}p{0.14\linewidth}>{\raggedright\arraybackslash}p{0.29\linewidth}>{\raggedleft\arraybackslash}p{0.13\linewidth}>{\raggedright\arraybackslash}p{0.22\linewidth}>{\raggedleft\arraybackslash}p{0.14\linewidth}@{}}
\toprule
Block & Parameter & Estimate & Interval & $P$ \\
\midrule\endhead
Symmetric window, $\pm$90 days & $\beta$: adjusted standardized slope for GlycA $\to$ seated diastolic blood pressure & $0.08383$ & $[0.03421$, $0.1334]$ & \mbox{$0.0009403$} \\
\bottomrule
\end{longtable}}

\paragraph{5.\ Pairwise association: diastolic blood pressure $\to$ carotid IMT (30--1825 days)}
\textit{Design.} Carotid ultrasound performed 30--1825 days after the seated blood pressure measurement (inclusive); one measurement pair per participant, $n = 2195$. Adjustment for approximate age on the blood pressure measurement date, sex, recruitment cohort, and the most recent BMI on or within 730 days before the blood pressure measurement date; continuous variables standardized within the analysis sample, with sex and cohort as categorical terms.

\textit{Model.} $Y = \beta_0 + \beta \cdot M_2 + \gamma^\top C + \varepsilon$ (linear regression; $M_2$ is seated diastolic blood pressure (mean of two readings) and $Y$ is carotid IMT (mean of left and right sides), both standardized, and $C$ denotes the covariates). $\beta$ is the SD difference in IMT per 1 SD higher diastolic blood pressure.

\textit{Intervals.} 95\% confidence interval.
{\footnotesize\setlength{\tabcolsep}{3pt}
\begin{longtable}{@{}>{\raggedright\arraybackslash}p{0.14\linewidth}>{\raggedright\arraybackslash}p{0.29\linewidth}>{\raggedleft\arraybackslash}p{0.13\linewidth}>{\raggedright\arraybackslash}p{0.22\linewidth}>{\raggedleft\arraybackslash}p{0.14\linewidth}@{}}
\toprule
Block & Parameter & Estimate & Interval & $P$ \\
\midrule\endhead
Forward window, 30--1825 days & $\beta$: adjusted standardized slope for seated diastolic blood pressure $\to$ carotid IMT & $0.08635$ & $[0.04715$, $0.1256]$ & \mbox{$1.634\times10^{-5}$} \\
\bottomrule
\end{longtable}}

\paragraph{6.\ Pairwise total association: grip strength $\to$ carotid IMT ($\pm$730 days)}
\textit{Design.} Carotid ultrasound performed within 730 days before or after the grip strength measurement (inclusive, same day allowed); one measurement pair per participant, $n = 7926$. Adjustment for approximate age on the grip strength measurement date, sex, recruitment cohort, and the most recent BMI on or within 730 days before the earlier of the two measurement dates; continuous variables standardized within the analysis sample, with sex and cohort as categorical terms.

\textit{Model.} $Y = \beta_0 + \beta \cdot A + \gamma^\top C + \varepsilon$ (linear regression; $A$ is grip strength (larger of left and right hand) and $Y$ is carotid IMT (mean of left and right sides), both standardized, and $C$ denotes the covariates). $\beta$ is the SD difference in IMT per 1 SD higher grip strength.

\textit{Intervals.} 95\% confidence interval.
{\footnotesize\setlength{\tabcolsep}{3pt}
\begin{longtable}{@{}>{\raggedright\arraybackslash}p{0.14\linewidth}>{\raggedright\arraybackslash}p{0.29\linewidth}>{\raggedleft\arraybackslash}p{0.13\linewidth}>{\raggedright\arraybackslash}p{0.22\linewidth}>{\raggedleft\arraybackslash}p{0.14\linewidth}@{}}
\toprule
Block & Parameter & Estimate & Interval & $P$ \\
\midrule\endhead
Symmetric window, $\pm$730 days & $\beta$: adjusted standardized total association slope for grip strength $\to$ carotid IMT & $0.06931$ & $[0.03507$, $0.1035]$ & \mbox{$7.288\times10^{-5}$} \\
\bottomrule
\end{longtable}}

\paragraph{7.\ Pairwise association: grip strength $\to$ diastolic blood pressure ($\pm$90 days)}
\textit{Design.} Seated blood pressure measured within 90 days before or after the grip strength measurement (inclusive, same day allowed); one measurement pair per participant, $n = 10344$. Adjustment for approximate age on the grip strength measurement date, sex, recruitment cohort, and the most recent BMI on or within 730 days before the earlier of the two measurement dates; continuous variables standardized within the analysis sample, with sex and cohort as categorical terms.

\textit{Model.} $M_2 = \beta_0 + \beta \cdot A + \gamma^\top C + \varepsilon$ (linear regression; $A$ is grip strength (larger of left and right hand) and $M_2$ is seated diastolic blood pressure (mean of two readings), both standardized, and $C$ denotes the covariates). $\beta$ is the SD difference in diastolic blood pressure per 1 SD higher grip strength. Bias bound $= \sqrt{R^2_{Y\sim U} \cdot R^2_{D\sim U} / (1 - R^2_{D\sim U})} \cdot \hat{\sigma}_Y / \hat{\sigma}_D$, where $R^2_{D\sim U}$ and $R^2_{Y\sim U}$ are the partial $R^2$ values of a hypothetical omitted confounder $U$ with the coefficient's regressor $D$ and with the equation's outcome, $\hat{\sigma}_Y$ is the residual SD of the equation, and $\hat{\sigma}_D$ is the residual SD from regressing $D$ on the remaining terms of the equation; each of the two partial $R^2$ values takes 0.01, 0.05, 0.10, 0.20, 0.30 and 0.50.

\textit{Intervals.} $\beta$: 95\% confidence interval. Bias-bound grid: lower and upper limits are the coefficient $\pm$ the bias bound.
{\footnotesize\setlength{\tabcolsep}{3pt}
\begin{longtable}{@{}>{\raggedright\arraybackslash}p{0.14\linewidth}>{\raggedright\arraybackslash}p{0.29\linewidth}>{\raggedleft\arraybackslash}p{0.13\linewidth}>{\raggedright\arraybackslash}p{0.22\linewidth}>{\raggedleft\arraybackslash}p{0.14\linewidth}@{}}
\toprule
Block & Parameter & Estimate & Interval & $P$ \\
\midrule\endhead
Symmetric window, $\pm$90 days & $\beta$: adjusted standardized slope for grip strength $\to$ seated diastolic blood pressure & $0.0931$ & $[0.06195$, $0.1243]$ & \mbox{$4.812\times10^{-9}$} \\
\bottomrule
\end{longtable}}

{\scriptsize\setlength{\tabcolsep}{3pt}
\begin{longtable}{@{}lc@{}}
\toprule
Partial $R^2$ of U (regressor; outcome) & $\beta$ \\
\midrule\endhead
Coefficient & $0.0931$ \\
0.01; 0.01 & $[0.0771, 0.1091]$ \\
0.01; 0.05 & $[0.05732, 0.1289]$ \\
0.01; 0.10 & $[0.0425, 0.1437]$ \\
0.01; 0.20 & $[0.02154, 0.1647]$ \\
0.01; 0.30 & $[0.005451, 0.1808]$ \\
0.01; 0.50 & $[-0.02005, 0.2063]$ \\
0.05; 0.01 & $[0.05657, 0.1296]$ \\
0.05; 0.05 & $[0.01142, 0.1748]$ \\
0.05; 0.10 & $[-0.02241, 0.2086]$ \\
0.05; 0.20 & $[-0.07026, 0.2565]$ \\
0.05; 0.30 & $[-0.107, 0.2932]$ \\
0.05; 0.50 & $[-0.1652, 0.3514]$ \\
0.10; 0.01 & $[0.04003, 0.1462]$ \\
0.10; 0.05 & $[-0.02558, 0.2118]$ \\
0.10; 0.10 & $[-0.07474, 0.2609]$ \\
0.10; 0.20 & $[-0.1443, 0.3305]$ \\
0.10; 0.30 & $[-0.1976, 0.3838]$ \\
0.10; 0.50 & $[-0.2822, 0.4684]$ \\
0.20; 0.01 & $[0.01349, 0.1727]$ \\
0.20; 0.05 & $[-0.08492, 0.2711]$ \\
0.20; 0.10 & $[-0.1587, 0.3449]$ \\
0.20; 0.20 & $[-0.2629, 0.4491]$ \\
0.20; 0.30 & $[-0.343, 0.5292]$ \\
0.20; 0.50 & $[-0.4698, 0.6561]$ \\
0.30; 0.01 & $[-0.01114, 0.1973]$ \\
0.30; 0.05 & $[-0.14, 0.3262]$ \\
0.30; 0.10 & $[-0.2365, 0.4227]$ \\
0.30; 0.20 & $[-0.3731, 0.5593]$ \\
0.30; 0.30 & $[-0.4778, 0.664]$ \\
0.30; 0.50 & $[-0.644, 0.8302]$ \\
0.50; 0.01 & $[-0.06612, 0.2523]$ \\
0.50; 0.05 & $[-0.2629, 0.4491]$ \\
0.50; 0.10 & $[-0.4104, 0.5966]$ \\
0.50; 0.20 & $[-0.619, 0.8052]$ \\
0.50; 0.30 & $[-0.779, 0.9652]$ \\
0.50; 0.50 & $[-1.033, 1.219]$ \\
\bottomrule
\end{longtable}}

\paragraph{8.\ Pairwise association: GlycA $\to$ carotid IMT (30--1825 days)}
\textit{Design.} Carotid ultrasound performed 30--1825 days after the GlycA measurement (inclusive); one measurement pair per participant, $n = 458$. Adjustment for approximate age on the GlycA measurement date, sex, recruitment cohort, and the most recent BMI on or within 730 days before the GlycA measurement date; continuous variables standardized within the analysis sample, with sex and cohort as categorical terms.

\textit{Model.} $Y = \beta_0 + \beta \cdot M_1 + \gamma^\top C + \varepsilon$ (linear regression; $M_1$ is GlycA and $Y$ is carotid IMT (mean of left and right sides), both standardized, and $C$ denotes the covariates). $\beta$ is the SD difference in IMT per 1 SD higher GlycA.

\textit{Intervals.} 95\% confidence interval.
{\footnotesize\setlength{\tabcolsep}{3pt}
\begin{longtable}{@{}>{\raggedright\arraybackslash}p{0.14\linewidth}>{\raggedright\arraybackslash}p{0.29\linewidth}>{\raggedleft\arraybackslash}p{0.13\linewidth}>{\raggedright\arraybackslash}p{0.22\linewidth}>{\raggedleft\arraybackslash}p{0.14\linewidth}@{}}
\toprule
Block & Parameter & Estimate & Interval & $P$ \\
\midrule\endhead
Forward window, 30--1825 days & $\beta$: adjusted standardized slope for GlycA $\to$ carotid IMT & $-0.09355$ & $[-0.1776$, $-0.009476]$ & \mbox{$0.02927$} \\
\bottomrule
\end{longtable}}

\paragraph{9.\ Single-mediator path: grip strength $\to$ GlycA $\to$ carotid IMT}
\textit{Design.} GlycA measured within 365 days before or after grip strength (same day allowed) and carotid IMT 30--1825 days after GlycA (all bounds inclusive); one chain per participant, $n = 450$. All three equations share the same sample, with adjustment for approximate age on the grip strength measurement date, sex, recruitment cohort, and the most recent BMI on or within 730 days before the earliest measurement date in the chain; continuous variables standardized within the analysis sample, with sex and cohort as categorical terms.

\textit{Model.} $M = a A + \gamma_1^\top C + \varepsilon_1$; $Y = c' A + b M + \gamma_2^\top C + \varepsilon_2$; $Y = c A + \gamma_T^\top C + \varepsilon_T$. All three are linear regressions (with intercept); $A$, $M$ and $Y$ are standardized grip strength (larger of left and right hand), GlycA and carotid IMT (mean of left and right sides), respectively, and $C$ denotes the covariates. $a$ is the SD change in GlycA per 1 SD higher grip strength; $b$ and $c'$ are the mutually adjusted slopes of IMT on GlycA and on grip strength, respectively; $c$ is the total slope of IMT on grip strength adjusted for $C$ only; the path product is $\theta = a \cdot b$.

\textit{Intervals.} $a$, $b$, $c'$, $c$: 95\% confidence intervals; $\theta$: bootstrap 95\% percentile interval. $P$ values not adjusted for multiple comparisons.
{\footnotesize\setlength{\tabcolsep}{3pt}
\begin{longtable}{@{}>{\raggedright\arraybackslash}p{0.14\linewidth}>{\raggedright\arraybackslash}p{0.29\linewidth}>{\raggedleft\arraybackslash}p{0.13\linewidth}>{\raggedright\arraybackslash}p{0.22\linewidth}>{\raggedleft\arraybackslash}p{0.14\linewidth}@{}}
\toprule
Block & Parameter & Estimate & Interval & $P$ \\
\midrule\endhead
Single-mediator path index & $\theta = a \cdot b$: single-mediator path product association index & $0.01095$ & $[-0.003799$, $0.03254]$ &  \\
Grip strength $\to$ IMT association & $c$: adjusted total standardized slope for grip strength $\to$ carotid IMT ($A \to Y$) & $0.02134$ & $[-0.1352$, $0.1779]$ & \mbox{$0.7888$} \\
 & $c'$: standardized slope for grip strength $\to$ carotid IMT adjusted for GlycA ($A \to Y \mid M$) & $0.0104$ & $[-0.1475$, $0.1683]$ & \mbox{$0.8971$} \\
Path components & $a$: standardized slope for grip strength $\to$ GlycA ($A \to M$) & $-0.1134$ & $[-0.2662$, $0.03944]$ & \mbox{$0.1455$} \\
 & $b$: standardized slope for GlycA $\to$ carotid IMT adjusted for grip strength ($M \to Y \mid A$) & $-0.09655$ & $[-0.1824$, $-0.01071]$ & \mbox{$0.02758$} \\
\bottomrule
\end{longtable}}

\paragraph{10.\ Single-mediator path: grip strength $\to$ diastolic blood pressure $\to$ carotid IMT}
\textit{Design.} Seated blood pressure measured within 90 days before or after grip strength (same day allowed) and carotid IMT 30--1825 days after the blood pressure measurement (all bounds inclusive); one chain per participant. All three equations share the same sample, with adjustment for approximate age on the grip strength measurement date, sex, recruitment cohort, and the most recent BMI on or within 730 days before the earliest measurement date in the chain; continuous variables standardized within the analysis sample, with sex and cohort as categorical terms.

\textit{Model.} $M = a A + \gamma_1^\top C + \varepsilon_1$; $Y = c' A + b M + \gamma_2^\top C + \varepsilon_2$; $Y = c A + \gamma_T^\top C + \varepsilon_T$. All three are linear regressions (with intercept); $A$, $M$ and $Y$ are standardized grip strength (larger of left and right hand), seated diastolic blood pressure (mean of two readings) and carotid IMT (mean of left and right sides), respectively, and $C$ denotes the covariates. $a$ is the SD change in seated diastolic blood pressure per 1 SD higher grip strength; $b$ and $c'$ are the mutually adjusted slopes of IMT on seated diastolic blood pressure and on grip strength, respectively; $c$ is the total slope of IMT on grip strength adjusted for $C$ only; the path product is $\theta = a \cdot b$. Bias bound $= \sqrt{R^2_{Y\sim U} \cdot R^2_{D\sim U} / (1 - R^2_{D\sim U})} \cdot \hat{\sigma}_Y / \hat{\sigma}_D$, where $R^2_{D\sim U}$ and $R^2_{Y\sim U}$ are the partial $R^2$ values of a hypothetical omitted confounder $U$ with the coefficient's regressor $D$ and with the equation's outcome, $\hat{\sigma}_Y$ is the residual SD of the equation, and $\hat{\sigma}_D$ is the residual SD from regressing $D$ on the remaining terms of the equation; each of the two partial $R^2$ values takes 0.01, 0.05, 0.10, 0.20, 0.30 and 0.50.

\textit{Intervals.} $a$, $b$, $c'$, $c$: 95\% confidence intervals; $\theta$: bootstrap 95\% percentile interval. $P$ values not adjusted for multiple comparisons. Bias-bound grid: lower and upper limits are the coefficient $\pm$ the bias bound.
{\footnotesize\setlength{\tabcolsep}{3pt}
\begin{longtable}{@{}>{\raggedright\arraybackslash}p{0.14\linewidth}>{\raggedright\arraybackslash}p{0.29\linewidth}>{\raggedleft\arraybackslash}p{0.13\linewidth}>{\raggedright\arraybackslash}p{0.22\linewidth}>{\raggedleft\arraybackslash}p{0.14\linewidth}@{}}
\toprule
Block & Parameter & Estimate & Interval & $P$ \\
\midrule\endhead
Single-mediator path index & $\theta = a \cdot b$: single-mediator path product association index & $0.006348$ & $[0.0001121$, $0.01371]$ &  \\
Grip strength $\to$ IMT association & $c$: adjusted total standardized slope for grip strength $\to$ carotid IMT ($A \to Y$) & $0.04613$ & $[-0.01922$, $0.1115]$ & \mbox{$0.1664$} \\
 & $c'$: standardized slope for grip strength $\to$ carotid IMT adjusted for seated diastolic blood pressure ($A \to Y \mid M$) & $0.03979$ & $[-0.02522$, $0.1048]$ & \mbox{$0.2302$} \\
Path components & $a$: standardized slope for grip strength $\to$ seated diastolic blood pressure ($A \to M$) & $0.0748$ & $[0.0036$, $0.146]$ & \mbox{$0.0395$} \\
 & $b$: standardized slope for seated diastolic blood pressure $\to$ carotid IMT adjusted for grip strength ($M \to Y \mid A$) & $0.08486$ & $[0.04541$, $0.1243]$ & \mbox{$2.557\times10^{-5}$} \\
\bottomrule
\end{longtable}}

{\scriptsize\setlength{\tabcolsep}{3pt}
\begin{longtable}{@{}lcc@{}}
\toprule
Partial $R^2$ of U (regressor; outcome) & $a$ & $b$ \\
\midrule\endhead
Coefficient & $0.0748$ & $0.08486$ \\
0.01; 0.01 & $[0.05766, 0.09195]$ & $[0.07543, 0.09429]$ \\
0.01; 0.05 & $[0.03647, 0.1131]$ & $[0.06377, 0.1059]$ \\
0.01; 0.10 & $[0.02059, 0.129]$ & $[0.05503, 0.1147]$ \\
0.01; 0.20 & $[-0.001864, 0.1515]$ & $[0.04268, 0.127]$ \\
0.01; 0.30 & $[-0.01909, 0.1687]$ & $[0.0332, 0.1365]$ \\
0.01; 0.50 & $[-0.04642, 0.196]$ & $[0.01817, 0.1515]$ \\
0.05; 0.01 & $[0.03567, 0.1139]$ & $[0.06333, 0.1064]$ \\
\bottomrule
\end{longtable}}

\paragraph{11.\ Pairwise association: grip strength--GlycA ($\pm$365 days)}
\textit{Design.} Grip strength and GlycA measured in the same period: each GlycA measurement date serves as the anchor, with the closest grip strength measurement within 365 days before or after it (inclusive); the same participant may contribute multiple anchor dates. Grip strength taken as the mean of the available left and right values. Adjustment for approximate age (GlycA measurement year minus birth year), the closest BMI within $\pm$365 days of the anchor date, and sex; continuous variables kept on the original scale.

\textit{Model.} $M_1 = \beta_0 + \beta_A \cdot A + \beta_{\mathrm{age}} \cdot \mathrm{age} + \beta_{\mathrm{BMI}} \cdot \mathrm{BMI} + \gamma_{\mathrm{sex}} + \varepsilon$ (linear regression; $A$ is grip strength and $M_1$ is GlycA, both on the original scale). $\beta_A$ is the difference in GlycA per 1 original unit higher grip strength with age, BMI and sex held fixed.

\textit{Intervals.} 95\% confidence interval.
{\footnotesize\setlength{\tabcolsep}{3pt}
\begin{longtable}{@{}>{\raggedright\arraybackslash}p{0.14\linewidth}>{\raggedright\arraybackslash}p{0.29\linewidth}>{\raggedleft\arraybackslash}p{0.13\linewidth}>{\raggedright\arraybackslash}p{0.22\linewidth}>{\raggedleft\arraybackslash}p{0.14\linewidth}@{}}
\toprule
Block & Parameter & Estimate & Interval & $P$ \\
\midrule\endhead
$\pm$365-day window & $\beta_A$: adjusted grip strength--GlycA association coefficient & $-0.003362$ & $[-0.004277$, $-0.002447]$ & \mbox{$6.046\times10^{-13}$} \\
\bottomrule
\end{longtable}}

\paragraph{12.\ Pairwise association: grip strength--seated diastolic blood pressure ($\pm$365 days)}
\textit{Design.} Grip strength and seated diastolic blood pressure measured in the same period: each blood pressure measurement date serves as the anchor, with the closest grip strength measurement within 365 days before or after it (inclusive); the same participant may contribute multiple anchor dates, $n = 10346$. Grip strength taken as the mean of the available left and right values, and diastolic blood pressure as the mean of the available values of the two seated readings. Adjustment for approximate age (blood pressure measurement year minus birth year), the closest BMI within $\pm$365 days of the anchor date, and sex; continuous variables kept on the original scale.

\textit{Model.} $M_2 = \beta_0 + \beta_A \cdot A + \beta_{\mathrm{age}} \cdot \mathrm{age} + \beta_{\mathrm{BMI}} \cdot \mathrm{BMI} + \gamma_{\mathrm{sex}} + \varepsilon$ (linear regression; $A$ is grip strength and $M_2$ is seated diastolic blood pressure, both on the original scale). $\beta_A$ is the difference in seated diastolic blood pressure per 1 original unit higher grip strength with age, BMI and sex held fixed.

\textit{Intervals.} 95\% confidence interval.
{\footnotesize\setlength{\tabcolsep}{3pt}
\begin{longtable}{@{}>{\raggedright\arraybackslash}p{0.14\linewidth}>{\raggedright\arraybackslash}p{0.29\linewidth}>{\raggedleft\arraybackslash}p{0.13\linewidth}>{\raggedright\arraybackslash}p{0.22\linewidth}>{\raggedleft\arraybackslash}p{0.14\linewidth}@{}}
\toprule
Block & Parameter & Estimate & Interval & $P$ \\
\midrule\endhead
$\pm$365-day window & $\beta_A$: adjusted grip strength--seated diastolic blood pressure association coefficient & $0.07306$ & $[0.04774$, $0.09838]$ & \mbox{$1.554\times10^{-8}$} \\
\bottomrule
\end{longtable}}

\paragraph{13.\ Pairwise association: GlycA--carotid IMT ($\pm$365 days)}
\textit{Design.} GlycA and carotid IMT measured in the same period: each IMT measurement date serves as the anchor, with the closest GlycA measurement within 365 days before or after it (inclusive); the same participant may contribute multiple anchor dates, $n = 821$. IMT taken as the mean of the available left and right values. Adjustment for approximate age (IMT measurement year minus birth year), the closest BMI within $\pm$365 days of the anchor date, and sex; continuous variables kept on the original scale.

\textit{Model.} $Y = \beta_0 + \beta_{M_1} \cdot M_1 + \beta_{\mathrm{age}} \cdot \mathrm{age} + \beta_{\mathrm{BMI}} \cdot \mathrm{BMI} + \gamma_{\mathrm{sex}} + \varepsilon$ (linear regression; $M_1$ is GlycA and $Y$ is carotid IMT, both on the original scale). $\beta_{M_1}$ is the difference in carotid IMT per 1 original unit higher GlycA with age, BMI and sex held fixed.

\textit{Intervals.} 95\% confidence interval.
{\footnotesize\setlength{\tabcolsep}{3pt}
\begin{longtable}{@{}>{\raggedright\arraybackslash}p{0.14\linewidth}>{\raggedright\arraybackslash}p{0.29\linewidth}>{\raggedleft\arraybackslash}p{0.13\linewidth}>{\raggedright\arraybackslash}p{0.22\linewidth}>{\raggedleft\arraybackslash}p{0.14\linewidth}@{}}
\toprule
Block & Parameter & Estimate & Interval & $P$ \\
\midrule\endhead
$\pm$365-day window & $\beta_{M_1}$: adjusted GlycA--carotid IMT association coefficient & $-0.07468$ & $[-0.1205$, $-0.02888]$ & \mbox{$0.001395$} \\
\bottomrule
\end{longtable}}

\paragraph{14.\ Pairwise association: seated diastolic blood pressure--carotid IMT ($\pm$365 days)}
\textit{Design.} Seated diastolic blood pressure and carotid IMT measured in the same period: each IMT measurement date serves as the anchor, with the closest seated diastolic blood pressure measurement within 365 days before or after it (inclusive); the same participant may contribute multiple anchor dates, $n = 7940$. Diastolic blood pressure taken as the mean of the available values of the two seated readings, and IMT as the mean of the available left and right values. Adjustment for approximate age (IMT measurement year minus birth year), the closest BMI within $\pm$365 days of the anchor date, and sex; continuous variables kept on the original scale.

\textit{Model.} $Y = \beta_0 + \beta_{M_2} \cdot M_2 + \beta_{\mathrm{age}} \cdot \mathrm{age} + \beta_{\mathrm{BMI}} \cdot \mathrm{BMI} + \gamma_{\mathrm{sex}} + \varepsilon$ (linear regression; $M_2$ is seated diastolic blood pressure and $Y$ is carotid IMT, both on the original scale). $\beta_{M_2}$ is the difference in carotid IMT per 1 original unit higher seated diastolic blood pressure with age, BMI and sex held fixed.

\textit{Intervals.} 95\% confidence interval.
{\footnotesize\setlength{\tabcolsep}{3pt}
\begin{longtable}{@{}>{\raggedright\arraybackslash}p{0.14\linewidth}>{\raggedright\arraybackslash}p{0.29\linewidth}>{\raggedleft\arraybackslash}p{0.13\linewidth}>{\raggedright\arraybackslash}p{0.22\linewidth}>{\raggedleft\arraybackslash}p{0.14\linewidth}@{}}
\toprule
Block & Parameter & Estimate & Interval & $P$ \\
\midrule\endhead
$\pm$365-day window & $\beta_{M_2}$: adjusted seated diastolic blood pressure--carotid IMT association coefficient & $0.0007998$ & $[0.0005816$, $0.001018]$ & \mbox{$6.655\times10^{-13}$} \\
\bottomrule
\end{longtable}}

\paragraph{15.\ Pairwise association: GlycA--seated diastolic blood pressure ($\pm$365 days)}
\textit{Design.} GlycA and seated diastolic blood pressure measured in the same period: each blood pressure measurement date serves as the anchor, with the closest GlycA measurement within 365 days before or after it (inclusive); the same participant may contribute multiple anchor dates. Diastolic blood pressure taken as the mean of the available values of the two seated readings. Adjustment for approximate age (blood pressure measurement year minus birth year), the closest BMI within $\pm$365 days of the anchor date, and sex; continuous variables kept on the original scale.

\textit{Model.} $M_2 = \beta_0 + \beta_{M_1} \cdot M_1 + \beta_{\mathrm{age}} \cdot \mathrm{age} + \beta_{\mathrm{BMI}} \cdot \mathrm{BMI} + \gamma_{\mathrm{sex}} + \varepsilon$ (linear regression; $M_1$ is GlycA and $M_2$ is seated diastolic blood pressure, both on the original scale). $\beta_{M_1}$ is the difference in seated diastolic blood pressure per 1 original unit higher GlycA with age, BMI and sex held fixed.

\textit{Intervals.} 95\% confidence interval.
{\footnotesize\setlength{\tabcolsep}{3pt}
\begin{longtable}{@{}>{\raggedright\arraybackslash}p{0.14\linewidth}>{\raggedright\arraybackslash}p{0.29\linewidth}>{\raggedleft\arraybackslash}p{0.13\linewidth}>{\raggedright\arraybackslash}p{0.22\linewidth}>{\raggedleft\arraybackslash}p{0.14\linewidth}@{}}
\toprule
Block & Parameter & Estimate & Interval & $P$ \\
\midrule\endhead
$\pm$365-day window & $\beta_{M_1}$: adjusted GlycA--seated diastolic blood pressure association coefficient & $5.827$ & $[2.3$, $9.354]$ & \mbox{$0.001203$} \\
\bottomrule
\end{longtable}}

\subsubsection*{Joint models}

\paragraph{16.\ Joint model: grip strength, GlycA, diastolic blood pressure and IMT ($\pm$365 days)}
\textit{Design.} Grip strength, GlycA, seated diastolic blood pressure and carotid IMT measured in the same period: each IMT measurement date serves as the anchor, with grip strength, GlycA and diastolic blood pressure each taken as the closest measurement within 365 days before or after the anchor (inclusive); the same participant may contribute multiple anchor dates, $n = 804$. The base and joint models use the same sample, both adjusted for approximate age (IMT measurement year minus birth year), the closest BMI within $\pm$365 days of the anchor date, and sex; continuous variables kept on the original scale.

\textit{Model.} Two linear regressions. Base model: $Y = \alpha_0 + \beta_A^{\mathrm{base}} \cdot A + \alpha_{\mathrm{age}} \cdot \mathrm{age} + \alpha_{\mathrm{BMI}} \cdot \mathrm{BMI} + \alpha_{\mathrm{sex}} + \varepsilon$; joint model: $Y = \beta_0 + \beta_A \cdot A + \beta_{M_1} \cdot M_1 + \beta_{M_2} \cdot M_2 + \gamma_{\mathrm{age}} \cdot \mathrm{age} + \gamma_{\mathrm{BMI}} \cdot \mathrm{BMI} + \gamma_{\mathrm{sex}} + \varepsilon$. $A$, $M_1$, $M_2$ and $Y$ are grip strength, GlycA, seated diastolic blood pressure and carotid IMT, respectively, all on the original scale. $\beta_A$, $\beta_{M_1}$ and $\beta_{M_2}$ are the differences in IMT per 1 original unit higher value of the respective variable with the other terms of the joint model held fixed; $\Delta_A = \beta_A^{\mathrm{base}} - \beta_A$ is the signed change in the grip strength coefficient on the same sample.

\textit{Intervals.} 95\% confidence intervals. $P$ values not adjusted for multiple comparisons.
{\footnotesize\setlength{\tabcolsep}{3pt}
\begin{longtable}{@{}>{\raggedright\arraybackslash}p{0.14\linewidth}>{\raggedright\arraybackslash}p{0.29\linewidth}>{\raggedleft\arraybackslash}p{0.13\linewidth}>{\raggedright\arraybackslash}p{0.22\linewidth}>{\raggedleft\arraybackslash}p{0.14\linewidth}@{}}
\toprule
Block & Parameter & Estimate & Interval & $P$ \\
\midrule\endhead
$\pm$365-day window; joint model & $\beta_A$: grip strength coefficient in the joint model & $-0.0001898$ & $[-0.001046$, $0.0006666]$ & \mbox{$0.6641$} \\
 & $\beta_{M_1}$: GlycA coefficient in the joint model & $-0.07773$ & $[-0.1249$, $-0.03055]$ & \mbox{$0.001243$} \\
 & $\beta_{M_2}$: seated diastolic blood pressure coefficient in the joint model & $0.0006717$ & $[4.816\times10^{-5}$, $0.001295]$ & \mbox{$0.03474$} \\
$\pm$365-day window; change in grip strength coefficient & $\Delta_A = \beta_A^{\mathrm{base}} - \beta_A$: difference between the base-model and joint-model grip strength coefficients & $0.0002664$ & $[8.864\times10^{-5}$, $0.0004442]$ & \mbox{$0.003313$} \\
\bottomrule
\end{longtable}}

\subsubsection*{Reverse direction}

\paragraph{17.\ Reversed measurement order: serial equations with IMT preceding diastolic blood pressure}
\textit{Design.} GlycA measured within 365 days before or after grip strength, diastolic blood pressure within 90 days before or after GlycA (same day allowed), and carotid IMT measured 30--1825 days before diastolic blood pressure (all bounds inclusive); one chain per participant, $n = 40$. All four equations share the same sample, with adjustment for approximate age on the grip strength measurement date, sex, recruitment cohort, and the most recent BMI on or within 730 days before the earliest measurement date in the chain; continuous variables standardized within the sample, with sex and cohort as categorical terms.

\textit{Model.} $M_1 = a_1 A + \gamma_1^\top C + \varepsilon_1$; $M_2 = a_2 A + d_{21} M_1 + \gamma_2^\top C + \varepsilon_2$; $Y = c' A + b_1 M_1 + b_2 M_2 + \gamma_3^\top C + \varepsilon_3$; $Y = c A + \gamma_T^\top C + \varepsilon_T$. All four are linear regressions (with intercept); $A$, $M_1$, $M_2$ and $Y$ are standardized grip strength (larger of left and right hand), GlycA, seated diastolic blood pressure (mean of two readings) and carotid IMT (mean of left and right sides), respectively, and $C$ denotes the covariates, with $Y$ taken as the IMT measured before diastolic blood pressure. $a_1$ is the SD change in GlycA per 1 SD higher grip strength; $d_{21}$ is the slope of diastolic blood pressure on GlycA adjusted for grip strength; $b_2$ is the slope of IMT on diastolic blood pressure adjusted for grip strength and GlycA; $c'$ is the slope of IMT on grip strength adjusted for both mediators; $c$ is the total slope of IMT on grip strength adjusted for $C$ only. Bias bound $= \sqrt{R^2_{Y\sim U} \cdot R^2_{D\sim U} / (1 - R^2_{D\sim U})} \cdot \hat{\sigma}_Y / \hat{\sigma}_D$, where $R^2_{D\sim U}$ and $R^2_{Y\sim U}$ are the partial $R^2$ values of a hypothetical omitted confounder $U$ with the coefficient's regressor $D$ and with the equation's outcome, $\hat{\sigma}_Y$ is the residual SD of the equation, and $\hat{\sigma}_D$ is the residual SD from regressing $D$ on the remaining terms of the equation; each of the two partial $R^2$ values takes 0.01, 0.05, 0.10, 0.20, 0.30 and 0.50.

\textit{Intervals.} $a_1$, $d_{21}$, $b_2$, $c'$, $c$: 95\% confidence intervals. $P$ values not adjusted for multiple comparisons. Bias-bound grid: lower and upper limits are the coefficient $\pm$ the bias bound.
{\footnotesize\setlength{\tabcolsep}{3pt}
\begin{longtable}{@{}>{\raggedright\arraybackslash}p{0.14\linewidth}>{\raggedright\arraybackslash}p{0.29\linewidth}>{\raggedleft\arraybackslash}p{0.13\linewidth}>{\raggedright\arraybackslash}p{0.22\linewidth}>{\raggedleft\arraybackslash}p{0.14\linewidth}@{}}
\toprule
Block & Parameter & Estimate & Interval & $P$ \\
\midrule\endhead
Grip strength $\to$ IMT association & $c'$: standardized slope for grip strength $\to$ carotid IMT adjusted for GlycA and diastolic blood pressure ($A \to Y \mid M_1, M_2$) & $-0.0704$ & $[-0.664$, $0.5232]$ & \mbox{$0.8108$} \\
 & $c$: adjusted total standardized slope for grip strength $\to$ carotid IMT ($A \to Y$) & $-0.05035$ & $[-0.5857$, $0.485]$ & \mbox{$0.8497$} \\
Path components & $a_1$: standardized slope for grip strength $\to$ GlycA ($A \to M_1$) & $-0.4205$ & $[-0.8959$, $0.05495]$ & \mbox{$0.08122$} \\
 & $d_{21}$: standardized slope for GlycA $\to$ seated diastolic blood pressure adjusted for grip strength ($M_1 \to M_2 \mid A$) & $0.0182$ & $[-0.4648$, $0.5012]$ & \mbox{$0.9394$} \\
 & $b_2$: standardized slope for seated diastolic blood pressure $\to$ carotid IMT adjusted for grip strength and GlycA ($M_2 \to Y \mid A, M_1$) & $-0.04171$ & $[-0.3699$, $0.2865]$ & \mbox{$0.7976$} \\
\bottomrule
\end{longtable}}

{\scriptsize\setlength{\tabcolsep}{3pt}
\begin{longtable}{@{}lccc@{}}
\toprule
Partial $R^2$ of U (regressor; outcome) & $a_1$ & $d_{21}$ & $b_2$ \\
\midrule\endhead
Coefficient & $-0.4205$ & $0.0182$ & $-0.04171$ \\
0.01; 0.01 & $[-0.4372, -0.4037]$ & $[0.007439, 0.02895]$ & $[-0.05108, -0.03234]$ \\
0.01; 0.05 & $[-0.4579, -0.383]$ & $[-0.005858, 0.04225]$ & $[-0.06266, -0.02076]$ \\
0.01; 0.10 & $[-0.4734, -0.3675]$ & $[-0.01582, 0.05221]$ & $[-0.07134, -0.01208]$ \\
0.01; 0.20 & $[-0.4953, -0.3456]$ & $[-0.02991, 0.0663]$ & $[-0.08362, 0.0001925]$ \\
0.01; 0.30 & $[-0.5122, -0.3288]$ & $[-0.04072, 0.07712]$ & $[-0.09303, 0.00961]$ \\
0.01; 0.50 & $[-0.5389, -0.3021]$ & $[-0.05787, 0.09426]$ & $[-0.108, 0.02454]$ \\
0.05; 0.01 & $[-0.4587, -0.3823]$ & $[-0.00636, 0.04275]$ & $[-0.0631, -0.02032]$ \\
\bottomrule
\end{longtable}}

\paragraph{18.\ Reversed time order: IMT and grip strength measured 366--730 days later}
\textit{Design.} Carotid IMT and grip strength measured in strict date order: each IMT measurement date serves as the anchor, with the closest grip strength measurement 366--730 days after it (inclusive); the same participant may contribute multiple anchor dates. Grip strength and IMT defined as in the main association. Adjustment for approximate age (IMT measurement year minus birth year), the closest BMI within $\pm$365 days of the anchor date, and sex; continuous variables kept on the original scale.

\textit{Model.} $Y = \beta_0 + \beta_A \cdot A' + \beta_{\mathrm{age}} \cdot \mathrm{age} + \beta_{\mathrm{BMI}} \cdot \mathrm{BMI} + \gamma_{\mathrm{sex}} + \varepsilon$ (linear regression; $A'$ is grip strength measured 366--730 days after IMT and $Y$ is carotid IMT, both on the original scale). $\beta_A$ is the difference in carotid IMT per 1 original unit higher subsequently measured grip strength with age, BMI and sex held fixed.

\textit{Intervals.} 95\% confidence interval.
{\footnotesize\setlength{\tabcolsep}{3pt}
\begin{longtable}{@{}>{\raggedright\arraybackslash}p{0.14\linewidth}>{\raggedright\arraybackslash}p{0.29\linewidth}>{\raggedleft\arraybackslash}p{0.13\linewidth}>{\raggedright\arraybackslash}p{0.22\linewidth}>{\raggedleft\arraybackslash}p{0.14\linewidth}@{}}
\toprule
Block & Parameter & Estimate & Interval & $P$ \\
\midrule\endhead
Grip strength 366--730 days after IMT & $\beta_A$: adjusted association coefficient of IMT with grip strength measured 366--730 days later & $0.0006266$ & $[-0.0007474$, $0.002001]$ & \mbox{$0.3714$} \\
\bottomrule
\end{longtable}}

\subsubsection*{Heterogeneity and interaction}

\paragraph{19.\ Heterogeneity by sex: grip strength--carotid IMT slope ($\pm$365 days)}
\textit{Design.} Grip strength and carotid IMT measured in the same period: each IMT measurement date serves as the anchor, with the closest grip strength measurement within 365 days before or after it (inclusive); the same participant may contribute multiple anchor dates. Sex (first and second sex categories) entered as a main effect and as an interaction with grip strength; additional adjustment for approximate age (IMT measurement year minus birth year) and the closest BMI within $\pm$365 days of the anchor date; continuous variables kept on the original scale.

\textit{Model.} $Y = \sum_s \mathbf{1}(\mathrm{sex} = s) \cdot (\beta_{0,s} + \beta_{A,s} \cdot A) + \beta_{\mathrm{age}} \cdot \mathrm{age} + \beta_{\mathrm{BMI}} \cdot \mathrm{BMI} + \varepsilon$ (linear regression with a grip strength $\times$ sex interaction; $A$ is grip strength and $Y$ is carotid IMT, both on the original scale). $\beta_{A,s}$ is the difference in carotid IMT per 1 original unit higher grip strength within sex category $s$ ($s = 1, 2$ for the first and second sex categories, respectively) with age and BMI held fixed.

\textit{Intervals.} 95\% confidence intervals. $P$ values not adjusted for multiple comparisons.
{\footnotesize\setlength{\tabcolsep}{3pt}
\begin{longtable}{@{}>{\raggedright\arraybackslash}p{0.14\linewidth}>{\raggedright\arraybackslash}p{0.29\linewidth}>{\raggedleft\arraybackslash}p{0.13\linewidth}>{\raggedright\arraybackslash}p{0.22\linewidth}>{\raggedleft\arraybackslash}p{0.14\linewidth}@{}}
\toprule
Block & Parameter & Estimate & Interval & $P$ \\
\midrule\endhead
First sex category & $\beta_{A,1}$: adjusted grip strength--carotid IMT slope within the first sex category & $0.0009764$ & $[0.0004786$, $0.001474]$ & \mbox{$0.0001208$} \\
Second sex category & $\beta_{A,2}$: adjusted grip strength--carotid IMT slope within the second sex category & $0.0004249$ & $[8.285\times10^{-5}$, $0.000767]$ & \mbox{$0.0149$} \\
\bottomrule
\end{longtable}}

\subsubsection*{Sensitivity analyses}

\paragraph{20.\ Narrow-window serial path: grip strength $\to$ GlycA $\to$ diastolic blood pressure $\to$ IMT}
\textit{Design.} GlycA measured within 180 days before or after grip strength, diastolic blood pressure within 30 days before or after GlycA (same day allowed), and carotid IMT measured 30--1095 days after diastolic blood pressure (all bounds inclusive); one chain per participant, $n = 444$. All four equations share the same sample, with adjustment for approximate age on the grip strength measurement date, sex, recruitment cohort, and the most recent BMI on or within 730 days before the earliest measurement date in the chain; continuous variables standardized within the sample, with sex and cohort as categorical terms.

\textit{Model.} $M_1 = a_1 A + \gamma_1^\top C + \varepsilon_1$; $M_2 = a_2 A + d_{21} M_1 + \gamma_2^\top C + \varepsilon_2$; $Y = c' A + b_1 M_1 + b_2 M_2 + \gamma_3^\top C + \varepsilon_3$; $Y = c A + \gamma_T^\top C + \varepsilon_T$. All four are linear regressions (with intercept); $A$, $M_1$, $M_2$ and $Y$ are standardized grip strength (larger of left and right hand), GlycA, seated diastolic blood pressure (mean of two readings) and carotid IMT (mean of left and right sides), respectively, and $C$ denotes the covariates. $a_1$ is the SD change in GlycA per 1 SD higher grip strength; $d_{21}$ is the slope of diastolic blood pressure on GlycA adjusted for grip strength; $b_2$ is the slope of IMT on diastolic blood pressure adjusted for grip strength and GlycA; $c'$ is the slope of IMT on grip strength adjusted for both mediators; $c$ is the total slope of IMT on grip strength adjusted for $C$ only; the serial path product is $\theta = a_1 \cdot d_{21} \cdot b_2$.

\textit{Intervals.} $a_1$, $d_{21}$, $b_2$, $c'$, $c$: 95\% confidence intervals; $\theta$: bootstrap 95\% percentile interval. $P$ values not adjusted for multiple comparisons.
{\footnotesize\setlength{\tabcolsep}{3pt}
\begin{longtable}{@{}>{\raggedright\arraybackslash}p{0.14\linewidth}>{\raggedright\arraybackslash}p{0.29\linewidth}>{\raggedleft\arraybackslash}p{0.13\linewidth}>{\raggedright\arraybackslash}p{0.22\linewidth}>{\raggedleft\arraybackslash}p{0.14\linewidth}@{}}
\toprule
Block & Parameter & Estimate & Interval & $P$ \\
\midrule\endhead
Serial path index & $\theta = a_1 \cdot d_{21} \cdot b_2$: serial path product association index & $-0.002228$ & $[-0.006997$, $0.0004698]$ &  \\
Grip strength $\to$ IMT association & $c'$: standardized slope for grip strength $\to$ carotid IMT adjusted for GlycA and diastolic blood pressure ($A \to Y \mid M_1, M_2$) & $0.003259$ & $[-0.1522$, $0.1587]$ & \mbox{$0.9672$} \\
 & $c$: adjusted total standardized slope for grip strength $\to$ carotid IMT ($A \to Y$) & $0.02021$ & $[-0.1368$, $0.1772]$ & \mbox{$0.8004$} \\
Path components & $a_1$: standardized slope for grip strength $\to$ GlycA ($A \to M_1$) & $-0.1187$ & $[-0.2728$, $0.03546]$ & \mbox{$0.1309$} \\
 & $d_{21}$: standardized slope for GlycA $\to$ seated diastolic blood pressure adjusted for grip strength ($M_1 \to M_2 \mid A$) & $0.1628$ & $[0.06057$, $0.265]$ & \mbox{$0.001866$} \\
 & $b_2$: standardized slope for seated diastolic blood pressure $\to$ carotid IMT adjusted for grip strength and GlycA ($M_2 \to Y \mid A, M_1$) & $0.1153$ & $[0.03365$, $0.197]$ & \mbox{$0.005758$} \\
\bottomrule
\end{longtable}}

\paragraph{21.\ Additional adjustment for education and smoking: grip strength $\to$ GlycA $\to$ diastolic blood pressure $\to$ IMT}
\textit{Design.} GlycA measured within 365 days before or after grip strength, diastolic blood pressure within 90 days before or after GlycA (same day allowed), and IMT measured 30--1825 days after diastolic blood pressure (all bounds inclusive); one chain per participant, $n = 203$. Adjustment for approximate age on the grip strength measurement date, sex, recruitment cohort, the most recent BMI and smoking status on or within 730 days before the earliest measurement date in the chain, and the most recent educational level within 3650 days; continuous variables standardized within the sample, with the remaining covariates as categorical terms.

\textit{Model.} $M_1 = a_1 A + \gamma_1^\top C + \varepsilon_1$; $M_2 = a_2 A + d_{21} M_1 + \gamma_2^\top C + \varepsilon_2$; $Y = c' A + b_1 M_1 + b_2 M_2 + \gamma_3^\top C + \varepsilon_3$; $Y = c A + \gamma_T^\top C + \varepsilon_T$. All four are linear regressions (with intercept); $A$, $M_1$, $M_2$ and $Y$ are standardized grip strength (larger of left and right hand), GlycA, seated diastolic blood pressure (mean of two readings) and carotid IMT (mean of left and right sides), respectively, and $C$ denotes the covariates (including educational level and smoking status). $a_1$ is the SD change in GlycA per 1 SD higher grip strength; $d_{21}$ is the slope of diastolic blood pressure on GlycA adjusted for grip strength; $b_2$ is the slope of IMT on diastolic blood pressure adjusted for grip strength and GlycA; $c$ is the total slope of IMT on grip strength adjusted for $C$ only. Bias bound $= \sqrt{R^2_{Y\sim U} \cdot R^2_{D\sim U} / (1 - R^2_{D\sim U})} \cdot \hat{\sigma}_Y / \hat{\sigma}_D$, where $R^2_{D\sim U}$ and $R^2_{Y\sim U}$ are the partial $R^2$ values of a hypothetical omitted confounder $U$ with the coefficient's regressor $D$ and with the equation's outcome, $\hat{\sigma}_Y$ is the residual SD of the equation, and $\hat{\sigma}_D$ is the residual SD from regressing $D$ on the remaining terms of the equation; each of the two partial $R^2$ values takes 0.01, 0.05, 0.10, 0.20, 0.30 and 0.50.

\textit{Intervals.} $a_1$, $c$: 95\% confidence intervals. $P$ values not adjusted for multiple comparisons. Bias-bound grid: lower and upper limits are the coefficient $\pm$ the bias bound.
{\footnotesize\setlength{\tabcolsep}{3pt}
\begin{longtable}{@{}>{\raggedright\arraybackslash}p{0.14\linewidth}>{\raggedright\arraybackslash}p{0.29\linewidth}>{\raggedleft\arraybackslash}p{0.13\linewidth}>{\raggedright\arraybackslash}p{0.22\linewidth}>{\raggedleft\arraybackslash}p{0.14\linewidth}@{}}
\toprule
Block & Parameter & Estimate & Interval & $P$ \\
\midrule\endhead
Grip strength $\to$ IMT association & $c$: adjusted total standardized slope for grip strength $\to$ carotid IMT ($A \to Y$) & $-0.04095$ & $[-0.2532$, $0.1713]$ & \mbox{$0.7039$} \\
Path components & $a_1$: standardized slope for grip strength $\to$ GlycA ($A \to M_1$) & $-0.2222$ & $[-0.4565$, $0.01214]$ & \mbox{$0.06297$} \\
\bottomrule
\end{longtable}}

{\scriptsize\setlength{\tabcolsep}{3pt}
\begin{longtable}{@{}lccc@{}}
\toprule
Partial $R^2$ of U (regressor; outcome) & $a_1$ & $d_{21}$ & $b_2$ \\
\midrule\endhead
Coefficient & $-0.2222$ & $0.08964$ & $0.03483$ \\
0.01; 0.01 & $[-0.2387, -0.2057]$ & $[0.07953, 0.09975]$ & $[0.02558, 0.04408]$ \\
0.01; 0.05 & $[-0.2591, -0.1853]$ & $[0.06704, 0.1123]$ & $[0.01414, 0.05551]$ \\
0.01; 0.10 & $[-0.2743, -0.17]$ & $[0.05767, 0.1216]$ & $[0.005575, 0.06408]$ \\
0.01; 0.20 & $[-0.2959, -0.1484]$ & $[0.04443, 0.1349]$ & $[-0.006542, 0.0762]$ \\
0.01; 0.30 & $[-0.3125, -0.1318]$ & $[0.03427, 0.145]$ & $[-0.01584, 0.08549]$ \\
0.01; 0.50 & $[-0.3388, -0.1055]$ & $[0.01816, 0.1611]$ & $[-0.03058, 0.1002]$ \\
0.05; 0.01 & $[-0.2598, -0.1845]$ & $[0.06657, 0.1127]$ & $[0.01371, 0.05594]$ \\
\bottomrule
\end{longtable}}

\paragraph{22.\ Narrow window: grip strength--carotid IMT ($\pm$90 days)}
\textit{Design.} Grip strength and carotid IMT measured in the same period: each IMT measurement date serves as the anchor, with the closest grip strength measurement within 90 days before or after it (inclusive); the same participant may contribute multiple anchor dates. Variable definitions and adjustment as in the main $\pm$365-day association: approximate age, the closest BMI within $\pm$365 days of the anchor date, and sex; continuous variables kept on the original scale.

\textit{Model.} As in the main $\pm$365-day association: $Y = \beta_0 + \beta_A \cdot A + \beta_{\mathrm{age}} \cdot \mathrm{age} + \beta_{\mathrm{BMI}} \cdot \mathrm{BMI} + \gamma_{\mathrm{sex}} + \varepsilon$ (linear regression; $A$ is grip strength and $Y$ is carotid IMT, both on the original scale). $\beta_A$ is the difference in carotid IMT per 1 original unit higher grip strength with age, BMI and sex held fixed.

\textit{Intervals.} 95\% confidence interval.
{\footnotesize\setlength{\tabcolsep}{3pt}
\begin{longtable}{@{}>{\raggedright\arraybackslash}p{0.14\linewidth}>{\raggedright\arraybackslash}p{0.29\linewidth}>{\raggedleft\arraybackslash}p{0.13\linewidth}>{\raggedright\arraybackslash}p{0.22\linewidth}>{\raggedleft\arraybackslash}p{0.14\linewidth}@{}}
\toprule
Block & Parameter & Estimate & Interval & $P$ \\
\midrule\endhead
$\pm$90-day window & $\beta_A$: adjusted grip strength--carotid IMT association coefficient & $0.0005631$ & $[0.0002759$, $0.0008502]$ & \mbox{$0.0001216$} \\
\bottomrule
\end{longtable}}

\paragraph{23.\ Additional adjustment for smoking status: grip strength--carotid IMT ($\pm$365 days)}
\textit{Design.} Grip strength and carotid IMT measured in the same period: each IMT measurement date serves as the anchor, with the closest grip strength measurement within 365 days before or after it (inclusive); the same participant may contribute multiple anchor dates. In addition to approximate age, the closest BMI within $\pm$365 days of the anchor date, and sex, adjustment for the closest smoking status record within $\pm$730 days of the anchor date (categorical term); continuous variables kept on the original scale.

\textit{Model.} $Y = \beta_0 + \beta_A \cdot A + \beta_{\mathrm{age}} \cdot \mathrm{age} + \beta_{\mathrm{BMI}} \cdot \mathrm{BMI} + \gamma_{\mathrm{sex}} + \gamma_{\mathrm{smk}} + \varepsilon$ (linear regression; $A$ is grip strength and $Y$ is carotid IMT, both on the original scale; $\gamma_{\mathrm{smk}}$ denotes the effects of the smoking status categories). $\beta_A$ is the difference in carotid IMT per 1 original unit higher grip strength with age, BMI, sex and smoking status held fixed.

\textit{Intervals.} 95\% confidence interval.
{\footnotesize\setlength{\tabcolsep}{3pt}
\begin{longtable}{@{}>{\raggedright\arraybackslash}p{0.14\linewidth}>{\raggedright\arraybackslash}p{0.29\linewidth}>{\raggedleft\arraybackslash}p{0.13\linewidth}>{\raggedright\arraybackslash}p{0.22\linewidth}>{\raggedleft\arraybackslash}p{0.14\linewidth}@{}}
\toprule
Block & Parameter & Estimate & Interval & $P$ \\
\midrule\endhead
$\pm$365-day window; additional adjustment for smoking status & $\beta_A$: grip strength--carotid IMT association coefficient with additional adjustment for smoking status & $0.0005596$ & $[0.0002452$, $0.000874]$ & \mbox{$0.0004865$} \\
\bottomrule
\end{longtable}}

\paragraph{24.\ Joint model: grip strength, GlycA, diastolic blood pressure and IMT (tightened windows)}
\textit{Design.} Grip strength, GlycA, seated diastolic blood pressure and carotid IMT measured in the same period: each IMT measurement date serves as the anchor, with grip strength and GlycA each taken as the closest measurement within 180 days before or after the anchor and diastolic blood pressure within 30 days before or after it (all bounds inclusive); the same participant may contribute multiple anchor dates. The base and joint models use the same sample, both adjusted for approximate age (IMT measurement year minus birth year), the closest BMI within $\pm$365 days of the anchor date, and sex; continuous variables kept on the original scale.

\textit{Model.} Two linear regressions, of the same form as the $\pm$365-day joint model. Base model: $Y = \alpha_0 + \beta_A^{\mathrm{base}} \cdot A + \alpha_{\mathrm{age}} \cdot \mathrm{age} + \alpha_{\mathrm{BMI}} \cdot \mathrm{BMI} + \alpha_{\mathrm{sex}} + \varepsilon$; joint model: $Y = \beta_0 + \beta_A \cdot A + \beta_{M_1} \cdot M_1 + \beta_{M_2} \cdot M_2 + \gamma_{\mathrm{age}} \cdot \mathrm{age} + \gamma_{\mathrm{BMI}} \cdot \mathrm{BMI} + \gamma_{\mathrm{sex}} + \varepsilon$. $A$, $M_1$, $M_2$ and $Y$ are grip strength, GlycA, seated diastolic blood pressure and carotid IMT, respectively, all on the original scale. $\beta_A$, $\beta_{M_1}$ and $\beta_{M_2}$ are the differences in IMT per 1 original unit higher value of the respective variable with the other terms of the joint model held fixed; $\Delta_A = \beta_A^{\mathrm{base}} - \beta_A$ is the signed change in the grip strength coefficient on the same sample.

\textit{Intervals.} 95\% confidence intervals. $P$ values not adjusted for multiple comparisons.
{\footnotesize\setlength{\tabcolsep}{3pt}
\begin{longtable}{@{}>{\raggedright\arraybackslash}p{0.14\linewidth}>{\raggedright\arraybackslash}p{0.29\linewidth}>{\raggedleft\arraybackslash}p{0.13\linewidth}>{\raggedright\arraybackslash}p{0.22\linewidth}>{\raggedleft\arraybackslash}p{0.14\linewidth}@{}}
\toprule
Block & Parameter & Estimate & Interval & $P$ \\
\midrule\endhead
Grip strength and GlycA $\pm$180 days, diastolic blood pressure $\pm$30 days; joint model & $\beta_A$: grip strength coefficient in the joint model & $-0.000188$ & $[-0.001045$, $0.0006687]$ & \mbox{$0.6672$} \\
 & $\beta_{M_1}$: GlycA coefficient in the joint model & $-0.07794$ & $[-0.1252$, $-0.03065]$ & \mbox{$0.001237$} \\
 & $\beta_{M_2}$: seated diastolic blood pressure coefficient in the joint model & $0.0006783$ & $[5.108\times10^{-5}$, $0.001306]$ & \mbox{$0.03404$} \\
Grip strength and GlycA $\pm$180 days, diastolic blood pressure $\pm$30 days; change in grip strength coefficient & $\Delta_A = \beta_A^{\mathrm{base}} - \beta_A$: difference between the base-model and joint-model grip strength coefficients & $0.0002659$ & $[8.788\times10^{-5}$, $0.000444]$ & \mbox{$0.003419$} \\
\bottomrule
\end{longtable}}

\paragraph{25.\ Joint model with additional adjustment for smoking status: grip strength, GlycA, diastolic blood pressure and IMT}
\textit{Design.} Grip strength, GlycA, seated diastolic blood pressure and carotid IMT measured in the same period: each IMT measurement date serves as the anchor, with the first three each taken as the closest measurement within 365 days before or after the anchor (inclusive); the same participant may contribute multiple anchor dates. The base and joint models use the same sample; in addition to approximate age, the closest BMI within $\pm$365 days of the anchor date, and sex, both adjust for the closest smoking status record within $\pm$730 days of the anchor date (categorical term); continuous variables kept on the original scale.

\textit{Model.} Two linear regressions. Base model: $Y = \alpha_0 + \beta_A^{\mathrm{base}} \cdot A + \alpha_{\mathrm{age}} \cdot \mathrm{age} + \alpha_{\mathrm{BMI}} \cdot \mathrm{BMI} + \alpha_{\mathrm{sex}} + \alpha_{\mathrm{smk}} + \varepsilon$; joint model: $Y = \beta_0 + \beta_A \cdot A + \beta_{M_1} \cdot M_1 + \beta_{M_2} \cdot M_2 + \gamma_{\mathrm{age}} \cdot \mathrm{age} + \gamma_{\mathrm{BMI}} \cdot \mathrm{BMI} + \gamma_{\mathrm{sex}} + \gamma_{\mathrm{smk}} + \varepsilon$. $A$, $M_1$, $M_2$ and $Y$ are grip strength, GlycA, seated diastolic blood pressure and carotid IMT, respectively, all on the original scale. $\alpha_{\mathrm{smk}}$ and $\gamma_{\mathrm{smk}}$ denote the effects of the smoking status categories. $\beta_A$, $\beta_{M_1}$ and $\beta_{M_2}$ are the differences in IMT per 1 original unit higher value of the respective variable with the other terms of the joint model held fixed; $\Delta_A = \beta_A^{\mathrm{base}} - \beta_A$ is the signed change in the grip strength coefficient on the same sample.

\textit{Intervals.} 95\% confidence intervals. $P$ values not adjusted for multiple comparisons.
{\footnotesize\setlength{\tabcolsep}{3pt}
\begin{longtable}{@{}>{\raggedright\arraybackslash}p{0.14\linewidth}>{\raggedright\arraybackslash}p{0.29\linewidth}>{\raggedleft\arraybackslash}p{0.13\linewidth}>{\raggedright\arraybackslash}p{0.22\linewidth}>{\raggedleft\arraybackslash}p{0.14\linewidth}@{}}
\toprule
Block & Parameter & Estimate & Interval & $P$ \\
\midrule\endhead
$\pm$365-day window; additional adjustment for smoking status; joint model & $\beta_A$: grip strength coefficient in the joint model & $-0.000144$ & $[-0.001056$, $0.0007678]$ & \mbox{$0.7569$} \\
 & $\beta_{M_1}$: GlycA coefficient in the joint model & $-0.0668$ & $[-0.117$, $-0.01665]$ & \mbox{$0.009037$} \\
 & $\beta_{M_2}$: seated diastolic blood pressure coefficient in the joint model & $0.0007493$ & $[6.73\times10^{-5}$, $0.001431]$ & \mbox{$0.03129$} \\
$\pm$365-day window; additional adjustment for smoking status; change in grip strength coefficient & $\Delta_A = \beta_A^{\mathrm{base}} - \beta_A$: difference between the base-model and joint-model grip strength coefficients & $0.0002514$ & $[6.337\times10^{-5}$, $0.0004395]$ & \mbox{$0.008784$} \\
\bottomrule
\end{longtable}}

\FloatBarrier

\subsection{Hip circumference--hepatic steatosis--DBP--PWV}
\label{app:rq-designs:d6c6}

\paragraph{Research question and measures.}
The research question concerns the association of hip circumference with lower-limb arterial stiffness via hepatic steatosis and diastolic blood pressure, with data from the HPP cohort. The exposure is hip circumference; the mediators are the hepatic ultrasound attenuation coefficient (a continuous proxy for hepatic steatosis) and seated diastolic blood pressure (DBP); the outcome is thigh-to-ankle pulse wave velocity (PWV). Unless otherwise stated, coefficients are SD changes per 1 SD after standardization within the analysis sample of each unit.

\subsubsection*{Primary pathways}

\paragraph{1.\ Single-mediator path: hip circumference $\to$ hepatic ultrasound attenuation $\to$ thigh-to-ankle PWV}
\textit{Design.} Hip circumference, hepatic ultrasound and PWV measured in strict date order: ultrasound 30--730 days after hip circumference and PWV 30--730 days after ultrasound (endpoints inclusive); one measurement chain per participant, $n = 43$. Adjustment for approximate age at hip circumference measurement, sex and height measured on the same day as hip circumference. PWV as the mean of the available sides. Hip circumference, hepatic attenuation, PWV, age and height standardized within the analysis sample; sex as a categorical term.

\textit{Model.} $M = \alpha_0 + \alpha_A \cdot A + \alpha_C^{\top} \cdot C + \varepsilon_M$; $Y = \beta_0 + \beta_A \cdot A + \beta_M \cdot M + \beta_C^{\top} \cdot C + \varepsilon_Y$; $Y = \gamma_0 + \gamma_A \cdot A + \gamma_C^{\top} \cdot C + \varepsilon_T$. All three equations are linear regressions; $A$, $M$ and $Y$ are standardized hip circumference, hepatic ultrasound attenuation coefficient and thigh-to-ankle PWV, and $C$ the covariates. $\alpha_A$ is the SD change in hepatic attenuation per 1 SD higher hip circumference; $\beta_M$ and $\beta_A$ are the standardized slopes of hepatic attenuation and of hip circumference on PWV, each adjusted for the other; $\gamma_A$ is the total standardized slope of hip circumference on PWV adjusted only for $C$; path product $\theta = \alpha_A \cdot \beta_M$.

\textit{Intervals.} $\alpha_A$, $\beta_M$, $\beta_A$, $\gamma_A$: 97.5\% confidence intervals; $\theta$: bootstrap 97.5\% percentile interval.
{\footnotesize\setlength{\tabcolsep}{3pt}
\begin{longtable}{@{}>{\raggedright\arraybackslash}p{0.15\linewidth}>{\raggedright\arraybackslash}p{0.40\linewidth}>{\raggedleft\arraybackslash}p{0.14\linewidth}>{\raggedright\arraybackslash}p{0.24\linewidth}@{}}
\toprule
Block & Parameter & Estimate & Interval \\
\midrule\endhead
Path components (97.5\% confidence intervals) & $\alpha_A$: standardized slope, hip circumference $\to$ hepatic ultrasound attenuation ($A \to M$) & $0.4586$ & $[-0.0008323$, $0.918]$ \\
 & $\beta_M$: standardized slope, hepatic ultrasound attenuation $\to$ thigh-to-ankle PWV, adjusted for hip circumference ($M \to Y \mid A$) & $0.1275$ & $[-0.2714$, $0.5265]$ \\
 & $\beta_A$: standardized slope, hip circumference $\to$ PWV, adjusted for hepatic attenuation ($A \to Y \mid M$) & $0.07654$ & $[-0.305$, $0.4581]$ \\
 & $\gamma_A$: adjusted total standardized slope, hip circumference $\to$ PWV ($A \to Y$) & $0.135$ & $[-0.265$, $0.5351]$ \\
Path index (bootstrap 97.5\% percentile interval) & $\theta = \alpha_A \cdot \beta_M$: path-product association index & $0.05848$ & $[-0.09385$, $0.2616]$ \\
\bottomrule
\end{longtable}}

\paragraph{2.\ Serial path coefficients and bias bounds: hip circumference $\to$ hepatic attenuation $\to$ seated DBP $\to$ PWV}
\textit{Design.} Hip circumference, hepatic ultrasound, seated DBP and PWV measured in date order (same day allowed), with adjacent gaps of 0--180 days and a total span of 0--365 days (endpoints inclusive); one chain per participant. The four equations share the same sample and covariates: approximate age at hip circumference measurement, sex, and the most recent height on or within 730 days before the hip circumference measurement day. DBP as the mean of the available values of two seated readings; PWV as the mean of the available sides. Continuous variables standardized within the analysis sample; sex as a categorical term.

\textit{Model.} $M_1 = \alpha_1 + a_1 A + \gamma_1^{\top} C + \varepsilon_1$; $M_2 = \alpha_2 + a_2 A + d_{21} M_1 + \gamma_2^{\top} C + \varepsilon_2$; $Y = \alpha_3 + c' A + b_1 M_1 + b_2 M_2 + \gamma_3^{\top} C + \varepsilon_3$; $Y = \alpha_T + c A + \gamma_T^{\top} C + \varepsilon_T$. All four equations are linear regressions; $A$, $M_1$, $M_2$ and $Y$ are standardized hip circumference, hepatic ultrasound attenuation coefficient, seated DBP and thigh-to-ankle PWV, and $C$ the covariates. Reported: bias bounds $\hat{\beta} \pm \mathrm{se} \cdot \sqrt{\nu \cdot R^2_{Y\sim U} \cdot R^2_{D\sim U} / (1 - R^2_{D\sim U})}$ for coefficients $\hat{\beta} \in \{a_1, a_2, d_{21}, b_1, b_2, c', c\}$, where $\mathrm{se}$ is the conventional standard error of the coefficient, $\nu$ the residual degrees of freedom, and $R^2_{D\sim U}$ and $R^2_{Y\sim U}$ the partial $R^2$ of a hypothetical omitted covariate $U$ with the regressor of that coefficient and with the outcome of the equation, each set to 0.01, 0.05, 0.10 and 0.20.

\textit{Intervals.} Bias-bound grid: each row corresponds to one pair $(R^2_{D\sim U}, R^2_{Y\sim U})$, with lower and upper limits equal to the coefficient $\pm$ the bias bound.
{\scriptsize\setlength{\tabcolsep}{3pt}
\begin{longtable}{@{}lcccc@{}}
\toprule
Partial $R^2$ of U (regressor; outcome) & $a_1$ & $a_2$ & $d_{21}$ & $b_1$ \\
\midrule\endhead
Coefficient & $0.2035$ & $0.2633$ & $0.08416$ & $0.04859$ \\
0.01; 0.01 & $[0.1934, 0.2135]$ & $[0.2535, 0.2731]$ & $[0.07455, 0.09378]$ & $[0.04011, 0.05707]$ \\
0.01; 0.05 & $[0.181, 0.2259]$ & $[0.2414, 0.2852]$ & $[0.06267, 0.1057]$ & $[0.02962, 0.06756]$ \\
0.01; 0.10 & $[0.1717, 0.2352]$ & $[0.2323, 0.2943]$ & $[0.05376, 0.1146]$ & $[0.02177, 0.07541]$ \\
0.01; 0.20 & $[0.1585, 0.2484]$ & $[0.2194, 0.3072]$ & $[0.04117, 0.1272]$ & $[0.01066, 0.08652]$ \\
0.05; 0.01 & $[0.1805, 0.2264]$ & $[0.2409, 0.2857]$ & $[0.06222, 0.1061]$ & $[0.02923, 0.06795]$ \\
0.05; 0.05 & $[0.1522, 0.2547]$ & $[0.2132, 0.3134]$ & $[0.03509, 0.1332]$ & $[0.005294, 0.09188]$ \\
0.05; 0.10 & $[0.1309, 0.276]$ & $[0.1925, 0.3341]$ & $[0.01476, 0.1536]$ & $[-0.01264, 0.1098]$ \\
0.05; 0.20 & $[0.1009, 0.306]$ & $[0.1632, 0.3634]$ & $[-0.01398, 0.1823]$ & $[-0.038, 0.1352]$ \\
0.10; 0.01 & $[0.1701, 0.2368]$ & $[0.2308, 0.2958]$ & $[0.05228, 0.1161]$ & $[0.02046, 0.07672]$ \\
0.10; 0.05 & $[0.1289, 0.278]$ & $[0.1906, 0.3361]$ & $[0.01286, 0.1555]$ & $[-0.01432, 0.1115]$ \\
0.10; 0.10 & $[0.09807, 0.3088]$ & $[0.1604, 0.3662]$ & $[-0.01667, 0.185]$ & $[-0.04037, 0.1376]$ \\
0.10; 0.20 & $[0.05442, 0.3525]$ & $[0.1178, 0.4088]$ & $[-0.05844, 0.2268]$ & $[-0.07722, 0.1744]$ \\
0.20; 0.01 & $[0.1535, 0.2534]$ & $[0.2145, 0.3121]$ & $[0.03633, 0.132]$ & $[0.006391, 0.09079]$ \\
0.20; 0.05 & $[0.09168, 0.3152]$ & $[0.1542, 0.3724]$ & $[-0.02279, 0.1911]$ & $[-0.04577, 0.1429]$ \\
0.20; 0.10 & $[0.04538, 0.3615]$ & $[0.109, 0.4176]$ & $[-0.06709, 0.2354]$ & $[-0.08485, 0.182]$ \\
0.20; 0.20 & $[-0.0201, 0.427]$ & $[0.04507, 0.4815]$ & $[-0.1297, 0.2981]$ & $[-0.1401, 0.2373]$ \\
\bottomrule
\end{longtable}}

{\scriptsize\setlength{\tabcolsep}{3pt}
\begin{longtable}{@{}lccc@{}}
\toprule
Partial $R^2$ of U (regressor; outcome) & $b_2$ & $c'$ & Total-association slope $c$ \\
\midrule\endhead
Coefficient & $0.4338$ & $-0.02049$ & $0.111$ \\
0.01; 0.01 & $[0.425, 0.4426]$ & $[-0.02942, -0.01156]$ & $[0.1016, 0.1205]$ \\
0.01; 0.05 & $[0.414, 0.4535]$ & $[-0.04045, -0.000524]$ & $[0.08989, 0.1322]$ \\
0.01; 0.10 & $[0.4059, 0.4617]$ & $[-0.04872, 0.007746]$ & $[0.08113, 0.141]$ \\
0.01; 0.20 & $[0.3943, 0.4733]$ & $[-0.06042, 0.01944]$ & $[0.06874, 0.1533]$ \\
0.05; 0.01 & $[0.4136, 0.454]$ & $[-0.04087, -0.000108]$ & $[0.08945, 0.1326]$ \\
0.05; 0.05 & $[0.3887, 0.4789]$ & $[-0.06606, 0.02508]$ & $[0.06276, 0.1593]$ \\
0.05; 0.10 & $[0.37, 0.4975]$ & $[-0.08494, 0.04396]$ & $[0.04276, 0.1793]$ \\
0.05; 0.20 & $[0.3436, 0.524]$ & $[-0.1116, 0.07066]$ & $[0.01448, 0.2076]$ \\
0.10; 0.01 & $[0.4045, 0.4631]$ & $[-0.0501, 0.009124]$ & $[0.07967, 0.1424]$ \\
0.10; 0.05 & $[0.3683, 0.4993]$ & $[-0.08671, 0.04573]$ & $[0.04089, 0.1812]$ \\
0.10; 0.10 & $[0.3412, 0.5264]$ & $[-0.1141, 0.07316]$ & $[0.01184, 0.2103]$ \\
0.10; 0.20 & $[0.3028, 0.5648]$ & $[-0.1529, 0.1119]$ & $[-0.02926, 0.2513]$ \\
0.20; 0.01 & $[0.3899, 0.4777]$ & $[-0.06491, 0.02393]$ & $[0.06399, 0.1581]$ \\
0.20; 0.05 & $[0.3355, 0.532]$ & $[-0.1198, 0.07884]$ & $[0.005818, 0.2163]$ \\
0.20; 0.10 & $[0.2948, 0.5727]$ & $[-0.161, 0.12]$ & $[-0.03777, 0.2599]$ \\
0.20; 0.20 & $[0.2373, 0.6303]$ & $[-0.2191, 0.1782]$ & $[-0.09941, 0.3215]$ \\
\bottomrule
\end{longtable}}

\paragraph{3.\ Exposure--outcome association: hip circumference and bilateral thigh-to-ankle PWV}
\textit{Design.} Hip circumference and PWV measured in the same period (PWV within 365 days before or after hip circumference, endpoints inclusive); the pair with the shortest gap per participant, $n = 8956$. PWV as the bilateral mean when both left and right sides are valid. Adjustment for approximate age (computed at the later date of the pair) and sex. Hip circumference, PWV and age standardized within the analysis sample; sex as a categorical term.

\textit{Model.} $z(Y) = \beta_0 + \beta_A \cdot z(A) + \beta_{\mathrm{age}} \cdot z(\mathrm{age}) + \gamma_{\mathrm{sex}} + \varepsilon$, a linear regression, with $A$ hip circumference and $Y$ bilateral thigh-to-ankle PWV. $\beta_A$ is the difference in PWV, in SD, per 1 SD higher hip circumference at fixed age and sex.

\textit{Intervals.} $\beta_A$: 95\% confidence interval.
{\footnotesize\setlength{\tabcolsep}{3pt}
\begin{longtable}{@{}>{\raggedright\arraybackslash}p{0.43\linewidth}>{\raggedleft\arraybackslash}p{0.13\linewidth}>{\raggedright\arraybackslash}p{0.22\linewidth}>{\raggedleft\arraybackslash}p{0.14\linewidth}@{}}
\toprule
Parameter & Estimate & Interval & $P$ \\
\midrule\endhead
Hip circumference--PWV standardized association coefficient $\beta_A$ & $0.1314$ & $[0.1124$, $0.1504]$ & \mbox{$2.091\times10^{-41}$} \\
\bottomrule
\end{longtable}}

\subsubsection*{Adjacent segments and pairwise associations}

\paragraph{4.\ Serial segment: hip circumference $\to$ hepatic ultrasound attenuation $\to$ seated DBP}
\textit{Design.} Hip circumference, hepatic ultrasound and seated blood pressure measured in strict date order: hepatic ultrasound 30--730 days after hip circumference measurement and seated blood pressure 30--365 days after hepatic ultrasound (endpoints inclusive in both); one chain per participant, $n = 19$. Adjustment for approximate age on the hip circumference measurement day, recorded sex and height measured on the same day; hip circumference, hepatic attenuation, DBP, age and height standardized within the analysis sample.

\textit{Model.} $M_z = \alpha_0 + \alpha_A \cdot A_z + \alpha_C \cdot C + \varepsilon_M$; $Y_z = \beta_0 + \beta_A \cdot A_z + \beta_M \cdot M_z + \beta_C \cdot C + \varepsilon_Y$; $Y_z = \gamma_0 + \gamma_A \cdot A_z + \gamma_C \cdot C + \varepsilon_T$ (linear regressions; $A$ hip circumference, $M$ hepatic ultrasound attenuation, $Y$ seated DBP, $C$ age, sex and height, subscript $z$ denoting standardization). $\alpha_A$ is the hip circumference $\to$ hepatic attenuation slope, $\beta_M$ the hepatic attenuation $\to$ DBP slope adjusted for hip circumference, $\beta_A$ the hip circumference $\to$ DBP slope adjusted for hepatic attenuation, $\gamma_A$ the total hip circumference $\to$ DBP slope, and $\theta = \alpha_A \cdot \beta_M$ the segment path index.

\textit{Intervals.} $\alpha_A$, $\beta_M$, $\beta_A$, $\gamma_A$: 95\% confidence intervals; $\theta$: bootstrap 95\% percentile interval.
{\footnotesize\setlength{\tabcolsep}{3pt}
\begin{longtable}{@{}>{\raggedright\arraybackslash}p{0.15\linewidth}>{\raggedright\arraybackslash}p{0.40\linewidth}>{\raggedleft\arraybackslash}p{0.14\linewidth}>{\raggedright\arraybackslash}p{0.24\linewidth}@{}}
\toprule
Block & Parameter & Estimate & Interval \\
\midrule\endhead
Path components (95\% confidence intervals) & $\alpha_A$: standardized slope, hip circumference $\to$ hepatic ultrasound attenuation ($A \to M$) & $0.4136$ & $[-0.3709$, $1.198]$ \\
 & $\beta_M$: standardized slope, hepatic ultrasound attenuation $\to$ seated DBP, adjusted for hip circumference ($M \to Y \mid A$) & $0.8824$ & $[0.1972$, $1.568]$ \\
 & $\beta_A$: standardized slope, hip circumference $\to$ seated DBP, adjusted for hepatic attenuation ($A \to Y \mid M$) & $-0.05954$ & $[-0.7601$, $0.641]$ \\
 & $\gamma_A$: adjusted total standardized slope, hip circumference $\to$ seated DBP ($A \to Y$) & $0.3054$ & $[-0.6448$, $1.256]$ \\
Path index (bootstrap 95\% percentile interval) & $\theta = \alpha_A \cdot \beta_M$: segment path-product association index & $0.3649$ & $[-0.1193$, $1.029]$ \\
\bottomrule
\end{longtable}}

\paragraph{5.\ Pairwise association: hip circumference $\to$ hepatic ultrasound attenuation (0--180 days)}
\textit{Design.} Hepatic ultrasound measured 0--180 days after hip circumference measurement (same day included); participants with both measurements and all covariates included, one pair of measurements per person. Adjustment for approximate age on the hip circumference measurement day, recorded sex and the most recent height on or within 730 days before that day; hip circumference, hepatic attenuation, age and height standardized within the analysis sample.

\textit{Model.} $M_{1,z} = \beta_0 + \beta \cdot A_z + \beta_{\mathrm{age}} \cdot \mathrm{Age}_z + \gamma_{\mathrm{Sex}} + \beta_h \cdot \mathrm{Height}_z + \varepsilon$ (linear regression; $A$ hip circumference, $M_1$ hepatic ultrasound attenuation). $\beta$ is the SD difference in hepatic ultrasound attenuation per 1 SD higher hip circumference.

\textit{Intervals.} Bootstrap 95\% percentile interval.
{\footnotesize\setlength{\tabcolsep}{3pt}
\begin{longtable}{@{}>{\raggedright\arraybackslash}p{0.15\linewidth}>{\raggedright\arraybackslash}p{0.40\linewidth}>{\raggedleft\arraybackslash}p{0.14\linewidth}>{\raggedright\arraybackslash}p{0.24\linewidth}@{}}
\toprule
Block & Parameter & Estimate & Interval \\
\midrule\endhead
Forward window 0--180 days & Adjusted standardized slope $\beta$ (hip circumference $\to$ hepatic ultrasound attenuation) & $0.2038$ & $[0.182$, $0.2263]$ \\
\bottomrule
\end{longtable}}

\paragraph{6.\ Pairwise association: hip circumference $\to$ seated DBP (0--180 days)}
\textit{Design.} Seated blood pressure measured 0--180 days after hip circumference measurement (same day included); participants with both measurements and all covariates included, one pair of measurements per person. Adjustment for approximate age on the hip circumference measurement day, recorded sex and the most recent height on or within 730 days before that day; hip circumference, DBP, age and height standardized within the analysis sample.

\textit{Model.} $M_{2,z} = \beta_0 + \beta \cdot A_z + \beta_{\mathrm{age}} \cdot \mathrm{Age}_z + \gamma_{\mathrm{Sex}} + \beta_h \cdot \mathrm{Height}_z + \varepsilon$ (linear regression; $A$ hip circumference, $M_2$ seated DBP). $\beta$ is the SD difference in seated DBP per 1 SD higher hip circumference.

\textit{Intervals.} Bootstrap 95\% percentile interval.
{\footnotesize\setlength{\tabcolsep}{3pt}
\begin{longtable}{@{}>{\raggedright\arraybackslash}p{0.15\linewidth}>{\raggedright\arraybackslash}p{0.40\linewidth}>{\raggedleft\arraybackslash}p{0.14\linewidth}>{\raggedright\arraybackslash}p{0.24\linewidth}@{}}
\toprule
Block & Parameter & Estimate & Interval \\
\midrule\endhead
Forward window 0--180 days & Adjusted standardized slope $\beta$ (hip circumference $\to$ seated DBP) & $0.2745$ & $[0.2562$, $0.2928]$ \\
\bottomrule
\end{longtable}}

\paragraph{7.\ Pairwise association: hepatic ultrasound attenuation $\to$ seated DBP (0--180 days)}
\textit{Design.} Seated blood pressure measured 0--180 days after hepatic ultrasound (same day included); participants with both measurements and all covariates included, one pair of measurements per person. Adjustment for approximate age on the hepatic ultrasound day, recorded sex and the most recent height on or within 730 days before that day; hepatic attenuation, DBP, age and height standardized within the analysis sample.

\textit{Model.} $M_{2,z} = \beta_0 + \beta \cdot M_{1,z} + \beta_{\mathrm{age}} \cdot \mathrm{Age}_z + \gamma_{\mathrm{Sex}} + \beta_h \cdot \mathrm{Height}_z + \varepsilon$ (linear regression; $M_1$ hepatic ultrasound attenuation, $M_2$ seated DBP). $\beta$ is the SD difference in seated DBP per 1 SD higher hepatic ultrasound attenuation.

\textit{Intervals.} Bootstrap 95\% percentile interval.
{\footnotesize\setlength{\tabcolsep}{3pt}
\begin{longtable}{@{}>{\raggedright\arraybackslash}p{0.15\linewidth}>{\raggedright\arraybackslash}p{0.40\linewidth}>{\raggedleft\arraybackslash}p{0.14\linewidth}>{\raggedright\arraybackslash}p{0.24\linewidth}@{}}
\toprule
Block & Parameter & Estimate & Interval \\
\midrule\endhead
Forward window 0--180 days & Adjusted standardized slope $\beta$ (hepatic ultrasound attenuation $\to$ seated DBP) & $0.1363$ & $[0.1161$, $0.1567]$ \\
\bottomrule
\end{longtable}}

\paragraph{8.\ Pairwise association: hepatic ultrasound attenuation $\to$ PWV (0--365 days)}
\textit{Design.} PWV measured 0--365 days after hepatic ultrasound (same day included); participants with both measurements and all covariates included, one pair of measurements per person. Adjustment for approximate age on the hepatic ultrasound day, recorded sex and the most recent height on or within 730 days before that day; hepatic attenuation, PWV, age and height standardized within the analysis sample.

\textit{Model.} $Y_z = \beta_0 + \beta \cdot M_{1,z} + \beta_{\mathrm{age}} \cdot \mathrm{Age}_z + \gamma_{\mathrm{Sex}} + \beta_h \cdot \mathrm{Height}_z + \varepsilon$ (linear regression; $M_1$ hepatic ultrasound attenuation, $Y$ thigh-to-ankle PWV). $\beta$ is the SD difference in PWV per 1 SD higher hepatic ultrasound attenuation.

\textit{Intervals.} Bootstrap 95\% percentile interval.
{\footnotesize\setlength{\tabcolsep}{3pt}
\begin{longtable}{@{}>{\raggedright\arraybackslash}p{0.15\linewidth}>{\raggedright\arraybackslash}p{0.40\linewidth}>{\raggedleft\arraybackslash}p{0.14\linewidth}>{\raggedright\arraybackslash}p{0.24\linewidth}@{}}
\toprule
Block & Parameter & Estimate & Interval \\
\midrule\endhead
Forward window 0--365 days & Adjusted standardized slope $\beta$ (hepatic ultrasound attenuation $\to$ thigh-to-ankle PWV) & $0.1046$ & $[0.08525$, $0.1258]$ \\
\bottomrule
\end{longtable}}

\paragraph{9.\ Pairwise association: seated DBP $\to$ PWV (0--180 days)}
\textit{Design.} PWV measured 0--180 days after seated blood pressure measurement (same day included); participants with both measurements and all covariates included, one pair of measurements per person. Adjustment for approximate age on the blood pressure measurement day, recorded sex and the most recent height on or within 730 days before that day; DBP, PWV, age and height standardized within the analysis sample.

\textit{Model.} $Y_z = \beta_0 + \beta \cdot M_{2,z} + \beta_{\mathrm{age}} \cdot \mathrm{Age}_z + \gamma_{\mathrm{Sex}} + \beta_h \cdot \mathrm{Height}_z + \varepsilon$ (linear regression; $M_2$ seated DBP, $Y$ thigh-to-ankle PWV). $\beta$ is the SD difference in PWV per 1 SD higher seated DBP.

\textit{Intervals.} Bootstrap 95\% percentile interval.
{\footnotesize\setlength{\tabcolsep}{3pt}
\begin{longtable}{@{}>{\raggedright\arraybackslash}p{0.15\linewidth}>{\raggedright\arraybackslash}p{0.40\linewidth}>{\raggedleft\arraybackslash}p{0.14\linewidth}>{\raggedright\arraybackslash}p{0.24\linewidth}@{}}
\toprule
Block & Parameter & Estimate & Interval \\
\midrule\endhead
Forward window 0--180 days & Adjusted standardized slope $\beta$ (seated DBP $\to$ thigh-to-ankle PWV) & $0.4419$ & $[0.4225$, $0.4598]$ \\
\bottomrule
\end{longtable}}

\paragraph{10.\ Pairwise total association: hip circumference $\to$ PWV (0--365 days)}
\textit{Design.} PWV measured 0--365 days after hip circumference measurement (same day included); participants with both measurements and all covariates included, one pair of measurements per person. Adjustment for approximate age on the hip circumference measurement day, recorded sex and the most recent height on or within 730 days before that day; hip circumference, PWV, age and height standardized within the analysis sample.

\textit{Model.} $Y_z = \beta_0 + \beta \cdot A_z + \beta_{\mathrm{age}} \cdot \mathrm{Age}_z + \gamma_{\mathrm{Sex}} + \beta_h \cdot \mathrm{Height}_z + \varepsilon$ (linear regression; $A$ hip circumference, $Y$ thigh-to-ankle PWV). $\beta$ is the SD difference in PWV per 1 SD higher hip circumference.

\textit{Intervals.} Bootstrap 95\% percentile interval.
{\footnotesize\setlength{\tabcolsep}{3pt}
\begin{longtable}{@{}>{\raggedright\arraybackslash}p{0.15\linewidth}>{\raggedright\arraybackslash}p{0.40\linewidth}>{\raggedleft\arraybackslash}p{0.14\linewidth}>{\raggedright\arraybackslash}p{0.24\linewidth}@{}}
\toprule
Block & Parameter & Estimate & Interval \\
\midrule\endhead
Forward window 0--365 days & Adjusted standardized total-association slope $\beta$ (hip circumference $\to$ thigh-to-ankle PWV) & $0.1012$ & $[0.08312$, $0.121]$ \\
\bottomrule
\end{longtable}}

\paragraph{11.\ Pairwise association: hip circumference--hepatic ultrasound attenuation ($\pm 365$ days)}
\textit{Design.} Hip circumference and hepatic ultrasound measured in the same period ($\pm 365$ days, endpoints inclusive); participants with both measurements and all covariates included, the pair closest in date per person. Adjustment for approximate age (at the later of the two measurement dates) and recorded sex; hip circumference, hepatic attenuation and age standardized within the analysis sample.

\textit{Model.} $M_{1,z} = \beta_0 + \beta_A \cdot A_z + \beta_{\mathrm{age}} \cdot \mathrm{Age}_z + \gamma_{\mathrm{Sex}} + \varepsilon$ (linear regression; $A$ hip circumference, $M_1$ hepatic ultrasound attenuation). $\beta_A$ is the SD difference in hepatic ultrasound attenuation per 1 SD higher hip circumference at fixed age and sex.

\textit{Intervals.} 95\% confidence interval.
{\footnotesize\setlength{\tabcolsep}{3pt}
\begin{longtable}{@{}>{\raggedright\arraybackslash}p{0.43\linewidth}>{\raggedleft\arraybackslash}p{0.13\linewidth}>{\raggedright\arraybackslash}p{0.22\linewidth}>{\raggedleft\arraybackslash}p{0.14\linewidth}@{}}
\toprule
Parameter & Estimate & Interval & $P$ \\
\midrule\endhead
Hip circumference--hepatic ultrasound attenuation proxy standardized coefficient $\beta_A$ & $0.2062$ & $[0.1842$, $0.2282]$ & \mbox{$6.755\times10^{-74}$} \\
\bottomrule
\end{longtable}}

\paragraph{12.\ Pairwise association: hip circumference--seated DBP ($\pm 365$ days)}
\textit{Design.} Hip circumference and seated blood pressure measured in the same period ($\pm 365$ days, endpoints inclusive); participants with both measurements and all covariates included, the pair closest in date per person, $n = 2748$. Adjustment for approximate age (at the later of the two measurement dates) and recorded sex; hip circumference, DBP and age standardized within the analysis sample.

\textit{Model.} $M_{2,z} = \beta_0 + \beta_A \cdot A_z + \beta_{\mathrm{age}} \cdot \mathrm{Age}_z + \gamma_{\mathrm{Sex}} + \varepsilon$ (linear regression; $A$ hip circumference, $M_2$ seated DBP). $\beta_A$ is the SD difference in seated DBP per 1 SD higher hip circumference at fixed age and sex.

\textit{Intervals.} 95\% confidence interval.
{\footnotesize\setlength{\tabcolsep}{3pt}
\begin{longtable}{@{}>{\raggedright\arraybackslash}p{0.43\linewidth}>{\raggedleft\arraybackslash}p{0.13\linewidth}>{\raggedright\arraybackslash}p{0.22\linewidth}>{\raggedleft\arraybackslash}p{0.14\linewidth}@{}}
\toprule
Parameter & Estimate & Interval & $P$ \\
\midrule\endhead
Hip circumference--seated DBP standardized coefficient $\beta_A$ & $0.2566$ & $[0.2197$, $0.2936]$ & \mbox{$7.259\times10^{-41}$} \\
\bottomrule
\end{longtable}}

\paragraph{13.\ Pairwise association: hepatic ultrasound attenuation--PWV ($\pm 365$ days)}
\textit{Design.} Hepatic ultrasound and PWV measured in the same period ($\pm 365$ days, endpoints inclusive); participants with both measurements and all covariates included, the pair closest in date per person, $n = 7465$. Adjustment for approximate age (at the later of the two measurement dates) and recorded sex; hepatic attenuation, PWV and age standardized within the analysis sample.

\textit{Model.} $Y_z = \beta_0 + \beta_{M_1} \cdot M_{1,z} + \beta_{\mathrm{age}} \cdot \mathrm{Age}_z + \gamma_{\mathrm{Sex}} + \varepsilon$ (linear regression; $M_1$ hepatic ultrasound attenuation, $Y$ thigh-to-ankle PWV). $\beta_{M_1}$ is the SD difference in PWV per 1 SD higher hepatic ultrasound attenuation at fixed age and sex.

\textit{Intervals.} 95\% confidence interval.
{\footnotesize\setlength{\tabcolsep}{3pt}
\begin{longtable}{@{}>{\raggedright\arraybackslash}p{0.43\linewidth}>{\raggedleft\arraybackslash}p{0.13\linewidth}>{\raggedright\arraybackslash}p{0.22\linewidth}>{\raggedleft\arraybackslash}p{0.14\linewidth}@{}}
\toprule
Parameter & Estimate & Interval & $P$ \\
\midrule\endhead
Hepatic ultrasound attenuation proxy--PWV standardized coefficient $\beta_{M_1}$ & $0.09955$ & $[0.0786$, $0.1205]$ & \mbox{$1.613\times10^{-20}$} \\
\bottomrule
\end{longtable}}

\paragraph{14.\ Pairwise association: seated DBP--PWV ($\pm 365$ days)}
\textit{Design.} Seated blood pressure and PWV measured in the same period ($\pm 365$ days, endpoints inclusive); participants with both measurements and all covariates included, the pair closest in date per person, $n = 1974$. Adjustment for approximate age (at the later of the two measurement dates) and recorded sex; DBP, PWV and age standardized within the analysis sample.

\textit{Model.} $Y_z = \beta_0 + \beta_{M_2} \cdot M_{2,z} + \beta_{\mathrm{age}} \cdot \mathrm{Age}_z + \gamma_{\mathrm{Sex}} + \varepsilon$ (linear regression; $M_2$ seated DBP, $Y$ thigh-to-ankle PWV). $\beta_{M_2}$ is the SD difference in PWV per 1 SD higher seated DBP at fixed age and sex.

\textit{Intervals.} 95\% confidence interval.
{\footnotesize\setlength{\tabcolsep}{3pt}
\begin{longtable}{@{}>{\raggedright\arraybackslash}p{0.43\linewidth}>{\raggedleft\arraybackslash}p{0.13\linewidth}>{\raggedright\arraybackslash}p{0.22\linewidth}>{\raggedleft\arraybackslash}p{0.14\linewidth}@{}}
\toprule
Parameter & Estimate & Interval & $P$ \\
\midrule\endhead
Seated DBP--PWV standardized coefficient $\beta_{M_2}$ & $0.545$ & $[0.5104$, $0.5795]$ & \mbox{$8.767\times10^{-172}$} \\
\bottomrule
\end{longtable}}

\paragraph{15.\ Pairwise association: hepatic ultrasound attenuation--seated DBP ($\pm 365$ days)}
\textit{Design.} Hepatic ultrasound and seated blood pressure measured in the same period ($\pm 365$ days, endpoints inclusive); participants with both measurements and all covariates included, the pair closest in date per person, $n = 2200$. Adjustment for approximate age (at the later of the two measurement dates) and recorded sex; hepatic attenuation, DBP and age standardized within the analysis sample.

\textit{Model.} $M_{2,z} = \beta_0 + \beta_{M_1} \cdot M_{1,z} + \beta_{\mathrm{age}} \cdot \mathrm{Age}_z + \gamma_{\mathrm{Sex}} + \varepsilon$ (linear regression; $M_1$ hepatic ultrasound attenuation, $M_2$ seated DBP). $\beta_{M_1}$ is the SD difference in seated DBP per 1 SD higher hepatic ultrasound attenuation at fixed age and sex.

\textit{Intervals.} 95\% confidence interval.
{\footnotesize\setlength{\tabcolsep}{3pt}
\begin{longtable}{@{}>{\raggedright\arraybackslash}p{0.43\linewidth}>{\raggedleft\arraybackslash}p{0.13\linewidth}>{\raggedright\arraybackslash}p{0.22\linewidth}>{\raggedleft\arraybackslash}p{0.14\linewidth}@{}}
\toprule
Parameter & Estimate & Interval & $P$ \\
\midrule\endhead
Hepatic ultrasound attenuation proxy--seated DBP standardized coefficient $\beta_{M_1}$ & $0.1359$ & $[0.09389$, $0.1779]$ & \mbox{$2.726\times10^{-10}$} \\
\bottomrule
\end{longtable}}

\subsubsection*{Joint models}

\paragraph{16.\ Joint models: hip circumference, hepatic attenuation, DBP and PWV ($\pm 365$ days)}
\textit{Design.} Anchored on the hip circumference measurement day, hepatic ultrasound attenuation, seated DBP and PWV each taken as the measurement closest to that day, all within $-365$ to $+365$ days of the hip circumference day (in either order), with at most 730 days between any two; each participant counted once. The four models use the same sample and all adjust for age (hip circumference measurement year minus birth year) and recorded sex (categorical); hip circumference, hepatic attenuation, DBP, PWV and age standardized once within this sample and shared by the four models.

\textit{Model.} Four linear regressions. Base model: $z(Y) = \beta_0 + \beta_A \cdot z(A) + \beta_{\mathrm{age}} \cdot z(\mathrm{age}) + \gamma_{\mathrm{sex}} + \varepsilon$; the hepatic-proxy model additionally includes $\beta_{M_1} \cdot z(M_1)$, the DBP model additionally includes $\beta_{M_2} \cdot z(M_2)$, and the mutually adjusted model includes both. $A$ is hip circumference, $M_1$ hepatic ultrasound attenuation, $M_2$ seated DBP, $Y$ PWV, and $z(\cdot)$ within-sample standardization. Reported: $\beta_A$, $\beta_{M_1}$ and $\beta_{M_2}$ as contained in each model, each the difference in PWV (in SD) per 1 SD higher value of that variable with the other terms of the same model fixed.

\textit{Intervals.} 95\% confidence intervals; $P$ values not corrected for multiple comparisons.
{\footnotesize\setlength{\tabcolsep}{3pt}
\begin{longtable}{@{}>{\raggedright\arraybackslash}p{0.14\linewidth}>{\raggedright\arraybackslash}p{0.29\linewidth}>{\raggedleft\arraybackslash}p{0.13\linewidth}>{\raggedright\arraybackslash}p{0.22\linewidth}>{\raggedleft\arraybackslash}p{0.14\linewidth}@{}}
\toprule
Block & Parameter & Estimate & Interval & $P$ \\
\midrule\endhead
$\pm 365$-day window; base model: hip circumference $+$ age $+$ sex & Hip circumference coefficient $\beta_A$ (base model) & $0.187$ & $[0.1452$, $0.2288]$ & \mbox{$4.366\times10^{-18}$} \\
$\pm 365$-day window; hepatic-proxy model: hip circumference $+$ hepatic attenuation $+$ age $+$ sex & Hip circumference coefficient $\beta_A$ (hepatic-proxy model) & $0.1723$ & $[0.13$, $0.2146]$ & \mbox{$2.588\times10^{-15}$} \\
 & Hepatic attenuation proxy coefficient $\beta_{M_1}$ (hepatic-proxy model) & $0.08857$ & $[0.04337$, $0.1338]$ & \mbox{$0.0001263$} \\
$\pm 365$-day window; DBP model: hip circumference $+$ seated DBP $+$ age $+$ sex & Hip circumference coefficient $\beta_A$ (DBP model) & $0.05249$ & $[0.01591$, $0.08908]$ & \mbox{$0.004948$} \\
 & Seated DBP coefficient $\beta_{M_2}$ (DBP model) & $0.5268$ & $[0.4859$, $0.5676]$ & \mbox{$1.242\times10^{-118}$} \\
$\pm 365$-day window; mutually adjusted model: hip circumference $+$ hepatic attenuation $+$ seated DBP $+$ age $+$ sex & Hip circumference coefficient $\beta_A$ (mutually adjusted model) & $0.04765$ & $[0.01038$, $0.08493]$ & \mbox{$0.01225$} \\
 & Hepatic attenuation proxy coefficient $\beta_{M_1}$ (mutually adjusted model) & $0.0351$ & $[-0.001711$, $0.0719]$ & \mbox{$0.06163$} \\
 & Seated DBP coefficient $\beta_{M_2}$ (mutually adjusted model) & $0.5229$ & $[0.4823$, $0.5636]$ & \mbox{$2.736\times10^{-118}$} \\
\bottomrule
\end{longtable}}

\paragraph{17.\ Joint models: hip circumference, hepatic attenuation, DBP and PWV ($\pm 180$ days)}
\textit{Design.} Hepatic ultrasound attenuation, seated DBP and PWV each taken as the measurement closest to the hip circumference measurement day, all within $-180$ to $+180$ days of the hip circumference day (in either order), with at most 360 days between any two; each participant counted once. The four models use the same sample, with age and recorded-sex adjustment as in the $\pm 365$-day joint models; hip circumference, hepatic attenuation, DBP, PWV and age standardized once within this sample and shared by the four models.

\textit{Model.} Four linear regressions, as in the $\pm 365$-day joint models. Base model: $z(Y) = \beta_0 + \beta_A \cdot z(A) + \beta_{\mathrm{age}} \cdot z(\mathrm{age}) + \gamma_{\mathrm{sex}} + \varepsilon$; the hepatic-proxy model additionally includes $\beta_{M_1} \cdot z(M_1)$, the DBP model additionally includes $\beta_{M_2} \cdot z(M_2)$, and the mutually adjusted model includes both. Reported: $\beta_A$, $\beta_{M_1}$ and $\beta_{M_2}$ as contained in each model, each the difference in PWV (in SD) per 1 SD higher value of that variable with the other terms of the same model fixed.

\textit{Intervals.} 95\% confidence intervals; $P$ values not corrected for multiple comparisons.
{\footnotesize\setlength{\tabcolsep}{3pt}
\begin{longtable}{@{}>{\raggedright\arraybackslash}p{0.14\linewidth}>{\raggedright\arraybackslash}p{0.29\linewidth}>{\raggedleft\arraybackslash}p{0.13\linewidth}>{\raggedright\arraybackslash}p{0.22\linewidth}>{\raggedleft\arraybackslash}p{0.14\linewidth}@{}}
\toprule
Block & Parameter & Estimate & Interval & $P$ \\
\midrule\endhead
$\pm 180$-day window; base model: hip circumference $+$ age $+$ sex & Hip circumference coefficient $\beta_A$ (base model) & $0.1879$ & $[0.146$, $0.2298]$ & \mbox{$3.397\times10^{-18}$} \\
$\pm 180$-day window; hepatic-proxy model: hip circumference $+$ hepatic attenuation $+$ age $+$ sex & Hip circumference coefficient $\beta_A$ (hepatic-proxy model) & $0.1733$ & $[0.131$, $0.2157]$ & \mbox{$1.969\times10^{-15}$} \\
 & Hepatic attenuation proxy coefficient $\beta_{M_1}$ (hepatic-proxy model) & $0.08798$ & $[0.0427$, $0.1333]$ & \mbox{$0.0001434$} \\
$\pm 180$-day window; DBP model: hip circumference $+$ seated DBP $+$ age $+$ sex & Hip circumference coefficient $\beta_A$ (DBP model) & $0.05322$ & $[0.01657$, $0.08987]$ & \mbox{$0.004455$} \\
 & Seated DBP coefficient $\beta_{M_2}$ (DBP model) & $0.5263$ & $[0.4854$, $0.5673]$ & \mbox{$3.57\times10^{-118}$} \\
$\pm 180$-day window; mutually adjusted model: hip circumference $+$ hepatic attenuation $+$ seated DBP $+$ age $+$ sex & Hip circumference coefficient $\beta_A$ (mutually adjusted model) & $0.04841$ & $[0.01107$, $0.08575]$ & \mbox{$0.01108$} \\
 & Hepatic attenuation proxy coefficient $\beta_{M_1}$ (mutually adjusted model) & $0.03478$ & $[-0.002093$, $0.07165]$ & \mbox{$0.06449$} \\
 & Seated DBP coefficient $\beta_{M_2}$ (mutually adjusted model) & $0.5226$ & $[0.4819$, $0.5633]$ & \mbox{$7.7\times10^{-118}$} \\
\bottomrule
\end{longtable}}

\subsubsection*{Reverse direction}

\paragraph{18.\ Reverse hepatic path: PWV $\to$ hepatic ultrasound attenuation $\to$ hip circumference}
\textit{Design.} Endpoint roles reversed: thigh-to-ankle PWV, hepatic ultrasound attenuation and hip circumference measured in strict calendar-day order (PWV $<$ ultrasound $<$ hip circumference), with each adjacent gap 30--730 days (endpoints inclusive), the same windows as the forward hepatic path; the chain closest in time per participant. Adjustment for approximate age at PWV measurement, recorded sex and the most recent height measured on or within 730 days before the PWV measurement day; PWV, hepatic attenuation, hip circumference, age and height standardized within the analysis sample.

\textit{Model.} $A$ denotes PWV, $M$ hepatic ultrasound attenuation, $Y$ hip circumference and $C$ age, sex and height; linear regressions ($z$ denoting standardized values): $M_z = \alpha_0 + \alpha_A \cdot A_z + \alpha_C \cdot C + \varepsilon_M$; $Y_z = \beta_0 + \beta_A \cdot A_z + \beta_M \cdot M_z + \beta_C \cdot C + \varepsilon_Y$; $Y_z = \gamma_0 + \gamma_A \cdot A_z + \gamma_C \cdot C + \varepsilon_T$. Reported quantities: $\alpha_A$ (PWV $\to$ hepatic attenuation), $\beta_M$ (hepatic attenuation $\to$ hip circumference, adjusted for PWV), $\beta_A$ (PWV $\to$ hip circumference, adjusted for hepatic attenuation), $\gamma_A$ (total PWV $\to$ hip circumference slope), and the reverse path-product index $\theta = \alpha_A \cdot \beta_M$.

\textit{Intervals.} $\alpha_A$, $\beta_M$, $\beta_A$, $\gamma_A$: 95\% confidence intervals; $\theta$: bootstrap 95\% percentile interval.
{\footnotesize\setlength{\tabcolsep}{3pt}
\begin{longtable}{@{}>{\raggedright\arraybackslash}p{0.15\linewidth}>{\raggedright\arraybackslash}p{0.40\linewidth}>{\raggedleft\arraybackslash}p{0.14\linewidth}>{\raggedright\arraybackslash}p{0.24\linewidth}@{}}
\toprule
Block & Parameter & Estimate & Interval \\
\midrule\endhead
Reverse path components (95\% confidence intervals) & $\alpha_A$: standardized slope, PWV $\to$ hepatic ultrasound attenuation (reverse $A \to M$) & $-0.09958$ & $[-0.5657$, $0.3665]$ \\
 & $\beta_M$: standardized slope, hepatic ultrasound attenuation $\to$ hip circumference, adjusted for PWV (reverse $M \to Y \mid A$) & $0.481$ & $[0.1916$, $0.7704]$ \\
 & $\beta_A$: standardized slope, PWV $\to$ hip circumference, adjusted for hepatic attenuation (reverse $A \to Y \mid M$) & $0.1053$ & $[-0.2392$, $0.4499]$ \\
 & $\gamma_A$: adjusted total standardized slope, PWV $\to$ hip circumference (reverse $A \to Y$) & $0.05745$ & $[-0.3053$, $0.4202]$ \\
Reverse path index (bootstrap 95\% percentile interval) & $\theta = \alpha_A \cdot \beta_M$: reverse path-product association index & $-0.0479$ & $[-0.2778$, $0.1492]$ \\
\bottomrule
\end{longtable}}

\paragraph{19.\ Reverse pairing: earlier PWV and subsequently measured hip circumference}
\textit{Design.} Starting from the PWV measurement day, hip circumference measured 1--365 days later (endpoints inclusive); the pair with the shortest gap per participant. Adjustment for approximate age at hip circumference measurement and recorded sex; PWV, hip circumference and age standardized within the analysis sample. $n = 16$.

\textit{Model.} Linear regression: $\mathrm{PWV}_z(\mathrm{earlier}) = \beta_0 + \beta_A \cdot \mathrm{Hip}_z(\mathrm{later}) + \beta_{\mathrm{age}} \cdot \mathrm{Age}_z + \gamma_{\mathrm{sex}} + \varepsilon$, with $\mathrm{Hip}$ hip circumference. Reported quantity: $\beta_A$, the difference in earlier PWV (in SD units) per 1 SD higher later-measured hip circumference at fixed age and sex.

\textit{Intervals.} $\beta_A$: 95\% confidence interval.
{\footnotesize\setlength{\tabcolsep}{3pt}
\begin{longtable}{@{}>{\raggedright\arraybackslash}p{0.43\linewidth}>{\raggedleft\arraybackslash}p{0.13\linewidth}>{\raggedright\arraybackslash}p{0.22\linewidth}>{\raggedleft\arraybackslash}p{0.14\linewidth}@{}}
\toprule
Parameter & Estimate & Interval & $P$ \\
\midrule\endhead
Reverse-direction coefficient $\beta_A$ (hip circumference measured 1--365 days after PWV) & $-0.1983$ & $[-0.6196$, $0.223]$ & \mbox{$0.3253$} \\
\bottomrule
\end{longtable}}

\subsubsection*{Heterogeneity and interaction}

\paragraph{20.\ Sex heterogeneity: hip circumference $\to$ hepatic ultrasound attenuation $\to$ PWV}
\textit{Design.} Hip circumference, hepatic ultrasound and PWV measured in strict calendar-day order: hepatic ultrasound 30--730 days after hip circumference measurement and PWV 30--730 days after hepatic ultrasound (endpoints inclusive in both); one chain per person, $n = 43$. Intercepts and slopes estimated separately for the two recorded sex categories in the analysis sample (first category, second category), with adjustment terms for approximate age on the hip circumference measurement day and same-day height shared across categories. Continuous variables $z$-standardized on the pooled sample.

\textit{Model.} $M = \sum_s \mathbf{1}(\mathrm{sex}=s) \cdot (\alpha_{0s} + \alpha_{As} \cdot A) + \alpha_{\mathrm{age}} \cdot \mathrm{age} + \alpha_{\mathrm{ht}} \cdot \mathrm{ht} + \varepsilon_M$; $Y = \sum_s \mathbf{1}(\mathrm{sex}=s) \cdot (\beta_{0s} + \beta_{As} \cdot A + \beta_{Ms} \cdot M) + \beta_{\mathrm{age}} \cdot \mathrm{age} + \beta_{\mathrm{ht}} \cdot \mathrm{ht} + \varepsilon_Y$; $Y = \sum_s \mathbf{1}(\mathrm{sex}=s) \cdot (\gamma_{0s} + \gamma_{As} \cdot A) + \gamma_{\mathrm{age}} \cdot \mathrm{age} + \gamma_{\mathrm{ht}} \cdot \mathrm{ht} + \varepsilon_T$ (linear regressions; $A$ hip circumference, $M$ hepatic ultrasound attenuation, $Y$ PWV, all $z$ values). Reported quantities: within category $s$, the $A \to M$ slope $\alpha_{As}$, the $A$-adjusted $M \to Y$ slope $\beta_{Ms}$, the $M$-adjusted $A \to Y$ slope $\beta_{As}$ and the total $A \to Y$ slope $\gamma_{As}$ (in pooled-sample SD units), and the path-product index $\theta_s = \alpha_{As} \cdot \beta_{Ms}$.

\textit{Intervals.} Slopes $\alpha$, $\beta$, $\gamma$: 95\% confidence intervals; $\theta_s$: bootstrap 95\% percentile interval.
{\footnotesize\setlength{\tabcolsep}{3pt}
\begin{longtable}{@{}>{\raggedright\arraybackslash}p{0.14\linewidth}>{\raggedright\arraybackslash}p{0.29\linewidth}>{\raggedleft\arraybackslash}p{0.13\linewidth}>{\raggedright\arraybackslash}p{0.22\linewidth}>{\raggedleft\arraybackslash}p{0.14\linewidth}@{}}
\toprule
Block & Parameter & Estimate & Interval & $P$ \\
\midrule\endhead
First recorded-sex category (95\% confidence intervals) & $\alpha_{A,1}$: standardized slope within the first category, hip circumference $\to$ hepatic ultrasound attenuation & $0.4955$ & $[-0.03314$, $1.024]$ & \mbox{$0.06536$} \\
 & $\beta_{M,1}$: standardized slope within the first category, hepatic attenuation $\to$ PWV, adjusted for hip circumference & $0.1895$ & $[-0.2266$, $0.6056]$ & \mbox{$0.3616$} \\
 & $\beta_{A,1}$: standardized slope within the first category, hip circumference $\to$ PWV, adjusted for hepatic attenuation & $-0.04897$ & $[-0.4217$, $0.3238]$ & \mbox{$0.7912$} \\
 & $\gamma_{A,1}$: adjusted total standardized slope within the first category, hip circumference $\to$ PWV & $0.04318$ & $[-0.2429$, $0.3293]$ & \mbox{$0.7615$} \\
 & $\theta_1 = \alpha_{A,1} \cdot \beta_{M,1}$: first-category path-product association index & $0.09388$ & $[-0.08691$, $0.3422]$ &  \\
Second recorded-sex category (95\% confidence intervals) & $\alpha_{A,2}$: standardized slope within the second category, hip circumference $\to$ hepatic ultrasound attenuation & $0.4053$ & $[-0.4208$, $1.231]$ & \mbox{$0.3266$} \\
 & $\beta_{M,2}$: standardized slope within the second category, hepatic attenuation $\to$ PWV, adjusted for hip circumference & $0.05124$ & $[-0.6936$, $0.7961]$ & \mbox{$0.8897$} \\
 & $\beta_{A,2}$: standardized slope within the second category, hip circumference $\to$ PWV, adjusted for hepatic attenuation & $0.2433$ & $[-0.4336$, $0.9202]$ & \mbox{$0.4705$} \\
 & $\gamma_{A,2}$: adjusted total standardized slope within the second category, hip circumference $\to$ PWV & $0.2675$ & $[-0.4583$, $0.9933]$ & \mbox{$0.46$} \\
 & $\theta_2 = \alpha_{A,2} \cdot \beta_{M,2}$: second-category path-product association index & $0.02077$ & $[-0.2457$, $0.3786]$ &  \\
\bottomrule
\end{longtable}}

\paragraph{21.\ Conditional PWV paths under the hip circumference $\times$ hepatic ultrasound attenuation interaction}
\textit{Design.} Hip circumference, hepatic ultrasound and PWV measured in strict calendar-day order: hepatic ultrasound 30--730 days after hip circumference measurement and PWV 30--730 days after hepatic ultrasound (endpoints inclusive in both); one chain per person, $n = 43$. Adjustment for approximate age on the hip circumference measurement day, recorded sex and same-day height. Hip circumference, hepatic ultrasound attenuation, PWV, age and height $z$-standardized within the analysis sample.

\textit{Model.} $M = \alpha_0 + \alpha_A \cdot A + \alpha_C^{\top} \cdot C + \varepsilon_M$; $Y = \beta_0 + \beta_A \cdot A + \beta_M \cdot M + \beta_{AM} \cdot A \cdot M + \beta_C^{\top} \cdot C + \varepsilon_Y$; $Y = \gamma_0 + \gamma_A \cdot A + \gamma_C^{\top} \cdot C + \varepsilon_T$ (linear regressions; $A$ hip circumference, $M$ hepatic ultrasound attenuation, $Y$ PWV, all $z$ values; $C$ age, sex and height). Reported quantities: the $A \to M$ slope $\alpha_A$; the $M \to Y$ slope $\beta_M$ at mean $A$; the $A \to Y$ slope $\beta_A$ at mean $M$; the total $A \to Y$ slope $\gamma_A$; the interaction coefficient $\beta_{AM}$; and the conditional path-product index $\theta(a) = \alpha_A \cdot (\beta_M + \beta_{AM} \cdot a)$, evaluated at $a = 0$ (mean $A$) and at the 25th, 50th and 75th percentiles of $A$ in the analysis sample.

\textit{Intervals.} $\alpha_A$, $\beta_M$, $\beta_A$, $\gamma_A$: 95\% confidence intervals; $\theta(0)$, $\theta(a)$ at the three percentiles and $\beta_{AM}$: bootstrap 95\% percentile intervals.
{\footnotesize\setlength{\tabcolsep}{3pt}
\begin{longtable}{@{}>{\raggedright\arraybackslash}p{0.15\linewidth}>{\raggedright\arraybackslash}p{0.40\linewidth}>{\raggedleft\arraybackslash}p{0.14\linewidth}>{\raggedright\arraybackslash}p{0.24\linewidth}@{}}
\toprule
Block & Parameter & Estimate & Interval \\
\midrule\endhead
Path components of the model with the interaction term (95\% confidence intervals) & $\alpha_A$: standardized slope, hip circumference $\to$ hepatic ultrasound attenuation ($A \to M$) & $0.4586$ & $[0.06006$, $0.8571]$ \\
 & $\beta_M$: conditional standardized slope, hepatic attenuation $\to$ PWV, at the exposure mean ($A_z = 0$) & $0.1554$ & $[-0.3084$, $0.6192]$ \\
 & $\beta_A$: conditional standardized slope, hip circumference $\to$ PWV, at the mediator mean ($M_z = 0$) & $0.08855$ & $[-0.2535$, $0.4306]$ \\
 & $\gamma_A$: adjusted total standardized slope, hip circumference $\to$ PWV ($A \to Y$) & $0.135$ & $[-0.212$, $0.482]$ \\
Path index at the exposure mean (bootstrap 95\% percentile interval) & $\theta(0) = \alpha_A \cdot \beta_M$: conditional path-product index at the exposure mean & $0.07127$ & $[-0.107$, $0.3025]$ \\
Conditional paths under the exposure $\times$ mediator interaction (bootstrap 95\% percentile intervals) & $\theta(Q_{25}) = \alpha_A \cdot (\beta_M + \beta_{AM} \cdot a_{25})$: conditional path index at the 25th percentile of the exposure & $0.08427$ & $[-0.1448$, $0.4065]$ \\
 & $\theta(Q_{50}) = \alpha_A \cdot (\beta_M + \beta_{AM} \cdot a_{50})$: conditional path index at the exposure median & $0.07419$ & $[-0.1137$, $0.3238]$ \\
 & $\theta(Q_{75}) = \alpha_A \cdot (\beta_M + \beta_{AM} \cdot a_{75})$: conditional path index at the 75th percentile of the exposure & $0.06412$ & $[-0.09558$, $0.2653]$ \\
 & $\beta_{AM}$: hip circumference $\times$ hepatic attenuation interaction coefficient in the conditional outcome equation & $-0.04257$ & $[-0.2992$, $0.3505]$ \\
\bottomrule
\end{longtable}}

\paragraph{22.\ Sex heterogeneity: hip circumference $\to$ hepatic ultrasound attenuation $\to$ DBP $\to$ PWV}
\textit{Design.} Hip circumference, hepatic ultrasound, seated blood pressure and PWV measured sequentially by date (same day allowed), with adjacent gaps of 0--180 days and a total span of 0--365 days from hip circumference to PWV (endpoints inclusive throughout); one chain per person. Recorded sex (first category, second category) enters as a main effect and through interaction terms with each upstream variable; adjustment for approximate age on the hip circumference measurement day and the most recent height on or within 730 days before that day. Continuous variables $z$-standardized within the analysis sample.

\textit{Model.} $M_1 = \alpha_1 + a_1 \cdot A + f(\mathrm{sex}) + A \times \mathrm{sex} + \delta_1^{\top} C + \varepsilon_1$; $M_2 = \alpha_2 + a_2 \cdot A + d_{21} \cdot M_1 + f(\mathrm{sex}) + (A, M_1) \times \mathrm{sex} + \delta_2^{\top} C + \varepsilon_2$; $Y = \alpha_3 + c' \cdot A + b_1 \cdot M_1 + b_2 \cdot M_2 + f(\mathrm{sex}) + (A, M_1, M_2) \times \mathrm{sex} + \delta_3^{\top} C + \varepsilon_3$ (pooled-sample linear regressions; $A$ hip circumference, $M_1$ hepatic ultrasound attenuation, $M_2$ seated DBP, $Y$ PWV, all $z$ values; $f(\mathrm{sex})$ the sex main effect and $\times\,\mathrm{sex}$ the interaction terms with sex; $C$ age and height). Reported quantities: within category $s$, the slopes of the fitted expectation on the corresponding upstream variable, $a_1(s)$, $a_2(s)$, $d_{21}(s)$, $b_1(s)$, $b_2(s)$ and $c'(s)$, and the serial path product $a_1(s) \cdot d_{21}(s) \cdot b_2(s)$.

\textit{Intervals.} All category-specific slopes and serial path products: bootstrap 95\% percentile intervals.
{\footnotesize\setlength{\tabcolsep}{3pt}
\begin{longtable}{@{}>{\raggedright\arraybackslash}p{0.15\linewidth}>{\raggedright\arraybackslash}p{0.40\linewidth}>{\raggedleft\arraybackslash}p{0.14\linewidth}>{\raggedright\arraybackslash}p{0.24\linewidth}@{}}
\toprule
Block & Parameter & Estimate & Interval \\
\midrule\endhead
First sex category & Category-specific serial path product $a_1(s) \cdot d_{21}(s) \cdot b_2(s)$ & $0.005615$ & $[0.003191$, $0.008495]$ \\
 & Category-specific slope $a_1(s)$ (hip circumference $\to$ hepatic ultrasound attenuation) & $0.1695$ & $[0.1411$, $0.1977]$ \\
 & Category-specific slope $a_2(s)$ (hip circumference $\to$ DBP, controlling for hepatic ultrasound attenuation) & $0.2404$ & $[0.2143$, $0.2664]$ \\
 & Category-specific slope $d_{21}(s)$ (hepatic ultrasound attenuation $\to$ DBP, controlling for hip circumference) & $0.06953$ & $[0.04064$, $0.1007]$ \\
 & Category-specific slope $b_1(s)$ (hepatic ultrasound attenuation $\to$ PWV, controlling for hip circumference and DBP) & $0.03948$ & $[0.01201$, $0.06483]$ \\
 & Category-specific slope $b_2(s)$ (DBP $\to$ PWV, controlling for hip circumference and hepatic ultrasound attenuation) & $0.4765$ & $[0.448$, $0.5041]$ \\
 & Category-specific conditional slope $c'(s)$ (hip circumference $\to$ PWV, controlling for both mediators) & $-0.03224$ & $[-0.05611$, $-0.007816]$ \\
Second sex category & Category-specific serial path product $a_1(s) \cdot d_{21}(s) \cdot b_2(s)$ & $0.009779$ & $[0.006492$, $0.01331]$ \\
 & Category-specific slope $a_1(s)$ (hip circumference $\to$ hepatic ultrasound attenuation) & $0.2597$ & $[0.2173$, $0.2999]$ \\
 & Category-specific slope $a_2(s)$ (hip circumference $\to$ DBP, controlling for hepatic ultrasound attenuation) & $0.302$ & $[0.2641$, $0.3395]$ \\
 & Category-specific slope $d_{21}(s)$ (hepatic ultrasound attenuation $\to$ DBP, controlling for hip circumference) & $0.09645$ & $[0.06538$, $0.1253]$ \\
 & Category-specific slope $b_1(s)$ (hepatic ultrasound attenuation $\to$ PWV, controlling for hip circumference and DBP) & $0.059$ & $[0.03189$, $0.08482]$ \\
 & Category-specific slope $b_2(s)$ (DBP $\to$ PWV, controlling for hip circumference and hepatic ultrasound attenuation) & $0.3905$ & $[0.3592$, $0.4222]$ \\
 & Category-specific conditional slope $c'(s)$ (hip circumference $\to$ PWV, controlling for both mediators) & $-0.005606$ & $[-0.0385$, $0.02908]$ \\
\bottomrule
\end{longtable}}

\paragraph{23.\ Hip circumference $\times$ mediator interaction: hepatic ultrasound attenuation $\to$ DBP $\to$ PWV serial path}
\textit{Design.} Hip circumference, hepatic ultrasound, seated blood pressure and PWV measured sequentially by date (same day allowed), with adjacent gaps of 0--180 days and a total span of 0--365 days from hip circumference to PWV (endpoints inclusive throughout); one chain per person. Adjustment for approximate age on the hip circumference measurement day, recorded sex and the most recent height on or within 730 days before that day. Continuous variables $z$-standardized within the analysis sample; $a = -1$, $0$ and $+1$ correspond to hip circumference 1 SD below the mean, at the mean and 1 SD above the mean.

\textit{Model.} $M_1 = \alpha_1 + a_1 \cdot A + \gamma_1^{\top} C + \varepsilon_1$; $M_2 = \alpha_2 + a_2 \cdot A + d_{21} \cdot M_1 + \theta_{M2,AM1} \cdot A \cdot M_1 + \gamma_2^{\top} C + \varepsilon_2$; $Y = \alpha_3 + c' \cdot A + b_1 \cdot M_1 + b_2 \cdot M_2 + \theta_{Y,AM1} \cdot A \cdot M_1 + \theta_{Y,AM2} \cdot A \cdot M_2 + \gamma_3^{\top} C + \varepsilon_3$ (linear regressions; $A$ hip circumference, $M_1$ hepatic ultrasound attenuation, $M_2$ seated DBP, $Y$ PWV, all $z$ values; $C$ age, sex and height). Reported quantities: the three product-term coefficients $\theta_{M2,AM1}$, $\theta_{Y,AM1}$ and $\theta_{Y,AM2}$, and the conditional serial path product at hip circumference level $a$, $a_1 \cdot (d_{21} + \theta_{M2,AM1} \cdot a) \cdot (b_2 + \theta_{Y,AM2} \cdot a)$, $a \in \{-1, 0, +1\}$.

\textit{Intervals.} Three product-term coefficients and three conditional serial path products: bootstrap 95\% percentile intervals.
{\footnotesize\setlength{\tabcolsep}{3pt}
\begin{longtable}{@{}>{\raggedright\arraybackslash}p{0.15\linewidth}>{\raggedright\arraybackslash}p{0.40\linewidth}>{\raggedleft\arraybackslash}p{0.14\linewidth}>{\raggedright\arraybackslash}p{0.24\linewidth}@{}}
\toprule
Block & Parameter & Estimate & Interval \\
\midrule\endhead
Exposure $\times$ mediator product-term coefficients & Hip circumference $\times$ hepatic ultrasound attenuation product-term coefficient in the DBP equation, $\theta_{M2,AM1}$ & $0.01034$ & $[-0.01034$, $0.0313]$ \\
 & Hip circumference $\times$ hepatic ultrasound attenuation product-term coefficient in the PWV equation, $\theta_{Y,AM1}$ & $0.007362$ & $[-0.01132$, $0.02576]$ \\
 & Hip circumference $\times$ DBP product-term coefficient in the PWV equation, $\theta_{Y,AM2}$ & $-0.01752$ & $[-0.03722$, $0.00175]$ \\
Hip circumference 1 SD below the analysis-set mean ($a = -1$) & Conditional serial path product $a_1 \cdot (d_{21} + \theta_{M2,AM1} \cdot a) \cdot (b_2 + \theta_{Y,AM2} \cdot a)$ & $0.006668$ & $[0.003749$, $0.009758]$ \\
Hip circumference at the analysis-set mean ($a = 0$) & Conditional serial path product $a_1 \cdot (d_{21} + \theta_{M2,AM1} \cdot a) \cdot (b_2 + \theta_{Y,AM2} \cdot a)$ & $0.007324$ & $[0.005353$, $0.009523]$ \\
Hip circumference 1 SD above the analysis-set mean ($a = +1$) & Conditional serial path product $a_1 \cdot (d_{21} + \theta_{M2,AM1} \cdot a) \cdot (b_2 + \theta_{Y,AM2} \cdot a)$ & $0.007906$ & $[0.005474$, $0.0106]$ \\
\bottomrule
\end{longtable}}

\paragraph{24.\ Hip circumference $\times$ age interaction: variation of the hip circumference--PWV slope with age}
\textit{Design.} Hip circumference and PWV measured in the same period: PWV day within $-365$ to $+365$ days of the hip circumference day (endpoints inclusive, in either order); one row per person, $n = 8956$. Approximate age (year of the later measurement day in the pair minus birth year) entered in continuous form both as a main effect and as the modifier, with further adjustment for recorded sex. Hip circumference, PWV and age $z$-standardized within the analysis sample.

\textit{Model.} $z(Y) = \beta_0 + \beta_A \cdot z(A) + \beta_{\mathrm{age}} \cdot z(\mathrm{age}) + \beta_{A \times \mathrm{age}} \cdot z(A) \cdot z(\mathrm{age}) + \gamma_{\mathrm{sex}} + \varepsilon$ (linear regression; $A$ hip circumference, $Y$ PWV). Reported quantity: the interaction coefficient $\beta_{A \times \mathrm{age}}$, the change in the standardized hip circumference slope per 1 SD higher age.

\textit{Intervals.} 95\% confidence interval.
{\footnotesize\setlength{\tabcolsep}{3pt}
\begin{longtable}{@{}>{\raggedright\arraybackslash}p{0.43\linewidth}>{\raggedleft\arraybackslash}p{0.13\linewidth}>{\raggedright\arraybackslash}p{0.22\linewidth}>{\raggedleft\arraybackslash}p{0.14\linewidth}@{}}
\toprule
Parameter & Estimate & Interval & $P$ \\
\midrule\endhead
Hip circumference $\times$ age interaction coefficient $\beta_{A \times \mathrm{age}}$ (change in the standardized hip circumference slope per 1 SD higher age) & $-0.02959$ & $[-0.04883$, $-0.01034]$ & \mbox{$0.002585$} \\
\bottomrule
\end{longtable}}

\subsubsection*{Sensitivity analyses}

\paragraph{25.\ Wide-window serial path: hip circumference $\to$ hepatic ultrasound attenuation $\to$ DBP $\to$ PWV}
\textit{Design.} Hip circumference, hepatic ultrasound, seated blood pressure and thigh-to-ankle PWV measured in date order within one chain (same day allowed), with adjacent gaps of 0--365 days and a total span of 0--730 days from hip circumference to PWV (endpoints inclusive throughout); one chain per person. The four equations fitted on the same sample, with adjustment for approximate age on the hip circumference measurement day, recorded sex and the most recent height on or within 730 days before that day; continuous variables $z$-standardized within the analysis sample.

\textit{Model.} Linear regressions ($A$ hip circumference, $M_1$ hepatic ultrasound attenuation, $M_2$ seated DBP, $Y$ thigh-to-ankle PWV, $C$ covariates): $M_1 = \alpha_1 + a_1 A + \gamma_1^{\top} C$; $M_2 = \alpha_2 + a_2 A + d_{21} M_1 + \gamma_2^{\top} C$; $Y = \alpha_3 + c' A + b_1 M_1 + b_2 M_2 + \gamma_3^{\top} C$; $Y = \alpha_T + c A + \gamma_T^{\top} C$. Reported: the serial path index $a_1 d_{21} b_2$; the seven standardized coefficients $a_1$, $a_2$, $d_{21}$, $b_1$, $b_2$, $c'$, $c$; the additive decomposition $a_1 b_1$, $a_2 b_2$, $a_1 d_{21} b_2$, $c'$, the algebraic sum of these four, and $c$.

\textit{Intervals.} All rows (path index, component coefficients and additive decomposition): bootstrap 95\% percentile intervals.
{\footnotesize\setlength{\tabcolsep}{3pt}
\begin{longtable}{@{}>{\raggedright\arraybackslash}p{0.15\linewidth}>{\raggedright\arraybackslash}p{0.40\linewidth}>{\raggedleft\arraybackslash}p{0.14\linewidth}>{\raggedright\arraybackslash}p{0.24\linewidth}@{}}
\toprule
Block & Parameter & Estimate & Interval \\
\midrule\endhead
Serial path index (wide window) & Serial path index $a_1 \cdot d_{21} \cdot b_2$ & $0.007426$ & $[0.005351$, $0.009601]$ \\
Component coefficients (wide window) & Standardized slope $a_1$ (hip circumference $\to$ hepatic ultrasound attenuation) & $0.2035$ & $[0.1795$, $0.2283]$ \\
 & Standardized slope $a_2$ (hip circumference $\to$ DBP, controlling for hepatic ultrasound attenuation) & $0.2634$ & $[0.2415$, $0.2851]$ \\
 & Standardized slope $d_{21}$ (hepatic ultrasound attenuation $\to$ DBP, controlling for hip circumference) & $0.08413$ & $[0.06158$, $0.1042]$ \\
 & Standardized slope $b_1$ (hepatic ultrasound attenuation $\to$ PWV, controlling for hip circumference and DBP) & $0.04858$ & $[0.03015$, $0.06662]$ \\
 & Standardized slope $b_2$ (DBP $\to$ PWV, controlling for hip circumference and hepatic ultrasound attenuation) & $0.4338$ & $[0.4133$, $0.4557]$ \\
 & Conditional slope $c'$ (hip circumference $\to$ PWV, controlling for both mediators) & $-0.02044$ & $[-0.04091$, $-0.00164]$ \\
 & Total-association slope $c$ (hip circumference $\to$ PWV, total-association equation) & $0.1111$ & $[0.08975$, $0.1316]$ \\
Additive path decomposition (wide window) & Path product via hepatic ultrasound attenuation only, $a_1 \cdot b_1$ & $0.009885$ & $[0.00606$, $0.01394]$ \\
 & Path product via DBP only, $a_2 \cdot b_2$ & $0.1143$ & $[0.1036$, $0.1256]$ \\
 & Serial path product $a_1 \cdot d_{21} \cdot b_2$ (serial term of the decomposition) & $0.007426$ & $[0.005351$, $0.009601]$ \\
 & Remaining conditional component $c'$ & $-0.02044$ & $[-0.04091$, $-0.00164]$ \\
 & Algebraic sum $c' + a_1 b_1 + a_2 b_2 + a_1 d_{21} b_2$ & $0.1111$ & $[0.08975$, $0.1316]$ \\
 & Total-association equation slope $c$ & $0.1111$ & $[0.08975$, $0.1316]$ \\
\bottomrule
\end{longtable}}

\paragraph{26.\ Conditioning on same-day BMI: hip circumference $\to$ hepatic ultrasound attenuation $\to$ DBP $\to$ PWV}
\textit{Design.} Hip circumference, hepatic ultrasound, seated blood pressure and thigh-to-ankle PWV measured in date order within one chain (same day allowed), with adjacent gaps of 0--180 days and a total span of 0--365 days from hip circumference to PWV (endpoints inclusive throughout); one chain per person. The four equations fitted on the same sample, with adjustment for approximate age on the hip circumference measurement day, recorded sex, the most recent height on or within 730 days before that day, and additionally BMI on the hip circumference measurement day; continuous variables $z$-standardized within the analysis sample.

\textit{Model.} Linear regressions ($A$ hip circumference, $M_1$ hepatic ultrasound attenuation, $M_2$ seated DBP, $Y$ thigh-to-ankle PWV, $C$ covariates including same-day BMI): $M_1 = \alpha_1 + a_1 A + \gamma_1^{\top} C$; $M_2 = \alpha_2 + a_2 A + d_{21} M_1 + \gamma_2^{\top} C$; $Y = \alpha_3 + c' A + b_1 M_1 + b_2 M_2 + \gamma_3^{\top} C$; $Y = \alpha_T + c A + \gamma_T^{\top} C$. Reported: the serial path index $a_1 d_{21} b_2$, the seven standardized coefficients and the additive decomposition; the hip circumference coefficients $a_1$, $a_2$, $c'$ and $c$ are SD differences per 1 SD higher hip circumference at equal same-day BMI.

\textit{Intervals.} All rows (path index, component coefficients and additive decomposition): bootstrap 95\% percentile intervals.
{\footnotesize\setlength{\tabcolsep}{3pt}
\begin{longtable}{@{}>{\raggedright\arraybackslash}p{0.15\linewidth}>{\raggedright\arraybackslash}p{0.40\linewidth}>{\raggedleft\arraybackslash}p{0.14\linewidth}>{\raggedright\arraybackslash}p{0.24\linewidth}@{}}
\toprule
Block & Parameter & Estimate & Interval \\
\midrule\endhead
Serial path index (conditional on same-day BMI) & Serial path index $a_1 \cdot d_{21} \cdot b_2$ & $-0.001464$ & $[-0.00266$, $-0.0005181]$ \\
Component coefficients (conditional on same-day BMI) & Standardized slope $a_1$ (hip circumference $\to$ hepatic ultrasound attenuation) & $-0.06817$ & $[-0.111$, $-0.02624]$ \\
 & Standardized slope $a_2$ (hip circumference $\to$ DBP, controlling for hepatic ultrasound attenuation) & $-0.05015$ & $[-0.09067$, $-0.007894]$ \\
 & Standardized slope $d_{21}$ (hepatic ultrasound attenuation $\to$ DBP, controlling for hip circumference) & $0.04983$ & $[0.02831$, $0.07055]$ \\
 & Standardized slope $b_1$ (hepatic ultrasound attenuation $\to$ PWV, controlling for hip circumference and DBP) & $0.0465$ & $[0.02822$, $0.06489]$ \\
 & Standardized slope $b_2$ (DBP $\to$ PWV, controlling for hip circumference and hepatic ultrasound attenuation) & $0.431$ & $[0.408$, $0.4535]$ \\
 & Conditional slope $c'$ (hip circumference $\to$ PWV, controlling for both mediators) & $-0.04036$ & $[-0.0787$, $-0.002933]$ \\
 & Total-association slope $c$ (hip circumference $\to$ PWV, total-association equation) & $-0.06661$ & $[-0.1099$, $-0.02552]$ \\
Additive path decomposition (conditional on same-day BMI) & Path product via hepatic ultrasound attenuation only, $a_1 \cdot b_1$ & $-0.00317$ & $[-0.005929$, $-0.001121]$ \\
 & Path product via DBP only, $a_2 \cdot b_2$ & $-0.02162$ & $[-0.03936$, $-0.00343]$ \\
 & Serial path product $a_1 \cdot d_{21} \cdot b_2$ (serial term of the decomposition) & $-0.001464$ & $[-0.00266$, $-0.0005181]$ \\
 & Remaining conditional component $c'$ & $-0.04036$ & $[-0.0787$, $-0.002933]$ \\
 & Algebraic sum $c' + a_1 b_1 + a_2 b_2 + a_1 d_{21} b_2$ & $-0.06661$ & $[-0.1099$, $-0.02552]$ \\
 & Total-association equation slope $c$ & $-0.06661$ & $[-0.1099$, $-0.02552]$ \\
\bottomrule
\end{longtable}}

\paragraph{27.\ Hip circumference--PWV association conditional on same-day BMI}
\textit{Design.} Hip circumference and bilateral thigh-to-ankle PWV measured in the same period: PWV day within $-365$ to $+365$ days of the hip circumference day (endpoints inclusive, in either order); the closest pair per person, one row per person, $n = 8937$. BMI taken on the same day as the paired hip circumference measurement. Adjustment for approximate age (year of the later measurement day in the pair minus birth year), recorded sex and same-day BMI; continuous variables $z$-standardized within the analysis sample.

\textit{Model.} Linear regression: $z(Y) = \beta_0 + \beta_{A \mid \mathrm{BMI}} \cdot z(A) + \beta_{\mathrm{BMI}} \cdot z(\mathrm{BMI}) + \beta_{\mathrm{age}} \cdot z(\mathrm{age}) + \gamma_{\mathrm{sex}} + \varepsilon$ ($A$ hip circumference, $Y$ PWV). $\beta_{A \mid \mathrm{BMI}}$ is the SD difference in PWV per 1 SD higher hip circumference at fixed BMI, age and sex.

\textit{Intervals.} 95\% confidence interval.
{\footnotesize\setlength{\tabcolsep}{3pt}
\begin{longtable}{@{}>{\raggedright\arraybackslash}p{0.43\linewidth}>{\raggedleft\arraybackslash}p{0.13\linewidth}>{\raggedright\arraybackslash}p{0.22\linewidth}>{\raggedleft\arraybackslash}p{0.14\linewidth}@{}}
\toprule
Parameter & Estimate & Interval & $P$ \\
\midrule\endhead
Hip circumference coefficient conditional on same-day BMI, $\beta_{A \mid \mathrm{BMI}}$ & $0.028$ & $[-0.005853$, $0.06185]$ & \mbox{$0.105$} \\
\bottomrule
\end{longtable}}

\paragraph{28.\ Negative-control outcome: hip circumference and left-minus-right PWV difference}
\textit{Design.} Hip circumference and thigh-to-ankle PWV measured in the same period: PWV day within $-365$ to $+365$ days of the hip circumference day (endpoints inclusive, in either order); the closest pair per person, one row per person, $n = 8956$. Outcome: the signed difference of left-side minus right-side PWV within the same record. Adjustment for approximate age (year of the later measurement day in the pair minus birth year) and recorded sex; continuous variables $z$-standardized within the analysis sample.

\textit{Model.} Linear regression: $z(D) = \beta_0 + \beta_A \cdot z(A) + \beta_{\mathrm{age}} \cdot z(\mathrm{age}) + \gamma_{\mathrm{sex}} + \varepsilon$ ($A$ hip circumference, $D$ left-minus-right PWV difference). $\beta_A$ is the SD difference in $D$ per 1 SD higher hip circumference at fixed age and sex.

\textit{Intervals.} 95\% confidence interval.
{\footnotesize\setlength{\tabcolsep}{3pt}
\begin{longtable}{@{}>{\raggedright\arraybackslash}p{0.43\linewidth}>{\raggedleft\arraybackslash}p{0.13\linewidth}>{\raggedright\arraybackslash}p{0.22\linewidth}>{\raggedleft\arraybackslash}p{0.14\linewidth}@{}}
\toprule
Parameter & Estimate & Interval & $P$ \\
\midrule\endhead
Hip circumference--left-minus-right PWV difference (negative-control outcome) standardized coefficient $\beta_A$ & $-0.006622$ & $[-0.03011$, $0.01686]$ & \mbox{$0.5805$} \\
\bottomrule
\end{longtable}}

\FloatBarrier

\subsection{WHR--LDL--DBP--PWV}
\label{app:rq-designs:9ac6}

\paragraph{Research question and measures.}
The research question concerns the association of waist-to-hip ratio with lower-limb arterial stiffness through LDL cholesterol and then diastolic blood pressure, with data from the HPP cohort. The exposure is waist-to-hip ratio; the mediators are blood-test LDL cholesterol and seated diastolic blood pressure (mean of the available values of two seated readings); the outcome is thigh-to-ankle pulse wave velocity (PWV, mean of the available sides). Unless otherwise stated, coefficients are unstandardized regression coefficients on the original recorded scale, that is, the change in the downstream variable when the upstream variable is 1 recorded unit higher.

\subsubsection*{Primary pathways}

\paragraph{1.\ Single-mediator path: waist-to-hip ratio $\to$ LDL cholesterol $\to$ thigh-to-ankle PWV}
\textit{Design.} Waist-to-hip ratio, LDL cholesterol and thigh-to-ankle PWV measured in strict date order: LDL 30--730 days after waist-to-hip ratio and PWV 30--1095 days after LDL (endpoints inclusive); one measurement chain per participant. Adjustment for approximate age at waist-to-hip ratio measurement and its square, sex, sub-study membership, same-day height, and the most recent smoking status on the same day or within the preceding 730 days. PWV as the mean of the available sides. All variables on the original recorded scale, unstandardized.

\textit{Model.} $M = \alpha_0 + \alpha_A A + \alpha_C^\top C + \varepsilon_M$; $Y = \beta_0 + \beta_A A + \beta_M M + \beta_{AM} A M + \beta_C^\top C + \varepsilon_Y$. Both equations are linear regressions, with $A$, $M$ and $Y$ the waist-to-hip ratio, LDL cholesterol and thigh-to-ankle PWV, and $C$ the covariates. $\alpha_A$ is the change in LDL per 1 recorded unit higher waist-to-hip ratio; $\beta_M(q) = \beta_M + \beta_{AM} q$ is the conditional slope of PWV on LDL at waist-to-hip ratio $q$, and $\beta_{AM}$ is the interaction coefficient; the path product is $\theta(q) = \alpha_A \beta_M(q)$, with $q$ at the 25th, 50th and 75th percentiles of waist-to-hip ratio in the analysis sample ($q_{25}$, $q_{50}$, $q_{75}$). The bias bound is $\delta = \mathrm{se}\sqrt{\nu R^2_{Y\sim U} R^2_{D\sim U}/(1 - R^2_{D\sim U})}$, reported as $\hat\beta \pm \delta$ with $\hat\beta \in \{\alpha_A, \beta_M(q_{50})\}$, where $\mathrm{se}$ is the standard error of that coefficient, $\nu$ the residual degrees of freedom, and $R^2_{D\sim U}$ and $R^2_{Y\sim U}$ the partial $R^2$ of a hypothetical omitted covariate $U$ with the regressor of that coefficient and with the outcome of its equation, each taking 0.01, 0.05, 0.10, 0.20 and 0.30.

\textit{Intervals.} $\alpha_A$, $\beta_M(q)$, $\beta_{AM}$: bootstrap 95\% percentile intervals; $\theta(q)$: bootstrap 95\% simultaneous confidence band (joint over the three percentile points). Bias-bound grid: each row corresponds to one pair $(R^2_{D\sim U}, R^2_{Y\sim U})$, with lower and upper limits equal to the coefficient $\pm$ the bias bound.
{\footnotesize\setlength{\tabcolsep}{3pt}
\begin{longtable}{@{}>{\raggedright\arraybackslash}p{0.15\linewidth}>{\raggedright\arraybackslash}p{0.40\linewidth}>{\raggedleft\arraybackslash}p{0.14\linewidth}>{\raggedright\arraybackslash}p{0.24\linewidth}@{}}
\toprule
Block & Parameter & Estimate & Interval \\
\midrule\endhead
Path-product curve & $\theta(q_{25}) = \alpha_A \beta_M(q_{25})$: path product at the 25th percentile of waist-to-hip ratio & $-0.01432$ & $[-0.3669$, $0.3383]$ \\
 & $\theta(q_{50}) = \alpha_A \beta_M(q_{50})$: path product at the median of waist-to-hip ratio & $-0.001852$ & $[-0.2854$, $0.2816]$ \\
 & $\theta(q_{75}) = \alpha_A \beta_M(q_{75})$: path product at the 75th percentile of waist-to-hip ratio & $0.01126$ & $[-0.4548$, $0.4773]$ \\
Path components & $\alpha_A$: waist-to-hip ratio $\to$ LDL cholesterol slope ($A \to M$) & $-20.55$ & $[-86.85$, $68.7]$ \\
 & $\beta_M(q_{25})$: conditional LDL $\to$ PWV slope at the 25th percentile of waist-to-hip ratio ($M \to Y \mid A$) & $0.0006969$ & $[-0.004273$, $0.005451]$ \\
 & $\beta_M(q_{50})$: conditional LDL $\to$ PWV slope at the median of waist-to-hip ratio ($M \to Y \mid A$) & $9.015\times10^{-5}$ & $[-0.004473$, $0.003375]$ \\
 & $\beta_M(q_{75})$: conditional LDL $\to$ PWV slope at the 75th percentile of waist-to-hip ratio ($M \to Y \mid A$) & $-0.0005478$ & $[-0.007203$, $0.002018]$ \\
 & $\beta_{AM}$: waist-to-hip ratio $\times$ LDL interaction coefficient in the PWV equation & $-0.01127$ & $[-0.08272$, $0.01296]$ \\
\bottomrule
\end{longtable}}

{\scriptsize\setlength{\tabcolsep}{3pt}
\begin{longtable}{@{}lcc@{}}
\toprule
Partial $R^2$ of U (regressor; outcome) & Omitted-confounding for $\alpha_A$ & Omitted-confounding for $\beta_M(q_{50})$ \\
\midrule\endhead
Coefficient & $-20.55$ & $9.015\times10^{-5}$ \\
0.01; 0.01 & $[-26.43, -14.66]$ & $[-0.0003009, 0.0004812]$ \\
0.01; 0.05 & $[-33.71, -7.389]$ & $[-0.0007843, 0.0009646]$ \\
0.01; 0.10 & $[-39.16, -1.938]$ & $[-0.001147, 0.001327]$ \\
0.01; 0.20 & $[-46.87, 5.771]$ & $[-0.001659, 0.001839]$ \\
0.01; 0.30 & $[-52.78, 11.69]$ & $[-0.002052, 0.002232]$ \\
0.05; 0.01 & $[-33.98, -7.115]$ & $[-0.0008025, 0.0009828]$ \\
0.05; 0.05 & $[-50.59, 9.49]$ & $[-0.001906, 0.002086]$ \\
0.05; 0.10 & $[-63.03, 21.93]$ & $[-0.002733, 0.002913]$ \\
0.05; 0.20 & $[-80.63, 39.53]$ & $[-0.003902, 0.004082]$ \\
0.05; 0.30 & $[-94.13, 53.03]$ & $[-0.004799, 0.00498]$ \\
0.10; 0.01 & $[-40.07, -1.03]$ & $[-0.001207, 0.001387]$ \\
0.10; 0.05 & $[-64.19, 23.1]$ & $[-0.00281, 0.00299]$ \\
0.10; 0.10 & $[-82.27, 41.17]$ & $[-0.004011, 0.004192]$ \\
0.10; 0.20 & $[-107.8, 66.74]$ & $[-0.00571, 0.005891]$ \\
0.10; 0.30 & $[-127.5, 86.36]$ & $[-0.007014, 0.007194]$ \\
0.20; 0.01 & $[-49.83, 8.73]$ & $[-0.001855, 0.002036]$ \\
0.20; 0.05 & $[-86.02, 44.92]$ & $[-0.00426, 0.004441]$ \\
0.20; 0.10 & $[-113.1, 72.04]$ & $[-0.006062, 0.006243]$ \\
0.20; 0.20 & $[-151.5, 110.4]$ & $[-0.008611, 0.008791]$ \\
0.20; 0.30 & $[-180.9, 139.8]$ & $[-0.01057, 0.01075]$ \\
0.30; 0.01 & $[-58.88, 17.79]$ & $[-0.002457, 0.002637]$ \\
0.30; 0.05 & $[-106.3, 65.17]$ & $[-0.005606, 0.005786]$ \\
0.30; 0.10 & $[-141.8, 100.7]$ & $[-0.007965, 0.008146]$ \\
0.30; 0.20 & $[-192, 150.9]$ & $[-0.0113, 0.01148]$ \\
0.30; 0.30 & $[-230.5, 189.4]$ & $[-0.01386, 0.01404]$ \\
\bottomrule
\end{longtable}}

\paragraph{2.\ Main serial path: waist-to-hip ratio $\to$ LDL $\to$ diastolic blood pressure $\to$ PWV}
\textit{Design.} Waist-to-hip ratio, LDL cholesterol, seated diastolic blood pressure and PWV measured in the same period: LDL within 365 days before or after waist-to-hip ratio, diastolic blood pressure within 90 days before or after LDL, and PWV within 365 days before or after diastolic blood pressure (endpoints inclusive); one set of measurements per person. Adjustment for approximate age (with a squared term), sex, study membership, and the most recent smoking status on the earliest measurement day or within the preceding 730 days; missing age multiply imputed, missing categorical covariates as a separate category. The four measurements standardized within the analysis sample.

\textit{Model.} $L = \alpha_1 + \beta_{AL} A + \gamma_1^\top C + \varepsilon_1$; $D = \alpha_2 + \beta_{AD} A + \beta_{LD} L + \gamma_2^\top C + \varepsilon_2$; $Y = \alpha_3 + \beta_{AY} A + \beta_{LY} L + \beta_{DY} D + \gamma_3^\top C + \varepsilon_3$. All three equations are linear regressions, with $A$, $L$, $D$ and $Y$ the standardized waist-to-hip ratio, LDL cholesterol, seated diastolic blood pressure (mean of the available values of two readings) and thigh-to-ankle PWV (mean of the available sides), and $C$ the covariates. $\beta_{AL}$ is the SD difference in LDL per 1 SD higher waist-to-hip ratio; $\beta_{LD}$ is the standardized slope of diastolic blood pressure on LDL adjusted for waist-to-hip ratio; $\beta_{DY}$ is the standardized slope of PWV on diastolic blood pressure adjusted for waist-to-hip ratio and LDL; the serial path product is $\theta = \beta_{AL} \beta_{LD} \beta_{DY}$.

\textit{Intervals.} $\beta_{AL}$, $\beta_{LD}$, $\beta_{DY}$: 95\% confidence intervals; $\theta$: 95\% percentile interval from joint normal simulation of the coefficients. $P$ values not corrected for multiple comparisons.
{\footnotesize\setlength{\tabcolsep}{3pt}
\begin{longtable}{@{}>{\raggedright\arraybackslash}p{0.14\linewidth}>{\raggedright\arraybackslash}p{0.29\linewidth}>{\raggedleft\arraybackslash}p{0.13\linewidth}>{\raggedright\arraybackslash}p{0.22\linewidth}>{\raggedleft\arraybackslash}p{0.14\linewidth}@{}}
\toprule
Block & Parameter & Estimate & Interval & $P$ \\
\midrule\endhead
Path components & $\beta_{AL}$: standardized waist-to-hip ratio $\to$ LDL cholesterol slope ($A \to L$) & $0.06237$ & $[0.007262$, $0.1175]$ & \mbox{$0.02654$} \\
 & $\beta_{LD}$: standardized LDL $\to$ seated diastolic blood pressure slope adjusted for waist-to-hip ratio ($L \to D \mid A$) & $0.08058$ & $[0.04736$, $0.1138]$ & \mbox{$1.993\times10^{-6}$} \\
 & $\beta_{DY}$: standardized diastolic blood pressure $\to$ PWV slope adjusted for waist-to-hip ratio and LDL ($D \to Y \mid A, L$) & $0.4273$ & $[0.3909$, $0.4637]$ & \mbox{$2.494\times10^{-117}$} \\
Path index & $\theta = \beta_{AL} \beta_{LD} \beta_{DY}$: serial path product & $0.002147$ & $[0.0002414$, $0.00442]$ &  \\
\bottomrule
\end{longtable}}

\paragraph{3.\ Exposure--outcome association: waist-to-hip ratio and thigh-to-ankle PWV ($\pm$365 days)}
\textit{Design.} Waist-to-hip ratio and PWV measured in the same period: each PWV measurement day as anchor, with the waist-to-hip ratio closest in date within 365 days before or after it (endpoints inclusive); one row per PWV measurement day, with possibly multiple rows per participant. PWV as the bilateral mean when both left and right sides are valid. Adjustment for approximate age (PWV measurement year minus birth year), sex and recruitment study category; all variables kept on the original recorded scale.

\textit{Model.} $Y = \beta_0 + \beta_A A + \beta_{\mathrm{age}}\,\mathrm{age} + \gamma_{\mathrm{sex}} + \delta_{\mathrm{study}} + \varepsilon$, linear regression; $A$ is waist-to-hip ratio, $Y$ the bilateral thigh-to-ankle PWV, and $\gamma_{\mathrm{sex}}$ and $\delta_{\mathrm{study}}$ the categorical effects of sex and recruitment study. Reported: $0.1\beta_A$, the PWV difference per 0.1 higher waist-to-hip ratio at fixed age, sex and recruitment study (original scale).

\textit{Intervals.} 95\% confidence interval.
{\footnotesize\setlength{\tabcolsep}{3pt}
\begin{longtable}{@{}>{\raggedright\arraybackslash}p{0.14\linewidth}>{\raggedright\arraybackslash}p{0.29\linewidth}>{\raggedleft\arraybackslash}p{0.13\linewidth}>{\raggedright\arraybackslash}p{0.22\linewidth}>{\raggedleft\arraybackslash}p{0.14\linewidth}@{}}
\toprule
Block & Parameter & Estimate & Interval & $P$ \\
\midrule\endhead
Exposure--outcome association & $0.1\beta_A$: PWV difference per 0.1 higher waist-to-hip ratio ($A \to Y$) & $0.3471$ & $[0.3016$, $0.3927]$ & \mbox{$4.98\times10^{-48}$} \\
\bottomrule
\end{longtable}}

\subsubsection*{Adjacent segments and pairwise associations}

\paragraph{4.\ Serial segment: waist-to-hip ratio $\to$ LDL cholesterol $\to$ seated diastolic blood pressure}
\textit{Design.} Waist-to-hip ratio, LDL cholesterol and seated diastolic blood pressure measured in strict date order: LDL 30--730 days after waist-to-hip ratio and blood pressure 1--180 days after LDL (endpoints inclusive); one measurement chain per participant. Adjustment for approximate age at waist-to-hip ratio measurement and its square, sex, sub-study membership, same-day height, and the most recent smoking status on the same day or within the preceding 730 days. Diastolic blood pressure as the mean of the available values of two seated readings. All variables on the original recorded scale, unstandardized.

\textit{Model.} $M = \alpha_0 + \alpha_A A + \alpha_C^\top C + \varepsilon_M$; $Y = \beta_0 + \beta_A A + \beta_M M + \beta_{AM} A M + \beta_C^\top C + \varepsilon_Y$. Both equations are linear regressions, with $A$, $M$ and $Y$ the waist-to-hip ratio, LDL cholesterol and seated diastolic blood pressure, and $C$ the covariates. $\alpha_A$ is the change in LDL per 1 recorded unit higher waist-to-hip ratio; $\beta_M(q) = \beta_M + \beta_{AM} q$ is the conditional slope of diastolic blood pressure on LDL at waist-to-hip ratio $q$, and $\beta_{AM}$ is the interaction coefficient; the segment path product is $\theta(q) = \alpha_A \beta_M(q)$, with $q$ at the 25th, 50th and 75th percentiles of waist-to-hip ratio in the analysis sample ($q_{25}$, $q_{50}$, $q_{75}$).

\textit{Intervals.} $\alpha_A$, $\beta_M(q)$, $\beta_{AM}$: bootstrap 95\% percentile intervals; $\theta(q)$: bootstrap 95\% simultaneous confidence band (joint over the three percentile points).
{\footnotesize\setlength{\tabcolsep}{3pt}
\begin{longtable}{@{}>{\raggedright\arraybackslash}p{0.15\linewidth}>{\raggedright\arraybackslash}p{0.40\linewidth}>{\raggedleft\arraybackslash}p{0.14\linewidth}>{\raggedright\arraybackslash}p{0.24\linewidth}@{}}
\toprule
Block & Parameter & Estimate & Interval \\
\midrule\endhead
Segment path-product curve & $\theta(q_{25}) = \alpha_A \beta_M(q_{25})$: segment path product at the 25th percentile of waist-to-hip ratio & $-0.6658$ & $[-2.983$, $1.651]$ \\
 & $\theta(q_{50}) = \alpha_A \beta_M(q_{50})$: segment path product at the median of waist-to-hip ratio & $-0.646$ & $[-2.8$, $1.508]$ \\
 & $\theta(q_{75}) = \alpha_A \beta_M(q_{75})$: segment path product at the 75th percentile of waist-to-hip ratio & $-0.6288$ & $[-2.908$, $1.651]$ \\
Segment path components & $\alpha_A$: waist-to-hip ratio $\to$ LDL cholesterol slope ($A \to M$) & $-36.73$ & $[-109.8$, $41.39]$ \\
 & $\beta_M(q_{25})$: conditional LDL $\to$ diastolic blood pressure slope at the 25th percentile of waist-to-hip ratio ($M \to Y \mid A$) & $0.01813$ & $[-0.0116$, $0.04322]$ \\
 & $\beta_M(q_{50})$: conditional LDL $\to$ diastolic blood pressure slope at the median of waist-to-hip ratio ($M \to Y \mid A$) & $0.01759$ & $[-0.01039$, $0.04006]$ \\
 & $\beta_M(q_{75})$: conditional LDL $\to$ diastolic blood pressure slope at the 75th percentile of waist-to-hip ratio ($M \to Y \mid A$) & $0.01712$ & $[-0.01592$, $0.04232]$ \\
 & $\beta_{AM}$: waist-to-hip ratio $\times$ LDL interaction coefficient in the diastolic blood pressure equation & $-0.008463$ & $[-0.2424$, $0.2112]$ \\
\bottomrule
\end{longtable}}

\paragraph{5.\ Pairwise association: waist-to-hip ratio--LDL cholesterol ($\pm$365 days; one pair per participant)}
\textit{Design.} Waist-to-hip ratio and LDL cholesterol measured in the same period (LDL within 365 days before or after waist-to-hip ratio, endpoints inclusive); one pair of measurements per person. Adjustment for approximate age (by the year of waist-to-hip ratio measurement, with a squared term), sex, study membership, and the most recent smoking status on the earlier measurement day or within the preceding 730 days; missing age multiply imputed, missing categorical covariates as a separate category. Both measurements standardized within the analysis sample.

\textit{Model.} $L = \alpha + \beta A + \gamma^\top C + \varepsilon$ (linear regression; $A$ and $L$ the standardized waist-to-hip ratio and LDL cholesterol, $C$ the covariates). $\beta$ is the SD difference in LDL cholesterol per 1 SD higher waist-to-hip ratio.

\textit{Intervals.} 95\% confidence interval.
{\footnotesize\setlength{\tabcolsep}{3pt}
\begin{longtable}{@{}>{\raggedright\arraybackslash}p{0.14\linewidth}>{\raggedright\arraybackslash}p{0.29\linewidth}>{\raggedleft\arraybackslash}p{0.13\linewidth}>{\raggedright\arraybackslash}p{0.22\linewidth}>{\raggedleft\arraybackslash}p{0.14\linewidth}@{}}
\toprule
Block & Parameter & Estimate & Interval & $P$ \\
\midrule\endhead
$\pm$365-day window & $\beta$: adjusted standardized waist-to-hip ratio $\to$ LDL cholesterol slope & $0.05173$ & $[0.0187$, $0.08475]$ & \mbox{$0.002143$} \\
\bottomrule
\end{longtable}}

\paragraph{6.\ Pairwise association: waist-to-hip ratio--seated diastolic blood pressure ($\pm$90 days)}
\textit{Design.} Waist-to-hip ratio and seated diastolic blood pressure measured in the same period (diastolic blood pressure within 90 days before or after waist-to-hip ratio, endpoints inclusive); one pair of measurements per person. Adjustment for approximate age (by the year of waist-to-hip ratio measurement, with a squared term), sex, study membership, and the most recent smoking status on the earlier measurement day or within the preceding 730 days; missing age multiply imputed, missing categorical covariates as a separate category. Both measurements standardized within the analysis sample.

\textit{Model.} $D = \alpha + \beta A + \gamma^\top C + \varepsilon$ (linear regression; $A$ and $D$ the standardized waist-to-hip ratio and seated diastolic blood pressure, $C$ the covariates). $\beta$ is the SD difference in diastolic blood pressure per 1 SD higher waist-to-hip ratio.

\textit{Intervals.} 95\% confidence interval.
{\footnotesize\setlength{\tabcolsep}{3pt}
\begin{longtable}{@{}>{\raggedright\arraybackslash}p{0.14\linewidth}>{\raggedright\arraybackslash}p{0.29\linewidth}>{\raggedleft\arraybackslash}p{0.13\linewidth}>{\raggedright\arraybackslash}p{0.22\linewidth}>{\raggedleft\arraybackslash}p{0.14\linewidth}@{}}
\toprule
Block & Parameter & Estimate & Interval & $P$ \\
\midrule\endhead
$\pm$90-day window & $\beta$: adjusted standardized waist-to-hip ratio $\to$ seated diastolic blood pressure slope & $0.2558$ & $[0.2317$, $0.28]$ & \mbox{$8.85\times10^{-96}$} \\
\bottomrule
\end{longtable}}

\paragraph{7.\ Pairwise association: LDL cholesterol--PWV ($\pm$365 days; one pair per participant)}
\textit{Design.} LDL cholesterol and thigh-to-ankle PWV measured in the same period (PWV within 365 days before or after LDL, endpoints inclusive); one pair of measurements per person, $n = 5690$. Adjustment for approximate age (by the year of LDL measurement, with a squared term), sex, study membership, and the most recent smoking status on the earlier measurement day or within the preceding 730 days; missing age multiply imputed, missing categorical covariates as a separate category. Both measurements standardized within the analysis sample.

\textit{Model.} $Y = \alpha + \beta L + \gamma^\top C + \varepsilon$ (linear regression; $L$ and $Y$ the standardized LDL cholesterol and thigh-to-ankle PWV, $C$ the covariates). $\beta$ is the SD difference in PWV per 1 SD higher LDL cholesterol.

\textit{Intervals.} 95\% confidence interval.
{\footnotesize\setlength{\tabcolsep}{3pt}
\begin{longtable}{@{}>{\raggedright\arraybackslash}p{0.14\linewidth}>{\raggedright\arraybackslash}p{0.29\linewidth}>{\raggedleft\arraybackslash}p{0.13\linewidth}>{\raggedright\arraybackslash}p{0.22\linewidth}>{\raggedleft\arraybackslash}p{0.14\linewidth}@{}}
\toprule
Block & Parameter & Estimate & Interval & $P$ \\
\midrule\endhead
$\pm$365-day window & $\beta$: adjusted standardized LDL cholesterol $\to$ thigh-to-ankle PWV slope & $0.05213$ & $[0.02795$, $0.07632]$ & \mbox{$2.389\times10^{-5}$} \\
\bottomrule
\end{longtable}}

\paragraph{8.\ Pairwise association: seated diastolic blood pressure--PWV ($\pm$90 days)}
\textit{Design.} Seated diastolic blood pressure and thigh-to-ankle PWV measured in the same period (PWV within 90 days before or after diastolic blood pressure, endpoints inclusive); one pair of measurements per person. Adjustment for approximate age (by the year of blood pressure measurement, with a squared term), sex, study membership, and the most recent smoking status on the earlier measurement day or within the preceding 730 days; missing age multiply imputed, missing categorical covariates as a separate category. Both measurements standardized within the analysis sample.

\textit{Model.} $Y = \alpha + \beta D + \gamma^\top C + \varepsilon$ (linear regression; $D$ and $Y$ the standardized seated diastolic blood pressure and thigh-to-ankle PWV, $C$ the covariates). $\beta$ is the SD difference in PWV per 1 SD higher diastolic blood pressure.

\textit{Intervals.} 95\% confidence interval.
{\footnotesize\setlength{\tabcolsep}{3pt}
\begin{longtable}{@{}>{\raggedright\arraybackslash}p{0.14\linewidth}>{\raggedright\arraybackslash}p{0.29\linewidth}>{\raggedleft\arraybackslash}p{0.13\linewidth}>{\raggedright\arraybackslash}p{0.22\linewidth}>{\raggedleft\arraybackslash}p{0.14\linewidth}@{}}
\toprule
Block & Parameter & Estimate & Interval & $P$ \\
\midrule\endhead
$\pm$90-day window & $\beta$: adjusted standardized seated diastolic blood pressure $\to$ thigh-to-ankle PWV slope & $0.4441$ & $[0.4253$, $0.4629]$ & \mbox{$0$} \\
\bottomrule
\end{longtable}}

\paragraph{9.\ Pairwise total association: waist-to-hip ratio--PWV ($\pm$365 days)}
\textit{Design.} Waist-to-hip ratio and thigh-to-ankle PWV measured in the same period (PWV within 365 days before or after waist-to-hip ratio, endpoints inclusive); one pair of measurements per person, $n = 9385$. Adjustment for approximate age (by the year of waist-to-hip ratio measurement, with a squared term), sex, study membership, and the most recent smoking status on the earlier measurement day or within the preceding 730 days; missing age multiply imputed, missing categorical covariates as a separate category. Both measurements standardized within the analysis sample.

\textit{Model.} $Y = \alpha + \beta A + \gamma^\top C + \varepsilon$ (linear regression; $A$ and $Y$ the standardized waist-to-hip ratio and thigh-to-ankle PWV, $C$ the covariates). $\beta$ is the SD difference in PWV per 1 SD higher waist-to-hip ratio.

\textit{Intervals.} 95\% confidence interval.
{\footnotesize\setlength{\tabcolsep}{3pt}
\begin{longtable}{@{}>{\raggedright\arraybackslash}p{0.14\linewidth}>{\raggedright\arraybackslash}p{0.29\linewidth}>{\raggedleft\arraybackslash}p{0.13\linewidth}>{\raggedright\arraybackslash}p{0.22\linewidth}>{\raggedleft\arraybackslash}p{0.14\linewidth}@{}}
\toprule
Block & Parameter & Estimate & Interval & $P$ \\
\midrule\endhead
$\pm$365-day window & $\beta$: adjusted standardized total-association slope, waist-to-hip ratio $\to$ thigh-to-ankle PWV & $0.1678$ & $[0.1438$, $0.1918]$ & \mbox{$8.378\times10^{-43}$} \\
\bottomrule
\end{longtable}}

\paragraph{10.\ Pairwise association: LDL cholesterol--seated diastolic blood pressure ($\pm$90 days)}
\textit{Design.} LDL cholesterol and seated diastolic blood pressure measured in the same period (diastolic blood pressure within 90 days before or after LDL, endpoints inclusive); one pair of measurements per person. Adjustment for approximate age (by the year of LDL measurement, with a squared term), sex, study membership, and the most recent smoking status on the earlier measurement day or within the preceding 730 days; missing age multiply imputed, missing categorical covariates as a separate category. Both measurements standardized within the analysis sample.

\textit{Model.} $D = \alpha + \beta L + \gamma^\top C + \varepsilon$ (linear regression; $L$ and $D$ the standardized LDL cholesterol and seated diastolic blood pressure, $C$ the covariates). $\beta$ is the SD difference in diastolic blood pressure per 1 SD higher LDL cholesterol.

\textit{Intervals.} 95\% confidence interval.
{\footnotesize\setlength{\tabcolsep}{3pt}
\begin{longtable}{@{}>{\raggedright\arraybackslash}p{0.14\linewidth}>{\raggedright\arraybackslash}p{0.29\linewidth}>{\raggedleft\arraybackslash}p{0.13\linewidth}>{\raggedright\arraybackslash}p{0.22\linewidth}>{\raggedleft\arraybackslash}p{0.14\linewidth}@{}}
\toprule
Block & Parameter & Estimate & Interval & $P$ \\
\midrule\endhead
$\pm$90-day window & $\beta$: adjusted standardized LDL cholesterol $\to$ seated diastolic blood pressure slope & $0.09179$ & $[0.06117$, $0.1224]$ & \mbox{$4.226\times10^{-9}$} \\
\bottomrule
\end{longtable}}

\paragraph{11.\ Pairwise association: waist-to-hip ratio--LDL cholesterol ($\pm$365 days; LDL-day anchors)}
\textit{Design.} Waist-to-hip ratio and LDL cholesterol measured in the same period: each LDL measurement day as anchor, with the waist-to-hip ratio closest in date within 365 days before or after it (endpoints inclusive); one row per LDL measurement day, with possibly multiple rows per participant. Adjustment for approximate age (LDL measurement year minus birth year), sex and recruitment study category; all variables kept on the original recorded scale.

\textit{Model.} $M_1 = \beta_0 + \beta_A A + \beta_{\mathrm{age}}\,\mathrm{age} + \gamma_{\mathrm{sex}} + \delta_{\mathrm{study}} + \varepsilon$, linear regression; $A$ is waist-to-hip ratio and $M_1$ LDL cholesterol. Reported: $0.1\beta_A$, the LDL difference per 0.1 higher waist-to-hip ratio at fixed age, sex and recruitment study (original scale).

\textit{Intervals.} 95\% confidence interval.
{\footnotesize\setlength{\tabcolsep}{3pt}
\begin{longtable}{@{}>{\raggedright\arraybackslash}p{0.14\linewidth}>{\raggedright\arraybackslash}p{0.29\linewidth}>{\raggedleft\arraybackslash}p{0.13\linewidth}>{\raggedright\arraybackslash}p{0.22\linewidth}>{\raggedleft\arraybackslash}p{0.14\linewidth}@{}}
\toprule
Block & Parameter & Estimate & Interval & $P$ \\
\midrule\endhead
Pairwise association & $0.1\beta_A$: LDL cholesterol difference per 0.1 higher waist-to-hip ratio ($A \to M_1$) & $0.6117$ & $[-0.6544$, $1.878]$ & \mbox{$0.3434$} \\
\bottomrule
\end{longtable}}

\paragraph{12.\ Pairwise association: waist-to-hip ratio--seated diastolic blood pressure ($\pm$365 days)}
\textit{Design.} Waist-to-hip ratio and seated diastolic blood pressure measured in the same period: each blood pressure measurement day as anchor, with the waist-to-hip ratio closest in date within 365 days before or after it (endpoints inclusive); one row per blood pressure measurement day, with possibly multiple rows per participant. Diastolic blood pressure as the mean of two seated readings. Adjustment for approximate age (blood pressure measurement year minus birth year), sex and recruitment study category; all variables kept on the original recorded scale.

\textit{Model.} $M_2 = \beta_0 + \beta_A A + \beta_{\mathrm{age}}\,\mathrm{age} + \gamma_{\mathrm{sex}} + \delta_{\mathrm{study}} + \varepsilon$, linear regression; $A$ is waist-to-hip ratio and $M_2$ seated diastolic blood pressure. Reported: $0.1\beta_A$, the diastolic blood pressure difference per 0.1 higher waist-to-hip ratio at fixed age, sex and recruitment study (original scale).

\textit{Intervals.} 95\% confidence interval.
{\footnotesize\setlength{\tabcolsep}{3pt}
\begin{longtable}{@{}>{\raggedright\arraybackslash}p{0.14\linewidth}>{\raggedright\arraybackslash}p{0.29\linewidth}>{\raggedleft\arraybackslash}p{0.13\linewidth}>{\raggedright\arraybackslash}p{0.22\linewidth}>{\raggedleft\arraybackslash}p{0.14\linewidth}@{}}
\toprule
Block & Parameter & Estimate & Interval & $P$ \\
\midrule\endhead
Pairwise association & $0.1\beta_A$: seated diastolic blood pressure difference per 0.1 higher waist-to-hip ratio ($A \to M_2$) & $3.134$ & $[2.635$, $3.634]$ & \mbox{$9.745\times10^{-32}$} \\
\bottomrule
\end{longtable}}

\paragraph{13.\ Pairwise association: LDL cholesterol--PWV ($\pm$365 days; PWV-day anchors)}
\textit{Design.} LDL cholesterol and PWV measured in the same period: each PWV measurement day as anchor, with the LDL closest in date within 365 days before or after it (endpoints inclusive); one row per PWV measurement day, with possibly multiple rows per participant. PWV as the bilateral mean when both left and right sides are valid. Adjustment for approximate age (PWV measurement year minus birth year), sex and recruitment study category; all variables kept on the original recorded scale.

\textit{Model.} $Y = \beta_0 + \beta_{M1} M_1 + \beta_{\mathrm{age}}\,\mathrm{age} + \gamma_{\mathrm{sex}} + \delta_{\mathrm{study}} + \varepsilon$, linear regression; $M_1$ is LDL cholesterol and $Y$ the bilateral thigh-to-ankle PWV. $\beta_{M1}$ is the PWV difference per 1 original unit higher LDL at fixed age, sex and recruitment study.

\textit{Intervals.} 95\% confidence interval.
{\footnotesize\setlength{\tabcolsep}{3pt}
\begin{longtable}{@{}>{\raggedright\arraybackslash}p{0.14\linewidth}>{\raggedright\arraybackslash}p{0.29\linewidth}>{\raggedleft\arraybackslash}p{0.13\linewidth}>{\raggedright\arraybackslash}p{0.22\linewidth}>{\raggedleft\arraybackslash}p{0.14\linewidth}@{}}
\toprule
Block & Parameter & Estimate & Interval & $P$ \\
\midrule\endhead
Pairwise association & $\beta_{M1}$: PWV difference per 1 original unit higher LDL cholesterol ($M_1 \to Y$) & $0.003064$ & $[0.001761$, $0.004367]$ & \mbox{$5.859\times10^{-6}$} \\
\bottomrule
\end{longtable}}

\paragraph{14.\ Pairwise association: seated diastolic blood pressure--PWV ($\pm$365 days)}
\textit{Design.} Seated diastolic blood pressure and PWV measured in the same period: each PWV measurement day as anchor, with the blood pressure closest in date within 365 days before or after it (endpoints inclusive); one row per PWV measurement day, with possibly multiple rows per participant. Diastolic blood pressure as the mean of two seated readings and PWV as the bilateral mean. Adjustment for approximate age (PWV measurement year minus birth year), sex and recruitment study category; all variables kept on the original recorded scale.

\textit{Model.} $Y = \beta_0 + \beta_{M2} M_2 + \beta_{\mathrm{age}}\,\mathrm{age} + \gamma_{\mathrm{sex}} + \delta_{\mathrm{study}} + \varepsilon$, linear regression; $M_2$ is seated diastolic blood pressure and $Y$ the bilateral thigh-to-ankle PWV. $\beta_{M2}$ is the PWV difference per 1 original unit higher diastolic blood pressure at fixed age, sex and recruitment study.

\textit{Intervals.} 95\% confidence interval.
{\footnotesize\setlength{\tabcolsep}{3pt}
\begin{longtable}{@{}>{\raggedright\arraybackslash}p{0.14\linewidth}>{\raggedright\arraybackslash}p{0.29\linewidth}>{\raggedleft\arraybackslash}p{0.13\linewidth}>{\raggedright\arraybackslash}p{0.22\linewidth}>{\raggedleft\arraybackslash}p{0.14\linewidth}@{}}
\toprule
Block & Parameter & Estimate & Interval & $P$ \\
\midrule\endhead
Pairwise association & $\beta_{M2}$: PWV difference per 1 original unit higher seated diastolic blood pressure ($M_2 \to Y$) & $0.08038$ & $[0.07529$, $0.08547]$ & \mbox{$3.381\times10^{-120}$} \\
\bottomrule
\end{longtable}}

\subsubsection*{Joint models}

\paragraph{15.\ Joint path: waist-to-hip ratio $\to$ LDL $\to$ diastolic blood pressure $\to$ PWV ($\pm$365 days)}
\textit{Design.} Waist-to-hip ratio, LDL cholesterol, seated diastolic blood pressure and PWV measured in the same period: each PWV measurement day as anchor, with each of the other three taken as the measurement closest in date within 365 days before or after it (endpoints inclusive), in any order; one row per PWV measurement day, with possibly multiple rows per participant. The three equations share one sample, with adjustment for approximate age (PWV measurement year minus birth year), sex and recruitment study category. Diastolic blood pressure as the mean of two seated readings and PWV as the bilateral mean; all variables kept on the original recorded scale.

\textit{Model.} $M_1 = \alpha_0 + \alpha_A A + \alpha_C^\top C + \varepsilon_1$; $M_2 = \gamma_0 + \gamma_A A + \gamma_{M1} M_1 + \gamma_C^\top C + \varepsilon_2$; $Y = \theta_0 + \theta_A A + \theta_{M1} M_1 + \theta_{M2} M_2 + \theta_C^\top C + \varepsilon_3$. All three equations are linear regressions, with $A$, $M_1$, $M_2$ and $Y$ the waist-to-hip ratio, LDL cholesterol, seated diastolic blood pressure and bilateral thigh-to-ankle PWV, and $C$ the covariates. Reported as outcome differences with the other terms of the same equation fixed: $0.1\alpha_A$, $0.1\gamma_A$ and $0.1\theta_A$ per 0.1 higher waist-to-hip ratio, and $\gamma_{M1}$, $\theta_{M1}$ and $\theta_{M2}$ per 1 original unit higher value of that variable (all on the original scale).

\textit{Intervals.} 95\% confidence intervals; $P$ values not corrected for multiple comparisons.
{\footnotesize\setlength{\tabcolsep}{3pt}
\begin{longtable}{@{}>{\raggedright\arraybackslash}p{0.14\linewidth}>{\raggedright\arraybackslash}p{0.29\linewidth}>{\raggedleft\arraybackslash}p{0.13\linewidth}>{\raggedright\arraybackslash}p{0.22\linewidth}>{\raggedleft\arraybackslash}p{0.14\linewidth}@{}}
\toprule
Block & Parameter & Estimate & Interval & $P$ \\
\midrule\endhead
LDL equation & $0.1\alpha_A$: waist-to-hip ratio $\to$ LDL cholesterol ($A \to M_1$) & $1.97$ & $[-1.538$, $5.478]$ & \mbox{$0.2695$} \\
Diastolic blood pressure equation & $0.1\gamma_A$: waist-to-hip ratio $\to$ seated diastolic blood pressure, adjusted for LDL ($A \to M_2 \mid M_1$) & $3.282$ & $[2.274$, $4.29]$ & \mbox{$9.034\times10^{-10}$} \\
 & $\gamma_{M1}$: LDL cholesterol $\to$ seated diastolic blood pressure, adjusted for waist-to-hip ratio ($M_1 \to M_2 \mid A$) & $0.01895$ & $[0.0005562$, $0.03735]$ & \mbox{$0.04362$} \\
PWV equation & $0.1\theta_A$: waist-to-hip ratio $\to$ PWV, adjusted for LDL and diastolic blood pressure ($A \to Y \mid M_1, M_2$) & $0.09884$ & $[-0.0151$, $0.2128]$ & \mbox{$0.08873$} \\
 & $\theta_{M1}$: LDL cholesterol $\to$ PWV, adjusted for waist-to-hip ratio and diastolic blood pressure ($M_1 \to Y \mid A, M_2$) & $0.0004654$ & $[-0.001966$, $0.002897]$ & \mbox{$0.7042$} \\
 & $\theta_{M2}$: seated diastolic blood pressure $\to$ PWV, adjusted for waist-to-hip ratio and LDL ($M_2 \to Y \mid A, M_1$) & $0.07667$ & $[0.06841$, $0.08492]$ & \mbox{$2.567\times10^{-44}$} \\
\bottomrule
\end{longtable}}

\subsubsection*{Reverse direction}

\paragraph{16.\ Reversed measurement order: PWV $\to$ LDL cholesterol $\to$ waist-to-hip ratio}
\textit{Design.} Reversed measurement order, with thigh-to-ankle PWV, LDL cholesterol and waist-to-hip ratio measured in strict date order: LDL 30--1095 days after PWV and waist-to-hip ratio 30--730 days after LDL (endpoints inclusive); one measurement chain per participant. Adjustment for approximate age at PWV measurement and its square, sex, sub-study membership, and the most recent height on the PWV measurement day or within the preceding 365 days and the most recent smoking status within 730 days. All variables on the original recorded scale.

\textit{Model.} $M = \alpha_0 + \alpha_A A + \alpha_C^\top C + \varepsilon_M$; $Y = \beta_0 + \beta_A A + \beta_M M + \beta_{AM} A M + \beta_C^\top C + \varepsilon_Y$. Both equations are linear regressions, with $A$ the thigh-to-ankle PWV, $M$ the LDL cholesterol, $Y$ the waist-to-hip ratio and $C$ the covariates. $\alpha_A$ is the change in LDL per 1 recorded unit higher PWV; $\beta_M(q) = \beta_M + \beta_{AM} q$ is the conditional slope of waist-to-hip ratio on LDL at PWV $q$, and $\beta_{AM}$ is the interaction coefficient; the reverse path product is $\theta(q) = \alpha_A \beta_M(q)$, with $q$ at the 25th, 50th and 75th percentiles of PWV in the analysis sample ($q_{25}$, $q_{50}$, $q_{75}$).

\textit{Intervals.} $\alpha_A$, $\beta_M(q)$, $\beta_{AM}$: bootstrap 95\% percentile intervals; $\theta(q)$: bootstrap 95\% simultaneous confidence band (joint over the three percentile points).
{\footnotesize\setlength{\tabcolsep}{3pt}
\begin{longtable}{@{}>{\raggedright\arraybackslash}p{0.15\linewidth}>{\raggedright\arraybackslash}p{0.40\linewidth}>{\raggedleft\arraybackslash}p{0.14\linewidth}>{\raggedright\arraybackslash}p{0.24\linewidth}@{}}
\toprule
Block & Parameter & Estimate & Interval \\
\midrule\endhead
Reverse path-product curve & $\theta(q_{25}) = \alpha_A \beta_M(q_{25})$: reverse path product at the 25th percentile of PWV & $8.423\times10^{-5}$ & $[-0.001015$, $0.001183]$ \\
 & $\theta(q_{50}) = \alpha_A \beta_M(q_{50})$: reverse path product at the median of PWV & $-0.0001484$ & $[-0.001$, $0.0007036]$ \\
 & $\theta(q_{75}) = \alpha_A \beta_M(q_{75})$: reverse path product at the 75th percentile of PWV & $-0.0004005$ & $[-0.001324$, $0.0005231]$ \\
Reverse path components & $\alpha_A$: PWV $\to$ LDL cholesterol slope ($A \to M$) & $2.526$ & $[-0.6467$, $6.831]$ \\
 & $\beta_M(q_{25})$: conditional LDL $\to$ waist-to-hip ratio slope at the 25th percentile of PWV ($M \to Y \mid A$) & $3.335\times10^{-5}$ & $[-0.0002058$, $0.0003012]$ \\
 & $\beta_M(q_{50})$: conditional LDL $\to$ waist-to-hip ratio slope at the median of PWV ($M \to Y \mid A$) & $-5.874\times10^{-5}$ & $[-0.0002458$, $0.0001378]$ \\
 & $\beta_M(q_{75})$: conditional LDL $\to$ waist-to-hip ratio slope at the 75th percentile of PWV ($M \to Y \mid A$) & $-0.0001586$ & $[-0.0003248$, $-1.928\times10^{-5}]$ \\
 & $\beta_{AM}$: PWV $\times$ LDL interaction coefficient in the waist-to-hip ratio equation & $-9.905\times10^{-5}$ & $[-0.000225$, $-1.461\times10^{-5}]$ \\
\bottomrule
\end{longtable}}

\paragraph{17.\ Reverse temporal order: PWV and seated diastolic blood pressure 1--365 days later}
\textit{Design.} PWV and seated diastolic blood pressure measured in strict date order, reversed relative to the main path: diastolic blood pressure 1--365 days after PWV (endpoints inclusive), taking the one closest to the PWV day; one row per PWV measurement day, with possibly multiple rows per participant. Diastolic blood pressure as the mean of two seated readings and PWV as the bilateral mean. Adjustment for approximate age (PWV measurement year minus birth year), sex and recruitment study category; all variables kept on the original recorded scale.

\textit{Model.} $Y(\mathrm{earlier}) = \beta_0 + \beta_{M2} M_2(\mathrm{later}) + \beta_{\mathrm{age}}\,\mathrm{age} + \gamma_{\mathrm{sex}} + \delta_{\mathrm{study}} + \varepsilon$, linear regression; $M_2$ is seated diastolic blood pressure and $Y$ the bilateral thigh-to-ankle PWV. $\beta_{M2}$ is the difference in earlier PWV per 1 original unit higher later-measured diastolic blood pressure at fixed age, sex and recruitment study.

\textit{Intervals.} 95\% confidence interval.
{\footnotesize\setlength{\tabcolsep}{3pt}
\begin{longtable}{@{}>{\raggedright\arraybackslash}p{0.14\linewidth}>{\raggedright\arraybackslash}p{0.29\linewidth}>{\raggedleft\arraybackslash}p{0.13\linewidth}>{\raggedright\arraybackslash}p{0.22\linewidth}>{\raggedleft\arraybackslash}p{0.14\linewidth}@{}}
\toprule
Block & Parameter & Estimate & Interval & $P$ \\
\midrule\endhead
Reverse temporal association & $\beta_{M2}$: difference in earlier PWV per 1 original unit higher later diastolic blood pressure (diastolic blood pressure 1--365 days after PWV) & $0.01285$ & $[-0.06697$, $0.09267]$ & \mbox{$0.5659$} \\
\bottomrule
\end{longtable}}

\subsubsection*{Heterogeneity and interaction}

\paragraph{18.\ Sex heterogeneity: waist-to-hip ratio $\to$ LDL cholesterol $\to$ thigh-to-ankle PWV}
\textit{Design.} LDL measured 30--730 days after waist-to-hip ratio and PWV 30--1095 days after LDL (strict order, endpoints inclusive); one measurement chain per participant. Sex (first and second sex categories) as a main effect and in interaction with waist-to-hip ratio, with additional interactions of sex with LDL and with waist-to-hip ratio $\times$ LDL in the PWV equation; further adjustment for approximate age at waist-to-hip ratio measurement and its square, sub-study membership, same-day height, and the most recent smoking status on the same day or within the preceding 730 days. All variables on the original recorded scale.

\textit{Model.} $M = \alpha_0 + \sum_s \mathbf{1}(\mathrm{sex}=s)\,\alpha_A(s) A + f(\mathrm{sex}) + \alpha_C^\top C + \varepsilon_M$; $Y = \beta_0 + \sum_s \mathbf{1}(\mathrm{sex}=s)\,[\beta_A(s) A + \beta_M(s) M + \beta_{AM}(s) A M] + f(\mathrm{sex}) + \beta_C^\top C + \varepsilon_Y$. Both equations are linear regressions, with $A$, $M$ and $Y$ the waist-to-hip ratio, LDL cholesterol and thigh-to-ankle PWV, $f(\mathrm{sex})$ the sex main effect and $C$ the remaining covariates. Reported within sex category $s$: $\alpha_A(s)$, $\beta_M(s)(q) = \beta_M(s) + \beta_{AM}(s) q$, $\beta_{AM}(s)$ and the path product $\theta_s(q) = \alpha_A(s) \beta_M(s)(q)$, with $q$ at the 25th, 50th and 75th percentiles of waist-to-hip ratio in the pooled sample ($q_{25}$, $q_{50}$, $q_{75}$).

\textit{Intervals.} $\alpha_A(s)$, $\beta_M(s)(q)$, $\beta_{AM}(s)$: bootstrap 95\% percentile intervals; $\theta_s(q)$: bootstrap 95\% simultaneous confidence band (joint over two sex categories $\times$ three percentile points).
{\footnotesize\setlength{\tabcolsep}{3pt}
\begin{longtable}{@{}>{\raggedright\arraybackslash}p{0.15\linewidth}>{\raggedright\arraybackslash}p{0.40\linewidth}>{\raggedleft\arraybackslash}p{0.14\linewidth}>{\raggedright\arraybackslash}p{0.24\linewidth}@{}}
\toprule
Block & Parameter & Estimate & Interval \\
\midrule\endhead
First sex category: path-product curve & $\theta_1(q_{25}) = \alpha_A(1) \beta_M(1)(q_{25})$: path product at the 25th percentile of waist-to-hip ratio & $-0.01927$ & $[-0.4271$, $0.3886]$ \\
 & $\theta_1(q_{50}) = \alpha_A(1) \beta_M(1)(q_{50})$: path product at the median of waist-to-hip ratio & $0.1576$ & $[-0.4084$, $0.7236]$ \\
 & $\theta_1(q_{75}) = \alpha_A(1) \beta_M(1)(q_{75})$: path product at the 75th percentile of waist-to-hip ratio & $0.3435$ & $[-0.6833$, $1.37]$ \\
Second sex category: path-product curve & $\theta_2(q_{25}) = \alpha_A(2) \beta_M(2)(q_{25})$: path product at the 25th percentile of waist-to-hip ratio & $-0.03152$ & $[-1.121$, $1.058]$ \\
 & $\theta_2(q_{50}) = \alpha_A(2) \beta_M(2)(q_{50})$: path product at the median of waist-to-hip ratio & $-0.02555$ & $[-0.9318$, $0.8807]$ \\
 & $\theta_2(q_{75}) = \alpha_A(2) \beta_M(2)(q_{75})$: path product at the 75th percentile of waist-to-hip ratio & $-0.01926$ & $[-0.808$, $0.7695]$ \\
First sex category: path components & $\alpha_A(1)$: waist-to-hip ratio $\to$ LDL cholesterol slope & $-41.46$ & $[-94.16$, $14.53]$ \\
 & $\beta_M(1)(q_{25})$: conditional LDL $\to$ PWV slope at the 25th percentile of waist-to-hip ratio & $0.0004648$ & $[-0.005007$, $0.005633]$ \\
 & $\beta_M(1)(q_{50})$: conditional LDL $\to$ PWV slope at the median of waist-to-hip ratio & $-0.003801$ & $[-0.00974$, $0.001949]$ \\
 & $\beta_M(1)(q_{75})$: conditional LDL $\to$ PWV slope at the 75th percentile of waist-to-hip ratio & $-0.008286$ & $[-0.01671$, $0.0005223]$ \\
 & $\beta_{AM}(1)$: waist-to-hip ratio $\times$ LDL interaction coefficient in the PWV equation & $-0.07924$ & $[-0.1428$, $0.005875]$ \\
Second sex category: path components & $\alpha_A(2)$: waist-to-hip ratio $\to$ LDL cholesterol slope & $9.668$ & $[-122.5$, $195.7]$ \\
 & $\beta_M(2)(q_{25})$: conditional LDL $\to$ PWV slope at the 25th percentile of waist-to-hip ratio & $-0.003261$ & $[-0.008865$, $0.004227]$ \\
 & $\beta_M(2)(q_{50})$: conditional LDL $\to$ PWV slope at the median of waist-to-hip ratio & $-0.002642$ & $[-0.007203$, $0.002195]$ \\
 & $\beta_M(2)(q_{75})$: conditional LDL $\to$ PWV slope at the 75th percentile of waist-to-hip ratio & $-0.001992$ & $[-0.006358$, $0.001432]$ \\
 & $\beta_{AM}(2)$: waist-to-hip ratio $\times$ LDL interaction coefficient in the PWV equation & $0.01149$ & $[-0.06074$, $0.03596]$ \\
\bottomrule
\end{longtable}}

\paragraph{19.\ Sex heterogeneity: waist-to-hip ratio $\to$ LDL $\to$ diastolic blood pressure $\to$ PWV}
\textit{Design.} Time windows as in the main serial path: LDL within 365 days before or after waist-to-hip ratio, diastolic blood pressure within 90 days before or after LDL, and PWV within 365 days before or after diastolic blood pressure (endpoints inclusive); one set of measurements per person, persons with missing sex not included. Slopes of the three paths and the intercept of each equation estimated separately by sex category, with shared coefficients for age (with a squared term), study membership and smoking status; missing age multiply imputed, missing categorical covariates as a separate category. The four measurements standardized within the analysis sample.

\textit{Model.} $L = \alpha_1(s) + \beta_{AL}(s) A + \gamma_1^\top C' + \varepsilon_1$; $D = \alpha_2(s) + \beta_{AD} A + \beta_{LD}(s) L + \gamma_2^\top C' + \varepsilon_2$; $Y = \alpha_3(s) + \beta_{AY} A + \beta_{LY} L + \beta_{DY}(s) D + \gamma_3^\top C' + \varepsilon_3$. Pooled-sample linear regressions, with $s$ the sex category, $A$, $L$, $D$ and $Y$ the standardized waist-to-hip ratio, LDL cholesterol, seated diastolic blood pressure and thigh-to-ankle PWV, and $C'$ the covariates excluding sex. Reported: the within-category slopes $\beta_{AL}(s)$, $\beta_{LD}(s)$, $\beta_{DY}(s)$ and the category-specific serial path product $\theta(s) = \beta_{AL}(s) \beta_{LD}(s) \beta_{DY}(s)$.

\textit{Intervals.} $\beta_{AL}(s)$, $\beta_{LD}(s)$, $\beta_{DY}(s)$: 95\% confidence intervals; $\theta(s)$: 95\% percentile interval from joint normal simulation of the coefficients. $P$ values not corrected for multiple comparisons.
{\footnotesize\setlength{\tabcolsep}{3pt}
\begin{longtable}{@{}>{\raggedright\arraybackslash}p{0.14\linewidth}>{\raggedright\arraybackslash}p{0.29\linewidth}>{\raggedleft\arraybackslash}p{0.13\linewidth}>{\raggedright\arraybackslash}p{0.22\linewidth}>{\raggedleft\arraybackslash}p{0.14\linewidth}@{}}
\toprule
Block & Parameter & Estimate & Interval & $P$ \\
\midrule\endhead
First sex category & $\beta_{AL}(s)$: category-specific waist-to-hip ratio $\to$ LDL cholesterol slope & $0.1281$ & $[0.07075$, $0.1855]$ & \mbox{$1.206\times10^{-5}$} \\
 & $\beta_{LD}(s)$: category-specific LDL $\to$ seated diastolic blood pressure slope adjusted for waist-to-hip ratio & $0.05017$ & $[-0.00422$, $0.1046]$ & \mbox{$0.07061$} \\
 & $\beta_{DY}(s)$: category-specific diastolic blood pressure $\to$ PWV slope adjusted for waist-to-hip ratio and LDL & $0.4437$ & $[0.3947$, $0.4926]$ & \mbox{$1.338\times10^{-70}$} \\
 & $\theta(s) = \beta_{AL}(s) \beta_{LD}(s) \beta_{DY}(s)$: category-specific serial path product & $0.002852$ & $[-0.0002408$, $0.006532]$ &  \\
Second sex category & $\beta_{AL}(s)$: category-specific waist-to-hip ratio $\to$ LDL cholesterol slope & $-0.009232$ & $[-0.1008$, $0.08234]$ & \mbox{$0.8434$} \\
 & $\beta_{LD}(s)$: category-specific LDL $\to$ seated diastolic blood pressure slope adjusted for waist-to-hip ratio & $0.1037$ & $[0.06022$, $0.1472]$ & \mbox{$2.976\times10^{-6}$} \\
 & $\beta_{DY}(s)$: category-specific diastolic blood pressure $\to$ PWV slope adjusted for waist-to-hip ratio and LDL & $0.4116$ & $[0.3602$, $0.4631]$ & \mbox{$1.861\times10^{-55}$} \\
 & $\theta(s) = \beta_{AL}(s) \beta_{LD}(s) \beta_{DY}(s)$: category-specific serial path product & $-0.0003942$ & $[-0.004827$, $0.003392]$ &  \\
\bottomrule
\end{longtable}}

\paragraph{20.\ Sex heterogeneity: sex-specific waist-to-hip ratio--PWV slopes ($\pm$365 days)}
\textit{Design.} Waist-to-hip ratio and PWV measured in the same period: each PWV measurement day as anchor, with the waist-to-hip ratio closest in date within 365 days before or after it (endpoints inclusive); one row per PWV measurement day, with possibly multiple rows per participant. Waist-to-hip ratio slopes estimated separately for the two sex categories in the analysis sample, with shared adjustment terms for approximate age (PWV measurement year minus birth year) and recruitment study category; all variables kept on the original recorded scale.

\textit{Model.} $Y = \eta_0 + \eta_{A,s} A + \eta_{\mathrm{age}}\,\mathrm{age} + \gamma_s + \delta_{\mathrm{study}} + \varepsilon$, linear regression; $s$ is the sex category, the waist-to-hip ratio slope $\eta_{A,s}$ varies by category, $A$ is waist-to-hip ratio and $Y$ the bilateral thigh-to-ankle PWV. Reported: $0.1\eta_{A,s}$, the PWV difference per 0.1 higher waist-to-hip ratio within sex category $s$ at fixed age and recruitment study (original scale).

\textit{Intervals.} 95\% confidence interval.
{\footnotesize\setlength{\tabcolsep}{3pt}
\begin{longtable}{@{}>{\raggedright\arraybackslash}p{0.14\linewidth}>{\raggedright\arraybackslash}p{0.29\linewidth}>{\raggedleft\arraybackslash}p{0.13\linewidth}>{\raggedright\arraybackslash}p{0.22\linewidth}>{\raggedleft\arraybackslash}p{0.14\linewidth}@{}}
\toprule
Block & Parameter & Estimate & Interval & $P$ \\
\midrule\endhead
First sex category & $0.1\eta_{A,1}$: PWV difference per 0.1 higher waist-to-hip ratio within the first sex category & $0.3715$ & $[0.3117$, $0.4314]$ & \mbox{$4.437\times10^{-33}$} \\
Second sex category & $0.1\eta_{A,2}$: PWV difference per 0.1 higher waist-to-hip ratio within the second sex category & $0.3177$ & $[0.2514$, $0.3841]$ & \mbox{$1.477\times10^{-20}$} \\
\bottomrule
\end{longtable}}

\paragraph{21.\ Age interaction: change in the waist-to-hip ratio--PWV slope with age ($\pm$365 days)}
\textit{Design.} Waist-to-hip ratio and PWV measured in the same period: each PWV measurement day as anchor, with the waist-to-hip ratio closest in date within 365 days before or after it (endpoints inclusive); one row per PWV measurement day, with possibly multiple rows per participant. Approximate age (PWV measurement year minus birth year) centered within the analysis sample and entered in continuous form as both a main effect and a modifier, with additional adjustment for sex and recruitment study category; other variables kept on the original recorded scale.

\textit{Model.} $Y = \kappa_0 + \kappa_A A + \kappa_{\mathrm{age}}\,\mathrm{age}_c + \kappa_{A\times\mathrm{age}} A\,\mathrm{age}_c + \gamma_{\mathrm{sex}} + \delta_{\mathrm{study}} + \varepsilon$, linear regression; $A$ is waist-to-hip ratio, $Y$ the bilateral thigh-to-ankle PWV and $\mathrm{age}_c$ the centered approximate age. Reported: $0.1\kappa_{A\times\mathrm{age}}$, the change, per 1 year higher approximate age, in the PWV difference per 0.1 higher waist-to-hip ratio (original scale).

\textit{Intervals.} 95\% confidence interval.
{\footnotesize\setlength{\tabcolsep}{3pt}
\begin{longtable}{@{}>{\raggedright\arraybackslash}p{0.14\linewidth}>{\raggedright\arraybackslash}p{0.29\linewidth}>{\raggedleft\arraybackslash}p{0.13\linewidth}>{\raggedright\arraybackslash}p{0.22\linewidth}>{\raggedleft\arraybackslash}p{0.14\linewidth}@{}}
\toprule
Block & Parameter & Estimate & Interval & $P$ \\
\midrule\endhead
Age interaction & $0.1\kappa_{A\times\mathrm{age}}$: waist-to-hip ratio $\times$ age interaction (change, per 1 year higher age, in the PWV difference per 0.1 higher waist-to-hip ratio) & $-0.01151$ & $[-0.01595$, $-0.00708]$ & \mbox{$4.065\times10^{-7}$} \\
\bottomrule
\end{longtable}}

\subsubsection*{Sensitivity analyses}

\paragraph{22.\ Narrower time windows: waist-to-hip ratio $\to$ LDL cholesterol $\to$ thigh-to-ankle PWV}
\textit{Design.} Waist-to-hip ratio, LDL cholesterol and thigh-to-ankle PWV measured in strict date order, with time windows narrowed relative to the main path: LDL 90--365 days after waist-to-hip ratio and PWV 90--730 days after LDL (endpoints inclusive); one measurement chain per participant. Adjustment for approximate age at waist-to-hip ratio measurement and its square, sex, sub-study membership, same-day height, and the most recent smoking status on the same day or within the preceding 730 days. PWV as the mean of the available sides. All variables on the original recorded scale.

\textit{Model.} $M = \alpha_0 + \alpha_A A + \alpha_C^\top C + \varepsilon_M$; $Y = \beta_0 + \beta_A A + \beta_M M + \beta_{AM} A M + \beta_C^\top C + \varepsilon_Y$. Both equations are linear regressions, with $A$, $M$ and $Y$ the waist-to-hip ratio, LDL cholesterol and thigh-to-ankle PWV, and $C$ the covariates. $\alpha_A$ is the change in LDL per 1 recorded unit higher waist-to-hip ratio; $\beta_M(q) = \beta_M + \beta_{AM} q$ is the conditional slope of PWV on LDL at waist-to-hip ratio $q$, with $q$ at the 25th, 50th and 75th percentiles of waist-to-hip ratio in the analysis sample ($q_{25}$, $q_{50}$, $q_{75}$); $\beta_{AM}$ is the interaction coefficient.

\textit{Intervals.} Bootstrap 95\% percentile intervals.
{\footnotesize\setlength{\tabcolsep}{3pt}
\begin{longtable}{@{}>{\raggedright\arraybackslash}p{0.15\linewidth}>{\raggedright\arraybackslash}p{0.40\linewidth}>{\raggedleft\arraybackslash}p{0.14\linewidth}>{\raggedright\arraybackslash}p{0.24\linewidth}@{}}
\toprule
Block & Parameter & Estimate & Interval \\
\midrule\endhead
Path components & $\alpha_A$: waist-to-hip ratio $\to$ LDL cholesterol slope ($A \to M$) & $-29.62$ & $[-102.9$, $40.69]$ \\
 & $\beta_M(q_{25})$: conditional LDL $\to$ PWV slope at the 25th percentile of waist-to-hip ratio ($M \to Y \mid A$) & $-0.001455$ & $[-0.0068$, $0.004806]$ \\
 & $\beta_M(q_{50})$: conditional LDL $\to$ PWV slope at the median of waist-to-hip ratio ($M \to Y \mid A$) & $-0.002968$ & $[-0.008336$, $0.001091]$ \\
 & $\beta_M(q_{75})$: conditional LDL $\to$ PWV slope at the 75th percentile of waist-to-hip ratio ($M \to Y \mid A$) & $-0.004228$ & $[-0.01154$, $-0.0001317]$ \\
 & $\beta_{AM}$: waist-to-hip ratio $\times$ LDL interaction coefficient in the PWV equation & $-0.02342$ & $[-0.09473$, $0.0108]$ \\
\bottomrule
\end{longtable}}

\paragraph{23.\ Expanded adjustment: waist-to-hip ratio $\to$ LDL cholesterol $\to$ thigh-to-ankle PWV}
\textit{Design.} LDL measured 30--730 days after waist-to-hip ratio and PWV 30--1095 days after LDL (strict order, endpoints inclusive); one measurement chain per participant. Beyond the main-path covariates, additional adjustment for the most recent BMI, HbA1c, creatinine and lipid-lowering medication (ATC C10) report on the waist-to-hip ratio measurement day or within the preceding 365 days (three categories: recorded, not seen in available reports, no available report). All variables on the original recorded scale.

\textit{Model.} $M = \alpha_0 + \alpha_A A + \alpha_C^\top C + \varepsilon_M$; $Y = \beta_0 + \beta_A A + \beta_M M + \beta_{AM} A M + \beta_C^\top C + \varepsilon_Y$. Both equations are linear regressions, with $A$, $M$ and $Y$ the waist-to-hip ratio, LDL cholesterol and thigh-to-ankle PWV, and $C$ the expanded covariate set. $\alpha_A$ is the change in LDL per 1 recorded unit higher waist-to-hip ratio; $\beta_M(q) = \beta_M + \beta_{AM} q$ is the conditional slope of PWV on LDL at waist-to-hip ratio $q$, and $\beta_{AM}$ is the interaction coefficient; the path product is $\theta(q) = \alpha_A \beta_M(q)$, with $q$ at the 25th, 50th and 75th percentiles of waist-to-hip ratio in the analysis sample ($q_{25}$, $q_{50}$, $q_{75}$).

\textit{Intervals.} $\alpha_A$, $\beta_M(q)$, $\beta_{AM}$: bootstrap 95\% percentile intervals; $\theta(q)$: bootstrap 95\% simultaneous confidence band (joint over the three percentile points).
{\footnotesize\setlength{\tabcolsep}{3pt}
\begin{longtable}{@{}>{\raggedright\arraybackslash}p{0.15\linewidth}>{\raggedright\arraybackslash}p{0.40\linewidth}>{\raggedleft\arraybackslash}p{0.14\linewidth}>{\raggedright\arraybackslash}p{0.24\linewidth}@{}}
\toprule
Block & Parameter & Estimate & Interval \\
\midrule\endhead
Path-product curve & $\theta(q_{25}) = \alpha_A \beta_M(q_{25})$: path product at the 25th percentile of waist-to-hip ratio & $-0.03832$ & $[-0.4758$, $0.3991]$ \\
 & $\theta(q_{50}) = \alpha_A \beta_M(q_{50})$: path product at the median of waist-to-hip ratio & $0.03915$ & $[-0.3046$, $0.3829]$ \\
 & $\theta(q_{75}) = \alpha_A \beta_M(q_{75})$: path product at the 75th percentile of waist-to-hip ratio & $0.1169$ & $[-0.3218$, $0.5556]$ \\
Path components & $\alpha_A$: waist-to-hip ratio $\to$ LDL cholesterol slope ($A \to M$) & $-52.58$ & $[-117.4$, $13.18]$ \\
 & $\beta_M(q_{25})$: conditional LDL $\to$ PWV slope at the 25th percentile of waist-to-hip ratio ($M \to Y \mid A$) & $0.0007288$ & $[-0.004242$, $0.00662]$ \\
 & $\beta_M(q_{50})$: conditional LDL $\to$ PWV slope at the median of waist-to-hip ratio ($M \to Y \mid A$) & $-0.0007446$ & $[-0.004803$, $0.003412]$ \\
 & $\beta_M(q_{75})$: conditional LDL $\to$ PWV slope at the 75th percentile of waist-to-hip ratio ($M \to Y \mid A$) & $-0.002223$ & $[-0.00724$, $0.002629]$ \\
 & $\beta_{AM}$: waist-to-hip ratio $\times$ LDL interaction coefficient in the PWV equation & $-0.02624$ & $[-0.08896$, $0.02265]$ \\
\bottomrule
\end{longtable}}

\paragraph{24.\ Narrow-window serial path: waist-to-hip ratio $\to$ LDL $\to$ diastolic blood pressure $\to$ PWV}
\textit{Design.} Waist-to-hip ratio, LDL cholesterol, seated diastolic blood pressure and PWV measured in the same period, with narrowed windows: LDL within 180 days before or after waist-to-hip ratio, diastolic blood pressure within 30 days before or after LDL, and PWV within 180 days before or after diastolic blood pressure (endpoints inclusive); one set of measurements per person, $n = 1449$. Adjustment covariates (age with a squared term, sex, study membership, smoking status), missing-data handling and standardization as in the main serial path.

\textit{Model.} $L = \alpha_1 + \beta_{AL} A + \gamma_1^\top C + \varepsilon_1$; $D = \alpha_2 + \beta_{AD} A + \beta_{LD} L + \gamma_2^\top C + \varepsilon_2$; $Y = \alpha_3 + \beta_{AY} A + \beta_{LY} L + \beta_{DY} D + \gamma_3^\top C + \varepsilon_3$. All three equations are linear regressions, with $A$, $L$, $D$ and $Y$ the standardized waist-to-hip ratio, LDL cholesterol, seated diastolic blood pressure and thigh-to-ankle PWV, and $C$ the covariates. $\beta_{AL}$ is the SD difference in LDL per 1 SD higher waist-to-hip ratio; $\beta_{LD}$ is the standardized slope of diastolic blood pressure on LDL adjusted for waist-to-hip ratio; $\beta_{DY}$ is the standardized slope of PWV on diastolic blood pressure adjusted for waist-to-hip ratio and LDL; the serial path product is $\theta = \beta_{AL} \beta_{LD} \beta_{DY}$.

\textit{Intervals.} $\beta_{AL}$, $\beta_{LD}$, $\beta_{DY}$: 95\% confidence intervals; $\theta$: 95\% percentile interval from joint normal simulation of the coefficients. $P$ values not corrected for multiple comparisons.
{\footnotesize\setlength{\tabcolsep}{3pt}
\begin{longtable}{@{}>{\raggedright\arraybackslash}p{0.14\linewidth}>{\raggedright\arraybackslash}p{0.29\linewidth}>{\raggedleft\arraybackslash}p{0.13\linewidth}>{\raggedright\arraybackslash}p{0.22\linewidth}>{\raggedleft\arraybackslash}p{0.14\linewidth}@{}}
\toprule
Block & Parameter & Estimate & Interval & $P$ \\
\midrule\endhead
Path components (narrow window) & $\beta_{AL}$: standardized waist-to-hip ratio $\to$ LDL cholesterol slope ($A \to L$) & $0.06008$ & $[-0.005711$, $0.1259]$ & \mbox{$0.07348$} \\
 & $\beta_{LD}$: standardized LDL $\to$ seated diastolic blood pressure slope adjusted for waist-to-hip ratio ($L \to D \mid A$) & $0.08287$ & $[0.03146$, $0.1343]$ & \mbox{$0.001582$} \\
 & $\beta_{DY}$: standardized diastolic blood pressure $\to$ PWV slope adjusted for waist-to-hip ratio and LDL ($D \to Y \mid A, L$) & $0.4096$ & $[0.3559$, $0.4633]$ & \mbox{$1.518\times10^{-50}$} \\
Path index (narrow window) & $\theta = \beta_{AL} \beta_{LD} \beta_{DY}$: serial path product & $0.00204$ & $[-0.0001736$, $0.005199]$ &  \\
\bottomrule
\end{longtable}}

\paragraph{25.\ Additional adjustment for BMI and height: waist-to-hip ratio $\to$ LDL $\to$ diastolic blood pressure $\to$ PWV}
\textit{Design.} Time windows as in the main serial path: LDL within 365 days before or after waist-to-hip ratio, diastolic blood pressure within 90 days before or after LDL, and PWV within 365 days before or after diastolic blood pressure (endpoints inclusive); one set of measurements per person. Beyond age (with a squared term), sex, study membership and smoking status, additional adjustment for BMI and height measured on the same day as waist-to-hip ratio; missing age, BMI and height multiply imputed, missing categorical covariates as a separate category. The four measurements, BMI and height standardized within the analysis sample.

\textit{Model.} $L = \alpha_1 + \beta_{AL} A + \gamma_1^\top C + \varepsilon_1$; $D = \alpha_2 + \beta_{AD} A + \beta_{LD} L + \gamma_2^\top C + \varepsilon_2$; $Y = \alpha_3 + \beta_{AY} A + \beta_{LY} L + \beta_{DY} D + \gamma_3^\top C + \varepsilon_3$. All three equations are linear regressions, with $A$, $L$, $D$ and $Y$ the standardized waist-to-hip ratio, LDL cholesterol, seated diastolic blood pressure and thigh-to-ankle PWV, and $C$ the covariates (including same-day BMI and height). $\beta_{AL}$ is the SD difference in LDL per 1 SD higher waist-to-hip ratio at equal BMI and height; $\beta_{LD}$ and $\beta_{DY}$ are defined as in the main serial path; the serial path product is $\theta = \beta_{AL} \beta_{LD} \beta_{DY}$.

\textit{Intervals.} $\beta_{AL}$, $\beta_{LD}$, $\beta_{DY}$: 95\% confidence intervals; $\theta$: 95\% percentile interval from joint normal simulation of the coefficients. $P$ values not corrected for multiple comparisons.
{\footnotesize\setlength{\tabcolsep}{3pt}
\begin{longtable}{@{}>{\raggedright\arraybackslash}p{0.14\linewidth}>{\raggedright\arraybackslash}p{0.29\linewidth}>{\raggedleft\arraybackslash}p{0.13\linewidth}>{\raggedright\arraybackslash}p{0.22\linewidth}>{\raggedleft\arraybackslash}p{0.14\linewidth}@{}}
\toprule
Block & Parameter & Estimate & Interval & $P$ \\
\midrule\endhead
Path components (additional adjustment for BMI and height) & $\beta_{AL}$: standardized waist-to-hip ratio $\to$ LDL cholesterol slope ($A \to L$) & $0.0359$ & $[-0.03976$, $0.1116]$ & \mbox{$0.3524$} \\
 & $\beta_{LD}$: standardized LDL $\to$ seated diastolic blood pressure slope adjusted for waist-to-hip ratio ($L \to D \mid A$) & $0.06941$ & $[0.0376$, $0.1012]$ & \mbox{$1.894\times10^{-5}$} \\
 & $\beta_{DY}$: standardized diastolic blood pressure $\to$ PWV slope adjusted for waist-to-hip ratio and LDL ($D \to Y \mid A, L$) & $0.4379$ & $[0.3997$, $0.4762]$ & \mbox{$1.37\times10^{-111}$} \\
Path index (additional adjustment for BMI and height) & $\theta = \beta_{AL} \beta_{LD} \beta_{DY}$: serial path product & $0.001091$ & $[-0.001113$, $0.003964]$ &  \\
\bottomrule
\end{longtable}}

\paragraph{26.\ Narrow-window joint path: waist-to-hip ratio $\to$ LDL $\to$ diastolic blood pressure $\to$ PWV ($\pm$90 days)}
\textit{Design.} As the $\pm$365-day joint path, with the time window narrowed to 90 days before or after the PWV measurement day (endpoints inclusive): waist-to-hip ratio, LDL cholesterol and seated diastolic blood pressure each taken as the measurement closest in date within the window, in any order; one row per PWV measurement day, with possibly multiple rows per participant. The three equations share one sample, with covariates and measurement definitions as in the $\pm$365-day joint path.

\textit{Model.} Equations as in the $\pm$365-day joint path: $M_1 = \alpha_0 + \alpha_A A + \alpha_C^\top C + \varepsilon_1$; $M_2 = \gamma_0 + \gamma_A A + \gamma_{M1} M_1 + \gamma_C^\top C + \varepsilon_2$; $Y = \theta_0 + \theta_A A + \theta_{M1} M_1 + \theta_{M2} M_2 + \theta_C^\top C + \varepsilon_3$ (linear regressions). Reported quantities as before: $0.1\alpha_A$, $0.1\gamma_A$ and $0.1\theta_A$ per 0.1 higher waist-to-hip ratio, and $\gamma_{M1}$, $\theta_{M1}$ and $\theta_{M2}$ per 1 original unit higher value of that variable, all as outcome differences with the other terms of the same equation fixed.

\textit{Intervals.} 95\% confidence intervals; $P$ values not corrected for multiple comparisons.
{\footnotesize\setlength{\tabcolsep}{3pt}
\begin{longtable}{@{}>{\raggedright\arraybackslash}p{0.14\linewidth}>{\raggedright\arraybackslash}p{0.29\linewidth}>{\raggedleft\arraybackslash}p{0.13\linewidth}>{\raggedright\arraybackslash}p{0.22\linewidth}>{\raggedleft\arraybackslash}p{0.14\linewidth}@{}}
\toprule
Block & Parameter & Estimate & Interval & $P$ \\
\midrule\endhead
LDL equation ($\pm$90 days) & $0.1\alpha_A$: waist-to-hip ratio $\to$ LDL cholesterol ($A \to M_1$) & $1.267$ & $[-5.008$, $7.542]$ & \mbox{$0.689$} \\
Diastolic blood pressure equation ($\pm$90 days) & $0.1\gamma_A$: waist-to-hip ratio $\to$ seated diastolic blood pressure, adjusted for LDL ($A \to M_2 \mid M_1$) & $2.666$ & $[0.9526$, $4.38]$ & \mbox{$0.002688$} \\
 & $\gamma_{M1}$: LDL cholesterol $\to$ seated diastolic blood pressure, adjusted for waist-to-hip ratio ($M_1 \to M_2 \mid A$) & $0.01524$ & $[-0.01431$, $0.04478]$ & \mbox{$0.2938$} \\
PWV equation ($\pm$90 days) & $0.1\theta_A$: waist-to-hip ratio $\to$ PWV, adjusted for LDL and diastolic blood pressure ($A \to Y \mid M_1, M_2$) & $0.1529$ & $[-0.03714$, $0.343]$ & \mbox{$0.1133$} \\
 & $\theta_{M1}$: LDL cholesterol $\to$ PWV, adjusted for waist-to-hip ratio and diastolic blood pressure ($M_1 \to Y \mid A, M_2$) & $-0.0002391$ & $[-0.004808$, $0.004329]$ & \mbox{$0.9138$} \\
 & $\theta_{M2}$: seated diastolic blood pressure $\to$ PWV, adjusted for waist-to-hip ratio and LDL ($M_2 \to Y \mid A, M_1$) & $0.07543$ & $[0.06199$, $0.08886]$ & \mbox{$5.042\times10^{-17}$} \\
\bottomrule
\end{longtable}}

\paragraph{27.\ Broad-window joint path: waist-to-hip ratio $\to$ LDL $\to$ diastolic blood pressure $\to$ PWV ($\pm$730 days)}
\textit{Design.} As the $\pm$365-day joint path, with the time window widened to 730 days before or after the PWV measurement day (endpoints inclusive): waist-to-hip ratio, LDL cholesterol and seated diastolic blood pressure each taken as the measurement closest in date within the window, in any order; one row per PWV measurement day, with possibly multiple rows per participant. The three equations share one sample, with covariates and measurement definitions as in the $\pm$365-day joint path.

\textit{Model.} Equations as in the $\pm$365-day joint path: $M_1 = \alpha_0 + \alpha_A A + \alpha_C^\top C + \varepsilon_1$; $M_2 = \gamma_0 + \gamma_A A + \gamma_{M1} M_1 + \gamma_C^\top C + \varepsilon_2$; $Y = \theta_0 + \theta_A A + \theta_{M1} M_1 + \theta_{M2} M_2 + \theta_C^\top C + \varepsilon_3$ (linear regressions). Reported quantities as before: $0.1\alpha_A$, $0.1\gamma_A$ and $0.1\theta_A$ per 0.1 higher waist-to-hip ratio, and $\gamma_{M1}$, $\theta_{M1}$ and $\theta_{M2}$ per 1 original unit higher value of that variable, all as outcome differences with the other terms of the same equation fixed.

\textit{Intervals.} 95\% confidence intervals; $P$ values not corrected for multiple comparisons.
{\footnotesize\setlength{\tabcolsep}{3pt}
\begin{longtable}{@{}>{\raggedright\arraybackslash}p{0.14\linewidth}>{\raggedright\arraybackslash}p{0.29\linewidth}>{\raggedleft\arraybackslash}p{0.13\linewidth}>{\raggedright\arraybackslash}p{0.22\linewidth}>{\raggedleft\arraybackslash}p{0.14\linewidth}@{}}
\toprule
Block & Parameter & Estimate & Interval & $P$ \\
\midrule\endhead
LDL equation ($\pm$730 days) & $0.1\alpha_A$: waist-to-hip ratio $\to$ LDL cholesterol ($A \to M_1$) & $0.1238$ & $[-3.17$, $3.418]$ & \mbox{$0.9411$} \\
Diastolic blood pressure equation ($\pm$730 days) & $0.1\gamma_A$: waist-to-hip ratio $\to$ seated diastolic blood pressure, adjusted for LDL ($A \to M_2 \mid M_1$) & $3.374$ & $[2.53$, $4.217]$ & \mbox{$8.252\times10^{-14}$} \\
 & $\gamma_{M1}$: LDL cholesterol $\to$ seated diastolic blood pressure, adjusted for waist-to-hip ratio ($M_1 \to M_2 \mid A$) & $0.01731$ & $[0.00156$, $0.03307]$ & \mbox{$0.03196$} \\
PWV equation ($\pm$730 days) & $0.1\theta_A$: waist-to-hip ratio $\to$ PWV, adjusted for LDL and diastolic blood pressure ($A \to Y \mid M_1, M_2$) & $0.1518$ & $[0.04221$, $0.2614]$ & \mbox{$0.006814$} \\
 & $\theta_{M1}$: LDL cholesterol $\to$ PWV, adjusted for waist-to-hip ratio and diastolic blood pressure ($M_1 \to Y \mid A, M_2$) & $0.0006378$ & $[-0.001423$, $0.002699]$ & \mbox{$0.5363$} \\
 & $\theta_{M2}$: seated diastolic blood pressure $\to$ PWV, adjusted for waist-to-hip ratio and LDL ($M_2 \to Y \mid A, M_1$) & $0.07203$ & $[0.06389$, $0.08016]$ & \mbox{$1.515\times10^{-46}$} \\
\bottomrule
\end{longtable}}

\paragraph{28.\ Joint path with additional adjustment for BMI and HbA1c ($\pm$365 days)}
\textit{Design.} Time window and matching as in the $\pm$365-day joint path: each PWV measurement day as anchor, with waist-to-hip ratio, LDL cholesterol and seated diastolic blood pressure each taken as the measurement closest in date within 365 days before or after it (endpoints inclusive), in any order. The three equations share one sample, with additional adjustment, beyond approximate age, sex and recruitment study category, for BMI measured on the same day as the matched waist-to-hip ratio and HbA1c measured on the same day as the matched LDL; all variables kept on the original recorded scale.

\textit{Model.} $M_1 = \alpha_0 + \alpha_A A + \alpha_C^\top C + \varepsilon_1$; $M_2 = \gamma_0 + \gamma_A A + \gamma_{M1} M_1 + \gamma_C^\top C + \varepsilon_2$; $Y = \theta_0 + \theta_A A + \theta_{M1} M_1 + \theta_{M2} M_2 + \theta_C^\top C + \varepsilon_3$ (linear regressions), with $C$ approximate age, sex, recruitment study category, same-day BMI and same-day HbA1c. Reported quantities as in the $\pm$365-day joint path: $0.1\alpha_A$, $0.1\gamma_A$ and $0.1\theta_A$ per 0.1 higher waist-to-hip ratio, and $\gamma_{M1}$, $\theta_{M1}$ and $\theta_{M2}$ per 1 original unit higher value of that variable, all as outcome differences with the other terms of the same equation fixed.

\textit{Intervals.} 95\% confidence intervals; $P$ values not corrected for multiple comparisons.
{\footnotesize\setlength{\tabcolsep}{3pt}
\begin{longtable}{@{}>{\raggedright\arraybackslash}p{0.14\linewidth}>{\raggedright\arraybackslash}p{0.29\linewidth}>{\raggedleft\arraybackslash}p{0.13\linewidth}>{\raggedright\arraybackslash}p{0.22\linewidth}>{\raggedleft\arraybackslash}p{0.14\linewidth}@{}}
\toprule
Block & Parameter & Estimate & Interval & $P$ \\
\midrule\endhead
LDL equation (additional adjustment for BMI, HbA1c) & $0.1\alpha_A$: waist-to-hip ratio $\to$ LDL cholesterol ($A \to M_1$) & $1.887$ & $[-4.02$, $7.793]$ & \mbox{$0.5282$} \\
Diastolic blood pressure equation (additional adjustment for BMI, HbA1c) & $0.1\gamma_A$: waist-to-hip ratio $\to$ seated diastolic blood pressure, adjusted for LDL ($A \to M_2 \mid M_1$) & $1.051$ & $[-0.4278$, $2.53]$ & \mbox{$0.1619$} \\
 & $\gamma_{M1}$: LDL cholesterol $\to$ seated diastolic blood pressure, adjusted for waist-to-hip ratio ($M_1 \to M_2 \mid A$) & $0.008258$ & $[-0.01652$, $0.03303]$ & \mbox{$0.4995$} \\
PWV equation (additional adjustment for BMI, HbA1c) & $0.1\theta_A$: waist-to-hip ratio $\to$ PWV, adjusted for LDL and diastolic blood pressure ($A \to Y \mid M_1, M_2$) & $0.1769$ & $[-0.009872$, $0.3637]$ & \mbox{$0.06318$} \\
 & $\theta_{M1}$: LDL cholesterol $\to$ PWV, adjusted for waist-to-hip ratio and diastolic blood pressure ($M_1 \to Y \mid A, M_2$) & $0.001067$ & $[-0.003241$, $0.005376]$ & \mbox{$0.6151$} \\
 & $\theta_{M2}$: seated diastolic blood pressure $\to$ PWV, adjusted for waist-to-hip ratio and LDL ($M_2 \to Y \mid A, M_1$) & $0.07043$ & $[0.05806$, $0.08281]$ & \mbox{$1.501\times10^{-17}$} \\
\bottomrule
\end{longtable}}

\FloatBarrier

\subsection{Hip circumference--GGT--SBP--PWV}
\label{app:rq-designs:43d6}

\paragraph{Research question and measures.}
The research question concerns the association of hip circumference, via $\gamma$-glutamyl transferase (GGT) and systolic blood pressure, with lower-limb arterial stiffness, with data from the HPP cohort. The exposure is hip circumference; the mediators are blood-test GGT and seated systolic blood pressure; the outcome is bilateral thigh-to-ankle pulse wave velocity (PWV). Unless stated otherwise, coefficients are SD changes per 1 SD after standardization within the analysis sample of each unit.

\subsubsection*{Primary pathways}

\paragraph{1.\ Single-mediator pathway: hip circumference $\to$ GGT $\to$ thigh-to-ankle PWV}
\textit{Design.} Hip circumference, GGT and PWV are measured in strict date order: GGT 30--730 days after hip circumference and PWV 90--1095 days after GGT (bounds inclusive); one measurement chain per participant. Adjustment for approximate age at the hip measurement, sex and recruitment cohort. PWV is the bilateral mean, taken when both left and right values are valid. Hip circumference, GGT and PWV are standardized within the analysis sample, with sex and cohort as categorical terms.

\textit{Model.} $M = \alpha_0 + \alpha_A \cdot A + \alpha_C^\top \cdot C + \varepsilon_M$; $Y = \beta_0 + \beta_A \cdot A + \beta_M \cdot M + \beta_C^\top \cdot C + \varepsilon_Y$; $Y = \tau_0 + \tau_A \cdot A + \tau_C^\top \cdot C + \varepsilon_T$. All three are linear regressions, with $A$, $M$ and $Y$ the standardized hip circumference, GGT and thigh-to-ankle PWV, and $C$ the covariates. $\alpha_A$ is the SD change in GGT per 1 SD higher hip circumference; $\beta_M$ and $\beta_A$ are the mutually adjusted standardized slopes of PWV on GGT and on hip circumference; $\tau_A$ is the total standardized slope of PWV on hip circumference, adjusted for $C$ only; the path product is $\theta = \alpha_A \cdot \beta_M$; $\mathrm{RV}(b) = \tfrac{1}{2}[\sqrt{f^4 + 4f^2} - f^2]$, $f = |t_b|/\sqrt{\nu}$, is the equal-strength partial $R^2$ that a single omitted confounder must have with the regressor and with the equation outcome to reduce coefficient $b$ to 0 ($t_b$ is the $t$ statistic of $b$ and $\nu$ the residual degrees of freedom).

\textit{Intervals.} Bootstrap 95\% percentile intervals.
{\footnotesize\setlength{\tabcolsep}{3pt}
\begin{longtable}{@{}>{\raggedright\arraybackslash}p{0.15\linewidth}>{\raggedright\arraybackslash}p{0.40\linewidth}>{\raggedleft\arraybackslash}p{0.14\linewidth}>{\raggedright\arraybackslash}p{0.24\linewidth}@{}}
\toprule
Block & Parameter & Estimate & Interval \\
\midrule\endhead
Path components & $\alpha_A$: standardized slope of hip circumference $\to$ GGT ($A \to M$) & $0.06359$ & $[-0.05313$, $0.1825]$ \\
 & $\beta_M$: standardized slope of GGT $\to$ thigh-to-ankle PWV, adjusted for hip circumference ($M \to Y \mid A$) & $0.09861$ & $[-0.04392$, $0.2407]$ \\
 & $\beta_A$: standardized slope of hip circumference $\to$ PWV, adjusted for GGT ($A \to Y \mid M$) & $0.05268$ & $[-0.08222$, $0.1939]$ \\
 & $\tau_A$: adjusted total standardized slope of hip circumference $\to$ PWV ($A \to Y$) & $0.05895$ & $[-0.07739$, $0.2036]$ \\
Path index & $\theta = \alpha_A \cdot \beta_M$: path-product association index & $0.006271$ & $[-0.00697$, $0.02438]$ \\
Omitted-confounder robustness values & $\mathrm{RV}(\alpha_A)$: equal-strength omitted-confounder partial $R^2$ required to reduce $\alpha_A$ to 0 & $0.06309$ & $[0.00442$, $0.1715]$ \\
 & $\mathrm{RV}(\beta_M)$: equal-strength omitted-confounder partial $R^2$ required to reduce $\beta_M$ to 0 & $0.09655$ & $[0.004159$, $0.2252]$ \\
 & $\mathrm{RV}(\beta_A)$: equal-strength omitted-confounder partial $R^2$ required to reduce $\beta_A$ to 0 & $0.05399$ & $[0.003224$, $0.189]$ \\
\bottomrule
\end{longtable}}

\paragraph{2.\ Single-mediator pathway: hip circumference $\to$ seated systolic blood pressure $\to$ thigh-to-ankle PWV}
\textit{Design.} Hip circumference, seated blood pressure and PWV are measured in strict date order: blood pressure 30--730 days after hip circumference and PWV 180--1095 days after blood pressure (bounds inclusive); one measurement chain per participant. Adjustment for approximate age at the hip measurement, sex and recruitment cohort. Systolic blood pressure is the mean of the available values of the two seated readings. PWV is the bilateral mean, taken when both left and right values are valid. Hip circumference, systolic blood pressure and PWV are standardized within the analysis sample, with sex and cohort as categorical terms.

\textit{Model.} $M = \alpha_0 + \alpha_A \cdot A + \alpha_C^\top \cdot C + \varepsilon_M$; $Y = \beta_0 + \beta_A \cdot A + \beta_M \cdot M + \beta_C^\top \cdot C + \varepsilon_Y$; $Y = \tau_0 + \tau_A \cdot A + \tau_C^\top \cdot C + \varepsilon_T$. All three are linear regressions, with $A$, $M$ and $Y$ the standardized hip circumference, seated systolic blood pressure and thigh-to-ankle PWV, and $C$ the covariates. $\alpha_A$ is the SD change in systolic blood pressure per 1 SD higher hip circumference; $\beta_M$ and $\beta_A$ are the mutually adjusted standardized slopes of PWV on systolic blood pressure and on hip circumference; $\tau_A$ is the total standardized slope of PWV on hip circumference, adjusted for $C$ only; the path product is $\theta = \alpha_A \cdot \beta_M$; $\mathrm{RV}(b) = \tfrac{1}{2}[\sqrt{f^4 + 4f^2} - f^2]$, $f = |t_b|/\sqrt{\nu}$, is the equal-strength partial $R^2$ that a single omitted confounder must have with the regressor and with the equation outcome to reduce coefficient $b$ to 0 ($t_b$ is the $t$ statistic of $b$ and $\nu$ the residual degrees of freedom).

\textit{Intervals.} Bootstrap 95\% percentile intervals.
{\footnotesize\setlength{\tabcolsep}{3pt}
\begin{longtable}{@{}>{\raggedright\arraybackslash}p{0.15\linewidth}>{\raggedright\arraybackslash}p{0.40\linewidth}>{\raggedleft\arraybackslash}p{0.14\linewidth}>{\raggedright\arraybackslash}p{0.24\linewidth}@{}}
\toprule
Block & Parameter & Estimate & Interval \\
\midrule\endhead
Path components & $\alpha_A$: standardized slope of hip circumference $\to$ seated systolic blood pressure ($A \to M$) & $-1.224$ & $[-7.58$, $1.748]$ \\
 & $\beta_M$: standardized slope of seated systolic blood pressure $\to$ thigh-to-ankle PWV, adjusted for hip circumference ($M \to Y \mid A$) & $1.494$ & $[-0.1309$, $13.99]$ \\
 & $\beta_A$: standardized slope of hip circumference $\to$ PWV, adjusted for seated systolic blood pressure ($A \to Y \mid M$) & $0.9852$ & $[-0.7504$, $18.87]$ \\
 & $\tau_A$: adjusted total standardized slope of hip circumference $\to$ PWV ($A \to Y$) & $-0.8437$ & $[-9.884$, $3.008]$ \\
Path index & $\theta = \alpha_A \cdot \beta_M$: path-product association index & $-1.829$ & $[-15.9$, $1.813]$ \\
Omitted-confounder robustness values & $\mathrm{RV}(\alpha_A)$: equal-strength omitted-confounder partial $R^2$ required to reduce $\alpha_A$ to 0 & $0.4731$ & $[0$, $0.9854]$ \\
 & $\mathrm{RV}(\beta_M)$: equal-strength omitted-confounder partial $R^2$ required to reduce $\beta_M$ to 0 & $0.9606$ & $[0$, $0.9606]$ \\
 & $\mathrm{RV}(\beta_A)$: equal-strength omitted-confounder partial $R^2$ required to reduce $\beta_A$ to 0 & $0.7342$ & $[0$, $0.7342]$ \\
\bottomrule
\end{longtable}}

\paragraph{3.\ Serial pathway and bias bounds: hip circumference $\to$ GGT $\to$ seated systolic blood pressure $\to$ PWV}
\textit{Design.} Each GGT test serves as the anchor: hip circumference on the GGT day or within the 0--365 days before it, seated systolic blood pressure within 30 days before or after GGT (either order), and PWV on the systolic blood pressure day or within the 0--365 days after it (bounds inclusive); a participant may contribute multiple chains, $n = 64$. Adjustment for approximate age at the hip measurement and sex. Systolic blood pressure is the mean of the available values of the two seated readings, and PWV the mean of the available sides. Continuous variables are standardized within the analysis sample, with sex as a categorical term.

\textit{Model.} $M_1 = \alpha_1 + a_1 A + \gamma_1^\top C + \varepsilon_1$; $M_2 = \alpha_2 + a_2 A + d_{21} M_1 + \gamma_2^\top C + \varepsilon_2$; $Y = \alpha_3 + c' A + b_1 M_1 + b_2 M_2 + \gamma_3^\top C + \varepsilon_3$. All three are linear regressions, with $A$, $M_1$, $M_2$ and $Y$ the standardized hip circumference, $\gamma$-glutamyl transferase (GGT), seated systolic blood pressure and thigh-to-ankle PWV, and $C$ age and sex. Reported are the serial path index $a_1 \cdot d_{21} \cdot b_2$, six standardized coefficients (the SD change in the outcome per 1 SD higher corresponding variable in each equation), and the bias bounds of $a_1$, $d_{21}$ and $b_2$, $\hat{\beta} \pm \mathrm{se} \cdot \sqrt{\nu \cdot R^2_{Y\sim U} \cdot R^2_{D\sim U}/(1 - R^2_{D\sim U})}$, where $\mathrm{se}$ is the conventional standard error of the coefficient, $\nu$ the residual degrees of freedom, and $R^2_{D\sim U}$ and $R^2_{Y\sim U}$ the partial $R^2$ of a hypothetical omitted covariate $U$ with the regressor of the coefficient and with the equation outcome, each taking 0.01, 0.05, 0.10 and 0.20.

\textit{Intervals.} Path index and path coefficients: 95\% confidence intervals. Bias-bound grid: each row corresponds to one $(R^2_{D\sim U}, R^2_{Y\sim U})$ pair, with lower and upper limits equal to the coefficient $\pm$ the bias bound.
{\footnotesize\setlength{\tabcolsep}{3pt}
\begin{longtable}{@{}>{\raggedright\arraybackslash}p{0.15\linewidth}>{\raggedright\arraybackslash}p{0.40\linewidth}>{\raggedleft\arraybackslash}p{0.14\linewidth}>{\raggedright\arraybackslash}p{0.24\linewidth}@{}}
\toprule
Block & Parameter & Estimate & Interval \\
\midrule\endhead
Serial path index & Serial path index $a_1 \cdot d_{21} \cdot b_2$ (hip circumference $\to$ GGT $\to$ seated systolic blood pressure $\to$ PWV) & $-0.002988$ & $[-0.01834$, $0.01236]$ \\
Path coefficients & Standardized slope $a_1$ (hip circumference $\to$ GGT) & $0.08544$ & $[-0.1068$, $0.2777]$ \\
 & Standardized slope $d_{21}$ (GGT $\to$ seated systolic blood pressure, controlling for hip circumference) & $-0.05347$ & $[-0.3095$, $0.2026]$ \\
 & Standardized slope $b_2$ (seated systolic blood pressure $\to$ PWV, controlling for hip circumference and GGT) & $0.6541$ & $[0.4989$, $0.8093]$ \\
 & Standardized slope $a_2$ (hip circumference $\to$ seated systolic blood pressure, controlling for GGT) & $0.1618$ & $[-0.01997$, $0.3436]$ \\
 & Standardized slope $b_1$ (GGT $\to$ PWV, controlling for hip circumference and seated systolic blood pressure) & $0.1966$ & $[0.007984$, $0.3852]$ \\
 & Conditional slope $c'$ (hip circumference $\to$ PWV, controlling for both mediators) & $-0.09191$ & $[-0.3101$, $0.1263]$ \\
\bottomrule
\end{longtable}}

{\scriptsize\setlength{\tabcolsep}{3pt}
\begin{longtable}{@{}lccc@{}}
\toprule
Partial $R^2$ of U (regressor; outcome) & $a_1$ & $d_{21}$ & $b_2$ \\
\midrule\endhead
Coefficient & $0.08544$ & $-0.05347$ & $0.6541$ \\
0.01; 0.01 & $[0.07613, 0.09476]$ & $[-0.06347, -0.04346]$ & $[0.6467, 0.6615]$ \\
0.01; 0.05 & $[0.06461, 0.1063]$ & $[-0.07583, -0.0311]$ & $[0.6376, 0.6706]$ \\
0.01; 0.10 & $[0.05598, 0.1149]$ & $[-0.08509, -0.02184]$ & $[0.6308, 0.6774]$ \\
0.01; 0.20 & $[0.04378, 0.1271]$ & $[-0.09819, -0.00874]$ & $[0.6212, 0.687]$ \\
0.05; 0.01 & $[0.06418, 0.1067]$ & $[-0.07629, -0.03064]$ & $[0.6373, 0.6709]$ \\
0.05; 0.05 & $[0.03789, 0.133]$ & $[-0.1045, -0.002419]$ & $[0.6165, 0.6917]$ \\
0.05; 0.10 & $[0.0182, 0.1527]$ & $[-0.1257, 0.01873]$ & $[0.601, 0.7072]$ \\
0.05; 0.20 & $[-0.009654, 0.1805]$ & $[-0.1556, 0.04863]$ & $[0.579, 0.7292]$ \\
0.10; 0.01 & $[0.05455, 0.1163]$ & $[-0.08664, -0.0203]$ & $[0.6297, 0.6785]$ \\
0.10; 0.05 & $[0.01636, 0.1545]$ & $[-0.1276, 0.0207]$ & $[0.5995, 0.7087]$ \\
0.10; 0.10 & $[-0.01226, 0.1831]$ & $[-0.1584, 0.05143]$ & $[0.5769, 0.7313]$ \\
0.10; 0.20 & $[-0.05273, 0.2236]$ & $[-0.2018, 0.09487]$ & $[0.5449, 0.7633]$ \\
0.20; 0.01 & $[0.0391, 0.1318]$ & $[-0.1032, -0.003711]$ & $[0.6175, 0.6907]$ \\
0.20; 0.05 & $[-0.01819, 0.1891]$ & $[-0.1647, 0.05779]$ & $[0.5722, 0.736]$ \\
0.20; 0.10 & $[-0.06111, 0.232]$ & $[-0.2108, 0.1039]$ & $[0.5383, 0.7699]$ \\
0.20; 0.20 & $[-0.1218, 0.2927]$ & $[-0.276, 0.169]$ & $[0.4904, 0.8178]$ \\
\bottomrule
\end{longtable}}

\paragraph{4.\ Exposure--outcome association: hip circumference and thigh-to-ankle PWV}
\textit{Design.} Hip circumference and PWV are measured in the same period (PWV within 90 days before or after hip circumference, bounds inclusive); each participant contributes the pair with the shortest gap, $n = 9385$. PWV is the mean of the available sides. Adjustment for approximate age (calendar year of the hip-measurement day minus birth year), recorded sex and recruitment cohort; hip circumference, PWV and age are kept on their original numerical scales, with sex and recruitment cohort as categorical terms.

\textit{Model.} $Y = \beta_0 + \beta_A \cdot A + \beta_{\mathrm{age}} \cdot \mathrm{age} + \gamma_{\mathrm{sex}} + \delta_{\mathrm{cohort}} + \varepsilon$ (linear regression; $A$ is hip circumference and $Y$ thigh-to-ankle PWV, both on original scales). $\beta_A$ is the difference in PWV per 1 unit higher hip circumference at fixed age, sex and recruitment cohort.

\textit{Intervals.} 95\% confidence intervals.
{\footnotesize\setlength{\tabcolsep}{3pt}
\begin{longtable}{@{}>{\raggedright\arraybackslash}p{0.15\linewidth}>{\raggedright\arraybackslash}p{0.40\linewidth}>{\raggedleft\arraybackslash}p{0.14\linewidth}>{\raggedright\arraybackslash}p{0.24\linewidth}@{}}
\toprule
Block & Parameter & Estimate & Interval \\
\midrule\endhead
Hip circumference--PWV association & $\beta_A$: adjusted slope of hip circumference $\to$ thigh-to-ankle PWV ($A \to Y$) & $0.02428$ & $[0.02041$, $0.02816]$ \\
\bottomrule
\end{longtable}}

\subsubsection*{Adjacent segments and pairwise associations}

\paragraph{5.\ Serial segment: hip circumference $\to$ GGT $\to$ seated systolic blood pressure}
\textit{Design.} Hip circumference, GGT and seated blood pressure are measured in strict date order: GGT 30--730 days after hip circumference and blood pressure 30--730 days after GGT (bounds inclusive); one measurement chain per participant, $n = 222$. Adjustment for approximate age at the hip measurement, sex and recruitment cohort. Systolic blood pressure is the mean of the available values of the two seated readings. Hip circumference, GGT and systolic blood pressure are standardized within the analysis sample, with sex and cohort as categorical terms.

\textit{Model.} $M = \alpha_0 + \alpha_A \cdot A + \alpha_C^\top \cdot C + \varepsilon_M$; $Y = \beta_0 + \beta_A \cdot A + \beta_M \cdot M + \beta_C^\top \cdot C + \varepsilon_Y$; $Y = \tau_0 + \tau_A \cdot A + \tau_C^\top \cdot C + \varepsilon_T$. All three are linear regressions, with $A$, $M$ and $Y$ the standardized hip circumference, GGT and seated systolic blood pressure, and $C$ the covariates. $\alpha_A$ is the SD change in GGT per 1 SD higher hip circumference; $\beta_M$ and $\beta_A$ are the mutually adjusted standardized slopes of systolic blood pressure on GGT and on hip circumference; $\tau_A$ is the total standardized slope of systolic blood pressure on hip circumference, adjusted for $C$ only; the segment path product is $\theta = \alpha_A \cdot \beta_M$; $\mathrm{RV}(b) = \tfrac{1}{2}[\sqrt{f^4 + 4f^2} - f^2]$, $f = |t_b|/\sqrt{\nu}$, is the equal-strength partial $R^2$ that a single omitted confounder must have with the regressor and with the equation outcome to reduce coefficient $b$ to 0 ($t_b$ is the $t$ statistic of $b$ and $\nu$ the residual degrees of freedom).

\textit{Intervals.} Bootstrap 95\% percentile intervals.
{\footnotesize\setlength{\tabcolsep}{3pt}
\begin{longtable}{@{}>{\raggedright\arraybackslash}p{0.15\linewidth}>{\raggedright\arraybackslash}p{0.40\linewidth}>{\raggedleft\arraybackslash}p{0.14\linewidth}>{\raggedright\arraybackslash}p{0.24\linewidth}@{}}
\toprule
Block & Parameter & Estimate & Interval \\
\midrule\endhead
Path components & $\alpha_A$: standardized slope of hip circumference $\to$ GGT ($A \to M$) & $0.09437$ & $[-0.02007$, $0.1924]$ \\
 & $\beta_M$: standardized slope of GGT $\to$ seated systolic blood pressure, adjusted for hip circumference ($M \to Y \mid A$) & $0.01108$ & $[-0.09675$, $0.105]$ \\
 & $\beta_A$: standardized slope of hip circumference $\to$ systolic blood pressure, adjusted for GGT ($A \to Y \mid M$) & $0.2145$ & $[0.1057$, $0.3254]$ \\
 & $\tau_A$: adjusted total standardized slope of hip circumference $\to$ systolic blood pressure ($A \to Y$) & $0.2155$ & $[0.105$, $0.3248]$ \\
Path index & $\theta = \alpha_A \cdot \beta_M$: segment path-product association index & $0.001046$ & $[-0.01027$, $0.01296]$ \\
Omitted-confounder robustness values & $\mathrm{RV}(\alpha_A)$: equal-strength omitted-confounder partial $R^2$ required to reduce $\alpha_A$ to 0 & $0.09104$ & $[0.005738$, $0.1791]$ \\
 & $\mathrm{RV}(\beta_M)$: equal-strength omitted-confounder partial $R^2$ required to reduce $\beta_M$ to 0 & $0.01187$ & $[0.002561$, $0.114]$ \\
 & $\mathrm{RV}(\beta_A)$: equal-strength omitted-confounder partial $R^2$ required to reduce $\beta_A$ to 0 & $0.2073$ & $[0.105$, $0.3048]$ \\
\bottomrule
\end{longtable}}

\paragraph{6.\ Pairwise association: hip circumference $\to$ GGT (0--365 days)}
\textit{Design.} Hip circumference measured on the GGT test day or within the 0--365 days before it (bounds inclusive); one row per GGT test, and a participant may contribute multiple rows, $n = 238$. Adjustment for approximate age at the hip measurement and sex; hip circumference, GGT and age are standardized within the analysis sample, with sex as a categorical term.

\textit{Model.} $M1_z = \beta_0 + \beta \cdot A_z + \beta_{\mathrm{age}} \cdot \mathrm{Age}_z + \gamma_{\mathrm{Sex}} + \varepsilon$ (linear regression; $A$ is hip circumference and $M1$ is $\gamma$-glutamyl transferase GGT). $\beta$ is the SD difference in GGT per 1 SD higher hip circumference.

\textit{Intervals.} 95\% confidence intervals.
{\footnotesize\setlength{\tabcolsep}{3pt}
\begin{longtable}{@{}>{\raggedright\arraybackslash}p{0.15\linewidth}>{\raggedright\arraybackslash}p{0.40\linewidth}>{\raggedleft\arraybackslash}p{0.14\linewidth}>{\raggedright\arraybackslash}p{0.24\linewidth}@{}}
\toprule
Block & Parameter & Estimate & Interval \\
\midrule\endhead
Forward window 0--365 days & Adjusted standardized slope $\beta$ (hip circumference $\to$ GGT) & $0.1205$ & $[-0.02022$, $0.2612]$ \\
\bottomrule
\end{longtable}}

\paragraph{7.\ Pairwise association: GGT--seated systolic blood pressure ($\pm$30 days)}
\textit{Design.} Seated systolic blood pressure measured within 30 days before or after the GGT test (either order, bounds inclusive); one row per GGT test, and a participant may contribute multiple rows, $n = 873$. Adjustment for approximate age at the GGT test and sex; systolic blood pressure is the mean of the available values of the two seated readings; GGT, systolic blood pressure and age are standardized within the analysis sample, with sex as a categorical term.

\textit{Model.} $M2_z = \beta_0 + \beta \cdot M1_z + \beta_{\mathrm{age}} \cdot \mathrm{Age}_z + \gamma_{\mathrm{Sex}} + \varepsilon$ (linear regression; $M1$ is $\gamma$-glutamyl transferase GGT and $M2$ is seated systolic blood pressure). $\beta$ is the SD difference in seated systolic blood pressure per 1 SD higher GGT.

\textit{Intervals.} 95\% confidence intervals.
{\footnotesize\setlength{\tabcolsep}{3pt}
\begin{longtable}{@{}>{\raggedright\arraybackslash}p{0.15\linewidth}>{\raggedright\arraybackslash}p{0.40\linewidth}>{\raggedleft\arraybackslash}p{0.14\linewidth}>{\raggedright\arraybackslash}p{0.24\linewidth}@{}}
\toprule
Block & Parameter & Estimate & Interval \\
\midrule\endhead
Two-sided window $\pm$30 days & Adjusted standardized slope $\beta$ (GGT $\to$ seated systolic blood pressure) & $-0.005974$ & $[-0.07491$, $0.06296]$ \\
\bottomrule
\end{longtable}}

\paragraph{8.\ Pairwise association: seated systolic blood pressure $\to$ PWV (0--365 days)}
\textit{Design.} PWV measured on the seated blood pressure measurement day or within the 0--365 days after it (bounds inclusive); one row per blood pressure measurement day, and a participant may contribute multiple rows, $n = 9397$. Adjustment for approximate age at the blood pressure measurement and sex; systolic blood pressure is the mean of the available values of the two seated readings, and PWV the mean of the available sides; systolic blood pressure, PWV and age are standardized within the analysis sample, with sex as a categorical term.

\textit{Model.} $Y_z = \beta_0 + \beta \cdot M2_z + \beta_{\mathrm{age}} \cdot \mathrm{Age}_z + \gamma_{\mathrm{Sex}} + \varepsilon$ (linear regression; $M2$ is seated systolic blood pressure and $Y$ is thigh-to-ankle PWV). $\beta$ is the SD difference in PWV per 1 SD higher seated systolic blood pressure.

\textit{Intervals.} 95\% confidence intervals.
{\footnotesize\setlength{\tabcolsep}{3pt}
\begin{longtable}{@{}>{\raggedright\arraybackslash}p{0.15\linewidth}>{\raggedright\arraybackslash}p{0.40\linewidth}>{\raggedleft\arraybackslash}p{0.14\linewidth}>{\raggedright\arraybackslash}p{0.24\linewidth}@{}}
\toprule
Block & Parameter & Estimate & Interval \\
\midrule\endhead
Forward window 0--365 days & Adjusted standardized slope $\beta$ (seated systolic blood pressure $\to$ thigh-to-ankle PWV) & $0.4844$ & $[0.4651$, $0.5036]$ \\
\bottomrule
\end{longtable}}

\paragraph{9.\ Pairwise association: hip circumference--GGT ($\pm$90 days)}
\textit{Design.} Hip circumference and $\gamma$-glutamyl transferase (GGT) are measured in the same period (GGT within 90 days before or after hip circumference, bounds inclusive); each participant contributes the pair with the closest dates, $n = 1650$. Adjustment for approximate age (calendar year of the hip-measurement day minus birth year), recorded sex and recruitment cohort; hip circumference, GGT and age are kept on their original numerical scales, with sex and recruitment cohort as categorical terms.

\textit{Model.} $M_1 = \beta_0 + \beta_A \cdot A + \beta_{\mathrm{age}} \cdot \mathrm{age} + \gamma_{\mathrm{sex}} + \delta_{\mathrm{cohort}} + \varepsilon$ (linear regression; $A$ is hip circumference and $M_1$ GGT, both on original scales). $\beta_A$ is the difference in GGT per 1 unit higher hip circumference at fixed age, sex and recruitment cohort.

\textit{Intervals.} 95\% confidence intervals.
{\footnotesize\setlength{\tabcolsep}{3pt}
\begin{longtable}{@{}>{\raggedright\arraybackslash}p{0.15\linewidth}>{\raggedright\arraybackslash}p{0.40\linewidth}>{\raggedleft\arraybackslash}p{0.14\linewidth}>{\raggedright\arraybackslash}p{0.24\linewidth}@{}}
\toprule
Block & Parameter & Estimate & Interval \\
\midrule\endhead
Pairwise association & $\beta_A$: adjusted slope of hip circumference $\to$ GGT ($A \to M_1$) & $0.2991$ & $[0.1901$, $0.4081]$ \\
\bottomrule
\end{longtable}}

\paragraph{10.\ Pairwise association: seated systolic blood pressure--PWV ($\pm$90 days)}
\textit{Design.} Seated blood pressure and PWV are measured in the same period (PWV within 90 days before or after blood pressure, bounds inclusive); each participant contributes the pair with the closest dates, $n = 9397$. Systolic blood pressure is the mean of the available values of the two seated readings, and PWV the mean of the available sides. Adjustment for approximate age (calendar year of the blood pressure measurement day minus birth year), recorded sex and recruitment cohort; systolic blood pressure, PWV and age are kept on their original numerical scales, with sex and recruitment cohort as categorical terms.

\textit{Model.} $Y = \beta_0 + \beta_{M2} \cdot M_2 + \beta_{\mathrm{age}} \cdot \mathrm{age} + \gamma_{\mathrm{sex}} + \delta_{\mathrm{cohort}} + \varepsilon$ (linear regression; $M_2$ is seated systolic blood pressure and $Y$ thigh-to-ankle PWV, both on original scales). $\beta_{M2}$ is the difference in PWV per 1 unit higher systolic blood pressure at fixed age, sex and recruitment cohort.

\textit{Intervals.} 95\% confidence intervals.
{\footnotesize\setlength{\tabcolsep}{3pt}
\begin{longtable}{@{}>{\raggedright\arraybackslash}p{0.15\linewidth}>{\raggedright\arraybackslash}p{0.40\linewidth}>{\raggedleft\arraybackslash}p{0.14\linewidth}>{\raggedright\arraybackslash}p{0.24\linewidth}@{}}
\toprule
Block & Parameter & Estimate & Interval \\
\midrule\endhead
Pairwise association & $\beta_{M2}$: adjusted slope of seated systolic blood pressure $\to$ thigh-to-ankle PWV ($M_2 \to Y$) & $0.05098$ & $[0.04878$, $0.05318]$ \\
\bottomrule
\end{longtable}}

\paragraph{11.\ Pairwise association: hip circumference--seated systolic blood pressure ($\pm$90 days)}
\textit{Design.} Hip circumference and seated blood pressure are measured in the same period (blood pressure within 90 days before or after hip circumference, bounds inclusive); each participant contributes the pair with the closest dates, $n = 10380$. Systolic blood pressure is the mean of the available values of the two seated readings. Adjustment for approximate age (calendar year of the hip-measurement day minus birth year), recorded sex and recruitment cohort; hip circumference, systolic blood pressure and age are kept on their original numerical scales, with sex and recruitment cohort as categorical terms.

\textit{Model.} $M_2 = \beta_0 + \beta_A \cdot A + \beta_{\mathrm{age}} \cdot \mathrm{age} + \gamma_{\mathrm{sex}} + \delta_{\mathrm{cohort}} + \varepsilon$ (linear regression; $A$ is hip circumference and $M_2$ seated systolic blood pressure, both on original scales). $\beta_A$ is the difference in systolic blood pressure per 1 unit higher hip circumference at fixed age, sex and recruitment cohort.

\textit{Intervals.} 95\% confidence intervals.
{\footnotesize\setlength{\tabcolsep}{3pt}
\begin{longtable}{@{}>{\raggedright\arraybackslash}p{0.15\linewidth}>{\raggedright\arraybackslash}p{0.40\linewidth}>{\raggedleft\arraybackslash}p{0.14\linewidth}>{\raggedright\arraybackslash}p{0.24\linewidth}@{}}
\toprule
Block & Parameter & Estimate & Interval \\
\midrule\endhead
Pairwise association & $\beta_A$: adjusted slope of hip circumference $\to$ seated systolic blood pressure ($A \to M_2$) & $0.3904$ & $[0.357$, $0.4237]$ \\
\bottomrule
\end{longtable}}

\paragraph{12.\ Pairwise association: GGT--thigh-to-ankle PWV ($\pm$90 days)}
\textit{Design.} GGT and PWV are measured in the same period (PWV within 90 days before or after GGT, bounds inclusive); each participant contributes the pair with the closest dates, $n = 1340$. PWV is the mean of the available sides. Adjustment for approximate age (calendar year of the GGT measurement day minus birth year), recorded sex and recruitment cohort; GGT, PWV and age are kept on their original numerical scales, with sex and recruitment cohort as categorical terms.

\textit{Model.} $Y = \beta_0 + \beta_{M1} \cdot M_1 + \beta_{\mathrm{age}} \cdot \mathrm{age} + \gamma_{\mathrm{sex}} + \delta_{\mathrm{cohort}} + \varepsilon$ (linear regression; $M_1$ is GGT and $Y$ thigh-to-ankle PWV, both on original scales). $\beta_{M1}$ is the difference in PWV per 1 unit higher GGT at fixed age, sex and recruitment cohort.

\textit{Intervals.} 95\% confidence intervals.
{\footnotesize\setlength{\tabcolsep}{3pt}
\begin{longtable}{@{}>{\raggedright\arraybackslash}p{0.15\linewidth}>{\raggedright\arraybackslash}p{0.40\linewidth}>{\raggedleft\arraybackslash}p{0.14\linewidth}>{\raggedright\arraybackslash}p{0.24\linewidth}@{}}
\toprule
Block & Parameter & Estimate & Interval \\
\midrule\endhead
Pairwise association & $\beta_{M1}$: adjusted slope of GGT $\to$ thigh-to-ankle PWV ($M_1 \to Y$) & $0.003657$ & $[-0.0007756$, $0.00809]$ \\
\bottomrule
\end{longtable}}

\subsubsection*{Joint models}

\paragraph{13.\ Joint path model: hip circumference, GGT, systolic blood pressure and PWV ($\pm$90 days)}
\textit{Design.} Hip circumference, GGT, seated systolic blood pressure and PWV are measured in the same period, with each adjacent pair in this order differing by $-90$ to $+90$ days (bounds inclusive, either order); one measurement chain per participant. All four equations share one sample and one covariate set: approximate age (calendar year of the hip-measurement day minus birth year), recorded sex and recruitment cohort. Systolic blood pressure is the mean of the available values of the two seated readings, and PWV the mean of the available sides; continuous variables are kept on their original numerical scales.

\textit{Model.} $M_1 = a_{10} + a_1 A + \gamma_1^\top C + \varepsilon_1$; $M_2 = a_{20} + a_2 A + d_{21} M_1 + \gamma_2^\top C + \varepsilon_2$; $Y = b_0 + c' A + b_1 M_1 + b_2 M_2 + \gamma_3^\top C + \varepsilon_3$; $Y = r_0 + c A + \gamma_T^\top C + \varepsilon_T$. All four are linear regressions, with $A$, $M_1$, $M_2$ and $Y$ the hip circumference, GGT, seated systolic blood pressure and thigh-to-ankle PWV (original scales), and $C$ the covariates. $a_1$, $a_2$, $d_{21}$, $c'$, $b_1$ and $b_2$ are the differences in the response per 1 unit higher value of that variable, with the other terms of its equation held fixed; $c$ is the total slope of PWV on hip circumference, adjusted for $C$ only.

\textit{Intervals.} 95\% confidence intervals.
{\footnotesize\setlength{\tabcolsep}{3pt}
\begin{longtable}{@{}>{\raggedright\arraybackslash}p{0.15\linewidth}>{\raggedright\arraybackslash}p{0.40\linewidth}>{\raggedleft\arraybackslash}p{0.14\linewidth}>{\raggedright\arraybackslash}p{0.24\linewidth}@{}}
\toprule
Block & Parameter & Estimate & Interval \\
\midrule\endhead
$\pm$90-day window; GGT equation: hip circumference + covariates & $a_1$: slope of hip circumference $\to$ GGT ($A \to M_1$) & $0.3248$ & $[0.1972$, $0.4524]$ \\
$\pm$90-day window; systolic blood pressure equation: hip circumference + GGT + covariates & $a_2$: slope of hip circumference $\to$ seated systolic blood pressure, adjusted for GGT ($A \to M_2 \mid M_1$) & $0.4647$ & $[0.3737$, $0.5556]$ \\
 & $d_{21}$: slope of GGT $\to$ seated systolic blood pressure, adjusted for hip circumference ($M_1 \to M_2 \mid A$) & $0.03139$ & $[-0.00894$, $0.07171]$ \\
$\pm$90-day window; joint PWV equation: hip circumference + GGT + systolic blood pressure + covariates & $c'$: slope of hip circumference $\to$ PWV, adjusted for GGT and systolic blood pressure ($A \to Y \mid M_1, M_2$) & $0.00814$ & $[-0.001682$, $0.01796]$ \\
 & $b_1$: slope of GGT $\to$ PWV, adjusted for hip circumference and systolic blood pressure ($M_1 \to Y \mid A, M_2$) & $0.0005692$ & $[-0.002862$, $0.004001]$ \\
 & $b_2$: slope of seated systolic blood pressure $\to$ PWV, adjusted for hip circumference and GGT ($M_2 \to Y \mid A, M_1$) & $0.04917$ & $[0.04291$, $0.05543]$ \\
$\pm$90-day window; reduced PWV equation: hip circumference + covariates & $c$: total slope of hip circumference $\to$ PWV, adjusted for covariates only ($A \to Y$) & $0.03168$ & $[0.02133$, $0.04202]$ \\
\bottomrule
\end{longtable}}

\paragraph{14.\ Joint path model: hip circumference, GGT, systolic blood pressure and PWV ($\pm$30 days)}
\textit{Design.} Hip circumference, GGT, seated systolic blood pressure and PWV are measured in the same period, with each adjacent pair in this order differing by $-30$ to $+30$ days (bounds inclusive, either order); one measurement chain per participant, $n = 736$. All four equations share one sample and one covariate set: approximate age (calendar year of the hip-measurement day minus birth year), recorded sex and recruitment cohort. Systolic blood pressure is the mean of the available values of the two seated readings, and PWV the mean of the available sides; continuous variables are kept on their original numerical scales.

\textit{Model.} $M_1 = a_{10} + a_1 A + \gamma_1^\top C + \varepsilon_1$; $M_2 = a_{20} + a_2 A + d_{21} M_1 + \gamma_2^\top C + \varepsilon_2$; $Y = b_0 + c' A + b_1 M_1 + b_2 M_2 + \gamma_3^\top C + \varepsilon_3$; $Y = r_0 + c A + \gamma_T^\top C + \varepsilon_T$. All four are linear regressions, with $A$, $M_1$, $M_2$ and $Y$ the hip circumference, GGT, seated systolic blood pressure and thigh-to-ankle PWV (original scales), and $C$ the covariates. $a_1$, $a_2$, $d_{21}$, $c'$, $b_1$ and $b_2$ are the differences in the response per 1 unit higher value of that variable, with the other terms of its equation held fixed; $c$ is the total slope of PWV on hip circumference, adjusted for $C$ only.

\textit{Intervals.} 95\% confidence intervals.
{\footnotesize\setlength{\tabcolsep}{3pt}
\begin{longtable}{@{}>{\raggedright\arraybackslash}p{0.15\linewidth}>{\raggedright\arraybackslash}p{0.40\linewidth}>{\raggedleft\arraybackslash}p{0.14\linewidth}>{\raggedright\arraybackslash}p{0.24\linewidth}@{}}
\toprule
Block & Parameter & Estimate & Interval \\
\midrule\endhead
$\pm$30-day window; GGT equation: hip circumference + covariates & $a_1$: slope of hip circumference $\to$ GGT ($A \to M_1$) & $0.3805$ & $[0.1981$, $0.5629]$ \\
$\pm$30-day window; systolic blood pressure equation: hip circumference + GGT + covariates & $a_2$: slope of hip circumference $\to$ seated systolic blood pressure, adjusted for GGT ($A \to M_2 \mid M_1$) & $0.4444$ & $[0.3216$, $0.5672]$ \\
 & $d_{21}$: slope of GGT $\to$ seated systolic blood pressure, adjusted for hip circumference ($M_1 \to M_2 \mid A$) & $-0.03075$ & $[-0.07313$, $0.01163]$ \\
$\pm$30-day window; joint PWV equation: hip circumference + GGT + systolic blood pressure + covariates & $c'$: slope of hip circumference $\to$ PWV, adjusted for GGT and systolic blood pressure ($A \to Y \mid M_1, M_2$) & $0.006072$ & $[-0.007779$, $0.01992]$ \\
 & $b_1$: slope of GGT $\to$ PWV, adjusted for hip circumference and systolic blood pressure ($M_1 \to Y \mid A, M_2$) & $0.001025$ & $[-0.002878$, $0.004928]$ \\
 & $b_2$: slope of seated systolic blood pressure $\to$ PWV, adjusted for hip circumference and GGT ($M_2 \to Y \mid A, M_1$) & $0.05006$ & $[0.0427$, $0.05742]$ \\
$\pm$30-day window; reduced PWV equation: hip circumference + covariates & $c$: total slope of hip circumference $\to$ PWV, adjusted for covariates only ($A \to Y$) & $0.02812$ & $[0.01315$, $0.04309]$ \\
\bottomrule
\end{longtable}}

\subsubsection*{Reverse direction}

\paragraph{15.\ Reverse measurement order: earlier PWV and subsequently measured hip circumference}
\textit{Design.} PWV is measured first and hip circumference 31--365 days later (bounds inclusive); each participant contributes the pair with the shortest gap, $n = 18$. PWV is the mean of the available sides. Adjustment for approximate age (calendar year of the PWV measurement day minus birth year), recorded sex and recruitment cohort; PWV, hip circumference and age are kept on their original numerical scales, with sex and recruitment cohort as categorical terms.

\textit{Model.} Linear regression: $Y(\mathrm{earlier}) = \beta_0 + \beta_A \cdot A(\mathrm{later}) + \beta_{\mathrm{age}} \cdot \mathrm{age} + \gamma_{\mathrm{sex}} + \delta_{\mathrm{cohort}} + \varepsilon$ ($A$ is hip circumference and $Y$ thigh-to-ankle PWV, both on original scales). $\beta_A$ is the difference in the earlier PWV per 1 unit higher subsequently measured hip circumference at fixed age, sex and recruitment cohort.

\textit{Intervals.} 95\% confidence intervals.
{\footnotesize\setlength{\tabcolsep}{3pt}
\begin{longtable}{@{}>{\raggedright\arraybackslash}p{0.15\linewidth}>{\raggedright\arraybackslash}p{0.40\linewidth}>{\raggedleft\arraybackslash}p{0.14\linewidth}>{\raggedright\arraybackslash}p{0.24\linewidth}@{}}
\toprule
Block & Parameter & Estimate & Interval \\
\midrule\endhead
Reverse measurement order & $\beta_A$: reverse-measurement-order coefficient (hip circumference measured 31--365 days after PWV) & $-0.003739$ & $[-0.08307$, $0.07559]$ \\
\bottomrule
\end{longtable}}

\subsubsection*{Heterogeneity and interaction}

\paragraph{16.\ Hip circumference $\times$ GGT interaction: GGT $\to$ PWV slope at different hip-circumference levels}
\textit{Design.} Hip circumference, GGT and PWV are measured in strict date order: GGT 30--730 days after hip circumference and PWV 90--1095 days after GGT (bounds inclusive); one measurement chain per participant. Adjustment for approximate age at the hip measurement, sex and recruitment cohort. Hip circumference, GGT and PWV are standardized within the analysis sample, with sex and cohort as categorical terms; $a = -1$, $0$, $+1$ correspond to hip circumference 1 SD below the mean, at the mean and 1 SD above the mean.

\textit{Model.} $Y = \beta_0 + \beta_A \cdot A + \beta_M \cdot M + \beta_{AM} \cdot A \cdot M + \beta_C^\top \cdot C + \varepsilon_Y$ (linear regression; $A$, $M$ and $Y$ are the standardized hip circumference, GGT and thigh-to-ankle PWV, and $C$ is age, sex and recruitment cohort). $\beta_{AM}$ is the hip circumference $\times$ GGT interaction coefficient; the conditional standardized slope of GGT $\to$ PWV at hip circumference $a$ is $\beta_M + \beta_{AM} \cdot a$, $a \in \{-1, 0, +1\}$.

\textit{Intervals.} Bootstrap 95\% percentile intervals.
{\footnotesize\setlength{\tabcolsep}{3pt}
\begin{longtable}{@{}>{\raggedright\arraybackslash}p{0.15\linewidth}>{\raggedright\arraybackslash}p{0.40\linewidth}>{\raggedleft\arraybackslash}p{0.14\linewidth}>{\raggedright\arraybackslash}p{0.24\linewidth}@{}}
\toprule
Block & Parameter & Estimate & Interval \\
\midrule\endhead
Interaction coefficient & $\beta_{AM}$: hip circumference $\times$ GGT interaction coefficient in the PWV equation & $0.04719$ & $[-0.1142$, $0.2216]$ \\
Hip circumference 1 SD below the mean ($a = -1$) & $\beta_M - \beta_{AM}$: conditional standardized slope of GGT $\to$ PWV & $0.05179$ & $[-0.1865$, $0.2919]$ \\
Hip circumference at the mean ($a = 0$) & $\beta_M$: conditional standardized slope of GGT $\to$ PWV & $0.09898$ & $[-0.05159$, $0.2619]$ \\
Hip circumference 1 SD above the mean ($a = +1$) & $\beta_M + \beta_{AM}$: conditional standardized slope of GGT $\to$ PWV & $0.1462$ & $[-0.04735$, $0.3858]$ \\
\bottomrule
\end{longtable}}

\paragraph{17.\ Age heterogeneity: hip circumference $\to$ GGT $\to$ PWV path product}
\textit{Design.} Hip circumference, GGT and PWV are measured in strict date order: GGT 30--730 days after hip circumference and PWV 90--1095 days after GGT (bounds inclusive); one measurement chain per participant. Approximate age at the hip measurement enters both as a main effect and as the modifier, with additional adjustment for sex and recruitment cohort (categorical terms). Hip circumference, GGT, PWV and age are standardized within the analysis sample; $g = -1$, $0$, $+1$ correspond to age 1 SD below the mean, at the mean and 1 SD above the mean.

\textit{Model.} $M = \alpha_0 + \alpha_A \cdot A + \alpha_G \cdot G + \alpha_{AG} \cdot A \cdot G + \alpha_C^\top \cdot C + \varepsilon_M$; $Y = \beta_0 + \beta_A \cdot A + \beta_M \cdot M + \beta_G \cdot G + \beta_{AG} \cdot A \cdot G + \beta_{MG} \cdot M \cdot G + \beta_C^\top \cdot C + \varepsilon_Y$ (linear regressions; $A$, $M$, $Y$ and $G$ are the standardized hip circumference, GGT, thigh-to-ankle PWV and approximate age, and $C$ is sex and recruitment cohort). The path product at age $g$ is $\theta(g) = (\alpha_A + \alpha_{AG} \cdot g) \cdot (\beta_M + \beta_{MG} \cdot g)$, $g \in \{-1, 0, +1\}$; the age contrast is $\theta(+1) - \theta(-1)$.

\textit{Intervals.} Bootstrap 95\% percentile intervals.
{\footnotesize\setlength{\tabcolsep}{3pt}
\begin{longtable}{@{}>{\raggedright\arraybackslash}p{0.15\linewidth}>{\raggedright\arraybackslash}p{0.40\linewidth}>{\raggedleft\arraybackslash}p{0.14\linewidth}>{\raggedright\arraybackslash}p{0.24\linewidth}@{}}
\toprule
Block & Parameter & Estimate & Interval \\
\midrule\endhead
Age 1 SD below the mean ($g = -1$) & $\theta(-1) = (\alpha_A - \alpha_{AG}) \cdot (\beta_M - \beta_{MG})$: path product & $0.008454$ & $[-0.007739$, $0.04569]$ \\
Age at the mean ($g = 0$) & $\theta(0) = \alpha_A \cdot \beta_M$: path product & $0.006302$ & $[-0.008361$, $0.02503]$ \\
Age 1 SD above the mean ($g = +1$) & $\theta(+1) = (\alpha_A + \alpha_{AG}) \cdot (\beta_M + \beta_{MG})$: path product & $0.002198$ & $[-0.03604$, $0.0293]$ \\
Age-heterogeneity contrast & Age contrast $\theta(+1) - \theta(-1)$ & $-0.006255$ & $[-0.06058$, $0.02482]$ \\
\bottomrule
\end{longtable}}

\paragraph{18.\ Age heterogeneity: hip circumference $\to$ GGT $\to$ seated systolic blood pressure $\to$ PWV}
\textit{Design.} Each GGT test serves as the anchor: hip circumference on the GGT day or within the 0--365 days before it, seated systolic blood pressure within 30 days before or after GGT (either order), and PWV on the systolic blood pressure day or within the 0--365 days after it (bounds inclusive); a participant may contribute multiple chains, $n = 64$. Approximate age at the hip measurement is the only prespecified modifier, interacting with all path slopes in each equation; sex enters as an adjustment term only. Continuous variables are standardized within the analysis sample.

\textit{Model.} $M_1 = \alpha_1 + (a_{10} + a_{11} \cdot \mathrm{Age}) A + u_1 \cdot \mathrm{Age} + \gamma_1^\top \mathrm{Sex} + \varepsilon_1$; $M_2 = \alpha_2 + (d_{210} + d_{211} \cdot \mathrm{Age}) M_1 + (a_{20} + a_{21} \cdot \mathrm{Age}) A + u_2 \cdot \mathrm{Age} + \gamma_2^\top \mathrm{Sex} + \varepsilon_2$; $Y = \alpha_3 + (b_{20} + b_{21} \cdot \mathrm{Age}) M_2 + (b_{10} + b_{11} \cdot \mathrm{Age}) M_1 + (c_0 + c_1 \cdot \mathrm{Age}) A + u_3 \cdot \mathrm{Age} + \gamma_3^\top \mathrm{Sex} + \varepsilon_3$ (linear regression; $A$ hip circumference, $M_1$ GGT, $M_2$ seated systolic blood pressure, $Y$ thigh-to-ankle PWV, $\mathrm{Age}$ age, all as $z$-scores). Reported are the age-conditional serial path index $\theta(q) = (a_{10} + a_{11} q)(d_{210} + d_{211} q)(b_{20} + b_{21} q)$ at the 25th, 50th and 75th percentiles of standardized age in the analysis sample (Q25, Q50, Q75), and the difference $\theta(Q75) - \theta(Q25)$.

\textit{Intervals.} 95\% confidence intervals.
{\footnotesize\setlength{\tabcolsep}{3pt}
\begin{longtable}{@{}>{\raggedright\arraybackslash}p{0.15\linewidth}>{\raggedright\arraybackslash}p{0.40\linewidth}>{\raggedleft\arraybackslash}p{0.14\linewidth}>{\raggedright\arraybackslash}p{0.24\linewidth}@{}}
\toprule
Block & Parameter & Estimate & Interval \\
\midrule\endhead
Age-conditional serial path index & $\theta(Q25)$: serial path index at the 25th age percentile & $-0.0006807$ & $[-0.0119$, $0.01054]$ \\
 & $\theta(Q50)$: serial path index at the median age & $-0.002768$ & $[-0.01772$, $0.01218]$ \\
 & $\theta(Q75)$: serial path index at the 75th age percentile & $-0.003511$ & $[-0.04825$, $0.04123]$ \\
Age-heterogeneity contrast & $\theta(Q75) - \theta(Q25)$: difference between the serial path indices at the 75th and 25th age percentiles & $-0.00283$ & $[-0.04701$, $0.04135]$ \\
\bottomrule
\end{longtable}}

\paragraph{19.\ Sex modification: hip circumference--PWV slope within each sex category}
\textit{Design.} Hip circumference and PWV are measured in the same period (PWV within 90 days before or after hip circumference, bounds inclusive); each participant contributes the pair with the shortest gap, $n = 9385$ (first sex category 4894, second 4491). PWV is the mean of the available sides. The model includes the sex main effect and the hip circumference $\times$ sex interaction, with additional adjustment for approximate age (calendar year of the hip-measurement day minus birth year) and recruitment cohort; continuous variables are kept on their original numerical scales, with recruitment cohort as a categorical term.

\textit{Model.} $Y = \beta_0 + \beta_A \cdot A + \gamma_{\mathrm{sex}} + \beta_{A\times\mathrm{sex}} \cdot A \cdot \mathrm{sex} + \beta_{\mathrm{age}} \cdot \mathrm{age} + \delta_{\mathrm{cohort}} + \varepsilon$ (linear regression; $A$ is hip circumference and $Y$ thigh-to-ankle PWV, both on original scales). $\beta_A(s) = \beta_A + \beta_{A\times s}$ is the difference in PWV per 1 unit higher hip circumference within sex category $s$ at fixed age and recruitment cohort.

\textit{Intervals.} 95\% confidence intervals.
{\footnotesize\setlength{\tabcolsep}{3pt}
\begin{longtable}{@{}>{\raggedright\arraybackslash}p{0.15\linewidth}>{\raggedright\arraybackslash}p{0.40\linewidth}>{\raggedleft\arraybackslash}p{0.14\linewidth}>{\raggedright\arraybackslash}p{0.24\linewidth}@{}}
\toprule
Block & Parameter & Estimate & Interval \\
\midrule\endhead
First sex category & $\beta_A$(first category): slope of hip circumference $\to$ PWV in the first sex category & $0.0205$ & $[0.01571$, $0.02529]$ \\
Second sex category & $\beta_A$(second category): slope of hip circumference $\to$ PWV in the second sex category & $0.0306$ & $[0.02402$, $0.03719]$ \\
\bottomrule
\end{longtable}}

\subsubsection*{Sensitivity analyses}

\paragraph{20.\ Narrow-window single-mediator pathway: hip circumference $\to$ GGT $\to$ PWV}
\textit{Design.} Hip circumference, GGT and PWV are measured in strict date order, with windows narrower than those of the main pathway: GGT 90--365 days after hip circumference and PWV 180--730 days after GGT (bounds inclusive); one measurement chain per participant. Adjustment for approximate age at the hip measurement, sex and recruitment cohort. PWV is the bilateral mean, taken when both left and right values are valid. Hip circumference, GGT and PWV are standardized within the analysis sample, with sex and cohort as categorical terms.

\textit{Model.} $M = \alpha_0 + \alpha_A \cdot A + \alpha_C^\top \cdot C + \varepsilon_M$; $Y = \beta_0 + \beta_A \cdot A + \beta_M \cdot M + \beta_C^\top \cdot C + \varepsilon_Y$; $Y = \tau_0 + \tau_A \cdot A + \tau_C^\top \cdot C + \varepsilon_T$. All three are linear regressions, with $A$, $M$ and $Y$ the standardized hip circumference, GGT and thigh-to-ankle PWV, and $C$ the covariates. $\alpha_A$ is the SD change in GGT per 1 SD higher hip circumference; $\beta_M$ and $\beta_A$ are the mutually adjusted standardized slopes of PWV on GGT and on hip circumference; $\tau_A$ is the total standardized slope of PWV on hip circumference, adjusted for $C$ only; the path product is $\theta = \alpha_A \cdot \beta_M$; $\mathrm{RV}(b) = \tfrac{1}{2}[\sqrt{f^4 + 4f^2} - f^2]$, $f = |t_b|/\sqrt{\nu}$, is the equal-strength partial $R^2$ that a single omitted confounder must have with the regressor and with the equation outcome to reduce coefficient $b$ to 0 ($t_b$ is the $t$ statistic of $b$ and $\nu$ the residual degrees of freedom).

\textit{Intervals.} Bootstrap 95\% percentile intervals.
{\footnotesize\setlength{\tabcolsep}{3pt}
\begin{longtable}{@{}>{\raggedright\arraybackslash}p{0.15\linewidth}>{\raggedright\arraybackslash}p{0.40\linewidth}>{\raggedleft\arraybackslash}p{0.14\linewidth}>{\raggedright\arraybackslash}p{0.24\linewidth}@{}}
\toprule
Block & Parameter & Estimate & Interval \\
\midrule\endhead
Path components (narrow window) & $\alpha_A$: standardized slope of hip circumference $\to$ GGT ($A \to M$) & $-0.03482$ & $[-0.1934$, $0.1222]$ \\
 & $\beta_M$: standardized slope of GGT $\to$ thigh-to-ankle PWV, adjusted for hip circumference ($M \to Y \mid A$) & $0.0342$ & $[-0.1344$, $0.2758]$ \\
 & $\beta_A$: standardized slope of hip circumference $\to$ PWV, adjusted for GGT ($A \to Y \mid M$) & $0.004702$ & $[-0.1981$, $0.2241]$ \\
 & $\tau_A$: adjusted total standardized slope of hip circumference $\to$ PWV ($A \to Y$) & $0.003511$ & $[-0.196$, $0.2235]$ \\
Path index (narrow window) & $\theta = \alpha_A \cdot \beta_M$: path-product association index & $-0.001191$ & $[-0.01615$, $0.02347]$ \\
Omitted-confounder robustness values (narrow window) & $\mathrm{RV}(\alpha_A)$: equal-strength omitted-confounder partial $R^2$ required to reduce $\alpha_A$ to 0 & $0.03538$ & $[0.002378$, $0.1905]$ \\
 & $\mathrm{RV}(\beta_M)$: equal-strength omitted-confounder partial $R^2$ required to reduce $\beta_M$ to 0 & $0.03552$ & $[0.003001$, $0.2679]$ \\
 & $\mathrm{RV}(\beta_A)$: equal-strength omitted-confounder partial $R^2$ required to reduce $\beta_A$ to 0 & $0.005129$ & $[0.002594$, $0.2402]$ \\
\bottomrule
\end{longtable}}

\paragraph{21.\ Additional adjustment for smoking and alcohol: hip circumference $\to$ GGT $\to$ PWV}
\textit{Design.} Hip circumference, GGT and PWV are measured in strict date order: GGT 30--730 days after hip circumference and PWV 90--1095 days after GGT (bounds inclusive); one measurement chain per participant, $n = 167$. Adjustment for approximate age at the hip measurement, sex, recruitment cohort, and the most recent smoking status and alcohol-consumption frequency on the hip-measurement day or within the preceding 730 days. Hip circumference, GGT and PWV are standardized within the analysis sample, with sex, cohort, smoking and alcohol as categorical terms.

\textit{Model.} $M = \alpha_0 + \alpha_A \cdot A + \alpha_C^\top \cdot C + \varepsilon_M$; $Y = \beta_0 + \beta_A \cdot A + \beta_M \cdot M + \beta_C^\top \cdot C + \varepsilon_Y$; $Y = \tau_0 + \tau_A \cdot A + \tau_C^\top \cdot C + \varepsilon_T$. All three are linear regressions, with $A$, $M$ and $Y$ the standardized hip circumference, GGT and thigh-to-ankle PWV, and $C$ age, sex, recruitment cohort, smoking status and alcohol-consumption frequency. $\alpha_A$ is the SD change in GGT per 1 SD higher hip circumference; $\beta_M$ and $\beta_A$ are the mutually adjusted standardized slopes of PWV on GGT and on hip circumference; $\tau_A$ is the total standardized slope of PWV on hip circumference, adjusted for $C$ only; the path product is $\theta = \alpha_A \cdot \beta_M$; $\mathrm{RV}(b) = \tfrac{1}{2}[\sqrt{f^4 + 4f^2} - f^2]$, $f = |t_b|/\sqrt{\nu}$, is the equal-strength partial $R^2$ that a single omitted confounder must have with the regressor and with the equation outcome to reduce coefficient $b$ to 0 ($t_b$ is the $t$ statistic of $b$ and $\nu$ the residual degrees of freedom).

\textit{Intervals.} Bootstrap 95\% percentile intervals.
{\footnotesize\setlength{\tabcolsep}{3pt}
\begin{longtable}{@{}>{\raggedright\arraybackslash}p{0.15\linewidth}>{\raggedright\arraybackslash}p{0.40\linewidth}>{\raggedleft\arraybackslash}p{0.14\linewidth}>{\raggedright\arraybackslash}p{0.24\linewidth}@{}}
\toprule
Block & Parameter & Estimate & Interval \\
\midrule\endhead
Path components (additionally adjusted for smoking and alcohol) & $\alpha_A$: standardized slope of hip circumference $\to$ GGT ($A \to M$) & $0.0772$ & $[-0.05856$, $0.2324]$ \\
 & $\beta_M$: standardized slope of GGT $\to$ thigh-to-ankle PWV, adjusted for hip circumference ($M \to Y \mid A$) & $0.1271$ & $[-0.02644$, $0.2707]$ \\
 & $\beta_A$: standardized slope of hip circumference $\to$ PWV, adjusted for GGT ($A \to Y \mid M$) & $0.05328$ & $[-0.08786$, $0.2035]$ \\
 & $\tau_A$: adjusted total standardized slope of hip circumference $\to$ PWV ($A \to Y$) & $0.0631$ & $[-0.08415$, $0.2215]$ \\
Path index (additionally adjusted for smoking and alcohol) & $\theta = \alpha_A \cdot \beta_M$: path-product association index & $0.009816$ & $[-0.01051$, $0.03508]$ \\
Omitted-confounder robustness values (additionally adjusted for smoking and alcohol) & $\mathrm{RV}(\alpha_A)$: equal-strength omitted-confounder partial $R^2$ required to reduce $\alpha_A$ to 0 & $0.07492$ & $[0.005551$, $0.2074]$ \\
 & $\mathrm{RV}(\beta_M)$: equal-strength omitted-confounder partial $R^2$ required to reduce $\beta_M$ to 0 & $0.1239$ & $[0.009596$, $0.2511]$ \\
 & $\mathrm{RV}(\beta_A)$: equal-strength omitted-confounder partial $R^2$ required to reduce $\beta_A$ to 0 & $0.05424$ & $[0.003884$, $0.1964]$ \\
\bottomrule
\end{longtable}}

\paragraph{22.\ Wide-window serial pathway: hip circumference $\to$ GGT $\to$ seated systolic blood pressure $\to$ PWV}
\textit{Design.} Each GGT test serves as the anchor: hip circumference on the GGT day or within the 0--730 days before it, seated systolic blood pressure within 90 days before or after GGT (either order), and PWV on the systolic blood pressure day or within the 0--730 days after it (bounds inclusive); a participant may contribute multiple chains. Adjustment for approximate age at the hip measurement and sex. Systolic blood pressure is the mean of the available values of the two seated readings, and PWV the mean of the available sides. Continuous variables are standardized within the analysis sample, with sex as a categorical term.

\textit{Model.} $M_1 = \alpha_1 + a_1 A + \gamma_1^\top C + \varepsilon_1$; $M_2 = \alpha_2 + a_2 A + d_{21} M_1 + \gamma_2^\top C + \varepsilon_2$; $Y = \alpha_3 + c' A + b_1 M_1 + b_2 M_2 + \gamma_3^\top C + \varepsilon_3$ (linear regression; $A$ hip circumference, $M_1$ GGT, $M_2$ seated systolic blood pressure, $Y$ thigh-to-ankle PWV, all as $z$-scores; $C$ is age and sex). Reported are the serial path index $a_1 \cdot d_{21} \cdot b_2$ and six standardized coefficients $a_1$, $a_2$, $d_{21}$, $b_1$, $b_2$, $c'$ (the SD change in the outcome per 1 SD higher corresponding variable in each equation).

\textit{Intervals.} 95\% confidence intervals.
{\footnotesize\setlength{\tabcolsep}{3pt}
\begin{longtable}{@{}>{\raggedright\arraybackslash}p{0.15\linewidth}>{\raggedright\arraybackslash}p{0.40\linewidth}>{\raggedleft\arraybackslash}p{0.14\linewidth}>{\raggedright\arraybackslash}p{0.24\linewidth}@{}}
\toprule
Block & Parameter & Estimate & Interval \\
\midrule\endhead
Serial path index (wide window) & Serial path index $a_1 \cdot d_{21} \cdot b_2$ (hip circumference $\to$ GGT $\to$ seated systolic blood pressure $\to$ PWV) & $0.00223$ & $[-0.009568$, $0.01403]$ \\
Path coefficients (wide window) & Standardized slope $a_1$ (hip circumference $\to$ GGT) & $0.1702$ & $[-0.02541$, $0.3658]$ \\
 & Standardized slope $d_{21}$ (GGT $\to$ seated systolic blood pressure, controlling for hip circumference) & $0.02221$ & $[-0.08878$, $0.1332]$ \\
 & Standardized slope $b_2$ (seated systolic blood pressure $\to$ PWV, controlling for hip circumference and GGT) & $0.5898$ & $[0.4362$, $0.7433]$ \\
 & Standardized slope $a_2$ (hip circumference $\to$ seated systolic blood pressure, controlling for GGT) & $0.2066$ & $[0.06747$, $0.3458]$ \\
 & Standardized slope $b_1$ (GGT $\to$ PWV, controlling for hip circumference and seated systolic blood pressure) & $0.01626$ & $[-0.1075$, $0.14]$ \\
 & Conditional slope $c'$ (hip circumference $\to$ PWV, controlling for both mediators) & $0.06357$ & $[-0.1205$, $0.2476]$ \\
\bottomrule
\end{longtable}}

\paragraph{23.\ Additional adjustment for alcohol and smoking: hip circumference $\to$ GGT $\to$ seated systolic blood pressure $\to$ PWV}
\textit{Design.} Each GGT test serves as the anchor: hip circumference on the GGT day or within the 0--365 days before it, seated systolic blood pressure within 30 days before or after GGT (either order), and PWV on the systolic blood pressure day or within the 0--365 days after it (bounds inclusive); a participant may contribute multiple chains. Adjustment for approximate age on the hip-measurement day, sex, and the most recent alcohol-consumption frequency and smoking status on that day or within the preceding 730 days (the latter three as unordered categorical terms). Continuous variables are standardized within the analysis sample.

\textit{Model.} $M_1 = \alpha_1 + a_1 A + \gamma_1^\top C + \varepsilon_1$; $M_2 = \alpha_2 + a_2 A + d_{21} M_1 + \gamma_2^\top C + \varepsilon_2$; $Y = \alpha_3 + c' A + b_1 M_1 + b_2 M_2 + \gamma_3^\top C + \varepsilon_3$ (linear regression; $A$ hip circumference, $M_1$ GGT, $M_2$ seated systolic blood pressure, $Y$ thigh-to-ankle PWV, all as $z$-scores; $C$ is age, sex, alcohol-consumption frequency and smoking status). Reported are the serial path index $a_1 \cdot d_{21} \cdot b_2$ and six standardized coefficients $a_1$, $a_2$, $d_{21}$, $b_1$, $b_2$, $c'$ (the SD change in the outcome per 1 SD higher corresponding variable at equal age, sex, alcohol consumption and smoking).

\textit{Intervals.} 95\% confidence intervals.
{\footnotesize\setlength{\tabcolsep}{3pt}
\begin{longtable}{@{}>{\raggedright\arraybackslash}p{0.15\linewidth}>{\raggedright\arraybackslash}p{0.40\linewidth}>{\raggedleft\arraybackslash}p{0.14\linewidth}>{\raggedright\arraybackslash}p{0.24\linewidth}@{}}
\toprule
Block & Parameter & Estimate & Interval \\
\midrule\endhead
Serial path index (additionally adjusted for alcohol and smoking) & Serial path index $a_1 \cdot d_{21} \cdot b_2$ (hip circumference $\to$ GGT $\to$ seated systolic blood pressure $\to$ PWV) & $-0.04819$ & $[-0.1661$, $0.0697]$ \\
Path coefficients (additionally adjusted for alcohol and smoking) & Standardized slope $a_1$ (hip circumference $\to$ GGT) & $0.1521$ & $[-0.2224$, $0.5266]$ \\
 & Standardized slope $d_{21}$ (GGT $\to$ seated systolic blood pressure, controlling for hip circumference) & $-0.3989$ & $[-0.7473$, $-0.05043]$ \\
 & Standardized slope $b_2$ (seated systolic blood pressure $\to$ PWV, controlling for hip circumference and GGT) & $0.7943$ & $[0.6614$, $0.9272]$ \\
 & Standardized slope $a_2$ (hip circumference $\to$ seated systolic blood pressure, controlling for GGT) & $0.162$ & $[-0.2194$, $0.5435]$ \\
 & Standardized slope $b_1$ (GGT $\to$ PWV, controlling for hip circumference and seated systolic blood pressure) & $0.3219$ & $[0.1398$, $0.5041]$ \\
 & Conditional slope $c'$ (hip circumference $\to$ PWV, controlling for both mediators) & $-0.1082$ & $[-0.2386$, $0.02217]$ \\
\bottomrule
\end{longtable}}

\paragraph{24.\ Joint path model with same-day height adjustment ($\pm$90 days)}
\textit{Design.} Hip circumference, GGT, seated systolic blood pressure and PWV are measured in the same period, with each adjacent pair in this order differing by $-90$ to $+90$ days (bounds inclusive, either order); one measurement chain per participant. All four equations share one sample and one covariate set: approximate age (calendar year of the hip-measurement day minus birth year), recorded sex, recruitment cohort, and height measured on the same day as the selected hip circumference. Systolic blood pressure is the mean of the available values of the two seated readings, and PWV the mean of the available sides; continuous variables are kept on their original numerical scales.

\textit{Model.} $M_1 = a_{10} + a_1 A + \gamma_1^\top C + \varepsilon_1$; $M_2 = a_{20} + a_2 A + d_{21} M_1 + \gamma_2^\top C + \varepsilon_2$; $Y = b_0 + c' A + b_1 M_1 + b_2 M_2 + \gamma_3^\top C + \varepsilon_3$; $Y = r_0 + c A + \gamma_T^\top C + \varepsilon_T$. All four are linear regressions, with $A$, $M_1$, $M_2$ and $Y$ the hip circumference, GGT, seated systolic blood pressure and thigh-to-ankle PWV (original scales), and $C$ the covariates (including same-day height). $a_1$, $a_2$, $d_{21}$, $c'$, $b_1$ and $b_2$ are the differences in the response per 1 unit higher value of that variable, with the other terms of its equation held fixed; $c$ is the total slope of PWV on hip circumference, adjusted for $C$ only.

\textit{Intervals.} 95\% confidence intervals.
{\footnotesize\setlength{\tabcolsep}{3pt}
\begin{longtable}{@{}>{\raggedright\arraybackslash}p{0.15\linewidth}>{\raggedright\arraybackslash}p{0.40\linewidth}>{\raggedleft\arraybackslash}p{0.14\linewidth}>{\raggedright\arraybackslash}p{0.24\linewidth}@{}}
\toprule
Block & Parameter & Estimate & Interval \\
\midrule\endhead
$\pm$90-day window, with height; GGT equation: hip circumference + covariates & $a_1$: slope of hip circumference $\to$ GGT ($A \to M_1$) & $0.346$ & $[0.2166$, $0.4754]$ \\
$\pm$90-day window, with height; systolic blood pressure equation: hip circumference + GGT + covariates & $a_2$: slope of hip circumference $\to$ seated systolic blood pressure, adjusted for GGT ($A \to M_2 \mid M_1$) & $0.463$ & $[0.3684$, $0.5575]$ \\
 & $d_{21}$: slope of GGT $\to$ seated systolic blood pressure, adjusted for hip circumference ($M_1 \to M_2 \mid A$) & $0.03118$ & $[-0.009068$, $0.07144]$ \\
$\pm$90-day window, with height; joint PWV equation: hip circumference + GGT + systolic blood pressure + covariates & $c'$: slope of hip circumference $\to$ PWV, adjusted for GGT and systolic blood pressure ($A \to Y \mid M_1, M_2$) & $0.003461$ & $[-0.006608$, $0.01353]$ \\
 & $b_1$: slope of GGT $\to$ PWV, adjusted for hip circumference and systolic blood pressure ($M_1 \to Y \mid A, M_2$) & $0.0008489$ & $[-0.002652$, $0.00435]$ \\
 & $b_2$: slope of seated systolic blood pressure $\to$ PWV, adjusted for hip circumference and GGT ($M_2 \to Y \mid A, M_1$) & $0.04906$ & $[0.04287$, $0.05525]$ \\
$\pm$90-day window, with height; reduced PWV equation: hip circumference + covariates & $c$: total slope of hip circumference $\to$ PWV, adjusted for covariates only ($A \to Y$) & $0.027$ & $[0.01632$, $0.03768]$ \\
\bottomrule
\end{longtable}}

\paragraph{25.\ Hip-measurement-day panel: hip circumference--PWV association}
\textit{Design.} Hip circumference and PWV are measured in the same period (PWV within 90 days before or after hip circumference, bounds inclusive); each hip-measurement day is paired once with the PWV of the closest date, and a participant may contribute multiple rows, $n = 10917$ rows. PWV is the mean of the available sides. Adjustment for approximate age (calendar year of the hip-measurement day minus birth year), recorded sex and recruitment cohort; hip circumference, PWV and age are kept on their original numerical scales, with sex and recruitment cohort as categorical terms.

\textit{Model.} $Y = \beta_0 + \beta_A \cdot A + \beta_{\mathrm{age}} \cdot \mathrm{age} + \gamma_{\mathrm{sex}} + \delta_{\mathrm{cohort}} + \varepsilon$ (linear regression; each row is one participant $\times$ hip-measurement day, $A$ is hip circumference and $Y$ thigh-to-ankle PWV, both on original scales). $\beta_A$ is the difference in PWV per 1 unit higher hip circumference at fixed age, sex and recruitment cohort.

\textit{Intervals.} 95\% confidence intervals.
{\footnotesize\setlength{\tabcolsep}{3pt}
\begin{longtable}{@{}>{\raggedright\arraybackslash}p{0.15\linewidth}>{\raggedright\arraybackslash}p{0.40\linewidth}>{\raggedleft\arraybackslash}p{0.14\linewidth}>{\raggedright\arraybackslash}p{0.24\linewidth}@{}}
\toprule
Block & Parameter & Estimate & Interval \\
\midrule\endhead
Hip-measurement-day panel & $\beta_A$: adjusted slope of hip circumference $\to$ thigh-to-ankle PWV (one row per hip-measurement day) & $0.0254$ & $[0.02174$, $0.02906]$ \\
\bottomrule
\end{longtable}}

\FloatBarrier

\subsection{Gynoid fat--LDL--SBP--PWV}
\label{app:rq-designs:bdd6}

\paragraph{Research question and measures.}
The research question concerns the association of gynoid fat mass with lower-limb arterial stiffness through low-density lipoprotein cholesterol and systolic blood pressure, with data from the HPP cohort. The exposure is gynoid fat mass from the body-composition measurement; the mediators are low-density lipoprotein cholesterol (LDL-C) from blood tests and seated systolic blood pressure (mean of the available values of two seated readings); the outcome is thigh-to-ankle pulse wave velocity (PWV, mean of the available sides). Unless stated otherwise, coefficients are unstandardized slopes in original units, that is, the change in the dependent variable per 1-unit increase in the independent variable.

\subsubsection*{Primary pathways}

\paragraph{1.\ Single-mediator path: gynoid fat mass $\to$ LDL-C $\to$ thigh-to-ankle PWV}
\textit{Design.} Gynoid fat mass, LDL-C and PWV measured in strict date order: LDL-C 30--730 days after the gynoid fat mass measurement, PWV 90--1095 days after LDL-C (all bounds inclusive); one measurement chain per participant. Adjustment for approximate age at the gynoid fat mass measurement (computed from calendar years) and sex (categorical term). PWV is the mean of the available sides.

\textit{Model.} $M = \alpha_0 + \alpha_A \cdot A + \alpha_C^{\top} C + \varepsilon_M$; $Y = \beta_0 + \beta_A \cdot A + \beta_M \cdot M + \beta_C^{\top} C + \varepsilon_Y$; $Y = \gamma_0 + \gamma_A \cdot A + \gamma_C^{\top} C + \varepsilon_T$. All three equations are linear regressions; $A$, $M$ and $Y$ are gynoid fat mass, LDL-C and thigh-to-ankle PWV (original units), and $C$ is age and sex. $\alpha_A$ is the change in LDL-C per 1-unit increase in gynoid fat mass; $\beta_M$ and $\beta_A$ are the mutually adjusted slopes of LDL-C and gynoid fat mass on PWV; $\gamma_A$ is the total slope of gynoid fat mass on PWV with adjustment for $C$ only; path product $\theta = \alpha_A \cdot \beta_M$.

\textit{Intervals.} Bootstrap 95\% percentile intervals.
{\footnotesize\setlength{\tabcolsep}{3pt}
\begin{longtable}{@{}>{\raggedright\arraybackslash}p{0.15\linewidth}>{\raggedright\arraybackslash}p{0.40\linewidth}>{\raggedleft\arraybackslash}p{0.14\linewidth}>{\raggedright\arraybackslash}p{0.24\linewidth}@{}}
\toprule
Block & Parameter & Estimate & Interval \\
\midrule\endhead
Path components & $\alpha_A$: slope of gynoid fat mass $\to$ LDL-C ($A \to M$) & $-0.00476$ & $[-0.01748$, $0.0039]$ \\
 & $\beta_M$: slope of LDL-C $\to$ PWV, adjusted for gynoid fat mass ($M \to Y \mid A$) & $0.0001116$ & $[-0.009117$, $0.002605]$ \\
 & $\beta_A$: slope of gynoid fat mass $\to$ PWV, adjusted for LDL-C ($A \to Y \mid M$) & $0.0001196$ & $[3.986\times10^{-6}$, $0.0002663]$ \\
 & $\gamma_A$: adjusted total slope of gynoid fat mass $\to$ PWV ($A \to Y$) & $0.000119$ & $[-1.639\times10^{-6}$, $0.000255]$ \\
Path index & $\theta = \alpha_A \cdot \beta_M$: path-product association index & $-5.313\times10^{-7}$ & $[-4.118\times10^{-5}$, $1.331\times10^{-5}]$ \\
\bottomrule
\end{longtable}}

\paragraph{2.\ Serial path: gynoid fat mass $\to$ LDL cholesterol $\to$ systolic blood pressure $\to$ PWV}
\textit{Design.} Each gynoid fat measurement date as anchor; nearest measurement of LDL cholesterol within 180 days before or after the anchor, of seated systolic blood pressure within 90 days before or after that LDL, and of PWV within 14 days before or after that systolic blood pressure (either order, bounds inclusive); one row per anchor date. All four equations share the same sample, with adjustment for approximate age (anchor year minus birth year) and recorded sex. Systolic blood pressure is the mean of the available values of two seated readings, PWV the mean of the available sides; variables not standardized.

\textit{Model.} $M_1 = i_1 + a \cdot A + \gamma_1^{\top} C + \varepsilon_1$; $M_2 = i_2 + d \cdot A + b \cdot M_1 + \gamma_2^{\top} C + \varepsilon_2$; $Y = i_3 + c' \cdot A + e \cdot M_1 + f \cdot M_2 + \gamma_3^{\top} C + \varepsilon_3$; $Y = i_T + \tau \cdot A + \gamma_T^{\top} C + \varepsilon_T$. All four equations are linear regressions; $A$, $M_1$, $M_2$ and $Y$ are gynoid fat mass, LDL cholesterol, seated systolic blood pressure and thigh-to-ankle PWV (original units), and $C$ is age and sex. Coefficients are the change in the dependent variable per 1 original unit higher in the independent variable; reported: the seven coefficients, plus the serial path product $a \cdot b \cdot f$, the LDL-only $a \cdot e$, the systolic-blood-pressure-only $d \cdot f$, the conditional slope $c'$ and the total association slope $\tau$.

\textit{Intervals.} 95\% confidence intervals. $P$ values not corrected for multiple comparisons.
{\footnotesize\setlength{\tabcolsep}{3pt}
\begin{longtable}{@{}>{\raggedright\arraybackslash}p{0.14\linewidth}>{\raggedright\arraybackslash}p{0.29\linewidth}>{\raggedleft\arraybackslash}p{0.13\linewidth}>{\raggedright\arraybackslash}p{0.22\linewidth}>{\raggedleft\arraybackslash}p{0.14\linewidth}@{}}
\toprule
Block & Parameter & Estimate & Interval & $P$ \\
\midrule\endhead
Path decomposition & $a \cdot b \cdot f$: serial path product gynoid fat mass $\to$ LDL cholesterol $\to$ seated systolic blood pressure $\to$ PWV & $2.462\times10^{-7}$ & $[-1.462\times10^{-6}$, $1.954\times10^{-6}]$ & \mbox{$0.7775$} \\
 & $a \cdot e$: path product through LDL cholesterol only & $3.904\times10^{-6}$ & $[-3.341\times10^{-7}$, $8.141\times10^{-6}]$ & \mbox{$0.071$} \\
 & $d \cdot f$: path product through seated systolic blood pressure only & $0.0001138$ & $[8.628\times10^{-5}$, $0.0001414]$ & \mbox{$5.797\times10^{-16}$} \\
 & $c'$: conditional slope of gynoid fat mass $\to$ PWV, adjusted for both mediators & $2.84\times10^{-5}$ & $[-2.536\times10^{-5}$, $8.217\times10^{-5}]$ & \mbox{$0.3005$} \\
 & $\tau$: total association slope of gynoid fat mass $\to$ PWV & $0.0001464$ & $[8.821\times10^{-5}$, $0.0002046]$ & \mbox{$8.165\times10^{-7}$} \\
Path components & $a$: slope of gynoid fat mass $\to$ LDL cholesterol ($A \to M_1$) & $0.001877$ & $[0.0005897$, $0.003165]$ & \mbox{$0.004268$} \\
 & $d$: slope of gynoid fat mass $\to$ seated systolic blood pressure, adjusted for LDL ($A \to M_2 \mid M_1$) & $0.002203$ & $[0.001703$, $0.002704]$ & \mbox{$6.093\times10^{-18}$} \\
 & $b$: slope of LDL cholesterol $\to$ seated systolic blood pressure, adjusted for gynoid fat mass ($M_1 \to M_2 \mid A$) & $0.002538$ & $[-0.01541$, $0.02049]$ & \mbox{$0.7817$} \\
 & $c'$: slope of gynoid fat mass $\to$ PWV, adjusted for LDL and systolic blood pressure ($A \to Y \mid M_1, M_2$) & $2.84\times10^{-5}$ & $[-2.536\times10^{-5}$, $8.217\times10^{-5}]$ & \mbox{$0.3005$} \\
 & $e$: slope of LDL cholesterol $\to$ PWV, adjusted for gynoid fat mass and systolic blood pressure ($M_1 \to Y \mid A, M_2$) & $0.002079$ & $[0.000158$, $0.004001]$ & \mbox{$0.03391$} \\
 & $f$: slope of seated systolic blood pressure $\to$ PWV, adjusted for gynoid fat mass and LDL ($M_2 \to Y \mid A, M_1$) & $0.05167$ & $[0.04649$, $0.05684]$ & \mbox{$3.191\times10^{-85}$} \\
 & $\tau$: total association slope of gynoid fat mass $\to$ PWV ($A \to Y$) & $0.0001464$ & $[8.821\times10^{-5}$, $0.0002046]$ & \mbox{$8.165\times10^{-7}$} \\
\bottomrule
\end{longtable}}

\paragraph{3.\ Exposure--outcome association: gynoid fat mass and thigh-to-ankle PWV}
\textit{Design.} Gynoid fat mass and PWV measured in the same period (gynoid fat mass within 365 days before or after PWV, bounds inclusive); for each participant the pair with the shortest interval. Adjustment for approximate age at the gynoid fat mass measurement (computed from calendar years) and sex (categorical term). PWV is the mean of the available sides.

\textit{Model.} $Y = \beta_0 + \beta_A \cdot A + \beta_{\mathrm{age}} \cdot \mathrm{age} + \gamma_{\mathrm{sex}} + \varepsilon$, linear regression; $A$ is gynoid fat mass and $Y$ thigh-to-ankle PWV, both in original units. $\beta_A$ is the change in PWV per 1-unit increase in gynoid fat mass, at fixed age and sex.

\textit{Intervals.} 95\% confidence intervals.
{\footnotesize\setlength{\tabcolsep}{3pt}
\begin{longtable}{@{}>{\raggedright\arraybackslash}p{0.14\linewidth}>{\raggedright\arraybackslash}p{0.29\linewidth}>{\raggedleft\arraybackslash}p{0.13\linewidth}>{\raggedright\arraybackslash}p{0.22\linewidth}>{\raggedleft\arraybackslash}p{0.14\linewidth}@{}}
\toprule
Block & Parameter & Estimate & Interval & $P$ \\
\midrule\endhead
Same-period pairing ($\pm$365 days) & $\beta_A$: adjusted gynoid fat mass--PWV slope & $0.0001872$ & $[0.0001598$, $0.0002145]$ & \mbox{$1.237\times10^{-40}$} \\
\bottomrule
\end{longtable}}

\subsubsection*{Adjacent segments and pairwise associations}

\paragraph{4.\ First segment of the systolic blood pressure chain and total association: gynoid fat mass, seated systolic blood pressure and PWV}
\textit{Design.} Gynoid fat mass, seated systolic blood pressure and PWV measured in strict date order: systolic blood pressure 30--730 days after the gynoid fat mass measurement, PWV 90--1095 days after systolic blood pressure (all bounds inclusive); one measurement chain per participant. Adjustment for approximate age at the gynoid fat mass measurement (computed from calendar years) and sex (categorical term). Systolic blood pressure is the mean of the available values of two seated readings, PWV the mean of the available sides; gynoid fat mass centered within the analysis sample.

\textit{Model.} $M = \alpha_0 + \alpha_A \cdot (A - \bar{A}) + \alpha_C^{\top} C + \varepsilon_M$; $Y = \gamma_0 + \gamma_A \cdot (A - \bar{A}) + \gamma_C^{\top} C + \varepsilon_T$. Both equations are linear regressions; $A$, $M$ and $Y$ are gynoid fat mass, seated systolic blood pressure and thigh-to-ankle PWV (original units), $\bar{A}$ is the sample mean, and $C$ is age and sex. $\alpha_A$ is the change in systolic blood pressure per 1-unit increase in gynoid fat mass; $\gamma_A$ is the adjusted total slope of gynoid fat mass on PWV.

\textit{Intervals.} Bootstrap 95\% percentile intervals.
{\footnotesize\setlength{\tabcolsep}{3pt}
\begin{longtable}{@{}>{\raggedright\arraybackslash}p{0.15\linewidth}>{\raggedright\arraybackslash}p{0.40\linewidth}>{\raggedleft\arraybackslash}p{0.14\linewidth}>{\raggedright\arraybackslash}p{0.24\linewidth}@{}}
\toprule
Block & Parameter & Estimate & Interval \\
\midrule\endhead
Path components & $\alpha_A$: slope of gynoid fat mass $\to$ seated systolic blood pressure ($A \to M$) & $-0.005525$ & $[-1.311$, $0.02516]$ \\
 & $\gamma_A$: adjusted total slope of gynoid fat mass $\to$ PWV ($A \to Y$) & $0.001685$ & $[-0.2857$, $0.005267]$ \\
\bottomrule
\end{longtable}}

\paragraph{5.\ Downstream serial fragment: LDL-C $\to$ seated systolic blood pressure $\to$ thigh-to-ankle PWV}
\textit{Design.} LDL-C, seated systolic blood pressure and PWV measured in strict date order: systolic blood pressure 90--730 days after LDL-C, PWV 90--1095 days after systolic blood pressure (all bounds inclusive); one measurement chain per participant. Adjustment for approximate age at the LDL-C measurement (computed from calendar years) and sex (categorical term). Systolic blood pressure is the mean of the available values of two seated readings, PWV the mean of the available sides.

\textit{Model.} $M = \alpha_0 + \alpha_A \cdot A + \alpha_C^{\top} C + \varepsilon_M$; $Y = \beta_0 + \beta_A \cdot A + \beta_M \cdot M + \beta_C^{\top} C + \varepsilon_Y$; $Y = \gamma_0 + \gamma_A \cdot A + \gamma_C^{\top} C + \varepsilon_T$. All three equations are linear regressions; $A$, $M$ and $Y$ are LDL-C, seated systolic blood pressure and thigh-to-ankle PWV (original units), and $C$ is age and sex. $\alpha_A$ is the change in systolic blood pressure per 1-unit increase in LDL-C; $\beta_M$ and $\beta_A$ are the mutually adjusted slopes of systolic blood pressure and LDL-C on PWV; $\gamma_A$ is the total slope of LDL-C on PWV with adjustment for $C$ only; fragment path product $\theta = \alpha_A \cdot \beta_M$.

\textit{Intervals.} Bootstrap 95\% percentile intervals.
{\footnotesize\setlength{\tabcolsep}{3pt}
\begin{longtable}{@{}>{\raggedright\arraybackslash}p{0.15\linewidth}>{\raggedright\arraybackslash}p{0.40\linewidth}>{\raggedleft\arraybackslash}p{0.14\linewidth}>{\raggedright\arraybackslash}p{0.24\linewidth}@{}}
\toprule
Block & Parameter & Estimate & Interval \\
\midrule\endhead
Path components & $\alpha_A$: slope of LDL-C $\to$ seated systolic blood pressure ($A \to M$) & $0.0165$ & $[-0.0102$, $0.04338]$ \\
 & $\beta_M$: slope of seated systolic blood pressure $\to$ PWV, adjusted for LDL-C ($M \to Y \mid A$) & $0.03409$ & $[0.02897$, $0.03929]$ \\
 & $\beta_A$: slope of LDL-C $\to$ PWV, adjusted for seated systolic blood pressure ($A \to Y \mid M$) & $2.462\times10^{-5}$ & $[-0.002287$, $0.002514]$ \\
 & $\gamma_A$: adjusted total slope of LDL-C $\to$ PWV ($A \to Y$) & $0.0005869$ & $[-0.001882$, $0.003282]$ \\
Path index & $\theta = \alpha_A \cdot \beta_M$: fragment path-product association index & $0.0005623$ & $[-0.0003356$, $0.001487]$ \\
\bottomrule
\end{longtable}}

\paragraph{6.\ Pairwise association: gynoid fat mass $\to$ LDL cholesterol ($\pm$180 days)}
\textit{Design.} Nearest LDL cholesterol measurement within 180 days before or after the gynoid fat measurement date (either order, bounds inclusive); one row per gynoid fat measurement date, $n = 3315$. Adjustment for approximate age (gynoid fat measurement year minus birth year) and recorded sex; variables not standardized.

\textit{Model.} $M_1 = \beta_0 + \beta \cdot A + \beta_{\mathrm{age}} \cdot \mathrm{age} + \gamma_{\mathrm{sex}} + \varepsilon$ (linear regression; $A$ is gynoid fat mass and $M_1$ LDL cholesterol, original units). $\beta$ is the difference in LDL cholesterol per 1 original unit higher gynoid fat mass, at fixed age and sex.

\textit{Intervals.} 95\% confidence intervals.
{\footnotesize\setlength{\tabcolsep}{3pt}
\begin{longtable}{@{}>{\raggedright\arraybackslash}p{0.14\linewidth}>{\raggedright\arraybackslash}p{0.29\linewidth}>{\raggedleft\arraybackslash}p{0.13\linewidth}>{\raggedright\arraybackslash}p{0.22\linewidth}>{\raggedleft\arraybackslash}p{0.14\linewidth}@{}}
\toprule
Block & Parameter & Estimate & Interval & $P$ \\
\midrule\endhead
$\pm$180-day window & $\beta$: adjusted slope of gynoid fat mass $\to$ LDL cholesterol & $0.001683$ & $[0.000745$, $0.002622]$ & \mbox{$0.0004384$} \\
\bottomrule
\end{longtable}}

\paragraph{7.\ Pairwise association: LDL cholesterol $\to$ seated systolic blood pressure ($\pm$90 days)}
\textit{Design.} Nearest seated systolic blood pressure measurement within 90 days before or after the LDL cholesterol test date (either order, bounds inclusive); one row per LDL test date, $n = 3945$. Adjustment for approximate age (LDL test year minus birth year) and recorded sex. Systolic blood pressure is the mean of the available values of two seated readings; variables not standardized.

\textit{Model.} $M_2 = \beta_0 + \beta \cdot M_1 + \beta_{\mathrm{age}} \cdot \mathrm{age} + \gamma_{\mathrm{sex}} + \varepsilon$ (linear regression; $M_1$ is LDL cholesterol and $M_2$ seated systolic blood pressure, original units). $\beta$ is the difference in seated systolic blood pressure per 1 original unit higher LDL cholesterol, at fixed age and sex.

\textit{Intervals.} 95\% confidence intervals.
{\footnotesize\setlength{\tabcolsep}{3pt}
\begin{longtable}{@{}>{\raggedright\arraybackslash}p{0.14\linewidth}>{\raggedright\arraybackslash}p{0.29\linewidth}>{\raggedleft\arraybackslash}p{0.13\linewidth}>{\raggedright\arraybackslash}p{0.22\linewidth}>{\raggedleft\arraybackslash}p{0.14\linewidth}@{}}
\toprule
Block & Parameter & Estimate & Interval & $P$ \\
\midrule\endhead
$\pm$90-day window & $\beta$: adjusted slope of LDL cholesterol $\to$ seated systolic blood pressure & $0.01847$ & $[0.005048$, $0.0319]$ & \mbox{$0.006998$} \\
\bottomrule
\end{longtable}}

\paragraph{8.\ Pairwise association: seated systolic blood pressure $\to$ PWV ($\pm$14 days)}
\textit{Design.} Nearest thigh-to-ankle PWV measurement within 14 days before or after the seated blood pressure measurement date (either order, bounds inclusive); one row per blood pressure measurement date, $n = 10933$. Adjustment for approximate age (blood pressure measurement year minus birth year) and recorded sex. Systolic blood pressure is the mean of the available values of two seated readings, PWV the mean of the available sides; variables not standardized.

\textit{Model.} $Y = \beta_0 + \beta \cdot M_2 + \beta_{\mathrm{age}} \cdot \mathrm{age} + \gamma_{\mathrm{sex}} + \varepsilon$ (linear regression; $M_2$ is seated systolic blood pressure and $Y$ thigh-to-ankle PWV, original units). $\beta$ is the difference in PWV per 1 original unit higher seated systolic blood pressure, at fixed age and sex.

\textit{Intervals.} 95\% confidence intervals.
{\footnotesize\setlength{\tabcolsep}{3pt}
\begin{longtable}{@{}>{\raggedright\arraybackslash}p{0.14\linewidth}>{\raggedright\arraybackslash}p{0.29\linewidth}>{\raggedleft\arraybackslash}p{0.13\linewidth}>{\raggedright\arraybackslash}p{0.22\linewidth}>{\raggedleft\arraybackslash}p{0.14\linewidth}@{}}
\toprule
Block & Parameter & Estimate & Interval & $P$ \\
\midrule\endhead
$\pm$14-day window & $\beta$: adjusted slope of seated systolic blood pressure $\to$ thigh-to-ankle PWV & $0.05133$ & $[0.04929$, $0.05337]$ & \mbox{$0$} \\
\bottomrule
\end{longtable}}

\paragraph{9.\ Pairwise total association: gynoid fat mass $\to$ PWV ($\pm$365 days)}
\textit{Design.} Nearest thigh-to-ankle PWV measurement within 365 days before or after the gynoid fat measurement date (either order, bounds inclusive); one row per gynoid fat measurement date, $n = 8470$. Adjustment for approximate age (gynoid fat measurement year minus birth year) and recorded sex. PWV is the mean of the available sides; variables not standardized.

\textit{Model.} $Y = \beta_0 + \beta \cdot A + \beta_{\mathrm{age}} \cdot \mathrm{age} + \gamma_{\mathrm{sex}} + \varepsilon$ (linear regression; $A$ is gynoid fat mass and $Y$ thigh-to-ankle PWV, original units). $\beta$ is the difference in PWV per 1 original unit higher gynoid fat mass, at fixed age and sex.

\textit{Intervals.} 95\% confidence intervals.
{\footnotesize\setlength{\tabcolsep}{3pt}
\begin{longtable}{@{}>{\raggedright\arraybackslash}p{0.14\linewidth}>{\raggedright\arraybackslash}p{0.29\linewidth}>{\raggedleft\arraybackslash}p{0.13\linewidth}>{\raggedright\arraybackslash}p{0.22\linewidth}>{\raggedleft\arraybackslash}p{0.14\linewidth}@{}}
\toprule
Block & Parameter & Estimate & Interval & $P$ \\
\midrule\endhead
$\pm$365-day window & $\beta$: adjusted total association slope of gynoid fat mass $\to$ thigh-to-ankle PWV & $0.000191$ & $[0.0001642$, $0.0002178]$ & \mbox{$2.246\times10^{-44}$} \\
\bottomrule
\end{longtable}}

\paragraph{10.\ Pairwise association: gynoid fat mass--LDL-C ($\pm$365 days)}
\textit{Design.} Gynoid fat mass and LDL-C measured in the same period (gynoid fat mass within 365 days before or after LDL-C, bounds inclusive); for each participant the pair with the shortest interval. Adjustment for approximate age at the gynoid fat mass measurement (computed from calendar years) and sex (categorical term).

\textit{Model.} $M_1 = \beta_0 + \beta_A \cdot A + \beta_{\mathrm{age}} \cdot \mathrm{age} + \gamma_{\mathrm{sex}} + \varepsilon$ (linear regression; $A$ is gynoid fat mass and $M_1$ LDL-C, both in original units). $\beta_A$ is the change in LDL-C per 1-unit increase in gynoid fat mass, at fixed age and sex.

\textit{Intervals.} 95\% confidence intervals.
{\footnotesize\setlength{\tabcolsep}{3pt}
\begin{longtable}{@{}>{\raggedright\arraybackslash}p{0.14\linewidth}>{\raggedright\arraybackslash}p{0.29\linewidth}>{\raggedleft\arraybackslash}p{0.13\linewidth}>{\raggedright\arraybackslash}p{0.22\linewidth}>{\raggedleft\arraybackslash}p{0.14\linewidth}@{}}
\toprule
Block & Parameter & Estimate & Interval & $P$ \\
\midrule\endhead
Same-period pairing ($\pm$365 days) & $\beta_A$: adjusted gynoid fat mass--LDL-C slope & $0.00176$ & $[0.001076$, $0.002444]$ & \mbox{$4.751\times10^{-7}$} \\
\bottomrule
\end{longtable}}

\paragraph{11.\ Pairwise association: LDL-C--seated systolic blood pressure ($\pm$365 days)}
\textit{Design.} LDL-C and seated systolic blood pressure measured in the same period (LDL-C within 365 days before or after systolic blood pressure, bounds inclusive); for each participant the pair with the shortest interval. Adjustment for approximate age at the LDL-C measurement (computed from calendar years) and sex (categorical term). Systolic blood pressure is the mean of the available values of two seated readings.

\textit{Model.} $M_2 = \beta_0 + \beta_{M_1} \cdot M_1 + \beta_{\mathrm{age}} \cdot \mathrm{age} + \gamma_{\mathrm{sex}} + \varepsilon$ (linear regression; $M_1$ is LDL-C and $M_2$ seated systolic blood pressure, both in original units). $\beta_{M_1}$ is the change in systolic blood pressure per 1-unit increase in LDL-C, at fixed age and sex.

\textit{Intervals.} 95\% confidence intervals.
{\footnotesize\setlength{\tabcolsep}{3pt}
\begin{longtable}{@{}>{\raggedright\arraybackslash}p{0.14\linewidth}>{\raggedright\arraybackslash}p{0.29\linewidth}>{\raggedleft\arraybackslash}p{0.13\linewidth}>{\raggedright\arraybackslash}p{0.22\linewidth}>{\raggedleft\arraybackslash}p{0.14\linewidth}@{}}
\toprule
Block & Parameter & Estimate & Interval & $P$ \\
\midrule\endhead
Same-period pairing ($\pm$365 days) & $\beta_{M_1}$: adjusted LDL-C--seated systolic blood pressure slope & $0.02593$ & $[0.0146$, $0.03725]$ & \mbox{$7.364\times10^{-6}$} \\
\bottomrule
\end{longtable}}

\paragraph{12.\ Pairwise association: gynoid fat mass--seated systolic blood pressure ($\pm$365 days)}
\textit{Design.} Gynoid fat mass and seated systolic blood pressure measured in the same period (gynoid fat mass within 365 days before or after systolic blood pressure, bounds inclusive); for each participant the pair with the shortest interval. Adjustment for approximate age at the gynoid fat mass measurement (computed from calendar years) and sex (categorical term). Systolic blood pressure is the mean of the available values of two seated readings.

\textit{Model.} $M_2 = \beta_0 + \beta_A \cdot A + \beta_{\mathrm{age}} \cdot \mathrm{age} + \gamma_{\mathrm{sex}} + \varepsilon$ (linear regression; $A$ is gynoid fat mass and $M_2$ seated systolic blood pressure, both in original units). $\beta_A$ is the change in systolic blood pressure per 1-unit increase in gynoid fat mass, at fixed age and sex.

\textit{Intervals.} 95\% confidence intervals.
{\footnotesize\setlength{\tabcolsep}{3pt}
\begin{longtable}{@{}>{\raggedright\arraybackslash}p{0.14\linewidth}>{\raggedright\arraybackslash}p{0.29\linewidth}>{\raggedleft\arraybackslash}p{0.13\linewidth}>{\raggedright\arraybackslash}p{0.22\linewidth}>{\raggedleft\arraybackslash}p{0.14\linewidth}@{}}
\toprule
Block & Parameter & Estimate & Interval & $P$ \\
\midrule\endhead
Same-period pairing ($\pm$365 days) & $\beta_A$: adjusted gynoid fat mass--seated systolic blood pressure slope & $0.002369$ & $[0.002132$, $0.002605]$ & \mbox{$8.464\times10^{-84}$} \\
\bottomrule
\end{longtable}}

\subsubsection*{Joint models}

\paragraph{13.\ $\pm$365-day joint model: gynoid fat, LDL-C, systolic blood pressure and PWV}
\textit{Design.} Gynoid fat mass, LDL-C, seated systolic blood pressure and PWV measured in the same period: date span of the four measurements $\le$ 365 days (bounds inclusive, any order); for each participant the set with the shortest span. Both models share the same sample, each adjusted for approximate age at the gynoid fat mass measurement (computed from calendar years) and sex (categorical term). Systolic blood pressure is the mean of the available values of two seated readings, PWV the mean of the available sides.

\textit{Model.} Two linear regressions, both with PWV ($Y$) as the outcome. Joint model: $Y = \beta_0 + \beta_A \cdot A + \beta_{\mathrm{LDL}} \cdot M_1 + \beta_{\mathrm{SBP}} \cdot M_2 + \beta_{\mathrm{age}} \cdot \mathrm{age} + \gamma_{\mathrm{sex}} + \varepsilon$; reduced model: $Y = \alpha_0 + \alpha_A \cdot A + \alpha_{\mathrm{age}} \cdot \mathrm{age} + \gamma_{\mathrm{sex}} + \varepsilon$. $A$, $M_1$ and $M_2$ are gynoid fat mass, LDL-C and seated systolic blood pressure, all in original units. $\beta_A$, $\beta_{\mathrm{LDL}}$ and $\beta_{\mathrm{SBP}}$ are the change in PWV per 1-unit increase in the respective variable with the other joint-model terms fixed; $\alpha_A$ is the gynoid fat mass slope in the same sample with adjustment for age and sex only.

\textit{Intervals.} 95\% confidence intervals.
{\footnotesize\setlength{\tabcolsep}{3pt}
\begin{longtable}{@{}>{\raggedright\arraybackslash}p{0.14\linewidth}>{\raggedright\arraybackslash}p{0.29\linewidth}>{\raggedleft\arraybackslash}p{0.13\linewidth}>{\raggedright\arraybackslash}p{0.22\linewidth}>{\raggedleft\arraybackslash}p{0.14\linewidth}@{}}
\toprule
Block & Parameter & Estimate & Interval & $P$ \\
\midrule\endhead
$\pm$365 days; joint model: gynoid fat mass + LDL-C + systolic blood pressure + age + sex & $\beta_A$: gynoid fat mass coefficient (joint model) & $6.276\times10^{-5}$ & $[2.662\times10^{-5}$, $9.889\times10^{-5}]$ & \mbox{$0.0006678$} \\
 & $\beta_{\mathrm{LDL}}$: LDL-C coefficient (joint model) & $0.001003$ & $[-0.0004678$, $0.002474]$ &  \\
 & $\beta_{\mathrm{SBP}}$: seated systolic blood pressure coefficient (joint model) & $0.05195$ & $[0.04844$, $0.05545]$ &  \\
$\pm$365 days; reduced model: gynoid fat mass + age + sex & $\alpha_A$: gynoid fat mass coefficient (reduced model, same sample) & $0.000187$ & $[0.0001481$, $0.0002259]$ &  \\
\bottomrule
\end{longtable}}

\paragraph{14.\ $\pm$180-day joint model: gynoid fat, LDL-C, systolic blood pressure and PWV}
\textit{Design.} Gynoid fat mass, LDL-C, seated systolic blood pressure and PWV measured in the same period: date span of the four measurements $\le$ 180 days (bounds inclusive, any order); for each participant the set with the shortest span. Both models share the same sample; covariates and variable handling as in the $\pm$365-day joint model.

\textit{Model.} Two linear regressions, as in the $\pm$365-day joint model. Joint model: $Y = \beta_0 + \beta_A \cdot A + \beta_{\mathrm{LDL}} \cdot M_1 + \beta_{\mathrm{SBP}} \cdot M_2 + \beta_{\mathrm{age}} \cdot \mathrm{age} + \gamma_{\mathrm{sex}} + \varepsilon$; reduced model: $Y = \alpha_0 + \alpha_A \cdot A + \alpha_{\mathrm{age}} \cdot \mathrm{age} + \gamma_{\mathrm{sex}} + \varepsilon$. $\beta_A$, $\beta_{\mathrm{LDL}}$ and $\beta_{\mathrm{SBP}}$ are the change in PWV per 1-unit increase in the respective variable with the other joint-model terms fixed; $\alpha_A$ is the gynoid fat mass slope in the same sample with adjustment for age and sex only.

\textit{Intervals.} 95\% confidence intervals.
{\footnotesize\setlength{\tabcolsep}{3pt}
\begin{longtable}{@{}>{\raggedright\arraybackslash}p{0.14\linewidth}>{\raggedright\arraybackslash}p{0.29\linewidth}>{\raggedleft\arraybackslash}p{0.13\linewidth}>{\raggedright\arraybackslash}p{0.22\linewidth}>{\raggedleft\arraybackslash}p{0.14\linewidth}@{}}
\toprule
Block & Parameter & Estimate & Interval & $P$ \\
\midrule\endhead
$\pm$180 days; joint model: gynoid fat mass + LDL-C + systolic blood pressure + age + sex & $\beta_A$: gynoid fat mass coefficient (joint model) & $5.593\times10^{-5}$ & $[1.3\times10^{-5}$, $9.886\times10^{-5}]$ & \mbox{$0.01068$} \\
 & $\beta_{\mathrm{LDL}}$: LDL-C coefficient (joint model) & $0.001401$ & $[-0.0003548$, $0.003156]$ &  \\
 & $\beta_{\mathrm{SBP}}$: seated systolic blood pressure coefficient (joint model) & $0.05253$ & $[0.04835$, $0.05671]$ &  \\
$\pm$180 days; reduced model: gynoid fat mass + age + sex & $\alpha_A$: gynoid fat mass coefficient (reduced model, same sample) & $0.0001817$ & $[0.000136$, $0.0002274]$ &  \\
\bottomrule
\end{longtable}}

\subsubsection*{Reverse direction}

\paragraph{15.\ Reverse path: PWV $\to$ LDL-C $\to$ gynoid fat mass}
\textit{Design.} Endpoint roles reversed; PWV, LDL-C and gynoid fat mass measured in strict date order: LDL-C 90--1095 days after PWV, gynoid fat mass 30--730 days after LDL-C (all bounds inclusive); one measurement chain per participant, $n = 419$. Adjustment for approximate age at the PWV measurement (computed from calendar years) and sex (categorical term). PWV is the mean of the available sides.

\textit{Model.} $A$ = PWV, $M$ = LDL-C, $Y$ = gynoid fat mass, $C$ = age, sex; linear regressions (original units): $M = \alpha_0 + \alpha_A \cdot A + \alpha_C^{\top} C + \varepsilon_M$; $Y = \beta_0 + \beta_A \cdot A + \beta_M \cdot M + \beta_C^{\top} C + \varepsilon_Y$; $Y = \gamma_0 + \gamma_A \cdot A + \gamma_C^{\top} C + \varepsilon_T$. Reported quantities: $\alpha_A$ (PWV $\to$ LDL-C), $\beta_M$ (LDL-C $\to$ gynoid fat mass, adjusted for PWV), $\beta_A$ (PWV $\to$ gynoid fat mass, adjusted for LDL-C), $\gamma_A$ (total slope of PWV $\to$ gynoid fat mass), and the reverse path product $\theta = \alpha_A \cdot \beta_M$.

\textit{Intervals.} Bootstrap 95\% percentile intervals.
{\footnotesize\setlength{\tabcolsep}{3pt}
\begin{longtable}{@{}>{\raggedright\arraybackslash}p{0.15\linewidth}>{\raggedright\arraybackslash}p{0.40\linewidth}>{\raggedleft\arraybackslash}p{0.14\linewidth}>{\raggedright\arraybackslash}p{0.24\linewidth}@{}}
\toprule
Block & Parameter & Estimate & Interval \\
\midrule\endhead
Reverse path components & $\alpha_A$: slope of PWV $\to$ LDL-C (reverse $A \to M$) & $1.399$ & $[-0.6785$, $3.682]$ \\
 & $\beta_M$: slope of LDL-C $\to$ gynoid fat mass, adjusted for PWV (reverse $M \to Y \mid A$) & $1.343$ & $[-2.669$, $5.388]$ \\
 & $\beta_A$: slope of PWV $\to$ gynoid fat mass, adjusted for LDL-C (reverse $A \to Y \mid M$) & $55.3$ & $[-17.53$, $132.9]$ \\
 & $\gamma_A$: adjusted total slope of PWV $\to$ gynoid fat mass (reverse $A \to Y$) & $57.18$ & $[-16.68$, $137.1]$ \\
Reverse path index & $\theta = \alpha_A \cdot \beta_M$: reverse path-product association index & $1.88$ & $[-4.99$, $10.94]$ \\
\bottomrule
\end{longtable}}

\paragraph{16.\ Reverse path: PWV $\to$ seated systolic blood pressure $\to$ gynoid fat mass}
\textit{Design.} Endpoint roles reversed; PWV, seated systolic blood pressure and gynoid fat mass measured in strict date order: systolic blood pressure 90--1095 days after PWV, gynoid fat mass 30--730 days after systolic blood pressure (all bounds inclusive); one measurement chain per participant. Adjustment for approximate age at the PWV measurement (computed from calendar years) and sex (categorical term). PWV is the mean of the available sides, systolic blood pressure the mean of the available values of two seated readings.

\textit{Model.} $A$ = PWV, $M$ = seated systolic blood pressure, $Y$ = gynoid fat mass, $C$ = age, sex; linear regressions (original units): $M = \alpha_0 + \alpha_A \cdot A + \alpha_C^{\top} C + \varepsilon_M$; $Y = \beta_0 + \beta_A \cdot A + \beta_M \cdot M + \beta_C^{\top} C + \varepsilon_Y$; $Y = \gamma_0 + \gamma_A \cdot A + \gamma_C^{\top} C + \varepsilon_T$. Reported quantities: $\alpha_A$ (PWV $\to$ systolic blood pressure), $\beta_M$ (systolic blood pressure $\to$ gynoid fat mass, adjusted for PWV), $\beta_A$ (PWV $\to$ gynoid fat mass, adjusted for systolic blood pressure), $\gamma_A$ (total slope of PWV $\to$ gynoid fat mass), and the reverse path product $\theta = \alpha_A \cdot \beta_M$.

\textit{Intervals.} Bootstrap 95\% percentile intervals.
{\footnotesize\setlength{\tabcolsep}{3pt}
\begin{longtable}{@{}>{\raggedright\arraybackslash}p{0.15\linewidth}>{\raggedright\arraybackslash}p{0.40\linewidth}>{\raggedleft\arraybackslash}p{0.14\linewidth}>{\raggedright\arraybackslash}p{0.24\linewidth}@{}}
\toprule
Block & Parameter & Estimate & Interval \\
\midrule\endhead
Reverse path components & $\alpha_A$: slope of PWV $\to$ seated systolic blood pressure (reverse $A \to M$) & $2.74$ & $[-12.73$, $17.38]$ \\
 & $\beta_M$: slope of seated systolic blood pressure $\to$ gynoid fat mass, adjusted for PWV (reverse $M \to Y \mid A$) & $-0.6826$ & $[-3720$, $3768]$ \\
 & $\beta_A$: slope of PWV $\to$ gynoid fat mass, adjusted for seated systolic blood pressure (reverse $A \to Y \mid M$) & $296.3$ & $[-2.915\times10^{4}$, $2.327\times10^{4}]$ \\
 & $\gamma_A$: adjusted total slope of PWV $\to$ gynoid fat mass (reverse $A \to Y$) & $294.4$ & $[-597.8$, $769.1]$ \\
Reverse path index & $\theta = \alpha_A \cdot \beta_M$: reverse path-product association index & $-1.87$ & $[-2.338\times10^{4}$, $2.952\times10^{4}]$ \\
\bottomrule
\end{longtable}}

\paragraph{17.\ Reverse pairing: earlier PWV and gynoid fat mass measured 1--365 days later}
\textit{Design.} Each thigh-to-ankle PWV measurement date as starting point; nearest gynoid fat mass measurement 1--365 days afterwards (bounds inclusive); one row per PWV measurement date, $n = 46$. Adjustment for approximate age (PWV measurement year minus birth year) and recorded sex; variables not standardized.

\textit{Model.} $A = \beta_0 + \beta \cdot Y + \beta_{\mathrm{age}} \cdot \mathrm{age} + \gamma_{\mathrm{sex}} + \varepsilon$ (linear regression; $Y$ is the earlier thigh-to-ankle PWV and $A$ the subsequently measured gynoid fat mass, original units). $\beta$ is the difference in later gynoid fat mass per 1 original unit higher PWV, at fixed age and sex.

\textit{Intervals.} 95\% confidence intervals.
{\footnotesize\setlength{\tabcolsep}{3pt}
\begin{longtable}{@{}>{\raggedright\arraybackslash}p{0.14\linewidth}>{\raggedright\arraybackslash}p{0.29\linewidth}>{\raggedleft\arraybackslash}p{0.13\linewidth}>{\raggedright\arraybackslash}p{0.22\linewidth}>{\raggedleft\arraybackslash}p{0.14\linewidth}@{}}
\toprule
Block & Parameter & Estimate & Interval & $P$ \\
\midrule\endhead
1--365 days after PWV & $\beta$: adjusted slope of earlier PWV $\to$ subsequently measured gynoid fat mass & $-207.4$ & $[-567.7$, $153]$ & \mbox{$0.2593$} \\
\bottomrule
\end{longtable}}

\paragraph{18.\ Reverse pairing: earlier PWV and gynoid fat mass measured 366--730 days later}
\textit{Design.} PWV and gynoid fat mass measured in date order: gynoid fat mass 366--730 days after PWV (bounds inclusive); for each participant the earliest eligible gynoid fat mass measurement, paired with the nearest PWV within that interval. Adjustment for approximate age at the gynoid fat mass measurement (computed from calendar years) and sex (categorical term). PWV is the mean of the available sides.

\textit{Model.} $A = \beta_0 + \beta_Y \cdot Y + \beta_{\mathrm{age}} \cdot \mathrm{age} + \gamma_{\mathrm{sex}} + \varepsilon$ (linear regression; $Y$ is the earlier thigh-to-ankle PWV and $A$ the subsequently measured gynoid fat mass, both in original units). $\beta_Y$ is the change in later gynoid fat mass per 1-unit increase in earlier PWV, at fixed age and sex.

\textit{Intervals.} 95\% confidence intervals.
{\footnotesize\setlength{\tabcolsep}{3pt}
\begin{longtable}{@{}>{\raggedright\arraybackslash}p{0.14\linewidth}>{\raggedright\arraybackslash}p{0.29\linewidth}>{\raggedleft\arraybackslash}p{0.13\linewidth}>{\raggedright\arraybackslash}p{0.22\linewidth}>{\raggedleft\arraybackslash}p{0.14\linewidth}@{}}
\toprule
Block & Parameter & Estimate & Interval & $P$ \\
\midrule\endhead
Reverse pairing (366--730 days) & $\beta_Y$: adjusted earlier PWV--later gynoid fat mass slope & $74.36$ & $[5.462$, $143.3]$ & \mbox{$0.03447$} \\
\bottomrule
\end{longtable}}

\subsubsection*{Heterogeneity and interaction}

\paragraph{19.\ Conditional path to PWV under the gynoid fat mass $\times$ LDL-C interaction}
\textit{Design.} Gynoid fat mass, LDL-C and PWV measured in strict date order: LDL-C 30--730 days after the gynoid fat mass measurement, PWV 90--1095 days after LDL-C (all bounds inclusive); one measurement chain per participant, $n = 138$. Adjustment for approximate age at the gynoid fat mass measurement (computed from calendar years) and sex (categorical term). PWV is the mean of the available sides; gynoid fat mass and LDL-C centered within the analysis sample.

\textit{Model.} $M = \alpha_0 + \alpha_A \cdot a + \alpha_C^{\top} C + \varepsilon_M$; $Y = \beta_0 + \beta_A \cdot a + \beta_M \cdot m + \beta_{AM} \cdot a \cdot m + \beta_C^{\top} C + \varepsilon_Y$; $Y = \gamma_0 + \gamma_A \cdot a + \gamma_C^{\top} C + \varepsilon_T$ (linear regressions; $a = A - \bar{A}$ and $m = M - \bar{M}$ are centered gynoid fat mass and LDL-C, $M$ and $Y$ are LDL-C and PWV as raw values, and $C$ is age and sex). Reported quantities: $\alpha_A$; slope $\beta_M$ of LDL-C $\to$ PWV at mean gynoid fat mass; slope $\beta_A$ of gynoid fat mass $\to$ PWV at mean LDL-C; total slope $\gamma_A$; interaction coefficient $\beta_{AM}$; path product at the exposure mean $\theta(0) = \alpha_A \cdot \beta_M$.

\textit{Intervals.} Bootstrap 95\% percentile intervals.
{\footnotesize\setlength{\tabcolsep}{3pt}
\begin{longtable}{@{}>{\raggedright\arraybackslash}p{0.15\linewidth}>{\raggedright\arraybackslash}p{0.40\linewidth}>{\raggedleft\arraybackslash}p{0.14\linewidth}>{\raggedright\arraybackslash}p{0.24\linewidth}@{}}
\toprule
Block & Parameter & Estimate & Interval \\
\midrule\endhead
Path components of the model with interaction term & $\alpha_A$: slope of gynoid fat mass $\to$ LDL-C ($A \to M$) & $-0.00476$ & $[-0.0177$, $0.003754]$ \\
 & $\beta_M$: conditional slope of LDL-C $\to$ PWV at the exposure mean ($a = 0$) & $-0.00321$ & $[-0.009316$, $0.001293]$ \\
 & $\beta_A$: conditional slope of gynoid fat mass $\to$ PWV at the mediator mean ($m = 0$) & $0.0001173$ & $[2.139\times10^{-6}$, $0.0002693]$ \\
 & $\gamma_A$: adjusted total slope of gynoid fat mass $\to$ PWV ($A \to Y$) & $0.000119$ & $[-1.926\times10^{-6}$, $0.0002515]$ \\
Exposure $\times$ mediator interaction & $\beta_{AM}$: gynoid fat mass $\times$ LDL-C interaction coefficient in the outcome equation & $-1.679\times10^{-6}$ & $[-3.114\times10^{-6}$, $4.237\times10^{-6}]$ \\
Path index at the exposure mean & $\theta(0) = \alpha_A \cdot \beta_M$: conditional path-product index at the exposure mean & $1.528\times10^{-5}$ & $[-2.564\times10^{-5}$, $7.985\times10^{-5}]$ \\
\bottomrule
\end{longtable}}

\paragraph{20.\ Sex heterogeneity: gynoid fat mass $\to$ LDL-C $\to$ PWV}
\textit{Design.} Gynoid fat mass, LDL-C and PWV measured in strict date order: LDL-C 30--730 days after the gynoid fat mass measurement, PWV 90--1095 days after LDL-C (all bounds inclusive); one measurement chain per participant. All models fitted separately within the first and second sex categories, each stratum adjusted for approximate age at the gynoid fat mass measurement (computed from calendar years). PWV is the mean of the available sides.

\textit{Model.} Within category $s$: $M = \alpha_{0s} + \alpha_{A,s} \cdot A + \alpha_{\mathrm{age},s} \cdot \mathrm{age} + \varepsilon_M$; $Y = \beta_{0s} + \beta_{A,s} \cdot A + \beta_{M,s} \cdot M + \beta_{\mathrm{age},s} \cdot \mathrm{age} + \varepsilon_Y$; $Y = \gamma_{0s} + \gamma_{A,s} \cdot A + \gamma_{\mathrm{age},s} \cdot \mathrm{age} + \varepsilon_T$ (linear regressions; $A$ gynoid fat mass, $M$ LDL-C, $Y$ PWV, original units). Reported quantities: per-category $\alpha_{A,s}$, $\beta_{M,s}$, $\beta_{A,s}$, $\gamma_{A,s}$ and path product $\theta_s = \alpha_{A,s} \cdot \beta_{M,s}$, and for each quantity the difference $\Delta$ of the first minus the second category.

\textit{Intervals.} Bootstrap 95\% percentile intervals.
{\footnotesize\setlength{\tabcolsep}{3pt}
\begin{longtable}{@{}>{\raggedright\arraybackslash}p{0.15\linewidth}>{\raggedright\arraybackslash}p{0.40\linewidth}>{\raggedleft\arraybackslash}p{0.14\linewidth}>{\raggedright\arraybackslash}p{0.24\linewidth}@{}}
\toprule
Block & Parameter & Estimate & Interval \\
\midrule\endhead
First sex category & $\alpha_{A,1}$: slope of gynoid fat mass $\to$ LDL-C within the first category & $0.00142$ & $[-0.003394$, $0.006793]$ \\
 & $\beta_{M,1}$: slope of LDL-C $\to$ PWV adjusted for gynoid fat mass, within the first category & $-0.005076$ & $[-0.01053$, $0.00189]$ \\
 & $\beta_{A,1}$: slope of gynoid fat mass $\to$ PWV adjusted for LDL-C, within the first category & $8.431\times10^{-5}$ & $[-5.177\times10^{-5}$, $0.0002455]$ \\
 & $\gamma_{A,1}$: adjusted total slope of gynoid fat mass $\to$ PWV within the first category & $7.71\times10^{-5}$ & $[-5.978\times10^{-5}$, $0.0002394]$ \\
 & $\theta_1 = \alpha_{A,1} \cdot \beta_{M,1}$: path-product association index, first category & $-7.207\times10^{-6}$ & $[-4.691\times10^{-5}$, $2.042\times10^{-5}]$ \\
Second sex category & $\alpha_{A,2}$: slope of gynoid fat mass $\to$ LDL-C within the second category & $-0.01657$ & $[-0.05232$, $0.005407]$ \\
 & $\beta_{M,2}$: slope of LDL-C $\to$ PWV adjusted for gynoid fat mass, within the second category & $0.001762$ & $[-0.01292$, $0.004243]$ \\
 & $\beta_{A,2}$: slope of gynoid fat mass $\to$ PWV adjusted for LDL-C, within the second category & $0.0002059$ & $[-4.947\times10^{-5}$, $0.000482]$ \\
 & $\gamma_{A,2}$: adjusted total slope of gynoid fat mass $\to$ PWV within the second category & $0.0001767$ & $[-8.36\times10^{-5}$, $0.0004429]$ \\
 & $\theta_2 = \alpha_{A,2} \cdot \beta_{M,2}$: path-product association index, second category & $-2.919\times10^{-5}$ & $[-0.0001765$, $4.192\times10^{-5}]$ \\
Sex difference (first $-$ second category) & $\Delta\alpha_A = \alpha_{A,1} - \alpha_{A,2}$: difference in gynoid fat mass $\to$ LDL-C slopes & $0.01799$ & $[-0.005462$, $0.05257]$ \\
 & $\Delta\beta_M = \beta_{M,1} - \beta_{M,2}$: difference in LDL-C $\to$ PWV conditional slopes & $-0.006837$ & $[-0.01275$, $0.009719]$ \\
 & $\Delta\beta_A = \beta_{A,1} - \beta_{A,2}$: difference in gynoid fat mass $\to$ PWV conditional slopes & $-0.0001216$ & $[-0.0004264$, $0.0001903]$ \\
 & $\Delta\gamma_A = \gamma_{A,1} - \gamma_{A,2}$: difference in gynoid fat mass $\to$ PWV total slopes & $-9.965\times10^{-5}$ & $[-0.0004027$, $0.0002111]$ \\
 & $\Delta\theta = \theta_1 - \theta_2$: difference in path-product association indices & $2.198\times10^{-5}$ & $[-5.713\times10^{-5}$, $0.0001699]$ \\
\bottomrule
\end{longtable}}

\paragraph{21.\ Sex stratification: gynoid fat mass $\to$ seated systolic blood pressure $\to$ PWV}
\textit{Design.} Gynoid fat mass, seated systolic blood pressure and PWV measured in strict date order: systolic blood pressure 30--730 days after the gynoid fat mass measurement, PWV 90--1095 days after systolic blood pressure (all bounds inclusive); one measurement chain per participant. Fitted separately by sex category, each stratum adjusted for approximate age at the gynoid fat mass measurement (computed from calendar years). Systolic blood pressure is the mean of the available values of two seated readings, PWV the mean of the available sides.

\textit{Model.} Within category $s$: $M = \alpha_{0s} + \alpha_{A,s} \cdot A + \alpha_{\mathrm{age},s} \cdot \mathrm{age} + \varepsilon_M$; $Y = \gamma_{0s} + \gamma_{A,s} \cdot A + \gamma_{\mathrm{age},s} \cdot \mathrm{age} + \varepsilon_T$ (linear regressions; $A$ gynoid fat mass, $M$ seated systolic blood pressure, $Y$ PWV, original units). $\alpha_{A,s}$ is the change in systolic blood pressure per 1-unit increase in gynoid fat mass within category $s$; $\gamma_{A,s}$ is the adjusted total slope of gynoid fat mass on PWV within category $s$.

\textit{Intervals.} Bootstrap 95\% percentile intervals.
{\footnotesize\setlength{\tabcolsep}{3pt}
\begin{longtable}{@{}>{\raggedright\arraybackslash}p{0.15\linewidth}>{\raggedright\arraybackslash}p{0.40\linewidth}>{\raggedleft\arraybackslash}p{0.14\linewidth}>{\raggedright\arraybackslash}p{0.24\linewidth}@{}}
\toprule
Block & Parameter & Estimate & Interval \\
\midrule\endhead
First sex category & $\alpha_{A,1}$: slope of gynoid fat mass $\to$ seated systolic blood pressure within the first category & $-0.005525$ & $[-1.311$, $0.02516]$ \\
 & $\gamma_{A,1}$: adjusted total slope of gynoid fat mass $\to$ PWV within the first category & $0.001685$ & $[-0.2857$, $0.005267]$ \\
\bottomrule
\end{longtable}}

\paragraph{22.\ Age modification: gynoid fat mass $\to$ LDL $\to$ systolic blood pressure $\to$ PWV serial path}
\textit{Design.} Windows as in the primary path: LDL cholesterol within 180 days before or after the gynoid fat measurement date, seated systolic blood pressure within 90 days before or after that LDL, PWV within 14 days before or after that systolic blood pressure, nearest measurement each (either order, bounds inclusive); one row per gynoid fat measurement date. Approximate age (gynoid fat measurement year minus birth year), centered at 50 years with 10 years as one unit, entered as a main effect and in interaction with each upstream variable; additional adjustment for recorded sex. Variables not standardized.

\textit{Model.} $M_1 = i_1 + a \cdot A + \kappa_1 G + a_G \cdot A \cdot G + s_1 + \varepsilon_1$; $M_2 = i_2 + d \cdot A + b \cdot M_1 + \kappa_2 G + d_G \cdot A \cdot G + b_G \cdot M_1 \cdot G + s_2 + \varepsilon_2$; $Y = i_3 + c' \cdot A + e \cdot M_1 + f \cdot M_2 + \kappa_3 G + c'_G \cdot A \cdot G + e_G \cdot M_1 \cdot G + f_G \cdot M_2 \cdot G + s_3 + \varepsilon_3$; $Y = i_T + \tau \cdot A + \kappa_T \cdot G + \tau_G \cdot A \cdot G + s_T + \varepsilon_T$. All four equations are linear regressions; $A$, $M_1$, $M_2$ and $Y$ are gynoid fat mass, LDL cholesterol, seated systolic blood pressure and thigh-to-ankle PWV (original units), $G = (\mathrm{age} - 50)/10$, and $s$ is the sex main effect. $a$, $d$, $b$, $c'$, $e$, $f$ and $\tau$ are slopes at age 50, and the coefficients with subscript $G$ are the change in the corresponding slope per 10 years higher age; serial path product at age $g$: $(a + a_G \cdot G)(b + b_G \cdot G)(f + f_G \cdot G)$ with $G = (g - 50)/10$, for $g$ = 40, 50 and 60 years.

\textit{Intervals.} 95\% confidence intervals. $P$ values not corrected for multiple comparisons.
{\footnotesize\setlength{\tabcolsep}{3pt}
\begin{longtable}{@{}>{\raggedright\arraybackslash}p{0.14\linewidth}>{\raggedright\arraybackslash}p{0.29\linewidth}>{\raggedleft\arraybackslash}p{0.13\linewidth}>{\raggedright\arraybackslash}p{0.22\linewidth}>{\raggedleft\arraybackslash}p{0.14\linewidth}@{}}
\toprule
Block & Parameter & Estimate & Interval & $P$ \\
\midrule\endhead
Slopes at age 50 & $a$: slope of gynoid fat mass $\to$ LDL cholesterol ($A \to M_1$) & $0.001675$ & $[0.0004082$, $0.002942]$ & \mbox{$0.009555$} \\
 & $d$: slope of gynoid fat mass $\to$ seated systolic blood pressure, adjusted for LDL ($A \to M_2 \mid M_1$) & $0.002135$ & $[0.001636$, $0.002634]$ & \mbox{$5.023\times10^{-17}$} \\
 & $b$: slope of LDL cholesterol $\to$ seated systolic blood pressure, adjusted for gynoid fat mass ($M_1 \to M_2 \mid A$) & $0.007799$ & $[-0.009909$, $0.02551]$ & \mbox{$0.388$} \\
 & $c'$: slope of gynoid fat mass $\to$ PWV, adjusted for LDL and systolic blood pressure ($A \to Y \mid M_1, M_2$) & $2.849\times10^{-5}$ & $[-2.481\times10^{-5}$, $8.178\times10^{-5}]$ & \mbox{$0.2948$} \\
 & $e$: slope of LDL cholesterol $\to$ PWV, adjusted for gynoid fat mass and systolic blood pressure ($M_1 \to Y \mid A, M_2$) & $0.001237$ & $[-0.000883$, $0.003357]$ & \mbox{$0.2528$} \\
 & $f$: slope of seated systolic blood pressure $\to$ PWV, adjusted for gynoid fat mass and LDL ($M_2 \to Y \mid A, M_1$) & $0.05463$ & $[0.04925$, $0.06001]$ & \mbox{$4.72\times10^{-88}$} \\
 & $\tau$: total association slope of gynoid fat mass $\to$ PWV ($A \to Y$) & $0.0001473$ & $[8.877\times10^{-5}$, $0.0002059]$ & \mbox{$8.198\times10^{-7}$} \\
Age interactions (per 10 years) & $a_G$: change in $a$ with age (gynoid fat mass $\times$ age) & $0.00104$ & $[-6.872\times10^{-5}$, $0.002148]$ & \mbox{$0.066$} \\
 & $d_G$: change in $d$ with age (gynoid fat mass $\times$ age) & $0.0002952$ & $[-0.0003021$, $0.0008924]$ & \mbox{$0.3328$} \\
 & $b_G$: change in $b$ with age (LDL cholesterol $\times$ age) & $-0.01768$ & $[-0.04249$, $0.007119]$ & \mbox{$0.1623$} \\
 & $c'_G$: change in $c'$ with age (gynoid fat mass $\times$ age) & $-2.126\times10^{-5}$ & $[-8.21\times10^{-5}$, $3.959\times10^{-5}]$ & \mbox{$0.4935$} \\
 & $e_G$: change in $e$ with age (LDL cholesterol $\times$ age) & $0.001652$ & $[-0.001035$, $0.004339]$ & \mbox{$0.2283$} \\
 & $f_G$: change in $f$ with age (seated systolic blood pressure $\times$ age) & $-0.00776$ & $[-0.01332$, $-0.002199]$ & \mbox{$0.006238$} \\
 & $\tau_G$: change in $\tau$ with age (gynoid fat mass $\times$ age) & $-4.819\times10^{-6}$ & $[-7.293\times10^{-5}$, $6.33\times10^{-5}]$ & \mbox{$0.8897$} \\
Age 40 years & Age-specific serial path product $(a + a_G \cdot G)(b + b_G \cdot G)(f + f_G \cdot G)$ & $1.01\times10^{-6}$ & $[-1.892\times10^{-6}$, $3.912\times10^{-6}]$ & \mbox{$0.4951$} \\
Age 50 years & Age-specific serial path product $(a + a_G \cdot G)(b + b_G \cdot G)(f + f_G \cdot G)$ & $7.137\times10^{-7}$ & $[-8.764\times10^{-7}$, $2.304\times10^{-6}]$ & \mbox{$0.379$} \\
Age 60 years & Age-specific serial path product $(a + a_G \cdot G)(b + b_G \cdot G)(f + f_G \cdot G)$ & $-1.258\times10^{-6}$ & $[-4.882\times10^{-6}$, $2.366\times10^{-6}]$ & \mbox{$0.4964$} \\
\bottomrule
\end{longtable}}

\paragraph{23.\ Gynoid fat mass $\times$ age interaction: PWV slope in the joint model}
\textit{Design.} Timing design and sample rules as in the $\pm$365-day joint model: date span of the four measurements $\le$ 365 days (bounds inclusive, any order), for each participant the set with the shortest span. Approximate age (gynoid fat mass measurement year minus birth year) mean-centered within the analysis sample, entered both as a main effect and as the modifier; additional adjustment for LDL-C, seated systolic blood pressure and sex (categorical term).

\textit{Model.} $Y = \beta_0 + \beta_A \cdot A + \beta_{\mathrm{LDL}} \cdot M_1 + \beta_{\mathrm{SBP}} \cdot M_2 + \beta_{\mathrm{age}} \cdot \mathrm{age}_c + \beta_{A \times \mathrm{age}} \cdot A \cdot \mathrm{age}_c + \gamma_{\mathrm{sex}} + \varepsilon$, linear regression; $A$, $M_1$, $M_2$ and $Y$ are gynoid fat mass, LDL-C, seated systolic blood pressure and PWV, all in original units, and $\mathrm{age}_c$ is age minus the sample mean age. $\beta_{A \times \mathrm{age}}$ is the change in the gynoid fat mass--PWV slope per 1-year increase in age; $\beta_A$ is the change in PWV per 1-unit increase in gynoid fat mass at the sample mean age.

\textit{Intervals.} 95\% confidence intervals.
{\footnotesize\setlength{\tabcolsep}{3pt}
\begin{longtable}{@{}>{\raggedright\arraybackslash}p{0.14\linewidth}>{\raggedright\arraybackslash}p{0.29\linewidth}>{\raggedleft\arraybackslash}p{0.13\linewidth}>{\raggedright\arraybackslash}p{0.22\linewidth}>{\raggedleft\arraybackslash}p{0.14\linewidth}@{}}
\toprule
Block & Parameter & Estimate & Interval & $P$ \\
\midrule\endhead
Interaction term & $\beta_{A \times \mathrm{age}}$: gynoid fat mass $\times$ age interaction coefficient (change in slope per 1-year increase in age) & $-1.512\times10^{-6}$ & $[-5.647\times10^{-6}$, $2.624\times10^{-6}]$ & \mbox{$0.4737$} \\
Slope at mean age & $\beta_A$: gynoid fat mass slope at the sample mean age & $6.172\times10^{-5}$ & $[2.515\times10^{-5}$, $9.828\times10^{-5}]$ &  \\
\bottomrule
\end{longtable}}

\subsubsection*{Sensitivity analyses}

\paragraph{24.\ Shorter-window single-mediator path: gynoid fat mass $\to$ LDL-C $\to$ PWV}
\textit{Design.} Gynoid fat mass, LDL-C and PWV measured in strict date order: LDL-C 90--365 days after the gynoid fat mass measurement, PWV 180--730 days after LDL-C (all bounds inclusive); one measurement chain per participant. Both windows nested within the primary-path windows. Adjustment for approximate age at the gynoid fat mass measurement (computed from calendar years) and sex (categorical term). PWV is the mean of the available sides.

\textit{Model.} Linear regressions with the same equations as the primary path: $M = \alpha_0 + \alpha_A \cdot A + \alpha_C^{\top} C + \varepsilon_M$; $Y = \beta_0 + \beta_A \cdot A + \beta_M \cdot M + \beta_C^{\top} C + \varepsilon_Y$; $Y = \gamma_0 + \gamma_A \cdot A + \gamma_C^{\top} C + \varepsilon_T$. $A$, $M$ and $Y$ are gynoid fat mass, LDL-C and thigh-to-ankle PWV (original units), and $C$ is age and sex. $\alpha_A$ is the change in LDL-C per 1-unit increase in gynoid fat mass; $\beta_M$ and $\beta_A$ are the mutually adjusted slopes of LDL-C and gynoid fat mass on PWV; $\gamma_A$ is the total slope of gynoid fat mass on PWV; path product $\theta = \alpha_A \cdot \beta_M$.

\textit{Intervals.} Bootstrap 95\% percentile intervals.
{\footnotesize\setlength{\tabcolsep}{3pt}
\begin{longtable}{@{}>{\raggedright\arraybackslash}p{0.15\linewidth}>{\raggedright\arraybackslash}p{0.40\linewidth}>{\raggedleft\arraybackslash}p{0.14\linewidth}>{\raggedright\arraybackslash}p{0.24\linewidth}@{}}
\toprule
Block & Parameter & Estimate & Interval \\
\midrule\endhead
Path components (shorter windows) & $\alpha_A$: slope of gynoid fat mass $\to$ LDL-C ($A \to M$) & $-0.001178$ & $[-0.0064$, $0.005374]$ \\
 & $\beta_M$: slope of LDL-C $\to$ PWV, adjusted for gynoid fat mass ($M \to Y \mid A$) & $-0.007099$ & $[-0.0126$, $-0.001129]$ \\
 & $\beta_A$: slope of gynoid fat mass $\to$ PWV, adjusted for LDL-C ($A \to Y \mid M$) & $0.0001702$ & $[4.211\times10^{-5}$, $0.0003379]$ \\
 & $\gamma_A$: adjusted total slope of gynoid fat mass $\to$ PWV ($A \to Y$) & $0.0001785$ & $[5.063\times10^{-5}$, $0.0003445]$ \\
Path index (shorter windows) & $\theta = \alpha_A \cdot \beta_M$: path-product association index & $8.365\times10^{-6}$ & $[-3.865\times10^{-5}$, $5.58\times10^{-5}]$ \\
\bottomrule
\end{longtable}}

\paragraph{25.\ Broad-window serial path: gynoid fat mass $\to$ LDL $\to$ systolic blood pressure $\to$ PWV}
\textit{Design.} Each gynoid fat measurement date as anchor; nearest measurement of LDL cholesterol within 365 days before or after the anchor, of seated systolic blood pressure within 180 days before or after that LDL, and of PWV within 30 days before or after that systolic blood pressure (either order, bounds inclusive); one row per anchor date. All four equations share the same sample, with adjustment for approximate age (anchor year minus birth year) and recorded sex; variables not standardized.

\textit{Model.} $M_1 = i_1 + a \cdot A + \gamma_1^{\top} C + \varepsilon_1$; $M_2 = i_2 + d \cdot A + b \cdot M_1 + \gamma_2^{\top} C + \varepsilon_2$; $Y = i_3 + c' \cdot A + e \cdot M_1 + f \cdot M_2 + \gamma_3^{\top} C + \varepsilon_3$; $Y = i_T + \tau \cdot A + \gamma_T^{\top} C + \varepsilon_T$. All four equations are linear regressions; $A$, $M_1$, $M_2$ and $Y$ are gynoid fat mass, LDL cholesterol, seated systolic blood pressure and thigh-to-ankle PWV (original units), and $C$ is age and sex. Coefficients are the change in the dependent variable per 1 original unit higher in the independent variable; reported: the seven coefficients, plus the serial path product $a \cdot b \cdot f$, the LDL-only $a \cdot e$, the systolic-blood-pressure-only $d \cdot f$, the conditional slope $c'$ and the total association slope $\tau$.

\textit{Intervals.} 95\% confidence intervals. $P$ values not corrected for multiple comparisons.
{\footnotesize\setlength{\tabcolsep}{3pt}
\begin{longtable}{@{}>{\raggedright\arraybackslash}p{0.14\linewidth}>{\raggedright\arraybackslash}p{0.29\linewidth}>{\raggedleft\arraybackslash}p{0.13\linewidth}>{\raggedright\arraybackslash}p{0.22\linewidth}>{\raggedleft\arraybackslash}p{0.14\linewidth}@{}}
\toprule
Block & Parameter & Estimate & Interval & $P$ \\
\midrule\endhead
Path decomposition (broad windows) & $a \cdot b \cdot f$: serial path product gynoid fat mass $\to$ LDL cholesterol $\to$ seated systolic blood pressure $\to$ PWV & $1.351\times10^{-6}$ & $[-1.314\times10^{-7}$, $2.834\times10^{-6}]$ & \mbox{$0.07407$} \\
 & $a \cdot e$: path product through LDL cholesterol only & $2.404\times10^{-6}$ & $[-4.178\times10^{-7}$, $5.226\times10^{-6}]$ & \mbox{$0.09496$} \\
 & $d \cdot f$: path product through seated systolic blood pressure only & $0.0001213$ & $[9.783\times10^{-5}$, $0.0001448]$ & \mbox{$4.38\times10^{-24}$} \\
 & $c'$: conditional slope of gynoid fat mass $\to$ PWV, adjusted for both mediators & $5.172\times10^{-5}$ & $[9.388\times10^{-6}$, $9.405\times10^{-5}]$ & \mbox{$0.01664$} \\
 & $\tau$: total association slope of gynoid fat mass $\to$ PWV & $0.0001768$ & $[0.0001315$, $0.0002221]$ & \mbox{$2.005\times10^{-14}$} \\
Path components (broad windows) & $a$: slope of gynoid fat mass $\to$ LDL cholesterol ($A \to M_1$) & $0.001679$ & $[0.0007121$, $0.002646]$ & \mbox{$0.000666$} \\
 & $d$: slope of gynoid fat mass $\to$ seated systolic blood pressure, adjusted for LDL ($A \to M_2 \mid M_1$) & $0.002301$ & $[0.001895$, $0.002708]$ & \mbox{$1.297\times10^{-28}$} \\
 & $b$: slope of LDL cholesterol $\to$ seated systolic blood pressure, adjusted for gynoid fat mass ($M_1 \to M_2 \mid A$) & $0.01527$ & $[0.0002821$, $0.03025]$ & \mbox{$0.04584$} \\
 & $c'$: slope of gynoid fat mass $\to$ PWV, adjusted for LDL and systolic blood pressure ($A \to Y \mid M_1, M_2$) & $5.172\times10^{-5}$ & $[9.388\times10^{-6}$, $9.405\times10^{-5}]$ & \mbox{$0.01664$} \\
 & $e$: slope of LDL cholesterol $\to$ PWV, adjusted for gynoid fat mass and systolic blood pressure ($M_1 \to Y \mid A, M_2$) & $0.001432$ & $[-0.0001611$, $0.003025]$ & \mbox{$0.07812$} \\
 & $f$: slope of seated systolic blood pressure $\to$ PWV, adjusted for gynoid fat mass and LDL ($M_2 \to Y \mid A, M_1$) & $0.05272$ & $[0.04862$, $0.05682]$ & \mbox{$4.583\times10^{-140}$} \\
 & $\tau$: total association slope of gynoid fat mass $\to$ PWV ($A \to Y$) & $0.0001768$ & $[0.0001315$, $0.0002221]$ & \mbox{$2.005\times10^{-14}$} \\
\bottomrule
\end{longtable}}

\paragraph{26.\ Partially ordered serial path: gynoid fat mass $\to$ LDL $\to$ systolic blood pressure $\to$ PWV}
\textit{Design.} Each gynoid fat measurement date as anchor; nearest LDL cholesterol measurement 1--365 days after the anchor; nearest seated systolic blood pressure within 90 days before or after that LDL, and nearest PWV within 14 days before or after that systolic blood pressure (either order; all bounds inclusive); one row per anchor date. All four equations share the same sample, with adjustment for approximate age (anchor year minus birth year) and recorded sex; variables not standardized.

\textit{Model.} $M_1 = i_1 + a \cdot A + \gamma_1^{\top} C + \varepsilon_1$; $M_2 = i_2 + d \cdot A + b \cdot M_1 + \gamma_2^{\top} C + \varepsilon_2$; $Y = i_3 + c' \cdot A + e \cdot M_1 + f \cdot M_2 + \gamma_3^{\top} C + \varepsilon_3$; $Y = i_T + \tau \cdot A + \gamma_T^{\top} C + \varepsilon_T$. All four equations are linear regressions; $A$, $M_1$, $M_2$ and $Y$ are gynoid fat mass, LDL cholesterol, seated systolic blood pressure and thigh-to-ankle PWV (original units), and $C$ is age and sex. Coefficients are the change in the dependent variable per 1 original unit higher in the independent variable; reported: the seven coefficients, plus the serial path product $a \cdot b \cdot f$, the LDL-only $a \cdot e$, the systolic-blood-pressure-only $d \cdot f$, the conditional slope $c'$ and the total association slope $\tau$.

\textit{Intervals.} 95\% confidence intervals. $P$ values not corrected for multiple comparisons.
{\footnotesize\setlength{\tabcolsep}{3pt}
\begin{longtable}{@{}>{\raggedright\arraybackslash}p{0.14\linewidth}>{\raggedright\arraybackslash}p{0.29\linewidth}>{\raggedleft\arraybackslash}p{0.13\linewidth}>{\raggedright\arraybackslash}p{0.22\linewidth}>{\raggedleft\arraybackslash}p{0.14\linewidth}@{}}
\toprule
Block & Parameter & Estimate & Interval & $P$ \\
\midrule\endhead
Path decomposition (partial ordering) & $c'$: conditional slope of gynoid fat mass $\to$ PWV, adjusted for both mediators & $3.624\times10^{-5}$ & $[-0.0001309$, $0.0002034]$ & \mbox{$0.6709$} \\
 & $a \cdot b \cdot f$: serial path product gynoid fat mass $\to$ LDL cholesterol $\to$ seated systolic blood pressure $\to$ PWV & $3.066\times10^{-6}$ & $[-6.329\times10^{-6}$, $1.246\times10^{-5}]$ & \mbox{$0.5224$} \\
 & $\tau$: total association slope of gynoid fat mass $\to$ PWV & $0.0001554$ & $[-4.404\times10^{-5}$, $0.0003547]$ & \mbox{$0.1267$} \\
 & $a \cdot e$: path product through LDL cholesterol only & $-6.388\times10^{-6}$ & $[-4.095\times10^{-5}$, $2.818\times10^{-5}]$ & \mbox{$0.7172$} \\
 & $d \cdot f$: path product through seated systolic blood pressure only & $0.0001224$ & $[-2.592\times10^{-5}$, $0.0002708]$ & \mbox{$0.1058$} \\
Path components (partial ordering) & $a$: slope of gynoid fat mass $\to$ LDL cholesterol ($A \to M_1$) & $-0.004014$ & $[-0.01628$, $0.008249]$ & \mbox{$0.5212$} \\
 & $b$: slope of LDL cholesterol $\to$ seated systolic blood pressure, adjusted for gynoid fat mass ($M_1 \to M_2 \mid A$) & $-0.01098$ & $[-0.05575$, $0.03378]$ & \mbox{$0.6306$} \\
 & $d$: slope of gynoid fat mass $\to$ seated systolic blood pressure, adjusted for LDL ($A \to M_2 \mid M_1$) & $0.00176$ & $[-0.0003251$, $0.003846]$ & \mbox{$0.09804$} \\
 & $c'$: slope of gynoid fat mass $\to$ PWV, adjusted for LDL and systolic blood pressure ($A \to Y \mid M_1, M_2$) & $3.624\times10^{-5}$ & $[-0.0001309$, $0.0002034]$ & \mbox{$0.6709$} \\
 & $e$: slope of LDL cholesterol $\to$ PWV, adjusted for gynoid fat mass and systolic blood pressure ($M_1 \to Y \mid A, M_2$) & $0.001592$ & $[-0.002627$, $0.00581]$ & \mbox{$0.4596$} \\
 & $f$: slope of seated systolic blood pressure $\to$ PWV, adjusted for gynoid fat mass and LDL ($M_2 \to Y \mid A, M_1$) & $0.06955$ & $[0.05699$, $0.08212]$ & \mbox{$2.025\times10^{-27}$} \\
 & $\tau$: total association slope of gynoid fat mass $\to$ PWV ($A \to Y$) & $0.0001554$ & $[-4.404\times10^{-5}$, $0.0003547]$ & \mbox{$0.1267$} \\
\bottomrule
\end{longtable}}

\paragraph{27.\ Extended adjustment: gynoid fat mass $\to$ LDL $\to$ systolic blood pressure $\to$ PWV}
\textit{Design.} Windows as in the primary path: LDL cholesterol within 180 days before or after the gynoid fat measurement date, seated systolic blood pressure within 90 days before or after that LDL, PWV within 14 days before or after that systolic blood pressure, nearest measurement each (either order, bounds inclusive); one row per gynoid fat measurement date. In addition to approximate age and recorded sex, adjustment for education (most recent on that date or within the preceding 3650 days), smoking status and alcohol consumption frequency (most recent on that date or within the preceding 365 days); variables not standardized.

\textit{Model.} $M_1 = i_1 + a \cdot A + \gamma_1^{\top} C + \varepsilon_1$; $M_2 = i_2 + d \cdot A + b \cdot M_1 + \gamma_2^{\top} C + \varepsilon_2$; $Y = i_3 + c' \cdot A + e \cdot M_1 + f \cdot M_2 + \gamma_3^{\top} C + \varepsilon_3$; $Y = i_T + \tau \cdot A + \gamma_T^{\top} C + \varepsilon_T$. All four equations are linear regressions; $A$, $M_1$, $M_2$ and $Y$ are gynoid fat mass, LDL cholesterol, seated systolic blood pressure and thigh-to-ankle PWV (original units), and $C$ is age, sex, education, smoking status and alcohol consumption frequency. Coefficients are the change in the dependent variable per 1 original unit higher in the independent variable; reported: the seven coefficients, plus the serial path product $a \cdot b \cdot f$, the LDL-only $a \cdot e$, the systolic-blood-pressure-only $d \cdot f$, the conditional slope $c'$ and the total association slope $\tau$.

\textit{Intervals.} 95\% confidence intervals. $P$ values not corrected for multiple comparisons.
{\footnotesize\setlength{\tabcolsep}{3pt}
\begin{longtable}{@{}>{\raggedright\arraybackslash}p{0.14\linewidth}>{\raggedright\arraybackslash}p{0.29\linewidth}>{\raggedleft\arraybackslash}p{0.13\linewidth}>{\raggedright\arraybackslash}p{0.22\linewidth}>{\raggedleft\arraybackslash}p{0.14\linewidth}@{}}
\toprule
Block & Parameter & Estimate & Interval & $P$ \\
\midrule\endhead
Path decomposition (extended adjustment) & $a \cdot b \cdot f$: serial path product gynoid fat mass $\to$ LDL cholesterol $\to$ seated systolic blood pressure $\to$ PWV & $-1.325\times10^{-6}$ & $[-4.321\times10^{-6}$, $1.672\times10^{-6}]$ & \mbox{$0.3862$} \\
 & $a \cdot e$: path product through LDL cholesterol only & $5.805\times10^{-6}$ & $[-1.096\times10^{-6}$, $1.271\times10^{-5}]$ & \mbox{$0.09919$} \\
 & $d \cdot f$: path product through seated systolic blood pressure only & $0.0001145$ & $[7.637\times10^{-5}$, $0.0001527]$ & \mbox{$4.085\times10^{-9}$} \\
 & $c'$: conditional slope of gynoid fat mass $\to$ PWV, adjusted for both mediators & $2.459\times10^{-5}$ & $[-4.387\times10^{-5}$, $9.304\times10^{-5}]$ & \mbox{$0.4814$} \\
 & $\tau$: total association slope of gynoid fat mass $\to$ PWV & $0.0001436$ & $[6.752\times10^{-5}$, $0.0002197]$ & \mbox{$0.0002163$} \\
Path components (extended adjustment) & $a$: slope of gynoid fat mass $\to$ LDL cholesterol ($A \to M_1$) & $0.001877$ & $[-2.68\times10^{-5}$, $0.00378]$ & \mbox{$0.05331$} \\
 & $d$: slope of gynoid fat mass $\to$ seated systolic blood pressure, adjusted for LDL ($A \to M_2 \mid M_1$) & $0.002198$ & $[0.001505$, $0.002892]$ & \mbox{$5.263\times10^{-10}$} \\
 & $b$: slope of LDL cholesterol $\to$ seated systolic blood pressure, adjusted for gynoid fat mass ($M_1 \to M_2 \mid A$) & $-0.01355$ & $[-0.03615$, $0.009053]$ & \mbox{$0.24$} \\
 & $c'$: slope of gynoid fat mass $\to$ PWV, adjusted for LDL and systolic blood pressure ($A \to Y \mid M_1, M_2$) & $2.459\times10^{-5}$ & $[-4.387\times10^{-5}$, $9.304\times10^{-5}]$ & \mbox{$0.4814$} \\
 & $e$: slope of LDL cholesterol $\to$ PWV, adjusted for gynoid fat mass and systolic blood pressure ($M_1 \to Y \mid A, M_2$) & $0.003094$ & $[0.000476$, $0.005711]$ & \mbox{$0.02054$} \\
 & $f$: slope of seated systolic blood pressure $\to$ PWV, adjusted for gynoid fat mass and LDL ($M_2 \to Y \mid A, M_1$) & $0.0521$ & $[0.04517$, $0.05904]$ & \mbox{$4.61\times10^{-49}$} \\
 & $\tau$: total association slope of gynoid fat mass $\to$ PWV ($A \to Y$) & $0.0001436$ & $[6.752\times10^{-5}$, $0.0002197]$ & \mbox{$0.0002163$} \\
\bottomrule
\end{longtable}}

\paragraph{28.\ Same-region bone mass comparator: gynoid bone mass $\to$ LDL $\to$ systolic blood pressure $\to$ PWV}
\textit{Design.} Exposure replaced by the bone mass of the same DXA gynoid region, otherwise as in the primary path: LDL cholesterol within 180 days before or after the bone mass measurement date, seated systolic blood pressure within 90 days before or after that LDL, PWV within 14 days before or after that systolic blood pressure, nearest measurement each (either order, bounds inclusive); one row per bone mass measurement date. Adjustment for approximate age (bone mass measurement year minus birth year) and recorded sex; variables not standardized.

\textit{Model.} $M_1 = i_1 + a \cdot A + \gamma_1^{\top} C + \varepsilon_1$; $M_2 = i_2 + d \cdot A + b \cdot M_1 + \gamma_2^{\top} C + \varepsilon_2$; $Y = i_3 + c' \cdot A + e \cdot M_1 + f \cdot M_2 + \gamma_3^{\top} C + \varepsilon_3$; $Y = i_T + \tau \cdot A + \gamma_T^{\top} C + \varepsilon_T$. All four equations are linear regressions; $A$, $M_1$, $M_2$ and $Y$ are gynoid bone mass, LDL cholesterol, seated systolic blood pressure and thigh-to-ankle PWV (original units), and $C$ is age and sex. Coefficients are the change in the dependent variable per 1 original unit higher in the independent variable; reported: the seven coefficients, plus the serial path product $a \cdot b \cdot f$, the LDL-only $a \cdot e$, the systolic-blood-pressure-only $d \cdot f$, the conditional slope $c'$ and the total association slope $\tau$.

\textit{Intervals.} 95\% confidence intervals. $P$ values not corrected for multiple comparisons.
{\footnotesize\setlength{\tabcolsep}{3pt}
\begin{longtable}{@{}>{\raggedright\arraybackslash}p{0.14\linewidth}>{\raggedright\arraybackslash}p{0.29\linewidth}>{\raggedleft\arraybackslash}p{0.13\linewidth}>{\raggedright\arraybackslash}p{0.22\linewidth}>{\raggedleft\arraybackslash}p{0.14\linewidth}@{}}
\toprule
Block & Parameter & Estimate & Interval & $P$ \\
\midrule\endhead
Path decomposition (gynoid bone mass) & $a \cdot b \cdot f$: serial path product gynoid bone mass $\to$ LDL cholesterol $\to$ seated systolic blood pressure $\to$ PWV & $-8.195\times10^{-6}$ & $[-3.659\times10^{-5}$, $2.02\times10^{-5}]$ & \mbox{$0.5716$} \\
 & $a \cdot e$: path product through LDL cholesterol only & $-3.906\times10^{-5}$ & $[-0.0001521$, $7.396\times10^{-5}]$ & \mbox{$0.4982$} \\
 & $d \cdot f$: path product through seated systolic blood pressure only & $0.002389$ & $[0.001455$, $0.003323]$ & \mbox{$5.406\times10^{-7}$} \\
 & $c'$: conditional slope of gynoid bone mass $\to$ PWV, adjusted for both mediators & $0.002679$ & $[0.001008$, $0.004349]$ & \mbox{$0.001672$} \\
 & $\tau$: total association slope of gynoid bone mass $\to$ PWV & $0.005021$ & $[0.003131$, $0.006911]$ & \mbox{$1.923\times10^{-7}$} \\
Path components (gynoid bone mass) & $a$: slope of gynoid bone mass $\to$ LDL cholesterol ($A \to M_1$) & $-0.01766$ & $[-0.0641$, $0.02878]$ & \mbox{$0.4561$} \\
 & $d$: slope of gynoid bone mass $\to$ seated systolic blood pressure, adjusted for LDL ($A \to M_2 \mid M_1$) & $0.04662$ & $[0.02945$, $0.06379]$ & \mbox{$1.028\times10^{-7}$} \\
 & $b$: slope of LDL cholesterol $\to$ seated systolic blood pressure, adjusted for gynoid bone mass ($M_1 \to M_2 \mid A$) & $0.009056$ & $[-0.008582$, $0.02669]$ & \mbox{$0.3143$} \\
 & $c'$: slope of gynoid bone mass $\to$ PWV, adjusted for LDL and systolic blood pressure ($A \to Y \mid M_1, M_2$) & $0.002679$ & $[0.001008$, $0.004349]$ & \mbox{$0.001672$} \\
 & $e$: slope of LDL cholesterol $\to$ PWV, adjusted for gynoid bone mass and systolic blood pressure ($M_1 \to Y \mid A, M_2$) & $0.002212$ & $[0.0002764$, $0.004148]$ & \mbox{$0.0251$} \\
 & $f$: slope of seated systolic blood pressure $\to$ PWV, adjusted for gynoid bone mass and LDL ($M_2 \to Y \mid A, M_1$) & $0.05124$ & $[0.0462$, $0.05629]$ & \mbox{$3.153\times10^{-88}$} \\
 & $\tau$: total association slope of gynoid bone mass $\to$ PWV ($A \to Y$) & $0.005021$ & $[0.003131$, $0.006911]$ & \mbox{$1.923\times10^{-7}$} \\
\bottomrule
\end{longtable}}

\paragraph{29.\ Joint model with extended adjustment: additional adjustment for total fat mass, cohort and height}
\textit{Design.} Timing design and sample rules as in the $\pm$365-day joint model: date span of the four measurements $\le$ 365 days (bounds inclusive, any order), for each participant the set with the shortest span. Both models share the same sample, with adjustment beyond age and sex for total fat mass on the day of the gynoid fat mass measurement, recruitment cohort (categorical term) and the nearest height within 365 days before or after PWV.

\textit{Model.} Two linear regressions, both with PWV ($Y$) as the outcome. Joint model: $Y = \beta_0 + \beta_A \cdot A + \beta_{\mathrm{LDL}} \cdot M_1 + \beta_{\mathrm{SBP}} \cdot M_2 + \delta^{\top} C + \varepsilon$; reduced model: $Y = \alpha_0 + \alpha_A \cdot A + \delta^{\prime\top} C + \varepsilon$; $C$ is age, sex, cohort, total fat mass and height. $\beta_A$, $\beta_{\mathrm{LDL}}$ and $\beta_{\mathrm{SBP}}$ are the change in PWV per 1-unit increase in the respective variable with the other joint-model terms fixed; $\alpha_A$ is the gynoid fat mass slope in the same sample with adjustment for $C$ only.

\textit{Intervals.} 95\% confidence intervals.
{\footnotesize\setlength{\tabcolsep}{3pt}
\begin{longtable}{@{}>{\raggedright\arraybackslash}p{0.14\linewidth}>{\raggedright\arraybackslash}p{0.29\linewidth}>{\raggedleft\arraybackslash}p{0.13\linewidth}>{\raggedright\arraybackslash}p{0.22\linewidth}>{\raggedleft\arraybackslash}p{0.14\linewidth}@{}}
\toprule
Block & Parameter & Estimate & Interval & $P$ \\
\midrule\endhead
$\pm$365 days, extended adjustment; joint model: gynoid fat mass + LDL-C + systolic blood pressure + $C$ & $\beta_A$: gynoid fat mass coefficient (joint model) & $-0.0001695$ & $[-0.0002638$, $-7.514\times10^{-5}]$ & \mbox{$0.0004329$} \\
 & $\beta_{\mathrm{LDL}}$: LDL-C coefficient (joint model) & $0.001019$ & $[-0.0004437$, $0.002481]$ &  \\
 & $\beta_{\mathrm{SBP}}$: seated systolic blood pressure coefficient (joint model) & $0.05006$ & $[0.04644$, $0.05367]$ &  \\
$\pm$365 days, extended adjustment; reduced model: gynoid fat mass + $C$ & $\alpha_A$: gynoid fat mass coefficient (reduced model, same sample) & $-0.0003733$ & $[-0.0004767$, $-0.0002699]$ &  \\
\bottomrule
\end{longtable}}

\paragraph{30.\ Repeated PWV measurements: joint model ($\pm$365 days)}
\textit{Design.} Each PWV measurement of each participant as one row, each row matched with gynoid fat mass, LDL-C and seated systolic blood pressure, with a date span of the four measurements $\le$ 365 days (bounds inclusive, any order), taking the set with the shortest span. Both models share the same sample; covariates and variable handling as in the $\pm$365-day joint model.

\textit{Model.} Two linear regressions, as in the $\pm$365-day joint model, with participant--PWV measurement date as the row. Joint model: $Y = \beta_0 + \beta_A \cdot A + \beta_{\mathrm{LDL}} \cdot M_1 + \beta_{\mathrm{SBP}} \cdot M_2 + \beta_{\mathrm{age}} \cdot \mathrm{age} + \gamma_{\mathrm{sex}} + \varepsilon$; reduced model: $Y = \alpha_0 + \alpha_A \cdot A + \alpha_{\mathrm{age}} \cdot \mathrm{age} + \gamma_{\mathrm{sex}} + \varepsilon$. $\beta_A$, $\beta_{\mathrm{LDL}}$ and $\beta_{\mathrm{SBP}}$ are the change in PWV per 1-unit increase in the respective variable with the other joint-model terms fixed; $\alpha_A$ is the gynoid fat mass slope in the same sample with adjustment for age and sex only.

\textit{Intervals.} 95\% confidence intervals.
{\footnotesize\setlength{\tabcolsep}{3pt}
\begin{longtable}{@{}>{\raggedright\arraybackslash}p{0.14\linewidth}>{\raggedright\arraybackslash}p{0.29\linewidth}>{\raggedleft\arraybackslash}p{0.13\linewidth}>{\raggedright\arraybackslash}p{0.22\linewidth}>{\raggedleft\arraybackslash}p{0.14\linewidth}@{}}
\toprule
Block & Parameter & Estimate & Interval & $P$ \\
\midrule\endhead
Repeated PWV rows; joint model: gynoid fat mass + LDL-C + systolic blood pressure + age + sex & $\beta_A$: gynoid fat mass coefficient (joint model) & $5.899\times10^{-5}$ & $[2.343\times10^{-5}$, $9.455\times10^{-5}]$ & \mbox{$0.001153$} \\
 & $\beta_{\mathrm{LDL}}$: LDL-C coefficient (joint model) & $0.001002$ & $[-0.0003809$, $0.002386]$ &  \\
 & $\beta_{\mathrm{SBP}}$: seated systolic blood pressure coefficient (joint model) & $0.05257$ & $[0.04912$, $0.05601]$ &  \\
Repeated PWV rows; reduced model: gynoid fat mass + age + sex & $\alpha_A$: gynoid fat mass coefficient (reduced model, same sample) & $0.0001841$ & $[0.0001455$, $0.0002227]$ &  \\
\bottomrule
\end{longtable}}

\FloatBarrier

\subsection{Gynoid fat--GGT--DBP--PWV}
\label{app:rq-designs:8e5f}

\paragraph{Research question and measures.}
The research question concerns the association of gynoid fat mass with lower-limb arterial stiffness via $\gamma$-glutamyl transferase and diastolic blood pressure, with data from the HPP cohort. The exposure is gynoid fat mass from body composition measurement; the mediators are blood-test $\gamma$-glutamyl transferase (GGT) and seated diastolic blood pressure (mean of the available values of two seated readings); the outcome is thigh-to-ankle pulse wave velocity (PWV, mean of the available sides). Unless otherwise stated, coefficients are SD changes per 1 SD after standardization within each unit's analysis sample.

\subsubsection*{Primary pathways}

\paragraph{1.\ Single-mediator path: gynoid fat mass $\to$ GGT $\to$ thigh-to-ankle PWV}
\textit{Design.} Gynoid fat mass, GGT and PWV measured in strict date order: GGT 30--365 days after the fat mass measurement and PWV 90--730 days after GGT (bounds inclusive); one measurement chain per participant. Adjustment for approximate age at the fat mass measurement, sex, cohort, and the most recent height on that day or within the preceding 365 days. Fat mass, GGT, PWV, age and height standardized within the analysis sample; sex and cohort as categorical terms.

\textit{Model.} $M = \alpha_0 + \alpha_A\cdot A + \alpha_C^\top C + \varepsilon_M$; $Y = \beta_0 + \beta_A\cdot A + \beta_M\cdot M + \beta_C^\top C + \varepsilon_Y$; $Y = \gamma_0 + \gamma_A\cdot A + \gamma_C^\top C + \varepsilon_T$. All three are linear regressions; $A$, $M$ and $Y$ are standardized gynoid fat mass, GGT and thigh-to-ankle PWV, and $C$ the covariates. $\alpha_A$: SD change in GGT per 1 SD higher fat mass; $\beta_M$, $\beta_A$: standardized slopes of PWV on GGT and on fat mass, respectively, under mutual adjustment; $\gamma_A$: total standardized slope of PWV on fat mass adjusted for $C$ only; path product $\theta = \alpha_A\cdot\beta_M$.

\textit{Intervals.} 95\% confidence intervals.
{\footnotesize\setlength{\tabcolsep}{3pt}
\begin{longtable}{@{}>{\raggedright\arraybackslash}p{0.15\linewidth}>{\raggedright\arraybackslash}p{0.40\linewidth}>{\raggedleft\arraybackslash}p{0.14\linewidth}>{\raggedright\arraybackslash}p{0.24\linewidth}@{}}
\toprule
Block & Parameter & Estimate & Interval \\
\midrule\endhead
Path components & $\alpha_A$: standardized slope, gynoid fat mass $\to$ GGT ($A \to M$) & $0.1001$ & $[-0.2836$, $0.4837]$ \\
 & $\beta_M$: standardized slope, GGT $\to$ thigh-to-ankle PWV, adjusted for fat mass ($M \to Y \mid A$) & $0.2973$ & $[-0.2258$, $0.8204]$ \\
 & $\beta_A$: standardized slope, gynoid fat mass $\to$ PWV, adjusted for GGT ($A \to Y \mid M$) & $0.08433$ & $[-0.1296$, $0.2983]$ \\
 & $\gamma_A$: adjusted total standardized slope, gynoid fat mass $\to$ PWV ($A \to Y$) & $0.1141$ & $[-0.0971$, $0.3253]$ \\
Path index & $\theta = \alpha_A\cdot\beta_M$: path-product association index & $0.02975$ & $[-0.04817$, $0.1077]$ \\
\bottomrule
\end{longtable}}

\paragraph{2.\ Serial path: gynoid fat mass $\to$ GGT $\to$ seated diastolic blood pressure $\to$ PWV}
\textit{Design.} GGT measured 0--365 days after gynoid fat mass, seated blood pressure within 90 days before or after GGT (in either order), and PWV 0--365 days after blood pressure (all bounds inclusive, same day allowed); one chain starting from each gynoid fat mass measurement day, with multiple chains allowed per participant. All four equations share the same sample, with adjustment for approximate age on that measurement day, recorded sex and recruitment cohort. Diastolic blood pressure as the mean of the available values of two seated readings; PWV as the mean of the available sides. The four path variables standardized within the analysis sample.

\textit{Model.} $M_1 = \alpha_1 + a_1 A + \gamma_1^\top C + \varepsilon_1$; $M_2 = \alpha_2 + a_2 A + d_{21} M_1 + \gamma_2^\top C + \varepsilon_2$; $Y = \alpha_3 + c' A + b_1 M_1 + b_2 M_2 + \gamma_3^\top C + \varepsilon_3$; $Y = \alpha_4 + c A + \gamma_4^\top C + \varepsilon_4$. All four are linear regressions; $A$, $M_1$, $M_2$ and $Y$ are standardized gynoid fat mass, $\gamma$-glutamyl transferase (GGT), seated diastolic blood pressure and thigh-to-ankle PWV, and $C$ comprises age, sex and recruitment cohort. Reported: the serial path product $a_1 d_{21} b_2$, $a_1 b_1$ through GGT only, $a_2 b_2$ through diastolic blood pressure only, the sum of the three, the conditional slope $c'$ controlling for both mediators, the total association slope $c$, and seven standardized coefficients (SD difference in the target variable per 1 SD of the predictor).

\textit{Intervals.} 95\% confidence intervals.
{\footnotesize\setlength{\tabcolsep}{3pt}
\begin{longtable}{@{}>{\raggedright\arraybackslash}p{0.15\linewidth}>{\raggedright\arraybackslash}p{0.40\linewidth}>{\raggedleft\arraybackslash}p{0.14\linewidth}>{\raggedright\arraybackslash}p{0.24\linewidth}@{}}
\toprule
Block & Parameter & Estimate & Interval \\
\midrule\endhead
Path decomposition & Serial path product $a_1\cdot d_{21}\cdot b_2$ (through GGT and diastolic blood pressure) & $0.01453$ & $[-0.01809$, $0.04715]$ \\
 & Path product through GGT only, $a_1\cdot b_1$ & $0.04162$ & $[-0.03614$, $0.1194]$ \\
 & Path product through diastolic blood pressure only, $a_2\cdot b_2$ & $0.1129$ & $[-0.0109$, $0.2368]$ \\
 & Sum of the three path products, $a_1 d_{21} b_2 + a_1 b_1 + a_2 b_2$ & $0.1691$ & $[0.03115$, $0.307]$ \\
 & Conditional slope $c'$ (gynoid fat mass $\to$ PWV, controlling for both mediators) & $-0.1573$ & $[-0.3593$, $0.04473]$ \\
 & Total association slope $c$ (gynoid fat mass $\to$ PWV) & $0.01179$ & $[-0.2061$, $0.2297]$ \\
Component coefficients & Standardized slope $a_1$ (gynoid fat mass $\to$ GGT) & $0.3021$ & $[-0.07528$, $0.6795]$ \\
 & Standardized slope $d_{21}$ (GGT $\to$ diastolic blood pressure, controlling for gynoid fat mass) & $0.08822$ & $[-0.09946$, $0.2759]$ \\
 & Standardized slope $a_2$ (gynoid fat mass $\to$ diastolic blood pressure, controlling for GGT) & $0.2072$ & $[-0.006459$, $0.4208]$ \\
 & Standardized slope $b_2$ (diastolic blood pressure $\to$ PWV, controlling for gynoid fat mass and GGT) & $0.5452$ & $[0.3958$, $0.6946]$ \\
 & Standardized slope $b_1$ (GGT $\to$ PWV, controlling for gynoid fat mass and diastolic blood pressure) & $0.1378$ & $[-0.02745$, $0.303]$ \\
 & Conditional slope $c'$ (gynoid fat mass $\to$ PWV, controlling for both mediators) & $-0.1573$ & $[-0.3593$, $0.04473]$ \\
 & Total association slope $c$ (gynoid fat mass $\to$ PWV) & $0.01179$ & $[-0.2061$, $0.2297]$ \\
\bottomrule
\end{longtable}}

\paragraph{3.\ Exposure--outcome association: gynoid fat mass and thigh-to-ankle PWV}
\textit{Design.} Gynoid fat mass and thigh-to-ankle PWV measured in the same period: each gynoid fat measurement day as an anchor, with the nearest PWV within 180 days before or after (bounds inclusive); multiple anchor days allowed per participant. PWV as the mean of the available values of the left and right sides. Adjustment for approximate age (anchor measurement year minus birth year), sex and sub-study; continuous variables kept on the original scale.

\textit{Model.} $Y = \beta_0 + \beta_1\cdot A + \beta_{\mathrm{age}}\cdot\mathrm{age} + \gamma_{\mathrm{sex}} + \gamma_{\mathrm{study}} + \varepsilon$ (linear regression; $A$ is gynoid fat mass and $Y$ thigh-to-ankle PWV, both on the original scale). $\beta_1$: difference in thigh-to-ankle PWV per 1 original unit higher gynoid fat mass, with age, sex and sub-study held fixed.

\textit{Intervals.} 95\% confidence intervals.
{\footnotesize\setlength{\tabcolsep}{3pt}
\begin{longtable}{@{}>{\raggedright\arraybackslash}p{0.14\linewidth}>{\raggedright\arraybackslash}p{0.29\linewidth}>{\raggedleft\arraybackslash}p{0.13\linewidth}>{\raggedright\arraybackslash}p{0.22\linewidth}>{\raggedleft\arraybackslash}p{0.14\linewidth}@{}}
\toprule
Block & Parameter & Estimate & Interval & $P$ \\
\midrule\endhead
$\pm$180-day window & $\beta_1$: adjusted association coefficient, gynoid fat mass--PWV & $0.0001889$ & $[0.000162$, $0.0002157]$ & \mbox{$3.045\times10^{-43}$} \\
\bottomrule
\end{longtable}}

\subsubsection*{Adjacent segments and pairwise associations}

\paragraph{4.\ Serial segment: gynoid fat mass $\to$ GGT $\to$ seated diastolic blood pressure}
\textit{Design.} Gynoid fat mass, GGT and seated blood pressure measured in strict date order: GGT 30--365 days after the fat mass measurement and seated blood pressure 30--365 days after GGT (bounds inclusive); one measurement chain per participant. Adjustment for approximate age at the fat mass measurement, sex, cohort, and the most recent height on that day or within the preceding 365 days. Fat mass, GGT, diastolic blood pressure, age and height standardized within the analysis sample; sex and cohort as categorical terms.

\textit{Model.} $M = \alpha_0 + \alpha_A\cdot A + \alpha_C^\top C + \varepsilon_M$; $Y = \gamma_0 + \gamma_A\cdot A + \gamma_C^\top C + \varepsilon_T$ (linear regressions; $A$, $M$ and $Y$ are standardized gynoid fat mass, GGT and seated diastolic blood pressure, and $C$ the covariates). $\alpha_A$: SD change in GGT per 1 SD higher fat mass; $\gamma_A$: adjusted total standardized slope of seated diastolic blood pressure on fat mass.

\textit{Intervals.} 95\% confidence intervals.
{\footnotesize\setlength{\tabcolsep}{3pt}
\begin{longtable}{@{}>{\raggedright\arraybackslash}p{0.15\linewidth}>{\raggedright\arraybackslash}p{0.40\linewidth}>{\raggedleft\arraybackslash}p{0.14\linewidth}>{\raggedright\arraybackslash}p{0.24\linewidth}@{}}
\toprule
Block & Parameter & Estimate & Interval \\
\midrule\endhead
Segment path components & $\alpha_A$: standardized slope, gynoid fat mass $\to$ GGT ($A \to M$) & $-0.4416$ & $[-2.23$, $1.346]$ \\
 & $\gamma_A$: adjusted total standardized slope, gynoid fat mass $\to$ seated diastolic blood pressure ($A \to Y$) & $1.645$ & $[1.505$, $1.785]$ \\
\bottomrule
\end{longtable}}

\paragraph{5.\ Pairwise association: gynoid fat mass $\to$ GGT (0--365 days)}
\textit{Design.} GGT measured 0--365 days after the gynoid fat mass measurement (bounds inclusive, same day allowed); each gynoid fat mass measurement day paired with the nearest subsequent GGT to form one row, with multiple rows allowed per participant, $n = 148$. Adjustment for approximate age on that measurement day (measurement year minus birth year), recorded sex and recruitment cohort. Gynoid fat mass and GGT standardized within the analysis sample; age in years; sex and cohort as categorical terms.

\textit{Model.} $M1_z = \beta_0 + \beta\cdot A_z + \beta_{\mathrm{age}}\cdot\mathrm{Age} + \gamma_{\mathrm{Sex}} + \gamma_{\mathrm{Cohort}} + \varepsilon$ (linear regression; $A$ is gynoid fat mass and $M1$ is $\gamma$-glutamyl transferase (GGT)). $\beta$: SD difference in GGT per 1 SD higher gynoid fat mass.

\textit{Intervals.} 95\% confidence intervals.
{\footnotesize\setlength{\tabcolsep}{3pt}
\begin{longtable}{@{}>{\raggedright\arraybackslash}p{0.15\linewidth}>{\raggedright\arraybackslash}p{0.40\linewidth}>{\raggedleft\arraybackslash}p{0.14\linewidth}>{\raggedright\arraybackslash}p{0.24\linewidth}@{}}
\toprule
Block & Parameter & Estimate & Interval \\
\midrule\endhead
Forward window 0--365 days & Adjusted standardized slope $\beta$ (gynoid fat mass $\to$ GGT) & $0.1924$ & $[-0.03711$, $0.4219]$ \\
\bottomrule
\end{longtable}}

\paragraph{6.\ Pairwise association: GGT--seated diastolic blood pressure ($\pm$90 days)}
\textit{Design.} Seated blood pressure measured within 90 days before or after the GGT assay (bounds inclusive, in either order); each GGT assay day paired with the seated blood pressure closest in date to form one row, with multiple rows allowed per participant, $n = 1760$. Adjustment for approximate age on the GGT assay day, recorded sex and recruitment cohort. Diastolic blood pressure as the mean of the available values of two seated readings. GGT and diastolic blood pressure standardized within the analysis sample; age in years; sex and cohort as categorical terms.

\textit{Model.} $M2_z = \beta_0 + \beta\cdot M1_z + \beta_{\mathrm{age}}\cdot\mathrm{Age} + \gamma_{\mathrm{Sex}} + \gamma_{\mathrm{Cohort}} + \varepsilon$ (linear regression; $M1$ is GGT and $M2$ is seated diastolic blood pressure). $\beta$: SD difference in seated diastolic blood pressure per 1 SD higher GGT.

\textit{Intervals.} 95\% confidence intervals.
{\footnotesize\setlength{\tabcolsep}{3pt}
\begin{longtable}{@{}>{\raggedright\arraybackslash}p{0.15\linewidth}>{\raggedright\arraybackslash}p{0.40\linewidth}>{\raggedleft\arraybackslash}p{0.14\linewidth}>{\raggedright\arraybackslash}p{0.24\linewidth}@{}}
\toprule
Block & Parameter & Estimate & Interval \\
\midrule\endhead
Symmetric window $\pm$90 days & Adjusted standardized slope $\beta$ (GGT $\to$ seated diastolic blood pressure) & $0.1124$ & $[0.04659$, $0.1782]$ \\
\bottomrule
\end{longtable}}

\paragraph{7.\ Pairwise association: seated diastolic blood pressure $\to$ PWV (0--365 days)}
\textit{Design.} PWV measured 0--365 days after the seated blood pressure measurement (bounds inclusive, same day allowed); each blood pressure measurement day paired with the nearest subsequent PWV to form one row, with multiple rows allowed per participant, $n = 10937$. Adjustment for approximate age on the blood pressure measurement day, recorded sex and recruitment cohort. Diastolic blood pressure as the mean of the available values of two seated readings; PWV as the mean of the available sides. Diastolic blood pressure and PWV standardized within the analysis sample; age in years; sex and cohort as categorical terms.

\textit{Model.} $Y_z = \beta_0 + \beta\cdot M2_z + \beta_{\mathrm{age}}\cdot\mathrm{Age} + \gamma_{\mathrm{Sex}} + \gamma_{\mathrm{Cohort}} + \varepsilon$ (linear regression; $M2$ is seated diastolic blood pressure and $Y$ is thigh-to-ankle PWV). $\beta$: SD difference in PWV per 1 SD higher seated diastolic blood pressure.

\textit{Intervals.} 95\% confidence intervals.
{\footnotesize\setlength{\tabcolsep}{3pt}
\begin{longtable}{@{}>{\raggedright\arraybackslash}p{0.15\linewidth}>{\raggedright\arraybackslash}p{0.40\linewidth}>{\raggedleft\arraybackslash}p{0.14\linewidth}>{\raggedright\arraybackslash}p{0.24\linewidth}@{}}
\toprule
Block & Parameter & Estimate & Interval \\
\midrule\endhead
Forward window 0--365 days & Adjusted standardized slope $\beta$ (seated diastolic blood pressure $\to$ thigh-to-ankle PWV) & $0.4478$ & $[0.4304$, $0.4651]$ \\
\bottomrule
\end{longtable}}

\paragraph{8.\ Pairwise total association: gynoid fat mass $\to$ PWV (0--730 days)}
\textit{Design.} PWV measured 0--730 days after the gynoid fat mass measurement (bounds inclusive, same day allowed); each gynoid fat mass measurement day paired with the nearest subsequent PWV to form one row, with multiple rows allowed per participant, $n = 8461$. Adjustment for approximate age on that measurement day, recorded sex and recruitment cohort. PWV as the mean of the available sides. Gynoid fat mass and PWV standardized within the analysis sample; age in years; sex and cohort as categorical terms.

\textit{Model.} $Y_z = \beta_0 + \beta\cdot A_z + \beta_{\mathrm{age}}\cdot\mathrm{Age} + \gamma_{\mathrm{Sex}} + \gamma_{\mathrm{Cohort}} + \varepsilon$ (linear regression; $A$ is gynoid fat mass and $Y$ is thigh-to-ankle PWV). $\beta$: SD difference in PWV per 1 SD higher gynoid fat mass.

\textit{Intervals.} 95\% confidence intervals.
{\footnotesize\setlength{\tabcolsep}{3pt}
\begin{longtable}{@{}>{\raggedright\arraybackslash}p{0.15\linewidth}>{\raggedright\arraybackslash}p{0.40\linewidth}>{\raggedleft\arraybackslash}p{0.14\linewidth}>{\raggedright\arraybackslash}p{0.24\linewidth}@{}}
\toprule
Block & Parameter & Estimate & Interval \\
\midrule\endhead
Forward window 0--730 days & Adjusted standardized total-association slope $\beta$ (gynoid fat mass $\to$ thigh-to-ankle PWV) & $0.1557$ & $[0.1341$, $0.1774]$ \\
\bottomrule
\end{longtable}}

\paragraph{9.\ Pairwise association: gynoid fat mass--GGT ($\pm$180 days)}
\textit{Design.} Gynoid fat mass and $\gamma$-glutamyl transferase (GGT) measured in the same period: each gynoid fat measurement day as an anchor, with the nearest GGT within 180 days before or after (bounds inclusive); multiple anchor days allowed per participant. Adjustment for approximate age (anchor measurement year minus birth year), sex and sub-study; continuous variables kept on the original scale.

\textit{Model.} $M1 = \beta_0 + \beta_1\cdot A + \beta_{\mathrm{age}}\cdot\mathrm{age} + \gamma_{\mathrm{sex}} + \gamma_{\mathrm{study}} + \varepsilon$ (linear regression; $A$ is gynoid fat mass and $M1$ GGT, both on the original scale). $\beta_1$: difference in GGT per 1 original unit higher gynoid fat mass, with age, sex and sub-study held fixed.

\textit{Intervals.} 95\% confidence intervals.
{\footnotesize\setlength{\tabcolsep}{3pt}
\begin{longtable}{@{}>{\raggedright\arraybackslash}p{0.14\linewidth}>{\raggedright\arraybackslash}p{0.29\linewidth}>{\raggedleft\arraybackslash}p{0.13\linewidth}>{\raggedright\arraybackslash}p{0.22\linewidth}>{\raggedleft\arraybackslash}p{0.14\linewidth}@{}}
\toprule
Block & Parameter & Estimate & Interval & $P$ \\
\midrule\endhead
$\pm$180-day window & $\beta_1$: adjusted association coefficient, gynoid fat mass--GGT & $0.001741$ & $[0.001015$, $0.002468]$ & \mbox{$2.646\times10^{-6}$} \\
\bottomrule
\end{longtable}}

\paragraph{10.\ Pairwise association: GGT--seated diastolic blood pressure ($\pm$180 days)}
\textit{Design.} GGT and seated diastolic blood pressure measured in the same period: each GGT measurement day as an anchor, with the nearest seated blood pressure within 180 days before or after (bounds inclusive); multiple anchor days allowed per participant. Diastolic blood pressure as the mean of the available values of two seated readings. Adjustment for approximate age (anchor measurement year minus birth year), sex and sub-study; continuous variables kept on the original scale.

\textit{Model.} $M2 = \beta_0 + \beta_1\cdot M1 + \beta_{\mathrm{age}}\cdot\mathrm{age} + \gamma_{\mathrm{sex}} + \gamma_{\mathrm{study}} + \varepsilon$ (linear regression; $M1$ is GGT and $M2$ seated diastolic blood pressure, both on the original scale). $\beta_1$: difference in seated diastolic blood pressure per 1 original unit higher GGT, with age, sex and sub-study held fixed.

\textit{Intervals.} 95\% confidence intervals.
{\footnotesize\setlength{\tabcolsep}{3pt}
\begin{longtable}{@{}>{\raggedright\arraybackslash}p{0.14\linewidth}>{\raggedright\arraybackslash}p{0.29\linewidth}>{\raggedleft\arraybackslash}p{0.13\linewidth}>{\raggedright\arraybackslash}p{0.22\linewidth}>{\raggedleft\arraybackslash}p{0.14\linewidth}@{}}
\toprule
Block & Parameter & Estimate & Interval & $P$ \\
\midrule\endhead
$\pm$180-day window & $\beta_1$: adjusted association coefficient, GGT--seated diastolic blood pressure & $0.05243$ & $[0.02754$, $0.07732]$ & \mbox{$3.646\times10^{-5}$} \\
\bottomrule
\end{longtable}}

\paragraph{11.\ Pairwise association: seated diastolic blood pressure--thigh-to-ankle PWV ($\pm$180 days)}
\textit{Design.} Seated diastolic blood pressure and thigh-to-ankle PWV measured in the same period: each seated blood pressure measurement day as an anchor, with the nearest PWV within 180 days before or after (bounds inclusive); multiple anchor days allowed per participant. Diastolic blood pressure as the mean of the available values of two seated readings. PWV as the mean of the available values of the left and right sides. Adjustment for approximate age (anchor measurement year minus birth year), sex and sub-study; continuous variables kept on the original scale.

\textit{Model.} $Y = \beta_0 + \beta_1\cdot M2 + \beta_{\mathrm{age}}\cdot\mathrm{age} + \gamma_{\mathrm{sex}} + \gamma_{\mathrm{study}} + \varepsilon$ (linear regression; $M2$ is seated diastolic blood pressure and $Y$ thigh-to-ankle PWV, both on the original scale). $\beta_1$: difference in thigh-to-ankle PWV per 1 original unit higher seated diastolic blood pressure, with age, sex and sub-study held fixed.

\textit{Intervals.} 95\% confidence intervals.
{\footnotesize\setlength{\tabcolsep}{3pt}
\begin{longtable}{@{}>{\raggedright\arraybackslash}p{0.14\linewidth}>{\raggedright\arraybackslash}p{0.29\linewidth}>{\raggedleft\arraybackslash}p{0.13\linewidth}>{\raggedright\arraybackslash}p{0.22\linewidth}>{\raggedleft\arraybackslash}p{0.14\linewidth}@{}}
\toprule
Block & Parameter & Estimate & Interval & $P$ \\
\midrule\endhead
$\pm$180-day window & $\beta_1$: adjusted association coefficient, seated diastolic blood pressure--PWV & $0.07883$ & $[0.07577$, $0.08188]$ & \mbox{$0$} \\
\bottomrule
\end{longtable}}

\paragraph{12.\ Pairwise association: gynoid fat mass--seated diastolic blood pressure ($\pm$180 days)}
\textit{Design.} Gynoid fat mass and seated diastolic blood pressure measured in the same period: each gynoid fat measurement day as an anchor, with the nearest seated blood pressure within 180 days before or after (bounds inclusive); multiple anchor days allowed per participant, $n = 8927$. Diastolic blood pressure as the mean of the available values of two seated readings. Adjustment for approximate age (anchor measurement year minus birth year), sex and sub-study; continuous variables kept on the original scale.

\textit{Model.} $M2 = \beta_0 + \beta_1\cdot A + \beta_{\mathrm{age}}\cdot\mathrm{age} + \gamma_{\mathrm{sex}} + \gamma_{\mathrm{study}} + \varepsilon$ (linear regression; $A$ is gynoid fat mass and $M2$ seated diastolic blood pressure, both on the original scale). $\beta_1$: difference in seated diastolic blood pressure per 1 original unit higher gynoid fat mass, with age, sex and sub-study held fixed.

\textit{Intervals.} 95\% confidence intervals.
{\footnotesize\setlength{\tabcolsep}{3pt}
\begin{longtable}{@{}>{\raggedright\arraybackslash}p{0.14\linewidth}>{\raggedright\arraybackslash}p{0.29\linewidth}>{\raggedleft\arraybackslash}p{0.13\linewidth}>{\raggedright\arraybackslash}p{0.22\linewidth}>{\raggedleft\arraybackslash}p{0.14\linewidth}@{}}
\toprule
Block & Parameter & Estimate & Interval & $P$ \\
\midrule\endhead
$\pm$180-day window & $\beta_1$: adjusted association coefficient, gynoid fat mass--seated diastolic blood pressure & $0.002006$ & $[0.001857$, $0.002155]$ & \mbox{$5.432\times10^{-154}$} \\
\bottomrule
\end{longtable}}

\paragraph{13.\ Pairwise association: GGT--thigh-to-ankle PWV ($\pm$180 days)}
\textit{Design.} GGT and thigh-to-ankle PWV measured in the same period: each GGT measurement day as an anchor, with the nearest PWV within 180 days before or after (bounds inclusive); multiple anchor days allowed per participant. PWV as the mean of the available values of the left and right sides. Adjustment for approximate age (anchor measurement year minus birth year), sex and sub-study; continuous variables kept on the original scale.

\textit{Model.} $Y = \beta_0 + \beta_1\cdot M1 + \beta_{\mathrm{age}}\cdot\mathrm{age} + \gamma_{\mathrm{sex}} + \gamma_{\mathrm{study}} + \varepsilon$ (linear regression; $M1$ is GGT and $Y$ thigh-to-ankle PWV, both on the original scale). $\beta_1$: difference in thigh-to-ankle PWV per 1 original unit higher GGT, with age, sex and sub-study held fixed.

\textit{Intervals.} 95\% confidence intervals.
{\footnotesize\setlength{\tabcolsep}{3pt}
\begin{longtable}{@{}>{\raggedright\arraybackslash}p{0.14\linewidth}>{\raggedright\arraybackslash}p{0.29\linewidth}>{\raggedleft\arraybackslash}p{0.13\linewidth}>{\raggedright\arraybackslash}p{0.22\linewidth}>{\raggedleft\arraybackslash}p{0.14\linewidth}@{}}
\toprule
Block & Parameter & Estimate & Interval & $P$ \\
\midrule\endhead
$\pm$180-day window & $\beta_1$: adjusted association coefficient, GGT--PWV & $0.003041$ & $[-0.0003577$, $0.00644]$ & \mbox{$0.07948$} \\
\bottomrule
\end{longtable}}

\subsubsection*{Joint models}

\paragraph{14.\ Joint model: gynoid fat mass, GGT, diastolic blood pressure, PWV ($\pm$180 days)}
\textit{Design.} Each gynoid fat measurement day as an anchor, with one measurement each of GGT, seated diastolic blood pressure and PWV, all four measurements pairwise no more than 180 days apart (in either order, bounds inclusive), taking the set with the shortest total span; multiple anchor days allowed per participant. Diastolic blood pressure as the mean of the available values of two seated readings. PWV as the mean of the available values of the left and right sides. Adjustment for approximate age (anchor measurement year minus birth year), sex and sub-study; continuous variables kept on the original scale.

\textit{Model.} $Y = \beta_0 + \beta_A\cdot A + \beta_{M1}\cdot M1 + \beta_{M2}\cdot M2 + \beta_{\mathrm{age}}\cdot\mathrm{age} + \gamma_{\mathrm{sex}} + \gamma_{\mathrm{study}} + \varepsilon$ (linear regression; $A$ is gynoid fat mass, $M1$ GGT, $M2$ seated diastolic blood pressure and $Y$ thigh-to-ankle PWV, all on the original scale). $\beta_A$, $\beta_{M1}$, $\beta_{M2}$: difference in PWV per 1 original unit higher value of the respective variable, with the other terms of the same model held fixed.

\textit{Intervals.} 95\% confidence intervals; $P$ values not adjusted for multiple comparisons.
{\footnotesize\setlength{\tabcolsep}{3pt}
\begin{longtable}{@{}>{\raggedright\arraybackslash}p{0.14\linewidth}>{\raggedright\arraybackslash}p{0.29\linewidth}>{\raggedleft\arraybackslash}p{0.13\linewidth}>{\raggedright\arraybackslash}p{0.22\linewidth}>{\raggedleft\arraybackslash}p{0.14\linewidth}@{}}
\toprule
Block & Parameter & Estimate & Interval & $P$ \\
\midrule\endhead
Mutually adjusted model ($\pm$180 days) & $\beta_A$: gynoid fat mass coefficient (mutually adjusted model) & $6.411\times10^{-5}$ & $[-2.883\times10^{-6}$, $0.0001311]$ & \mbox{$0.06071$} \\
 & $\beta_{M1}$: GGT coefficient (mutually adjusted model) & $-0.0003819$ & $[-0.003828$, $0.003064]$ & \mbox{$0.8281$} \\
 & $\beta_{M2}$: seated diastolic blood pressure coefficient (mutually adjusted model) & $0.07577$ & $[0.06564$, $0.0859]$ & \mbox{$1.189\times10^{-48}$} \\
\bottomrule
\end{longtable}}

\paragraph{15.\ Gynoid fat mass coefficient: before and after adding GGT and diastolic blood pressure ($\pm$180 days)}
\textit{Design.} Same sample as the $\pm$180-day joint model: each gynoid fat measurement day as an anchor, with one measurement each of GGT, seated diastolic blood pressure and PWV, all four measurements pairwise no more than 180 days apart (in either order, bounds inclusive). A reduced model (gynoid fat mass and covariates only) and a mutually adjusted model additionally including GGT and seated diastolic blood pressure, both fitted on this same sample; covariates are approximate age, sex and sub-study, and continuous variables are kept on the original scale.

\textit{Model.} Reduced model: $Y = \alpha_0 + \alpha_A\cdot A + \alpha_{\mathrm{age}}\cdot\mathrm{age} + \gamma_{\mathrm{sex}} + \gamma_{\mathrm{study}} + \varepsilon$; mutually adjusted model: $Y = \beta_0 + \beta_A\cdot A + \beta_{M1}\cdot M1 + \beta_{M2}\cdot M2 + \beta_{\mathrm{age}}\cdot\mathrm{age} + \gamma_{\mathrm{sex}} + \gamma_{\mathrm{study}} + \varepsilon$ (both linear regressions; $A$ is gynoid fat mass, $M1$ GGT, $M2$ seated diastolic blood pressure and $Y$ thigh-to-ankle PWV, all on the original scale). $\alpha_A$, $\beta_A$: difference in PWV per 1 original unit higher gynoid fat mass in the two models, respectively, with the other terms held fixed.

\textit{Intervals.} 95\% confidence intervals; $P$ values not adjusted for multiple comparisons.
{\footnotesize\setlength{\tabcolsep}{3pt}
\begin{longtable}{@{}>{\raggedright\arraybackslash}p{0.14\linewidth}>{\raggedright\arraybackslash}p{0.29\linewidth}>{\raggedleft\arraybackslash}p{0.13\linewidth}>{\raggedright\arraybackslash}p{0.22\linewidth}>{\raggedleft\arraybackslash}p{0.14\linewidth}@{}}
\toprule
Block & Parameter & Estimate & Interval & $P$ \\
\midrule\endhead
Reduced model: gynoid fat mass + covariates & $\alpha_A$: gynoid fat mass coefficient (reduced model) & $0.000239$ & $[0.0001709$, $0.0003072]$ & \mbox{$6.354\times10^{-12}$} \\
Mutually adjusted model: additionally including GGT and seated diastolic blood pressure & $\beta_A$: gynoid fat mass coefficient (mutually adjusted model) & $6.411\times10^{-5}$ & $[-2.883\times10^{-6}$, $0.0001311]$ & \mbox{$0.06071$} \\
\bottomrule
\end{longtable}}

\subsubsection*{Reverse direction}

\paragraph{16.\ Reverse path: PWV $\to$ GGT $\to$ gynoid fat mass}
\textit{Design.} Endpoint roles reversed: thigh-to-ankle PWV, GGT and gynoid fat mass measured in strict date order, with GGT 90--730 days after PWV and fat mass 30--365 days after GGT (bounds inclusive); one measurement chain per participant. Adjustment for approximate age at the PWV measurement, sex, cohort, and the most recent height on that day or within the preceding 365 days. Continuous variables standardized within the analysis sample; sex and cohort as categorical terms.

\textit{Model.} $A$ = PWV, $M$ = GGT, $Y$ = gynoid fat mass (all standardized values), $C$ the covariates; linear regressions: $M = \alpha_0 + \alpha_A\cdot A + \alpha_C^\top C + \varepsilon_M$; $Y = \beta_0 + \beta_A\cdot A + \beta_M\cdot M + \beta_C^\top C + \varepsilon_Y$; $Y = \gamma_0 + \gamma_A\cdot A + \gamma_C^\top C + \varepsilon_T$. Reported quantities: $\alpha_A$ (PWV $\to$ GGT), $\beta_M$ (GGT $\to$ fat mass, adjusted for PWV), $\beta_A$ (PWV $\to$ fat mass, adjusted for GGT), $\gamma_A$ (total slope, PWV $\to$ fat mass), the reverse path product $\theta = \alpha_A\cdot\beta_M$, and bias bounds for $\alpha_A$ and $\beta_M$, $\hat\beta \pm \mathrm{se}\cdot\sqrt{\nu r^2/(1 - r)}$, where $\mathrm{se}$ is the conventional standard error of the coefficient, $\nu$ the residual degrees of freedom, and $r$ the partial $R^2$ of a hypothetical omitted covariate $U$ with the coefficient's regressor and with the equation's outcome (the same value for both), set to 0.05, 0.10, 0.20 and 0.30.

\textit{Intervals.} $\alpha_A$, $\beta_M$, $\beta_A$, $\gamma_A$, $\theta$: 95\% confidence intervals. Bias-bound grid: one row per $r$; lower and upper limits are the coefficient $\pm$ the bias bound.
{\footnotesize\setlength{\tabcolsep}{3pt}
\begin{longtable}{@{}>{\raggedright\arraybackslash}p{0.15\linewidth}>{\raggedright\arraybackslash}p{0.40\linewidth}>{\raggedleft\arraybackslash}p{0.14\linewidth}>{\raggedright\arraybackslash}p{0.24\linewidth}@{}}
\toprule
Block & Parameter & Estimate & Interval \\
\midrule\endhead
Reverse path components & $\alpha_A$: standardized slope, PWV $\to$ GGT (reverse $A \to M$) & $0.09561$ & $[-0.08353$, $0.2747]$ \\
 & $\beta_M$: standardized slope, GGT $\to$ gynoid fat mass, adjusted for PWV (reverse $M \to Y \mid A$) & $-0.1229$ & $[-0.2265$, $-0.01928]$ \\
 & $\beta_A$: standardized slope, PWV $\to$ gynoid fat mass, adjusted for GGT (reverse $A \to Y \mid M$) & $0.06048$ & $[-0.09303$, $0.214]$ \\
 & $\gamma_A$: adjusted total standardized slope, PWV $\to$ gynoid fat mass (reverse $A \to Y$) & $0.04873$ & $[-0.1067$, $0.2042]$ \\
Reverse path index & $\theta = \alpha_A\cdot\beta_M$: reverse path-product association index & $-0.01175$ & $[-0.03721$, $0.01372]$ \\
\bottomrule
\end{longtable}}

{\scriptsize\setlength{\tabcolsep}{3pt}
\begin{longtable}{@{}lcc@{}}
\toprule
Partial $R^2$ of U (regressor; outcome) & $\alpha_A$ & $\beta_M$ \\
\midrule\endhead
Coefficient & $0.09561$ & $-0.1229$ \\
0.05; 0.05 & $[0.03997, 0.1512]$ & $[-0.1642, -0.08154]$ \\
0.10; 0.10 & $[-0.01871, 0.2099]$ & $[-0.2079, -0.03791]$ \\
0.20; 0.20 & $[-0.1469, 0.3381]$ & $[-0.3032, 0.05737]$ \\
0.30; 0.30 & $[-0.2933, 0.4845]$ & $[-0.412, 0.1662]$ \\
\bottomrule
\end{longtable}}

\paragraph{17.\ Reverse serial path: PWV $\to$ diastolic blood pressure $\to$ GGT $\to$ gynoid fat mass}
\textit{Design.} Seated blood pressure measured 0--365 days after PWV, GGT within 90 days before or after seated blood pressure (in either order), and gynoid fat mass 0--365 days after GGT (all bounds inclusive, same day allowed); one chain starting from each PWV measurement day, with multiple chains allowed per participant, $n = 838$. Adjustment for approximate age on the PWV measurement day, recorded sex and recruitment cohort. Diastolic blood pressure and PWV derived as in the primary serial path; the four path variables standardized within the analysis sample.

\textit{Model.} $M_2 = \alpha_1 + a_{r1} Y + \gamma_1^\top C + \varepsilon_1$; $M_1 = \alpha_2 + a_{r2} Y + d_{r21} M_2 + \gamma_2^\top C + \varepsilon_2$; $A = \alpha_3 + c'_{r} Y + b_{r1} M_2 + b_{r2} M_1 + \gamma_3^\top C + \varepsilon_3$ (linear regressions; $Y$ is thigh-to-ankle PWV, $M_2$ seated diastolic blood pressure, $M_1$ GGT and $A$ gynoid fat mass, all standardized values; $C$ age, sex and recruitment cohort). Reported: the reverse serial path product $a_{r1}\cdot d_{r21}\cdot b_{r2}$ (PWV $\to$ diastolic blood pressure $\to$ GGT $\to$ gynoid fat mass).

\textit{Intervals.} 95\% confidence intervals.
{\footnotesize\setlength{\tabcolsep}{3pt}
\begin{longtable}{@{}>{\raggedright\arraybackslash}p{0.15\linewidth}>{\raggedright\arraybackslash}p{0.40\linewidth}>{\raggedleft\arraybackslash}p{0.14\linewidth}>{\raggedright\arraybackslash}p{0.24\linewidth}@{}}
\toprule
Block & Parameter & Estimate & Interval \\
\midrule\endhead
Reverse serial path & Reverse serial path product $a_{r1}\cdot d_{r21}\cdot b_{r2}$ (PWV $\to$ diastolic blood pressure $\to$ GGT $\to$ gynoid fat mass) & $0.003255$ & $[-0.0004637$, $0.006973]$ \\
\bottomrule
\end{longtable}}

\paragraph{18.\ Reverse pairing: earlier PWV and subsequently measured gynoid fat mass}
\textit{Design.} PWV and gynoid fat mass measured in strict date order: each PWV measurement day as an anchor, with the nearest gynoid fat mass 366--1095 days afterwards (bounds inclusive); multiple anchor days allowed per participant. PWV as the mean of the available values of the left and right sides. Adjustment for approximate age (PWV measurement year minus birth year), sex and sub-study; continuous variables kept on the original scale.

\textit{Model.} $A_{\mathrm{later}} = \gamma_0 + \gamma_Y\cdot Y_{\mathrm{earlier}} + \gamma_{\mathrm{age}}\cdot\mathrm{age} + \gamma_{\mathrm{sex}} + \gamma_{\mathrm{study}} + \varepsilon$ (linear regression; $Y_{\mathrm{earlier}}$ is the earlier thigh-to-ankle PWV and $A_{\mathrm{later}}$ the subsequently measured gynoid fat mass, both on the original scale). $\gamma_Y$: difference in later gynoid fat mass per 1 original unit higher earlier PWV, with age, sex and sub-study held fixed.

\textit{Intervals.} 95\% confidence intervals.
{\footnotesize\setlength{\tabcolsep}{3pt}
\begin{longtable}{@{}>{\raggedright\arraybackslash}p{0.14\linewidth}>{\raggedright\arraybackslash}p{0.29\linewidth}>{\raggedleft\arraybackslash}p{0.13\linewidth}>{\raggedright\arraybackslash}p{0.22\linewidth}>{\raggedleft\arraybackslash}p{0.14\linewidth}@{}}
\toprule
Block & Parameter & Estimate & Interval & $P$ \\
\midrule\endhead
Gynoid fat mass 366--1095 days after PWV & $\gamma_Y$: adjusted association coefficient, earlier PWV--subsequent gynoid fat mass & $76.66$ & $[36.56$, $116.8]$ & \mbox{$0.0001792$} \\
\bottomrule
\end{longtable}}

\subsubsection*{Heterogeneity and interaction}

\paragraph{19.\ Age modification: gynoid fat mass $\to$ GGT $\to$ PWV}
\textit{Design.} Gynoid fat mass, GGT and PWV measured in strict date order: GGT 30--365 days after the fat mass measurement and PWV 90--730 days after GGT (bounds inclusive); one measurement chain per participant. Approximate age at the fat mass measurement as a continuous modifier; further adjustment for sex, cohort, and the most recent height on that day or within the preceding 365 days. Fat mass, GGT, PWV, age and height standardized within the analysis sample; sex and cohort as categorical terms.

\textit{Model.} $M = \alpha_0 + \alpha_A\cdot A + \alpha_{A\times\mathrm{age}}\cdot A\cdot\mathrm{age} + \alpha_C^\top C + \varepsilon_M$; $Y = \beta_0 + \beta_A\cdot A + \beta_M\cdot M + \beta_{A\times\mathrm{age}}\cdot A\cdot\mathrm{age} + \beta_{M\times\mathrm{age}}\cdot M\cdot\mathrm{age} + \beta_C^\top C + \varepsilon_Y$; $Y = \gamma_0 + \gamma_A\cdot A + \gamma_{A\times\mathrm{age}}\cdot A\cdot\mathrm{age} + \gamma_C^\top C + \varepsilon_T$ (linear regressions; $A$, $M$ and $Y$ are standardized gynoid fat mass, GGT and PWV, $\mathrm{age}$ is standardized age, and $C$ contains the age main effect, height, sex and cohort). $\alpha_A$, $\beta_M$, $\beta_A$, $\gamma_A$: slopes at mean age ($\mathrm{age} = 0$), with $\theta = \alpha_A\cdot\beta_M$; the four interaction coefficients: change in the corresponding slope per 1 SD higher age; $\theta(g) = (\alpha_A + \alpha_{A\times\mathrm{age}}\cdot h)\cdot(\beta_M + \beta_{M\times\mathrm{age}}\cdot h)$, $h = (g - \mbox{mean age})/\mbox{age SD}$, computed at each integer age $g$ present in the sample.

\textit{Intervals.} 95\% confidence intervals.
{\footnotesize\setlength{\tabcolsep}{3pt}
\begin{longtable}{@{}>{\raggedright\arraybackslash}p{0.15\linewidth}>{\raggedright\arraybackslash}p{0.40\linewidth}>{\raggedleft\arraybackslash}p{0.14\linewidth}>{\raggedright\arraybackslash}p{0.24\linewidth}@{}}
\toprule
Block & Parameter & Estimate & Interval \\
\midrule\endhead
Path components at mean age & $\alpha_A$: standardized slope at mean age, gynoid fat mass $\to$ GGT ($A \to M$) & $0.1436$ & $[-0.2447$, $0.5318]$ \\
 & $\beta_M$: standardized slope at mean age, GGT $\to$ PWV, adjusted for fat mass ($M \to Y \mid A$) & $0.01402$ & $[-0.2856$, $0.3136]$ \\
 & $\beta_A$: standardized slope at mean age, gynoid fat mass $\to$ PWV, adjusted for GGT ($A \to Y \mid M$) & $0.08875$ & $[-0.0909$, $0.2684]$ \\
 & $\gamma_A$: adjusted total standardized slope at mean age, gynoid fat mass $\to$ PWV ($A \to Y$) & $0.1352$ & $[-0.04493$, $0.3153]$ \\
Path index at mean age & $\theta = \alpha_A\cdot\beta_M$: path-product association index at mean age & $0.002012$ & $[-0.03731$, $0.04134]$ \\
Age interaction coefficients & $\alpha_{A\times\mathrm{age}}$: fat mass $\times$ age interaction coefficient in the GGT equation & $-0.1008$ & $[-0.3929$, $0.1913]$ \\
 & $\beta_{A\times\mathrm{age}}$: fat mass $\times$ age interaction coefficient in the PWV equation & $0.1131$ & $[-0.1131$, $0.3394]$ \\
 & $\beta_{M\times\mathrm{age}}$: GGT $\times$ age interaction coefficient in the PWV equation & $0.5728$ & $[0.1698$, $0.9757]$ \\
 & $\gamma_{A\times\mathrm{age}}$: fat mass $\times$ age interaction coefficient in the total-association equation & $-0.04886$ & $[-0.3386$, $0.2409]$ \\
Age-conditional path products & $\theta(g_{1})$: conditional path product at the 1st integer age in the sample, ordered from lowest to highest age (lowest age in the sample) & $-0.2368$ & $[-0.9084$, $0.4348]$ \\
 & $\theta(g_{2})$: conditional path product at the 2nd integer age in the sample, ordered from lowest to highest age & $-0.2065$ & $[-0.7971$, $0.3841]$ \\
 & $\theta(g_{3})$: conditional path product at the 3rd integer age in the sample, ordered from lowest to highest age & $-0.1266$ & $[-0.5064$, $0.2531]$ \\
 & $\theta(g_{4})$: conditional path product at the 4th integer age in the sample, ordered from lowest to highest age & $-0.1036$ & $[-0.4234$, $0.2161]$ \\
 & $\theta(g_{5})$: conditional path product at the 5th integer age in the sample, ordered from lowest to highest age & $-0.08246$ & $[-0.3472$, $0.1822]$ \\
 & $\theta(g_{6})$: conditional path product at the 6th integer age in the sample, ordered from lowest to highest age & $-0.06311$ & $[-0.2774$, $0.1512]$ \\
 & $\theta(g_{7})$: conditional path product at the 7th integer age in the sample, ordered from lowest to highest age & $-0.04558$ & $[-0.2139$, $0.1228]$ \\
 & $\theta(g_{8})$: conditional path product at the 8th integer age in the sample, ordered from lowest to highest age & $-0.02987$ & $[-0.1564$, $0.09668]$ \\
 & $\theta(g_{9})$: conditional path product at the 9th integer age in the sample, ordered from lowest to highest age & $0.01477$ & $[-0.01537$, $0.04492]$ \\
 & $\theta(g_{10})$: conditional path product at the 10th integer age in the sample, ordered from lowest to highest age & $0.02138$ & $[-0.03256$, $0.07532]$ \\
 & $\theta(g_{11})$: conditional path product at the 11th integer age in the sample, ordered from lowest to highest age & $0.02617$ & $[-0.05659$, $0.1089]$ \\
 & $\theta(g_{12})$: conditional path product at the 12th integer age in the sample, ordered from lowest to highest age & $0.02914$ & $[-0.08447$, $0.1428]$ \\
 & $\theta(g_{13})$: conditional path product at the 13th integer age in the sample, ordered from lowest to highest age & $0.03029$ & $[-0.1163$, $0.1768]$ \\
 & $\theta(g_{14})$: conditional path product at the 14th integer age in the sample, ordered from lowest to highest age & $0.02281$ & $[-0.2405$, $0.2861]$ \\
 & $\theta(g_{15})$: conditional path product at the 15th integer age in the sample, ordered from lowest to highest age & $0.01667$ & $[-0.2932$, $0.3266]$ \\
 & $\theta(g_{16})$: conditional path product at the 16th integer age in the sample, ordered from lowest to highest age & $-0.01265$ & $[-0.4909$, $0.4656]$ \\
 & $\theta(g_{17})$: conditional path product at the 17th integer age in the sample, ordered from lowest to highest age (highest age in the sample) & $-0.04131$ & $[-0.6578$, $0.5752]$ \\
\bottomrule
\end{longtable}}

\paragraph{20.\ Slopes by sex: gynoid fat mass--thigh-to-ankle PWV}
\textit{Design.} Measurement pairing as in the primary gynoid fat mass--PWV association: each gynoid fat measurement day as an anchor, with the nearest PWV within 180 days before or after (bounds inclusive); multiple anchor days allowed per participant. Sex as both a main effect and a modifier, with further adjustment for approximate age (anchor measurement year minus birth year) and sub-study; continuous variables kept on the original scale. First sex category $n = 4304$; second sex category $n = 4145$.

\textit{Model.} $Y = \theta_0 + \theta_A\cdot A + \sum_k \theta_{A\times k}\cdot A\cdot \mathbf{1}[\mathrm{sex} = k] + \theta_{\mathrm{age}}\cdot\mathrm{age} + \gamma_{\mathrm{sex}} + \gamma_{\mathrm{study}} + \varepsilon$ (linear regression; $A$ is gynoid fat mass and $Y$ thigh-to-ankle PWV, both on the original scale). The slope in each sex category is the sum of $\theta_A$ and that category's interaction term, that is, the difference in PWV per 1 original unit higher gynoid fat mass within that category, with age and sub-study held fixed.

\textit{Intervals.} 95\% confidence intervals.
{\footnotesize\setlength{\tabcolsep}{3pt}
\begin{longtable}{@{}>{\raggedright\arraybackslash}p{0.15\linewidth}>{\raggedright\arraybackslash}p{0.40\linewidth}>{\raggedleft\arraybackslash}p{0.14\linewidth}>{\raggedright\arraybackslash}p{0.24\linewidth}@{}}
\toprule
Block & Parameter & Estimate & Interval \\
\midrule\endhead
First sex category & Gynoid fat mass--PWV slope in the first sex category & $0.0001622$ & $[0.0001283$, $0.0001962]$ \\
Second sex category & Gynoid fat mass--PWV slope in the second sex category & $0.0002295$ & $[0.0001858$, $0.0002733]$ \\
\bottomrule
\end{longtable}}

\subsubsection*{Sensitivity analyses}

\paragraph{21.\ Narrower windows: gynoid fat mass $\to$ GGT $\to$ thigh-to-ankle PWV}
\textit{Design.} Gynoid fat mass, GGT and PWV measured in strict date order, with windows narrower than in the primary single-mediator path analysis: GGT 90--365 days after the fat mass measurement and PWV 180--730 days after GGT (bounds inclusive); one measurement chain per participant. Adjustment for approximate age at the fat mass measurement, sex, cohort, and the most recent height on that day or within the preceding 365 days. Continuous variables standardized within the analysis sample; sex and cohort as categorical terms.

\textit{Model.} $M = \alpha_0 + \alpha_A\cdot A + \alpha_C^\top C + \varepsilon_M$; $Y = \beta_0 + \beta_A\cdot A + \beta_M\cdot M + \beta_C^\top C + \varepsilon_Y$; $Y = \gamma_0 + \gamma_A\cdot A + \gamma_C^\top C + \varepsilon_T$ (linear regressions; $A$, $M$ and $Y$ are standardized gynoid fat mass, GGT and thigh-to-ankle PWV, and $C$ the covariates). $\alpha_A$: $A \to M$ slope; $\beta_M$: $M \to Y$ slope adjusted for $A$; $\beta_A$: $A \to Y$ slope adjusted for $M$; $\gamma_A$: total $A \to Y$ slope; path product $\theta = \alpha_A\cdot\beta_M$.

\textit{Intervals.} 95\% confidence intervals.
{\footnotesize\setlength{\tabcolsep}{3pt}
\begin{longtable}{@{}>{\raggedright\arraybackslash}p{0.15\linewidth}>{\raggedright\arraybackslash}p{0.40\linewidth}>{\raggedleft\arraybackslash}p{0.14\linewidth}>{\raggedright\arraybackslash}p{0.24\linewidth}@{}}
\toprule
Block & Parameter & Estimate & Interval \\
\midrule\endhead
Path components & $\alpha_A$: standardized slope, gynoid fat mass $\to$ GGT ($A \to M$) & $0.1191$ & $[-0.2892$, $0.5274]$ \\
 & $\beta_M$: standardized slope, GGT $\to$ thigh-to-ankle PWV, adjusted for fat mass ($M \to Y \mid A$) & $0.3196$ & $[-0.2033$, $0.8425]$ \\
 & $\beta_A$: standardized slope, gynoid fat mass $\to$ PWV, adjusted for GGT ($A \to Y \mid M$) & $0.04984$ & $[-0.1741$, $0.2737]$ \\
 & $\gamma_A$: adjusted total standardized slope, gynoid fat mass $\to$ PWV ($A \to Y$) & $0.08791$ & $[-0.1296$, $0.3055]$ \\
Path index & $\theta = \alpha_A\cdot\beta_M$: path-product association index & $0.03807$ & $[-0.0521$, $0.1282]$ \\
\bottomrule
\end{longtable}}

\paragraph{22.\ Narrow-window serial path: gynoid fat mass $\to$ GGT $\to$ diastolic blood pressure $\to$ PWV}
\textit{Design.} GGT measured 0--180 days after gynoid fat mass, seated blood pressure within 30 days before or after GGT (in either order), and PWV 0--180 days after blood pressure (all bounds inclusive, same day allowed); one chain starting from each gynoid fat mass measurement day, with multiple chains allowed per participant. All four equations share the same sample, with adjustment for approximate age on that measurement day, recorded sex and recruitment cohort. Diastolic blood pressure as the mean of the available values of two seated readings; PWV as the mean of the available sides. The four path variables standardized within the analysis sample.

\textit{Model.} Linear regressions, with equations and reported quantities as in the primary serial path: $M_1 = \alpha_1 + a_1 A + \gamma_1^\top C$; $M_2 = \alpha_2 + a_2 A + d_{21} M_1 + \gamma_2^\top C$; $Y = \alpha_3 + c' A + b_1 M_1 + b_2 M_2 + \gamma_3^\top C$; $Y = \alpha_4 + c A + \gamma_4^\top C$ ($A$ gynoid fat mass, $M_1$ GGT, $M_2$ seated diastolic blood pressure, $Y$ thigh-to-ankle PWV, all standardized values; $C$ age, sex and recruitment cohort). Reported: $a_1 d_{21} b_2$, $a_1 b_1$, $a_2 b_2$, the sum of the three, $c'$, $c$ and seven standardized coefficients.

\textit{Intervals.} 95\% confidence intervals.
{\footnotesize\setlength{\tabcolsep}{3pt}
\begin{longtable}{@{}>{\raggedright\arraybackslash}p{0.15\linewidth}>{\raggedright\arraybackslash}p{0.40\linewidth}>{\raggedleft\arraybackslash}p{0.14\linewidth}>{\raggedright\arraybackslash}p{0.24\linewidth}@{}}
\toprule
Block & Parameter & Estimate & Interval \\
\midrule\endhead
Path decomposition (narrow window) & Serial path product $a_1\cdot d_{21}\cdot b_2$ (through GGT and diastolic blood pressure) & $-0.002383$ & $[-0.01872$, $0.01396]$ \\
 & Path product through GGT only, $a_1\cdot b_1$ & $-0.009157$ & $[-0.05363$, $0.03531]$ \\
 & Path product through diastolic blood pressure only, $a_2\cdot b_2$ & $0.1645$ & $[0.01235$, $0.3166]$ \\
 & Sum of the three path products, $a_1 d_{21} b_2 + a_1 b_1 + a_2 b_2$ & $0.1529$ & $[-0.01586$, $0.3217]$ \\
 & Conditional slope $c'$ (gynoid fat mass $\to$ PWV, controlling for both mediators) & $-0.2234$ & $[-0.4547$, $0.008018]$ \\
 & Total association slope $c$ (gynoid fat mass $\to$ PWV) & $-0.07044$ & $[-0.3268$, $0.1859]$ \\
Component coefficients (narrow window) & Standardized slope $a_1$ (gynoid fat mass $\to$ GGT) & $-0.0515$ & $[-0.3293$, $0.2263]$ \\
 & Standardized slope $d_{21}$ (GGT $\to$ diastolic blood pressure, controlling for gynoid fat mass) & $0.08141$ & $[-0.1975$, $0.3603]$ \\
 & Standardized slope $a_2$ (gynoid fat mass $\to$ diastolic blood pressure, controlling for GGT) & $0.2894$ & $[0.05068$, $0.5281]$ \\
 & Standardized slope $b_2$ (diastolic blood pressure $\to$ PWV, controlling for gynoid fat mass and GGT) & $0.5683$ & $[0.4113$, $0.7252]$ \\
 & Standardized slope $b_1$ (GGT $\to$ PWV, controlling for gynoid fat mass and diastolic blood pressure) & $0.1778$ & $[-0.05693$, $0.4125]$ \\
 & Conditional slope $c'$ (gynoid fat mass $\to$ PWV, controlling for both mediators) & $-0.2234$ & $[-0.4547$, $0.008018]$ \\
 & Total association slope $c$ (gynoid fat mass $\to$ PWV) & $-0.07044$ & $[-0.3268$, $0.1859]$ \\
\bottomrule
\end{longtable}}

\paragraph{23.\ Additional adjustment for BMI, smoking and alcohol consumption: gynoid fat mass serial path}
\textit{Design.} GGT measured 0--365 days after gynoid fat mass, seated blood pressure within 90 days before or after GGT, and PWV 0--365 days after blood pressure (all bounds inclusive, same day allowed). BMI, smoking status and alcohol consumption frequency taken as the most recent record on or within the 365 days before the gynoid fat mass measurement day; only chains with all three available included, $n = 22$. Extended model adding these three to age, sex and cohort; minimal model without them; both on the same sample. Path variables and BMI standardized within the sample.

\textit{Model.} Two sets of four-equation linear regressions, of the same form as the primary serial path: $M_1 = \alpha_1 + a_1 A + \gamma_1^\top C$; $M_2 = \alpha_2 + a_2 A + d_{21} M_1 + \gamma_2^\top C$; $Y = \alpha_3 + c' A + b_1 M_1 + b_2 M_2 + \gamma_3^\top C$; $Y = \alpha_4 + c A + \gamma_4^\top C$. In the extended model $C$ contains age, sex, cohort, BMI, smoking status and alcohol consumption frequency (the last two as categorical terms); in the minimal model $C$ contains only age, sex and cohort. Reported for each model: $a_1 d_{21} b_2$, $a_1 b_1$, $a_2 b_2$, the sum of the three, $c'$, $c$ and seven standardized coefficients, plus the extended-minus-minimal difference for each quantity.

\textit{Intervals.} 95\% confidence intervals.
{\footnotesize\setlength{\tabcolsep}{3pt}
\begin{longtable}{@{}>{\raggedright\arraybackslash}p{0.15\linewidth}>{\raggedright\arraybackslash}p{0.40\linewidth}>{\raggedleft\arraybackslash}p{0.14\linewidth}>{\raggedright\arraybackslash}p{0.24\linewidth}@{}}
\toprule
Block & Parameter & Estimate & Interval \\
\midrule\endhead
Extended adjustment model: path decomposition & Serial path product $a_1\cdot d_{21}\cdot b_2$ (through GGT and diastolic blood pressure) & $0.125$ & $[-0.03923$, $0.2892]$ \\
 & Path product through GGT only, $a_1\cdot b_1$ & $-0.2854$ & $[-0.5342$, $-0.03669]$ \\
 & Path product through diastolic blood pressure only, $a_2\cdot b_2$ & $-0.2215$ & $[-0.5686$, $0.1255]$ \\
 & Sum of the three path products, $a_1 d_{21} b_2 + a_1 b_1 + a_2 b_2$ & $-0.382$ & $[-0.7171$, $-0.04683]$ \\
 & Conditional slope $c'$ (gynoid fat mass $\to$ PWV, controlling for both mediators) & $0.3325$ & $[0.008261$, $0.6567]$ \\
 & Total association slope $c$ (gynoid fat mass $\to$ PWV) & $-0.04948$ & $[-0.3606$, $0.2617]$ \\
Extended adjustment model: component coefficients & Standardized slope $a_1$ (gynoid fat mass $\to$ GGT) & $-0.658$ & $[-1.131$, $-0.1855]$ \\
 & Standardized slope $d_{21}$ (GGT $\to$ diastolic blood pressure, controlling for gynoid fat mass) & $-0.3228$ & $[-0.6705$, $0.02485]$ \\
 & Standardized slope $a_2$ (gynoid fat mass $\to$ diastolic blood pressure, controlling for GGT) & $-0.3764$ & $[-0.9407$, $0.1878]$ \\
 & Standardized slope $b_2$ (diastolic blood pressure $\to$ PWV, controlling for gynoid fat mass and GGT) & $0.5885$ & $[0.3149$, $0.8622]$ \\
 & Standardized slope $b_1$ (GGT $\to$ PWV, controlling for gynoid fat mass and diastolic blood pressure) & $0.4338$ & $[0.1365$, $0.7311]$ \\
 & Conditional slope $c'$ (gynoid fat mass $\to$ PWV, controlling for both mediators) & $0.3325$ & $[0.008261$, $0.6567]$ \\
 & Total association slope $c$ (gynoid fat mass $\to$ PWV) & $-0.04948$ & $[-0.3606$, $0.2617]$ \\
Minimal adjustment model (same sample): path decomposition & Serial path product $a_1\cdot d_{21}\cdot b_2$ (through GGT and diastolic blood pressure) & $-0.029$ & $[-0.1246$, $0.06658]$ \\
 & Path product through GGT only, $a_1\cdot b_1$ & $-0.04195$ & $[-0.1902$, $0.1063]$ \\
 & Path product through diastolic blood pressure only, $a_2\cdot b_2$ & $0.1295$ & $[-0.1089$, $0.3678]$ \\
 & Sum of the three path products, $a_1 d_{21} b_2 + a_1 b_1 + a_2 b_2$ & $0.05851$ & $[-0.2186$, $0.3356]$ \\
 & Conditional slope $c'$ (gynoid fat mass $\to$ PWV, controlling for both mediators) & $-0.1871$ & $[-0.4985$, $0.1242]$ \\
 & Total association slope $c$ (gynoid fat mass $\to$ PWV) & $-0.1286$ & $[-0.3687$, $0.1115]$ \\
Minimal adjustment model (same sample): component coefficients & Standardized slope $a_1$ (gynoid fat mass $\to$ GGT) & $-0.1977$ & $[-0.791$, $0.3957]$ \\
 & Standardized slope $d_{21}$ (GGT $\to$ diastolic blood pressure, controlling for gynoid fat mass) & $0.3302$ & $[-0.0885$, $0.749]$ \\
 & Standardized slope $a_2$ (gynoid fat mass $\to$ diastolic blood pressure, controlling for GGT) & $0.2914$ & $[-0.1422$, $0.7251]$ \\
 & Standardized slope $b_2$ (diastolic blood pressure $\to$ PWV, controlling for gynoid fat mass and GGT) & $0.4443$ & $[0.07607$, $0.8125]$ \\
 & Standardized slope $b_1$ (GGT $\to$ PWV, controlling for gynoid fat mass and diastolic blood pressure) & $0.2122$ & $[-0.08648$, $0.511]$ \\
 & Conditional slope $c'$ (gynoid fat mass $\to$ PWV, controlling for both mediators) & $-0.1871$ & $[-0.4985$, $0.1242]$ \\
 & Total association slope $c$ (gynoid fat mass $\to$ PWV) & $-0.1286$ & $[-0.3687$, $0.1115]$ \\
Extended minus minimal adjustment: path decomposition & Serial path product $a_1\cdot d_{21}\cdot b_2$ (through GGT and diastolic blood pressure): extended minus minimal & $0.154$ & $[-0.0639$, $0.3719]$ \\
 & Path product through GGT only, $a_1\cdot b_1$: extended minus minimal & $-0.2435$ & $[-0.436$, $-0.05095]$ \\
 & Path product through diastolic blood pressure only, $a_2\cdot b_2$: extended minus minimal & $-0.351$ & $[-0.6609$, $-0.04114]$ \\
 & Sum of the three path products, $a_1 d_{21} b_2 + a_1 b_1 + a_2 b_2$: extended minus minimal & $-0.4405$ & $[-0.8214$, $-0.05957]$ \\
 & Conditional slope $c'$ (gynoid fat mass $\to$ PWV, controlling for both mediators): extended minus minimal & $0.5196$ & $[0.2603$, $0.7789]$ \\
 & Total association slope $c$ (gynoid fat mass $\to$ PWV): extended minus minimal & $0.07915$ & $[-0.2754$, $0.4337]$ \\
Extended minus minimal adjustment: component coefficients & Standardized slope $a_1$ (gynoid fat mass $\to$ GGT): extended minus minimal & $-0.4603$ & $[-0.9136$, $-0.00706]$ \\
 & Standardized slope $d_{21}$ (GGT $\to$ diastolic blood pressure, controlling for gynoid fat mass): extended minus minimal & $-0.6531$ & $[-1.064$, $-0.2419]$ \\
 & Standardized slope $a_2$ (gynoid fat mass $\to$ diastolic blood pressure, controlling for GGT): extended minus minimal & $-0.6678$ & $[-1.184$, $-0.1515]$ \\
 & Standardized slope $b_2$ (diastolic blood pressure $\to$ PWV, controlling for gynoid fat mass and GGT): extended minus minimal & $0.1443$ & $[-0.1898$, $0.4783]$ \\
 & Standardized slope $b_1$ (GGT $\to$ PWV, controlling for gynoid fat mass and diastolic blood pressure): extended minus minimal & $0.2216$ & $[-0.05595$, $0.4991]$ \\
 & Conditional slope $c'$ (gynoid fat mass $\to$ PWV, controlling for both mediators): extended minus minimal & $0.5196$ & $[0.2603$, $0.7789]$ \\
 & Total association slope $c$ (gynoid fat mass $\to$ PWV): extended minus minimal & $0.07915$ & $[-0.2754$, $0.4337]$ \\
\bottomrule
\end{longtable}}

\paragraph{24.\ Joint model: gynoid fat mass, GGT, diastolic blood pressure, PWV ($\pm$90 days)}
\textit{Design.} Each gynoid fat measurement day as an anchor, with one measurement each of GGT, seated diastolic blood pressure and PWV, all four measurements pairwise no more than 90 days apart (in either order, bounds inclusive), taking the set with the shortest total span; multiple anchor days allowed per participant, $n = 905$. Diastolic blood pressure and PWV values and covariates as in the $\pm$180-day joint model; continuous variables kept on the original scale.

\textit{Model.} $Y = \beta_0 + \beta_A\cdot A + \beta_{M1}\cdot M1 + \beta_{M2}\cdot M2 + \beta_{\mathrm{age}}\cdot\mathrm{age} + \gamma_{\mathrm{sex}} + \gamma_{\mathrm{study}} + \varepsilon$ (linear regression, as in the $\pm$180-day joint model; $A$ is gynoid fat mass, $M1$ GGT, $M2$ seated diastolic blood pressure and $Y$ thigh-to-ankle PWV, all on the original scale). $\beta_A$, $\beta_{M1}$, $\beta_{M2}$: difference in PWV per 1 original unit higher value of the respective variable, with the other terms of the same model held fixed.

\textit{Intervals.} 95\% confidence intervals; $P$ values not adjusted for multiple comparisons.
{\footnotesize\setlength{\tabcolsep}{3pt}
\begin{longtable}{@{}>{\raggedright\arraybackslash}p{0.14\linewidth}>{\raggedright\arraybackslash}p{0.29\linewidth}>{\raggedleft\arraybackslash}p{0.13\linewidth}>{\raggedright\arraybackslash}p{0.22\linewidth}>{\raggedleft\arraybackslash}p{0.14\linewidth}@{}}
\toprule
Block & Parameter & Estimate & Interval & $P$ \\
\midrule\endhead
Mutually adjusted model ($\pm$90 days) & $\beta_A$: gynoid fat mass coefficient (mutually adjusted model) & $7.933\times10^{-5}$ & $[7.876\times10^{-7}$, $0.0001579]$ & \mbox{$0.04775$} \\
 & $\beta_{M1}$: GGT coefficient (mutually adjusted model) & $-0.0007563$ & $[-0.004707$, $0.003195]$ & \mbox{$0.7075$} \\
 & $\beta_{M2}$: seated diastolic blood pressure coefficient (mutually adjusted model) & $0.07522$ & $[0.06418$, $0.08626]$ & \mbox{$1.108\times10^{-40}$} \\
\bottomrule
\end{longtable}}

\paragraph{25.\ Joint model with additional adjustment for lifestyle, education and height ($\pm$180 days)}
\textit{Design.} Measurement sets and variable values as in the $\pm$180-day joint model. Beyond approximate age, sex and sub-study, additional adjustment for three categorical variables, namely smoking status and alcohol consumption frequency (each the nearest record within 365 days before or after the anchor day) and education level (the nearest record within 3650 days before or after), and for the nearest height within 365 days before or after the anchor day; continuous variables kept on the original scale.

\textit{Model.} $Y = \beta_0 + \beta_A\cdot A + \beta_{M1}\cdot M1 + \beta_{M2}\cdot M2 + \beta_{\mathrm{age}}\cdot\mathrm{age} + \delta\cdot\mathrm{height} + \gamma_{\mathrm{sex}} + \gamma_{\mathrm{study}} + \gamma_{\mathrm{smk}} + \gamma_{\mathrm{alc}} + \gamma_{\mathrm{edu}} + \varepsilon$ (linear regression; $A$ is gynoid fat mass, $M1$ GGT, $M2$ seated diastolic blood pressure and $Y$ thigh-to-ankle PWV, all on the original scale). $\beta_A$, $\beta_{M1}$, $\beta_{M2}$: difference in PWV per 1 original unit higher value of the respective variable, with the other terms of the same model held fixed.

\textit{Intervals.} 95\% confidence intervals; $P$ values not adjusted for multiple comparisons.
{\footnotesize\setlength{\tabcolsep}{3pt}
\begin{longtable}{@{}>{\raggedright\arraybackslash}p{0.14\linewidth}>{\raggedright\arraybackslash}p{0.29\linewidth}>{\raggedleft\arraybackslash}p{0.13\linewidth}>{\raggedright\arraybackslash}p{0.22\linewidth}>{\raggedleft\arraybackslash}p{0.14\linewidth}@{}}
\toprule
Block & Parameter & Estimate & Interval & $P$ \\
\midrule\endhead
Mutually adjusted model + extended adjustment ($\pm$180 days) & $\beta_A$: gynoid fat mass coefficient (extended adjustment model) & $6.274\times10^{-5}$ & $[-2.533\times10^{-5}$, $0.0001508]$ & \mbox{$0.1627$} \\
 & $\beta_{M1}$: GGT coefficient (extended adjustment model) & $0.0001896$ & $[-0.00567$, $0.006049]$ & \mbox{$0.9494$} \\
 & $\beta_{M2}$: seated diastolic blood pressure coefficient (extended adjustment model) & $0.06999$ & $[0.05696$, $0.08301]$ & \mbox{$6.091\times10^{-26}$} \\
\bottomrule
\end{longtable}}

\FloatBarrier

\subsection{Waist circumference--TG--DBP--PWV}
\label{app:rq-designs:72d2}

\paragraph{Research question and measures.}
The research question concerns the association of waist circumference with lower-limb arterial stiffness via triglycerides and diastolic blood pressure, with data from the HPP cohort. The exposure is waist circumference; the mediators are blood triglycerides and seated diastolic blood pressure; the outcome is thigh-to-ankle pulse wave velocity (PWV). Unless stated otherwise, coefficients are unstandardized slopes on the original recorded scale of each variable, i.e., the change in the response variable, in recorded units, per 1 recorded unit higher value of the predictor.

\subsubsection*{Primary pathways}

\paragraph{1.\ Single-mediator path: waist circumference $\to$ triglycerides $\to$ thigh-to-ankle PWV}
\textit{Design.} Waist circumference, triglycerides and PWV measured in strict date order: triglycerides 30--365 days after waist circumference, PWV 90--1095 days after triglycerides (endpoints inclusive); one measurement chain per participant. Adjustment for approximate age at the waist circumference measurement and sex. PWV taken as the mean of the available left and right values. All variables entered as their original recorded values, with sex as a categorical term.

\textit{Model.} $M = \alpha_0 + \alpha_A \cdot A + \alpha_C^{\top} \cdot C + \varepsilon_M$; $Y = \beta_0 + \beta_A \cdot A + \beta_M \cdot M + \beta_C^{\top} \cdot C + \varepsilon_Y$; $Y = \gamma_0 + \gamma_A \cdot A + \gamma_C^{\top} \cdot C + \varepsilon_T$. All three equations are linear regressions; $A$, $M$ and $Y$ are waist circumference, triglycerides and thigh-to-ankle PWV, respectively, and $C$ is approximate age and sex. $\alpha_A$ is the change in triglycerides per 1 recorded unit higher waist circumference; $\beta_M$ and $\beta_A$ are the mutually adjusted slopes of PWV on triglycerides and on waist circumference, respectively; $\gamma_A$ is the total slope of PWV on waist circumference with adjustment for $C$ only; path product $\theta = \alpha_A \cdot \beta_M$.

\textit{Intervals.} $\alpha_A$, $\beta_M$, $\beta_A$, $\gamma_A$: bootstrap 95\% percentile intervals; $\theta$: bootstrap 97.5\% percentile interval.
{\footnotesize\setlength{\tabcolsep}{3pt}
\begin{longtable}{@{}>{\raggedright\arraybackslash}p{0.15\linewidth}>{\raggedright\arraybackslash}p{0.40\linewidth}>{\raggedleft\arraybackslash}p{0.14\linewidth}>{\raggedright\arraybackslash}p{0.24\linewidth}@{}}
\toprule
Block & Parameter & Estimate & Interval \\
\midrule\endhead
Path index & $\theta = \alpha_A \cdot \beta_M$: path-product association index & $0.003205$ & $[-0.003298$, $0.01037]$ \\
Path components & $\alpha_A$: waist circumference $\to$ triglycerides slope ($A \to M$) & $1.738$ & $[1.291$, $2.224]$ \\
 & $\beta_M$: triglycerides $\to$ thigh-to-ankle PWV slope, adjusted for waist circumference ($M \to Y \mid A$) & $0.001844$ & $[-0.001566$, $0.005293]$ \\
 & $\beta_A$: waist circumference $\to$ PWV slope, adjusted for triglycerides ($A \to Y \mid M$) & $0.02077$ & $[0.006894$, $0.03415]$ \\
 & $\gamma_A$: adjusted total waist circumference $\to$ PWV slope ($A \to Y$) & $0.02397$ & $[0.01147$, $0.03589]$ \\
\bottomrule
\end{longtable}}

\paragraph{2.\ Single-mediator path and bias bounds: waist circumference $\to$ seated diastolic blood pressure $\to$ PWV}
\textit{Design.} Waist circumference, seated blood pressure and PWV measured in strict date order: blood pressure 30--730 days after waist circumference, PWV 90--1095 days after blood pressure (endpoints inclusive); one measurement chain per participant. Adjustment for approximate age at the waist circumference measurement and sex. Diastolic blood pressure taken as the mean of the available values of the two seated readings, PWV as the mean of the available left and right values. All variables entered as their original recorded values, with sex as a categorical term.

\textit{Model.} $M = \alpha_0 + \alpha_A \cdot A + \alpha_C^{\top} \cdot C + \varepsilon_M$; $Y = \beta_0 + \beta_A \cdot A + \beta_M \cdot M + \beta_C^{\top} \cdot C + \varepsilon_Y$; $Y = \gamma_0 + \gamma_A \cdot A + \gamma_C^{\top} \cdot C + \varepsilon_T$. All three equations are linear regressions; $A$, $M$ and $Y$ are waist circumference, seated diastolic blood pressure and thigh-to-ankle PWV, respectively, and $C$ is approximate age and sex. $\alpha_A$ is the change in diastolic blood pressure per 1 recorded unit higher waist circumference; $\beta_M$ is the slope of PWV on diastolic blood pressure adjusted for waist circumference; $\gamma_A$ is the total slope of PWV on waist circumference with adjustment for $C$ only; $\theta = \alpha_A \cdot \beta_M$. Bias bounds are $\hat{\beta} \pm \mathrm{se} \cdot \sqrt{\nu \cdot R^2_{Y\sim U} \cdot R^2_{D\sim U} / (1 - R^2_{D\sim U})}$, $\hat{\beta} \in \{\alpha_A, \beta_M\}$, where $\mathrm{se}$ is the conventional standard error of the coefficient, $\nu$ the residual degrees of freedom, and $R^2_{D\sim U}$ and $R^2_{Y\sim U}$ the partial $R^2$ of a hypothetical omitted covariate $U$ with the regressor of that coefficient and with the outcome of the equation, each taking 0.01, 0.05, 0.10 and 0.20; the bounds for $\theta$ are the minimum and maximum of the four products of the $\alpha_A$ and $\beta_M$ bound endpoints at the same grid point.

\textit{Intervals.} $\alpha_A$, $\gamma_A$: bootstrap 95\% percentile intervals. Bias-bound grid: each row corresponds to one pair $(R^2_{D\sim U}, R^2_{Y\sim U})$, with lower and upper limits equal to the coefficient $\pm$ the bias bound; the lower and upper limits for $\theta$ are the minimum and maximum of the products of the $\alpha_A$ and $\beta_M$ bound endpoints.
{\footnotesize\setlength{\tabcolsep}{3pt}
\begin{longtable}{@{}>{\raggedright\arraybackslash}p{0.15\linewidth}>{\raggedright\arraybackslash}p{0.40\linewidth}>{\raggedleft\arraybackslash}p{0.14\linewidth}>{\raggedright\arraybackslash}p{0.24\linewidth}@{}}
\toprule
Block & Parameter & Estimate & Interval \\
\midrule\endhead
Path components & $\alpha_A$: waist circumference $\to$ seated diastolic blood pressure slope ($A \to M$) & $0.2641$ & $[-6.006$, $1.952]$ \\
 & $\gamma_A$: adjusted total waist circumference $\to$ PWV slope ($A \to Y$) & $0.1108$ & $[-0.7152$, $0.3966]$ \\
\bottomrule
\end{longtable}}

{\scriptsize\setlength{\tabcolsep}{3pt}
\begin{longtable}{@{}lccc@{}}
\toprule
Partial $R^2$ of U (regressor; outcome) & $\alpha_A$ & $\beta_M$ & $\theta$ \\
\midrule\endhead
Coefficient & $0.2641$ & $0.08402$ & $0.02219$ \\
0.01; 0.01 & $[0.2516, 0.2766]$ & $[0.08295, 0.08509]$ & $[0.02087, 0.02353]$ \\
0.01; 0.05 & $[0.2362, 0.292]$ & $[0.08162, 0.08641]$ & $[0.01928, 0.02523]$ \\
0.01; 0.10 & $[0.2247, 0.3036]$ & $[0.08063, 0.08741]$ & $[0.01811, 0.02653]$ \\
0.01; 0.20 & $[0.2083, 0.3199]$ & $[0.07922, 0.08881]$ & $[0.0165, 0.02841]$ \\
0.05; 0.01 & $[0.2356, 0.2926]$ & $[0.08157, 0.08646]$ & $[0.01922, 0.0253]$ \\
0.05; 0.05 & $[0.2004, 0.3278]$ & $[0.07854, 0.08949]$ & $[0.01574, 0.02933]$ \\
0.05; 0.10 & $[0.174, 0.3542]$ & $[0.07628, 0.09176]$ & $[0.01328, 0.0325]$ \\
0.05; 0.20 & $[0.1367, 0.3915]$ & $[0.07307, 0.09496]$ & $[0.009992, 0.03718]$ \\
0.10; 0.01 & $[0.2227, 0.3055]$ & $[0.08046, 0.08757]$ & $[0.01792, 0.02675]$ \\
0.10; 0.05 & $[0.1716, 0.3566]$ & $[0.07607, 0.09197]$ & $[0.01305, 0.0328]$ \\
0.10; 0.10 & $[0.1333, 0.395]$ & $[0.07277, 0.09526]$ & $[0.009697, 0.03763]$ \\
0.10; 0.20 & $[0.07905, 0.4492]$ & $[0.06811, 0.09992]$ & $[0.005385, 0.04488]$ \\
0.20; 0.01 & $[0.202, 0.3262]$ & $[0.07868, 0.08935]$ & $[0.0159, 0.02914]$ \\
0.20; 0.05 & $[0.1253, 0.4029]$ & $[0.07209, 0.09594]$ & $[0.009034, 0.03866]$ \\
0.20; 0.10 & $[0.06783, 0.4604]$ & $[0.06715, 0.1009]$ & $[0.004555, 0.04645]$ \\
0.20; 0.20 & $[-0.01347, 0.5417]$ & $[0.06016, 0.1079]$ & $[-0.001453, 0.05843]$ \\
\bottomrule
\end{longtable}}

\paragraph{3.\ Serial path: waist circumference $\to$ triglycerides $\to$ diastolic blood pressure $\to$ PWV}
\textit{Design.} Waist circumference, triglycerides, seated diastolic blood pressure and thigh-to-ankle PWV measured in the same period: starting from each waist circumference measurement date, triglycerides, diastolic blood pressure and PWV are taken in turn as the measurement closest to the preceding item, with adjacent items no more than 365 days apart (endpoints inclusive, in either order); a participant may contribute multiple chains, $n = 6266$. Adjustment for approximate age at the waist circumference measurement date and recorded sex. Diastolic blood pressure taken as the mean of the available values of the two seated readings, PWV as the mean of the available sides; all variables taken as their original recorded values, unstandardized.

\textit{Model.} $M_1 = \alpha_1 + \beta_{AM_1} A + \gamma_1^{\top} C + \varepsilon_1$; $M_2 = \alpha_2 + \delta_2 A + \beta_{M_1M_2} M_1 + \gamma_2^{\top} C + \varepsilon_2$; $Y = \alpha_3 + \delta_3 A + \delta_4 M_1 + \beta_{M_2Y} M_2 + \gamma_3^{\top} C + \varepsilon_3$. All three equations are linear regressions; $A$, $M_1$, $M_2$ and $Y$ are waist circumference, triglycerides, seated diastolic blood pressure and thigh-to-ankle PWV, respectively (original recorded scale), and $C$ is approximate age and recorded sex. $\beta_{AM_1}$, $\beta_{M_1M_2}$ and $\beta_{M_2Y}$ are the focal slopes of the respective equations (the change in the dependent variable, in recorded units, per 1 recorded unit higher independent variable); serial path product $\theta = \beta_{AM_1} \cdot \beta_{M_1M_2} \cdot \beta_{M_2Y}$.

\textit{Intervals.} Bootstrap 95\% percentile intervals; $P$ values not corrected for multiple comparisons.
{\footnotesize\setlength{\tabcolsep}{3pt}
\begin{longtable}{@{}>{\raggedright\arraybackslash}p{0.14\linewidth}>{\raggedright\arraybackslash}p{0.29\linewidth}>{\raggedleft\arraybackslash}p{0.13\linewidth}>{\raggedright\arraybackslash}p{0.22\linewidth}>{\raggedleft\arraybackslash}p{0.14\linewidth}@{}}
\toprule
Block & Parameter & Estimate & Interval & $P$ \\
\midrule\endhead
Path components & $\beta_{AM_1}$: adjusted waist circumference $\to$ triglycerides slope ($A \to M_1$) & $2.001$ & $[1.85$, $2.146]$ & \mbox{$0$} \\
 & $\beta_{M_1M_2}$: triglycerides $\to$ seated diastolic blood pressure slope, adjusted for waist circumference ($M_1 \to M_2 \mid A$) & $0.02143$ & $[0.01711$, $0.02619]$ & \mbox{$0$} \\
 & $\beta_{M_2Y}$: seated diastolic blood pressure $\to$ PWV slope, adjusted for waist circumference and triglycerides ($M_2 \to Y \mid A, M_1$) & $0.07551$ & $[0.07124$, $0.07996]$ & \mbox{$0$} \\
Path index & $\theta = \beta_{AM_1} \cdot \beta_{M_1M_2} \cdot \beta_{M_2Y}$: serial path product & $0.003237$ & $[0.002525$, $0.003986]$ & \mbox{$0$} \\
\bottomrule
\end{longtable}}

\paragraph{4.\ Exposure--outcome association: waist circumference and thigh-to-ankle PWV ($\pm$180 days)}
\textit{Design.} Waist circumference and PWV measured in the same period: with each waist circumference measurement date as an anchor, the PWV measurement closest in date within 180 days before or after (endpoints inclusive) is taken; a participant may contribute multiple anchor dates, $n = 10898$. PWV taken as the mean of the available sides of the left and right thigh-to-ankle readings. Adjustment for approximate age (anchor-date year minus birth year), sex and height measured on the same day as waist circumference. All variables on their original recorded scale, unstandardized; sex as a categorical term.

\textit{Model.} $Y = \beta_0 + \beta_A \cdot A + \beta_{\mathrm{age}} \cdot \mathrm{age10} + \beta_H \cdot H + \gamma_{\mathrm{sex}} + \varepsilon$, linear regression; $A$ is waist circumference, $Y$ thigh-to-ankle PWV, $H$ height, and $\mathrm{age10} = (\text{approximate age} - 50)/10$. $\beta_A$ is the difference in PWV (PWV original recorded units) per 1 original recorded unit higher waist circumference, with age, sex and height held fixed.

\textit{Intervals.} 95\% confidence interval.
{\footnotesize\setlength{\tabcolsep}{3pt}
\begin{longtable}{@{}>{\raggedright\arraybackslash}p{0.14\linewidth}>{\raggedright\arraybackslash}p{0.29\linewidth}>{\raggedleft\arraybackslash}p{0.13\linewidth}>{\raggedright\arraybackslash}p{0.22\linewidth}>{\raggedleft\arraybackslash}p{0.14\linewidth}@{}}
\toprule
Block & Parameter & Estimate & Interval & $P$ \\
\midrule\endhead
Exposure--outcome association & $\beta_A$: adjusted waist circumference--PWV association coefficient & $0.0272$ & $[0.02439$, $0.03001]$ & \mbox{$1.065\times10^{-78}$} \\
\bottomrule
\end{longtable}}

\subsubsection*{Adjacent segments and pairwise associations}

\paragraph{5.\ Serial segment: waist circumference $\to$ triglycerides $\to$ seated diastolic blood pressure}
\textit{Design.} Waist circumference, triglycerides and seated blood pressure measured in strict date order: triglycerides 30--365 days after waist circumference, blood pressure 30--365 days after triglycerides (endpoints inclusive); one measurement chain per participant. Adjustment for approximate age at the waist circumference measurement and sex. Diastolic blood pressure taken as the mean of the available values of the two seated readings. All variables entered as their original recorded values, with sex as a categorical term.

\textit{Model.} $M = \alpha_0 + \alpha_A \cdot A + \alpha_C^{\top} \cdot C + \varepsilon_M$; $Y = \beta_0 + \beta_A \cdot A + \beta_M \cdot M + \beta_C^{\top} \cdot C + \varepsilon_Y$; $Y = \gamma_0 + \gamma_A \cdot A + \gamma_C^{\top} \cdot C + \varepsilon_T$. All three equations are linear regressions; $A$, $M$ and $Y$ are waist circumference, triglycerides and seated diastolic blood pressure, respectively, and $C$ is approximate age and sex. $\alpha_A$ is the waist circumference $\to$ triglycerides slope, $\beta_M$ the triglycerides $\to$ diastolic blood pressure slope adjusted for waist circumference, $\beta_A$ the waist circumference $\to$ diastolic blood pressure slope adjusted for triglycerides, $\gamma_A$ the total waist circumference $\to$ diastolic blood pressure slope, and segment path index $\theta = \alpha_A \cdot \beta_M$.

\textit{Intervals.} $\alpha_A$, $\beta_M$, $\beta_A$, $\gamma_A$: bootstrap 95\% percentile intervals; $\theta$: bootstrap 97.5\% percentile interval.
{\footnotesize\setlength{\tabcolsep}{3pt}
\begin{longtable}{@{}>{\raggedright\arraybackslash}p{0.15\linewidth}>{\raggedright\arraybackslash}p{0.40\linewidth}>{\raggedleft\arraybackslash}p{0.14\linewidth}>{\raggedright\arraybackslash}p{0.24\linewidth}@{}}
\toprule
Block & Parameter & Estimate & Interval \\
\midrule\endhead
Path index & $\theta = \alpha_A \cdot \beta_M$: segment path-product association index & $-0.01189$ & $[-0.1359$, $0.2203]$ \\
Path components & $\alpha_A$: waist circumference $\to$ triglycerides slope ($A \to M$) & $1.868$ & $[-0.4008$, $5.042]$ \\
 & $\beta_M$: triglycerides $\to$ seated diastolic blood pressure slope, adjusted for waist circumference ($M \to Y \mid A$) & $-0.006368$ & $[-0.06319$, $0.05166]$ \\
 & $\beta_A$: waist circumference $\to$ seated diastolic blood pressure slope, adjusted for triglycerides ($A \to Y \mid M$) & $0.2391$ & $[-0.1192$, $0.4615]$ \\
 & $\gamma_A$: adjusted total waist circumference $\to$ seated diastolic blood pressure slope ($A \to Y$) & $0.2272$ & $[-0.06735$, $0.4516]$ \\
\bottomrule
\end{longtable}}

\paragraph{6.\ Serial segment: triglycerides $\to$ seated diastolic blood pressure $\to$ thigh-to-ankle PWV}
\textit{Design.} Triglycerides, seated blood pressure and PWV measured in strict date order: blood pressure 30--365 days after triglycerides, PWV 90--1095 days after blood pressure (endpoints inclusive); one measurement chain per participant. Adjustment for approximate age at the triglycerides measurement and sex. Diastolic blood pressure taken as the mean of the available values of the two seated readings, PWV as the mean of the available left and right values. All variables entered as their original recorded values, with sex as a categorical term.

\textit{Model.} $M = \alpha_0 + \alpha_A \cdot A + \alpha_C^{\top} \cdot C + \varepsilon_M$; $Y = \beta_0 + \beta_A \cdot A + \beta_M \cdot M + \beta_C^{\top} \cdot C + \varepsilon_Y$; $Y = \gamma_0 + \gamma_A \cdot A + \gamma_C^{\top} \cdot C + \varepsilon_T$. All three equations are linear regressions; $A$, $M$ and $Y$ are triglycerides, seated diastolic blood pressure and thigh-to-ankle PWV, respectively, and $C$ is approximate age and sex. $\alpha_A$ is the triglycerides $\to$ diastolic blood pressure slope, $\beta_M$ the diastolic blood pressure $\to$ PWV slope adjusted for triglycerides, $\beta_A$ the triglycerides $\to$ PWV slope adjusted for diastolic blood pressure, $\gamma_A$ the total triglycerides $\to$ PWV slope, and segment path index $\theta = \alpha_A \cdot \beta_M$.

\textit{Intervals.} $\alpha_A$, $\beta_M$, $\beta_A$, $\gamma_A$: bootstrap 95\% percentile intervals; $\theta$: bootstrap 97.5\% percentile interval.
{\footnotesize\setlength{\tabcolsep}{3pt}
\begin{longtable}{@{}>{\raggedright\arraybackslash}p{0.15\linewidth}>{\raggedright\arraybackslash}p{0.40\linewidth}>{\raggedleft\arraybackslash}p{0.14\linewidth}>{\raggedright\arraybackslash}p{0.24\linewidth}@{}}
\toprule
Block & Parameter & Estimate & Interval \\
\midrule\endhead
Path index & $\theta = \alpha_A \cdot \beta_M$: segment path-product association index & $0.002001$ & $[0.001399$, $0.002696]$ \\
Path components & $\alpha_A$: triglycerides $\to$ seated diastolic blood pressure slope ($A \to M$) & $0.03716$ & $[0.02827$, $0.04686]$ \\
 & $\beta_M$: seated diastolic blood pressure $\to$ thigh-to-ankle PWV slope, adjusted for triglycerides ($M \to Y \mid A$) & $0.05384$ & $[0.04659$, $0.06157]$ \\
 & $\beta_A$: triglycerides $\to$ PWV slope, adjusted for seated diastolic blood pressure ($A \to Y \mid M$) & $0.0006683$ & $[-0.0005066$, $0.001887]$ \\
 & $\gamma_A$: adjusted total triglycerides $\to$ PWV slope ($A \to Y$) & $0.002669$ & $[0.001503$, $0.00399]$ \\
\bottomrule
\end{longtable}}

\paragraph{7.\ Pairwise association: waist circumference--triglycerides ($\pm$90 days)}
\textit{Design.} Waist circumference and triglycerides measured in the same period: with each waist circumference measurement date as an anchor, the triglycerides measurement closest in date within 90 days before or after (endpoints inclusive) is taken; a participant may contribute multiple anchor dates, $n = 4048$. Adjustment for approximate age (anchor-date year minus birth year), sex and height measured on the same day as waist circumference. All variables on their original recorded scale, unstandardized; sex as a categorical term.

\textit{Model.} $M_1 = \beta_0 + \beta_A \cdot A + \beta_{\mathrm{age}} \cdot \mathrm{age10} + \beta_H \cdot H + \gamma_{\mathrm{sex}} + \varepsilon$, linear regression; $A$ is waist circumference, $M_1$ triglycerides, $H$ height, and $\mathrm{age10} = (\text{approximate age} - 50)/10$. $\beta_A$ is the difference in triglycerides (original recorded units) per 1 original recorded unit higher waist circumference, with age, sex and height held fixed.

\textit{Intervals.} 95\% confidence interval.
{\footnotesize\setlength{\tabcolsep}{3pt}
\begin{longtable}{@{}>{\raggedright\arraybackslash}p{0.14\linewidth}>{\raggedright\arraybackslash}p{0.29\linewidth}>{\raggedleft\arraybackslash}p{0.13\linewidth}>{\raggedright\arraybackslash}p{0.22\linewidth}>{\raggedleft\arraybackslash}p{0.14\linewidth}@{}}
\toprule
Block & Parameter & Estimate & Interval & $P$ \\
\midrule\endhead
Pairwise association & $\beta_A$: adjusted waist circumference--triglycerides association coefficient & $2.029$ & $[1.846$, $2.211]$ & \mbox{$2.824\times10^{-99}$} \\
\bottomrule
\end{longtable}}

\paragraph{8.\ Pairwise association: waist circumference--seated diastolic blood pressure ($\pm$90 days)}
\textit{Design.} Waist circumference and seated diastolic blood pressure measured in the same period: with each waist circumference measurement date as an anchor, the diastolic blood pressure measurement closest in date within 90 days before or after (endpoints inclusive) is taken; a participant may contribute multiple anchor dates, $n = 12899$. Diastolic blood pressure taken as the mean of the available values of the two seated readings. Adjustment for approximate age (anchor-date year minus birth year), sex and height measured on the same day as waist circumference. All variables on their original recorded scale, unstandardized; sex as a categorical term.

\textit{Model.} $M_2 = \beta_0 + \beta_A \cdot A + \beta_{\mathrm{age}} \cdot \mathrm{age10} + \beta_H \cdot H + \gamma_{\mathrm{sex}} + \varepsilon$, linear regression; $A$ is waist circumference, $M_2$ seated diastolic blood pressure, $H$ height, and $\mathrm{age10} = (\text{approximate age} - 50)/10$. $\beta_A$ is the difference in diastolic blood pressure (original recorded units) per 1 original recorded unit higher waist circumference, with age, sex and height held fixed.

\textit{Intervals.} 95\% confidence interval.
{\footnotesize\setlength{\tabcolsep}{3pt}
\begin{longtable}{@{}>{\raggedright\arraybackslash}p{0.14\linewidth}>{\raggedright\arraybackslash}p{0.29\linewidth}>{\raggedleft\arraybackslash}p{0.13\linewidth}>{\raggedright\arraybackslash}p{0.22\linewidth}>{\raggedleft\arraybackslash}p{0.14\linewidth}@{}}
\toprule
Block & Parameter & Estimate & Interval & $P$ \\
\midrule\endhead
Pairwise association & $\beta_A$: adjusted waist circumference--seated diastolic blood pressure association coefficient & $0.2874$ & $[0.2714$, $0.3034]$ & \mbox{$1.841\times10^{-256}$} \\
\bottomrule
\end{longtable}}

\paragraph{9.\ Pairwise association: triglycerides--seated diastolic blood pressure ($\pm$30 days)}
\textit{Design.} Triglycerides and seated diastolic blood pressure measured in the same period: with each triglycerides measurement date as an anchor, the diastolic blood pressure measurement closest in date within 30 days before or after (endpoints inclusive) is taken; a participant may contribute multiple anchor dates, $n = 1916$. Diastolic blood pressure taken as the mean of the available values of the two seated readings. Adjustment for approximate age (anchor-date year minus birth year), sex and the height closest in date within 365 days before or after the anchor date. All variables on their original recorded scale, unstandardized; sex as a categorical term.

\textit{Model.} $M_2 = \beta_0 + \beta_{M1} \cdot M_1 + \beta_{\mathrm{age}} \cdot \mathrm{age10} + \beta_H \cdot H + \gamma_{\mathrm{sex}} + \varepsilon$, linear regression; $M_1$ is triglycerides, $M_2$ seated diastolic blood pressure, $H$ height, and $\mathrm{age10} = (\text{approximate age} - 50)/10$. $\beta_{M1}$ is the difference in diastolic blood pressure (original recorded units) per 1 original recorded unit higher triglycerides, with age, sex and height held fixed.

\textit{Intervals.} 95\% confidence interval.
{\footnotesize\setlength{\tabcolsep}{3pt}
\begin{longtable}{@{}>{\raggedright\arraybackslash}p{0.14\linewidth}>{\raggedright\arraybackslash}p{0.29\linewidth}>{\raggedleft\arraybackslash}p{0.13\linewidth}>{\raggedright\arraybackslash}p{0.22\linewidth}>{\raggedleft\arraybackslash}p{0.14\linewidth}@{}}
\toprule
Block & Parameter & Estimate & Interval & $P$ \\
\midrule\endhead
Pairwise association & $\beta_{M1}$: adjusted triglycerides--seated diastolic blood pressure association coefficient & $0.03667$ & $[0.02809$, $0.04525]$ & \mbox{$1.02\times10^{-16}$} \\
\bottomrule
\end{longtable}}

\paragraph{10.\ Pairwise association: triglycerides--PWV ($\pm$90 days)}
\textit{Design.} Triglycerides and PWV measured in the same period: with each triglycerides measurement date as an anchor, the PWV measurement closest in date within 90 days before or after (endpoints inclusive) is taken; a participant may contribute multiple anchor dates, $n = 3296$. PWV taken as the mean of the available sides. Adjustment for approximate age (anchor-date year minus birth year), sex and the height closest in date within 365 days before or after the anchor date. All variables on their original recorded scale, unstandardized; sex as a categorical term.

\textit{Model.} $Y = \beta_0 + \beta_{M1} \cdot M_1 + \beta_{\mathrm{age}} \cdot \mathrm{age10} + \beta_H \cdot H + \gamma_{\mathrm{sex}} + \varepsilon$, linear regression; $M_1$ is triglycerides, $Y$ thigh-to-ankle PWV, $H$ height, and $\mathrm{age10} = (\text{approximate age} - 50)/10$. $\beta_{M1}$ is the difference in PWV (PWV original recorded units) per 1 original recorded unit higher triglycerides, with age, sex and height held fixed.

\textit{Intervals.} 95\% confidence interval.
{\footnotesize\setlength{\tabcolsep}{3pt}
\begin{longtable}{@{}>{\raggedright\arraybackslash}p{0.14\linewidth}>{\raggedright\arraybackslash}p{0.29\linewidth}>{\raggedleft\arraybackslash}p{0.13\linewidth}>{\raggedright\arraybackslash}p{0.22\linewidth}>{\raggedleft\arraybackslash}p{0.14\linewidth}@{}}
\toprule
Block & Parameter & Estimate & Interval & $P$ \\
\midrule\endhead
Pairwise association & $\beta_{M1}$: adjusted triglycerides--PWV association coefficient & $0.001753$ & $[1.251\times10^{-5}$, $0.003493]$ & \mbox{$0.04838$} \\
\bottomrule
\end{longtable}}

\paragraph{11.\ Pairwise association: seated diastolic blood pressure--PWV ($\pm$30 days)}
\textit{Design.} Seated diastolic blood pressure and PWV measured in the same period: with each diastolic blood pressure measurement date as an anchor, the PWV measurement closest in date within 30 days before or after (endpoints inclusive) is taken; a participant may contribute multiple anchor dates, $n = 10904$. Diastolic blood pressure taken as the mean of the available values of the two seated readings, PWV as the mean of the available sides. Adjustment for approximate age (anchor-date year minus birth year), sex and the height closest in date within 365 days before or after the anchor date. All variables on their original recorded scale, unstandardized; sex as a categorical term.

\textit{Model.} $Y = \beta_0 + \beta_{M2} \cdot M_2 + \beta_{\mathrm{age}} \cdot \mathrm{age10} + \beta_H \cdot H + \gamma_{\mathrm{sex}} + \varepsilon$, linear regression; $M_2$ is seated diastolic blood pressure, $Y$ thigh-to-ankle PWV, $H$ height, and $\mathrm{age10} = (\text{approximate age} - 50)/10$. $\beta_{M2}$ is the difference in PWV (PWV original recorded units) per 1 original recorded unit higher diastolic blood pressure, with age, sex and height held fixed.

\textit{Intervals.} 95\% confidence interval.
{\footnotesize\setlength{\tabcolsep}{3pt}
\begin{longtable}{@{}>{\raggedright\arraybackslash}p{0.14\linewidth}>{\raggedright\arraybackslash}p{0.29\linewidth}>{\raggedleft\arraybackslash}p{0.13\linewidth}>{\raggedright\arraybackslash}p{0.22\linewidth}>{\raggedleft\arraybackslash}p{0.14\linewidth}@{}}
\toprule
Block & Parameter & Estimate & Interval & $P$ \\
\midrule\endhead
Pairwise association & $\beta_{M2}$: adjusted seated diastolic blood pressure--PWV association coefficient & $0.07859$ & $[0.07555$, $0.08163]$ & \mbox{$0$} \\
\bottomrule
\end{longtable}}

\subsubsection*{Joint models}

\paragraph{12.\ Joint model: waist circumference, triglycerides, diastolic blood pressure and PWV ($\pm$90 days)}
\textit{Design.} All four measured in the same period: with each waist circumference measurement date as an anchor, triglycerides, seated diastolic blood pressure and PWV are each taken as the measurement closest in date within 90 days before or after the anchor (endpoints inclusive), with all three required; a participant may contribute multiple anchor dates, $n = 3212$. Diastolic blood pressure taken as the mean of the available values of the two seated readings, PWV as the mean of the available sides. Adjustment for approximate age (anchor-date year minus birth year), sex and same-day height at the waist circumference measurement. All variables on their original recorded scale, unstandardized; sex as a categorical term.

\textit{Model.} $Y = \beta_0 + \beta_A \cdot A + \beta_{M1} \cdot M_1 + \beta_{M2} \cdot M_2 + \beta_{\mathrm{age}} \cdot \mathrm{age10} + \beta_H \cdot H + \gamma_{\mathrm{sex}} + \varepsilon$, linear regression; $A$, $M_1$, $M_2$ and $Y$ are waist circumference, triglycerides, seated diastolic blood pressure and thigh-to-ankle PWV, respectively, $H$ is height, and $\mathrm{age10} = (\text{approximate age} - 50)/10$. $\beta_A$, $\beta_{M1}$ and $\beta_{M2}$ are the differences in PWV (PWV original recorded units) per 1 original recorded unit higher value of the respective variable, with the remaining model terms held fixed.

\textit{Intervals.} 95\% confidence intervals; $P$ values not corrected for multiple comparisons.
{\footnotesize\setlength{\tabcolsep}{3pt}
\begin{longtable}{@{}>{\raggedright\arraybackslash}p{0.14\linewidth}>{\raggedright\arraybackslash}p{0.29\linewidth}>{\raggedleft\arraybackslash}p{0.13\linewidth}>{\raggedright\arraybackslash}p{0.22\linewidth}>{\raggedleft\arraybackslash}p{0.14\linewidth}@{}}
\toprule
Block & Parameter & Estimate & Interval & $P$ \\
\midrule\endhead
$\pm$90-day window; mutually adjusted model: waist circumference + triglycerides + diastolic blood pressure + age + sex + height & $\beta_A$: waist circumference coefficient (adjusted for triglycerides and diastolic blood pressure) & $-0.000454$ & $[-0.005548$, $0.00464]$ & \mbox{$0.8613$} \\
 & $\beta_{M1}$: triglycerides coefficient (adjusted for waist circumference and diastolic blood pressure) & $0.0008425$ & $[3.615\times10^{-5}$, $0.001649]$ & \mbox{$0.04058$} \\
 & $\beta_{M2}$: seated diastolic blood pressure coefficient (adjusted for waist circumference and triglycerides) & $0.07457$ & $[0.06845$, $0.0807]$ & \mbox{$8.644\times10^{-116}$} \\
\bottomrule
\end{longtable}}

\subsubsection*{Reverse direction}

\paragraph{13.\ Reverse path: PWV $\to$ triglycerides $\to$ waist circumference}
\textit{Design.} Endpoint roles reversed: PWV, triglycerides and waist circumference measured in strict date order, triglycerides 1--730 days after PWV, waist circumference 1--730 days after triglycerides (endpoints inclusive); one measurement chain per participant. Adjustment for approximate age at the PWV measurement and sex. PWV taken as the mean of the available left and right values. All variables entered as their original recorded values, with sex as a categorical term.

\textit{Model.} $M = \alpha_0 + \alpha_A \cdot A + \alpha_C^{\top} \cdot C + \varepsilon_M$; $Y = \beta_0 + \beta_A \cdot A + \beta_M \cdot M + \beta_C^{\top} \cdot C + \varepsilon_Y$; $Y = \gamma_0 + \gamma_A \cdot A + \gamma_C^{\top} \cdot C + \varepsilon_T$. All three equations are linear regressions; $A$ is thigh-to-ankle PWV, $M$ triglycerides and $Y$ waist circumference, and $C$ is approximate age and sex. $\alpha_A$ is the PWV $\to$ triglycerides slope, $\beta_M$ the triglycerides $\to$ waist circumference slope adjusted for PWV, $\beta_A$ the PWV $\to$ waist circumference slope adjusted for triglycerides, $\gamma_A$ the total PWV $\to$ waist circumference slope, and reverse path product $\theta = \alpha_A \cdot \beta_M$.

\textit{Intervals.} $\alpha_A$, $\beta_M$, $\beta_A$, $\gamma_A$: bootstrap 95\% percentile intervals; $\theta$: bootstrap 97.5\% percentile interval.
{\footnotesize\setlength{\tabcolsep}{3pt}
\begin{longtable}{@{}>{\raggedright\arraybackslash}p{0.15\linewidth}>{\raggedright\arraybackslash}p{0.40\linewidth}>{\raggedleft\arraybackslash}p{0.14\linewidth}>{\raggedright\arraybackslash}p{0.24\linewidth}@{}}
\toprule
Block & Parameter & Estimate & Interval \\
\midrule\endhead
Reverse path index & $\theta = \alpha_A \cdot \beta_M$: reverse path-product association index & $0.5632$ & $[0.2413$, $0.9478]$ \\
Reverse path components & $\alpha_A$: PWV $\to$ triglycerides slope (reverse $A \to M$) & $8.065$ & $[3.949$, $12.7]$ \\
 & $\beta_M$: triglycerides $\to$ waist circumference slope, adjusted for PWV (reverse $M \to Y \mid A$) & $0.06983$ & $[0.04548$, $0.09814]$ \\
 & $\beta_A$: PWV $\to$ waist circumference slope, adjusted for triglycerides (reverse $A \to Y \mid M$) & $0.2526$ & $[-0.408$, $0.9484]$ \\
 & $\gamma_A$: adjusted total PWV $\to$ waist circumference slope (reverse $A \to Y$) & $0.8158$ & $[0.1361$, $1.522]$ \\
\bottomrule
\end{longtable}}

\paragraph{14.\ Reverse path: PWV $\to$ seated diastolic blood pressure $\to$ waist circumference}
\textit{Design.} Endpoint roles reversed: PWV, seated blood pressure and waist circumference measured in strict date order, blood pressure 1--730 days after PWV, waist circumference 1--730 days after blood pressure (endpoints inclusive); one measurement chain per participant. Adjustment for approximate age at the PWV measurement and sex. Diastolic blood pressure taken as the mean of the available values of the two seated readings, PWV as the mean of the available left and right values. All variables entered as their original recorded values, with sex as a categorical term.

\textit{Model.} $M = \alpha_0 + \alpha_A \cdot A + \alpha_C^{\top} \cdot C + \varepsilon_M$; $Y = \beta_0 + \beta_A \cdot A + \beta_M \cdot M + \beta_C^{\top} \cdot C + \varepsilon_Y$; $Y = \gamma_0 + \gamma_A \cdot A + \gamma_C^{\top} \cdot C + \varepsilon_T$. All three equations are linear regressions; $A$ is thigh-to-ankle PWV, $M$ seated diastolic blood pressure and $Y$ waist circumference, and $C$ is approximate age and sex. $\alpha_A$ is the PWV $\to$ diastolic blood pressure slope, $\beta_M$ the diastolic blood pressure $\to$ waist circumference slope adjusted for PWV, $\beta_A$ the PWV $\to$ waist circumference slope adjusted for diastolic blood pressure, $\gamma_A$ the total PWV $\to$ waist circumference slope, and reverse path product $\theta = \alpha_A \cdot \beta_M$.

\textit{Intervals.} $\alpha_A$, $\beta_M$, $\beta_A$, $\gamma_A$: bootstrap 95\% percentile intervals; $\theta$: bootstrap 97.5\% percentile interval.
{\footnotesize\setlength{\tabcolsep}{3pt}
\begin{longtable}{@{}>{\raggedright\arraybackslash}p{0.15\linewidth}>{\raggedright\arraybackslash}p{0.40\linewidth}>{\raggedleft\arraybackslash}p{0.14\linewidth}>{\raggedright\arraybackslash}p{0.24\linewidth}@{}}
\toprule
Block & Parameter & Estimate & Interval \\
\midrule\endhead
Reverse path index & $\theta = \alpha_A \cdot \beta_M$: reverse path-product association index & $-0.2471$ & $[-25.98$, $861.8]$ \\
Reverse path components & $\alpha_A$: PWV $\to$ seated diastolic blood pressure slope (reverse $A \to M$) & $4.865$ & $[-17.16$, $11.19]$ \\
 & $\beta_M$: seated diastolic blood pressure $\to$ waist circumference slope, adjusted for PWV (reverse $M \to Y \mid A$) & $-0.05079$ & $[-6.943$, $131.1]$ \\
 & $\beta_A$: PWV $\to$ waist circumference slope, adjusted for diastolic blood pressure (reverse $A \to Y \mid M$) & $4.742$ & $[-863.6$, $19.83]$ \\
 & $\gamma_A$: adjusted total PWV $\to$ waist circumference slope (reverse $A \to Y$) & $4.495$ & $[-5.609$, $11.31]$ \\
\bottomrule
\end{longtable}}

\paragraph{15.\ Reversed measurement order: serial path with triglycerides preceding waist circumference}
\textit{Design.} Starting from each triglycerides measurement date, waist circumference is taken as the closest measurement within 1--365 days afterwards (same day excluded); seated diastolic blood pressure is taken as the measurement closest to that triglycerides measurement and PWV as the measurement closest to that diastolic blood pressure, each no more than 365 days apart (in either order, endpoints inclusive throughout); a participant may contribute multiple chains, $n = 7071$. Adjustment for approximate age at the triglycerides measurement date and recorded sex. Diastolic blood pressure taken as the mean of the available values of the two seated readings, PWV as the mean of the available sides; all variables taken as their original recorded values, unstandardized.

\textit{Model.} $M_1 = \alpha_1 + \rho_{AM_1} A + \gamma_1^{\top} C + \varepsilon_1$; $M_2 = \alpha_2 + \delta_2 A + \rho_{M_1M_2} M_1 + \gamma_2^{\top} C + \varepsilon_2$; $Y = \alpha_3 + \delta_3 A + \delta_4 M_1 + \rho_{M_2Y} M_2 + \gamma_3^{\top} C + \varepsilon_3$. Equation directions as in the forward chain; all three equations are linear regressions; $A$, $M_1$, $M_2$ and $Y$ are waist circumference, triglycerides, seated diastolic blood pressure and thigh-to-ankle PWV, respectively (original recorded scale), and $C$ is approximate age and recorded sex. $\rho_{AM_1}$, $\rho_{M_1M_2}$ and $\rho_{M_2Y}$ are the focal slopes of the respective equations (the change in the dependent variable, in recorded units, per 1 recorded unit higher independent variable); serial path product $\theta = \rho_{AM_1} \cdot \rho_{M_1M_2} \cdot \rho_{M_2Y}$.

\textit{Intervals.} Bootstrap 95\% percentile intervals; $P$ values not corrected for multiple comparisons.
{\footnotesize\setlength{\tabcolsep}{3pt}
\begin{longtable}{@{}>{\raggedright\arraybackslash}p{0.14\linewidth}>{\raggedright\arraybackslash}p{0.29\linewidth}>{\raggedleft\arraybackslash}p{0.13\linewidth}>{\raggedright\arraybackslash}p{0.22\linewidth}>{\raggedleft\arraybackslash}p{0.14\linewidth}@{}}
\toprule
Block & Parameter & Estimate & Interval & $P$ \\
\midrule\endhead
Path components (triglycerides before waist circumference) & $\rho_{AM_1}$: adjusted waist circumference $\to$ triglycerides slope ($A \to M_1$) & $1.955$ & $[1.801$, $2.121]$ & \mbox{$0$} \\
 & $\rho_{M_1M_2}$: triglycerides $\to$ seated diastolic blood pressure slope, adjusted for waist circumference ($M_1 \to M_2 \mid A$) & $0.01547$ & $[0.009197$, $0.02428]$ & \mbox{$0$} \\
 & $\rho_{M_2Y}$: seated diastolic blood pressure $\to$ PWV slope, adjusted for waist circumference and triglycerides ($M_2 \to Y \mid A, M_1$) & $0.07573$ & $[0.07052$, $0.08051]$ & \mbox{$0$} \\
Path index (triglycerides before waist circumference) & $\theta = \rho_{AM_1} \cdot \rho_{M_1M_2} \cdot \rho_{M_2Y}$: serial path product & $0.002291$ & $[0.001387$, $0.003601]$ & \mbox{$0$} \\
\bottomrule
\end{longtable}}

\subsubsection*{Heterogeneity and interaction}

\paragraph{16.\ Sex heterogeneity: waist circumference $\to$ triglycerides $\to$ diastolic blood pressure $\to$ PWV}
\textit{Design.} Timing design as in the main analysis: starting from each waist circumference measurement date, triglycerides, seated diastolic blood pressure and PWV are taken in turn as the measurement closest to the preceding item, with adjacent items no more than 365 days apart (endpoints inclusive, in either order); a participant may contribute multiple chains. Recorded sex enters as a main effect and through interaction terms with the focal variable of each equation, with additional adjustment for approximate age at the waist circumference measurement date. Diastolic blood pressure taken as the mean of the available values of the two seated readings, PWV as the mean of the available sides; all variables taken as their original recorded values, unstandardized.

\textit{Model.} $M_1 = \alpha_1 + \beta_{AM_1} A + \lambda_1 \mathrm{Age} + f_1(S) + \eta_1 A \cdot S + \varepsilon_1$; $M_2 = \alpha_2 + \delta_2 A + \beta_{M_1M_2} M_1 + \lambda_2 \mathrm{Age} + f_2(S) + \eta_2 M_1 \cdot S + \varepsilon_2$; $Y = \alpha_3 + \delta_3 A + \delta_4 M_1 + \beta_{M_2Y} M_2 + \lambda_3 \mathrm{Age} + f_3(S) + \eta_3 M_2 \cdot S + \varepsilon_3$. All three equations are pooled-sample linear regressions; $A$, $M_1$, $M_2$ and $Y$ are waist circumference, triglycerides, seated diastolic blood pressure and thigh-to-ankle PWV, respectively (original recorded scale), $S$ is recorded sex, $f(S)$ the sex main effect, and the $\eta$ terms the interactions between the focal variables and sex. $\beta_{AM_1}(s)$, $\beta_{M_1M_2}(s)$ and $\beta_{M_2Y}(s)$ are the focal slopes within sex category $s$ (main effect plus the corresponding interaction term); within-category serial path product $\theta(s) = \beta_{AM_1}(s) \cdot \beta_{M_1M_2}(s) \cdot \beta_{M_2Y}(s)$.

\textit{Intervals.} Bootstrap 95\% percentile intervals; $P$ values not corrected for multiple comparisons.
{\footnotesize\setlength{\tabcolsep}{3pt}
\begin{longtable}{@{}>{\raggedright\arraybackslash}p{0.14\linewidth}>{\raggedright\arraybackslash}p{0.29\linewidth}>{\raggedleft\arraybackslash}p{0.13\linewidth}>{\raggedright\arraybackslash}p{0.22\linewidth}>{\raggedleft\arraybackslash}p{0.14\linewidth}@{}}
\toprule
Block & Parameter & Estimate & Interval & $P$ \\
\midrule\endhead
First sex category & $\beta_{AM_1}(s)$: category-specific waist circumference $\to$ triglycerides slope & $1.867$ & $[1.71$, $2.02]$ & \mbox{$0$} \\
 & $\beta_{M_1M_2}(s)$: category-specific triglycerides $\to$ seated diastolic blood pressure slope, adjusted for waist circumference & $0.02936$ & $[0.02248$, $0.03637]$ & \mbox{$0$} \\
 & $\beta_{M_2Y}(s)$: category-specific seated diastolic blood pressure $\to$ PWV slope, adjusted for waist circumference and triglycerides & $0.07881$ & $[0.07315$, $0.08428]$ & \mbox{$0$} \\
 & $\theta(s) = \beta_{AM_1}(s) \cdot \beta_{M_1M_2}(s) \cdot \beta_{M_2Y}(s)$: within-category serial path product & $0.004319$ & $[0.003187$, $0.005513]$ & \mbox{$0$} \\
Second sex category & $\beta_{AM_1}(s)$: category-specific waist circumference $\to$ triglycerides slope & $2.154$ & $[1.905$, $2.428]$ & \mbox{$0$} \\
 & $\beta_{M_1M_2}(s)$: category-specific triglycerides $\to$ seated diastolic blood pressure slope, adjusted for waist circumference & $0.01784$ & $[0.01271$, $0.02299]$ & \mbox{$0$} \\
 & $\beta_{M_2Y}(s)$: category-specific seated diastolic blood pressure $\to$ PWV slope, adjusted for waist circumference and triglycerides & $0.07233$ & $[0.06592$, $0.07845]$ & \mbox{$0$} \\
 & $\theta(s) = \beta_{AM_1}(s) \cdot \beta_{M_1M_2}(s) \cdot \beta_{M_2Y}(s)$: within-category serial path product & $0.002779$ & $[0.001935$, $0.003649]$ & \mbox{$0$} \\
\bottomrule
\end{longtable}}

\paragraph{17.\ Sex heterogeneity: sex-specific waist circumference--PWV slopes ($\pm$90 days)}
\textit{Design.} All four measured in the same period: with each waist circumference measurement date as an anchor, triglycerides, seated diastolic blood pressure and PWV are each taken as the measurement closest in date within 90 days before or after (endpoints inclusive), with all three required; a participant may contribute multiple anchor dates. A waist circumference $\times$ sex interaction is added to the covariates of the $\pm$90-day joint model (triglycerides, diastolic blood pressure, approximate age, sex, same-day height at the waist circumference measurement). All variables on their original recorded scale, unstandardized.

\textit{Model.} $Y = \beta_0 + \beta_{A,s} \cdot A + \beta_{M1} \cdot M_1 + \beta_{M2} \cdot M_2 + \beta_{\mathrm{age}} \cdot \mathrm{age10} + \beta_H \cdot H + \gamma_s + \varepsilon$, linear regression, where $s$ is the sex category and the waist circumference slope varies by category; remaining notation as in the $\pm$90-day joint model. $\beta_{A,s}$ is the difference in PWV (PWV original recorded units) per 1 original recorded unit higher waist circumference within sex category $s$, with triglycerides, diastolic blood pressure, age and height held fixed.

\textit{Intervals.} 95\% confidence interval.
{\footnotesize\setlength{\tabcolsep}{3pt}
\begin{longtable}{@{}>{\raggedright\arraybackslash}p{0.15\linewidth}>{\raggedright\arraybackslash}p{0.40\linewidth}>{\raggedleft\arraybackslash}p{0.14\linewidth}>{\raggedright\arraybackslash}p{0.24\linewidth}@{}}
\toprule
Block & Parameter & Estimate & Interval \\
\midrule\endhead
First sex category & $\beta_{A,1}$: waist circumference--PWV slope in the first sex category & $0.003328$ & $[-0.003239$, $0.009894]$ \\
Second sex category & $\beta_{A,2}$: waist circumference--PWV slope in the second sex category & $-0.004506$ & $[-0.01145$, $0.002435]$ \\
\bottomrule
\end{longtable}}

\subsubsection*{Sensitivity analyses}

\paragraph{18.\ Longer minimum intervals: waist circumference $\to$ triglycerides $\to$ PWV}
\textit{Design.} Waist circumference, triglycerides and PWV measured in strict date order, with minimum intervals longer than in the main analysis: triglycerides 90--365 days after waist circumference, PWV 180--730 days after triglycerides (endpoints inclusive); one measurement chain per participant. Adjustment for approximate age at the waist circumference measurement and sex. PWV taken as the mean of the available left and right values. All variables entered as their original recorded values, with sex as a categorical term.

\textit{Model.} $M = \alpha_0 + \alpha_A \cdot A + \alpha_C^{\top} \cdot C + \varepsilon_M$; $Y = \beta_0 + \beta_A \cdot A + \beta_M \cdot M + \beta_C^{\top} \cdot C + \varepsilon_Y$; $Y = \gamma_0 + \gamma_A \cdot A + \gamma_C^{\top} \cdot C + \varepsilon_T$. All three equations are linear regressions; $A$, $M$ and $Y$ are waist circumference, triglycerides and thigh-to-ankle PWV, respectively, and $C$ is approximate age and sex. $\alpha_A$ is the waist circumference $\to$ triglycerides slope, $\beta_M$ the triglycerides $\to$ PWV slope adjusted for waist circumference, $\beta_A$ the waist circumference $\to$ PWV slope adjusted for triglycerides, $\gamma_A$ the total waist circumference $\to$ PWV slope, and path product $\theta = \alpha_A \cdot \beta_M$.

\textit{Intervals.} $\alpha_A$, $\beta_M$, $\beta_A$, $\gamma_A$: bootstrap 95\% percentile intervals; $\theta$: bootstrap 97.5\% percentile interval.
{\footnotesize\setlength{\tabcolsep}{3pt}
\begin{longtable}{@{}>{\raggedright\arraybackslash}p{0.15\linewidth}>{\raggedright\arraybackslash}p{0.40\linewidth}>{\raggedleft\arraybackslash}p{0.14\linewidth}>{\raggedright\arraybackslash}p{0.24\linewidth}@{}}
\toprule
Block & Parameter & Estimate & Interval \\
\midrule\endhead
Path index & $\theta = \alpha_A \cdot \beta_M$: path-product association index & $0.003839$ & $[-0.002721$, $0.01121]$ \\
Path components & $\alpha_A$: waist circumference $\to$ triglycerides slope ($A \to M$) & $1.694$ & $[1.171$, $2.266]$ \\
 & $\beta_M$: triglycerides $\to$ thigh-to-ankle PWV slope, adjusted for waist circumference ($M \to Y \mid A$) & $0.002267$ & $[-0.001211$, $0.005696]$ \\
 & $\beta_A$: waist circumference $\to$ PWV slope, adjusted for triglycerides ($A \to Y \mid M$) & $0.01961$ & $[0.00545$, $0.03404]$ \\
 & $\gamma_A$: adjusted total waist circumference $\to$ PWV slope ($A \to Y$) & $0.02344$ & $[0.01092$, $0.03668]$ \\
\bottomrule
\end{longtable}}

\paragraph{19.\ Genetic principal component adjustment: waist circumference $\to$ triglycerides $\to$ PWV}
\textit{Design.} Waist circumference, triglycerides and PWV measured in strict date order: triglycerides 30--365 days after waist circumference, PWV 90--1095 days after triglycerides (endpoints inclusive); one measurement chain per participant. Adjustment for approximate age at the waist circumference measurement, sex and genetic principal components PC1--PC10. PWV taken as the mean of the available left and right values. All variables entered as their original recorded values, with sex as a categorical term.

\textit{Model.} $M = \alpha_0 + \alpha_A \cdot A + \alpha_C^{\top} \cdot C + \varepsilon_M$; $Y = \beta_0 + \beta_A \cdot A + \beta_M \cdot M + \beta_C^{\top} \cdot C + \varepsilon_Y$; $Y = \gamma_0 + \gamma_A \cdot A + \gamma_C^{\top} \cdot C + \varepsilon_T$. All three equations are linear regressions; $A$, $M$ and $Y$ are waist circumference, triglycerides and thigh-to-ankle PWV, respectively, and $C$ is approximate age, sex and genetic principal components PC1--PC10. $\alpha_A$ is the waist circumference $\to$ triglycerides slope, $\beta_M$ the triglycerides $\to$ PWV slope adjusted for waist circumference, $\beta_A$ the waist circumference $\to$ PWV slope adjusted for triglycerides, $\gamma_A$ the total waist circumference $\to$ PWV slope, and path product $\theta = \alpha_A \cdot \beta_M$.

\textit{Intervals.} $\alpha_A$, $\beta_M$, $\beta_A$, $\gamma_A$: bootstrap 95\% percentile intervals; $\theta$: bootstrap 97.5\% percentile interval.
{\footnotesize\setlength{\tabcolsep}{3pt}
\begin{longtable}{@{}>{\raggedright\arraybackslash}p{0.15\linewidth}>{\raggedright\arraybackslash}p{0.40\linewidth}>{\raggedleft\arraybackslash}p{0.14\linewidth}>{\raggedright\arraybackslash}p{0.24\linewidth}@{}}
\toprule
Block & Parameter & Estimate & Interval \\
\midrule\endhead
Path index & $\theta = \alpha_A \cdot \beta_M$: path-product association index & $0.003784$ & $[-0.003173$, $0.0112]$ \\
Path components & $\alpha_A$: waist circumference $\to$ triglycerides slope ($A \to M$) & $1.801$ & $[1.305$, $2.321]$ \\
 & $\beta_M$: triglycerides $\to$ thigh-to-ankle PWV slope, adjusted for waist circumference ($M \to Y \mid A$) & $0.002102$ & $[-0.001155$, $0.005571]$ \\
 & $\beta_A$: waist circumference $\to$ PWV slope, adjusted for triglycerides ($A \to Y \mid M$) & $0.01763$ & $[0.003698$, $0.03176]$ \\
 & $\gamma_A$: adjusted total waist circumference $\to$ PWV slope ($A \to Y$) & $0.02142$ & $[0.009592$, $0.03402]$ \\
\bottomrule
\end{longtable}}

\paragraph{20.\ $\pm$180-day window serial path: waist circumference $\to$ triglycerides $\to$ diastolic blood pressure $\to$ PWV}
\textit{Design.} Chain construction as in the main analysis, with a narrower window: starting from each waist circumference measurement date, triglycerides, seated diastolic blood pressure and PWV are taken in turn as the measurement closest to the preceding item, with adjacent items no more than 180 days apart (endpoints inclusive, in either order); a participant may contribute multiple chains, $n = 4846$. Adjustment for approximate age at the waist circumference measurement date and recorded sex. Diastolic blood pressure taken as the mean of the available values of the two seated readings, PWV as the mean of the available sides; all variables taken as their original recorded values, unstandardized.

\textit{Model.} $M_1 = \alpha_1 + \beta_{AM_1} A + \gamma_1^{\top} C + \varepsilon_1$; $M_2 = \alpha_2 + \delta_2 A + \beta_{M_1M_2} M_1 + \gamma_2^{\top} C + \varepsilon_2$; $Y = \alpha_3 + \delta_3 A + \delta_4 M_1 + \beta_{M_2Y} M_2 + \gamma_3^{\top} C + \varepsilon_3$. All three equations are linear regressions; $A$, $M_1$, $M_2$ and $Y$ are waist circumference, triglycerides, seated diastolic blood pressure and thigh-to-ankle PWV, respectively (original recorded scale), and $C$ is approximate age and recorded sex. $\beta_{AM_1}$, $\beta_{M_1M_2}$ and $\beta_{M_2Y}$ are the focal slopes of the respective equations (the change in the dependent variable, in recorded units, per 1 recorded unit higher independent variable); serial path product $\theta = \beta_{AM_1} \cdot \beta_{M_1M_2} \cdot \beta_{M_2Y}$.

\textit{Intervals.} Bootstrap 95\% percentile intervals; $P$ values not corrected for multiple comparisons.
{\footnotesize\setlength{\tabcolsep}{3pt}
\begin{longtable}{@{}>{\raggedright\arraybackslash}p{0.14\linewidth}>{\raggedright\arraybackslash}p{0.29\linewidth}>{\raggedleft\arraybackslash}p{0.13\linewidth}>{\raggedright\arraybackslash}p{0.22\linewidth}>{\raggedleft\arraybackslash}p{0.14\linewidth}@{}}
\toprule
Block & Parameter & Estimate & Interval & $P$ \\
\midrule\endhead
Path components ($\pm$180 days) & $\beta_{AM_1}$: adjusted waist circumference $\to$ triglycerides slope ($A \to M_1$) & $2.03$ & $[1.853$, $2.209]$ & \mbox{$0$} \\
 & $\beta_{M_1M_2}$: triglycerides $\to$ seated diastolic blood pressure slope, adjusted for waist circumference ($M_1 \to M_2 \mid A$) & $0.02075$ & $[0.01606$, $0.02611]$ & \mbox{$0$} \\
 & $\beta_{M_2Y}$: seated diastolic blood pressure $\to$ PWV slope, adjusted for waist circumference and triglycerides ($M_2 \to Y \mid A, M_1$) & $0.07505$ & $[0.07006$, $0.08002]$ & \mbox{$0$} \\
Path index ($\pm$180 days) & $\theta = \beta_{AM_1} \cdot \beta_{M_1M_2} \cdot \beta_{M_2Y}$: serial path product & $0.003162$ & $[0.002424$, $0.004015]$ & \mbox{$0$} \\
\bottomrule
\end{longtable}}

\paragraph{21.\ Triglycerides after waist circumference: waist circumference $\to$ triglycerides $\to$ diastolic blood pressure $\to$ PWV}
\textit{Design.} Starting from each waist circumference measurement date, triglycerides are taken as the closest measurement within 1--365 days afterwards (same day excluded); seated diastolic blood pressure is taken as the measurement closest to that triglycerides measurement and PWV as the measurement closest to that diastolic blood pressure, each no more than 365 days apart (in either order, endpoints inclusive throughout); a participant may contribute multiple chains, $n = 567$. Adjustment for approximate age at the waist circumference measurement date and recorded sex. Diastolic blood pressure taken as the mean of the available values of the two seated readings, PWV as the mean of the available sides; all variables taken as their original recorded values, unstandardized.

\textit{Model.} $M_1 = \alpha_1 + \beta_{AM_1} A + \gamma_1^{\top} C + \varepsilon_1$; $M_2 = \alpha_2 + \delta_2 A + \beta_{M_1M_2} M_1 + \gamma_2^{\top} C + \varepsilon_2$; $Y = \alpha_3 + \delta_3 A + \delta_4 M_1 + \beta_{M_2Y} M_2 + \gamma_3^{\top} C + \varepsilon_3$. All three equations are linear regressions; $A$, $M_1$, $M_2$ and $Y$ are waist circumference, triglycerides, seated diastolic blood pressure and thigh-to-ankle PWV, respectively (original recorded scale), and $C$ is approximate age and recorded sex. $\beta_{AM_1}$, $\beta_{M_1M_2}$ and $\beta_{M_2Y}$ are the focal slopes of the respective equations (the change in the dependent variable, in recorded units, per 1 recorded unit higher independent variable); serial path product $\theta = \beta_{AM_1} \cdot \beta_{M_1M_2} \cdot \beta_{M_2Y}$.

\textit{Intervals.} Bootstrap 95\% percentile intervals; $P$ values not corrected for multiple comparisons.
{\footnotesize\setlength{\tabcolsep}{3pt}
\begin{longtable}{@{}>{\raggedright\arraybackslash}p{0.14\linewidth}>{\raggedright\arraybackslash}p{0.29\linewidth}>{\raggedleft\arraybackslash}p{0.13\linewidth}>{\raggedright\arraybackslash}p{0.22\linewidth}>{\raggedleft\arraybackslash}p{0.14\linewidth}@{}}
\toprule
Block & Parameter & Estimate & Interval & $P$ \\
\midrule\endhead
Path components (triglycerides after waist circumference) & $\beta_{AM_1}$: adjusted waist circumference $\to$ triglycerides slope ($A \to M_1$) & $1.963$ & $[1.57$, $2.37]$ & \mbox{$0$} \\
 & $\beta_{M_1M_2}$: triglycerides $\to$ seated diastolic blood pressure slope, adjusted for waist circumference ($M_1 \to M_2 \mid A$) & $0.02938$ & $[0.01507$, $0.04593]$ & \mbox{$0$} \\
 & $\beta_{M_2Y}$: seated diastolic blood pressure $\to$ PWV slope, adjusted for waist circumference and triglycerides ($M_2 \to Y \mid A, M_1$) & $0.07855$ & $[0.06476$, $0.09385]$ & \mbox{$0$} \\
Path index (triglycerides after waist circumference) & $\theta = \beta_{AM_1} \cdot \beta_{M_1M_2} \cdot \beta_{M_2Y}$: serial path product & $0.004529$ & $[0.002257$, $0.007131]$ & \mbox{$0$} \\
\bottomrule
\end{longtable}}

\paragraph{22.\ Added smoking adjustment: waist circumference $\to$ triglycerides $\to$ diastolic blood pressure $\to$ PWV}
\textit{Design.} Starting from each waist circumference measurement date, triglycerides, seated diastolic blood pressure and PWV are taken in turn as the measurement closest to the preceding item, with adjacent items no more than 365 days apart (endpoints inclusive, in either order); a participant may contribute multiple chains, $n = 4881$. In addition to approximate age and recorded sex, adjustment for current smoking status from the most recent questionnaire record on or within 730 days before the waist circumference measurement date. Diastolic blood pressure taken as the mean of the available values of the two seated readings, PWV as the mean of the available sides; all variables taken as their original recorded values, unstandardized.

\textit{Model.} $M_1 = \alpha_1 + \beta_{AM_1} A + \gamma_1^{\top} C + \varepsilon_1$; $M_2 = \alpha_2 + \delta_2 A + \beta_{M_1M_2} M_1 + \gamma_2^{\top} C + \varepsilon_2$; $Y = \alpha_3 + \delta_3 A + \delta_4 M_1 + \beta_{M_2Y} M_2 + \gamma_3^{\top} C + \varepsilon_3$. All three equations are linear regressions; $A$, $M_1$, $M_2$ and $Y$ are waist circumference, triglycerides, seated diastolic blood pressure and thigh-to-ankle PWV, respectively (original recorded scale), and $C$ is approximate age, recorded sex and current smoking status. $\beta_{AM_1}$, $\beta_{M_1M_2}$ and $\beta_{M_2Y}$ are the focal slopes of the respective equations (the change in the dependent variable, in recorded units, per 1 recorded unit higher independent variable); serial path product $\theta = \beta_{AM_1} \cdot \beta_{M_1M_2} \cdot \beta_{M_2Y}$.

\textit{Intervals.} Bootstrap 95\% percentile intervals; $P$ values not corrected for multiple comparisons.
{\footnotesize\setlength{\tabcolsep}{3pt}
\begin{longtable}{@{}>{\raggedright\arraybackslash}p{0.14\linewidth}>{\raggedright\arraybackslash}p{0.29\linewidth}>{\raggedleft\arraybackslash}p{0.13\linewidth}>{\raggedright\arraybackslash}p{0.22\linewidth}>{\raggedleft\arraybackslash}p{0.14\linewidth}@{}}
\toprule
Block & Parameter & Estimate & Interval & $P$ \\
\midrule\endhead
Path components (added smoking adjustment) & $\beta_{AM_1}$: adjusted waist circumference $\to$ triglycerides slope ($A \to M_1$) & $1.894$ & $[1.734$, $2.065]$ & \mbox{$0$} \\
 & $\beta_{M_1M_2}$: triglycerides $\to$ seated diastolic blood pressure slope, adjusted for waist circumference ($M_1 \to M_2 \mid A$) & $0.0199$ & $[0.01504$, $0.02504]$ & \mbox{$0$} \\
 & $\beta_{M_2Y}$: seated diastolic blood pressure $\to$ PWV slope, adjusted for waist circumference and triglycerides ($M_2 \to Y \mid A, M_1$) & $0.07424$ & $[0.06929$, $0.07903]$ & \mbox{$0$} \\
Path index (added smoking adjustment) & $\theta = \beta_{AM_1} \cdot \beta_{M_1M_2} \cdot \beta_{M_2Y}$: serial path product & $0.002798$ & $[0.002083$, $0.003586]$ & \mbox{$0$} \\
\bottomrule
\end{longtable}}

\paragraph{23.\ Joint model: waist circumference, triglycerides, diastolic blood pressure and PWV ($\pm$30 days)}
\textit{Design.} All four measured in the same period: with each waist circumference measurement date as an anchor, triglycerides, seated diastolic blood pressure and PWV are each taken as the measurement closest in date within 30 days before or after the anchor (endpoints inclusive), with all three required; a participant may contribute multiple anchor dates. Measurement definitions, covariates (approximate age, sex, same-day height at the waist circumference measurement) and original recorded scale as in the $\pm$90-day joint model.

\textit{Model.} As in the $\pm$90-day joint model: $Y = \beta_0 + \beta_A \cdot A + \beta_{M1} \cdot M_1 + \beta_{M2} \cdot M_2 + \beta_{\mathrm{age}} \cdot \mathrm{age10} + \beta_H \cdot H + \gamma_{\mathrm{sex}} + \varepsilon$ (linear regression). $\beta_A$, $\beta_{M1}$ and $\beta_{M2}$ are the differences in PWV (PWV original recorded units) per 1 original recorded unit higher value of the respective variable, with the remaining model terms held fixed.

\textit{Intervals.} 95\% confidence intervals; $P$ values not corrected for multiple comparisons.
{\footnotesize\setlength{\tabcolsep}{3pt}
\begin{longtable}{@{}>{\raggedright\arraybackslash}p{0.14\linewidth}>{\raggedright\arraybackslash}p{0.29\linewidth}>{\raggedleft\arraybackslash}p{0.13\linewidth}>{\raggedright\arraybackslash}p{0.22\linewidth}>{\raggedleft\arraybackslash}p{0.14\linewidth}@{}}
\toprule
Block & Parameter & Estimate & Interval & $P$ \\
\midrule\endhead
$\pm$30-day window; mutually adjusted model: waist circumference + triglycerides + diastolic blood pressure + age + sex + height & $\beta_A$: waist circumference coefficient (adjusted for triglycerides and diastolic blood pressure) & $-0.0001311$ & $[-0.007559$, $0.007297]$ & \mbox{$0.9724$} \\
 & $\beta_{M1}$: triglycerides coefficient (adjusted for waist circumference and diastolic blood pressure) & $0.001771$ & $[0.0005955$, $0.002946]$ & \mbox{$0.003172$} \\
 & $\beta_{M2}$: seated diastolic blood pressure coefficient (adjusted for waist circumference and triglycerides) & $0.07002$ & $[0.06083$, $0.07922]$ & \mbox{$3.423\times10^{-47}$} \\
\bottomrule
\end{longtable}}

\paragraph{24.\ Joint model with additional adjustment for education and current smoking ($\pm$90 days)}
\textit{Design.} All four measured in the same period: with each waist circumference measurement date as an anchor, triglycerides, seated diastolic blood pressure and PWV are each taken as the measurement closest in date within 90 days before or after (endpoints inclusive), with all three required; a participant may contribute multiple anchor dates, $n = 1643$. In addition to approximate age, sex and same-day height at the waist circumference measurement, adjustment for education level (closest in date within 730 days before or after) and current smoking status (closest in date within 365 days before or after), both as categorical terms. All variables on their original recorded scale, unstandardized.

\textit{Model.} $Y = \beta_0 + \beta_A \cdot A + \beta_{M1} \cdot M_1 + \beta_{M2} \cdot M_2 + \beta_{\mathrm{age}} \cdot \mathrm{age10} + \beta_H \cdot H + \gamma_{\mathrm{sex}} + \gamma_{\mathrm{edu}} + \gamma_{\mathrm{smk}} + \varepsilon$, linear regression; $\gamma_{\mathrm{edu}}$ and $\gamma_{\mathrm{smk}}$ are the categorical terms for education level and current smoking status; remaining notation as in the $\pm$90-day joint model. $\beta_A$, $\beta_{M1}$ and $\beta_{M2}$ are the differences in PWV (PWV original recorded units) per 1 original recorded unit higher value of the respective variable, with the remaining model terms held fixed.

\textit{Intervals.} 95\% confidence intervals; $P$ values not corrected for multiple comparisons.
{\footnotesize\setlength{\tabcolsep}{3pt}
\begin{longtable}{@{}>{\raggedright\arraybackslash}p{0.14\linewidth}>{\raggedright\arraybackslash}p{0.29\linewidth}>{\raggedleft\arraybackslash}p{0.13\linewidth}>{\raggedright\arraybackslash}p{0.22\linewidth}>{\raggedleft\arraybackslash}p{0.14\linewidth}@{}}
\toprule
Block & Parameter & Estimate & Interval & $P$ \\
\midrule\endhead
$\pm$90-day window; additionally adjusted model: waist circumference + triglycerides + diastolic blood pressure + age + sex + height + education + smoking & $\beta_A$: waist circumference coefficient (adjusted for triglycerides and diastolic blood pressure) & $-0.004073$ & $[-0.01156$, $0.003419]$ & \mbox{$0.2865$} \\
 & $\beta_{M1}$: triglycerides coefficient (adjusted for waist circumference and diastolic blood pressure) & $0.000656$ & $[-0.0005758$, $0.001888]$ & \mbox{$0.2964$} \\
 & $\beta_{M2}$: seated diastolic blood pressure coefficient (adjusted for waist circumference and triglycerides) & $0.07576$ & $[0.06672$, $0.0848]$ & \mbox{$3.005\times10^{-56}$} \\
\bottomrule
\end{longtable}}

\paragraph{25.\ PWV-anchored pairing: waist circumference--PWV ($\pm$180 days)}
\textit{Design.} Waist circumference and PWV measured in the same period: with each PWV measurement date as an anchor, the waist circumference measurement closest in date within 180 days before or after (endpoints inclusive) is taken; a participant may contribute multiple anchor dates, $n = 10898$. PWV taken as the mean of the available sides. Adjustment for approximate age (PWV measurement-date year minus birth year), sex and height measured on the same day as the selected waist circumference. All variables on their original recorded scale, unstandardized; sex as a categorical term.

\textit{Model.} $Y = \beta_0 + \beta_A \cdot A + \beta_{\mathrm{age}} \cdot \mathrm{age10} + \beta_H \cdot H + \gamma_{\mathrm{sex}} + \varepsilon$, linear regression; $A$ is waist circumference, $Y$ thigh-to-ankle PWV, $H$ height, and $\mathrm{age10} = (\text{approximate age} - 50)/10$. $\beta_A$ is the difference in PWV (PWV original recorded units) per 1 original recorded unit higher waist circumference, with age, sex and height held fixed.

\textit{Intervals.} 95\% confidence interval.
{\footnotesize\setlength{\tabcolsep}{3pt}
\begin{longtable}{@{}>{\raggedright\arraybackslash}p{0.14\linewidth}>{\raggedright\arraybackslash}p{0.29\linewidth}>{\raggedleft\arraybackslash}p{0.13\linewidth}>{\raggedright\arraybackslash}p{0.22\linewidth}>{\raggedleft\arraybackslash}p{0.14\linewidth}@{}}
\toprule
Block & Parameter & Estimate & Interval & $P$ \\
\midrule\endhead
PWV-anchored pairing & $\beta_A$: adjusted waist circumference--PWV association coefficient under PWV-anchored pairing & $0.0272$ & $[0.02439$, $0.03001]$ & \mbox{$1.065\times10^{-78}$} \\
\bottomrule
\end{longtable}}

\FloatBarrier

\subsection{Hip circumference--TG--DBP--PWV}
\label{app:rq-designs:4703}

\paragraph{Research question and measures.}
The research question concerns the association of hip circumference with lower-limb arterial stiffness along the serial path through triglycerides and diastolic pressure, with data from the HPP cohort. The exposure is hip circumference; the mediators are blood-test triglycerides and seated diastolic pressure (mean of the available values of two seated readings); the outcome is thigh-to-ankle pulse wave velocity (PWV, mean of the available sides). Unless stated otherwise, coefficients are SD changes per 1 SD after standardization within the analysis sample of each unit.

\subsubsection*{Primary pathways}

\paragraph{1.\ Single-mediator path: hip circumference $\to$ triglycerides $\to$ thigh-to-ankle PWV}
\textit{Design.} Hip circumference, triglycerides and PWV are measured in strict date order: triglycerides 30--730 days after hip circumference and PWV 90--1095 days after triglycerides (bounds inclusive); all eligible measurement chains of a participant are included and weighted to give each participant equal weight. Adjustment is for approximate age at the hip circumference measurement and sex. PWV is the mean of the available sides. Hip circumference, triglycerides, PWV and age are standardized within the analysis sample, with sex as a categorical term.

\textit{Model.} $M = \gamma_0 + \gamma_A \cdot A + \gamma_{\mathrm{age}} \cdot \mathrm{age} + \gamma_{\mathrm{sex}} + \varepsilon_M$; $Y = \beta_0 + \beta_A \cdot A + \beta_M \cdot M + \beta_{AM} \cdot A \cdot M + \beta_{\mathrm{age}} \cdot \mathrm{age} + \beta_{\mathrm{sex}} + \varepsilon_Y$. Both equations are weighted linear regressions, with $A$, $M$ and $Y$ the standardized hip circumference, triglycerides and thigh-to-ankle PWV, respectively. With $A = -0.5$ versus $+0.5$ (a difference of 1 SD) as the contrast, $m_0$ and $m_1$ are the predicted values of $M$ at $A = -0.5$ and $+0.5$, respectively, and $\hat{Y}(a, m)$ is the predicted value of $Y$ at $A = a$, $M = m$: $\mathrm{DE} = \hat{Y}(+0.5, m_0) - \hat{Y}(-0.5, m_0)$, $\mathrm{IE} = \hat{Y}(+0.5, m_1) - \hat{Y}(+0.5, m_0)$, $\mathrm{TE} = \mathrm{DE} + \mathrm{IE}$, each averaged with weights over the sample covariate distribution; also reported are $\gamma_A$, the $M \to Y$ slope $\beta_M - 0.5 \cdot \beta_{AM}$ at $A = -0.5$, and the interaction coefficient $\beta_{AM}$.

\textit{Intervals.} Bootstrap 95\% percentile intervals.
{\footnotesize\setlength{\tabcolsep}{3pt}
\begin{longtable}{@{}>{\raggedright\arraybackslash}p{0.15\linewidth}>{\raggedright\arraybackslash}p{0.40\linewidth}>{\raggedleft\arraybackslash}p{0.14\linewidth}>{\raggedright\arraybackslash}p{0.24\linewidth}@{}}
\toprule
Block & Parameter & Estimate & Interval \\
\midrule\endhead
Path association decomposition & Direct path association DE: PWV difference corresponding to a 1 SD difference in hip circumference with triglycerides fixed at $m_0$ & $0.08811$ & $[0.02347$, $0.1569]$ \\
 & Indirect path association IE: PWV difference corresponding to a change in triglycerides from $m_0$ to $m_1$ with hip circumference fixed at $+0.5$ SD & $0.02295$ & $[0.009727$, $0.0381]$ \\
 & Decomposed total association $\mathrm{TE} = \mathrm{DE} + \mathrm{IE}$ & $0.1111$ & $[0.04553$, $0.1779]$ \\
Path components & $\gamma_A$: standardized slope of hip circumference $\to$ triglycerides ($A \to M$) & $0.193$ & $[0.1219$, $0.2659]$ \\
 & $\beta_M - 0.5 \cdot \beta_{AM}$: standardized slope of triglycerides $\to$ PWV at hip circumference $-0.5$ SD ($M \to Y \mid A$) & $0.1272$ & $[0.06277$, $0.1963]$ \\
 & $\beta_{AM}$: hip circumference $\times$ triglycerides interaction coefficient in the outcome equation & $-0.008204$ & $[-0.06925$, $0.05715]$ \\
\bottomrule
\end{longtable}}

\paragraph{2.\ Single-mediator path: hip circumference $\to$ seated diastolic pressure $\to$ thigh-to-ankle PWV}
\textit{Design.} Hip circumference, seated blood pressure and PWV are measured in strict date order: blood pressure 30--730 days after hip circumference and PWV 90--1095 days after blood pressure (bounds inclusive); all eligible measurement chains of a participant are included and weighted to give each participant equal weight, $n = 8$. Adjustment is for approximate age at the hip circumference measurement and sex. Diastolic pressure is the mean of the available values of two seated readings, and PWV the mean of the available sides. Continuous variables are standardized within the analysis sample, with sex as a categorical term.

\textit{Model.} $M = \gamma_0 + \gamma_A \cdot A + \gamma_{\mathrm{age}} \cdot \mathrm{age} + \gamma_{\mathrm{sex}} + \varepsilon_M$; $Y = \beta_0 + \beta_A \cdot A + \beta_M \cdot M + \beta_{AM} \cdot A \cdot M + \beta_{\mathrm{age}} \cdot \mathrm{age} + \beta_{\mathrm{sex}} + \varepsilon_Y$. Both equations are weighted linear regressions, with $A$, $M$ and $Y$ the standardized hip circumference, seated diastolic pressure and thigh-to-ankle PWV, respectively. With $A = -0.5$ versus $+0.5$ (a difference of 1 SD) as the contrast, $m_0$ and $m_1$ are the predicted values of $M$ at $A = -0.5$ and $+0.5$, respectively, and $\hat{Y}(a, m)$ is the predicted value of $Y$ at $A = a$, $M = m$: $\mathrm{DE} = \hat{Y}(+0.5, m_0) - \hat{Y}(-0.5, m_0)$, $\mathrm{IE} = \hat{Y}(+0.5, m_1) - \hat{Y}(+0.5, m_0)$, $\mathrm{TE} = \mathrm{DE} + \mathrm{IE}$, each averaged with weights over the sample covariate distribution; also reported are $\gamma_A$, the $M \to Y$ slope $\beta_M - 0.5 \cdot \beta_{AM}$ at $A = -0.5$, and the interaction coefficient $\beta_{AM}$.

\textit{Intervals.} Bootstrap 95\% percentile intervals.
{\footnotesize\setlength{\tabcolsep}{3pt}
\begin{longtable}{@{}>{\raggedright\arraybackslash}p{0.15\linewidth}>{\raggedright\arraybackslash}p{0.40\linewidth}>{\raggedleft\arraybackslash}p{0.14\linewidth}>{\raggedright\arraybackslash}p{0.24\linewidth}@{}}
\toprule
Block & Parameter & Estimate & Interval \\
\midrule\endhead
Path association decomposition & Direct path association DE: PWV difference corresponding to a 1 SD difference in hip circumference with diastolic pressure fixed at $m_0$ & $0.5889$ & $[-3.598$, $30.1]$ \\
 & Indirect path association IE: PWV difference corresponding to a change in diastolic pressure from $m_0$ to $m_1$ with hip circumference fixed at $+0.5$ SD & $-0.02402$ & $[-29.6$, $3.554]$ \\
 & Decomposed total association $\mathrm{TE} = \mathrm{DE} + \mathrm{IE}$ & $0.5649$ & $[-12.43$, $3.232]$ \\
Path components & $\gamma_A$: standardized slope of hip circumference $\to$ diastolic pressure ($A \to M$) & $-0.03084$ & $[-4.948$, $1.994]$ \\
 & $\beta_M - 0.5 \cdot \beta_{AM}$: standardized slope of diastolic pressure $\to$ PWV at hip circumference $-0.5$ SD ($M \to Y \mid A$) & $0.6638$ & $[-2.24$, $22.72]$ \\
 & $\beta_{AM}$: hip circumference $\times$ diastolic pressure interaction coefficient in the outcome equation & $0.1149$ & $[-16.27$, $13.88]$ \\
\bottomrule
\end{longtable}}

\paragraph{3.\ Serial path: hip circumference $\to$ triglycerides $\to$ seated diastolic pressure $\to$ PWV}
\textit{Design.} Each hip circumference measurement day starts one chain, and a participant may contribute several chains: triglycerides 1--365 days after hip circumference, seated blood pressure within 180 days before or after triglycerides (same day allowed), and PWV 1--730 days after seated blood pressure (all bounds inclusive). Diastolic pressure is the mean of the available values of two seated readings, and PWV the mean of the available sides. The three equations share one sample, with adjustment for approximate age on the hip circumference measurement day, recorded sex and study membership; the four measurements are standardized across all eligible chains, and age within the analysis sample.

\textit{Model.} $M_1 = \alpha_1 + a_1 A + \gamma_1^{\top} C + \varepsilon_1$; $M_2 = \alpha_2 + a_2 A + d_{21} M_1 + \gamma_2^{\top} C + \varepsilon_2$; $Y = \alpha_3 + c' A + b_1 M_1 + b_2 M_2 + \gamma_3^{\top} C + \varepsilon_3$. All three are linear regressions, with $A$, $M_1$, $M_2$ and $Y$ the standardized hip circumference, triglycerides, seated diastolic pressure and thigh-to-ankle PWV, and $C$ age, sex and study membership. The coefficients $a_1$, $a_2$, $d_{21}$, $c'$, $b_1$ and $b_2$ are standardized slopes with the other terms of the same equation held fixed; the serial path product is $\theta = a_1 \cdot d_{21} \cdot b_2$.

\textit{Intervals.} $a_1$, $a_2$, $d_{21}$, $c'$, $b_1$, $b_2$: 95\% confidence intervals; $\theta$: bootstrap 95\% percentile interval. $P$ values are not adjusted for multiple comparisons.
{\footnotesize\setlength{\tabcolsep}{3pt}
\begin{longtable}{@{}>{\raggedright\arraybackslash}p{0.14\linewidth}>{\raggedright\arraybackslash}p{0.29\linewidth}>{\raggedleft\arraybackslash}p{0.13\linewidth}>{\raggedright\arraybackslash}p{0.22\linewidth}>{\raggedleft\arraybackslash}p{0.14\linewidth}@{}}
\toprule
Block & Parameter & Estimate & Interval & $P$ \\
\midrule\endhead
Path components & $a_1$: standardized slope of hip circumference $\to$ triglycerides ($A \to M_1$) & $0.1953$ & $[-0.2838$, $0.6744]$ & \mbox{$0.413$} \\
 & $a_2$: standardized slope of hip circumference $\to$ seated diastolic pressure, adjusted for triglycerides ($A \to M_2 \mid M_1$) & $0.0791$ & $[-0.7415$, $0.8997]$ & \mbox{$0.8457$} \\
 & $d_{21}$: standardized slope of triglycerides $\to$ seated diastolic pressure, adjusted for hip circumference ($M_1 \to M_2 \mid A$) & $0.05509$ & $[-0.7547$, $0.8649]$ & \mbox{$0.8908$} \\
 & $c'$: standardized slope of hip circumference $\to$ PWV, adjusted for both mediators ($A \to Y \mid M_1, M_2$) & $0.1639$ & $[-0.1948$, $0.5225]$ & \mbox{$0.3593$} \\
 & $b_1$: standardized slope of triglycerides $\to$ PWV, adjusted for hip circumference and diastolic pressure ($M_1 \to Y \mid A, M_2$) & $-0.2011$ & $[-0.6197$, $0.2176]$ & \mbox{$0.3356$} \\
 & $b_2$: standardized slope of seated diastolic pressure $\to$ PWV, adjusted for hip circumference and triglycerides ($M_2 \to Y \mid A, M_1$) & $0.4458$ & $[0.06639$, $0.8252]$ & \mbox{$0.02269$} \\
Path index & $\theta = a_1 \cdot d_{21} \cdot b_2$: serial path product association index & $0.004796$ & $[-0.03463$, $0.179]$ &  \\
\bottomrule
\end{longtable}}

\paragraph{4.\ Exposure--outcome association: hip circumference and thigh-to-ankle PWV ($\pm 180$ days)}
\textit{Design.} Hip circumference and PWV are measured in the same period: each PWV measurement day forms one row, taking the nearest hip circumference within 180 days before or after that day (bounds inclusive); a participant may contribute several rows, $n = 9385$. PWV is the mean of the available values of the left and right sides. Adjustment is for approximate age (PWV measurement year minus birth year) and recorded sex (categorical term); continuous variables are kept on their original scale.

\textit{Model.} $Y = \beta_0 + \beta_A \cdot A + \beta_{\mathrm{age}} \cdot \mathrm{age} + \gamma_{\mathrm{sex}} + \varepsilon$, linear regression, with $A$ hip circumference and $Y$ thigh-to-ankle PWV (original scale). $\beta_A$ is the PWV difference per 1 original unit higher hip circumference with age and sex held fixed.

\textit{Intervals.} 95\% confidence intervals.
{\footnotesize\setlength{\tabcolsep}{3pt}
\begin{longtable}{@{}>{\raggedright\arraybackslash}p{0.43\linewidth}>{\raggedleft\arraybackslash}p{0.13\linewidth}>{\raggedright\arraybackslash}p{0.22\linewidth}>{\raggedleft\arraybackslash}p{0.14\linewidth}@{}}
\toprule
Parameter & Estimate & Interval & $P$ \\
\midrule\endhead
$\beta_A$: adjusted hip circumference--PWV association coefficient (original scale) & $0.0254$ & $[0.02174$, $0.02906]$ & \mbox{$9.689\times10^{-42}$} \\
\bottomrule
\end{longtable}}

\subsubsection*{Adjacent segments and pairwise associations}

\paragraph{5.\ Serial segment: hip circumference $\to$ triglycerides $\to$ seated diastolic pressure}
\textit{Design.} Hip circumference, triglycerides and seated blood pressure are measured in strict date order: triglycerides 30--730 days after hip circumference and blood pressure 30--730 days after triglycerides (bounds inclusive); all eligible measurement chains of a participant are included and weighted to give each participant equal weight. Adjustment is for approximate age at the hip circumference measurement and sex. Diastolic pressure is the mean of the available values of two seated readings. Continuous variables are standardized within the analysis sample, with sex as a categorical term.

\textit{Model.} $M = \gamma_0 + \gamma_A \cdot A + \gamma_{\mathrm{age}} \cdot \mathrm{age} + \gamma_{\mathrm{sex}} + \varepsilon_M$; $Y = \beta_0 + \beta_A \cdot A + \beta_M \cdot M + \beta_{AM} \cdot A \cdot M + \beta_{\mathrm{age}} \cdot \mathrm{age} + \beta_{\mathrm{sex}} + \varepsilon_Y$. Both equations are weighted linear regressions, with $A$, $M$ and $Y$ the standardized hip circumference, triglycerides and seated diastolic pressure, respectively. With $A = -0.5$ versus $+0.5$ (a difference of 1 SD) as the contrast, $m_0$ and $m_1$ are the predicted values of $M$ at $A = -0.5$ and $+0.5$, respectively, and $\hat{Y}(a, m)$ is the predicted value of $Y$ at $A = a$, $M = m$: $\mathrm{DE} = \hat{Y}(+0.5, m_0) - \hat{Y}(-0.5, m_0)$, $\mathrm{IE} = \hat{Y}(+0.5, m_1) - \hat{Y}(+0.5, m_0)$, $\mathrm{TE} = \mathrm{DE} + \mathrm{IE}$, each averaged with weights over the sample covariate distribution; also reported are $\gamma_A$, the $M \to Y$ slope $\beta_M - 0.5 \cdot \beta_{AM}$ at $A = -0.5$, and the interaction coefficient $\beta_{AM}$.

\textit{Intervals.} Bootstrap 95\% percentile intervals.
{\footnotesize\setlength{\tabcolsep}{3pt}
\begin{longtable}{@{}>{\raggedright\arraybackslash}p{0.15\linewidth}>{\raggedright\arraybackslash}p{0.40\linewidth}>{\raggedleft\arraybackslash}p{0.14\linewidth}>{\raggedright\arraybackslash}p{0.24\linewidth}@{}}
\toprule
Block & Parameter & Estimate & Interval \\
\midrule\endhead
Path association decomposition & Direct path association DE: diastolic pressure difference corresponding to a 1 SD difference in hip circumference with triglycerides fixed at $m_0$ & $0.1878$ & $[0.1113$, $0.2661]$ \\
 & Indirect path association IE: diastolic pressure difference corresponding to a change in triglycerides from $m_0$ to $m_1$ with hip circumference fixed at $+0.5$ SD & $0.03541$ & $[0.01888$, $0.05435]$ \\
 & Decomposed total association $\mathrm{TE} = \mathrm{DE} + \mathrm{IE}$ & $0.2232$ & $[0.15$, $0.2987]$ \\
Path components & $\gamma_A$: standardized slope of hip circumference $\to$ triglycerides ($A \to M$) & $0.2038$ & $[0.1384$, $0.2671]$ \\
 & $\beta_M - 0.5 \cdot \beta_{AM}$: standardized slope of triglycerides $\to$ diastolic pressure at hip circumference $-0.5$ SD ($M \to Y \mid A$) & $0.1599$ & $[0.08353$, $0.2611]$ \\
 & $\beta_{AM}$: hip circumference $\times$ triglycerides interaction coefficient in the outcome equation & $0.01384$ & $[-0.07324$, $0.08038]$ \\
\bottomrule
\end{longtable}}

\paragraph{6.\ Serial segment: triglycerides $\to$ seated diastolic pressure $\to$ PWV}
\textit{Design.} Triglycerides, seated blood pressure and PWV are measured in strict date order: blood pressure 30--730 days after triglycerides and PWV 90--1095 days after blood pressure (bounds inclusive); all eligible measurement chains of a participant are included and weighted to give each participant equal weight. Adjustment is for approximate age at the triglyceride test and sex. Diastolic pressure is the mean of the available values of two seated readings, and PWV the mean of the available sides. Continuous variables are standardized within the analysis sample, with sex as a categorical term.

\textit{Model.} $M = \gamma_0 + \gamma_A \cdot A + \gamma_{\mathrm{age}} \cdot \mathrm{age} + \gamma_{\mathrm{sex}} + \varepsilon_M$; $Y = \beta_0 + \beta_A \cdot A + \beta_M \cdot M + \beta_{AM} \cdot A \cdot M + \beta_{\mathrm{age}} \cdot \mathrm{age} + \beta_{\mathrm{sex}} + \varepsilon_Y$. Both equations are weighted linear regressions, with $A$, $M$ and $Y$ the standardized triglycerides, seated diastolic pressure and thigh-to-ankle PWV, respectively. With $A = -0.5$ versus $+0.5$ (a difference of 1 SD) as the contrast, $m_0$ and $m_1$ are the predicted values of $M$ at $A = -0.5$ and $+0.5$, respectively, and $\hat{Y}(a, m)$ is the predicted value of $Y$ at $A = a$, $M = m$: $\mathrm{DE} = \hat{Y}(+0.5, m_0) - \hat{Y}(-0.5, m_0)$, $\mathrm{IE} = \hat{Y}(+0.5, m_1) - \hat{Y}(+0.5, m_0)$, $\mathrm{TE} = \mathrm{DE} + \mathrm{IE}$, each averaged with weights over the sample covariate distribution; also reported are $\gamma_A$, the $M \to Y$ slope $\beta_M - 0.5 \cdot \beta_{AM}$ at $A = -0.5$, and the interaction coefficient $\beta_{AM}$.

\textit{Intervals.} Bootstrap 95\% percentile intervals.
{\footnotesize\setlength{\tabcolsep}{3pt}
\begin{longtable}{@{}>{\raggedright\arraybackslash}p{0.15\linewidth}>{\raggedright\arraybackslash}p{0.40\linewidth}>{\raggedleft\arraybackslash}p{0.14\linewidth}>{\raggedright\arraybackslash}p{0.24\linewidth}@{}}
\toprule
Block & Parameter & Estimate & Interval \\
\midrule\endhead
Path association decomposition & Direct path association DE: PWV difference corresponding to a 1 SD difference in triglycerides with diastolic pressure fixed at $m_0$ & $0.04516$ & $[0.00304$, $0.08829]$ \\
 & Indirect path association IE: PWV difference corresponding to a change in diastolic pressure from $m_0$ to $m_1$ with triglycerides fixed at $+0.5$ SD & $0.07377$ & $[0.05691$, $0.09393]$ \\
 & Decomposed total association $\mathrm{TE} = \mathrm{DE} + \mathrm{IE}$ & $0.1189$ & $[0.07573$, $0.1607]$ \\
Path components & $\gamma_A$: standardized slope of triglycerides $\to$ diastolic pressure ($A \to M$) & $0.2257$ & $[0.1816$, $0.2757]$ \\
 & $\beta_M - 0.5 \cdot \beta_{AM}$: standardized slope of diastolic pressure $\to$ PWV at triglycerides $-0.5$ SD ($M \to Y \mid A$) & $0.3584$ & $[0.3116$, $0.408]$ \\
 & $\beta_{AM}$: triglycerides $\times$ diastolic pressure interaction coefficient in the outcome equation & $-0.03157$ & $[-0.0792$, $0.02046]$ \\
\bottomrule
\end{longtable}}

\paragraph{7.\ Pairwise association: hip circumference $\to$ triglycerides (1--365 days)}
\textit{Design.} Triglycerides are measured 1--365 days after the hip circumference measurement (bounds inclusive); each hip circumference measurement day is paired with the earliest triglyceride measurement in the window, and a participant may contribute several pairs. Adjustment is for approximate age on the hip circumference measurement day, recorded sex and study membership, the latter two as categorical terms; hip circumference and triglycerides are standardized across all pairs satisfying the window, and age within the analysis sample.

\textit{Model.} $M_{1,z} = \beta_0 + \beta \cdot A_z + \beta_{\mathrm{age}} \cdot \mathrm{Age}_z + \gamma_{\mathrm{Sex}} + \gamma_{\mathrm{Study}} + \varepsilon$ (linear regression; $A$ is hip circumference and $M_1$ triglycerides). $\beta$ is the SD difference in triglycerides per 1 SD higher hip circumference.

\textit{Intervals.} 95\% confidence intervals.
{\footnotesize\setlength{\tabcolsep}{3pt}
\begin{longtable}{@{}>{\raggedright\arraybackslash}p{0.14\linewidth}>{\raggedright\arraybackslash}p{0.29\linewidth}>{\raggedleft\arraybackslash}p{0.13\linewidth}>{\raggedright\arraybackslash}p{0.22\linewidth}>{\raggedleft\arraybackslash}p{0.14\linewidth}@{}}
\toprule
Block & Parameter & Estimate & Interval & $P$ \\
\midrule\endhead
Forward window 1--365 days & Adjusted standardized slope $\beta$ (hip circumference $\to$ triglycerides) & $0.2421$ & $[0.1701$, $0.314]$ & \mbox{$7.947\times10^{-11}$} \\
\bottomrule
\end{longtable}}

\paragraph{8.\ Pairwise association: triglycerides $\to$ seated diastolic pressure (1--365 days)}
\textit{Design.} Seated blood pressure is measured 1--365 days after the triglyceride measurement (bounds inclusive); each triglyceride measurement day is paired with the earliest seated blood pressure measurement in the window, and a participant may contribute several pairs, $n = 8784$. Diastolic pressure is the mean of the available values of two seated readings. Adjustment is for approximate age on the triglyceride measurement day, recorded sex and study membership, the latter two as categorical terms; triglycerides and diastolic pressure are standardized across all pairs satisfying the window, and age within the analysis sample.

\textit{Model.} $M_{2,z} = \beta_0 + \beta \cdot M_{1,z} + \beta_{\mathrm{age}} \cdot \mathrm{Age}_z + \gamma_{\mathrm{Sex}} + \gamma_{\mathrm{Study}} + \varepsilon$ (linear regression; $M_1$ is triglycerides and $M_2$ seated diastolic pressure). $\beta$ is the SD difference in seated diastolic pressure per 1 SD higher triglycerides.

\textit{Intervals.} 95\% confidence intervals.
{\footnotesize\setlength{\tabcolsep}{3pt}
\begin{longtable}{@{}>{\raggedright\arraybackslash}p{0.14\linewidth}>{\raggedright\arraybackslash}p{0.29\linewidth}>{\raggedleft\arraybackslash}p{0.13\linewidth}>{\raggedright\arraybackslash}p{0.22\linewidth}>{\raggedleft\arraybackslash}p{0.14\linewidth}@{}}
\toprule
Block & Parameter & Estimate & Interval & $P$ \\
\midrule\endhead
Forward window 1--365 days & Adjusted standardized slope $\beta$ (triglycerides $\to$ seated diastolic pressure) & $0.19$ & $[0.1091$, $0.2709]$ & \mbox{$4.243\times10^{-6}$} \\
\bottomrule
\end{longtable}}

\paragraph{9.\ Pairwise association: seated diastolic pressure $\to$ PWV (1--730 days)}
\textit{Design.} PWV is measured 1--730 days after the seated blood pressure measurement (bounds inclusive); each blood pressure measurement day is paired with the earliest PWV measurement in the window, and a participant may contribute several pairs, $n = 504$. Diastolic pressure is the mean of the available values of two seated readings, and PWV the mean of the available sides. Adjustment is for approximate age on the blood pressure measurement day, recorded sex and study membership, the latter two as categorical terms; diastolic pressure and PWV are standardized across all pairs satisfying the window, and age within the analysis sample.

\textit{Model.} $Y_z = \beta_0 + \beta \cdot M_{2,z} + \beta_{\mathrm{age}} \cdot \mathrm{Age}_z + \gamma_{\mathrm{Sex}} + \gamma_{\mathrm{Study}} + \varepsilon$ (linear regression; $M_2$ is seated diastolic pressure and $Y$ thigh-to-ankle PWV). $\beta$ is the SD difference in PWV per 1 SD higher seated diastolic pressure.

\textit{Intervals.} 95\% confidence intervals.
{\footnotesize\setlength{\tabcolsep}{3pt}
\begin{longtable}{@{}>{\raggedright\arraybackslash}p{0.14\linewidth}>{\raggedright\arraybackslash}p{0.29\linewidth}>{\raggedleft\arraybackslash}p{0.13\linewidth}>{\raggedright\arraybackslash}p{0.22\linewidth}>{\raggedleft\arraybackslash}p{0.14\linewidth}@{}}
\toprule
Block & Parameter & Estimate & Interval & $P$ \\
\midrule\endhead
Forward window 1--730 days & Adjusted standardized slope $\beta$ (seated diastolic pressure $\to$ thigh-to-ankle PWV) & $0.3717$ & $[0.2963$, $0.4471]$ & \mbox{$1.93\times10^{-20}$} \\
\bottomrule
\end{longtable}}

\paragraph{10.\ Pairwise association: hip circumference--triglycerides ($\pm 180$ days)}
\textit{Design.} Hip circumference and blood triglycerides are measured in the same period: each triglyceride test day forms one row, taking the nearest hip circumference within 180 days before or after that day (bounds inclusive); a participant may contribute several rows. Adjustment is for approximate age (test year minus birth year) and recorded sex (categorical term); continuous variables are kept on their original scale.

\textit{Model.} $M_1 = \beta_0 + \beta_A \cdot A + \beta_{\mathrm{age}} \cdot \mathrm{age} + \gamma_{\mathrm{sex}} + \varepsilon$, linear regression, with $A$ hip circumference and $M_1$ blood triglycerides (original scale). $\beta_A$ is the triglyceride difference per 1 original unit higher hip circumference with age and sex held fixed.

\textit{Intervals.} 95\% confidence intervals.
{\footnotesize\setlength{\tabcolsep}{3pt}
\begin{longtable}{@{}>{\raggedright\arraybackslash}p{0.43\linewidth}>{\raggedleft\arraybackslash}p{0.13\linewidth}>{\raggedright\arraybackslash}p{0.22\linewidth}>{\raggedleft\arraybackslash}p{0.14\linewidth}@{}}
\toprule
Parameter & Estimate & Interval & $P$ \\
\midrule\endhead
$\beta_A$: adjusted hip circumference--triglycerides association coefficient (original scale) & $1.772$ & $[1.495$, $2.049]$ & \mbox{$1.744\times10^{-35}$} \\
\bottomrule
\end{longtable}}

\paragraph{11.\ Pairwise association: hip circumference--seated diastolic pressure ($\pm 180$ days)}
\textit{Design.} Hip circumference and seated blood pressure are measured in the same period: each blood pressure measurement day forms one row, taking the nearest hip circumference within 180 days before or after that day (bounds inclusive); a participant may contribute several rows, $n = 12922$. Diastolic pressure is the mean of the available values of two seated readings. Adjustment is for approximate age (blood pressure measurement year minus birth year) and recorded sex (categorical term); continuous variables are kept on their original scale.

\textit{Model.} $M_2 = \beta_0 + \beta_A \cdot A + \beta_{\mathrm{age}} \cdot \mathrm{age} + \gamma_{\mathrm{sex}} + \varepsilon$, linear regression, with $A$ hip circumference and $M_2$ seated diastolic pressure (original scale). $\beta_A$ is the diastolic pressure difference per 1 original unit higher hip circumference with age and sex held fixed.

\textit{Intervals.} 95\% confidence intervals.
{\footnotesize\setlength{\tabcolsep}{3pt}
\begin{longtable}{@{}>{\raggedright\arraybackslash}p{0.43\linewidth}>{\raggedleft\arraybackslash}p{0.13\linewidth}>{\raggedright\arraybackslash}p{0.22\linewidth}>{\raggedleft\arraybackslash}p{0.14\linewidth}@{}}
\toprule
Parameter & Estimate & Interval & $P$ \\
\midrule\endhead
$\beta_A$: adjusted hip circumference--seated diastolic pressure association coefficient (original scale) & $0.3037$ & $[0.283$, $0.3245]$ & \mbox{$2.347\times10^{-174}$} \\
\bottomrule
\end{longtable}}

\paragraph{12.\ Pairwise association: triglycerides--PWV ($\pm 180$ days)}
\textit{Design.} Blood triglycerides and PWV are measured in the same period: each PWV measurement day forms one row, taking the nearest triglyceride test within 180 days before or after that day (bounds inclusive); a participant may contribute several rows. PWV is the mean of the available values of the left and right sides. Adjustment is for approximate age (PWV measurement year minus birth year) and recorded sex (categorical term); continuous variables are kept on their original scale.

\textit{Model.} $Y = \beta_0 + \beta_{M_1} \cdot M_1 + \beta_{\mathrm{age}} \cdot \mathrm{age} + \gamma_{\mathrm{sex}} + \varepsilon$, linear regression, with $M_1$ blood triglycerides and $Y$ thigh-to-ankle PWV (original scale). $\beta_{M_1}$ is the PWV difference per 1 original unit higher triglycerides with age and sex held fixed.

\textit{Intervals.} 95\% confidence intervals.
{\footnotesize\setlength{\tabcolsep}{3pt}
\begin{longtable}{@{}>{\raggedright\arraybackslash}p{0.43\linewidth}>{\raggedleft\arraybackslash}p{0.13\linewidth}>{\raggedright\arraybackslash}p{0.22\linewidth}>{\raggedleft\arraybackslash}p{0.14\linewidth}@{}}
\toprule
Parameter & Estimate & Interval & $P$ \\
\midrule\endhead
$\beta_{M_1}$: adjusted triglycerides--PWV association coefficient (original scale) & $0.003439$ & $[0.002708$, $0.004169]$ & \mbox{$4.103\times10^{-20}$} \\
\bottomrule
\end{longtable}}

\paragraph{13.\ Pairwise association: seated diastolic pressure--PWV ($\pm 180$ days)}
\textit{Design.} Seated blood pressure and PWV are measured in the same period: each PWV measurement day forms one row, taking the nearest blood pressure measurement within 180 days before or after that day (bounds inclusive); a participant may contribute several rows, $n = 9401$. Diastolic pressure is the mean of the available values of two seated readings, and PWV the mean of the available values of the left and right sides. Adjustment is for approximate age (PWV measurement year minus birth year) and recorded sex (categorical term); continuous variables are kept on their original scale.

\textit{Model.} $Y = \beta_0 + \beta_{M_2} \cdot M_2 + \beta_{\mathrm{age}} \cdot \mathrm{age} + \gamma_{\mathrm{sex}} + \varepsilon$, linear regression, with $M_2$ seated diastolic pressure and $Y$ thigh-to-ankle PWV (original scale). $\beta_{M_2}$ is the PWV difference per 1 original unit higher diastolic pressure with age and sex held fixed.

\textit{Intervals.} 95\% confidence intervals.
{\footnotesize\setlength{\tabcolsep}{3pt}
\begin{longtable}{@{}>{\raggedright\arraybackslash}p{0.43\linewidth}>{\raggedleft\arraybackslash}p{0.13\linewidth}>{\raggedright\arraybackslash}p{0.22\linewidth}>{\raggedleft\arraybackslash}p{0.14\linewidth}@{}}
\toprule
Parameter & Estimate & Interval & $P$ \\
\midrule\endhead
$\beta_{M_2}$: adjusted seated diastolic pressure--PWV association coefficient (original scale) & $0.07881$ & $[0.07575$, $0.08186]$ & \mbox{$0$} \\
\bottomrule
\end{longtable}}

\subsubsection*{Joint models}

\paragraph{14.\ Joint path: hip circumference $\to$ triglycerides $\to$ diastolic pressure $\to$ PWV ($\pm 180$ days)}
\textit{Design.} The four measurements are taken in the same period: each PWV measurement day forms one row, with hip circumference, blood triglycerides and seated blood pressure each taken as the nearest measurement within 180 days before or after that day (bounds inclusive), order unrestricted; a participant may contribute several rows, $n = 4846$. The four equations share one sample, all adjusted for approximate age (PWV measurement year minus birth year) and recorded sex (categorical term). Diastolic pressure is the mean of the available seated readings and PWV the mean of the available values of both sides; continuous variables are kept on their original scale.

\textit{Model.} $Y = c_0 + c \cdot A + \gamma_0^{\top} C + \varepsilon_0$; $M_1 = \alpha_1 + a_1 A + \gamma_1^{\top} C + \varepsilon_1$; $M_2 = \alpha_2 + a_2 A + d_{21} M_1 + \gamma_2^{\top} C + \varepsilon_2$; $Y = \alpha_3 + c' A + b_1 M_1 + b_2 M_2 + \gamma_3^{\top} C + \varepsilon_3$. All four are linear regressions, with $A$, $M_1$, $M_2$ and $Y$ hip circumference, blood triglycerides, seated diastolic pressure and thigh-to-ankle PWV (original scale), and $C$ age and sex. Each coefficient is the difference in the equation's outcome per 1 original unit higher value of that variable with the other terms of the same equation held fixed; $c$ is the hip circumference coefficient adjusted for $C$ only, and $\Delta = c - c'$.

\textit{Intervals.} 95\% confidence intervals; $P$ values are not adjusted for multiple comparisons.
{\footnotesize\setlength{\tabcolsep}{3pt}
\begin{longtable}{@{}>{\raggedright\arraybackslash}p{0.14\linewidth}>{\raggedright\arraybackslash}p{0.29\linewidth}>{\raggedleft\arraybackslash}p{0.13\linewidth}>{\raggedright\arraybackslash}p{0.22\linewidth}>{\raggedleft\arraybackslash}p{0.14\linewidth}@{}}
\toprule
Block & Parameter & Estimate & Interval & $P$ \\
\midrule\endhead
Path coefficients & $a_1$: hip circumference $\to$ triglycerides ($A \to M_1$) & $1.757$ & $[1.536$, $1.978]$ & \mbox{$2.375\times10^{-53}$} \\
 & $a_2$: hip circumference $\to$ seated diastolic pressure, adjusted for triglycerides ($A \to M_2 \mid M_1$) & $0.273$ & $[0.2401$, $0.3059]$ & \mbox{$7.117\times10^{-58}$} \\
 & $d_{21}$: triglycerides $\to$ seated diastolic pressure, adjusted for hip circumference ($M_1 \to M_2 \mid A$) & $0.02684$ & $[0.02187$, $0.0318]$ & \mbox{$6.772\times10^{-26}$} \\
 & $c'$: hip circumference $\to$ PWV, adjusted for both mediators ($A \to Y \mid M_1, M_2$) & $-0.004061$ & $[-0.009239$, $0.001117]$ & \mbox{$0.1242$} \\
 & $b_1$: triglycerides $\to$ PWV, adjusted for hip circumference and diastolic pressure ($M_1 \to Y \mid A, M_2$) & $0.000879$ & $[0.0002241$, $0.001534]$ & \mbox{$0.008532$} \\
 & $b_2$: seated diastolic pressure $\to$ PWV, adjusted for hip circumference and triglycerides ($M_2 \to Y \mid A, M_1$) & $0.07639$ & $[0.07139$, $0.08139]$ & \mbox{$1.527\times10^{-180}$} \\
Hip circumference coefficient comparison (same sample) & $c$: hip circumference $\to$ PWV, adjusted for age and sex only & $0.02194$ & $[0.0165$, $0.02737]$ & \mbox{$3.053\times10^{-15}$} \\
 & $c'$: hip circumference $\to$ PWV, adjusted for both mediators & $-0.004061$ & $[-0.009239$, $0.001117]$ & \mbox{$0.1242$} \\
 & $\Delta = c - c'$: signed change in the hip circumference coefficient after adding both mediators & $0.026$ & $[0.02289$, $0.0291]$ & \mbox{$7.041\times10^{-59}$} \\
\bottomrule
\end{longtable}}

\subsubsection*{Reverse direction}

\paragraph{15.\ Reverse path: PWV $\to$ triglycerides $\to$ hip circumference}
\textit{Design.} Endpoint roles reversed: PWV, triglycerides and hip circumference are measured in strict date order (PWV $<$ triglycerides $<$ hip circumference), with triglycerides 30--730 days after PWV and hip circumference 30--730 days after triglycerides (bounds inclusive); all eligible measurement chains of a participant are included and weighted to give each participant equal weight. Adjustment is for approximate age at the PWV measurement and sex. PWV is the mean of the available sides. Continuous variables are standardized within the analysis sample, with sex as a categorical term.

\textit{Model.} $A$ = thigh-to-ankle PWV, $M$ = triglycerides, $Y$ = hip circumference (all standardized values). $M = \gamma_0 + \gamma_A \cdot A + \gamma_{\mathrm{age}} \cdot \mathrm{age} + \gamma_{\mathrm{sex}} + \varepsilon_M$; $Y = \beta_0 + \beta_A \cdot A + \beta_M \cdot M + \beta_{AM} \cdot A \cdot M + \beta_{\mathrm{age}} \cdot \mathrm{age} + \beta_{\mathrm{sex}} + \varepsilon_Y$. Both equations are weighted linear regressions. With $A = -0.5$ versus $+0.5$ (a difference of 1 SD) as the contrast, $m_0$ and $m_1$ are the predicted values of $M$ at $A = -0.5$ and $+0.5$, respectively, and $\hat{Y}(a, m)$ is the predicted value of $Y$ at $A = a$, $M = m$: $\mathrm{DE} = \hat{Y}(+0.5, m_0) - \hat{Y}(-0.5, m_0)$, $\mathrm{IE} = \hat{Y}(+0.5, m_1) - \hat{Y}(+0.5, m_0)$, $\mathrm{TE} = \mathrm{DE} + \mathrm{IE}$, each averaged with weights over the sample covariate distribution; also reported are $\gamma_A$, the $M \to Y$ slope $\beta_M - 0.5 \cdot \beta_{AM}$ at $A = -0.5$, and the interaction coefficient $\beta_{AM}$.

\textit{Intervals.} Bootstrap 95\% percentile intervals.
{\footnotesize\setlength{\tabcolsep}{3pt}
\begin{longtable}{@{}>{\raggedright\arraybackslash}p{0.15\linewidth}>{\raggedright\arraybackslash}p{0.40\linewidth}>{\raggedleft\arraybackslash}p{0.14\linewidth}>{\raggedright\arraybackslash}p{0.24\linewidth}@{}}
\toprule
Block & Parameter & Estimate & Interval \\
\midrule\endhead
Reverse path association decomposition & Direct path association DE: hip circumference difference corresponding to a 1 SD difference in PWV with triglycerides fixed at $m_0$ & $0.03301$ & $[-0.06901$, $0.1341]$ \\
 & Indirect path association IE: hip circumference difference corresponding to a change in triglycerides from $m_0$ to $m_1$ with PWV fixed at $+0.5$ SD & $0.04734$ & $[0.02027$, $0.0877]$ \\
 & Decomposed total association $\mathrm{TE} = \mathrm{DE} + \mathrm{IE}$ & $0.08035$ & $[-0.02412$, $0.1819]$ \\
Reverse path components & $\gamma_A$: standardized slope of PWV $\to$ triglycerides ($A \to M$) & $0.2052$ & $[0.09105$, $0.3434]$ \\
 & $\beta_M - 0.5 \cdot \beta_{AM}$: standardized slope of triglycerides $\to$ hip circumference at PWV $-0.5$ SD ($M \to Y \mid A$) & $0.3$ & $[0.1818$, $0.4426]$ \\
 & $\beta_{AM}$: PWV $\times$ triglycerides interaction coefficient in the outcome equation & $-0.06921$ & $[-0.1833$, $0.07738]$ \\
\bottomrule
\end{longtable}}

\paragraph{16.\ Reverse path: PWV $\to$ seated diastolic pressure $\to$ hip circumference}
\textit{Design.} Endpoint roles reversed: PWV, seated blood pressure and hip circumference are measured in strict date order (PWV $<$ blood pressure $<$ hip circumference), with blood pressure 30--730 days after PWV and hip circumference 30--730 days after blood pressure (bounds inclusive); all eligible measurement chains of a participant are included and weighted to give each participant equal weight. Adjustment is for approximate age at the PWV measurement and sex. Diastolic pressure is the mean of the available values of two seated readings, and PWV the mean of the available sides. Continuous variables are standardized within the analysis sample, with sex as a categorical term.

\textit{Model.} $A$ = thigh-to-ankle PWV, $M$ = seated diastolic pressure, $Y$ = hip circumference (all standardized values). $M = \gamma_0 + \gamma_A \cdot A + \gamma_{\mathrm{age}} \cdot \mathrm{age} + \gamma_{\mathrm{sex}} + \varepsilon_M$; $Y = \beta_0 + \beta_A \cdot A + \beta_M \cdot M + \beta_{AM} \cdot A \cdot M + \beta_{\mathrm{age}} \cdot \mathrm{age} + \beta_{\mathrm{sex}} + \varepsilon_Y$. Both equations are weighted linear regressions. With $A = -0.5$ versus $+0.5$ (a difference of 1 SD) as the contrast, $m_0$ and $m_1$ are the predicted values of $M$ at $A = -0.5$ and $+0.5$, respectively, and $\hat{Y}(a, m)$ is the predicted value of $Y$ at $A = a$, $M = m$: $\mathrm{DE} = \hat{Y}(+0.5, m_0) - \hat{Y}(-0.5, m_0)$, $\mathrm{IE} = \hat{Y}(+0.5, m_1) - \hat{Y}(+0.5, m_0)$, $\mathrm{TE} = \mathrm{DE} + \mathrm{IE}$, each averaged with weights over the sample covariate distribution; also reported are $\gamma_A$, the $M \to Y$ slope $\beta_M - 0.5 \cdot \beta_{AM}$ at $A = -0.5$, and the interaction coefficient $\beta_{AM}$.

\textit{Intervals.} Bootstrap 95\% percentile intervals.
{\footnotesize\setlength{\tabcolsep}{3pt}
\begin{longtable}{@{}>{\raggedright\arraybackslash}p{0.15\linewidth}>{\raggedright\arraybackslash}p{0.40\linewidth}>{\raggedleft\arraybackslash}p{0.14\linewidth}>{\raggedright\arraybackslash}p{0.24\linewidth}@{}}
\toprule
Block & Parameter & Estimate & Interval \\
\midrule\endhead
Reverse path association decomposition & Direct path association DE: hip circumference difference corresponding to a 1 SD difference in PWV with diastolic pressure fixed at $m_0$ & $2.814$ & $[-5.498$, $123.2]$ \\
 & Indirect path association IE: hip circumference difference corresponding to a change in diastolic pressure from $m_0$ to $m_1$ with PWV fixed at $+0.5$ SD & $-2.064$ & $[-119.8$, $8.1]$ \\
 & Decomposed total association $\mathrm{TE} = \mathrm{DE} + \mathrm{IE}$ & $0.7495$ & $[-1.367$, $4.536]$ \\
Reverse path components & $\gamma_A$: standardized slope of PWV $\to$ diastolic pressure ($A \to M$) & $0.9454$ & $[-0.5054$, $2.174]$ \\
 & $\beta_M - 0.5 \cdot \beta_{AM}$: standardized slope of diastolic pressure $\to$ hip circumference at PWV $-0.5$ SD ($M \to Y \mid A$) & $-1.214$ & $[-81.97$, $8.584]$ \\
 & $\beta_{AM}$: PWV $\times$ diastolic pressure interaction coefficient in the outcome equation & $-0.9688$ & $[-6.57$, $3.232]$ \\
\bottomrule
\end{longtable}}

\paragraph{17.\ Reverse pairing: earlier PWV and subsequently measured hip circumference}
\textit{Design.} PWV and hip circumference are measured in strict date order: each PWV measurement day forms one row, taking the nearest hip circumference 365--730 days later (bounds inclusive); a participant may contribute several rows, $n = 434$. PWV is the mean of the available values of the left and right sides. Adjustment is for approximate age (PWV measurement year minus birth year) and recorded sex (categorical term); continuous variables are kept on their original scale.

\textit{Model.} $A = \beta_0 + \beta_Y \cdot Y + \beta_{\mathrm{age}} \cdot \mathrm{age} + \gamma_{\mathrm{sex}} + \varepsilon$, linear regression, with $Y$ the earlier thigh-to-ankle PWV and $A$ the hip circumference measured 365--730 days later (original scale). $\beta_Y$ is the difference in subsequent hip circumference per 1 original unit higher earlier PWV with age and sex held fixed.

\textit{Intervals.} 95\% confidence intervals.
{\footnotesize\setlength{\tabcolsep}{3pt}
\begin{longtable}{@{}>{\raggedright\arraybackslash}p{0.43\linewidth}>{\raggedleft\arraybackslash}p{0.13\linewidth}>{\raggedright\arraybackslash}p{0.22\linewidth}>{\raggedleft\arraybackslash}p{0.14\linewidth}@{}}
\toprule
Parameter & Estimate & Interval & $P$ \\
\midrule\endhead
$\beta_Y$: earlier PWV $\to$ hip circumference measured 365--730 days later (original scale) & $0.3772$ & $[-0.1471$, $0.9015]$ & \mbox{$0.1581$} \\
\bottomrule
\end{longtable}}

\subsubsection*{Heterogeneity and interaction}

\paragraph{18.\ Age modification: hip circumference $\to$ triglycerides $\to$ PWV path}
\textit{Design.} Hip circumference, triglycerides and PWV are measured in strict date order: triglycerides 30--730 days after hip circumference and PWV 90--1095 days after triglycerides (bounds inclusive); all eligible measurement chains of a participant are included and weighted to give each participant equal weight. Approximate age at the hip circumference measurement enters in continuous form as a main effect and in interaction terms with hip circumference, triglycerides and their product, with additional adjustment for sex (categorical term). PWV is the mean of the available sides. Continuous variables are standardized within the analysis sample, with age set to $-1$, $0$ and $+1$ SD.

\textit{Model.} $M = \gamma_0 + \gamma_A \cdot A + \gamma_{\mathrm{age}} \cdot \mathrm{age} + \gamma_{A \cdot \mathrm{age}} \cdot A \cdot \mathrm{age} + \gamma_{\mathrm{sex}} + \varepsilon_M$; $Y = \beta_0 + \beta_A \cdot A + \beta_M \cdot M + \beta_{AM} \cdot A \cdot M + \beta_{\mathrm{age}} \cdot \mathrm{age} + \beta_{A \cdot \mathrm{age}} \cdot A \cdot \mathrm{age} + \beta_{M \cdot \mathrm{age}} \cdot M \cdot \mathrm{age} + \beta_{AM \cdot \mathrm{age}} \cdot A \cdot M \cdot \mathrm{age} + \beta_{\mathrm{sex}} + \varepsilon_Y$. Both equations are weighted linear regressions, with $A$, $M$ and $Y$ the standardized hip circumference, triglycerides and thigh-to-ankle PWV, respectively. With age fixed at $k \in \{-1, 0, +1\}$ SD and sex averaged over the sample distribution, and with $A = -0.5$ versus $+0.5$ (a difference of 1 SD) as the contrast, $\mathrm{DE}(k) = \hat{Y}(+0.5, m_0) - \hat{Y}(-0.5, m_0)$, $\mathrm{IE}(k) = \hat{Y}(+0.5, m_1) - \hat{Y}(+0.5, m_0)$ and $\mathrm{TE}(k) = \mathrm{DE}(k) + \mathrm{IE}(k)$ are computed, where $m_0$ and $m_1$ are the predicted values of $M$ at $A = -0.5$ and $+0.5$, respectively; also reported are the conditional slope $\gamma_A + \gamma_{A \cdot \mathrm{age}} \cdot k$, the conditional $M \to Y$ slope at $A = -0.5$, the conditional interaction coefficient $\beta_{AM} + \beta_{AM \cdot \mathrm{age}} \cdot k$, and the differences in DE, IE and TE between $k = +1$ and $k = -1$.

\textit{Intervals.} Bootstrap 95\% percentile intervals.
{\footnotesize\setlength{\tabcolsep}{3pt}
\begin{longtable}{@{}>{\raggedright\arraybackslash}p{0.15\linewidth}>{\raggedright\arraybackslash}p{0.40\linewidth}>{\raggedleft\arraybackslash}p{0.14\linewidth}>{\raggedright\arraybackslash}p{0.24\linewidth}@{}}
\toprule
Block & Parameter & Estimate & Interval \\
\midrule\endhead
Age = mean $-$ 1 SD ($k = -1$) & $\mathrm{DE}(k)$: direct path association with age fixed at $k$ & $0.09709$ & $[0.002465$, $0.187]$ \\
 & $\mathrm{IE}(k)$: indirect path association with age fixed at $k$ & $0.02315$ & $[0.006662$, $0.04564]$ \\
 & $\mathrm{TE}(k) = \mathrm{DE}(k) + \mathrm{IE}(k)$: decomposed total association with age fixed at $k$ & $0.1202$ & $[0.02517$, $0.2101]$ \\
 & $\gamma_A + \gamma_{A \cdot \mathrm{age}} \cdot k$: conditional standardized slope of hip circumference $\to$ triglycerides ($A \to M$) & $0.1715$ & $[0.07615$, $0.2678]$ \\
 & $\beta_M - 0.5 \cdot \beta_{AM} + (\beta_{M \cdot \mathrm{age}} - 0.5 \cdot \beta_{AM \cdot \mathrm{age}}) \cdot k$: conditional standardized slope of triglycerides $\to$ PWV at hip circumference $-0.5$ SD ($M \to Y \mid A$) & $0.1728$ & $[0.08334$, $0.2725]$ \\
 & $\beta_{AM} + \beta_{AM \cdot \mathrm{age}} \cdot k$: conditional hip circumference $\times$ triglycerides interaction coefficient & $-0.03775$ & $[-0.1305$, $0.05916]$ \\
Age = mean ($k = 0$) & $\mathrm{DE}(k)$: direct path association with age fixed at $k$ & $0.08708$ & $[0.01809$, $0.1566]$ \\
 & $\mathrm{IE}(k)$: indirect path association with age fixed at $k$ & $0.02267$ & $[0.00959$, $0.03958]$ \\
 & $\mathrm{TE}(k) = \mathrm{DE}(k) + \mathrm{IE}(k)$: decomposed total association with age fixed at $k$ & $0.1098$ & $[0.0454$, $0.1781]$ \\
 & $\gamma_A + \gamma_{A \cdot \mathrm{age}} \cdot k$: conditional standardized slope of hip circumference $\to$ triglycerides ($A \to M$) & $0.1954$ & $[0.1216$, $0.2685]$ \\
 & $\beta_M - 0.5 \cdot \beta_{AM} + (\beta_{M \cdot \mathrm{age}} - 0.5 \cdot \beta_{AM \cdot \mathrm{age}}) \cdot k$: conditional standardized slope of triglycerides $\to$ PWV at hip circumference $-0.5$ SD ($M \to Y \mid A$) & $0.1126$ & $[0.04059$, $0.1876]$ \\
 & $\beta_{AM} + \beta_{AM \cdot \mathrm{age}} \cdot k$: conditional hip circumference $\times$ triglycerides interaction coefficient & $0.003397$ & $[-0.05677$, $0.06907]$ \\
Age = mean $+$ 1 SD ($k = +1$) & $\mathrm{DE}(k)$: direct path association with age fixed at $k$ & $0.07652$ & $[-0.03523$, $0.1869]$ \\
 & $\mathrm{IE}(k)$: indirect path association with age fixed at $k$ & $0.02128$ & $[-0.0001705$, $0.04881]$ \\
 & $\mathrm{TE}(k) = \mathrm{DE}(k) + \mathrm{IE}(k)$: decomposed total association with age fixed at $k$ & $0.0978$ & $[-0.003166$, $0.2026]$ \\
 & $\gamma_A + \gamma_{A \cdot \mathrm{age}} \cdot k$: conditional standardized slope of hip circumference $\to$ triglycerides ($A \to M$) & $0.2193$ & $[0.1216$, $0.3239]$ \\
 & $\beta_M - 0.5 \cdot \beta_{AM} + (\beta_{M \cdot \mathrm{age}} - 0.5 \cdot \beta_{AM \cdot \mathrm{age}}) \cdot k$: conditional standardized slope of triglycerides $\to$ PWV at hip circumference $-0.5$ SD ($M \to Y \mid A$) & $0.05249$ & $[-0.07354$, $0.1855]$ \\
 & $\beta_{AM} + \beta_{AM \cdot \mathrm{age}} \cdot k$: conditional hip circumference $\times$ triglycerides interaction coefficient & $0.04454$ & $[-0.06044$, $0.1536]$ \\
Difference between age $+1$ SD and $-1$ SD & $\mathrm{DE}(+1) - \mathrm{DE}(-1)$: age difference in the direct path association & $-0.02057$ & $[-0.1652$, $0.1307]$ \\
 & $\mathrm{IE}(+1) - \mathrm{IE}(-1)$: age difference in the indirect path association & $-0.001874$ & $[-0.03405$, $0.03116]$ \\
 & $\mathrm{TE}(+1) - \mathrm{TE}(-1)$: age difference in the decomposed total association & $-0.02245$ & $[-0.1582$, $0.122]$ \\
\bottomrule
\end{longtable}}

\paragraph{19.\ Age modification: hip circumference $\to$ seated diastolic pressure slope}
\textit{Design.} Hip circumference, seated blood pressure and PWV are measured in strict date order: blood pressure 30--730 days after hip circumference and PWV 90--1095 days after blood pressure (bounds inclusive); all eligible measurement chains of a participant are included and weighted to give each participant equal weight. Approximate age at the hip circumference measurement enters in continuous form as a main effect and in an interaction term with hip circumference, with additional adjustment for sex (categorical term). Diastolic pressure is the mean of the available values of two seated readings. Continuous variables are standardized within the analysis sample, with age set to $-1$, $0$ and $+1$ SD.

\textit{Model.} $M = \gamma_0 + \gamma_A \cdot A + \gamma_{\mathrm{age}} \cdot \mathrm{age} + \gamma_{A \cdot \mathrm{age}} \cdot A \cdot \mathrm{age} + \gamma_{\mathrm{sex}} + \varepsilon_M$ (weighted linear regression; $A$ and $M$ are the standardized hip circumference and seated diastolic pressure, respectively). Reported is the conditional slope $\gamma_A + \gamma_{A \cdot \mathrm{age}} \cdot k$ with age fixed at $k \in \{-1, 0, +1\}$ SD, that is, the SD change in diastolic pressure per 1 SD higher hip circumference at that age level.

\textit{Intervals.} Bootstrap 95\% percentile intervals.
{\footnotesize\setlength{\tabcolsep}{3pt}
\begin{longtable}{@{}>{\raggedright\arraybackslash}p{0.15\linewidth}>{\raggedright\arraybackslash}p{0.40\linewidth}>{\raggedleft\arraybackslash}p{0.14\linewidth}>{\raggedright\arraybackslash}p{0.24\linewidth}@{}}
\toprule
Block & Parameter & Estimate & Interval \\
\midrule\endhead
Age = mean $-$ 1 SD ($k = -1$) & $\gamma_A + \gamma_{A \cdot \mathrm{age}} \cdot k$: conditional standardized slope of hip circumference $\to$ seated diastolic pressure ($A \to M$) & $-0.676$ & $[-36.02$, $2.375]$ \\
Age = mean ($k = 0$) & $\gamma_A + \gamma_{A \cdot \mathrm{age}} \cdot k$: conditional standardized slope of hip circumference $\to$ seated diastolic pressure ($A \to M$) & $-0.2443$ & $[-11.16$, $0.8942]$ \\
Age = mean $+$ 1 SD ($k = +1$) & $\gamma_A + \gamma_{A \cdot \mathrm{age}} \cdot k$: conditional standardized slope of hip circumference $\to$ seated diastolic pressure ($A \to M$) & $0.1873$ & $[-2.026$, $5.346]$ \\
\bottomrule
\end{longtable}}

\paragraph{20.\ Sex heterogeneity: hip circumference $\to$ triglycerides $\to$ diastolic pressure $\to$ PWV}
\textit{Design.} Measurement chains and windows as in the main serial path: triglycerides 1--365 days after hip circumference, seated blood pressure within 180 days before or after triglycerides (same day allowed), and PWV 1--730 days after seated blood pressure (all bounds inclusive); one chain per hip circumference measurement day. Recorded sex (first sex category, second sex category) enters as a main effect and in interaction terms with the upstream variables of the three adjacent edges; additional adjustment is for approximate age on the hip circumference measurement day and study membership. The four measurements are standardized across all eligible chains, and age within the analysis sample.

\textit{Model.} $M_1 = \alpha_1 + a_1(s) \cdot A + \gamma_1^{\top} C + \varepsilon_1$; $M_2 = \alpha_2 + a_2 A + d_{21}(s) \cdot M_1 + \gamma_2^{\top} C + \varepsilon_2$; $Y = \alpha_3 + c' A + b_1 M_1 + b_2(s) \cdot M_2 + \gamma_3^{\top} C + \varepsilon_3$ (pooled-sample linear regressions, each equation with a sex main effect, $s$ the sex category; $A$, $M_1$, $M_2$ and $Y$ are the standardized hip circumference, triglycerides, seated diastolic pressure and thigh-to-ankle PWV, and $C$ age and study membership). The coefficients $a_1(s)$, $d_{21}(s)$ and $b_2(s)$ are the standardized slopes of the adjacent edges within category $s$; the category path product is $\theta(s) = a_1(s) \cdot d_{21}(s) \cdot b_2(s)$.

\textit{Intervals.} $a_1(s)$, $d_{21}(s)$, $b_2(s)$: 95\% confidence intervals; $\theta(s)$: bootstrap 95\% percentile interval. $P$ values are not adjusted for multiple comparisons.
{\footnotesize\setlength{\tabcolsep}{3pt}
\begin{longtable}{@{}>{\raggedright\arraybackslash}p{0.14\linewidth}>{\raggedright\arraybackslash}p{0.29\linewidth}>{\raggedleft\arraybackslash}p{0.13\linewidth}>{\raggedright\arraybackslash}p{0.22\linewidth}>{\raggedleft\arraybackslash}p{0.14\linewidth}@{}}
\toprule
Block & Parameter & Estimate & Interval & $P$ \\
\midrule\endhead
First sex category & $a_1(1)$: standardized slope of hip circumference $\to$ triglycerides within the first sex category & $0.01119$ & $[-1.729$, $1.751]$ & \mbox{$0.9896$} \\
 & $d_{21}(1)$: standardized slope of triglycerides $\to$ seated diastolic pressure within the first sex category, adjusted for hip circumference & $-0.1406$ & $[-1.28$, $0.9989]$ & \mbox{$0.8034$} \\
 & $b_2(1)$: standardized slope of seated diastolic pressure $\to$ PWV within the first sex category, adjusted for hip circumference and triglycerides & $0.2635$ & $[-0.3299$, $0.857]$ & \mbox{$0.3728$} \\
 & $\theta(1) = a_1(1) \cdot d_{21}(1) \cdot b_2(1)$: serial path product association index for the first sex category & $-0.0004146$ & $[-0.07446$, $0.6064]$ &  \\
Second sex category & $a_1(2)$: standardized slope of hip circumference $\to$ triglycerides within the second sex category & $0.2377$ & $[-0.2799$, $0.7553]$ & \mbox{$0.3569$} \\
 & $d_{21}(2)$: standardized slope of triglycerides $\to$ seated diastolic pressure within the second sex category, adjusted for hip circumference & $0.3632$ & $[-0.6662$, $1.393]$ & \mbox{$0.478$} \\
 & $b_2(2)$: standardized slope of seated diastolic pressure $\to$ PWV within the second sex category, adjusted for hip circumference and triglycerides & $0.5914$ & $[0.05517$, $1.128]$ & \mbox{$0.03167$} \\
 & $\theta(2) = a_1(2) \cdot d_{21}(2) \cdot b_2(2)$: serial path product association index for the second sex category & $0.05105$ & $[-0.06286$, $0.228]$ &  \\
\bottomrule
\end{longtable}}

\paragraph{21.\ Sex heterogeneity: hip circumference--PWV slope adjusted for both mediators}
\textit{Design.} The four measurements are taken in the same period: each PWV measurement day forms one row, with hip circumference, blood triglycerides and seated blood pressure each taken as the nearest measurement within 180 days before or after that day (bounds inclusive), order unrestricted; a participant may contribute several rows. The model includes a hip circumference $\times$ recorded sex interaction term, with additional adjustment for triglycerides, seated diastolic pressure and approximate age (PWV measurement year minus birth year); continuous variables are kept on their original scale.

\textit{Model.} $Y = \beta_0 + \beta_A \cdot A + \beta_{M_1} \cdot M_1 + \beta_{M_2} \cdot M_2 + \beta_{\mathrm{age}} \cdot \mathrm{age} + \gamma_s + \delta_s \cdot A + \varepsilon$, linear regression, with $s$ the recorded sex category and $A$, $M_1$, $M_2$ and $Y$ hip circumference, blood triglycerides, seated diastolic pressure and thigh-to-ankle PWV (original scale). Reported is the hip circumference slope $\beta_A + \delta_s$ for each sex category: the PWV difference per 1 original unit higher hip circumference in that category with triglycerides, diastolic pressure and age held fixed.

\textit{Intervals.} 95\% confidence intervals.
{\footnotesize\setlength{\tabcolsep}{3pt}
\begin{longtable}{@{}>{\raggedright\arraybackslash}p{0.15\linewidth}>{\raggedright\arraybackslash}p{0.40\linewidth}>{\raggedleft\arraybackslash}p{0.14\linewidth}>{\raggedright\arraybackslash}p{0.24\linewidth}@{}}
\toprule
Block & Parameter & Estimate & Interval \\
\midrule\endhead
First sex category & Hip circumference $\to$ PWV slope $\beta_A + \delta_s$ in the first sex category & $-0.004576$ & $[-0.01091$, $0.001757]$ \\
Second sex category & Hip circumference $\to$ PWV slope $\beta_A + \delta_s$ in the second sex category & $-0.003262$ & $[-0.01138$, $0.00486]$ \\
\bottomrule
\end{longtable}}

\subsubsection*{Sensitivity analyses}

\paragraph{22.\ Narrower time window: hip circumference $\to$ triglycerides $\to$ thigh-to-ankle PWV}
\textit{Design.} Hip circumference, triglycerides and PWV are measured in strict date order with narrowed time windows: triglycerides 90--365 days after hip circumference and PWV 180--730 days after triglycerides (bounds inclusive); all eligible measurement chains of a participant are included and weighted to give each participant equal weight. Adjustment is for approximate age at the hip circumference measurement and sex. PWV is the mean of the available sides. Continuous variables are standardized within the analysis sample, with sex as a categorical term.

\textit{Model.} Model as for the single-mediator path (hip circumference $\to$ triglycerides $\to$ PWV): $M = \gamma_0 + \gamma_A \cdot A + \gamma_{\mathrm{age}} \cdot \mathrm{age} + \gamma_{\mathrm{sex}} + \varepsilon_M$; $Y = \beta_0 + \beta_A \cdot A + \beta_M \cdot M + \beta_{AM} \cdot A \cdot M + \beta_{\mathrm{age}} \cdot \mathrm{age} + \beta_{\mathrm{sex}} + \varepsilon_Y$. Both equations are weighted linear regressions, with $A$, $M$ and $Y$ the standardized hip circumference, triglycerides and thigh-to-ankle PWV, respectively. With $A = -0.5$ versus $+0.5$ (a difference of 1 SD) as the contrast, $m_0$ and $m_1$ are the predicted values of $M$ at $A = -0.5$ and $+0.5$, respectively, and $\hat{Y}(a, m)$ is the predicted value of $Y$ at $A = a$, $M = m$: $\mathrm{DE} = \hat{Y}(+0.5, m_0) - \hat{Y}(-0.5, m_0)$, $\mathrm{IE} = \hat{Y}(+0.5, m_1) - \hat{Y}(+0.5, m_0)$, $\mathrm{TE} = \mathrm{DE} + \mathrm{IE}$, each averaged with weights over the sample covariate distribution; also reported are $\gamma_A$, the $M \to Y$ slope $\beta_M - 0.5 \cdot \beta_{AM}$ at $A = -0.5$, and the interaction coefficient $\beta_{AM}$.

\textit{Intervals.} Bootstrap 95\% percentile intervals.
{\footnotesize\setlength{\tabcolsep}{3pt}
\begin{longtable}{@{}>{\raggedright\arraybackslash}p{0.15\linewidth}>{\raggedright\arraybackslash}p{0.40\linewidth}>{\raggedleft\arraybackslash}p{0.14\linewidth}>{\raggedright\arraybackslash}p{0.24\linewidth}@{}}
\toprule
Block & Parameter & Estimate & Interval \\
\midrule\endhead
Path association decomposition (narrower time window) & Direct path association DE: PWV difference corresponding to a 1 SD difference in hip circumference with triglycerides fixed at $m_0$ & $0.06527$ & $[-0.04552$, $0.1802]$ \\
 & Indirect path association IE: PWV difference corresponding to a change in triglycerides from $m_0$ to $m_1$ with hip circumference fixed at $+0.5$ SD & $0.03096$ & $[0.0005886$, $0.07036]$ \\
 & Decomposed total association $\mathrm{TE} = \mathrm{DE} + \mathrm{IE}$ & $0.09623$ & $[-0.00465$, $0.2046]$ \\
Path components (narrower time window) & $\gamma_A$: standardized slope of hip circumference $\to$ triglycerides ($A \to M$) & $0.2617$ & $[0.143$, $0.378]$ \\
 & $\beta_M - 0.5 \cdot \beta_{AM}$: standardized slope of triglycerides $\to$ PWV at hip circumference $-0.5$ SD ($M \to Y \mid A$) & $0.1624$ & $[0.01903$, $0.3309]$ \\
 & $\beta_{AM}$: hip circumference $\times$ triglycerides interaction coefficient in the outcome equation & $-0.04405$ & $[-0.1635$, $0.07593]$ \\
\bottomrule
\end{longtable}}

\paragraph{23.\ Concurrent-window serial path: hip circumference $\to$ triglycerides $\to$ diastolic pressure $\to$ PWV}
\textit{Design.} Each hip circumference measurement day starts one chain, and a participant may contribute several chains: triglycerides within 180 days before or after hip circumference, seated blood pressure within 180 days before or after triglycerides, and PWV within 180 days before or after seated blood pressure (all bounds inclusive, same day allowed, order unrestricted). The three equations share one sample, with adjustment for approximate age on the hip circumference measurement day, recorded sex and study membership; the four measurements are standardized across all eligible chains, and age within the analysis sample.

\textit{Model.} $M_1 = \alpha_1 + a_1 A + \gamma_1^{\top} C + \varepsilon_1$; $M_2 = \alpha_2 + a_2 A + d_{21} M_1 + \gamma_2^{\top} C + \varepsilon_2$; $Y = \alpha_3 + c' A + b_1 M_1 + b_2 M_2 + \gamma_3^{\top} C + \varepsilon_3$. All three are linear regressions, with $A$, $M_1$, $M_2$ and $Y$ the standardized hip circumference, triglycerides, seated diastolic pressure and thigh-to-ankle PWV, and $C$ age, sex and study membership. The coefficients $a_1$, $a_2$, $d_{21}$, $c'$, $b_1$ and $b_2$ are standardized slopes with the other terms of the same equation held fixed; the serial path product is $\theta = a_1 \cdot d_{21} \cdot b_2$.

\textit{Intervals.} $a_1$, $a_2$, $d_{21}$, $c'$, $b_1$, $b_2$: 95\% confidence intervals; $\theta$: bootstrap 95\% percentile interval. $P$ values are not adjusted for multiple comparisons.
{\footnotesize\setlength{\tabcolsep}{3pt}
\begin{longtable}{@{}>{\raggedright\arraybackslash}p{0.14\linewidth}>{\raggedright\arraybackslash}p{0.29\linewidth}>{\raggedleft\arraybackslash}p{0.13\linewidth}>{\raggedright\arraybackslash}p{0.22\linewidth}>{\raggedleft\arraybackslash}p{0.14\linewidth}@{}}
\toprule
Block & Parameter & Estimate & Interval & $P$ \\
\midrule\endhead
Path components (adjacent nodes $\pm 180$ days) & $a_1$: standardized slope of hip circumference $\to$ triglycerides ($A \to M_1$) & $0.2287$ & $[0.1999$, $0.2575]$ & \mbox{$3.326\times10^{-53}$} \\
 & $a_2$: standardized slope of hip circumference $\to$ seated diastolic pressure, adjusted for triglycerides ($A \to M_2 \mid M_1$) & $0.2305$ & $[0.2026$, $0.2583]$ & \mbox{$9.479\times10^{-58}$} \\
 & $d_{21}$: standardized slope of triglycerides $\to$ seated diastolic pressure, adjusted for hip circumference ($M_1 \to M_2 \mid A$) & $0.1748$ & $[0.1425$, $0.207]$ & \mbox{$4.492\times10^{-26}$} \\
 & $c'$: standardized slope of hip circumference $\to$ PWV, adjusted for both mediators ($A \to Y \mid M_1, M_2$) & $-0.01977$ & $[-0.04495$, $0.005403]$ & \mbox{$0.1237$} \\
 & $b_1$: standardized slope of triglycerides $\to$ PWV, adjusted for hip circumference and diastolic pressure ($M_1 \to Y \mid A, M_2$) & $0.03246$ & $[0.007954$, $0.05697]$ & \mbox{$0.009439$} \\
 & $b_2$: standardized slope of seated diastolic pressure $\to$ PWV, adjusted for hip circumference and triglycerides ($M_2 \to Y \mid A, M_1$) & $0.4394$ & $[0.4107$, $0.4682]$ & \mbox{$4.326\times10^{-180}$} \\
Path index (adjacent nodes $\pm 180$ days) & $\theta = a_1 \cdot d_{21} \cdot b_2$: serial path product association index & $0.01757$ & $[0.01384$, $0.02164]$ &  \\
\bottomrule
\end{longtable}}

\paragraph{24.\ Extended-adjustment serial path: hip circumference $\to$ triglycerides $\to$ diastolic pressure $\to$ PWV}
\textit{Design.} Chains and windows as in the main serial path: triglycerides 1--365 days after hip circumference, seated blood pressure within 180 days before or after triglycerides, and PWV 1--730 days after seated blood pressure (all bounds inclusive), one chain per hip circumference measurement day. In addition to approximate age, recorded sex and study membership, adjustment includes the most recent self-reported current smoking status on or within the 730 days before the hip circumference measurement day (categorical term) and genetic principal components 1--5; the four measurements are standardized across all eligible chains, and age and principal components within the analysis sample.

\textit{Model.} $M_1 = \alpha_1 + a_1 A + \gamma_1^{\top} C + \varepsilon_1$; $M_2 = \alpha_2 + a_2 A + d_{21} M_1 + \gamma_2^{\top} C + \varepsilon_2$; $Y = \alpha_3 + c' A + b_1 M_1 + b_2 M_2 + \gamma_3^{\top} C + \varepsilon_3$ (linear regressions; $A$, $M_1$, $M_2$ and $Y$ are the standardized hip circumference, triglycerides, seated diastolic pressure and thigh-to-ankle PWV, and $C$ age, sex, study membership, smoking status and genetic principal components PC1--PC5). $\theta = a_1 \cdot d_{21} \cdot b_2$ is the serial path product under extended adjustment.

\textit{Intervals.} Bootstrap 95\% percentile intervals.
{\footnotesize\setlength{\tabcolsep}{3pt}
\begin{longtable}{@{}>{\raggedright\arraybackslash}p{0.15\linewidth}>{\raggedright\arraybackslash}p{0.40\linewidth}>{\raggedleft\arraybackslash}p{0.14\linewidth}>{\raggedright\arraybackslash}p{0.24\linewidth}@{}}
\toprule
Block & Parameter & Estimate & Interval \\
\midrule\endhead
Path index (extended adjustment) & $\theta = a_1 \cdot d_{21} \cdot b_2$: serial path product association index under extended adjustment & $-0.006951$ & $[-0.4041$, $0.4889]$ \\
\bottomrule
\end{longtable}}

\paragraph{25.\ Narrow-window joint path ($\pm 90$ days): hip circumference $\to$ triglycerides $\to$ diastolic pressure $\to$ PWV}
\textit{Design.} The four measurements are taken in the same period: each PWV measurement day forms one row, with hip circumference, blood triglycerides and seated blood pressure each taken as the nearest measurement within 90 days before or after that day (bounds inclusive), order unrestricted; a participant may contribute several rows, $n = 3219$. The four equations share one sample, all adjusted for approximate age (PWV measurement year minus birth year) and recorded sex (categorical term); variable definitions as in the $\pm 180$-day joint path, and continuous variables are kept on their original scale.

\textit{Model.} $Y = c_0 + c \cdot A + \gamma_0^{\top} C + \varepsilon_0$; $M_1 = \alpha_1 + a_1 A + \gamma_1^{\top} C + \varepsilon_1$; $M_2 = \alpha_2 + a_2 A + d_{21} M_1 + \gamma_2^{\top} C + \varepsilon_2$; $Y = \alpha_3 + c' A + b_1 M_1 + b_2 M_2 + \gamma_3^{\top} C + \varepsilon_3$. All four are linear regressions, with $A$, $M_1$, $M_2$ and $Y$ hip circumference, blood triglycerides, seated diastolic pressure and thigh-to-ankle PWV (original scale), and $C$ age and sex. Each coefficient is the difference in the equation's outcome per 1 original unit higher value of that variable with the other terms of the same equation held fixed; $c$ is the hip circumference coefficient adjusted for $C$ only, and $\Delta = c - c'$.

\textit{Intervals.} 95\% confidence intervals; $P$ values are not adjusted for multiple comparisons.
{\footnotesize\setlength{\tabcolsep}{3pt}
\begin{longtable}{@{}>{\raggedright\arraybackslash}p{0.14\linewidth}>{\raggedright\arraybackslash}p{0.29\linewidth}>{\raggedleft\arraybackslash}p{0.13\linewidth}>{\raggedright\arraybackslash}p{0.22\linewidth}>{\raggedleft\arraybackslash}p{0.14\linewidth}@{}}
\toprule
Block & Parameter & Estimate & Interval & $P$ \\
\midrule\endhead
Path coefficients & $a_1$: hip circumference $\to$ triglycerides ($A \to M_1$) & $1.678$ & $[1.422$, $1.934]$ & \mbox{$8.992\times10^{-37}$} \\
 & $a_2$: hip circumference $\to$ seated diastolic pressure, adjusted for triglycerides ($A \to M_2 \mid M_1$) & $0.2644$ & $[0.2227$, $0.306]$ & \mbox{$1.024\times10^{-34}$} \\
 & $d_{21}$: triglycerides $\to$ seated diastolic pressure, adjusted for hip circumference ($M_1 \to M_2 \mid A$) & $0.02577$ & $[0.01969$, $0.03185]$ & \mbox{$1.431\times10^{-16}$} \\
 & $c'$: hip circumference $\to$ PWV, adjusted for both mediators ($A \to Y \mid M_1, M_2$) & $-0.004615$ & $[-0.01108$, $0.001853]$ & \mbox{$0.1619$} \\
 & $b_1$: triglycerides $\to$ PWV, adjusted for hip circumference and diastolic pressure ($M_1 \to Y \mid A, M_2$) & $0.0008735$ & $[8.347\times10^{-5}$, $0.001664]$ & \mbox{$0.03024$} \\
 & $b_2$: seated diastolic pressure $\to$ PWV, adjusted for hip circumference and triglycerides ($M_2 \to Y \mid A, M_1$) & $0.0759$ & $[0.06971$, $0.08208]$ & \mbox{$1.877\times10^{-117}$} \\
Hip circumference coefficient comparison (same sample) & $c$: hip circumference $\to$ PWV, adjusted for age and sex only & $0.0202$ & $[0.01333$, $0.02706]$ & \mbox{$8.83\times10^{-9}$} \\
 & $c'$: hip circumference $\to$ PWV, adjusted for both mediators & $-0.004615$ & $[-0.01108$, $0.001853]$ & \mbox{$0.1619$} \\
 & $\Delta = c - c'$: signed change in the hip circumference coefficient after adding both mediators & $0.02481$ & $[0.02102$, $0.02861]$ & \mbox{$9.8\times10^{-37}$} \\
\bottomrule
\end{longtable}}

\paragraph{26.\ Extended-adjustment joint path: smoking and medication added ($\pm 180$ days)}
\textit{Design.} Measurement combination as in the $\pm 180$-day joint path (within 180 days before or after the PWV measurement day, bounds inclusive), $n = 1425$. Beyond age and recorded sex, adjustment includes the most recent smoking status on or within the 730 days before the PWV measurement day (categorical term), and indicators for antihypertensive (ATC C02/C03/C07/C08/C09) and lipid-lowering (C10) medications in the most recent medication record within the preceding 365 days. The four equations share one sample; continuous variables are kept on their original scale.

\textit{Model.} $Y = c_0 + c \cdot A + \gamma_0^{\top} C + \varepsilon_0$; $M_1 = \alpha_1 + a_1 A + \gamma_1^{\top} C + \varepsilon_1$; $M_2 = \alpha_2 + a_2 A + d_{21} M_1 + \gamma_2^{\top} C + \varepsilon_2$; $Y = \alpha_3 + c' A + b_1 M_1 + b_2 M_2 + \gamma_3^{\top} C + \varepsilon_3$. All four are linear regressions, with $A$, $M_1$, $M_2$ and $Y$ hip circumference, blood triglycerides, seated diastolic pressure and thigh-to-ankle PWV (original scale), and $C$ age, sex, smoking status and the two medication indicators. Each coefficient is the difference in the equation's outcome per 1 original unit higher value of that variable with the other terms of the same equation held fixed; $c$ is the hip circumference coefficient adjusted for $C$ only, and $\Delta = c - c'$.

\textit{Intervals.} 95\% confidence intervals; $P$ values are not adjusted for multiple comparisons.
{\footnotesize\setlength{\tabcolsep}{3pt}
\begin{longtable}{@{}>{\raggedright\arraybackslash}p{0.14\linewidth}>{\raggedright\arraybackslash}p{0.29\linewidth}>{\raggedleft\arraybackslash}p{0.13\linewidth}>{\raggedright\arraybackslash}p{0.22\linewidth}>{\raggedleft\arraybackslash}p{0.14\linewidth}@{}}
\toprule
Block & Parameter & Estimate & Interval & $P$ \\
\midrule\endhead
Path coefficients & $a_1$: hip circumference $\to$ triglycerides ($A \to M_1$) & $1.557$ & $[1.195$, $1.92]$ & \mbox{$8.065\times10^{-17}$} \\
 & $a_2$: hip circumference $\to$ seated diastolic pressure, adjusted for triglycerides ($A \to M_2 \mid M_1$) & $0.2751$ & $[0.2157$, $0.3344]$ & \mbox{$3.151\times10^{-19}$} \\
 & $d_{21}$: triglycerides $\to$ seated diastolic pressure, adjusted for hip circumference ($M_1 \to M_2 \mid A$) & $0.01889$ & $[0.01063$, $0.02715]$ & \mbox{$7.83\times10^{-6}$} \\
 & $c'$: hip circumference $\to$ PWV, adjusted for both mediators ($A \to Y \mid M_1, M_2$) & $-0.009385$ & $[-0.019$, $0.0002292]$ & \mbox{$0.05571$} \\
 & $b_1$: triglycerides $\to$ PWV, adjusted for hip circumference and diastolic pressure ($M_1 \to Y \mid A, M_2$) & $0.0004802$ & $[-0.0007165$, $0.001677]$ & \mbox{$0.4313$} \\
 & $b_2$: seated diastolic pressure $\to$ PWV, adjusted for hip circumference and triglycerides ($M_2 \to Y \mid A, M_1$) & $0.07903$ & $[0.07005$, $0.088]$ & \mbox{$1.088\times10^{-60}$} \\
Hip circumference coefficient comparison (same sample) & $c$: hip circumference $\to$ PWV without mediators, adjusted for age, sex, smoking and medication only & $0.01543$ & $[0.00525$, $0.0256]$ & \mbox{$0.002994$} \\
 & $c'$: hip circumference $\to$ PWV, adjusted for both mediators & $-0.009385$ & $[-0.019$, $0.0002292]$ & \mbox{$0.05571$} \\
 & $\Delta = c - c'$: signed change in the hip circumference coefficient after adding both mediators & $0.02481$ & $[0.01922$, $0.0304]$ & \mbox{$8.483\times10^{-18}$} \\
\bottomrule
\end{longtable}}

\FloatBarrier

\subsection{VAT--urate--PWV}
\label{app:rq-designs:233c}

\paragraph{Research question and measures.}
The research question is the association of visceral fat with lower-limb arterial stiffness through serum uric acid, with data from the HPP cohort. The exposure is visceral adipose tissue mass measured by whole-body dual-energy X-ray absorptiometry (DXA) scans; the mediator is serum uric acid; the outcome is thigh-to-ankle pulse wave velocity (PWV), taken as the mean of the available left and right sides. Unless otherwise stated, coefficients are SD changes per 1 SD after standardization within each unit's analysis sample.

\subsubsection*{Primary pathways}

\paragraph{1.\ Single-mediator path: visceral fat mass $\to$ serum uric acid $\to$ thigh-to-ankle PWV}
\textit{Design.} Visceral fat mass, serum uric acid and PWV measured in strict date order: uric acid 30--730 days after the visceral fat measurement and PWV 90--1095 days after uric acid (endpoints inclusive); one measurement chain per participant. Adjustment for approximate age at the visceral fat measurement, recorded sex, study of origin and the most recent serum creatinine on that date or within the preceding 365 days. PWV as the mean of available sides; models fitted on the original scale, with coefficients converted by analysis-sample standard deviations.

\textit{Model.} $M = \alpha_0 + \alpha_A \cdot A + \alpha_C^{\top} C + \varepsilon_M$; $Y = \beta_0 + \beta_A \cdot A + \beta_M \cdot M + \beta_C^{\top} C + \varepsilon_Y$; $Y = \gamma_0 + \gamma_A \cdot A + \gamma_C^{\top} C + \varepsilon_T$. All three are linear regressions, with $A$, $M$ and $Y$ denoting visceral fat mass, serum uric acid and thigh-to-ankle PWV, and $C$ the covariates. $\alpha_A$, $\beta_M$, $\beta_A$ and $\gamma_A$ are the slopes of $A \to M$, of $M \to Y$ adjusted for $A$, of $A \to Y$ adjusted for $M$, and of the total $A \to Y$ association, converted to standardized values by analysis-sample standard deviations; path product $\theta = \alpha_A \cdot \beta_M$, decomposition sum $\beta_A + \theta$.

\textit{Intervals.} Bootstrap 95\% percentile intervals.
{\footnotesize\setlength{\tabcolsep}{3pt}
\begin{longtable}{@{}>{\raggedright\arraybackslash}p{0.15\linewidth}>{\raggedright\arraybackslash}p{0.40\linewidth}>{\raggedleft\arraybackslash}p{0.14\linewidth}>{\raggedright\arraybackslash}p{0.24\linewidth}@{}}
\toprule
Block & Parameter & Estimate & Interval \\
\midrule\endhead
Path components & $\alpha_A$: standardized slope of visceral fat mass $\to$ serum uric acid ($A \to M$) & $0.2933$ & $[0.121$, $0.4496]$ \\
 & $\beta_M$: standardized slope of serum uric acid $\to$ PWV adjusted for visceral fat mass ($M \to Y \mid A$) & $0.1306$ & $[-0.1616$, $0.388]$ \\
 & $\beta_A$: standardized slope of visceral fat mass $\to$ PWV adjusted for serum uric acid ($A \to Y \mid M$) & $0.2887$ & $[0.01637$, $0.5378]$ \\
 & $\gamma_A$: adjusted total standardized slope of visceral fat mass $\to$ PWV ($A \to Y$) & $0.327$ & $[0.04982$, $0.5581]$ \\
Path index & $\theta = \alpha_A \cdot \beta_M$: path-product association index & $0.03831$ & $[-0.04437$, $0.1257]$ \\
Additive decomposition & $\beta_A + \theta$: decomposition sum & $0.327$ & $[0.04982$, $0.5581]$ \\
\bottomrule
\end{longtable}}

\paragraph{2.\ Single-mediator path: visceral fat mass $\to$ serum uric acid $\to$ PWV (within 365 days)}
\textit{Design.} Visceral fat mass, serum uric acid and PWV measured in the same period, with any two measurements at most 365 days apart (endpoints inclusive); one set of measurements per participant. All three equations share the same sample, with adjustment for approximate age at the visceral fat measurement and sex. PWV as the mean of available sides. Visceral fat mass, uric acid and PWV standardized within the analysis sample; sex as a categorical term.

\textit{Model.} $M = \alpha_0 + \alpha_A \cdot A + \alpha_C^{\top} C + \varepsilon_M$; $Y = \beta_0 + \beta_A \cdot A + \beta_M \cdot M + \beta_C^{\top} C + \varepsilon_Y$; $Y = \gamma_0 + \gamma_A \cdot A + \gamma_C^{\top} C + \varepsilon_T$. All three are linear regressions, with $A$, $M$ and $Y$ the standardized visceral fat mass, serum uric acid and thigh-to-ankle PWV, and $C$ age and sex. $\alpha_A$ is the SD change in uric acid per 1 SD higher visceral fat mass; $\beta_M$ and $\beta_A$ are the mutually adjusted standardized slopes of uric acid and visceral fat mass on PWV; $\gamma_A$ is the total standardized slope of visceral fat mass on PWV adjusted only for $C$; path product $\theta = \alpha_A \cdot \beta_M$.

\textit{Intervals.} $\alpha_A$, $\beta_M$, $\beta_A$, $\gamma_A$: 95\% confidence intervals; $\theta$: bootstrap 95\% percentile interval.
{\footnotesize\setlength{\tabcolsep}{3pt}
\begin{longtable}{@{}>{\raggedright\arraybackslash}p{0.15\linewidth}>{\raggedright\arraybackslash}p{0.40\linewidth}>{\raggedleft\arraybackslash}p{0.14\linewidth}>{\raggedright\arraybackslash}p{0.24\linewidth}@{}}
\toprule
Block & Parameter & Estimate & Interval \\
\midrule\endhead
Path components & $\alpha_A$: standardized slope of visceral fat mass $\to$ serum uric acid ($A \to M$) & $0.3147$ & $[0.2792$, $0.3502]$ \\
 & $\beta_M$: standardized slope of serum uric acid $\to$ PWV adjusted for visceral fat mass ($M \to Y \mid A$) & $0.05477$ & $[0.00806$, $0.1015]$ \\
 & $\beta_A$: standardized slope of visceral fat mass $\to$ PWV adjusted for serum uric acid ($A \to Y \mid M$) & $0.2115$ & $[0.1667$, $0.2563]$ \\
 & $\gamma_A$: adjusted total standardized slope of visceral fat mass $\to$ PWV ($A \to Y$) & $0.2287$ & $[0.1869$, $0.2705]$ \\
Path index & $\theta = \alpha_A \cdot \beta_M$: path-product association index & $0.01723$ & $[0.0024$, $0.03242]$ \\
\bottomrule
\end{longtable}}

\paragraph{3.\ Mutually adjusted model: visceral fat and uric acid on PWV ($\pm$180 days)}
\textit{Design.} Visceral adipose tissue mass, serum uric acid and thigh-to-ankle PWV measured in the same period: all three pairwise within 180 days (endpoints inclusive, either order); for each participant, the set closest in date, $n = 2063$. PWV as the mean of available sides. Adjustment for approximate age (visceral fat measurement year minus birth year) and recorded sex. Visceral fat, uric acid and PWV standardized within the analysis sample; sex as a categorical term.

\textit{Model.} $z(Y) = \beta_0 + \beta_A \cdot z(A) + \beta_M \cdot z(M) + \gamma_{\mathrm{age}} \cdot \mathrm{age} + \gamma_{\mathrm{sex}} + \varepsilon$ (linear regression; $A$ visceral adipose tissue mass, $M$ serum uric acid, $Y$ thigh-to-ankle PWV, $z(\cdot)$ standardization within the analysis sample). $\beta_A$ and $\beta_M$ are the SD differences in PWV per 1 SD higher visceral fat and uric acid, mutually adjusted and at fixed age and sex.

\textit{Intervals.} 95\% confidence intervals.
{\footnotesize\setlength{\tabcolsep}{3pt}
\begin{longtable}{@{}>{\raggedright\arraybackslash}p{0.15\linewidth}>{\raggedright\arraybackslash}p{0.40\linewidth}>{\raggedleft\arraybackslash}p{0.14\linewidth}>{\raggedright\arraybackslash}p{0.24\linewidth}@{}}
\toprule
Block & Parameter & Estimate & Interval \\
\midrule\endhead
Mutually adjusted model ($\pm$180 days) & $\beta_A$: standardized coefficient of visceral adipose tissue mass $\to$ PWV adjusted for uric acid ($A \to Y \mid M$) & $0.2098$ & $[0.1586$, $0.261]$ \\
 & $\beta_M$: standardized coefficient of serum uric acid $\to$ PWV adjusted for visceral fat ($M \to Y \mid A$) & $0.06302$ & $[0.009301$, $0.1167]$ \\
\bottomrule
\end{longtable}}

\subsubsection*{Adjacent segments and pairwise associations}

\paragraph{4.\ Pairwise association: visceral fat mass and serum uric acid (within 365 days)}
\textit{Design.} Visceral fat mass and serum uric acid measured in the same period (uric acid within 365 days before or after the visceral fat measurement, endpoints inclusive); for each participant, the pair with the shortest interval. Adjustment for approximate age at the visceral fat measurement and sex. Visceral fat mass and uric acid standardized within the analysis sample; sex as a categorical term.

\textit{Model.} $z(M) = \beta_0 + \beta \cdot z(A) + \beta_{\mathrm{age}} \cdot \mathrm{age} + \gamma_{\mathrm{sex}} + \varepsilon$, a linear regression, with $A$ visceral fat mass and $M$ serum uric acid. $\beta$ is the SD difference in uric acid per 1 SD higher visceral fat mass at fixed age and sex.

\textit{Intervals.} 95\% confidence intervals.
{\footnotesize\setlength{\tabcolsep}{3pt}
\begin{longtable}{@{}>{\raggedright\arraybackslash}p{0.15\linewidth}>{\raggedright\arraybackslash}p{0.40\linewidth}>{\raggedleft\arraybackslash}p{0.14\linewidth}>{\raggedright\arraybackslash}p{0.24\linewidth}@{}}
\toprule
Block & Parameter & Estimate & Interval \\
\midrule\endhead
Pairwise association (within 365 days) & $\beta$: adjusted standardized slope of visceral fat mass on serum uric acid ($A \to M$) & $0.3137$ & $[0.2792$, $0.3483]$ \\
\bottomrule
\end{longtable}}

\paragraph{5.\ Pairwise association: serum uric acid and thigh-to-ankle PWV (within 365 days)}
\textit{Design.} Serum uric acid and PWV measured in the same period (PWV within 365 days before or after the uric acid measurement, endpoints inclusive); for each participant, the pair with the shortest interval, $n = 3830$. Adjustment for approximate age at the uric acid measurement and sex. PWV as the mean of available sides. Uric acid and PWV standardized within the analysis sample; sex as a categorical term.

\textit{Model.} $z(Y) = \beta_0 + \beta \cdot z(M) + \beta_{\mathrm{age}} \cdot \mathrm{age} + \gamma_{\mathrm{sex}} + \varepsilon$, a linear regression, with $M$ serum uric acid and $Y$ thigh-to-ankle PWV. $\beta$ is the SD difference in PWV per 1 SD higher uric acid at fixed age and sex.

\textit{Intervals.} 95\% confidence intervals.
{\footnotesize\setlength{\tabcolsep}{3pt}
\begin{longtable}{@{}>{\raggedright\arraybackslash}p{0.15\linewidth}>{\raggedright\arraybackslash}p{0.40\linewidth}>{\raggedleft\arraybackslash}p{0.14\linewidth}>{\raggedright\arraybackslash}p{0.24\linewidth}@{}}
\toprule
Block & Parameter & Estimate & Interval \\
\midrule\endhead
Pairwise association (within 365 days) & $\beta$: adjusted standardized slope of serum uric acid on thigh-to-ankle PWV ($M \to Y$) & $0.1476$ & $[0.1105$, $0.1847]$ \\
\bottomrule
\end{longtable}}

\paragraph{6.\ Pairwise total association: visceral fat mass and PWV (within 365 days)}
\textit{Design.} Visceral fat mass and PWV measured in the same period (PWV within 365 days before or after the visceral fat measurement, endpoints inclusive); for each participant, the pair with the shortest interval, $n = 7849$. Adjustment for approximate age at the visceral fat measurement and sex. PWV as the mean of available sides. Visceral fat mass and PWV standardized within the analysis sample; sex as a categorical term.

\textit{Model.} $z(Y) = \beta_0 + \beta \cdot z(A) + \beta_{\mathrm{age}} \cdot \mathrm{age} + \gamma_{\mathrm{sex}} + \varepsilon$, a linear regression, with $A$ visceral fat mass and $Y$ thigh-to-ankle PWV. $\beta$ is the SD difference in PWV per 1 SD higher visceral fat mass at fixed age and sex.

\textit{Intervals.} 95\% confidence intervals.
{\footnotesize\setlength{\tabcolsep}{3pt}
\begin{longtable}{@{}>{\raggedright\arraybackslash}p{0.15\linewidth}>{\raggedright\arraybackslash}p{0.40\linewidth}>{\raggedleft\arraybackslash}p{0.14\linewidth}>{\raggedright\arraybackslash}p{0.24\linewidth}@{}}
\toprule
Block & Parameter & Estimate & Interval \\
\midrule\endhead
Pairwise total association (within 365 days) & $\beta$: adjusted standardized total-association slope of visceral fat mass on thigh-to-ankle PWV ($A \to Y$) & $0.2141$ & $[0.1903$, $0.2379]$ \\
\bottomrule
\end{longtable}}

\paragraph{7.\ Pairwise association: visceral adipose tissue mass--serum uric acid ($\pm$180 days)}
\textit{Design.} Visceral adipose tissue mass (whole-body scan) and serum uric acid measured in the same period: uric acid within 180 days of the visceral fat measurement date (endpoints inclusive, either order); for each participant, the pair closest in date. Adjustment for approximate age (visceral fat measurement year minus birth year) and recorded sex. Visceral fat and uric acid standardized within the analysis sample; sex as a categorical term.

\textit{Model.} $z(M) = \beta_0 + \beta_A \cdot z(A) + \gamma_{\mathrm{age}} \cdot \mathrm{age} + \gamma_{\mathrm{sex}} + \varepsilon$ (linear regression; $A$ visceral adipose tissue mass, $M$ serum uric acid, $z(\cdot)$ standardization within the analysis sample). $\beta_A$ is the SD difference in uric acid per 1 SD higher visceral fat at fixed age and sex.

\textit{Intervals.} 95\% confidence intervals.
{\footnotesize\setlength{\tabcolsep}{3pt}
\begin{longtable}{@{}>{\raggedright\arraybackslash}p{0.15\linewidth}>{\raggedright\arraybackslash}p{0.40\linewidth}>{\raggedleft\arraybackslash}p{0.14\linewidth}>{\raggedright\arraybackslash}p{0.24\linewidth}@{}}
\toprule
Block & Parameter & Estimate & Interval \\
\midrule\endhead
Pairwise association & $\beta_A$: standardized coefficient of visceral adipose tissue mass $\to$ serum uric acid ($A \to M$) & $0.3178$ & $[0.2771$, $0.3584]$ \\
\bottomrule
\end{longtable}}

\paragraph{8.\ Pairwise association: serum uric acid--thigh-to-ankle PWV ($\pm$180 days)}
\textit{Design.} Serum uric acid and thigh-to-ankle PWV measured in the same period: PWV within 180 days of the uric acid test date (endpoints inclusive, either order); for each participant, the pair closest in date. PWV as the mean of the available left and right values. Adjustment for approximate age (uric acid test year minus birth year) and recorded sex. Uric acid and PWV standardized within the analysis sample; sex as a categorical term.

\textit{Model.} $z(Y) = \beta_0 + \beta_M \cdot z(M) + \gamma_{\mathrm{age}} \cdot \mathrm{age} + \gamma_{\mathrm{sex}} + \varepsilon$ (linear regression; $M$ serum uric acid, $Y$ thigh-to-ankle PWV, $z(\cdot)$ standardization within the analysis sample). $\beta_M$ is the SD difference in PWV per 1 SD higher uric acid at fixed age and sex.

\textit{Intervals.} 95\% confidence intervals.
{\footnotesize\setlength{\tabcolsep}{3pt}
\begin{longtable}{@{}>{\raggedright\arraybackslash}p{0.15\linewidth}>{\raggedright\arraybackslash}p{0.40\linewidth}>{\raggedleft\arraybackslash}p{0.14\linewidth}>{\raggedright\arraybackslash}p{0.24\linewidth}@{}}
\toprule
Block & Parameter & Estimate & Interval \\
\midrule\endhead
Pairwise association & $\beta_M$: standardized coefficient of serum uric acid $\to$ thigh-to-ankle PWV ($M \to Y$) & $0.1575$ & $[0.1157$, $0.1993]$ \\
\bottomrule
\end{longtable}}

\subsubsection*{Reverse direction}

\paragraph{9.\ Reverse path: PWV $\to$ serum uric acid $\to$ visceral fat mass}
\textit{Design.} Measurement order reversed: PWV first, serum uric acid 90--1095 days after PWV, and visceral fat mass 30--730 days after uric acid (endpoints inclusive), in strict order; one chain per person, $n = 250$. Adjustment for approximate age at the PWV measurement, recorded sex, study of origin and the most recent serum creatinine on the PWV measurement date or within the preceding 365 days. PWV as the mean of available sides; models fitted on the original scale, with coefficients converted by analysis-sample standard deviations.

\textit{Model.} $A =$ the earlier PWV, $M =$ the subsequent serum uric acid, $Y =$ the later visceral fat mass, $C$ the covariates; linear regressions: $M = \alpha_0 + \alpha_A \cdot A + \alpha_C^{\top} C + \varepsilon_M$; $Y = \beta_0 + \beta_A \cdot A + \beta_M \cdot M + \beta_C^{\top} C + \varepsilon_Y$; $Y = \gamma_0 + \gamma_A \cdot A + \gamma_C^{\top} C + \varepsilon_T$. $\alpha_A$, $\beta_M$, $\beta_A$ and $\gamma_A$ converted to standardized values by analysis-sample standard deviations; reverse path product $\theta = \alpha_A \cdot \beta_M$, decomposition sum $\beta_A + \theta$.

\textit{Intervals.} Bootstrap 95\% percentile intervals.
{\footnotesize\setlength{\tabcolsep}{3pt}
\begin{longtable}{@{}>{\raggedright\arraybackslash}p{0.15\linewidth}>{\raggedright\arraybackslash}p{0.40\linewidth}>{\raggedleft\arraybackslash}p{0.14\linewidth}>{\raggedright\arraybackslash}p{0.24\linewidth}@{}}
\toprule
Block & Parameter & Estimate & Interval \\
\midrule\endhead
Reverse path components & $\alpha_A$: standardized slope of earlier PWV $\to$ subsequent serum uric acid (reverse $A \to M$) & $0.1289$ & $[0.002218$, $0.2482]$ \\
 & $\beta_M$: standardized slope of serum uric acid $\to$ later visceral fat mass adjusted for PWV (reverse $M \to Y \mid A$) & $0.3613$ & $[0.2373$, $0.4954]$ \\
 & $\beta_A$: standardized slope of PWV $\to$ later visceral fat mass adjusted for serum uric acid (reverse $A \to Y \mid M$) & $0.1848$ & $[0.05737$, $0.3095]$ \\
 & $\gamma_A$: adjusted total standardized slope of PWV $\to$ later visceral fat mass (reverse $A \to Y$) & $0.2314$ & $[0.09662$, $0.3532]$ \\
Reverse path index & $\theta = \alpha_A \cdot \beta_M$: reverse path-product association index & $0.04657$ & $[0.0009421$, $0.08922]$ \\
Reverse additive decomposition & $\beta_A + \theta$: reverse decomposition sum & $0.2314$ & $[0.09662$, $0.3532]$ \\
\bottomrule
\end{longtable}}

\paragraph{10.\ Reversed measurement order: PWV before visceral fat and serum uric acid}
\textit{Design.} Variable roles and equations as in the forward design, with the measurement order reversed: visceral fat 1--730 days after PWV, uric acid 1--365 days after PWV, and visceral fat and uric acid at most 365 days apart (all endpoints inclusive); one set of measurements per participant. Adjustment for approximate age at the visceral fat measurement and sex. PWV as the mean of available sides. Visceral fat mass, uric acid and PWV standardized within the analysis sample; sex as a categorical term.

\textit{Model.} $M = \alpha_0 + \alpha_A \cdot A + \alpha_C^{\top} C + \varepsilon_M$; $Y = \beta_0 + \beta_A \cdot A + \beta_M \cdot M + \beta_C^{\top} C + \varepsilon_Y$; $Y = \gamma_0 + \gamma_A \cdot A + \gamma_C^{\top} C + \varepsilon_T$. All three are linear regressions, with $A$, $M$ and $Y$ the standardized visceral fat mass, serum uric acid and thigh-to-ankle PWV, and $C$ age and sex. $\alpha_A$ is the SD change in uric acid per 1 SD higher visceral fat mass; $\beta_M$ and $\beta_A$ are the mutually adjusted standardized slopes of uric acid and visceral fat mass on PWV; $\gamma_A$ is the total standardized slope of visceral fat mass on PWV adjusted only for $C$; path product $\theta = \alpha_A \cdot \beta_M$.

\textit{Intervals.} $\alpha_A$, $\beta_M$, $\beta_A$, $\gamma_A$: 95\% confidence intervals; $\theta$: bootstrap 95\% percentile interval.
{\footnotesize\setlength{\tabcolsep}{3pt}
\begin{longtable}{@{}>{\raggedright\arraybackslash}p{0.15\linewidth}>{\raggedright\arraybackslash}p{0.40\linewidth}>{\raggedleft\arraybackslash}p{0.14\linewidth}>{\raggedright\arraybackslash}p{0.24\linewidth}@{}}
\toprule
Block & Parameter & Estimate & Interval \\
\midrule\endhead
Path components (PWV measured first) & $\alpha_A$: standardized slope of visceral fat mass $\to$ serum uric acid ($A \to M$) & $0.2704$ & $[-0.662$, $1.203]$ \\
 & $\beta_M$: standardized slope of serum uric acid $\to$ PWV adjusted for visceral fat mass ($M \to Y \mid A$) & $0.5706$ & $[-0.2617$, $1.403]$ \\
 & $\beta_A$: standardized slope of visceral fat mass $\to$ PWV adjusted for serum uric acid ($A \to Y \mid M$) & $0.2335$ & $[-0.5313$, $0.9983]$ \\
 & $\gamma_A$: adjusted total standardized slope of visceral fat mass $\to$ PWV ($A \to Y$) & $0.3878$ & $[-0.532$, $1.308]$ \\
Path index (PWV measured first) & $\theta = \alpha_A \cdot \beta_M$: path-product association index & $0.1543$ & $[-0.3598$, $0.6748]$ \\
\bottomrule
\end{longtable}}

\subsubsection*{Heterogeneity and interaction}

\paragraph{11.\ Conditional PWV slopes under the visceral fat mass $\times$ serum uric acid interaction}
\textit{Design.} Visceral fat mass, serum uric acid and PWV measured in strict date order: uric acid 30--730 days after the visceral fat measurement and PWV 90--1095 days after uric acid (endpoints inclusive); one chain per person. Adjustment for approximate age at the visceral fat measurement, recorded sex, study of origin and the most recent serum creatinine on that date or within the preceding 365 days; the outcome equation adds a visceral fat mass $\times$ serum uric acid product term. PWV as the mean of available sides; models fitted on the original scale, with coefficients converted by analysis-sample standard deviations.

\textit{Model.} $M = \alpha_0 + \alpha_A \cdot A + \alpha_C^{\top} C + \varepsilon_M$; $Y = \beta_0 + \beta_A \cdot A + \beta_M \cdot M + \beta_{AM} \cdot A \cdot M + \beta_C^{\top} C + \varepsilon_Y$ (linear regressions on the original scale; $A$ visceral fat mass, $M$ serum uric acid, $Y$ thigh-to-ankle PWV, $C$ covariates; $s$ the analysis-sample standard deviation, $\bar{A}$ the sample mean of $A$). Reported: $\tilde{\alpha}_A = \alpha_A \cdot s_A / s_M$; standardized interaction coefficient $\tilde{\beta}_{AM} = \beta_{AM} \cdot s_A \cdot s_M / s_Y$; conditional standardized $M \to Y$ slopes $(\beta_M + \beta_{AM} \cdot a) \cdot s_M / s_Y$ with $A$ set to $a = \bar{A}$ and $a = \bar{A} + s_A$.

\textit{Intervals.} Bootstrap 95\% percentile intervals.
{\footnotesize\setlength{\tabcolsep}{3pt}
\begin{longtable}{@{}>{\raggedright\arraybackslash}p{0.15\linewidth}>{\raggedright\arraybackslash}p{0.40\linewidth}>{\raggedleft\arraybackslash}p{0.14\linewidth}>{\raggedright\arraybackslash}p{0.24\linewidth}@{}}
\toprule
Block & Parameter & Estimate & Interval \\
\midrule\endhead
Interaction term & $\tilde{\beta}_{AM}$: standardized interaction coefficient of visceral fat mass $\times$ serum uric acid & $0.2426$ & $[-0.00345$, $0.4171]$ \\
Conditional $M \to Y$ slopes & Conditional standardized slope of serum uric acid $\to$ PWV at mean visceral fat mass ($a = \bar{A}$) & $0.2063$ & $[-0.1143$, $0.4813]$ \\
 & Conditional standardized slope of serum uric acid $\to$ PWV at visceral fat mass 1 SD above the mean ($a = \bar{A} + s_A$) & $0.4489$ & $[-0.05142$, $0.8079]$ \\
Path components & $\tilde{\alpha}_A$: standardized slope of visceral fat mass $\to$ serum uric acid ($A \to M$) & $0.2933$ & $[0.1257$, $0.4521]$ \\
\bottomrule
\end{longtable}}

\paragraph{12.\ Sex heterogeneity: visceral fat mass $\to$ serum uric acid $\to$ PWV (strict date order)}
\textit{Design.} Visceral fat mass, serum uric acid and PWV measured in strict date order: uric acid 30--730 days after the visceral fat measurement and PWV 90--1095 days after uric acid (endpoints inclusive); one chain per person. Intercepts and exposure and mediator slopes estimated separately for the two recorded sex categories (first and second sex categories), with shared adjustment terms for approximate age at the visceral fat measurement, study of origin and the most recent serum creatinine on that date or within the preceding 365 days. PWV as the mean of available sides; coefficients converted by pooled-sample standard deviations.

\textit{Model.} $M = \sum_s 1(\mathrm{sex}=s) \cdot (\alpha_{0s} + \alpha_{A,s} \cdot A) + \alpha_C^{\top} C + \varepsilon_M$; $Y = \sum_s 1(\mathrm{sex}=s) \cdot (\beta_{0s} + \beta_{A,s} \cdot A + \beta_{M,s} \cdot M) + \beta_C^{\top} C + \varepsilon_Y$ (linear regressions on the original scale; $A$ visceral fat mass, $M$ serum uric acid, $Y$ thigh-to-ankle PWV, $C$ age, study of origin and serum creatinine). Reported within category $s$: the $A \to M$ slope $\alpha_{A,s}$, the $A$-adjusted $M \to Y$ slope $\beta_{M,s}$ and the $M$-adjusted $A \to Y$ slope $\beta_{A,s}$ (all converted by pooled-sample standard deviations), and the path product $\theta_s = \alpha_{A,s} \cdot \beta_{M,s}$.

\textit{Intervals.} Bootstrap 95\% percentile intervals.
{\footnotesize\setlength{\tabcolsep}{3pt}
\begin{longtable}{@{}>{\raggedright\arraybackslash}p{0.15\linewidth}>{\raggedright\arraybackslash}p{0.40\linewidth}>{\raggedleft\arraybackslash}p{0.14\linewidth}>{\raggedright\arraybackslash}p{0.24\linewidth}@{}}
\toprule
Block & Parameter & Estimate & Interval \\
\midrule\endhead
First sex category & $\alpha_{A,1}$: standardized slope of visceral fat mass $\to$ serum uric acid in the first sex category & $0.5695$ & $[0.1195$, $1.144]$ \\
 & $\beta_{M,1}$: standardized slope of serum uric acid $\to$ PWV adjusted for visceral fat mass in the first sex category & $-0.07044$ & $[-0.4298$, $0.218]$ \\
 & $\beta_{A,1}$: standardized slope of visceral fat mass $\to$ PWV adjusted for serum uric acid in the first sex category & $0.415$ & $[-0.3006$, $1.201]$ \\
 & $\theta_1 = \alpha_{A,1} \cdot \beta_{M,1}$: path-product association index in the first sex category & $-0.04012$ & $[-0.2843$, $0.1475]$ \\
Second sex category & $\alpha_{A,2}$: standardized slope of visceral fat mass $\to$ serum uric acid in the second sex category & $0.247$ & $[0.04738$, $0.4148]$ \\
 & $\beta_{M,2}$: standardized slope of serum uric acid $\to$ PWV adjusted for visceral fat mass in the second sex category & $0.6185$ & $[0.1087$, $1.051]$ \\
 & $\beta_{A,2}$: standardized slope of visceral fat mass $\to$ PWV adjusted for serum uric acid in the second sex category & $0.1838$ & $[-0.08574$, $0.4157]$ \\
 & $\theta_2 = \alpha_{A,2} \cdot \beta_{M,2}$: path-product association index in the second sex category & $0.1528$ & $[0.00681$, $0.3343]$ \\
\bottomrule
\end{longtable}}

\paragraph{13.\ Age heterogeneity: visceral fat mass $\to$ serum uric acid $\to$ PWV}
\textit{Design.} Visceral fat mass, serum uric acid and PWV measured in strict date order: uric acid 30--730 days after the visceral fat measurement and PWV 90--1095 days after uric acid (endpoints inclusive); one chain per person. Approximate age at the visceral fat measurement entered in continuous form as a main effect and interacted with the exposure and the mediator; additional adjustment for recorded sex, study of origin and the most recent serum creatinine on that date or within the preceding 365 days. PWV as the mean of available sides; coefficients converted by analysis-sample standard deviations.

\textit{Model.} $M = \alpha_0 + \alpha_A \cdot A + \alpha_{\mathrm{age}} \cdot \mathrm{age} + \alpha_{A \times \mathrm{age}} \cdot A \cdot \mathrm{age} + \alpha_C^{\top} C + \varepsilon_M$; $Y = \beta_0 + \beta_A \cdot A + \beta_M \cdot M + \beta_{\mathrm{age}} \cdot \mathrm{age} + \beta_{A \times \mathrm{age}} \cdot A \cdot \mathrm{age} + \beta_{M \times \mathrm{age}} \cdot M \cdot \mathrm{age} + \beta_C^{\top} C + \varepsilon_Y$ (linear regressions on the original scale; $A$ visceral fat mass, $M$ serum uric acid, $Y$ thigh-to-ankle PWV, $\mathrm{age}$ approximate age, $C$ sex, study of origin and serum creatinine). Reported at ages $q = q_{-1}, q_0, q_{+1}$ (mean age $-$ 1 SD, mean age, mean age $+$ 1 SD): $\alpha_A(q) = (\alpha_A + \alpha_{A \times \mathrm{age}} \cdot q) \cdot s_A / s_M$, $\beta_M(q) = (\beta_M + \beta_{M \times \mathrm{age}} \cdot q) \cdot s_M / s_Y$, $\beta_A(q) = (\beta_A + \beta_{A \times \mathrm{age}} \cdot q) \cdot s_A / s_Y$ and $\theta(q) = \alpha_A(q) \cdot \beta_M(q)$.

\textit{Intervals.} Bootstrap 95\% percentile intervals.
{\footnotesize\setlength{\tabcolsep}{3pt}
\begin{longtable}{@{}>{\raggedright\arraybackslash}p{0.15\linewidth}>{\raggedright\arraybackslash}p{0.40\linewidth}>{\raggedleft\arraybackslash}p{0.14\linewidth}>{\raggedright\arraybackslash}p{0.24\linewidth}@{}}
\toprule
Block & Parameter & Estimate & Interval \\
\midrule\endhead
Age $=$ mean $-$ 1 SD & $\alpha_A(q_{-1})$: conditional standardized slope of visceral fat mass $\to$ serum uric acid & $0.3315$ & $[0.03207$, $0.6133]$ \\
 & $\beta_A(q_{-1})$: conditional standardized slope of visceral fat mass $\to$ PWV adjusted for serum uric acid & $0.09374$ & $[-0.3509$, $0.5081]$ \\
 & $\beta_M(q_{-1})$: conditional standardized slope of serum uric acid $\to$ PWV adjusted for visceral fat mass & $0.0826$ & $[-0.2763$, $0.4272]$ \\
 & $\theta(q_{-1}) = \alpha_A(q_{-1}) \cdot \beta_M(q_{-1})$: conditional path-product association index & $0.02738$ & $[-0.1$, $0.167]$ \\
Age $=$ mean & $\alpha_A(q_0)$: conditional standardized slope of visceral fat mass $\to$ serum uric acid & $0.3029$ & $[0.1196$, $0.4815]$ \\
 & $\beta_A(q_0)$: conditional standardized slope of visceral fat mass $\to$ PWV adjusted for serum uric acid & $0.2344$ & $[-0.04972$, $0.4937]$ \\
 & $\beta_M(q_0)$: conditional standardized slope of serum uric acid $\to$ PWV adjusted for visceral fat mass & $0.1502$ & $[-0.1828$, $0.4258]$ \\
 & $\theta(q_0) = \alpha_A(q_0) \cdot \beta_M(q_0)$: conditional path-product association index & $0.04549$ & $[-0.05777$, $0.1527]$ \\
Age $=$ mean $+$ 1 SD & $\alpha_A(q_{+1})$: conditional standardized slope of visceral fat mass $\to$ serum uric acid & $0.2743$ & $[0.07395$, $0.4617]$ \\
 & $\beta_A(q_{+1})$: conditional standardized slope of visceral fat mass $\to$ PWV adjusted for serum uric acid & $0.3751$ & $[0.03728$, $0.7669]$ \\
 & $\beta_M(q_{+1})$: conditional standardized slope of serum uric acid $\to$ PWV adjusted for visceral fat mass & $0.2178$ & $[-0.3079$, $0.6125]$ \\
 & $\theta(q_{+1}) = \alpha_A(q_{+1}) \cdot \beta_M(q_{+1})$: conditional path-product association index & $0.05974$ & $[-0.09548$, $0.1926]$ \\
\bottomrule
\end{longtable}}

\paragraph{14.\ Sex heterogeneity: visceral fat mass $\to$ serum uric acid $\to$ PWV (same period)}
\textit{Design.} Visceral fat mass, serum uric acid and PWV measured in the same period, with any two measurements at most 365 days apart (endpoints inclusive); one set of measurements per participant. Fitted on the pooled sample, with sex (first and second categories) entering as a main effect and through interaction terms with visceral fat mass and uric acid; adjustment for approximate age at the visceral fat measurement. PWV as the mean of available sides. Visceral fat mass, uric acid and PWV standardized within the analysis sample.

\textit{Model.} $M = \alpha_0 + \alpha_{\mathrm{age}} \cdot \mathrm{age} + f(\mathrm{sex}) + \alpha_A(s) \cdot A + \varepsilon_M$; $Y = \beta_0 + \beta_{\mathrm{age}} \cdot \mathrm{age} + f(\mathrm{sex}) + \beta_A(s) \cdot A + \beta_M(s) \cdot M + \varepsilon_Y$ (pooled-sample linear regressions; $A$, $M$ and $Y$ the standardized visceral fat mass, serum uric acid and thigh-to-ankle PWV; $f(\mathrm{sex})$ the sex main effect; $\alpha_A(s)$, $\beta_A(s)$ and $\beta_M(s)$ the slopes within sex category $s$, given by the interaction terms with sex). Reported quantities: within each category, $\alpha_A(s)$, $\beta_M(s)$, $\beta_A(s)$ and the path product $\theta(s) = \alpha_A(s) \cdot \beta_M(s)$.

\textit{Intervals.} $\alpha_A(s)$, $\beta_M(s)$, $\beta_A(s)$: 95\% confidence intervals; $\theta(s)$: bootstrap 95\% percentile interval.
{\footnotesize\setlength{\tabcolsep}{3pt}
\begin{longtable}{@{}>{\raggedright\arraybackslash}p{0.15\linewidth}>{\raggedright\arraybackslash}p{0.40\linewidth}>{\raggedleft\arraybackslash}p{0.14\linewidth}>{\raggedright\arraybackslash}p{0.24\linewidth}@{}}
\toprule
Block & Parameter & Estimate & Interval \\
\midrule\endhead
First sex category & $\alpha_A(s)$: within-category standardized slope of visceral fat mass $\to$ serum uric acid ($A \to M$) & $0.5181$ & $[0.4568$, $0.5794]$ \\
 & $\beta_M(s)$: within-category standardized slope of serum uric acid $\to$ PWV adjusted for visceral fat mass ($M \to Y \mid A$) & $0.072$ & $[-0.001531$, $0.1455]$ \\
 & $\beta_A(s)$: within-category standardized slope of visceral fat mass $\to$ PWV adjusted for serum uric acid ($A \to Y \mid M$) & $0.2952$ & $[0.2076$, $0.3827]$ \\
 & $\theta(s) = \alpha_A(s) \cdot \beta_M(s)$: within-category path-product association index & $0.0373$ & $[-0.001764$, $0.07738]$ \\
Second sex category & $\alpha_A(s)$: within-category standardized slope of visceral fat mass $\to$ serum uric acid ($A \to M$) & $0.2487$ & $[0.2073$, $0.2901]$ \\
 & $\beta_M(s)$: within-category standardized slope of serum uric acid $\to$ PWV adjusted for visceral fat mass ($M \to Y \mid A$) & $0.02729$ & $[-0.0343$, $0.08889]$ \\
 & $\beta_A(s)$: within-category standardized slope of visceral fat mass $\to$ PWV adjusted for serum uric acid ($A \to Y \mid M$) & $0.1889$ & $[0.1383$, $0.2395]$ \\
 & $\theta(s) = \alpha_A(s) \cdot \beta_M(s)$: within-category path-product association index & $0.006788$ & $[-0.007665$, $0.02186]$ \\
\bottomrule
\end{longtable}}

\paragraph{15.\ Exposure $\times$ mediator interaction: visceral fat mass $\to$ serum uric acid $\to$ PWV}
\textit{Design.} Visceral fat mass, serum uric acid and PWV measured in the same period, with any two measurements at most 365 days apart (endpoints inclusive); one set of measurements per participant. Adjustment for approximate age at the visceral fat measurement and sex. PWV as the mean of available sides. Visceral fat mass, uric acid and PWV standardized within the analysis sample; $a = -1$, $0$, $+1$ correspond to visceral fat mass 1 SD below the mean, at the mean and 1 SD above the mean.

\textit{Model.} $M = \alpha_0 + \alpha_A \cdot A + \alpha_C^{\top} C + \varepsilon_M$; $Y = \beta_0 + \beta_M \cdot M + \beta_A \cdot A + \beta_{AM} \cdot A \cdot M + \beta_C^{\top} C + \varepsilon_Y$ (linear regressions; $A$, $M$ and $Y$ the standardized visceral fat mass, serum uric acid and thigh-to-ankle PWV, $C$ age and sex). Reported quantities: the product-term coefficient $\beta_{AM}$, and the conditional path product $\theta(a) = \alpha_A \cdot (\beta_M + \beta_{AM} \cdot a)$ at visceral fat mass level $a$, $a \in \{-1, 0, +1\}$.

\textit{Intervals.} $\beta_{AM}$: 95\% confidence interval; $\theta(a)$: bootstrap 95\% percentile interval.
{\footnotesize\setlength{\tabcolsep}{3pt}
\begin{longtable}{@{}>{\raggedright\arraybackslash}p{0.15\linewidth}>{\raggedright\arraybackslash}p{0.40\linewidth}>{\raggedleft\arraybackslash}p{0.14\linewidth}>{\raggedright\arraybackslash}p{0.24\linewidth}@{}}
\toprule
Block & Parameter & Estimate & Interval \\
\midrule\endhead
Exposure $\times$ mediator product term & $\beta_{AM}$: coefficient of the visceral fat mass $\times$ serum uric acid product term in the PWV equation & $-0.04458$ & $[-0.07859$, $-0.01057]$ \\
Visceral fat mass 1 SD below the analysis-sample mean ($a = -1$) & $\theta(a) = \alpha_A \cdot (\beta_M + \beta_{AM} \cdot a)$: conditional path product & $0.03206$ & $[0.01433$, $0.05127]$ \\
Visceral fat mass at the analysis-sample mean ($a = 0$) & $\theta(a) = \alpha_A \cdot (\beta_M + \beta_{AM} \cdot a)$: conditional path product & $0.01804$ & $[0.003503$, $0.03269]$ \\
Visceral fat mass 1 SD above the analysis-sample mean ($a = +1$) & $\theta(a) = \alpha_A \cdot (\beta_M + \beta_{AM} \cdot a)$: conditional path product & $0.004008$ & $[-0.01376$, $0.02209]$ \\
\bottomrule
\end{longtable}}

\paragraph{16.\ Age interaction: variation of the PWV slopes of visceral fat and uric acid with age}
\textit{Design.} Temporal design as in the $\pm$180-day mutually adjusted model: visceral adipose tissue mass, serum uric acid and PWV all pairwise within 180 days (endpoints inclusive, either order); for each participant, the set closest in date. Approximate age (visceral fat measurement year minus birth year) centered at the analysis-sample mean, entered as a main effect and interacted separately with visceral fat and uric acid; additional adjustment for recorded sex. Visceral fat, uric acid and PWV standardized within the analysis sample.

\textit{Model.} $z(Y) = \beta_0 + \beta_A \cdot z(A) + \beta_M \cdot z(M) + \gamma_{\mathrm{age}} \cdot \mathrm{age}_c + \theta_A \cdot z(A) \cdot \mathrm{age}_c + \theta_M \cdot z(M) \cdot \mathrm{age}_c + \gamma_{\mathrm{sex}} + \varepsilon$ (linear regression; $\mathrm{age}_c$ centered age, in years). Reported: $10 \cdot \theta_A$ and $10 \cdot \theta_M$, the changes in the standardized slopes of visceral fat and uric acid on PWV per 10 years of higher age.

\textit{Intervals.} 95\% confidence intervals.
{\footnotesize\setlength{\tabcolsep}{3pt}
\begin{longtable}{@{}>{\raggedright\arraybackslash}p{0.15\linewidth}>{\raggedright\arraybackslash}p{0.40\linewidth}>{\raggedleft\arraybackslash}p{0.14\linewidth}>{\raggedright\arraybackslash}p{0.24\linewidth}@{}}
\toprule
Block & Parameter & Estimate & Interval \\
\midrule\endhead
Age interaction & $10 \cdot \theta_A$: change in the standardized visceral fat--PWV slope per 10 years of higher age & $-0.06721$ & $[-0.122$, $-0.01238]$ \\
 & $10 \cdot \theta_M$: change in the standardized uric acid--PWV slope per 10 years of higher age & $-0.02704$ & $[-0.08367$, $0.02959]$ \\
\bottomrule
\end{longtable}}

\paragraph{17.\ Visceral fat $\times$ uric acid interaction: non-additivity in the PWV model}
\textit{Design.} Temporal design as in the $\pm$180-day mutually adjusted model: visceral adipose tissue mass, serum uric acid and PWV all pairwise within 180 days (endpoints inclusive, either order); for each participant, the set closest in date. Product term constructed from visceral fat and uric acid after centering at the analysis-sample means; adjustment for approximate age (visceral fat measurement year minus birth year) and recorded sex. Visceral fat, uric acid and PWV standardized within the analysis sample.

\textit{Model.} $z(Y) = \beta_0 + \beta_A \cdot z(A) + \beta_M \cdot z(M) + \theta_{AM} \cdot z(A) \cdot z(M) + \gamma_{\mathrm{age}} \cdot \mathrm{age} + \gamma_{\mathrm{sex}} + \varepsilon$ (linear regression). $\theta_{AM}$ is the change in the standardized slope of visceral fat on PWV per 1 SD higher uric acid, equivalently the change in the uric acid slope per 1 SD higher visceral fat.

\textit{Intervals.} 95\% confidence intervals.
{\footnotesize\setlength{\tabcolsep}{3pt}
\begin{longtable}{@{}>{\raggedright\arraybackslash}p{0.15\linewidth}>{\raggedright\arraybackslash}p{0.40\linewidth}>{\raggedleft\arraybackslash}p{0.14\linewidth}>{\raggedright\arraybackslash}p{0.24\linewidth}@{}}
\toprule
Block & Parameter & Estimate & Interval \\
\midrule\endhead
Visceral fat $\times$ uric acid interaction & $\theta_{AM}$: standardized coefficient of the visceral adipose tissue mass $\times$ serum uric acid interaction & $-0.06174$ & $[-0.09944$, $-0.02404]$ \\
\bottomrule
\end{longtable}}

\subsubsection*{Sensitivity analyses}

\paragraph{18.\ Narrow-window path: visceral fat mass $\to$ serum uric acid $\to$ PWV}
\textit{Design.} Visceral fat mass, serum uric acid and PWV measured in strict date order with narrowed windows: uric acid 90--365 days after the visceral fat measurement and PWV 180--730 days after uric acid (endpoints inclusive); one measurement chain per participant, $n = 52$. Adjustment for approximate age at the visceral fat measurement, recorded sex, study of origin and the most recent serum creatinine on that date or within the preceding 365 days. PWV as the mean of available sides; models fitted on the original scale, with coefficients converted by analysis-sample standard deviations.

\textit{Model.} $M = \alpha_0 + \alpha_A \cdot A + \alpha_C^{\top} C + \varepsilon_M$; $Y = \beta_0 + \beta_A \cdot A + \beta_M \cdot M + \beta_C^{\top} C + \varepsilon_Y$; $Y = \gamma_0 + \gamma_A \cdot A + \gamma_C^{\top} C + \varepsilon_T$. All three are linear regressions, with $A$, $M$ and $Y$ denoting visceral fat mass, serum uric acid and thigh-to-ankle PWV, and $C$ the covariates. $\alpha_A$, $\beta_M$, $\beta_A$ and $\gamma_A$ are the slopes of $A \to M$, of $M \to Y$ adjusted for $A$, of $A \to Y$ adjusted for $M$, and of the total $A \to Y$ association, converted to standardized values by analysis-sample standard deviations; path product $\theta = \alpha_A \cdot \beta_M$, decomposition sum $\beta_A + \theta$.

\textit{Intervals.} Bootstrap 95\% percentile intervals.
{\footnotesize\setlength{\tabcolsep}{3pt}
\begin{longtable}{@{}>{\raggedright\arraybackslash}p{0.15\linewidth}>{\raggedright\arraybackslash}p{0.40\linewidth}>{\raggedleft\arraybackslash}p{0.14\linewidth}>{\raggedright\arraybackslash}p{0.24\linewidth}@{}}
\toprule
Block & Parameter & Estimate & Interval \\
\midrule\endhead
Path components & $\alpha_A$: standardized slope of visceral fat mass $\to$ serum uric acid ($A \to M$) & $0.3126$ & $[0.1195$, $0.498]$ \\
 & $\beta_M$: standardized slope of serum uric acid $\to$ PWV adjusted for visceral fat mass ($M \to Y \mid A$) & $0.1473$ & $[-0.1975$, $0.5562]$ \\
 & $\beta_A$: standardized slope of visceral fat mass $\to$ PWV adjusted for serum uric acid ($A \to Y \mid M$) & $0.4505$ & $[0.1193$, $0.7216]$ \\
 & $\gamma_A$: adjusted total standardized slope of visceral fat mass $\to$ PWV ($A \to Y$) & $0.4965$ & $[0.175$, $0.7436]$ \\
Path index & $\theta = \alpha_A \cdot \beta_M$: path-product association index & $0.04605$ & $[-0.06337$, $0.1736]$ \\
Additive decomposition & $\beta_A + \theta$: decomposition sum & $0.4965$ & $[0.175$, $0.7436]$ \\
\bottomrule
\end{longtable}}

\paragraph{19.\ Extended covariate adjustment: visceral fat mass $\to$ serum uric acid $\to$ PWV}
\textit{Design.} Three measurements in strict date order: uric acid 30--730 days after the visceral fat measurement and PWV 90--1095 days after uric acid (endpoints inclusive); one chain per person, $n = 77$. Adjustment for approximate age at the visceral fat measurement, recorded sex and study of origin, and for the most recent values on or before that date of serum creatinine and seated systolic blood pressure (within 365 days), smoking status (within 1825 days), and urate-lowering medication and antihypertensive or diuretic medication (within 730 days). PWV as the mean of available sides; coefficients converted by analysis-sample standard deviations.

\textit{Model.} $M = \alpha_0 + \alpha_A \cdot A + \alpha_C^{\top} C + \varepsilon_M$; $Y = \beta_0 + \beta_A \cdot A + \beta_M \cdot M + \beta_C^{\top} C + \varepsilon_Y$; $Y = \gamma_0 + \gamma_A \cdot A + \gamma_C^{\top} C + \varepsilon_T$. All three are linear regressions, with $A$, $M$ and $Y$ denoting visceral fat mass, serum uric acid and thigh-to-ankle PWV; $C$ comprises age, sex, study of origin, serum creatinine, seated systolic blood pressure, smoking status and two medication indicators. $\alpha_A$, $\beta_M$, $\beta_A$ and $\gamma_A$ are the slopes of $A \to M$, of $M \to Y$ adjusted for $A$, of $A \to Y$ adjusted for $M$, and of the total $A \to Y$ association, converted to standardized values by analysis-sample standard deviations; path product $\theta = \alpha_A \cdot \beta_M$, decomposition sum $\beta_A + \theta$.

\textit{Intervals.} Bootstrap 95\% percentile intervals.
{\footnotesize\setlength{\tabcolsep}{3pt}
\begin{longtable}{@{}>{\raggedright\arraybackslash}p{0.15\linewidth}>{\raggedright\arraybackslash}p{0.40\linewidth}>{\raggedleft\arraybackslash}p{0.14\linewidth}>{\raggedright\arraybackslash}p{0.24\linewidth}@{}}
\toprule
Block & Parameter & Estimate & Interval \\
\midrule\endhead
Path components & $\alpha_A$: standardized slope of visceral fat mass $\to$ serum uric acid ($A \to M$) & $0.194$ & $[-0.228$, $0.5944]$ \\
 & $\beta_M$: standardized slope of serum uric acid $\to$ PWV adjusted for visceral fat mass ($M \to Y \mid A$) & $-0.04134$ & $[-1.077$, $0.5076]$ \\
 & $\beta_A$: standardized slope of visceral fat mass $\to$ PWV adjusted for serum uric acid ($A \to Y \mid M$) & $-0.06223$ & $[-0.6025$, $0.7057]$ \\
 & $\gamma_A$: adjusted total standardized slope of visceral fat mass $\to$ PWV ($A \to Y$) & $-0.07025$ & $[-0.5604$, $0.6784]$ \\
Path index & $\theta = \alpha_A \cdot \beta_M$: path-product association index & $-0.008019$ & $[-0.231$, $0.1641]$ \\
Additive decomposition & $\beta_A + \theta$: decomposition sum & $-0.07025$ & $[-0.5604$, $0.6784]$ \\
\bottomrule
\end{longtable}}

\paragraph{20.\ Regional fat contrast: visceral and subcutaneous fat mass $\to$ serum uric acid $\to$ PWV}
\textit{Design.} Visceral and subcutaneous fat mass taken from the same body-composition measurement date; uric acid 30--730 days later and PWV 90--1095 days after uric acid (endpoints inclusive), in strict order; one chain per person, $n = 77$. The two exposures fitted separately on the same sample, with adjustment for approximate age at that measurement, recorded sex, study of origin and the most recent serum creatinine on that date or within the preceding 365 days. PWV as the mean of available sides; models fitted on the original scale, with coefficients converted by the same sample's standard deviations.

\textit{Model.} Fitted separately for each exposure $E \in \{\text{visceral fat mass}, \text{subcutaneous fat mass}\}$: $M = \alpha_0 + \alpha_A \cdot E + \alpha_C^{\top} C + \varepsilon_M$; $Y = \beta_0 + \beta_A \cdot E + \beta_M \cdot M + \beta_C^{\top} C + \varepsilon_Y$; $Y = \gamma_0 + \gamma_A \cdot E + \gamma_C^{\top} C + \varepsilon_T$ (linear regressions; $M$ serum uric acid, $Y$ thigh-to-ankle PWV, $C$ covariates). $\alpha_A$, $\beta_M$, $\beta_A$ and $\gamma_A$ converted to standardized values by the same sample's standard deviations; $\theta = \alpha_A \cdot \beta_M$, decomposition sum $\beta_A + \theta$; regional contrast $\theta_{\mathrm{VAT}} - \theta_{\mathrm{SAT}}$.

\textit{Intervals.} Bootstrap 95\% percentile intervals.
{\footnotesize\setlength{\tabcolsep}{3pt}
\begin{longtable}{@{}>{\raggedright\arraybackslash}p{0.15\linewidth}>{\raggedright\arraybackslash}p{0.40\linewidth}>{\raggedleft\arraybackslash}p{0.14\linewidth}>{\raggedright\arraybackslash}p{0.24\linewidth}@{}}
\toprule
Block & Parameter & Estimate & Interval \\
\midrule\endhead
Exposure: visceral fat mass & $\alpha_A$: standardized slope of visceral fat mass $\to$ serum uric acid ($A \to M$) & $0.2933$ & $[0.1085$, $0.4454]$ \\
 & $\beta_A + \theta$: decomposition sum & $0.327$ & $[0.04569$, $0.5557]$ \\
 & $\beta_A$: standardized slope of visceral fat mass $\to$ PWV adjusted for serum uric acid ($A \to Y \mid M$) & $0.2887$ & $[-0.002725$, $0.5449]$ \\
 & $\theta = \alpha_A \cdot \beta_M$: path-product association index & $0.03831$ & $[-0.0471$, $0.1239]$ \\
 & $\beta_M$: standardized slope of serum uric acid $\to$ PWV adjusted for visceral fat mass ($M \to Y \mid A$) & $0.1306$ & $[-0.154$, $0.3995]$ \\
 & $\gamma_A$: adjusted total standardized slope of visceral fat mass $\to$ PWV ($A \to Y$) & $0.327$ & $[0.04569$, $0.5557]$ \\
Exposure: subcutaneous fat mass & $\alpha_A$: standardized slope of subcutaneous fat mass $\to$ serum uric acid ($A \to M$) & $0.1265$ & $[-0.04722$, $0.266]$ \\
 & $\beta_A + \theta$: decomposition sum & $0.04052$ & $[-0.1264$, $0.2259]$ \\
 & $\beta_A$: standardized slope of subcutaneous fat mass $\to$ PWV adjusted for serum uric acid ($A \to Y \mid M$) & $0.0114$ & $[-0.1576$, $0.2069]$ \\
 & $\theta = \alpha_A \cdot \beta_M$: path-product association index & $0.02912$ & $[-0.01623$, $0.09113]$ \\
 & $\beta_M$: standardized slope of serum uric acid $\to$ PWV adjusted for subcutaneous fat mass ($M \to Y \mid A$) & $0.2302$ & $[-0.06568$, $0.5247]$ \\
 & $\gamma_A$: adjusted total standardized slope of subcutaneous fat mass $\to$ PWV ($A \to Y$) & $0.04052$ & $[-0.1264$, $0.2259]$ \\
Regional contrast & $\theta_{\mathrm{VAT}} - \theta_{\mathrm{SAT}}$: difference between the visceral and subcutaneous fat mass path products & $0.00919$ & $[-0.06009$, $0.08563]$ \\
\bottomrule
\end{longtable}}

\paragraph{21.\ Narrow-window path: visceral fat mass $\to$ serum uric acid $\to$ PWV (within 180 days)}
\textit{Design.} Visceral fat mass, serum uric acid and PWV measured in the same period, with any two measurements at most 180 days apart (endpoints inclusive); one set of measurements per participant. All three equations share the same sample, with adjustment for approximate age at the visceral fat measurement and sex. PWV as the mean of available sides. Visceral fat mass, uric acid and PWV standardized within the analysis sample; sex as a categorical term.

\textit{Model.} $M = \alpha_0 + \alpha_A \cdot A + \alpha_C^{\top} C + \varepsilon_M$; $Y = \beta_0 + \beta_A \cdot A + \beta_M \cdot M + \beta_C^{\top} C + \varepsilon_Y$; $Y = \gamma_0 + \gamma_A \cdot A + \gamma_C^{\top} C + \varepsilon_T$. All three are linear regressions, with $A$, $M$ and $Y$ the standardized visceral fat mass, serum uric acid and thigh-to-ankle PWV, and $C$ age and sex. $\alpha_A$ is the SD change in uric acid per 1 SD higher visceral fat mass; $\beta_M$ and $\beta_A$ are the mutually adjusted standardized slopes of uric acid and visceral fat mass on PWV; $\gamma_A$ is the total standardized slope of visceral fat mass on PWV adjusted only for $C$; path product $\theta = \alpha_A \cdot \beta_M$.

\textit{Intervals.} $\alpha_A$, $\beta_M$, $\beta_A$, $\gamma_A$: 95\% confidence intervals; $\theta$: bootstrap 95\% percentile interval.
{\footnotesize\setlength{\tabcolsep}{3pt}
\begin{longtable}{@{}>{\raggedright\arraybackslash}p{0.15\linewidth}>{\raggedright\arraybackslash}p{0.40\linewidth}>{\raggedleft\arraybackslash}p{0.14\linewidth}>{\raggedright\arraybackslash}p{0.24\linewidth}@{}}
\toprule
Block & Parameter & Estimate & Interval \\
\midrule\endhead
Path components (within 180 days) & $\alpha_A$: standardized slope of visceral fat mass $\to$ serum uric acid ($A \to M$) & $0.3151$ & $[0.2736$, $0.3566]$ \\
 & $\beta_M$: standardized slope of serum uric acid $\to$ PWV adjusted for visceral fat mass ($M \to Y \mid A$) & $0.06631$ & $[0.01255$, $0.1201]$ \\
 & $\beta_A$: standardized slope of visceral fat mass $\to$ PWV adjusted for serum uric acid ($A \to Y \mid M$) & $0.2086$ & $[0.1574$, $0.2599]$ \\
 & $\gamma_A$: adjusted total standardized slope of visceral fat mass $\to$ PWV ($A \to Y$) & $0.2295$ & $[0.182$, $0.2771]$ \\
Path index (within 180 days) & $\theta = \alpha_A \cdot \beta_M$: path-product association index & $0.0209$ & $[0.003782$, $0.03803]$ \\
\bottomrule
\end{longtable}}

\paragraph{22.\ Visceral fat measured first: visceral fat mass $\to$ serum uric acid $\to$ PWV}
\textit{Design.} Visceral fat mass measured before uric acid and PWV: uric acid and PWV 1--365 days and 1--730 days after the visceral fat measurement, respectively, and uric acid and PWV at most 365 days apart (all endpoints inclusive); one set of measurements per participant. All three equations share the same sample, with adjustment for approximate age at the visceral fat measurement and sex. PWV as the mean of available sides. Visceral fat mass, uric acid and PWV standardized within the analysis sample; sex as a categorical term.

\textit{Model.} $M = \alpha_0 + \alpha_A \cdot A + \alpha_C^{\top} C + \varepsilon_M$; $Y = \beta_0 + \beta_A \cdot A + \beta_M \cdot M + \beta_C^{\top} C + \varepsilon_Y$; $Y = \gamma_0 + \gamma_A \cdot A + \gamma_C^{\top} C + \varepsilon_T$. All three are linear regressions, with $A$, $M$ and $Y$ the standardized visceral fat mass, serum uric acid and thigh-to-ankle PWV, and $C$ age and sex. $\alpha_A$ is the SD change in uric acid per 1 SD higher visceral fat mass; $\beta_M$ and $\beta_A$ are the mutually adjusted standardized slopes of uric acid and visceral fat mass on PWV; $\gamma_A$ is the total standardized slope of visceral fat mass on PWV adjusted only for $C$; path product $\theta = \alpha_A \cdot \beta_M$.

\textit{Intervals.} $\alpha_A$, $\beta_M$, $\beta_A$, $\gamma_A$: 95\% confidence intervals; $\theta$: bootstrap 95\% percentile interval.
{\footnotesize\setlength{\tabcolsep}{3pt}
\begin{longtable}{@{}>{\raggedright\arraybackslash}p{0.15\linewidth}>{\raggedright\arraybackslash}p{0.40\linewidth}>{\raggedleft\arraybackslash}p{0.14\linewidth}>{\raggedright\arraybackslash}p{0.24\linewidth}@{}}
\toprule
Block & Parameter & Estimate & Interval \\
\midrule\endhead
Path components (visceral fat measured first) & $\alpha_A$: standardized slope of visceral fat mass $\to$ serum uric acid ($A \to M$) & $0.5909$ & $[0.2781$, $0.9036]$ \\
 & $\beta_M$: standardized slope of serum uric acid $\to$ PWV adjusted for visceral fat mass ($M \to Y \mid A$) & $0.5104$ & $[-0.2363$, $1.257]$ \\
 & $\beta_A$: standardized slope of visceral fat mass $\to$ PWV adjusted for serum uric acid ($A \to Y \mid M$) & $-0.0644$ & $[-0.8994$, $0.7706]$ \\
 & $\gamma_A$: adjusted total standardized slope of visceral fat mass $\to$ PWV ($A \to Y$) & $0.2372$ & $[-0.5125$, $0.9869]$ \\
Path index (visceral fat measured first) & $\theta = \alpha_A \cdot \beta_M$: path-product association index & $0.3016$ & $[-0.3011$, $1.209]$ \\
\bottomrule
\end{longtable}}

\paragraph{23.\ Additional adjustment for creatinine and brachial blood pressure: visceral fat mass $\to$ serum uric acid $\to$ PWV}
\textit{Design.} Visceral fat mass, serum uric acid and PWV measured in the same period, with any two measurements at most 365 days apart (endpoints inclusive); one set of measurements per participant. All three equations share the same sample, with adjustment for approximate age at the visceral fat measurement, sex, creatinine from the same blood test as uric acid, and brachial blood pressure from the same examination as PWV (mean of available sides). PWV as the mean of available sides. Visceral fat mass, uric acid, PWV, creatinine and brachial blood pressure standardized within the analysis sample; sex as a categorical term.

\textit{Model.} $M = \alpha_0 + \alpha_A \cdot A + \alpha_C^{\top} C + \varepsilon_M$; $Y = \beta_0 + \beta_A \cdot A + \beta_M \cdot M + \beta_C^{\top} C + \varepsilon_Y$; $Y = \gamma_0 + \gamma_A \cdot A + \gamma_C^{\top} C + \varepsilon_T$. All three are linear regressions, with $A$, $M$ and $Y$ the standardized visceral fat mass, serum uric acid and thigh-to-ankle PWV, and $C$ age, sex, creatinine and brachial blood pressure. $\alpha_A$ is the SD change in uric acid per 1 SD higher visceral fat mass; $\beta_M$ and $\beta_A$ are the mutually adjusted standardized slopes of uric acid and visceral fat mass on PWV; $\gamma_A$ is the total standardized slope of visceral fat mass on PWV adjusted only for $C$; path product $\theta = \alpha_A \cdot \beta_M$.

\textit{Intervals.} $\alpha_A$, $\beta_M$, $\beta_A$, $\gamma_A$: 95\% confidence intervals; $\theta$: bootstrap 95\% percentile interval.
{\footnotesize\setlength{\tabcolsep}{3pt}
\begin{longtable}{@{}>{\raggedright\arraybackslash}p{0.15\linewidth}>{\raggedright\arraybackslash}p{0.40\linewidth}>{\raggedleft\arraybackslash}p{0.14\linewidth}>{\raggedright\arraybackslash}p{0.24\linewidth}@{}}
\toprule
Block & Parameter & Estimate & Interval \\
\midrule\endhead
Path components (additional adjustment for creatinine and brachial blood pressure) & $\alpha_A$: standardized slope of visceral fat mass $\to$ serum uric acid ($A \to M$) & $0.3162$ & $[0.281$, $0.3514]$ \\
 & $\beta_M$: standardized slope of serum uric acid $\to$ PWV adjusted for visceral fat mass ($M \to Y \mid A$) & $0.03626$ & $[-0.00956$, $0.08208]$ \\
 & $\beta_A$: standardized slope of visceral fat mass $\to$ PWV adjusted for serum uric acid ($A \to Y \mid M$) & $0.09908$ & $[0.056$, $0.1422]$ \\
 & $\gamma_A$: adjusted total standardized slope of visceral fat mass $\to$ PWV ($A \to Y$) & $0.1105$ & $[0.07019$, $0.1509]$ \\
Path index (additional adjustment for creatinine and brachial blood pressure) & $\theta = \alpha_A \cdot \beta_M$: path-product association index & $0.01147$ & $[-0.002876$, $0.02695]$ \\
\bottomrule
\end{longtable}}

\paragraph{24.\ Mutually adjusted model: visceral fat and uric acid on PWV ($\pm$30 days)}
\textit{Design.} As in the $\pm$180-day mutually adjusted model, but with visceral adipose tissue mass, serum uric acid and PWV all pairwise within 30 days (endpoints inclusive, either order); for each participant, the set closest in date. PWV as the mean of available sides. Adjustment for approximate age (visceral fat measurement year minus birth year) and recorded sex; visceral fat, uric acid and PWV standardized within the analysis sample.

\textit{Model.} As in the $\pm$180-day mutually adjusted model: $z(Y) = \beta_0 + \beta_A \cdot z(A) + \beta_M \cdot z(M) + \gamma_{\mathrm{age}} \cdot \mathrm{age} + \gamma_{\mathrm{sex}} + \varepsilon$ (linear regression). $\beta_A$ and $\beta_M$ are the SD differences in PWV per 1 SD higher visceral fat and uric acid, mutually adjusted and at fixed age and sex.

\textit{Intervals.} 95\% confidence intervals.
{\footnotesize\setlength{\tabcolsep}{3pt}
\begin{longtable}{@{}>{\raggedright\arraybackslash}p{0.15\linewidth}>{\raggedright\arraybackslash}p{0.40\linewidth}>{\raggedleft\arraybackslash}p{0.14\linewidth}>{\raggedright\arraybackslash}p{0.24\linewidth}@{}}
\toprule
Block & Parameter & Estimate & Interval \\
\midrule\endhead
Mutually adjusted model ($\pm$30 days) & $\beta_A$: standardized coefficient of visceral adipose tissue mass $\to$ PWV adjusted for uric acid ($A \to Y \mid M$) & $0.1833$ & $[0.09173$, $0.2749]$ \\
 & $\beta_M$: standardized coefficient of serum uric acid $\to$ PWV adjusted for visceral fat ($M \to Y \mid A$) & $0.03308$ & $[-0.0628$, $0.129]$ \\
\bottomrule
\end{longtable}}

\paragraph{25.\ Mutually adjusted model: visceral fat and uric acid on PWV ($\pm$365 days)}
\textit{Design.} As in the $\pm$180-day mutually adjusted model, but with visceral adipose tissue mass, serum uric acid and PWV all pairwise within 365 days (endpoints inclusive, either order); for each participant, the set closest in date. PWV as the mean of available sides. Adjustment for approximate age (visceral fat measurement year minus birth year) and recorded sex; visceral fat, uric acid and PWV standardized within the analysis sample.

\textit{Model.} As in the $\pm$180-day mutually adjusted model: $z(Y) = \beta_0 + \beta_A \cdot z(A) + \beta_M \cdot z(M) + \gamma_{\mathrm{age}} \cdot \mathrm{age} + \gamma_{\mathrm{sex}} + \varepsilon$ (linear regression). $\beta_A$ and $\beta_M$ are the SD differences in PWV per 1 SD higher visceral fat and uric acid, mutually adjusted and at fixed age and sex.

\textit{Intervals.} 95\% confidence intervals.
{\footnotesize\setlength{\tabcolsep}{3pt}
\begin{longtable}{@{}>{\raggedright\arraybackslash}p{0.15\linewidth}>{\raggedright\arraybackslash}p{0.40\linewidth}>{\raggedleft\arraybackslash}p{0.14\linewidth}>{\raggedright\arraybackslash}p{0.24\linewidth}@{}}
\toprule
Block & Parameter & Estimate & Interval \\
\midrule\endhead
Mutually adjusted model ($\pm$365 days) & $\beta_A$: standardized coefficient of visceral adipose tissue mass $\to$ PWV adjusted for uric acid ($A \to Y \mid M$) & $0.2123$ & $[0.1676$, $0.2571]$ \\
 & $\beta_M$: standardized coefficient of serum uric acid $\to$ PWV adjusted for visceral fat ($M \to Y \mid A$) & $0.05217$ & $[0.005496$, $0.09884]$ \\
\bottomrule
\end{longtable}}

\paragraph{26.\ Extended adjustment: mutually adjusted model with BMI, systolic blood pressure and creatinine added}
\textit{Design.} Temporal design as in the $\pm$180-day mutually adjusted model (visceral fat, uric acid and PWV all pairwise within 180 days, endpoints inclusive, either order), $n = 2005$. Beyond age and sex, additional adjustment for BMI (the value nearest to the visceral fat measurement date, within 365 days before or after), seated systolic blood pressure (mean of the available values of two readings; the nearest within 180 days before or after), and creatinine from the same blood test as uric acid. Visceral fat, uric acid and PWV standardized within the analysis sample.

\textit{Model.} $z(Y) = \beta_0 + \beta_A \cdot z(A) + \beta_M \cdot z(M) + \gamma_{\mathrm{age}} \cdot \mathrm{age} + \gamma_{\mathrm{sex}} + \gamma_{\mathrm{BMI}} \cdot \mathrm{BMI} + \gamma_{\mathrm{SBP}} \cdot \mathrm{SBP} + \gamma_{\mathrm{Cr}} \cdot \mathrm{Cr} + \varepsilon$ (linear regression; SBP seated systolic blood pressure, Cr serum creatinine). $\beta_A$ and $\beta_M$ are the SD differences in PWV per 1 SD higher visceral fat and uric acid, mutually adjusted and at fixed age, sex, BMI, seated systolic blood pressure and creatinine.

\textit{Intervals.} 95\% confidence intervals.
{\footnotesize\setlength{\tabcolsep}{3pt}
\begin{longtable}{@{}>{\raggedright\arraybackslash}p{0.15\linewidth}>{\raggedright\arraybackslash}p{0.40\linewidth}>{\raggedleft\arraybackslash}p{0.14\linewidth}>{\raggedright\arraybackslash}p{0.24\linewidth}@{}}
\toprule
Block & Parameter & Estimate & Interval \\
\midrule\endhead
Extended adjustment model ($\pm$180 days) & $\beta_A$: standardized coefficient of visceral adipose tissue mass $\to$ PWV adjusted for uric acid ($A \to Y \mid M$) & $0.1098$ & $[0.04906$, $0.1706]$ \\
 & $\beta_M$: standardized coefficient of serum uric acid $\to$ PWV adjusted for visceral fat ($M \to Y \mid A$) & $0.03447$ & $[-0.01929$, $0.08822]$ \\
\bottomrule
\end{longtable}}

\paragraph{27.\ Negative-control outcome: visceral fat, uric acid and the right-minus-left PWV difference}
\textit{Design.} Visceral adipose tissue mass, serum uric acid and PWV measured in the same period: all three pairwise within 180 days (endpoints inclusive, either order); for each participant, the set closest in date. Outcome: the signed difference of right-side minus left-side thigh-to-ankle PWV within the same record, with both sides required to be valid. Adjustment for approximate age (visceral fat measurement year minus birth year) and recorded sex; visceral fat, uric acid and the difference standardized within the analysis sample.

\textit{Model.} $z(D) = \beta_0 + \beta_A \cdot z(A) + \beta_M \cdot z(M) + \gamma_{\mathrm{age}} \cdot \mathrm{age} + \gamma_{\mathrm{sex}} + \varepsilon$ (linear regression; $D$ the right-minus-left thigh-to-ankle PWV difference). $\beta_A$ and $\beta_M$ are the SD differences in $D$ per 1 SD higher visceral fat and uric acid, mutually adjusted and at fixed age and sex.

\textit{Intervals.} 95\% confidence intervals.
{\footnotesize\setlength{\tabcolsep}{3pt}
\begin{longtable}{@{}>{\raggedright\arraybackslash}p{0.15\linewidth}>{\raggedright\arraybackslash}p{0.40\linewidth}>{\raggedleft\arraybackslash}p{0.14\linewidth}>{\raggedright\arraybackslash}p{0.24\linewidth}@{}}
\toprule
Block & Parameter & Estimate & Interval \\
\midrule\endhead
Negative-control outcome & $\beta_A$: standardized coefficient of visceral adipose tissue mass $\to$ right-minus-left PWV difference adjusted for uric acid & $0.0126$ & $[-0.04193$, $0.06714]$ \\
 & $\beta_M$: standardized coefficient of serum uric acid $\to$ right-minus-left PWV difference adjusted for visceral fat & $0.0309$ & $[-0.03188$, $0.09369]$ \\
\bottomrule
\end{longtable}}

\paragraph{28.\ Fat-depot specificity: visceral and subcutaneous fat on PWV}
\textit{Design.} Temporal design as in the $\pm$180-day mutually adjusted model: visceral adipose tissue mass, serum uric acid and PWV all pairwise within 180 days (endpoints inclusive, either order); for each participant, the set closest in date, $n = 2063$. Subcutaneous adipose tissue mass taken from the record of the same whole-body scan as visceral fat. Adjustment for uric acid, approximate age (visceral fat measurement year minus birth year) and recorded sex. Visceral fat, subcutaneous fat, uric acid and PWV standardized within the analysis sample.

\textit{Model.} $z(Y) = \beta_0 + \beta_{\mathrm{VAT}} \cdot z(\mathrm{VAT}) + \beta_{\mathrm{SAT}} \cdot z(\mathrm{SAT}) + \beta_M \cdot z(M) + \gamma_{\mathrm{age}} \cdot \mathrm{age} + \gamma_{\mathrm{sex}} + \varepsilon$ (linear regression; VAT and SAT the visceral and subcutaneous adipose tissue mass). $\beta_{\mathrm{VAT}}$ and $\beta_{\mathrm{SAT}}$ are the SD differences in PWV per 1 SD higher value of each depot, with the other depot, uric acid, age and sex fixed; the difference $\beta_{\mathrm{VAT}} - \beta_{\mathrm{SAT}}$ is also reported.

\textit{Intervals.} 95\% confidence intervals.
{\footnotesize\setlength{\tabcolsep}{3pt}
\begin{longtable}{@{}>{\raggedright\arraybackslash}p{0.15\linewidth}>{\raggedright\arraybackslash}p{0.40\linewidth}>{\raggedleft\arraybackslash}p{0.14\linewidth}>{\raggedright\arraybackslash}p{0.24\linewidth}@{}}
\toprule
Block & Parameter & Estimate & Interval \\
\midrule\endhead
Fat-depot comparison & $\beta_{\mathrm{VAT}}$: standardized coefficient of visceral adipose tissue mass $\to$ PWV adjusted for subcutaneous fat and uric acid & $0.1567$ & $[0.09603$, $0.2173]$ \\
 & $\beta_{\mathrm{SAT}}$: standardized coefficient of subcutaneous adipose tissue mass $\to$ PWV adjusted for visceral fat and uric acid & $0.08028$ & $[0.03105$, $0.1295]$ \\
 & $\beta_{\mathrm{VAT}} - \beta_{\mathrm{SAT}}$: difference between the visceral and subcutaneous fat standardized coefficients & $0.07639$ & $[-0.02025$, $0.173]$ \\
\bottomrule
\end{longtable}}

\FloatBarrier

\subsection{ALT--TG--DBP--PWV}
\label{app:rq-designs:8439}

\paragraph{Research question and measures.}
The research question concerns the serial path association of alanine aminotransferase (ALT) $\to$ triglycerides (TG) $\to$ diastolic blood pressure (DBP) $\to$ pulse wave velocity (PWV), with data from the HPP cohort. The exposure is ALT from blood tests; the mediators are TG from blood tests and seated DBP; the outcome is thigh-to-ankle PWV. ALT, TG, DBP and PWV are all used in the numeric units recorded in the database. Unless otherwise stated, coefficients in analysis units whose variables were measured in date order (including units that allow the same day) are unstandardized slopes in recorded units (the change in the mediator or outcome per 1 unit higher exposure); coefficients in analysis units whose variables were measured within the same period are changes in SD per 1 SD after standardization within the analysis sample of each unit. This research question appears as L7 in Figures~\ref{fig:findings:experimental-atlas} and~\ref{fig:findings:conditional-validation} (Appendix~\ref{app:findings}); pathways T1, T2, X1, I1 and I2 there also come from its analyses.

\subsubsection*{Primary pathways}

\paragraph{1.\ Single-mediator path: ALT $\to$ TG $\to$ thigh-to-ankle PWV}
\textit{Design.} ALT, TG and PWV measured in strict date order: TG 1--365 days after ALT, PWV 30--1095 days after TG (all bounds inclusive); one chain per ALT test date (a participant may contribute several chains), with intervals from resampling of participants as clusters. Adjustment for approximate age in the year of the ALT test and recorded sex. PWV as the mean of available sides; all variables kept in recorded units.

\textit{Model.} $M = \alpha_0 + \alpha_A A + \alpha_{\mathrm{age}}\,\mathrm{age} + f(\mathrm{sex}) + \varepsilon_M$; $Y = \beta_0 + \beta_A A + \beta_M M + \beta_{\mathrm{age}}\,\mathrm{age} + f(\mathrm{sex}) + \varepsilon_Y$. Both equations are linear regressions on the same sample; $A$, $M$ and $Y$ are ALT, TG and thigh-to-ankle PWV (recorded units), $\mathrm{age}$ is approximate age and $f(\mathrm{sex})$ is a sex fixed effect. $\alpha_A$ is the change in TG per 1 unit higher ALT; $\beta_M$ and $\beta_A$ are the mutually adjusted slopes of PWV on TG and ALT; path product $\theta = \alpha_A \cdot \beta_M$, decomposed total $\tau = \beta_A + \theta$.

\textit{Intervals.} Bootstrap 95\% percentile intervals.
{\footnotesize\setlength{\tabcolsep}{3pt}
\begin{longtable}{@{}>{\raggedright\arraybackslash}p{0.15\linewidth}>{\raggedright\arraybackslash}p{0.40\linewidth}>{\raggedleft\arraybackslash}p{0.14\linewidth}>{\raggedright\arraybackslash}p{0.24\linewidth}@{}}
\toprule
Block & Parameter & Estimate & Interval \\
\midrule\endhead
Path components & $\alpha_A$: ALT $\to$ TG slope ($A \to M$) & $0.2205$ & $[0.03882$, $0.7663]$ \\
 & $\beta_M$: TG $\to$ PWV slope adjusted for ALT ($M \to Y \mid A$) & $0.003957$ & $[0.002663$, $0.005308]$ \\
 & $\beta_A$: ALT $\to$ PWV slope adjusted for TG ($A \to Y \mid M$) & $-0.001303$ & $[-0.003546$, $0.005373]$ \\
Path indices & $\theta = \alpha_A \cdot \beta_M$: path-product association index & $0.0008726$ & $[0.0001607$, $0.002892]$ \\
 & $\tau = \beta_A + \theta$: decomposed total & $-0.0004307$ & $[-0.003079$, $0.007891]$ \\
\bottomrule
\end{longtable}}

\paragraph{2.\ Single-mediator path: ALT $\to$ seated DBP $\to$ thigh-to-ankle PWV}
\textit{Design.} ALT, seated blood pressure and PWV measured in strict date order: blood pressure 7--365 days after ALT, PWV 30--1095 days after blood pressure (all bounds inclusive); one chain per ALT test date (a participant may contribute several chains), with intervals from resampling of participants as clusters. Adjustment for approximate age in the year of the ALT test and recorded sex. DBP as the mean of the available values of two seated readings, PWV as the mean of available sides; all variables kept in recorded units.

\textit{Model.} $M = \alpha_0 + \alpha_A A + \alpha_{\mathrm{age}}\,\mathrm{age} + f(\mathrm{sex}) + \varepsilon_M$; $Y = \beta_0 + \beta_A A + \beta_M M + \beta_{\mathrm{age}}\,\mathrm{age} + f(\mathrm{sex}) + \varepsilon_Y$. Both equations are linear regressions on the same sample; $A$, $M$ and $Y$ are ALT, seated DBP and thigh-to-ankle PWV (recorded units), $\mathrm{age}$ is approximate age and $f(\mathrm{sex})$ is a sex fixed effect. $\alpha_A$ is the change in DBP per 1 unit higher ALT; $\beta_M$ and $\beta_A$ are the mutually adjusted slopes of PWV on DBP and ALT; path product $\theta = \alpha_A \cdot \beta_M$, decomposed total $\tau = \beta_A + \theta$.

\textit{Intervals.} Bootstrap 95\% percentile intervals.
{\footnotesize\setlength{\tabcolsep}{3pt}
\begin{longtable}{@{}>{\raggedright\arraybackslash}p{0.15\linewidth}>{\raggedright\arraybackslash}p{0.40\linewidth}>{\raggedleft\arraybackslash}p{0.14\linewidth}>{\raggedright\arraybackslash}p{0.24\linewidth}@{}}
\toprule
Block & Parameter & Estimate & Interval \\
\midrule\endhead
Path components & $\alpha_A$: ALT $\to$ DBP slope ($A \to M$) & $0.1092$ & $[0.0656$, $0.1619]$ \\
 & $\beta_M$: DBP $\to$ PWV slope adjusted for ALT ($M \to Y \mid A$) & $0.05255$ & $[0.04525$, $0.06046]$ \\
 & $\beta_A$: ALT $\to$ PWV slope adjusted for DBP ($A \to Y \mid M$) & $0.005798$ & $[-0.0002208$, $0.01225]$ \\
Path indices & $\theta = \alpha_A \cdot \beta_M$: path-product association index & $0.005739$ & $[0.003361$, $0.008694]$ \\
 & $\tau = \beta_A + \theta$: decomposed total & $0.01154$ & $[0.005263$, $0.01876]$ \\
\bottomrule
\end{longtable}}

\paragraph{3.\ Exposure--outcome association: ALT and thigh-to-ankle PWV}
\textit{Design.} ALT and PWV measured in the same period: PWV within 30 days before or after the ALT blood draw date (inclusive, in either order), with the PWV nearest in date taken for each ALT measurement date, $n = 127$. Adjustment for approximate age at the ALT measurement, sex, and the most recent BMI on or within 365 days before the earlier of the two measurement dates. ALT, PWV, age and BMI standardized within the analysis sample, sex as a categorical term.

\textit{Model.} $z(Y) = \beta_0 + \beta_A z(A) + \beta_{\mathrm{age}} z(\mathrm{age}) + \beta_{\mathrm{BMI}} z(\mathrm{BMI}) + \gamma_{\mathrm{sex}} + \varepsilon$, a linear regression; $A$ is ALT and $Y$ is thigh-to-ankle PWV (mean of available sides). $\beta_A$ is the difference in PWV in SD per 1 SD higher ALT, with age, sex and BMI held fixed.

\textit{Intervals.} 95\% confidence intervals.
{\footnotesize\setlength{\tabcolsep}{3pt}
\begin{longtable}{@{}>{\raggedright\arraybackslash}p{0.14\linewidth}>{\raggedright\arraybackslash}p{0.29\linewidth}>{\raggedleft\arraybackslash}p{0.13\linewidth}>{\raggedright\arraybackslash}p{0.22\linewidth}>{\raggedleft\arraybackslash}p{0.14\linewidth}@{}}
\toprule
Block & Parameter & Estimate & Interval & $P$ \\
\midrule\endhead
ALT--PWV ($\pm 30$ days) & $\beta_A$: standardized ALT--PWV association coefficient & $0.07582$ & $[-0.03959$, $0.1912]$ & \mbox{$0.1959$} \\
\bottomrule
\end{longtable}}

\subsubsection*{Adjacent segments and pairwise associations}

\paragraph{4.\ Adjacent segment: ALT $\to$ TG $\to$ seated DBP}
\textit{Design.} ALT, TG and seated blood pressure measured in strict date order: TG 1--365 days after ALT, blood pressure 7--365 days after TG (all bounds inclusive); one chain per ALT test date (a participant may contribute several chains), with intervals from resampling of participants as clusters. Adjustment for approximate age in the year of the ALT test and recorded sex. DBP as the mean of the available values of two seated readings; all variables kept in recorded units.

\textit{Model.} $M = \alpha_0 + \alpha_A A + \alpha_{\mathrm{age}}\,\mathrm{age} + f(\mathrm{sex}) + \varepsilon_M$; $Y = \beta_0 + \beta_A A + \beta_M M + \beta_{\mathrm{age}}\,\mathrm{age} + f(\mathrm{sex}) + \varepsilon_Y$. Both equations are linear regressions on the same sample; $A$, $M$ and $Y$ are ALT, TG and seated DBP (recorded units), $\mathrm{age}$ is approximate age and $f(\mathrm{sex})$ is a sex fixed effect. $\alpha_A$ is the change in TG per 1 unit higher ALT; $\beta_M$ and $\beta_A$ are the mutually adjusted slopes of DBP on TG and ALT; path product $\theta = \alpha_A \cdot \beta_M$, decomposed total $\tau = \beta_A + \theta$.

\textit{Intervals.} Bootstrap 95\% percentile intervals.
{\footnotesize\setlength{\tabcolsep}{3pt}
\begin{longtable}{@{}>{\raggedright\arraybackslash}p{0.15\linewidth}>{\raggedright\arraybackslash}p{0.40\linewidth}>{\raggedleft\arraybackslash}p{0.14\linewidth}>{\raggedright\arraybackslash}p{0.24\linewidth}@{}}
\toprule
Block & Parameter & Estimate & Interval \\
\midrule\endhead
Path components & $\alpha_A$: ALT $\to$ TG slope ($A \to M$) & $0.07288$ & $[-0.01439$, $0.4265]$ \\
 & $\beta_M$: TG $\to$ DBP slope adjusted for ALT ($M \to Y \mid A$) & $0.03283$ & $[0.0192$, $0.0477]$ \\
 & $\beta_A$: ALT $\to$ DBP slope adjusted for TG ($A \to Y \mid M$) & $0.01594$ & $[-0.002627$, $0.07234]$ \\
Path indices & $\theta = \alpha_A \cdot \beta_M$: path-product association index & $0.002392$ & $[-0.0004518$, $0.01516]$ \\
 & $\tau = \beta_A + \theta$: decomposed total & $0.01833$ & $[-0.002025$, $0.08176]$ \\
\bottomrule
\end{longtable}}

\paragraph{5.\ Adjacent segment: TG $\to$ seated DBP $\to$ thigh-to-ankle PWV}
\textit{Design.} TG, seated blood pressure and PWV measured in strict date order: blood pressure 7--365 days after TG, PWV 30--1095 days after blood pressure (all bounds inclusive); one chain per TG test date, $n = 1663$, with intervals from resampling of participants as clusters. Adjustment for approximate age in the year of the TG test and recorded sex. DBP as the mean of the available values of two seated readings, PWV as the mean of available sides; all variables kept in recorded units.

\textit{Model.} $M = \alpha_0 + \alpha_A A + \alpha_{\mathrm{age}}\,\mathrm{age} + f(\mathrm{sex}) + \varepsilon_M$; $Y = \beta_0 + \beta_A A + \beta_M M + \beta_{\mathrm{age}}\,\mathrm{age} + f(\mathrm{sex}) + \varepsilon_Y$. Both equations are linear regressions on the same sample; $A$, $M$ and $Y$ are TG, seated DBP and thigh-to-ankle PWV (recorded units), $\mathrm{age}$ is approximate age and $f(\mathrm{sex})$ is a sex fixed effect. $\alpha_A$ is the change in DBP per 1 unit higher TG; $\beta_M$ and $\beta_A$ are the mutually adjusted slopes of PWV on DBP and TG; path product $\theta = \alpha_A \cdot \beta_M$, decomposed total $\tau = \beta_A + \theta$.

\textit{Intervals.} Bootstrap 95\% percentile intervals.
{\footnotesize\setlength{\tabcolsep}{3pt}
\begin{longtable}{@{}>{\raggedright\arraybackslash}p{0.15\linewidth}>{\raggedright\arraybackslash}p{0.40\linewidth}>{\raggedleft\arraybackslash}p{0.14\linewidth}>{\raggedright\arraybackslash}p{0.24\linewidth}@{}}
\toprule
Block & Parameter & Estimate & Interval \\
\midrule\endhead
Path components & $\alpha_A$: TG $\to$ DBP slope ($A \to M$) & $0.03807$ & $[0.03027$, $0.04673]$ \\
 & $\beta_M$: DBP $\to$ PWV slope adjusted for TG ($M \to Y \mid A$) & $0.051$ & $[0.04384$, $0.05853]$ \\
 & $\beta_A$: TG $\to$ PWV slope adjusted for DBP ($A \to Y \mid M$) & $0.001475$ & $[0.0003427$, $0.002539]$ \\
Path indices & $\theta = \alpha_A \cdot \beta_M$: path-product association index & $0.001942$ & $[0.001466$, $0.002483]$ \\
 & $\tau = \beta_A + \theta$: decomposed total & $0.003417$ & $[0.002249$, $0.004612]$ \\
\bottomrule
\end{longtable}}

\paragraph{6.\ Pairwise association: ALT $\to$ TG (same blood test)}
\textit{Design.} ALT and TG from the same blood test (measured on the same day, 0-day gap); one row per test date (a participant may contribute several rows), $n = 768$. Adjustment for approximate age on the test date, recorded sex, and the most recent BMI on or within 365 days before that date. ALT, TG, age and BMI on their original recorded scales, sex as a categorical term.

\textit{Model.} $M_1 = \beta_0 + \beta_{AM_1} A + \lambda\,\mathrm{Age} + f(\mathrm{Sex}) + \kappa\,\mathrm{BMI} + \varepsilon$ (linear regression; $A$ is alanine aminotransferase (ALT) and $M_1$ is TG, both on original recorded scales; $\mathrm{Age}$ is approximate age and $f(\mathrm{Sex})$ is the registered-sex main effect). $\beta_{AM_1}$ is the difference in recorded TG value per 1 recorded unit higher ALT, with age, sex and BMI held fixed.

\textit{Intervals.} Bootstrap 95\% percentile intervals.
{\footnotesize\setlength{\tabcolsep}{3pt}
\begin{longtable}{@{}>{\raggedright\arraybackslash}p{0.14\linewidth}>{\raggedright\arraybackslash}p{0.29\linewidth}>{\raggedleft\arraybackslash}p{0.13\linewidth}>{\raggedright\arraybackslash}p{0.22\linewidth}>{\raggedleft\arraybackslash}p{0.14\linewidth}@{}}
\toprule
Block & Parameter & Estimate & Interval & $P$ \\
\midrule\endhead
Same blood test (0-day gap) & Adjusted slope $\beta_{AM_1}$ (ALT $\to$ TG) & $0.5177$ & $[0.2391$, $0.9076]$ & \mbox{$0$} \\
\bottomrule
\end{longtable}}

\paragraph{7.\ Pairwise association: ALT--TG (same-day blood test)}
\textit{Design.} ALT and TG from blood tests on the same blood draw date (measured on the same day), one row per blood draw date, $n = 768$. Adjustment for approximate age at the blood draw, sex, and the most recent BMI on or within 365 days before the blood draw date. ALT, TG, age and BMI standardized within the analysis sample, sex as a categorical term.

\textit{Model.} $z(M_1) = \beta_0 + \beta_A z(A) + \beta_{\mathrm{age}} z(\mathrm{age}) + \beta_{\mathrm{BMI}} z(\mathrm{BMI}) + \gamma_{\mathrm{sex}} + \varepsilon$, a linear regression; $A$ is ALT and $M_1$ is TG. $\beta_A$ is the difference in TG in SD per 1 SD higher ALT, with age, sex and BMI held fixed.

\textit{Intervals.} 95\% confidence intervals.
{\footnotesize\setlength{\tabcolsep}{3pt}
\begin{longtable}{@{}>{\raggedright\arraybackslash}p{0.14\linewidth}>{\raggedright\arraybackslash}p{0.29\linewidth}>{\raggedleft\arraybackslash}p{0.13\linewidth}>{\raggedright\arraybackslash}p{0.22\linewidth}>{\raggedleft\arraybackslash}p{0.14\linewidth}@{}}
\toprule
Block & Parameter & Estimate & Interval & $P$ \\
\midrule\endhead
ALT--TG (same day) & $\beta_A$: standardized ALT--TG association coefficient & $0.156$ & $[0.05355$, $0.2584]$ & \mbox{$0.002892$} \\
\bottomrule
\end{longtable}}

\paragraph{8.\ Pairwise association: ALT--seated DBP ($\pm 30$ days)}
\textit{Design.} ALT and seated blood pressure measured in the same period: blood pressure within 30 days before or after the ALT blood draw date (inclusive, in either order), with the blood pressure nearest in date taken for each ALT measurement date, $n = 137$. Adjustment for approximate age at the ALT measurement, sex, and the most recent BMI on or within 365 days before the earlier of the two measurement dates. ALT, DBP, age and BMI standardized within the analysis sample, sex as a categorical term.

\textit{Model.} $z(M_2) = \beta_0 + \beta_A z(A) + \beta_{\mathrm{age}} z(\mathrm{age}) + \beta_{\mathrm{BMI}} z(\mathrm{BMI}) + \gamma_{\mathrm{sex}} + \varepsilon$, a linear regression; $A$ is ALT and $M_2$ is seated DBP (mean of the available values of two readings). $\beta_A$ is the difference in DBP in SD per 1 SD higher ALT, with age, sex and BMI held fixed.

\textit{Intervals.} 95\% confidence intervals.
{\footnotesize\setlength{\tabcolsep}{3pt}
\begin{longtable}{@{}>{\raggedright\arraybackslash}p{0.14\linewidth}>{\raggedright\arraybackslash}p{0.29\linewidth}>{\raggedleft\arraybackslash}p{0.13\linewidth}>{\raggedright\arraybackslash}p{0.22\linewidth}>{\raggedleft\arraybackslash}p{0.14\linewidth}@{}}
\toprule
Block & Parameter & Estimate & Interval & $P$ \\
\midrule\endhead
ALT--DBP ($\pm 30$ days) & $\beta_A$: standardized ALT--seated DBP association coefficient & $-0.0346$ & $[-0.166$, $0.09685]$ & \mbox{$0.6035$} \\
\bottomrule
\end{longtable}}

\paragraph{9.\ Pairwise association: TG--seated DBP ($\pm 30$ days)}
\textit{Design.} TG and seated blood pressure measured in the same period: blood pressure within 30 days before or after the TG blood draw date (inclusive, in either order), with the blood pressure nearest in date taken for each blood draw date. Adjustment for approximate age at the blood draw, sex, and the most recent BMI on or within 365 days before the earlier of the two measurement dates. TG, DBP, age and BMI standardized within the analysis sample, sex as a categorical term.

\textit{Model.} $z(M_2) = \beta_0 + \beta_{M_1} z(M_1) + \beta_{\mathrm{age}} z(\mathrm{age}) + \beta_{\mathrm{BMI}} z(\mathrm{BMI}) + \gamma_{\mathrm{sex}} + \varepsilon$, a linear regression; $M_1$ is TG and $M_2$ is seated DBP (mean of the available values of two readings). $\beta_{M_1}$ is the difference in DBP in SD per 1 SD higher TG, with age, sex and BMI held fixed.

\textit{Intervals.} 95\% confidence intervals.
{\footnotesize\setlength{\tabcolsep}{3pt}
\begin{longtable}{@{}>{\raggedright\arraybackslash}p{0.14\linewidth}>{\raggedright\arraybackslash}p{0.29\linewidth}>{\raggedleft\arraybackslash}p{0.13\linewidth}>{\raggedright\arraybackslash}p{0.22\linewidth}>{\raggedleft\arraybackslash}p{0.14\linewidth}@{}}
\toprule
Block & Parameter & Estimate & Interval & $P$ \\
\midrule\endhead
TG--DBP ($\pm 30$ days) & $\beta_{M_1}$: standardized TG--seated DBP association coefficient & $0.1163$ & $[-0.05628$, $0.2888]$ & \mbox{$0.1847$} \\
\bottomrule
\end{longtable}}

\paragraph{10.\ Pairwise association: TG--PWV ($\pm 30$ days)}
\textit{Design.} TG and PWV measured in the same period: PWV within 30 days before or after the TG blood draw date (inclusive, in either order), with the PWV nearest in date taken for each blood draw date, $n = 118$. Adjustment for approximate age at the blood draw, sex, and the most recent BMI on or within 365 days before the earlier of the two measurement dates. TG, PWV, age and BMI standardized within the analysis sample, sex as a categorical term.

\textit{Model.} $z(Y) = \beta_0 + \beta_{M_1} z(M_1) + \beta_{\mathrm{age}} z(\mathrm{age}) + \beta_{\mathrm{BMI}} z(\mathrm{BMI}) + \gamma_{\mathrm{sex}} + \varepsilon$, a linear regression; $M_1$ is TG and $Y$ is thigh-to-ankle PWV (mean of available sides). $\beta_{M_1}$ is the difference in PWV in SD per 1 SD higher TG, with age, sex and BMI held fixed.

\textit{Intervals.} 95\% confidence intervals.
{\footnotesize\setlength{\tabcolsep}{3pt}
\begin{longtable}{@{}>{\raggedright\arraybackslash}p{0.14\linewidth}>{\raggedright\arraybackslash}p{0.29\linewidth}>{\raggedleft\arraybackslash}p{0.13\linewidth}>{\raggedright\arraybackslash}p{0.22\linewidth}>{\raggedleft\arraybackslash}p{0.14\linewidth}@{}}
\toprule
Block & Parameter & Estimate & Interval & $P$ \\
\midrule\endhead
TG--PWV ($\pm 30$ days) & $\beta_{M_1}$: standardized TG--PWV association coefficient & $0.0318$ & $[-0.1351$, $0.1987]$ & \mbox{$0.7066$} \\
\bottomrule
\end{longtable}}

\paragraph{11.\ Pairwise association: seated DBP--PWV ($\pm 30$ days)}
\textit{Design.} Seated blood pressure and PWV measured in the same period: PWV within 30 days before or after the blood pressure measurement date (inclusive, in either order), with the PWV nearest in date taken for each blood pressure measurement date, $n = 10903$. Adjustment for approximate age at the blood pressure measurement, sex, and the most recent BMI on or within 365 days before the earlier of the two measurement dates. DBP, PWV, age and BMI standardized within the analysis sample, sex as a categorical term.

\textit{Model.} $z(Y) = \beta_0 + \beta_{M_2} z(M_2) + \beta_{\mathrm{age}} z(\mathrm{age}) + \beta_{\mathrm{BMI}} z(\mathrm{BMI}) + \gamma_{\mathrm{sex}} + \varepsilon$, a linear regression; $M_2$ is seated DBP (mean of the available values of two readings) and $Y$ is thigh-to-ankle PWV (mean of available sides). $\beta_{M_2}$ is the difference in PWV in SD per 1 SD higher DBP, with age, sex and BMI held fixed.

\textit{Intervals.} 95\% confidence intervals.
{\footnotesize\setlength{\tabcolsep}{3pt}
\begin{longtable}{@{}>{\raggedright\arraybackslash}p{0.14\linewidth}>{\raggedright\arraybackslash}p{0.29\linewidth}>{\raggedleft\arraybackslash}p{0.13\linewidth}>{\raggedright\arraybackslash}p{0.22\linewidth}>{\raggedleft\arraybackslash}p{0.14\linewidth}@{}}
\toprule
Block & Parameter & Estimate & Interval & $P$ \\
\midrule\endhead
DBP--PWV ($\pm 30$ days) & $\beta_{M_2}$: standardized seated DBP--PWV association coefficient & $0.4487$ & $[0.4299$, $0.4675]$ & \mbox{$0$} \\
\bottomrule
\end{longtable}}

\subsubsection*{Joint models}

\paragraph{12.\ Joint model: ALT, TG, DBP and PWV ($\pm 30$ days)}
\textit{Design.} ALT and TG as same-day blood test values; seated DBP as the nearest measurement within 30 days before or after the blood draw date, PWV as the measurement nearest to the DBP date and within 30 days of both the DBP date and the blood draw date (inclusive, in either order); one measurement chain per blood draw date, $n = 108$. Adjustment for approximate age at the blood draw, sex, and the most recent BMI on or within 365 days before the earliest date of the chain. Continuous variables standardized within the analysis sample, sex as a categorical term.

\textit{Model.} $z(Y) = \beta_0 + \beta_A z(A) + \beta_{M_1} z(M_1) + \beta_{M_2} z(M_2) + \beta_{\mathrm{age}} z(\mathrm{age}) + \beta_{\mathrm{BMI}} z(\mathrm{BMI}) + \gamma_{\mathrm{sex}} + \varepsilon$, a linear regression; $A$, $M_1$, $M_2$ and $Y$ are ALT, TG, seated DBP (mean of the available values of two readings) and thigh-to-ankle PWV (mean of available sides). $\beta_A$, $\beta_{M_1}$ and $\beta_{M_2}$ are the differences in PWV in SD per 1 SD higher value of that variable, with the other terms of the model held fixed.

\textit{Intervals.} 95\% confidence intervals; $P$ values not adjusted for multiple comparisons.
{\footnotesize\setlength{\tabcolsep}{3pt}
\begin{longtable}{@{}>{\raggedright\arraybackslash}p{0.14\linewidth}>{\raggedright\arraybackslash}p{0.29\linewidth}>{\raggedleft\arraybackslash}p{0.13\linewidth}>{\raggedright\arraybackslash}p{0.22\linewidth}>{\raggedleft\arraybackslash}p{0.14\linewidth}@{}}
\toprule
Block & Parameter & Estimate & Interval & $P$ \\
\midrule\endhead
Joint model ($\pm 30$ days) & $\beta_A$: ALT coefficient (adjusted for TG and DBP) & $0.09657$ & $[0.006087$, $0.1871]$ & \mbox{$0.03669$} \\
 & $\beta_{M_1}$: TG coefficient (adjusted for ALT and DBP) & $-0.01479$ & $[-0.1413$, $0.1117]$ & \mbox{$0.8172$} \\
 & $\beta_{M_2}$: seated DBP coefficient (adjusted for ALT and TG) & $0.5085$ & $[0.3682$, $0.6489]$ & \mbox{$9.524\times10^{-11}$} \\
\bottomrule
\end{longtable}}

\paragraph{13.\ ALT coefficient before and after adding TG and DBP ($\pm 30$ days)}
\textit{Design.} Measurement chains as in the $\pm 30$-day joint model: ALT and TG from the same-day blood test, DBP within 30 days before or after the blood draw date, PWV within 30 days of both the blood draw date and the DBP date (inclusive, in either order), $n = 108$. Both models use the same sample and adjust for approximate age at the blood draw, sex, and the most recent BMI on or within 365 days before the earliest date of the chain; continuous variables standardized once within this sample and shared by both models.

\textit{Model.} Reduced model: $z(Y) = \beta_0 + \beta_A^{R} z(A) + \beta_{\mathrm{age}} z(\mathrm{age}) + \beta_{\mathrm{BMI}} z(\mathrm{BMI}) + \gamma_{\mathrm{sex}} + \varepsilon$; the full model additionally includes $\beta_{M_1} z(M_1) + \beta_{M_2} z(M_2)$, with its ALT coefficient denoted $\beta_A^{F}$. Both are linear regressions; $A$, $M_1$, $M_2$ and $Y$ are ALT, TG, seated DBP and thigh-to-ankle PWV. $\beta_A^{R}$ and $\beta_A^{F}$ are the differences in PWV in SD per 1 SD higher ALT before and after adding TG and DBP.

\textit{Intervals.} 95\% confidence intervals; $P$ values not adjusted for multiple comparisons.
{\footnotesize\setlength{\tabcolsep}{3pt}
\begin{longtable}{@{}>{\raggedright\arraybackslash}p{0.14\linewidth}>{\raggedright\arraybackslash}p{0.29\linewidth}>{\raggedleft\arraybackslash}p{0.13\linewidth}>{\raggedright\arraybackslash}p{0.22\linewidth}>{\raggedleft\arraybackslash}p{0.14\linewidth}@{}}
\toprule
Block & Parameter & Estimate & Interval & $P$ \\
\midrule\endhead
Reduced model: ALT + age + sex + BMI & $\beta_A^{R}$: ALT coefficient (reduced model) & $0.06504$ & $[-0.06611$, $0.1962]$ & \mbox{$0.3278$} \\
Full model: ALT + TG + DBP + age + sex + BMI & $\beta_A^{F}$: ALT coefficient (full model) & $0.09657$ & $[0.006087$, $0.1871]$ & \mbox{$0.03669$} \\
\bottomrule
\end{longtable}}

\subsubsection*{Reverse direction}

\paragraph{14.\ Reverse path: thigh-to-ankle PWV $\to$ TG $\to$ ALT}
\textit{Design.} PWV, TG and ALT measured in strict date order: TG 30--1095 days after PWV, ALT 1--365 days after TG (all bounds inclusive); one chain per PWV measurement date (a participant may contribute several chains), with intervals from resampling of participants as clusters. Adjustment for approximate age in the year of the PWV measurement and recorded sex. PWV as the mean of available sides; all variables kept in recorded units.

\textit{Model.} $M = \alpha_0 + \alpha_A A + \alpha_{\mathrm{age}}\,\mathrm{age} + f(\mathrm{sex}) + \varepsilon_M$; $Y = \beta_0 + \beta_A A + \beta_M M + \beta_{\mathrm{age}}\,\mathrm{age} + f(\mathrm{sex}) + \varepsilon_Y$. Both equations are linear regressions on the same sample; $A$, $M$ and $Y$ are thigh-to-ankle PWV, TG and ALT (recorded units), $\mathrm{age}$ is approximate age and $f(\mathrm{sex})$ is a sex fixed effect. $\alpha_A$ is the change in TG per 1 unit higher PWV; $\beta_M$ and $\beta_A$ are the mutually adjusted slopes of ALT on TG and PWV; path product $\theta = \alpha_A \cdot \beta_M$, decomposed total $\tau = \beta_A + \theta$.

\textit{Intervals.} Bootstrap 95\% percentile intervals.
{\footnotesize\setlength{\tabcolsep}{3pt}
\begin{longtable}{@{}>{\raggedright\arraybackslash}p{0.15\linewidth}>{\raggedright\arraybackslash}p{0.40\linewidth}>{\raggedleft\arraybackslash}p{0.14\linewidth}>{\raggedright\arraybackslash}p{0.24\linewidth}@{}}
\toprule
Block & Parameter & Estimate & Interval \\
\midrule\endhead
Reverse path components & $\alpha_A$: PWV $\to$ TG slope ($A \to M$) & $3.88$ & $[-1.528$, $10.5]$ \\
 & $\beta_M$: TG $\to$ ALT slope adjusted for PWV ($M \to Y \mid A$) & $0.005503$ & $[0.002258$, $0.1143]$ \\
 & $\beta_A$: PWV $\to$ ALT slope adjusted for TG ($A \to Y \mid M$) & $0.4112$ & $[-0.5419$, $1.218]$ \\
Reverse path indices & $\theta = \alpha_A \cdot \beta_M$: path-product association index & $0.02135$ & $[-0.008912$, $0.5664]$ \\
 & $\tau = \beta_A + \theta$: decomposed total & $0.4325$ & $[-0.4403$, $1.369]$ \\
\bottomrule
\end{longtable}}

\paragraph{15.\ Reverse path: thigh-to-ankle PWV $\to$ seated DBP $\to$ ALT}
\textit{Design.} PWV, seated blood pressure and ALT measured in strict date order: blood pressure 30--1095 days after PWV, ALT 7--365 days after blood pressure (all bounds inclusive); one chain per PWV measurement date (a participant may contribute several chains), with intervals from resampling of participants as clusters. Adjustment for approximate age in the year of the PWV measurement and recorded sex. DBP as the mean of the available values of two seated readings, PWV as the mean of available sides; all variables kept in recorded units.

\textit{Model.} $M = \alpha_0 + \alpha_A A + \alpha_{\mathrm{age}}\,\mathrm{age} + f(\mathrm{sex}) + \varepsilon_M$; $Y = \beta_0 + \beta_A A + \beta_M M + \beta_{\mathrm{age}}\,\mathrm{age} + f(\mathrm{sex}) + \varepsilon_Y$. Both equations are linear regressions on the same sample; $A$, $M$ and $Y$ are thigh-to-ankle PWV, seated DBP and ALT (recorded units), $\mathrm{age}$ is approximate age and $f(\mathrm{sex})$ is a sex fixed effect. $\alpha_A$ is the change in DBP per 1 unit higher PWV; $\beta_M$ and $\beta_A$ are the mutually adjusted slopes of ALT on DBP and PWV; path product $\theta = \alpha_A \cdot \beta_M$, decomposed total $\tau = \beta_A + \theta$.

\textit{Intervals.} Bootstrap 95\% percentile intervals.
{\footnotesize\setlength{\tabcolsep}{3pt}
\begin{longtable}{@{}>{\raggedright\arraybackslash}p{0.15\linewidth}>{\raggedright\arraybackslash}p{0.40\linewidth}>{\raggedleft\arraybackslash}p{0.14\linewidth}>{\raggedright\arraybackslash}p{0.24\linewidth}@{}}
\toprule
Block & Parameter & Estimate & Interval \\
\midrule\endhead
Reverse path indices & $\tau = \beta_A + \theta$: decomposed total & $0.2407$ & $[-1.991$, $1.903]$ \\
 & $\theta = \alpha_A \cdot \beta_M$: path-product association index & $0.2099$ & $[-0.6153$, $1.018]$ \\
Reverse path components & $\alpha_A$: PWV $\to$ DBP slope ($A \to M$) & $2.152$ & $[0.8433$, $3.646]$ \\
 & $\beta_A$: PWV $\to$ ALT slope adjusted for DBP ($A \to Y \mid M$) & $0.03087$ & $[-1.908$, $1.601]$ \\
 & $\beta_M$: DBP $\to$ ALT slope adjusted for PWV ($M \to Y \mid A$) & $0.09753$ & $[-0.2659$, $0.403]$ \\
\bottomrule
\end{longtable}}

\paragraph{16.\ Serial path with reversed measurement order: ALT $\to$ TG $\to$ DBP $\to$ PWV}
\textit{Design.} Measurement order reversed, in strict date order: seated blood pressure 1--365 days before the blood test, PWV 1--730 days before that blood pressure (the most recent of each, all bounds inclusive); ALT and TG from the same test. One chain per test date, $n = 33$. Adjustment for approximate age on the PWV measurement date, recorded sex, and the most recent BMI on or within 365 days before that date. DBP as the mean of seated readings, PWV as the mean of available sides; all variables on original recorded scales.

\textit{Model.} $M_1 = \alpha_1 + \beta_{AM_1} A + \gamma_1^{\top} C + \varepsilon_1$; $M_2 = \alpha_2 + \delta_2 A + \beta_{M_1M_2} M_1 + \gamma_2^{\top} C + \varepsilon_2$; $Y = \alpha_3 + \delta_3 A + \delta_4 M_1 + \beta_{M_2Y} M_2 + \gamma_3^{\top} C + \varepsilon_3$. Variable roles unchanged; all three equations are linear regressions on the same sample; $A$, $M_1$, $M_2$ and $Y$ are alanine aminotransferase (ALT), TG, seated DBP and thigh-to-ankle PWV (original recorded scales), and $C$ comprises approximate age, recorded sex and BMI. $\beta_{AM_1}$, $\beta_{M_1M_2}$ and $\beta_{M_2Y}$ are the differences in the recorded value of the equation's outcome per 1 recorded unit higher value of that regressor, with the other terms of the same equation held fixed; serial path product $\theta = \beta_{AM_1} \cdot \beta_{M_1M_2} \cdot \beta_{M_2Y}$.

\textit{Intervals.} Bootstrap 95\% percentile intervals. $P$ values not adjusted for multiple comparisons.
{\footnotesize\setlength{\tabcolsep}{3pt}
\begin{longtable}{@{}>{\raggedright\arraybackslash}p{0.14\linewidth}>{\raggedright\arraybackslash}p{0.29\linewidth}>{\raggedleft\arraybackslash}p{0.13\linewidth}>{\raggedright\arraybackslash}p{0.22\linewidth}>{\raggedleft\arraybackslash}p{0.14\linewidth}@{}}
\toprule
Block & Parameter & Estimate & Interval & $P$ \\
\midrule\endhead
Path indices (reversed measurement order) & $\theta = \beta_{AM_1} \cdot \beta_{M_1M_2} \cdot \beta_{M_2Y}$: serial path product & $0.001286$ & $[-0.002474$, $0.007358]$ & \mbox{$0.559$} \\
Path components (reversed measurement order) & $\beta_{AM_1}$: adjusted ALT $\to$ TG slope ($A \to M_1$) & $1.198$ & $[-0.4278$, $3.571]$ & \mbox{$0.156$} \\
 & $\beta_{M_1M_2}$: TG $\to$ seated DBP slope adjusted for ALT ($M_1 \to M_2 \mid A$) & $0.02561$ & $[-0.04714$, $0.0689]$ & \mbox{$0.417$} \\
 & $\beta_{M_2Y}$: seated DBP $\to$ PWV slope adjusted for ALT and TG ($M_2 \to Y \mid A, M_1$) & $0.04193$ & $[-0.005673$, $0.08644]$ & \mbox{$0.088$} \\
\bottomrule
\end{longtable}}

\paragraph{17.\ Reverse pairing: earlier PWV and later ALT}
\textit{Design.} PWV measured 31--365 days before the ALT blood draw date (inclusive), with the most recent earlier PWV taken for each ALT measurement date, $n = 606$. Adjustment for approximate age at the ALT measurement, sex, and the most recent BMI on or within 365 days before the PWV measurement date. ALT, PWV, age and BMI standardized within the analysis sample, sex as a categorical term.

\textit{Model.} $z(Y_{\mathrm{early}}) = \beta_0 + \beta_A z(A_{\mathrm{late}}) + \beta_{\mathrm{age}} z(\mathrm{age}) + \beta_{\mathrm{BMI}} z(\mathrm{BMI}) + \gamma_{\mathrm{sex}} + \varepsilon$, a linear regression; $A$ is ALT and $Y$ is thigh-to-ankle PWV (mean of available sides). $\beta_A$ is the difference in earlier PWV in SD per 1 SD higher later-measured ALT, with age, sex and BMI held fixed.

\textit{Intervals.} 95\% confidence intervals.
{\footnotesize\setlength{\tabcolsep}{3pt}
\begin{longtable}{@{}>{\raggedright\arraybackslash}p{0.14\linewidth}>{\raggedright\arraybackslash}p{0.29\linewidth}>{\raggedleft\arraybackslash}p{0.13\linewidth}>{\raggedright\arraybackslash}p{0.22\linewidth}>{\raggedleft\arraybackslash}p{0.14\linewidth}@{}}
\toprule
Block & Parameter & Estimate & Interval & $P$ \\
\midrule\endhead
Reverse pairing (PWV 31--365 days before ALT) & $\beta_A$: standardized later ALT--earlier PWV association coefficient (PWV 31--365 days before ALT) & $0.02872$ & $[-0.07315$, $0.1306]$ & \mbox{$0.5799$} \\
\bottomrule
\end{longtable}}

\subsubsection*{Heterogeneity and interaction}

\paragraph{18.\ Sex heterogeneity: ALT $\to$ TG $\to$ thigh-to-ankle PWV}
\textit{Design.} ALT, TG and PWV measured in strict date order: TG 1--365 days after ALT, PWV 30--1095 days after TG (all bounds inclusive); one chain per ALT test date, with intervals from resampling of participants as clusters. In both equations the slopes of ALT and TG are estimated separately by recorded sex (first and second category), with a shared approximate-age term and sex main effect. PWV as the mean of available sides; all variables kept in recorded units.

\textit{Model.} $M = \alpha_0 + \sum_s \mathbf{1}(\mathrm{sex}=s)\,\alpha_{A,s} A + \alpha_{\mathrm{age}}\,\mathrm{age} + f(\mathrm{sex}) + \varepsilon_M$; $Y = \beta_0 + \sum_s \mathbf{1}(\mathrm{sex}=s)\,(\beta_{A,s} A + \beta_{M,s} M) + \beta_{\mathrm{age}}\,\mathrm{age} + f(\mathrm{sex}) + \varepsilon_Y$. Both equations are linear regressions on the same sample; $A$, $M$ and $Y$ are ALT, TG and thigh-to-ankle PWV (recorded units), and $s$ is the recorded sex category. Reported quantities: within category $s$, the ALT $\to$ TG slope $\alpha_{A,s}$, the TG $\to$ PWV slope $\beta_{M,s}$ adjusted for ALT and the ALT $\to$ PWV slope $\beta_{A,s}$ adjusted for TG; the path product $\theta_s = \alpha_{A,s} \cdot \beta_{M,s}$ and decomposed total $\tau_s = \beta_{A,s} + \theta_s$; and the sex difference of each quantity (second $-$ first category).

\textit{Intervals.} Bootstrap 95\% percentile intervals.
{\footnotesize\setlength{\tabcolsep}{3pt}
\begin{longtable}{@{}>{\raggedright\arraybackslash}p{0.15\linewidth}>{\raggedright\arraybackslash}p{0.40\linewidth}>{\raggedleft\arraybackslash}p{0.14\linewidth}>{\raggedright\arraybackslash}p{0.24\linewidth}@{}}
\toprule
Block & Parameter & Estimate & Interval \\
\midrule\endhead
First sex category & $\alpha_{A,1}$: ALT $\to$ TG slope within the first sex category & $-0.0054$ & $[-0.07181$, $0.3416]$ \\
 & $\beta_{M,1}$: TG $\to$ PWV slope adjusted for ALT within the first sex category & $0.004593$ & $[0.002675$, $0.006541]$ \\
 & $\beta_{A,1}$: ALT $\to$ PWV slope adjusted for TG within the first sex category & $-0.003222$ & $[-0.004911$, $0.003067]$ \\
 & $\theta_1 = \alpha_{A,1} \cdot \beta_{M,1}$: path-product association index, first sex category & $-2.481\times10^{-5}$ & $[-0.0003598$, $0.001549]$ \\
 & $\tau_1 = \beta_{A,1} + \theta_1$: decomposed total, first sex category & $-0.003247$ & $[-0.00478$, $0.004187]$ \\
Second sex category & $\alpha_{A,2}$: ALT $\to$ TG slope within the second sex category & $1.015$ & $[0.5418$, $1.588]$ \\
 & $\beta_{M,2}$: TG $\to$ PWV slope adjusted for ALT within the second sex category & $0.003239$ & $[0.001675$, $0.005109]$ \\
 & $\beta_{A,2}$: ALT $\to$ PWV slope adjusted for TG within the second sex category & $0.00617$ & $[-0.001837$, $0.01355]$ \\
 & $\theta_2 = \alpha_{A,2} \cdot \beta_{M,2}$: path-product association index, second sex category & $0.003287$ & $[0.00156$, $0.005673]$ \\
 & $\tau_2 = \beta_{A,2} + \theta_2$: decomposed total, second sex category & $0.009457$ & $[0.001339$, $0.01699]$ \\
Sex difference (second $-$ first category) & $\alpha_{A,2} - \alpha_{A,1}$: sex difference in the ALT $\to$ TG slope & $1.02$ & $[0.4391$, $1.576]$ \\
 & $\beta_{M,2} - \beta_{M,1}$: sex difference in the conditional TG $\to$ PWV slope & $-0.001355$ & $[-0.003893$, $0.001438]$ \\
 & $\beta_{A,2} - \beta_{A,1}$: sex difference in the conditional ALT $\to$ PWV slope & $0.009392$ & $[-0.0009787$, $0.01675]$ \\
 & $\theta_2 - \theta_1$: sex difference in the path product & $0.003311$ & $[0.000969$, $0.005602]$ \\
 & $\tau_2 - \tau_1$: sex difference in the decomposed total & $0.0127$ & $[0.002034$, $0.02023]$ \\
\bottomrule
\end{longtable}}

\paragraph{19.\ Sex heterogeneity: ALT $\to$ seated DBP $\to$ thigh-to-ankle PWV}
\textit{Design.} ALT, seated blood pressure and PWV measured in strict date order: blood pressure 7--365 days after ALT, PWV 30--1095 days after blood pressure (all bounds inclusive); one chain per ALT test date, with intervals from resampling of participants as clusters. In both equations the slopes of ALT and DBP are estimated separately by recorded sex (first and second category), with a shared approximate-age term and sex main effect. DBP as the mean of the available values of two seated readings, PWV as the mean of available sides; all variables kept in recorded units.

\textit{Model.} $M = \alpha_0 + \sum_s \mathbf{1}(\mathrm{sex}=s)\,\alpha_{A,s} A + \alpha_{\mathrm{age}}\,\mathrm{age} + f(\mathrm{sex}) + \varepsilon_M$; $Y = \beta_0 + \sum_s \mathbf{1}(\mathrm{sex}=s)\,(\beta_{A,s} A + \beta_{M,s} M) + \beta_{\mathrm{age}}\,\mathrm{age} + f(\mathrm{sex}) + \varepsilon_Y$. Both equations are linear regressions on the same sample; $A$, $M$ and $Y$ are ALT, seated DBP and thigh-to-ankle PWV (recorded units), and $s$ is the recorded sex category. Reported quantities: within category $s$, the ALT $\to$ DBP slope $\alpha_{A,s}$, the DBP $\to$ PWV slope $\beta_{M,s}$ adjusted for ALT and the ALT $\to$ PWV slope $\beta_{A,s}$ adjusted for DBP; the path product $\theta_s = \alpha_{A,s} \cdot \beta_{M,s}$ and decomposed total $\tau_s = \beta_{A,s} + \theta_s$; and the sex difference of each quantity (second $-$ first category).

\textit{Intervals.} Bootstrap 95\% percentile intervals.
{\footnotesize\setlength{\tabcolsep}{3pt}
\begin{longtable}{@{}>{\raggedright\arraybackslash}p{0.15\linewidth}>{\raggedright\arraybackslash}p{0.40\linewidth}>{\raggedleft\arraybackslash}p{0.14\linewidth}>{\raggedright\arraybackslash}p{0.24\linewidth}@{}}
\toprule
Block & Parameter & Estimate & Interval \\
\midrule\endhead
First sex category & $\alpha_{A,1}$: ALT $\to$ DBP slope within the first sex category & $0.09475$ & $[0.02666$, $0.1702]$ \\
 & $\beta_{M,1}$: DBP $\to$ PWV slope adjusted for ALT within the first sex category & $0.06031$ & $[0.05079$, $0.07116]$ \\
 & $\beta_{A,1}$: ALT $\to$ PWV slope adjusted for DBP within the first sex category & $0.008939$ & $[0.0007023$, $0.01882]$ \\
 & $\theta_1 = \alpha_{A,1} \cdot \beta_{M,1}$: path-product association index, first sex category & $0.005715$ & $[0.001575$, $0.01067]$ \\
 & $\tau_1 = \beta_{A,1} + \theta_1$: decomposed total, first sex category & $0.01465$ & $[0.006203$, $0.02578]$ \\
Second sex category & $\alpha_{A,2}$: ALT $\to$ DBP slope within the second sex category & $0.1189$ & $[0.06108$, $0.1866]$ \\
 & $\beta_{M,2}$: DBP $\to$ PWV slope adjusted for ALT within the second sex category & $0.04405$ & $[0.03324$, $0.05521]$ \\
 & $\beta_{A,2}$: ALT $\to$ PWV slope adjusted for DBP within the second sex category & $0.003923$ & $[-0.004242$, $0.01221]$ \\
 & $\theta_2 = \alpha_{A,2} \cdot \beta_{M,2}$: path-product association index, second sex category & $0.005238$ & $[0.002629$, $0.008789]$ \\
 & $\tau_2 = \beta_{A,2} + \theta_2$: decomposed total, second sex category & $0.009161$ & $[0.0008104$, $0.01822]$ \\
Sex difference (second $-$ first category) & $\alpha_{A,2} - \alpha_{A,1}$: sex difference in the ALT $\to$ DBP slope & $0.02417$ & $[-0.07285$, $0.1208]$ \\
 & $\beta_{M,2} - \beta_{M,1}$: sex difference in the conditional DBP $\to$ PWV slope & $-0.01626$ & $[-0.03177$, $-0.001897]$ \\
 & $\beta_{A,2} - \beta_{A,1}$: sex difference in the conditional ALT $\to$ PWV slope & $-0.005016$ & $[-0.01739$, $0.006736]$ \\
 & $\theta_2 - \theta_1$: sex difference in the path product & $-0.0004763$ & $[-0.006175$, $0.005117]$ \\
 & $\tau_2 - \tau_1$: sex difference in the decomposed total & $-0.005492$ & $[-0.01874$, $0.007006]$ \\
\bottomrule
\end{longtable}}

\paragraph{20.\ Exposure $\times$ mediator interaction: ALT, TG and PWV}
\textit{Design.} ALT, TG and PWV measured in strict date order: TG 1--365 days after ALT, PWV 30--1095 days after TG (all bounds inclusive); one chain per ALT test date, with intervals from resampling of participants as clusters. Adjustment for approximate age in the year of the ALT test and recorded sex; the PWV equation includes an ALT $\times$ TG product term. PWV as the mean of available sides; all variables kept in recorded units.

\textit{Model.} $M = \alpha_0 + \alpha_A A + \alpha_{\mathrm{age}}\,\mathrm{age} + f(\mathrm{sex}) + \varepsilon_M$; $Y = \beta_0 + \beta_A A + \beta_M M + \beta_{AM} A M + \beta_{\mathrm{age}}\,\mathrm{age} + f(\mathrm{sex}) + \varepsilon_Y$. Both equations are linear regressions on the same sample; $A$, $M$ and $Y$ are ALT, TG and thigh-to-ankle PWV (recorded units). $\alpha_A$ is the ALT $\to$ TG slope; $\beta_A$ is the slope of PWV on ALT at a recorded TG value of 0; $\beta_M$ is the slope of PWV on TG at a recorded ALT value of 0; $\beta_{AM}$ is the interaction coefficient, namely the change in the slope of one variable per 1 unit higher value of the other.

\textit{Intervals.} Bootstrap 95\% percentile intervals.
{\footnotesize\setlength{\tabcolsep}{3pt}
\begin{longtable}{@{}>{\raggedright\arraybackslash}p{0.15\linewidth}>{\raggedright\arraybackslash}p{0.40\linewidth}>{\raggedleft\arraybackslash}p{0.14\linewidth}>{\raggedright\arraybackslash}p{0.24\linewidth}@{}}
\toprule
Block & Parameter & Estimate & Interval \\
\midrule\endhead
Mediator equation & $\alpha_A$: ALT $\to$ TG slope ($A \to M$) & $0.2205$ & $[0.03518$, $0.751]$ \\
PWV equation with interaction term & $\beta_A$: ALT $\to$ PWV slope at TG $= 0$ & $0.0008482$ & $[-0.005051$, $0.01181]$ \\
 & $\beta_M$: TG $\to$ PWV slope at ALT $= 0$ & $0.004791$ & $[0.002646$, $0.006739]$ \\
 & $\beta_{AM}$: ALT $\times$ TG interaction coefficient & $-2.405\times10^{-5}$ & $[-7.113\times10^{-5}$, $4.604\times10^{-5}]$ \\
\bottomrule
\end{longtable}}

\paragraph{21.\ Exposure $\times$ mediator interaction: ALT, seated DBP and PWV}
\textit{Design.} ALT, seated blood pressure and PWV measured in strict date order: blood pressure 7--365 days after ALT, PWV 30--1095 days after blood pressure (all bounds inclusive); one chain per ALT test date, with intervals from resampling of participants as clusters. Adjustment for approximate age in the year of the ALT test and recorded sex; the PWV equation includes an ALT $\times$ DBP product term. DBP as the mean of the available values of two seated readings, PWV as the mean of available sides; all variables kept in recorded units.

\textit{Model.} $M = \alpha_0 + \alpha_A A + \alpha_{\mathrm{age}}\,\mathrm{age} + f(\mathrm{sex}) + \varepsilon_M$; $Y = \beta_0 + \beta_A A + \beta_M M + \beta_{AM} A M + \beta_{\mathrm{age}}\,\mathrm{age} + f(\mathrm{sex}) + \varepsilon_Y$. Both equations are linear regressions on the same sample; $A$, $M$ and $Y$ are ALT, seated DBP and thigh-to-ankle PWV (recorded units). $\alpha_A$ is the ALT $\to$ DBP slope; $\beta_A$ is the slope of PWV on ALT at a recorded DBP value of 0; $\beta_M$ is the slope of PWV on DBP at a recorded ALT value of 0; $\beta_{AM}$ is the interaction coefficient, namely the change in the slope of one variable per 1 unit higher value of the other.

\textit{Intervals.} Bootstrap 95\% percentile intervals.
{\footnotesize\setlength{\tabcolsep}{3pt}
\begin{longtable}{@{}>{\raggedright\arraybackslash}p{0.15\linewidth}>{\raggedright\arraybackslash}p{0.40\linewidth}>{\raggedleft\arraybackslash}p{0.14\linewidth}>{\raggedright\arraybackslash}p{0.24\linewidth}@{}}
\toprule
Block & Parameter & Estimate & Interval \\
\midrule\endhead
Mediator equation & $\alpha_A$: ALT $\to$ DBP slope ($A \to M$) & $0.1092$ & $[0.06578$, $0.1575]$ \\
PWV equation with interaction term & $\beta_A$: ALT $\to$ PWV slope at DBP $= 0$ & $0.07935$ & $[0.03182$, $0.1351]$ \\
 & $\beta_M$: DBP $\to$ PWV slope at ALT $= 0$ & $0.07269$ & $[0.05837$, $0.08958]$ \\
 & $\beta_{AM}$: ALT $\times$ DBP interaction coefficient & $-0.0009153$ & $[-0.001608$, $-0.0003361]$ \\
\bottomrule
\end{longtable}}

\paragraph{22.\ ALT $\times$ sex interaction: sex-specific ALT--PWV slopes}
\textit{Design.} Pairing as in the main ALT--PWV association: PWV within 30 days before or after the ALT blood draw date (inclusive, in either order), with the nearest PWV taken for each ALT measurement date, $n = 127$. Sex enters both as a main effect and as a modifier, with additional adjustment for approximate age at the ALT measurement and the most recent BMI on or within 365 days before the earlier of the two measurement dates; continuous variables standardized within the analysis sample.

\textit{Model.} $z(Y) = \beta_0 + \sum_s \beta_{A,s} z(A) \mathbf{1}[\mathrm{sex}=s] + \beta_{\mathrm{age}} z(\mathrm{age}) + \beta_{\mathrm{BMI}} z(\mathrm{BMI}) + \gamma_{\mathrm{sex}} + \varepsilon$, a linear regression; $A$ is ALT, $Y$ is thigh-to-ankle PWV and $s$ is the sex category. $\beta_{A,s}$ is the difference in PWV in SD per 1 SD higher ALT within sex category $s$, with age and BMI held fixed.

\textit{Intervals.} 95\% confidence intervals; $P$ values not adjusted for multiple comparisons.
{\footnotesize\setlength{\tabcolsep}{3pt}
\begin{longtable}{@{}>{\raggedright\arraybackslash}p{0.14\linewidth}>{\raggedright\arraybackslash}p{0.29\linewidth}>{\raggedleft\arraybackslash}p{0.13\linewidth}>{\raggedright\arraybackslash}p{0.22\linewidth}>{\raggedleft\arraybackslash}p{0.14\linewidth}@{}}
\toprule
Block & Parameter & Estimate & Interval & $P$ \\
\midrule\endhead
First sex category & $\beta_{A,1}$: standardized ALT--PWV slope, first sex category & $0.2298$ & $[-0.1591$, $0.6187]$ & \mbox{$0.2445$} \\
Second sex category & $\beta_{A,2}$: standardized ALT--PWV slope, second sex category & $0.0574$ & $[-0.06325$, $0.1781]$ & \mbox{$0.3482$} \\
\bottomrule
\end{longtable}}

\subsubsection*{Sensitivity analyses}

\paragraph{23.\ Narrow time windows: ALT $\to$ TG $\to$ thigh-to-ankle PWV}
\textit{Design.} ALT, TG and PWV measured in strict date order: TG 1--180 days after ALT, PWV 90--730 days after TG (all bounds inclusive); one chain per ALT test date (a participant may contribute several chains), with intervals from resampling of participants as clusters. Adjustment for approximate age in the year of the ALT test and recorded sex. PWV as the mean of available sides; all variables kept in recorded units.

\textit{Model.} $M = \alpha_0 + \alpha_A A + \alpha_{\mathrm{age}}\,\mathrm{age} + f(\mathrm{sex}) + \varepsilon_M$; $Y = \beta_0 + \beta_A A + \beta_M M + \beta_{\mathrm{age}}\,\mathrm{age} + f(\mathrm{sex}) + \varepsilon_Y$. Both equations are linear regressions on the same sample; $A$, $M$ and $Y$ are ALT, TG and thigh-to-ankle PWV (recorded units), $\mathrm{age}$ is approximate age and $f(\mathrm{sex})$ is a sex fixed effect. $\alpha_A$ is the change in TG per 1 unit higher ALT; $\beta_M$ and $\beta_A$ are the mutually adjusted slopes of PWV on TG and ALT; path product $\theta = \alpha_A \cdot \beta_M$, decomposed total $\tau = \beta_A + \theta$.

\textit{Intervals.} Bootstrap 95\% percentile intervals.
{\footnotesize\setlength{\tabcolsep}{3pt}
\begin{longtable}{@{}>{\raggedright\arraybackslash}p{0.15\linewidth}>{\raggedright\arraybackslash}p{0.40\linewidth}>{\raggedleft\arraybackslash}p{0.14\linewidth}>{\raggedright\arraybackslash}p{0.24\linewidth}@{}}
\toprule
Block & Parameter & Estimate & Interval \\
\midrule\endhead
Path components & $\alpha_A$: ALT $\to$ TG slope ($A \to M$) & $0.08791$ & $[-0.04794$, $1.1]$ \\
 & $\beta_M$: TG $\to$ PWV slope adjusted for ALT ($M \to Y \mid A$) & $0.003458$ & $[0.001458$, $0.006596]$ \\
 & $\beta_A$: ALT $\to$ PWV slope adjusted for TG ($A \to Y \mid M$) & $-0.004011$ & $[-0.006055$, $0.001082]$ \\
Path indices & $\theta = \alpha_A \cdot \beta_M$: path-product association index & $0.000304$ & $[-0.000247$, $0.003227]$ \\
 & $\tau = \beta_A + \theta$: decomposed total & $-0.003707$ & $[-0.005103$, $0.003405]$ \\
\bottomrule
\end{longtable}}

\paragraph{24.\ Narrow time windows: ALT $\to$ seated DBP $\to$ thigh-to-ankle PWV}
\textit{Design.} ALT, seated blood pressure and PWV measured in strict date order: blood pressure 7--180 days after ALT, PWV 90--730 days after blood pressure (all bounds inclusive); one chain per ALT test date (a participant may contribute several chains), with intervals from resampling of participants as clusters. Adjustment for approximate age in the year of the ALT test and recorded sex. DBP as the mean of the available values of two seated readings, PWV as the mean of available sides; all variables kept in recorded units.

\textit{Model.} $M = \alpha_0 + \alpha_A A + \alpha_{\mathrm{age}}\,\mathrm{age} + f(\mathrm{sex}) + \varepsilon_M$; $Y = \beta_0 + \beta_A A + \beta_M M + \beta_{\mathrm{age}}\,\mathrm{age} + f(\mathrm{sex}) + \varepsilon_Y$. Both equations are linear regressions on the same sample; $A$, $M$ and $Y$ are ALT, seated DBP and thigh-to-ankle PWV (recorded units), $\mathrm{age}$ is approximate age and $f(\mathrm{sex})$ is a sex fixed effect. $\alpha_A$ is the change in DBP per 1 unit higher ALT; $\beta_M$ and $\beta_A$ are the mutually adjusted slopes of PWV on DBP and ALT; path product $\theta = \alpha_A \cdot \beta_M$, decomposed total $\tau = \beta_A + \theta$.

\textit{Intervals.} Bootstrap 95\% percentile intervals.
{\footnotesize\setlength{\tabcolsep}{3pt}
\begin{longtable}{@{}>{\raggedright\arraybackslash}p{0.15\linewidth}>{\raggedright\arraybackslash}p{0.40\linewidth}>{\raggedleft\arraybackslash}p{0.14\linewidth}>{\raggedright\arraybackslash}p{0.24\linewidth}@{}}
\toprule
Block & Parameter & Estimate & Interval \\
\midrule\endhead
Path components & $\alpha_A$: ALT $\to$ DBP slope ($A \to M$) & $0.1482$ & $[0.06225$, $0.2542]$ \\
 & $\beta_M$: DBP $\to$ PWV slope adjusted for ALT ($M \to Y \mid A$) & $0.06156$ & $[0.04416$, $0.07816]$ \\
 & $\beta_A$: ALT $\to$ PWV slope adjusted for DBP ($A \to Y \mid M$) & $0.00742$ & $[-0.01213$, $0.0283]$ \\
Path indices & $\theta = \alpha_A \cdot \beta_M$: path-product association index & $0.009123$ & $[0.003557$, $0.01643]$ \\
 & $\tau = \beta_A + \theta$: decomposed total & $0.01654$ & $[-0.002814$, $0.03767]$ \\
\bottomrule
\end{longtable}}

\paragraph{25.\ Extended confounder adjustment: ALT $\to$ TG $\to$ thigh-to-ankle PWV}
\textit{Design.} ALT, TG and PWV measured in strict date order: TG 1--365 days after ALT, PWV 30--1095 days after TG (all bounds inclusive); one chain per ALT test date, with intervals from resampling of participants as clusters. Adjustment for approximate age, recorded sex, and the most recent BMI, smoking status and drinking frequency on or within 730 days before the ALT test date (the latter two as categorical terms by their original coding). All variables kept in recorded units.

\textit{Model.} $M = \alpha_0 + \alpha_A A + \alpha_C^{\top} C + \varepsilon_M$; $Y = \beta_0 + \beta_A A + \beta_M M + \beta_C^{\top} C + \varepsilon_Y$. Both equations are linear regressions on the same sample; $A$, $M$ and $Y$ are ALT, TG and thigh-to-ankle PWV (recorded units), and $C$ comprises approximate age, sex, BMI, smoking status and drinking frequency. $\alpha_A$ is the change in TG per 1 unit higher ALT; $\beta_M$ and $\beta_A$ are the mutually adjusted slopes of PWV on TG and ALT; path product $\theta = \alpha_A \cdot \beta_M$, decomposed total $\tau = \beta_A + \theta$.

\textit{Intervals.} Bootstrap 95\% percentile intervals.
{\footnotesize\setlength{\tabcolsep}{3pt}
\begin{longtable}{@{}>{\raggedright\arraybackslash}p{0.15\linewidth}>{\raggedright\arraybackslash}p{0.40\linewidth}>{\raggedleft\arraybackslash}p{0.14\linewidth}>{\raggedright\arraybackslash}p{0.24\linewidth}@{}}
\toprule
Block & Parameter & Estimate & Interval \\
\midrule\endhead
Path components & $\alpha_A$: ALT $\to$ TG slope ($A \to M$) & $0.6551$ & $[0.09791$, $1.56]$ \\
 & $\beta_A$: ALT $\to$ PWV slope adjusted for TG ($A \to Y \mid M$) & $0.00242$ & $[-0.01098$, $0.0155]$ \\
 & $\beta_M$: TG $\to$ PWV slope adjusted for ALT ($M \to Y \mid A$) & $0.004013$ & $[0.001439$, $0.007874]$ \\
Path indices & $\tau = \beta_A + \theta$: decomposed total & $0.005049$ & $[-0.007573$, $0.01796]$ \\
 & $\theta = \alpha_A \cdot \beta_M$: path-product association index & $0.002629$ & $[0.0002741$, $0.007509]$ \\
\bottomrule
\end{longtable}}

\paragraph{26.\ Adjustment for prior PWV: ALT $\to$ TG $\to$ thigh-to-ankle PWV}
\textit{Design.} ALT, TG and PWV measured in strict date order: TG 1--365 days after ALT, PWV 30--1095 days after TG (all bounds inclusive); one chain per ALT test date, with intervals from resampling of participants as clusters. Both equations adjust for approximate age and recorded sex; the PWV equation additionally adjusts for the most recent PWV 1--730 days before the ALT test date. PWV as the mean of available sides; all variables kept in recorded units.

\textit{Model.} $M = \alpha_0 + \alpha_A A + \alpha_{\mathrm{age}}\,\mathrm{age} + f(\mathrm{sex}) + \varepsilon_M$; $Y = \beta_0 + \beta_A A + \beta_M M + \beta_{\mathrm{age}}\,\mathrm{age} + f(\mathrm{sex}) + \beta_{\mathrm{prior}} Y_{\mathrm{prior}} + \varepsilon_Y$. Both equations are linear regressions on the same sample; $A$, $M$ and $Y$ are ALT, TG and thigh-to-ankle PWV (recorded units), and $Y_{\mathrm{prior}}$ is the most recent PWV before the ALT test. $\alpha_A$ is the change in TG per 1 unit higher ALT; $\beta_M$ and $\beta_A$ are the slopes of PWV on TG and ALT, mutually adjusted and conditional on prior PWV; path product $\theta = \alpha_A \cdot \beta_M$, decomposed total $\tau = \beta_A + \theta$.

\textit{Intervals.} Bootstrap 95\% percentile intervals.
{\footnotesize\setlength{\tabcolsep}{3pt}
\begin{longtable}{@{}>{\raggedright\arraybackslash}p{0.15\linewidth}>{\raggedright\arraybackslash}p{0.40\linewidth}>{\raggedleft\arraybackslash}p{0.14\linewidth}>{\raggedright\arraybackslash}p{0.24\linewidth}@{}}
\toprule
Block & Parameter & Estimate & Interval \\
\midrule\endhead
Path components & $\alpha_A$: ALT $\to$ TG slope ($A \to M$) & $0.8407$ & $[0.2433$, $1.822]$ \\
 & $\beta_M$: TG $\to$ PWV slope adjusted for ALT ($M \to Y \mid A$) & $0.004358$ & $[0.0007089$, $0.009099]$ \\
 & $\beta_A$: ALT $\to$ PWV slope adjusted for TG ($A \to Y \mid M$) & $-0.01369$ & $[-0.03081$, $0.003351]$ \\
Path indices & $\theta = \alpha_A \cdot \beta_M$: path-product association index & $0.003664$ & $[0.0003396$, $0.01125]$ \\
 & $\tau = \beta_A + \theta$: decomposed total & $-0.01003$ & $[-0.02526$, $0.007694]$ \\
\bottomrule
\end{longtable}}

\paragraph{27.\ Within-participant fixed effects: ALT $\to$ TG $\to$ PWV}
\textit{Design.} ALT, TG and PWV measured in strict date order: TG 1--365 days after ALT, PWV 30--1095 days after TG (all bounds inclusive); one chain per ALT test date (a participant may contribute several chains), $n = 1711$, with intervals from resampling of participants as clusters. Both equations contain only participant fixed effects and the path variables; the slopes come from differences between chains of the same participant. PWV as the mean of available sides; all variables kept in recorded units.

\textit{Model.} $M_{ij} = u_i + \alpha_A A_{ij} + \varepsilon_M$; $Y_{ij} = v_i + \beta_A A_{ij} + \beta_M M_{ij} + \varepsilon_Y$. Both equations are linear regressions with participant fixed effects ($u_i$, $v_i$) on the same sample; $i$ indexes participants and $j$ their chains; $A$, $M$ and $Y$ are ALT, TG and thigh-to-ankle PWV (recorded units). $\alpha_A$ is the within-participant change in TG per 1 unit higher ALT; $\beta_M$ and $\beta_A$ are the within-participant, mutually adjusted slopes of PWV on TG and ALT; path product $\theta = \alpha_A \cdot \beta_M$, decomposed total $\tau = \beta_A + \theta$.

\textit{Intervals.} Bootstrap 95\% percentile intervals.
{\footnotesize\setlength{\tabcolsep}{3pt}
\begin{longtable}{@{}>{\raggedright\arraybackslash}p{0.15\linewidth}>{\raggedright\arraybackslash}p{0.40\linewidth}>{\raggedleft\arraybackslash}p{0.14\linewidth}>{\raggedright\arraybackslash}p{0.24\linewidth}@{}}
\toprule
Block & Parameter & Estimate & Interval \\
\midrule\endhead
Within-participant path components & $\alpha_A$: ALT $\to$ TG slope ($A \to M$) & $0.05109$ & $[-0.02325$, $0.6038]$ \\
 & $\beta_M$: TG $\to$ PWV slope adjusted for ALT ($M \to Y \mid A$) & $0.0002791$ & $[-0.0001585$, $0.0008513]$ \\
 & $\beta_A$: ALT $\to$ PWV slope adjusted for TG ($A \to Y \mid M$) & $-4.53\times10^{-5}$ & $[-0.0009806$, $0.0005664]$ \\
Within-participant path indices & $\theta = \alpha_A \cdot \beta_M$: path-product association index & $1.426\times10^{-5}$ & $[-1.919\times10^{-5}$, $0.0001908]$ \\
 & $\tau = \beta_A + \theta$: decomposed total & $-3.104\times10^{-5}$ & $[-0.0009029$, $0.0006154]$ \\
\bottomrule
\end{longtable}}

\paragraph{28.\ Within-participant fixed effects: ALT $\to$ seated DBP $\to$ PWV}
\textit{Design.} ALT, seated blood pressure and PWV measured in strict date order: blood pressure 7--365 days after ALT, PWV 30--1095 days after blood pressure (all bounds inclusive); one chain per ALT test date (a participant may contribute several chains), with intervals from resampling of participants as clusters. Both equations contain only participant fixed effects and the path variables; the slopes come from differences between chains of the same participant. DBP as the mean of the available values of two seated readings, PWV as the mean of available sides; all variables kept in recorded units.

\textit{Model.} $M_{ij} = u_i + \alpha_A A_{ij} + \varepsilon_M$; $Y_{ij} = v_i + \beta_A A_{ij} + \beta_M M_{ij} + \varepsilon_Y$. Both equations are linear regressions with participant fixed effects ($u_i$, $v_i$) on the same sample; $i$ indexes participants and $j$ their chains; $A$, $M$ and $Y$ are ALT, seated DBP and thigh-to-ankle PWV (recorded units). $\alpha_A$ is the within-participant change in DBP per 1 unit higher ALT; $\beta_A$ is the within-participant slope of PWV on ALT adjusted for DBP.

\textit{Intervals.} Bootstrap 95\% percentile intervals.
{\footnotesize\setlength{\tabcolsep}{3pt}
\begin{longtable}{@{}>{\raggedright\arraybackslash}p{0.15\linewidth}>{\raggedright\arraybackslash}p{0.40\linewidth}>{\raggedleft\arraybackslash}p{0.14\linewidth}>{\raggedright\arraybackslash}p{0.24\linewidth}@{}}
\toprule
Block & Parameter & Estimate & Interval \\
\midrule\endhead
Within-participant path components & $\alpha_A$: ALT $\to$ DBP slope ($A \to M$) & $0.001252$ & $[-0.0003606$, $0.006746]$ \\
 & $\beta_A$: ALT $\to$ PWV slope adjusted for DBP ($A \to Y \mid M$) & $-0.0001135$ & $[-0.0004746$, $2.295\times10^{-19}]$ \\
\bottomrule
\end{longtable}}

\paragraph{29.\ Serial path with symmetric windows: ALT $\to$ TG $\to$ DBP $\to$ PWV}
\textit{Design.} ALT and TG from the same blood test; seated blood pressure as the nearest measurement within 180 days before or after the test date, PWV as the nearest measurement within 365 days before or after that blood-pressure date (all bounds inclusive, in either order, same day allowed). One chain per test date (a participant may contribute several chains). Adjustment for approximate age on the earliest measurement date of the chain, recorded sex, and the most recent BMI on or within 365 days before that date. DBP as the mean of seated readings, PWV as the mean of available sides; all variables on original recorded scales.

\textit{Model.} $M_1 = \alpha_1 + \beta_{AM_1} A + \gamma_1^{\top} C + \varepsilon_1$; $M_2 = \alpha_2 + \delta_2 A + \beta_{M_1M_2} M_1 + \gamma_2^{\top} C + \varepsilon_2$; $Y = \alpha_3 + \delta_3 A + \delta_4 M_1 + \beta_{M_2Y} M_2 + \gamma_3^{\top} C + \varepsilon_3$. All three equations are linear regressions on the same sample; $A$, $M_1$, $M_2$ and $Y$ are alanine aminotransferase (ALT), TG, seated DBP and thigh-to-ankle PWV (original recorded scales), and $C$ comprises approximate age, recorded sex and BMI. $\beta_{AM_1}$, $\beta_{M_1M_2}$ and $\beta_{M_2Y}$ are the differences in the recorded value of the equation's outcome per 1 recorded unit higher value of that regressor, with the other terms of the same equation held fixed; serial path product $\theta = \beta_{AM_1} \cdot \beta_{M_1M_2} \cdot \beta_{M_2Y}$.

\textit{Intervals.} Bootstrap 95\% percentile intervals. $P$ values not adjusted for multiple comparisons.
{\footnotesize\setlength{\tabcolsep}{3pt}
\begin{longtable}{@{}>{\raggedright\arraybackslash}p{0.14\linewidth}>{\raggedright\arraybackslash}p{0.29\linewidth}>{\raggedleft\arraybackslash}p{0.13\linewidth}>{\raggedright\arraybackslash}p{0.22\linewidth}>{\raggedleft\arraybackslash}p{0.14\linewidth}@{}}
\toprule
Block & Parameter & Estimate & Interval & $P$ \\
\midrule\endhead
Path indices (symmetric windows) & $\theta = \beta_{AM_1} \cdot \beta_{M_1M_2} \cdot \beta_{M_2Y}$: serial path product & $0.0002614$ & $[-0.0003314$, $0.0009597]$ & \mbox{$0.4$} \\
Path components (symmetric windows) & $\beta_{AM_1}$: adjusted ALT $\to$ TG slope ($A \to M_1$) & $0.3386$ & $[-0.09444$, $0.6628]$ & \mbox{$0.107$} \\
 & $\beta_{M_1M_2}$: TG $\to$ seated DBP slope adjusted for ALT ($M_1 \to M_2 \mid A$) & $0.009174$ & $[-0.01022$, $0.02731]$ & \mbox{$0.333$} \\
 & $\beta_{M_2Y}$: seated DBP $\to$ PWV slope adjusted for ALT and TG ($M_2 \to Y \mid A, M_1$) & $0.08414$ & $[0.06345$, $0.1019]$ & \mbox{$0$} \\
\bottomrule
\end{longtable}}

\paragraph{30.\ Joint model: ALT, TG, DBP and PWV ($\pm 90$ days)}
\textit{Design.} ALT and TG as same-day blood test values; seated DBP as the nearest measurement within 90 days before or after the blood draw date, PWV as the measurement nearest to the DBP date and within 90 days of both the DBP date and the blood draw date (inclusive, in either order); one measurement chain per blood draw date, $n = 167$. Adjustment for approximate age at the blood draw, sex, and the most recent BMI on or within 365 days before the earliest date of the chain. Continuous variables standardized within the analysis sample, sex as a categorical term.

\textit{Model.} $z(Y) = \beta_0 + \beta_A z(A) + \beta_{M_1} z(M_1) + \beta_{M_2} z(M_2) + \beta_{\mathrm{age}} z(\mathrm{age}) + \beta_{\mathrm{BMI}} z(\mathrm{BMI}) + \gamma_{\mathrm{sex}} + \varepsilon$, a linear regression, with notation as in the $\pm 30$-day joint model. $\beta_A$, $\beta_{M_1}$ and $\beta_{M_2}$ are the differences in PWV in SD per 1 SD higher value of that variable, with the other terms of the model held fixed.

\textit{Intervals.} 95\% confidence intervals; $P$ values not adjusted for multiple comparisons.
{\footnotesize\setlength{\tabcolsep}{3pt}
\begin{longtable}{@{}>{\raggedright\arraybackslash}p{0.14\linewidth}>{\raggedright\arraybackslash}p{0.29\linewidth}>{\raggedleft\arraybackslash}p{0.13\linewidth}>{\raggedright\arraybackslash}p{0.22\linewidth}>{\raggedleft\arraybackslash}p{0.14\linewidth}@{}}
\toprule
Block & Parameter & Estimate & Interval & $P$ \\
\midrule\endhead
Joint model ($\pm 90$ days) & $\beta_A$: ALT coefficient (adjusted for TG and DBP) & $0.1033$ & $[0.02722$, $0.1794]$ & \mbox{$0.008098$} \\
 & $\beta_{M_1}$: TG coefficient (adjusted for ALT and DBP) & $0.01386$ & $[-0.1007$, $0.1284]$ & \mbox{$0.8115$} \\
 & $\beta_{M_2}$: seated DBP coefficient (adjusted for ALT and TG) & $0.5217$ & $[0.4046$, $0.6387]$ & \mbox{$1.843\times10^{-15}$} \\
\bottomrule
\end{longtable}}

\paragraph{31.\ Joint model with additional adjustment for smoking status ($\pm 30$ days)}
\textit{Design.} Measurement chains as in the $\pm 30$-day joint model: ALT and TG from the same-day blood test, DBP within 30 days before or after the blood draw date, PWV within 30 days of both the blood draw date and the DBP date (inclusive, in either order), $n = 90$. In addition to age, sex and BMI, adjustment for the most recently recorded current smoking status on or within 365 days before the earliest date of the chain. Continuous variables standardized within the analysis sample, sex and smoking status as categorical terms.

\textit{Model.} $z(Y) = \beta_0 + \beta_A z(A) + \beta_{M_1} z(M_1) + \beta_{M_2} z(M_2) + \beta_{\mathrm{age}} z(\mathrm{age}) + \beta_{\mathrm{BMI}} z(\mathrm{BMI}) + \gamma_{\mathrm{sex}} + \gamma_{\mathrm{smk}} + \varepsilon$, a linear regression; $\gamma_{\mathrm{smk}}$ is the categorical smoking-status term, with other notation as in the $\pm 30$-day joint model. $\beta_A$, $\beta_{M_1}$ and $\beta_{M_2}$ are the differences in PWV in SD per 1 SD higher value of that variable, with the other terms of the model held fixed.

\textit{Intervals.} 95\% confidence intervals; $P$ values not adjusted for multiple comparisons.
{\footnotesize\setlength{\tabcolsep}{3pt}
\begin{longtable}{@{}>{\raggedright\arraybackslash}p{0.14\linewidth}>{\raggedright\arraybackslash}p{0.29\linewidth}>{\raggedleft\arraybackslash}p{0.13\linewidth}>{\raggedright\arraybackslash}p{0.22\linewidth}>{\raggedleft\arraybackslash}p{0.14\linewidth}@{}}
\toprule
Block & Parameter & Estimate & Interval & $P$ \\
\midrule\endhead
Joint model + smoking status ($\pm 30$ days) & $\beta_A$: ALT coefficient (adjusted for TG and DBP) & $0.1266$ & $[0.03439$, $0.2188]$ & \mbox{$0.007671$} \\
 & $\beta_{M_1}$: TG coefficient (adjusted for ALT and DBP) & $-0.005891$ & $[-0.1468$, $0.135]$ & \mbox{$0.934$} \\
 & $\beta_{M_2}$: seated DBP coefficient (adjusted for ALT and TG) & $0.4444$ & $[0.2873$, $0.6015]$ & \mbox{$2.156\times10^{-7}$} \\
\bottomrule
\end{longtable}}

\FloatBarrier

\fussy

\section{Interpretable Foundation Model}
\label{app:disease-foundation-construction}

\subsection{The equation system of the network}
\label{app:disease-foundation:equations}

\paragraph{Source contexts and applicability.}
A source context is the primary analysis of a pathway
(Table~\ref{tab:findings:pathway-codes}) or one of the extended specifications
listed in Table~\ref{tab:foundation:source-sequences}. The applicability
indicator $a_r$ of Equation~\ref{eq:foundation:task-head}
takes the value 1 when every input in $S_r$ is observed and each categorical input,
including the sex stratum of the extended L2 equations, takes a value present
in the estimation sample of that equation, and 0 otherwise; $f_{tj}$ is zero
when $x_j$ is unobserved, and $d_{tj}$ is the fitted offset for that missing
input. The source equations are estimated at the prediction time of
Appendix~\ref{app:disease-foundation:prediction}, with inputs and targets
taken as the latest measurements at or before it and with the equation forms
and covariates of the source analyses; $\beta$ is estimated on the HPP
training split and frozen. The task terms are fitted on the training split,
and the tuning split serves model selection.

\begin{table}[tbp]
  \centering
  \caption{\textbf{Source equations of the 15 network pathways.} In the
  covariate column, entries after ``extended'' are the covariates or terms added
  by the extended contexts; ---, no declared covariate beyond age and sex.}
  \label{tab:foundation:source-sequences}
  \footnotesize
  \setlength{\tabcolsep}{3pt}
  \begin{tabular*}{\linewidth}{@{\extracolsep{\fill}}llp{0.46\linewidth}r@{}}
    \toprule
    Pathway & Node chain & Declared covariates and terms & Equations \\
    \midrule
    H1 & WHR $\to$ ALT $\to$ DBP $\to$ PWV & ---; extended: smoking, alcohol, moderate activity & 6 \\
    H2 & Hip $\to$ ALT $\to$ DBP $\to$ PWV & study; age$^2$ & 3 \\
    H3 & WC $\to$ ALT $\to$ SBP $\to$ cIMT & cohort & 3 \\
    H4 & WHR $\to$ ATT $\to$ SBP $\to$ cIMT & ---; extended: BMI & 6 \\
    H5 & Neck $\to$ ATT $\to$ SBP $\to$ PWV & age$^2$ & 3 \\
    L1 & AF $\to$ TG $\to$ DBP $\to$ PWV & study & 3 \\
    L2 & VAT $\to$ ApoB $\to$ SBP $\to$ PWV & age$^2$; extended: sex-stratified & 9 \\
    L3 & Weight $\to$ ApoB $\to$ DBP $\to$ PWV & height, study; extended: node$\times$age & 6 \\
    L4 & Hip $\to$ ApoB $\to$ DBP $\to$ PWV & height, cohort, smoking; age$^2$ & 3 \\
    L5 & WHR $\to$ TG $\to$ SBP $\to$ cIMT & cohort; age$^2$ & 3 \\
    L6 & GGT $\to$ TG $\to$ SBP $\to$ PWV & --- & 3 \\
    L7 & ALT $\to$ TG $\to$ PWV & ---; extended: prior BMI, prior smoking, prior alcohol & 4 \\
    O1 & WC $\to$ SpO$_2$ $\to$ SBP $\to$ cIMT & cohort & 3 \\
    O2 & SpO$_2$ $\to$ GlycA $\to$ PWV & prior BMI; extended 1: SBP; extended 2: AHI & 6 \\
    O3 & BMI $\to$ SpO$_2$ $\to$ cIMT & cohort, smoking, education; BMI$\times$SpO$_2$ & 2 \\
    \bottomrule
  \end{tabular*}
\end{table}

\subsection{HPP multi-phenotype prediction}
\label{app:disease-foundation:prediction}

\paragraph{Tasks, splits and metrics.}
The prediction time follows the first anthropometric assessment, and the inputs
are measurements taken before it. Future PWV and cIMT are the first bilateral
means measured within the follow-up window after the prediction time, with a
median follow-up of about 1.9 years; the labels of the queried tasks are the latest
ATT, ApoB, GlycA and SpO$_2$ measurements before the prediction time. Baseline
PWV and cIMT enter no task input. The 10,401 participants are split by participant
into training (5,759), tuning (1,548), calibration (1,012) and test (2,082) sets; the
calibration set holds 71 future-PWV and 79 future-cIMT labels. The 12,492 HPP
predictions are the six task predictions of each of the 2,082 test participants.
The coefficient of determination is
$R^2=1-\sum_i(y_i-\hat y_i)^2/\sum_i(y_i-\bar y)^2$, with the sums and
$\bar y$ taken over the test labels. The 90\% prediction intervals are
split-conformal intervals \citep{lei2018distribution} constructed on the calibration set. The intervals in
Table~\ref{tab:foundation:hpp-errors} are 95\% percentile intervals over
1,000 bootstrap resamples of test participants, conditional on the fitted model.

\begin{table}[tbp]
  \centering
  \caption{\textbf{Errors of the six HPP tasks.} Label counts are the training
  and tuning labels; test counts and NRMSE point estimates are in
  Table~\ref{tab:foundation:real-performance}. The mean absolute error (MAE),
  root-mean-square error (RMSE), mean error (prediction minus observation) and
  the mean width of the 90\% interval are in original units: PWV m/s; cIMT mm; ATT dB/cm/MHz; ApoB g/L;
  GlycA mmol/L; SpO$_2$ percentage points.}
  \label{tab:foundation:hpp-errors}
  \footnotesize
  \setlength{\tabcolsep}{2pt}
  \begin{tabular*}{\linewidth}{@{\extracolsep{\fill}}lrrcrcrr@{}}
    \toprule
    Task & Labels & MAE & MAE 95\% CI & RMSE & NRMSE 95\% CI & Mean error & Width \\
    \midrule
    Future PWV & 494 / 131 & $0.993$ & [$0.838$, $1.17$] & $1.45$ & [0.728, 1.209] & $-0.140$ & $3.86$ \\
    Future cIMT & 518 / 145 & $0.0729$ & [$0.0652$, $0.0812$] & $0.0938$ & [0.826, 1.033] & $-0.0115$ & $0.248$ \\
    Queried ATT & 3{,}680 / 984 & $0.0738$ & [$0.0707$, $0.0769$] & $0.0953$ & [0.871, 0.953] & $-0.00451$ & $0.320$ \\
    Queried ApoB & 754 / 249 & $0.117$ & [$0.106$, $0.129$] & $0.155$ & [0.762, 0.967] & $-0.00752$ & $0.569$ \\
    Queried GlycA & 754 / 249 & $0.0832$ & [$0.0762$, $0.0910$] & $0.107$ & [0.710, 0.853] & $-0.00893$ & $0.307$ \\
    Queried SpO$_2$ & 3{,}193 / 876 & $0.774$ & [$0.741$, $0.810$] & $0.977$ & [0.753, 0.832] & $-0.0113$ & $3.17$ \\
    \bottomrule
  \end{tabular*}
\end{table}

\paragraph{Additional random masking.}
Beyond the natural missingness, each observed network node of each test
participant is hidden independently with probability $p$ (10\%, 30\%, 50\% or
70\%), together with the derived inputs that depend on it; age, sex and the
other covariates are not masked. The fitted predictor, the interval radius set
on the calibration set and the test labels stay fixed, and each probability uses
five fixed seeds (Table~\ref{tab:foundation:random-masking}).

\begin{table}[tbp]
  \centering
  \caption{\textbf{Future PWV and cIMT prediction under additional random
  masking.} Entries are five-seed means, with the minimum and maximum over the
  five seeds in parentheses.}
  \label{tab:foundation:random-masking}
  \footnotesize
  \setlength{\tabcolsep}{3pt}
  \begin{tabular*}{\linewidth}{@{\extracolsep{\fill}}lrccc@{}}
    \toprule
    Task & Masking & $R^2$ & NRMSE & Coverage (\%) \\
    \midrule
    Future PWV & 10\% & 0.353 (0.329--0.375) & 0.977 (0.960--0.995) & 89.0 (88.0--90.4) \\
     & 30\% & 0.321 (0.282--0.351) & 1.002 (0.979--1.030) & 87.6 (85.5--89.2) \\
     & 50\% & 0.290 (0.256--0.315) & 1.024 (1.006--1.048) & 86.0 (84.9--87.3) \\
     & 70\% & 0.248 (0.202--0.294) & 1.054 (1.021--1.086) & 84.3 (82.5--85.5) \\
    \addlinespace
    Future cIMT & 10\% & 0.165 (0.137--0.179) & 0.933 (0.925--0.948) & 83.0 (82.6--83.1) \\
     & 30\% & 0.149 (0.125--0.167) & 0.942 (0.932--0.955) & 83.1 (81.4--84.3) \\
     & 50\% & 0.143 (0.121--0.159) & 0.945 (0.936--0.957) & 82.8 (82.6--83.1) \\
     & 70\% & 0.136 (0.127--0.152) & 0.949 (0.940--0.954) & 82.9 (82.0--83.7) \\
    \bottomrule
  \end{tabular*}
\end{table}

\paragraph{Structural missingness.}
The same predictor is also evaluated under three kinds of structural
missingness: one network node hidden at a time; all nodes of one or two
measurement modalities hidden (NMR: ApoB and GlycA; DXA: AF and VAT; sleep
monitoring: SpO$_2$; liver ultrasound: ATT); and the network nodes restricted
to the nine measured in NHANES (WC, weight, BMI, ALT, GGT, TG, ApoB, SBP and
DBP), with the covariates unchanged
(Table~\ref{tab:foundation:structural-missingness}).

\begin{table}[tbp]
  \centering
  \caption{\textbf{Future PWV and cIMT prediction under structural
  missingness.} Results without additional missingness are in
  Table~\ref{tab:foundation:real-performance}; hiding sleep monitoring or liver
  ultrasound is identical to hiding SpO$_2$ or ATT, respectively; Cov., coverage
  of the 90\% prediction interval.}
  \label{tab:foundation:structural-missingness}
  \footnotesize
  \setlength{\tabcolsep}{3pt}
  \begin{tabular*}{\linewidth}{@{\extracolsep{\fill}}lrrrrrr@{}}
    \toprule
    & \multicolumn{3}{c}{Future PWV} & \multicolumn{3}{c}{Future cIMT} \\
    \cmidrule(lr){2-4}\cmidrule(l){5-7}
    Hidden & $R^2$ & NRMSE & Cov.\ (\%) & $R^2$ & NRMSE & Cov.\ (\%) \\
    \midrule
    WHR & 0.377 & 0.959 & 89.2 & 0.184 & 0.922 & 83.1 \\
    Hip & 0.369 & 0.966 & 89.2 & 0.178 & 0.925 & 82.6 \\
    WC & 0.371 & 0.964 & 89.2 & 0.177 & 0.926 & 83.1 \\
    Neck & 0.378 & 0.959 & 89.2 & 0.176 & 0.926 & 83.1 \\
    ALT & 0.389 & 0.950 & 89.8 & 0.180 & 0.924 & 84.3 \\
    ATT & 0.384 & 0.953 & 89.8 & 0.182 & 0.923 & 82.6 \\
    DBP & 0.331 & 0.994 & 88.6 & 0.174 & 0.928 & 84.3 \\
    SBP & 0.305 & 1.013 & 85.5 & 0.148 & 0.942 & 83.1 \\
    AF & 0.377 & 0.959 & 89.2 & 0.171 & 0.929 & 83.1 \\
    VAT & 0.381 & 0.956 & 89.8 & 0.182 & 0.923 & 83.1 \\
    Weight & 0.385 & 0.953 & 89.2 & 0.160 & 0.936 & 83.7 \\
    BMI & 0.361 & 0.971 & 88.6 & 0.172 & 0.929 & 83.1 \\
    TG & 0.377 & 0.959 & 90.4 & 0.153 & 0.939 & 83.1 \\
    ApoB & 0.384 & 0.954 & 88.6 & 0.168 & 0.931 & 83.7 \\
    GGT & 0.379 & 0.957 & 89.2 & 0.172 & 0.929 & 83.1 \\
    SpO$_2$ & 0.373 & 0.962 & 89.2 & 0.164 & 0.934 & 83.7 \\
    GlycA & 0.378 & 0.958 & 89.2 & 0.179 & 0.925 & 82.6 \\
    \addlinespace
    NMR & 0.383 & 0.954 & 88.6 & 0.178 & 0.925 & 83.7 \\
    DXA & 0.379 & 0.958 & 89.2 & 0.179 & 0.925 & 82.6 \\
    NMR and DXA & 0.388 & 0.951 & 90.4 & 0.179 & 0.925 & 82.6 \\
    Nine NHANES nodes only & 0.363 & 0.970 & 88.6 & 0.184 & 0.922 & 82.6 \\
    \bottomrule
  \end{tabular*}
\end{table}

\FloatBarrier

\subsection{The BMI--SpO\texorpdfstring{$_2$}{2}--cIMT decomposition and other effect queries}
\label{app:disease-foundation:effects}

\paragraph{Sample and contrast.}
The sample of the BMI--SpO$_2$--cIMT decomposition comprises the participants
with BMI, SpO$_2$ and cIMT measured in temporal order and a complete $C_e$; all
belong to one cohort, so the cohort term is constant. The contrast levels $a_0$
and $a_1$ are the 25th and 75th percentiles of BMI among the training-split
participants of Appendix~\ref{app:disease-foundation:prediction} with this
temporal sequence; they do not depend on outcome values and stay fixed across
resamples. The controlled direct contrast is
$\mathrm{CDE}(m^\ast)=E_{C_e}[\mu_Y(a_1,m^\ast,C_e)-\mu_Y(a_0,m^\ast,C_e)]
=\Delta a(\theta_A+\theta_{AM}m^\ast)$, where $m^\ast$ is the
analysis-sample median of SpO$_2$; the conditional BMI slope is
$\mathrm{CDE}(m^\ast)/\Delta a$, with the interval endpoints divided likewise.
With an exposure--mediator interaction, $T-\mathrm{CDE}(m^\ast)$ need not
equal $I$. The alternative decomposition switches the mediator distribution at
$a_0$: $D'=g(a_1,a_1)-g(a_0,a_1)$ and $I'=g(a_0,a_1)-g(a_0,a_0)$, with
$T=D'+I'$. The mediator shift is
$E_{C_e}[\mu_M(a_1,C_e)-\mu_M(a_0,C_e)]=\alpha_A\Delta a$. The 95\% CIs are
pointwise percentile intervals over 1,000 participant-level bootstrap resamples,
with both equations refitted in each resample and no multiplicity adjustment;
paired differences use the same sample and the same resamples.
Table~\ref{tab:foundation:o3-coefficients} lists every coefficient of the two
equations.

\paragraph{Identification conditions.}
The functional $g(a,a')$ of Equation~\ref{eq:foundation:o3-pathway-estimand}
is an interventional pathway effect \citep{vansteelandt2017interventional}: the
exposure is set to $a$ and the mediator is drawn from its conditional
distribution under exposure $a'$. The causal reading of $D$, $I$ and $T$
requires consistency; positivity, under which, given $C_e$, the exposure has
positive density at $a_0$ and $a_1$ and the mediator ranges under the two
exposure levels cover each other, with support for $m^\ast$ at both levels for the
controlled direct contrast; no unmeasured exposure--outcome or
exposure--mediator confounding given $C_e$, and no unmeasured mediator--outcome
confounding given the exposure and $C_e$; measurement of $C_e$ at or before
the exposure; no differential measurement error in the exposure, mediator,
outcome and $C_e$; entry into the analysis sample independent of the potential
outcomes given $C_e$; and correct specification of the two conditional-mean
models.
Because $\mu_Y$ is linear in $M$, the integral over $F_{M\mid A=a',C_e}$
equals substitution of $\mu_M(a',C_e)$. The same conditions apply to L7, and
the single exposure--outcome queries involve only their exposure--outcome part.
Contrast support is the proportion of participants for whom both contrast
levels lie within two residual standard deviations of the covariate-conditional
exposure mean (Tables~\ref{tab:foundation:controlled-direct}
and~\ref{tab:foundation:single-exposure}).

\begin{table}[tbp]
  \centering
  \caption{\textbf{Conditional-mean equations of the BMI--SpO$_2$--cIMT
  decomposition} (primary adjustment, 425 participants). Coefficients are in
  original units (SpO$_2$ equation: percentage points; cIMT equation: mm; per
  unit of each input), and each categorical coefficient multiplies its indicator.
  Reference categories: first sex category; current smoking on most or all days;
  first academic degree. ---, term absent from the equation.}
  \label{tab:foundation:o3-coefficients}
  \footnotesize
  \setlength{\tabcolsep}{3pt}
  \begin{tabular*}{\linewidth}{@{\extracolsep{\fill}}lcrcr@{}}
    \toprule
    Term & \multicolumn{2}{c}{SpO$_2$ equation} & \multicolumn{2}{c}{cIMT equation} \\
    \midrule
    Intercept & $\alpha_0$ & $100.7$ & $\theta_0$ & $-0.4883$ \\
    BMI ($A$; kg/m$^2$) & $\alpha_A$ & $-0.08190$ & $\theta_A$ & $-0.001297$ \\
    SpO$_2$ ($M$; \%) &  & --- & $\theta_M$ & $0.006467$ \\
    $A\times M$ &  & --- & $\theta_{AM}$ & $6.466\times10^{-5}$ \\
    Age (years) & $\gamma_M$ & $-0.04768$ & $\gamma_Y$ & $0.007096$ \\
    Sex: second category & $\gamma_M$ & $-0.1264$ & $\gamma_Y$ & $4.347\times10^{-5}$ \\
    Waist circumference (cm) & $\gamma_M$ & $-0.01028$ & $\gamma_Y$ & $-0.0003858$ \\
    Smoking: current occasional & $\gamma_M$ & $0.1834$ & $\gamma_Y$ & $-0.01095$ \\
    Smoking: not current & $\gamma_M$ & $0.3659$ & $\gamma_Y$ & $0.004144$ \\
    Education: high-school certificate & $\gamma_M$ & $0.2266$ & $\gamma_Y$ & $0.01742$ \\
    Education: high school without diploma & $\gamma_M$ & $-0.1474$ & $\gamma_Y$ & $0.03605$ \\
    Education: no certificate & $\gamma_M$ & $0.3273$ & $\gamma_Y$ & $0.01498$ \\
    Education: professional or other certificate & $\gamma_M$ & $-0.04736$ & $\gamma_Y$ & $0.002780$ \\
    Education: second academic degree & $\gamma_M$ & $0.02687$ & $\gamma_Y$ & $0.004264$ \\
    Education: third academic degree & $\gamma_M$ & $0.1151$ & $\gamma_Y$ & $0.01014$ \\
    \bottomrule
  \end{tabular*}
\end{table}

\paragraph{Other effect queries.}
The other effect queries are the decomposition of L7 (ALT--TG--PWV) and three
single exposure--outcome queries: GGT--PWV, TG--PWV and SpO$_2$--cIMT. The two
conditional-mean equations of L7 take the form of
Equation~\ref{eq:foundation:o3-conditional-model} without the
exposure--mediator interaction. The L7 sample consists of ALT, TG and PWV
measurement chains in temporal order with a complete primary adjustment set,
196 chains from 147 participants; the chains of each participant carry a total
weight of one, the bootstrap resamples participants, and the controlled direct
contrast fixes TG at the chain median of 92 mg/dL. Each single exposure--outcome query takes, for each
participant, the first exposure measurement with a complete primary adjustment
set and the first outcome measurement within the subsequent follow-up window;
its outcome equation is a linear regression of the outcome on the exposure and
the adjustment set, and its total contrast is
$T=E_C[\mu_Y(a_1,C)-\mu_Y(a_0,C)]$. The primary adjustment set is age and sex
for GGT--PWV, with prior ALT and WHR added for TG--PWV and prior waist
circumference added for SpO$_2$--cIMT; expanded adjustment of TG--PWV and
SpO$_2$--cIMT adds the prior outcome, BMI, smoking and alcohol. For L7, the
adjustment set is age, sex and prior hip circumference, waist circumference and
WHR. Each exposure contrast spans the 25th to
75th training-split percentiles: GGT 14 to 26 U/L, TG 71 to 131.5 mg/dL,
SpO$_2$ 94\% to 96\% and ALT 14 to 27 U/L. Tables~\ref{tab:foundation:mediation-decomposition}
and~\ref{tab:foundation:controlled-direct} report the decompositions and
Table~\ref{tab:foundation:single-exposure} the single exposure--outcome
queries. The covariate-omission sensitivity omits each covariate added by
expanded adjustment, one at a time, on the same expanded-adjustment sample and
compares the total contrast with that of full adjustment
(Table~\ref{tab:foundation:omission}).

\begin{table}[tbp]
  \centering
  \caption{\textbf{Decomposition of the two mediated pathways.} $T=D+I=D'+I'$;
  $n$, participants; 95\% CIs below; units: mm for O3 (BMI--SpO$_2$--cIMT) and
  m/s for L7 (ALT--TG--PWV). The L7 outcome equation has no interaction, so
  $D=D'$ and $I=I'$.}
  \label{tab:foundation:mediation-decomposition}
  \footnotesize
  \setlength{\tabcolsep}{2pt}
  \begin{tabular*}{\linewidth}{@{\extracolsep{\fill}}llrccccc@{}}
    \toprule
    Query & Adjustment & $n$ & $T$ & $D$ & $I$ & $D'$ & $I'$ \\
    \midrule
    O3 & Primary & 425 & \begin{tabular}[t]{@{}c@{}}$0.02306$\\{\tiny $[0.00400,0.04285]$}\end{tabular} & \begin{tabular}[t]{@{}c@{}}$0.02678$\\{\tiny $[0.00737,0.04683]$}\end{tabular} & \begin{tabular}[t]{@{}c@{}}$-0.00372$\\{\tiny $[-0.00883,-0.00055]$}\end{tabular} & \begin{tabular}[t]{@{}c@{}}$0.02662$\\{\tiny $[0.00786,0.04649]$}\end{tabular} & \begin{tabular}[t]{@{}c@{}}$-0.00356$\\{\tiny $[-0.00907,0.00018]$}\end{tabular} \\[2pt]
    O3 & Expanded & 294 & \begin{tabular}[t]{@{}c@{}}$0.01962$\\{\tiny $[-0.00090,0.03747]$}\end{tabular} & \begin{tabular}[t]{@{}c@{}}$0.02258$\\{\tiny $[0.00227,0.04057]$}\end{tabular} & \begin{tabular}[t]{@{}c@{}}$-0.00296$\\{\tiny $[-0.00769,0.00015]$}\end{tabular} & \begin{tabular}[t]{@{}c@{}}$0.02179$\\{\tiny $[0.00136,0.03993]$}\end{tabular} & \begin{tabular}[t]{@{}c@{}}$-0.00217$\\{\tiny $[-0.00689,0.00130]$}\end{tabular} \\[2pt]
    L7 & Primary & 147 & \begin{tabular}[t]{@{}c@{}}$-0.03061$\\{\tiny $[-0.25270,0.28523]$}\end{tabular} & \begin{tabular}[t]{@{}c@{}}$-0.08646$\\{\tiny $[-0.30193,0.20208]$}\end{tabular} & \begin{tabular}[t]{@{}c@{}}$0.05585$\\{\tiny $[-0.00035,0.14807]$}\end{tabular} & \begin{tabular}[t]{@{}c@{}}$-0.08646$\\{\tiny $[-0.30193,0.20208]$}\end{tabular} & \begin{tabular}[t]{@{}c@{}}$0.05585$\\{\tiny $[-0.00035,0.14807]$}\end{tabular} \\[2pt]
    \bottomrule
  \end{tabular*}
\end{table}

\begin{table}[tbp]
  \centering
  \caption{\textbf{Controlled direct contrast, $T-\mathrm{CDE}$ and mediator
  shift.} The controlled direct contrast fixes the mediator at the
  analysis-sample median (SpO$_2$ 96\%, TG 92 mg/dL); $\mathrm{CDE}$ and
  $T-\mathrm{CDE}$ are in the units of
  Table~\ref{tab:foundation:mediation-decomposition}, and mediator shifts in
  SpO$_2$ percentage points or TG mg/dL; 95\% CIs below.}
  \label{tab:foundation:controlled-direct}
  \footnotesize
  \setlength{\tabcolsep}{3pt}

\end{table}

\begin{table}[tbp]
  \centering
  \caption{\textbf{Total contrasts $T$ of the single exposure--outcome queries.}
  95\% CIs in brackets.}
  \label{tab:foundation:single-exposure}
  \footnotesize
  \setlength{\tabcolsep}{3pt}
  \begin{tabular*}{\linewidth}{@{\extracolsep{\fill}}llrrcr@{}}
    \toprule
    Query (unit) & Adjustment & $n$ & $T$ & 95\% CI & Support (\%) \\
    \midrule
    GGT--PWV (m/s) & Primary & 1{,}573 & $0.01505$ & $[-0.02200,0.05827]$ & 100.0 \\
    TG--PWV (m/s) & Primary & 281 & $0.14110$ & $[-0.04879,0.33341]$ & 98.6 \\
    TG--PWV (m/s) & Expanded & 150 & $0.04853$ & $[-0.24175,0.33510]$ & 94.0 \\
    SpO$_2$--cIMT (mm) & Primary & 311 & $0.01314$ & $[-0.00422,0.02922]$ & 90.0 \\
    SpO$_2$--cIMT (mm) & Expanded & 147 & $0.01267$ & $[-0.00704,0.03169]$ & 89.1 \\
    \bottomrule
  \end{tabular*}
\end{table}

\begin{table}[tbp]
  \centering
  \caption{\textbf{Covariate-omission sensitivity.} On the same
  expanded-adjustment sample, the total contrast after omitting one covariate
  added by expanded adjustment minus the total contrast under full adjustment;
  the primary adjustment set is always kept; units: m/s for TG--PWV and mm for
  SpO$_2$--cIMT and O3.}
  \label{tab:foundation:omission}
  \footnotesize
  \setlength{\tabcolsep}{3pt}
  \begin{tabular*}{\linewidth}{@{\extracolsep{\fill}}lrlrc@{}}
    \toprule
    Query & $n$ & Omitted covariate & Difference & 95\% CI \\
    \midrule
    TG--PWV & 150 & Prior PWV & $0.07246$ & $[-0.00931,0.15360]$ \\
     &  & Prior BMI & $0.06802$ & $[-0.00283,0.14554]$ \\
     &  & Smoking & $-0.00305$ & $[-0.04215,0.03887]$ \\
     &  & Alcohol & $-0.02507$ & $[-0.11975,0.04534]$ \\
    SpO$_2$--cIMT & 147 & Prior BMI & $-0.00103$ & $[-0.00691,0.00334]$ \\
     &  & Smoking & $0.00091$ & $[-0.00199,0.00468]$ \\
     &  & Alcohol & $-0.00012$ & $[-0.00782,0.00665]$ \\
     &  & Prior cIMT & $0.00103$ & $[-0.01044,0.01103]$ \\
    O3 & 294 & Alcohol & $0.00062$ & $[-0.00355,0.00623]$ \\
     &  & Prior cIMT & $0.01376$ & $[0.00554,0.02572]$ \\
    \bottomrule
  \end{tabular*}
\end{table}

\paragraph{Sex and age subgroups.}
Each subgroup analysis refits the equations within the subgroup and averages
the contrast over its participants; age at exposure defines the groups below 60
and at or above 60 years. The subgroup differences, second minus first sex
category and older minus younger, use the same resamples as the subgroup
effects (Table~\ref{tab:foundation:subgroups}).

\begin{table}[tbp]
  \centering
  \caption{\textbf{Sex and age subgroups.} $T$, total contrast; $I$, mediated
  component; $n$, subgroup participants; units as in
  Tables~\ref{tab:foundation:mediation-decomposition}
  and~\ref{tab:foundation:single-exposure}. For L7, age is taken at the exposure
  date of each chain, so one participant can contribute to both age groups.}
  \label{tab:foundation:subgroups}
  \scriptsize
  \setlength{\tabcolsep}{3pt}
  \renewcommand{\arraystretch}{0.92}
\begin{tabular*}{\linewidth}{@{\extracolsep{\fill}}lllrrc@{}}
    \toprule
    Query & Adjustment & Subgroup or difference & $n$ & Estimate & 95\% CI \\
    \midrule
    GGT--PWV ($T$) & Primary & First sex category & 880 & $-0.00622$ & $[-0.05413,0.10054]$ \\
     &  & Second sex category & 693 & $0.02791$ & $[-0.01648,0.06871]$ \\
     &  & Second $-$ first &  & $0.03413$ & $[-0.07938,0.10476]$ \\
     &  & Below 60 years & 1{,}278 & $0.01791$ & $[-0.02339,0.06963]$ \\
     &  & 60 years or above & 295 & $-0.00931$ & $[-0.09486,0.12346]$ \\
     &  & Older $-$ younger &  & $-0.02723$ & $[-0.12594,0.10696]$ \\
    \addlinespace[2pt]
    TG--PWV ($T$) & Primary & First sex category & 142 & $0.28005$ & $[-0.00519,0.56368]$ \\
\bottomrule
  \end{tabular*}
\end{table}

\paragraph{Residual-correlation sensitivity of L7.}
Let $\alpha_A$ be the exposure
coefficient of the mediator equation, $\theta_M$ the mediator coefficient of the
outcome equation, $s_M$ and $s_Y$ the residual standard deviations of the two
equations, and $\rho$ the correlation between the mediator and outcome
residuals, which represents unmeasured mediator--outcome confounding. The
mediated component (Table~\ref{tab:foundation:l7-rho}) is
$I(\rho)=\Delta a\,\alpha_A\bigl[\theta_M-\rho s_Y/\bigl(s_M\sqrt{1-\rho^2}\bigr)\bigr]$,
which equals $I$ of Table~\ref{tab:foundation:mediation-decomposition} at
$\rho=0$; $I(\rho)=0$ at
$\rho_0=\theta_Ms_M/\sqrt{s_Y^2+\theta_M^2s_M^2}=0.23801$ (95\% CI $[0.08147,0.37299]$).

\begin{table}[tbp]
  \centering
  \caption{\textbf{Mediated component of L7 under residual correlation
  $\rho$} (m/s). 95\% CIs in brackets.}
  \label{tab:foundation:l7-rho}
  \footnotesize
  \begin{tabular}{rrc}
    \toprule
    $\rho$ & $I(\rho)$ & 95\% CI \\
    \midrule
    $-0.3$ & $0.12752$ & $[0.00314,0.31006]$ \\
    $-0.1$ & $0.07875$ & $[0.00196,0.19750]$ \\
    $0$ & $0.05585$ & $[-0.00035,0.14807]$ \\
    $+0.1$ & $0.03294$ & $[-0.00596,0.10498]$ \\
    $+0.3$ & $-0.01582$ & $[-0.08266,0.02102]$ \\
    \bottomrule
  \end{tabular}
\end{table}

\FloatBarrier

\end{document}